\documentclass[School=Harvard]{Dissertate}

\usepackage{float}
\usepackage{datetime}

\RequirePackage{algorithm}
\RequirePackage{algorithmic}
\usepackage{etoolbox}\AtBeginEnvironment{algorithmic}{\small}

\usepackage{adjustbox}
\usepackage{makecell}

\title{Multi-Objective Reward and Preference Optimization: \hspace{2cm} Theory and Algorithms}
\author{Akhil Agnihotri}
\advisor{Rahul Jain}

\studydegree{Doctor of Philosophy in Electrical Engineering}
\field{Electrical and Computer Engineering}
\degreemonth{\monthname}
\degreeyear{\the\year{}}
\department{Ming Hsieh Department of Electrical and Computer Engineering}

\definecolor{TS}{RGB}{0, 125, 140}
\definecolor{LinTS}{RGB}{6, 75, 204}
\definecolor{naiveTS}{RGB}{201, 127, 0}
\definecolor{warmTS}{RGB}{128, 0, 32}
\definecolor{cellg}{RGB}{223, 247, 220}

\definecolor{Green}{rgb}{0.13, 0.65, 0.3}
\definecolor{Greenish}{RGB}{49, 158, 140}
\definecolor{Pinkish}{RGB}{184, 31, 120}
\definecolor{CellCol}{RGB}{218, 242, 199}
\definecolor{CellColLight}{RGB}{238, 245, 233}

\definecolor{blueish}{RGB}{46, 48, 146}
\definecolor{marigold}{RGB}{246, 145, 30}
\definecolor{purpleish}{RGB}{129, 2, 129}
\definecolor{greenish}{RGB}{0, 154, 85}
\definecolor{maroonish}{RGB}{130, 27, 1}
\definecolor{cream}{RGB}{255, 253, 208}
\definecolor{lavender}{RGB}{230, 230, 250}
\definecolor{sagegreen}{RGB}{157, 193, 131}
\definecolor{charcoal}{rgb}{0.25,0.25,0.25}
\definecolor{darkorange}{RGB}{209, 119, 2}

\definecolor{usc}{RGB}{153, 27, 30}
\definecolor{jpm}{RGB}{0, 53, 148}
\definecolor{eqcol}{RGB}{206, 240, 216}
\definecolor{backg_blue}{RGB}{225,236,244}
\definecolor{tagtxt_blue}{RGB}{88,115,159}
\definecolor{backg_red}{RGB}{250, 222, 222}
\definecolor{tagtxt_red}{RGB}{204, 71, 71}
\definecolor{darkgreen}{RGB}{21, 145, 1}
\definecolor{orange}{RGB}{230, 125, 14}
\definecolor{teal}{RGB}{0, 128, 128}
\definecolor{neon}{RGB}{25, 210, 227}
\definecolor{brightblue}{RGB}{102, 179, 255}
\definecolor{brightred}{RGB}{255, 102, 102}
\definecolor{maroon}{RGB}{143, 4, 4}
\definecolor{purple}{RGB}{119, 26, 240}
\definecolor{Pink}{RGB}{212, 38, 235}
\definecolor{crimsonred}{RGB}{220, 20, 60}
\definecolor{deeppink}{RGB}{176, 23, 106}

\definecolor{indigo}{RGB}{122, 93, 227}
\definecolor{violet}{RGB}{153, 153, 255}
\definecolor{dullpink}{RGB}{204, 171, 208}
\definecolor{lightorange}{RGB}{255, 191, 128}
\definecolor{lemon}{RGB}{255, 228, 113}
\definecolor{skyblue}{RGB}{141, 215, 247}

\newtcbox{\mybox}[2][]{on line,
  colback=#2,     
  colframe=#2!gray,    
  boxsep=1.5pt,      
  left=0pt,          
  right=0pt,         
  top=0pt,           
  bottom=0pt,        
  arc=1.25mm,           
  boxrule=0pt,       
  #1                 
}

\newcommand{\barrierf}{barrier function\xspace}
\newcommand{\sbarrierf}{strong-barrier function\xspace}
\newcommand{\subxy}{\substack{x\sim \mathcal X\\ y\sim\pi(\cdot\vert x)}}

\newtheorem{hypothesis}{Hypothesis}

\newcommand{\piref}{\pi_{\mathsf{ref}}}
\newcommand{\rlhf}{{RLHF}\hphantom{.}}
\newcommand{\dual}{\texttt{Dual$^{\star}$}}

\newcommand{\paretoset}{\Pi_{\pareto{P}}}

\definecolor{pareto}{RGB}{27, 224, 250}
\definecolor{D1}{RGB}{220, 20, 60}
\definecolor{D2}{RGB}{255, 140, 0}
\definecolor{DJ}{RGB}{50, 205, 50}
\definecolor{DC}{RGB}{138, 43, 226}
\definecolor{COP}{RGB}{255, 105, 180}
\definecolor{MOPO}{RGB}{219, 82, 18}

\newcommand{\pareto}[1]{#1}
\newcommand{\Done}[1]{\textcolor{D1!70}{{#1}}}
\newcommand{\Dtwo}[1]{\textcolor{D2!70!black}{{#1}}}
\newcommand{\Djoint}[1]{\textcolor{DJ!70!black}{{#1}}}
\newcommand{\Dcombined}[1]{\textcolor{DC!80}{{#1}}}
\newcommand{\COP}[1]{\textcolor{COP!70!black}{\texttt{#1}}}
\newcommand{\MOPO}[1]{\textcolor{MOPO!95!black}{\texttt{#1}}}

\definecolor{accgood}{RGB}{97, 207, 102}
\definecolor{accneutral}{RGB}{255, 238, 84}
\definecolor{accbad}{RGB}{250, 115, 105}
\definecolor{bestblue}{RGB}{25, 118, 210}
\definecolor{worstred}{RGB}{198, 40, 40}
\definecolor{mopogray}{RGB}{189, 200, 201}

\newcommand{\colmin}[1]{%
  \ifcase#1 0.0\or 0.21\or 0.24\or 0.34\or 0.30\else 0.0\fi}
\newcommand{\colmid}[1]{%
  \ifcase#1 0.0\or 0.37\or 0.39\or 0.47\or 0.42\else 0.0\fi}
\newcommand{\colmax}[1]{%
  \ifcase#1 0.0\or 0.53\or 0.53\or 0.63\or 0.52\else 0.0\fi}

\newcommand{\acc}[3]{%
  \pgfmathsetmacro{\Value}{#1}
  \pgfmathsetmacro{\MinNumber}{\colmin{#3}}
  \pgfmathsetmacro{\MidNumber}{\colmid{#3}}
  \pgfmathsetmacro{\MaxNumber}{\colmax{#3}}
  \ifdim \Value pt<\MidNumber pt
    \pgfmathsetmacro{\PercentColor}{100*((\MidNumber-\Value)/(\MidNumber-\MinNumber))}
    \xdef\PercentColorr{\PercentColor}
    {\tcbox[on line,
      boxsep=0.1pt, left=0.1pt, right=0.1pt, top=0pt, bottom=0pt, arc=0.8pt,
      colframe=accbad!\PercentColorr!accneutral!60,
      colback=accbad!\PercentColorr!accneutral!60]{\scriptsize $#1_{\text{\tiny $\pm#2$}}$}}
  \else
    \pgfmathsetmacro{\PercentColor}{100*((\Value-\MidNumber)/(\MaxNumber-\MidNumber))}
    \xdef\PercentColorr{\PercentColor}
    {\tcbox[on line,
      boxsep=0.1pt, left=0.1pt, right=0.1pt, top=0pt, bottom=0pt, arc=0.8pt,
      colframe=accgood!\PercentColorr!accneutral!60,
      colback=accgood!\PercentColorr!accneutral!60]{\scriptsize $#1_{\text{\tiny $\pm#2$}}$}}
  \fi
}

\newcommand{\accp}[3]{\hspace{-0.4cm}{\acc{#1}{#2}{#3}}\hspace{0.01cm}}

\newcommand{\PSPL}{\mathsf{PSPL}}

\newcommand{\SR}{\mathscr{SR}}

\newcommand{\Anzero}{\bar{A}_n^{(0)}}
\newcommand{\Anone}{\bar{A}_n^{(1)}}
\newcommand{\BR}{\mathscr{BR}}

\newcommand{\UD}{\Ucal_{\Dcal_{0}}}

\newcommand\code[1]{\colorbox{jpm!10}{\texttt{#1}}}

\newcommand{\pib}{\bm{\pi}}

\newcommand{\rp}{\rho_{\pi}}

\newcommand{\norm}[3]{\left\| #1 \right\|_{#2}^{#3}}

\renewcommand{\iota}{\textit{i}}
\renewcommand{\implies}{\; \Rightarrow \;}

\newcommand{\forAll}{{\;\; \forall \;}}

\newcommand{\suchthat}{\;\; \texttt{s.t.} \;\;}

\newcommand{\given}{\, | \,}
\newcommand{\Given}{\; \big| \;}

\renewcommand{\bar}[1]{\overline{#1}} 
\renewcommand{\tilde}[1]{\widetilde{#1}} 
\renewcommand{\hat}[1]{\widehat{#1}} 

\newcommand{\wb}[1]{\widebar{#1}}
\newcommand{\E}[1]{\underset{\begin{subarray}{c} #1 \end{subarray}}{\Ebb}}

\newcommand{\TV}[2]{D_\text{TV}(#1\parallel #2)}

\newcommand{\pik}{\pi_{k}}
\newcommand{\pikup}{\pi_{k+1}}
\newcommand{\pol}{\pi_{\theta}}

\newcommand{\polk}{\pi_{\theta_k}}
\newcommand{\polkup}{\pi_{\theta_{k+1}}}
\newcommand{\grad}{\nabla_{\theta}}

\newcommand{\vfunc}{\wb{V}^{\pi}}
\newcommand{\vfuncc}{\wb{V}_{C_{i}}^{\pi}}
\newcommand{\qfunc}{\wb{Q}^{\pi}}
\newcommand{\qfuncc}{\wb{Q}_{C_{i}}^{\pi}}

\newcommand{\vphi}{\wb{V}^{\pi}_{\phi}}
\newcommand{\adv}{\wb{A}^{\pi}}
\newcommand{\advc}{\wb{A}_{C_{i}}^{\pi}}

\newcommand{\jfuncp}{J_{C_{i}}(\pi)}

\newcommand{\advd}{A_{\gamma}^{\pi}}
\newcommand{\vtarg}{\wb{V}^{\text{target}}}

\newcommand{\advk}{\wb{A}^{\pik}}

\newcommand{\dpi}{d_{\pi}}
\newcommand{\dpip}{d_{\pi'}}
\newcommand{\dpid}{d_{\pi,\gamma}}

\newcommand{\dpik}{d_{\pik}}
\newcommand{\dpolk}{d_{\pi_{\theta_k}}}

\newcommand{\Zpi}{Z^{\pi}}
\newcommand{\Mpip}{M^{\pi'}}

\newcommand{\Zpip}{Z^{\pi'}}
\newcommand{\dg}{\text{dg}}

\newcommand{\KL}{\mathsf{KL}}

\newcommand{\avKL}[2]{\bar{D}_\text{KL}(#1\parallel #2)}

\newcommand{\Ebb}{{\mathbb E}}

\newcommand{\Ibb}{{\mathbb I}}

\newcommand{\Nbb}{{\mathbb N}}
\newcommand{\Pbb}{{\mathbb P}}

\newcommand{\Rbb}{{\mathbb R}}

\newcommand{\Ibf}{{\mathbf{I}}}

\newcommand{\Acal}{{\mathcal{A}}}
\newcommand{\Bcal}{{\mathcal{B}}}
\newcommand{\Ccal}{{\mathcal{C}}}
\newcommand{\Dcal}{{\mathcal{D}}}
\newcommand{\Ecal}{{\mathcal{E}}}
\newcommand{\Fcal}{{\mathcal{F}}}
\newcommand{\Gcal}{{\mathcal{G}}}
\newcommand{\Hcal}{{\mathcal{H}}}

\newcommand{\Lcal}{{\mathcal{L}}}
\newcommand{\Mcal}{{\mathcal{M}}}
\newcommand{\Ncal}{{\mathcal{N}}}
\newcommand{\Ocal}{{\mathcal{O}}}
\newcommand{\Pcal}{{\mathcal{P}}}

\newcommand{\Scal}{{\mathcal{S}}}
\newcommand{\Tcal}{{\mathcal{T}}}
\newcommand{\Ucal}{{\mathcal{U}}}

\newcommand{\Wcal}{{\mathcal{W}}}
\newcommand{\Xcal}{{\mathcal{X}}}
\newcommand{\Ycal}{{\mathcal{Y}}}

\DeclareMathOperator*{\argmax}{arg\,max}
\DeclareMathOperator*{\argmin}{arg\,min}

\makeatletter
\newcommand{\revisionhistory}[1]{%
\@ifundefined{showrevisionhistory}{\relax}{%
{#1}%
}%
}
\makeatother

\usepackage{mathabx}

\makeatletter \renewcommand\paragraph[1]{\@startsection{paragraph}{4}{\z@}{2.0ex \@plus1ex \@minus.2ex}{-1em}{\bfseries\small\textsc{#1}}} \makeatother

\begin{document}

\pagenumbering{roman}
\setcounter{page}{1}
\maketitle

\newpage
\addcontentsline{toc}{chapter}{Doctoral Committee}

\vspace*{\fill}

\begin{table}[h]
        \centering
        \begin{tabular}{cl}
           \textsc{Doctoral Committee} :  &  \textsc{Rahul Jain $\,$ (Chair)} \\
           & \textsc{Haipeng Luo} \\
           & \textsc{Ashutosh Nayyar}
        \end{tabular}
\end{table}

\vspace*{\fill}

\newpage
\addcontentsline{toc}{chapter}{Author Contributions}

\begin{table}[hbt!]
\fontsize{9}{10}\selectfont
    \renewcommand{\arraystretch}{2}
    \vspace{8cm}
    \centering
    \begin{tabular}{c c c} 
     \hline
      Chapter Number & Contributors & Venue\\
      \hline
      Chapter \ref{chapter:acpo} & \small Akhil Agnihotri, Rahul Jain, Haipeng Luo & ICML 2024 \\
      \multirow{2}{*}{Chapter \ref{chapter:ecop}} & \multirowcell{2}{\small Akhil Agnihotri, Rahul Jain, \\ Deepak Ramachandran, Sahil Singla} & \multirow{2}{*}{NeurIPS 2024} \\
      & & \\
      \multirow{2}{*}{Chapter \ref{chapter:warmPref-PS}} & \multirowcell{2}{\small Akhil Agnihotri, Rahul Jain, \\ Deepak Ramachandran, Zheng Wen} & \multirow{2}{*}{TMLR (submitted)} \\
      & & \\
      \multirow{2}{*}{Chapter \ref{chapter:PSPL}} & \multirowcell{2}{\small Akhil Agnihotri, Rahul Jain, \\ Deepak Ramachandran, Zheng Wen} & \multirow{2}{*}{AISTATS (submitted)} \\
      & & \\
      \multirow{2}{*}{Chapter \ref{chapter:mopo}} & \multirowcell{2}{\small Akhil Agnihotri, Rahul Jain, \\ Deepak Ramachandran, Zheng Wen} & \multirow{2}{*}{(to be submitted)} \\
      & & \\
      Chapter \ref{chapter:conclusion} & \small Akhil Agnihotri & n/a \\
      \hline 
    \end{tabular}
    \caption{\centering Author Contributions}
\end{table}

\newpage
\addcontentsline{toc}{chapter}{Acknowledgments}

	\chapter*{Acknowledgments}
	\noindent

\begin{adjustwidth}{0.85cm}{0.85cm}

\lettrine[lines=3,slope=-2pt,nindent=-4pt]{\textcolor{SchoolColor}{T}}{ hese few words} will never be enough to thank the people who guided, helped, and supported me throughout my Ph.D. It has been a journey full of ups and downs, but what ultimately matters is whether something meaningful came out of it. Over the last five years, I have grown both as a researcher and as an individual, and I am deeply grateful to everyone who has been part of that process.

First and foremost, my deepest gratitude goes to my advisor, Prof. Rahul Jain, who has guided me and remained a steady, positive influence throughout. He taught me how to conduct independent research and consistently encouraged me to learn new things and build the mindset needed for a career in research. I am equally grateful to Prof. Haipeng Luo, whose excellence in research and mentorship is truly unmatched. He patiently taught me the ropes during my first paper and continued to help whenever I needed his guidance.

I have also been extremely fortunate to be mentored by Zheng Wen during our collaborations. He had an uncanny ability to understand my strengths and weaknesses, and he offered me a clear roadmap for how to think about research problems, approach proofs, and grow into a better researcher. Without his patience and passion for helping young students develop their full potential, my Ph.D. would have been far more difficult. I learned an enormous amount during our two years of collaboration.

Before starting my Ph.D., I had the privilege of working with Prof. Ding Zhao at Carnegie Mellon University. He saw potential in me early on and welcomed me into the SafeAI Lab, where I gained my first real exposure to research. Without that experience, I cannot imagine being ready to begin a Ph.D. His kindness and support, both then and continuing even today, have meant the world to me, and I remain deeply grateful for his mentorship. I also appreciate the time I spent working with Zuxin Liu, which helped set the foundation for everything that followed.

During my undergraduate years at BITS Pilani, I was profoundly influenced by Prof. Bivudutta Mishra, Prof. N. L. Bhanu Murthy and Prof. Anil Nemili. Their 8 a.m. classes, dreaded by everyone else, were the ones I eagerly looked forward to. Their clarity, energy, and passion for teaching were unmatched, and the concepts they taught nearly a decade ago continue to guide me whenever I revisit the fundamentals.

Beyond academics, I have been blessed with wonderful friendships that have shaped me over the years. Starting with my Day $1$s - Chirag Dharmesh, Amrit Raheja, Reewan Aaron Alvares, and Hassan Laiq, I am grateful for a friendship that has lasted over 15 years. We may not talk every day, but I know they are always in my corner, a belief that has only grown stronger with time. I am equally thankful for my \emph{wingies} from college - Akshay Anand, Divyanshu Soni, Pragyan Shukla, Mridul Bhaskar, Akshat Chaurasia, and Prayansh Chauhan. Every time I returned to India, meeting them was non-negotiable. I hope that as we grow older, we remain just as silly and stay the same clueless 18-year-olds at heart.

I am also grateful to Utkarsh Gupta, Prathamesh Saraf, Gokul Karthik, Ahan Makanur, Ayush Yadav, and Tarunbir Singh Gambhir, who have been rays of sunshine during the sometimes-cloudy Ph.D. days. My first research project was with Prathamesh in 2019, and that small seed eventually grew into the journey of the last five years.

During my Ph.D., I was fortunate enough to make new friends who quietly kept my mental health in check. My time at USC would not have been the same without Disha Jain, Jeongmin Chae (Chai), Shardul Nazirkar, Hrushi Lavate, Tejas Shenoy, Aryan Gandhi, Aratrika Basu, Rajesh Shrestha, Akshita Namjoshi, Rishabh Agrawal, and Tejas Muthal. Whether it was our fountain-side study sessions or 4 a.m. Saturday night parties, these moments became some of my favorite memories. I can also never forget Christine Olive Dobard (sorry, CoCo). Our twice-a-year therapy sessions and our very occasional (read: constant) banter kept me alive in ways I didn’t fully appreciate at the time. And truly, I couldn’t have asked for a better companion for Sigur R\'os.

Just as importantly, I would like to thank my parents, Ajay Agnihotri and Mamta Sharma, and my entire family for always supporting me and believing in me. I am forever grateful for their sacrifices and dedication in helping me reach my goals. Thank you for making me possible.

\vspace{1cm}

\emph{Last, but not the least, I want to thank me for believing in me, I want to thank me for doing all this hard work. I wanna thank me for having no days off. I wanna thank me for never quitting. I wanna thank me for being me at all times.}

\end{adjustwidth}

	\vspace*{\fill} \newpage
    \newpage

\newpage
\thispagestyle{empty}
\vspace*{\fill}
\begin{center}
    \emph{"Remember that wherever your heart is, there you will find your treasure." \\ - The Alchemist, Paulo Coelho}
\end{center}
\vspace*{\fill}
\newpage

\tableofcontents

\newpage
\addcontentsline{toc}{chapter}{Listing of Figures}
\listoffigures

\newpage
\addcontentsline{toc}{chapter}{List of Tables}
\listoftables

\newpage

\addcontentsline{toc}{chapter}{Abstract}
\abstractpage

\newpage
\pagenumbering{arabic}
\setcounter{page}{1}

\setcounter{chapter}{-1}  
\chapter{Introduction}
\label{introduction}

Reinforcement Learning (RL) provides a mathematical framework for learning optimal sequential
decision-making policies through interactions with an unknown environment. The foundational
Markov Decision Process (MDP) abstraction has enabled RL to contribute significantly to fields such
as robotic control, autonomous systems, resource allocation, and generative modeling
\cite{levine2016end,akkaya2019solving,aractingi2023controlling}. Recent advances in deep RL have
produced highly scalable policy optimization algorithms capable of operating in continuous state and
action spaces, including TRPO and PPO \cite{silver2017mastering,vinyals2019grandmaster}, and have
fueled breakthroughs in complex tasks such as strategic gameplay, dexterous manipulation, and the
training of large language models (LLMs) via Reinforcement Learning from Human Feedback (RLHF)
\cite{ouyang2022training,chung2024scaling}. These algorithms commonly rely on variants of likelihood
ratio policy gradients, trust-region optimization principles, and generalized advantage estimation to
achieve stable policy improvement.

Although deep RL has achieved impressive performance, the challenge of satisfying \emph{safety
constraints} during the learning process remains largely unresolved. In many real-world systems,
agents must satisfy hard or soft constraints at all times in order to prevent catastrophic failures,
adhere to operational limits, ensure fairness, or guarantee that model behavior remains aligned with
human intent. For instance, robotic agents must avoid physical hazards, autonomous vehicles must
respect safety-critical rules, and LLMs must obey safety, toxicity, and honesty requirements. While
the control theory community has developed frameworks for safe control over several decades, the
design of safe, scalable, and statistically efficient RL algorithms is still an evolving research
frontier \cite{garcia2015comprehensive,brunke2022safe}. This thesis aims to address several
fundamental gaps that arise when constraints must be enforced in RL systems operating over long
horizons, finite horizons, or in environments where feedback is provided in the form of preferences.

A principled modeling abstraction for constrained decision-making is the \emph{Constrained Markov
Decision Process} (CMDP) \cite{altman1999constrained}. An average-reward CMDP augments the classical MDP with
cost functions $c_i(s,a)$ for $i = 1,\dots,m$, and the learning objective is to maximize a reward
criterion subject to constraints on long-term accumulated costs. For the average-reward setting, a
policy $\pi$ is considered feasible if
\[
\limsup_{T \to \infty} \frac{1}{T} \mathbb{E}_{\pi} \left[\sum_{t=1}^{T} c_i(s_t,a_t)\right]
\le d_i, \quad \text{for all } i.
\]
Although CMDPs provide the correct conceptual foundation, designing algorithms that can enforce such
constraints reliably, while remaining scalable under deep function approximation, is a challenging
problem. Lagrangian or penalty-based methods often suffer from weak constraint satisfaction or
unstable oscillatory behavior. Moreover, most existing methods are specialized for the infinite-horizon
discounted setting, and key analytical tools used in this setting do not generalize to alternative
objective formulations.

This thesis develops new theory and algorithms for constrained and preference-based RL across two broad themes:

\begin{enumerate}
    \item \textbf{Explicit constraints}, arising from environment-provided cost functions and
    thresholds (Chapters \ref{chapter:acpo} and \ref{chapter:ecop}).

    \item \textbf{Preference-induced constraints}, arising from human or rater judgments that impose
    implicit structure on the agent's policy (Chapters \ref{chapter:warmPref-PS},
    \ref{chapter:PSPL}, and \ref{chapter:mopo}).
\end{enumerate}

The resulting algorithms introduce several novel technical ingredients, including a new average-reward
performance bound based on Markov chain mixing properties, a finite-horizon policy difference lemma,
Bayesian posterior sampling methods for preference learning, the first simple-regret guarantees for
preference-based RL, and multi-objective preference optimization strategies for large-scale alignment
of LLMs.

\textbf{Average-Reward Constrained RL and the ACPO Algorithm.} Many real-world learning systems operate over long or continuous horizons, where the average reward
criterion is more suitable than the commonly used discounted objective. The average-reward setting
exhibits fundamentally different structure. For instance, the stationary distribution induced by a
policy plays a central role, and classical discounted tools such as the performance-difference lemma
degenerate when the discount factor approaches one. A key difficulty lies in characterizing how the
stationary distribution of an MDP changes when the policy changes. The discounted analysis implicitly
controls this effect through the discount factor, but such control is absent in the average case.

To address this gap, Chapter \ref{chapter:acpo} develops the \emph{Average-Constrained Policy
Optimization (ACPO)} algorithm, the first practical deep RL method for average-constrained MDPs
(ACMDPs). A central technical contribution is a new bound that relates the difference in average
rewards or costs of two policies to the \emph{worst-case mixture level} of the Markov chain induced by
the policies. Informally, the mixture level characterizes the largest possible change in the stationary
distribution when the policy changes by a small amount. This bound provides an analogue to the
discounted performance-difference lemma but tailored to the average setting.

In particular, the thesis derives an inequality of the form
\[
\eta(\pi) - \eta(\pi') \le L_{\mathrm{mix}}(\pi, \pi') \cdot
\mathbb{E}_{s \sim d_{\pi'}}\left[ A^{\pi'}(s,a) \right],
\]
where $L_{\mathrm{mix}}(\pi, \pi')$ is controlled by the spectral properties of the transition kernel
under $\pi'$, and $A^{\pi'}$ denotes the bias-based average-reward advantage function. This bound
leads to a trust-region constraint that enforces monotonicity in the average setting and supports
constraint satisfaction.

ACPO additionally introduces a new advantage estimator based on approximations of the bias function
rather than discounted returns, and incorporates sublevel sets of constraint functions to ensure that
all iterates remain feasible with respect to average constraint thresholds. A carefully constructed
line-search method provides additional stability. Empirically, ACPO achieves state-of-the-art
performance on Mujoco benchmarks, outperforming CPO \cite{achiam2017constrained}, PCPO \cite{zhang2022penalized}, PPO \cite{schulman2017proximal}, and ATRPO \cite{pmlr-v139-zhang21q}. These results demonstrate that average-constrained RL requires fundamentally different analysis and algorithmic mechanisms compared to the discounted setting.

\textbf{Episodic Constrained RL and the e-COP Algorithm.} Many RL tasks in robotics and generative modeling are naturally formulated as finite-horizon problems.
For example, episodic interactions arise in trajectory optimization, long-horizon manipulation tasks,
and diffusion-model denoising sequences \cite{black2023training,janner2022planning}. In episodic
settings, the optimal policy is non-stationary and varies with the time step. Classical reinforcement
learning results that depend on stationary distributions or infinite-horizon contraction properties are
therefore not applicable.

A key technical obstacle is the absence of an episodic analogue of the infinite-horizon performance
difference lemma. Chapter \ref{chapter:ecop} resolves this by introducing a new \emph{policy difference
lemma for finite-horizon MDPs}. Informally, if $\pi$ and $\pi'$ are two policies for an episodic MDP
of horizon $H$, then the difference in their performances can be decomposed into a sum of expectations
over time-indexed advantage functions:
\[
J(\pi) - J(\pi')
    = \sum_{h=1}^H \mathbb{E}_{s_h,a_h \sim \pi}
    \left[ A^{\pi'}_h(s_h,a_h) \right].
\]
Unlike infinite-horizon variants, the expectation is taken with respect to the non-stationary
trajectory distribution of $\pi$, and there is no discount factor involved. This lemma serves as the
foundation of the algorithmic design.

Based on this theoretical insight, the thesis develops \texttt{e-COP}, the first policy optimization
algorithm specifically tailored for finite-horizon CMDPs. The algorithm combines a KL-divergence-based
trust region, time-dependent advantage estimation, and a quadratic penalty method that avoids expensive
Hessian computations. The design ensures that constraint satisfaction is controlled at each time step
rather than only in a steady-state sense. Under appropriate scaling of the trust region size and
advantage estimation error, the algorithm satisfies a guarantee of the form
\[
\text{cost}_{\pi_k} \le d + O(\varepsilon_k),
\]
where $\varepsilon_k$ captures estimation error at iteration $k$. This produces stable convergence and
predictable constraint enforcement.

Experimental evaluations on Safety Gym benchmarks demonstrate that \texttt{e-COP} outperforms adapted
versions of CPO, FOCOPS, P3O, and PPO in terms of both reward and constraint satisfaction. The
algorithm's design is compatible with deep neural network policies, making it a viable candidate for
safety-constrained RLHF applications in LLMs and diffusion models.

\textbf{Preferences as Implicit Constraints: From Bandits to RLHF.} In many modern alignment settings, constraints arise not from environment-provided cost functions but
from human or rater preferences over system behavior. This motivates \emph{preference-based RL}
(PbRL), where the agent receives comparative feedback such as ``trajectory A is preferred over
trajectory B'' instead of numeric rewards. PbRL aligns well with human evaluation processes but
introduces new statistical and computational challenges, including rater noise, competence variation,
trajectory-level credit assignment, and the difficulty of inferring multi-objective trade-offs from
binary comparisons.

Chapter \ref{chapter:warmPref-PS} introduces \texttt{warmPref-PS}, a Bayesian posterior sampling
algorithm for linear bandits that integrates offline preference datasets with online numerical rewards.
A central idea is the modeling of \emph{rater competence}. Offline preference data may come from raters
whose reliability varies, and naively treating all comparisons as equally trustworthy can lead to poor
performance. The algorithm therefore maintains posterior distributions over model parameters and rater
ability, enabling selective incorporation of offline preference information. Theoretical analysis shows
that the algorithm obtains a Bayesian regret bound of order
\[
\tilde{O}\left(d \sqrt{T} + \frac{1}{\alpha_{\mathrm{rater}}}\right),
\]
where $\alpha_{\mathrm{rater}}$ quantifies rater competence. This is a significant improvement over
methods that ignore rater heterogeneity.

Moving beyond bandits, Chapter \ref{chapter:PSPL} develops $\PSPL$, a posterior sampling approach for
best-policy identification in preference-based RL. The objective shifts from cumulative regret to
\emph{simple regret}, which is appropriate for RLHF and alignment tasks where only the final learned
policy matters. A key challenge is that preferences are expressed over entire trajectories. To address
this, the algorithm jointly samples reward functions and transition dynamics, producing posterior samples
that induce candidate optimal policies. The chapter provides the first Bayesian simple-regret guarantees
for PbRL and introduces a bootstrapped variant that remains effective in high-dimensional or
function-approximation settings.

\textbf{Multi-Objective Preference Optimization for LLM Alignment.} Large language models operate under multiple alignment objectives including helpfulness, harmlessness,
truthfulness, and non-toxicity. These objectives often conflict. Standard RLHF pipelines collapse these
criteria into a single reward model, which can obscure trade-offs and lead to reward hacking,
out-of-distribution failures, or undesirable emergent behaviors \cite{ye2024theoretical}. To address
these issues, Chapter \ref{chapter:mopo} formulates alignment as a multi-objective constrained
preference optimization problem.

The resulting algorithm, \texttt{MOPO}, uses preference data to construct a concave constrained
optimization program. A primary preference objective is maximized subject to secondary objectives being
no less than user-specified thresholds. The optimization variables correspond to importance sampling
ratios and are updated using closed-form proximal steps. A behavior cloning step extracts a neural
policy that matches the optimal ratios. Under mild assumptions on the preference dataset, MOPO
approximates the Pareto front and avoids the need for inference-time preference conditioning. Extensive
experiments on multi-billion parameter LLMs show that MOPO achieves stable alignment performance across
tasks requiring multiple competing objectives.

Taken together, ACPO, \texttt{e-COP}, \texttt{warmPref-PS}, $\PSPL$, and \texttt{MOPO} form a unified framework for constrained reinforcement learning across explicit, implicit, and multi-objective constraint regimes. The technical contributions include: (i) new performance-difference bounds for average-reward CMDPs based on Markov chain mixture properties, (ii) a novel finite-horizon policy difference lemma for episodic MDPs, (iii) posterior sampling algorithms that integrate offline preferences with rater-competence modeling, (iv) the first simple-regret guarantees for preference-based RL, and (v) a scalable multi-objective preference optimization method for aligning large language models. These contributions address several key limitations of existing constrained RL methodologies and advance the theoretical and practical foundations required for safe, scalable, and aligned decision-making in high-stakes real-world applications.

\chapter{Policy Optimization for Average-Reward Criterion}
\label{chapter:acpo}

\paragraph{Abstract}.
Reinforcement Learning (RL) for constrained MDPs (CMDPs) is an increasingly important problem for various applications. Often, the average criterion is more suitable than the discounted criterion. Yet, RL for average-CMDPs (ACMDPs) remains a challenging problem. Algorithms designed for discounted constrained RL problems often do not perform well for the average CMDP setting. In this chapter, we introduce a new policy optimization with function approximation algorithm for constrained MDPs with the average criterion. The Average-Constrained Policy Optimization (ACPO) algorithm is inspired by trust region-based policy optimization algorithms. We develop basic sensitivity theory for average CMDPs, and then use the corresponding bounds in the design of the algorithm. We provide theoretical guarantees on its performance, and through extensive experimental work in various challenging OpenAI Gym environments, show its superior empirical performance when compared to other state-of-the-art algorithms adapted for the ACMDPs.

\section{Introduction}
\label{sec:intro}

Over the last decade, we have seen an enormous impact of RL techniques on a variety of problems, from mastering complex games like Go  \cite{silver2017mastering} and StarCraft \cite{vinyals2019grandmaster} to robotic control \cite{levine2016end,akkaya2019solving, aractingi2023controlling}. Many of these have used RL-policy  optimization algorithms such as \citet{schulman2017proximal} for discounted MDPs (DMDPs). These have come in handy even in generative AI, e.g., training large language models (LLMs) \cite{achiam2023gpt}. However, applications often need satisfaction of some \textit{constraints}, e.g., physical safety of mobile robots \cite{Hu_2022_CVPR}, safe language, image or multi-modal output generation. Furthermore, the \textit{average criterion} when long-term rewards and safety are of consideration is more suitable. Using discounted cost formulations (as a proxy for safety) incentivizes policy optimization algorithms to search for policies that are short-term safe but not long-term  because of future-discounting. 

The planning problem for MDPs with constraints is often formulated as a Constrained MDP (CMDP) model \cite{manne1960linear,hordijk1979linear,altman1999constrained}. Unfortunately, CMDP models do not satisfy Bellman’s principle of optimality, and hence dynamic programming (DP)-style algorithms cannot be developed for the setting. Instead, an alternative approach called the convex analytic approach \cite{borkar1988convex, altman1999constrained} is used by way of introducing occupation measures that leads to optimization formulations. This can be done for both discounted (DCMDPs) and average-criterion (ACMDPs) constrained MDPs.

Theory and algorithms for RL deal with  settings when the MDP model is unknown. While DP-inspired RL algorithms such as DQN, when combined with deep learning architectures for function approximation work remarkably effectively \cite{mnih2015human}, policy optimization algorithms such as TRPO \cite{schulman2015trust}, PPO \cite{schulman2017proximal} have proven even more effective in solving high dimensional problems. Since the discounted criterion is sometimes not suitable, policy optimization algorithms such as ATRPO \cite{zhang2021average, wan2021learning, liao2022batch} have been developed for AMDPs. Furthermore, as already mentioned, certain RL applications have multiple objectives, one of which is to be optimized and the rest constrained. Thus, the Constrained Policy Optimization (CPO) algorithm \cite{achiam2017constrained} was  introduced for infinite-horizon DCMDP problems. Unfortunately, as motivated above, not all such applications fit the discounted-criterion formulation: there are settings, for example where there may be safety requirements when the average-CMDP model is a better fit. No scalable RL algorithms are currently available for such settings. 

We note that the RL problem is usually harder than the corresponding planning problem;   average-MDPs are  more challenging than discounted MDPs; and constrained MDPs are more challenging than unconstrained ones. In this chapter, we present the first practical algorithm for policy optimization-based RL algorithm for average-constrained MDPs. We propose ACPO, a policy optimization algorithm for an average-CMDP  with deep learning for function approximation. Our approach is motivated by theoretical guarantees that bound the difference between the average long-run rewards or costs of different policies. It draws inspiration from CPO \cite{achiam2017constrained} (see also \citet{tessler2018reward} for DCMDPs), which uses a policy improvement theorem for the discounted setting based on the trust region methods of \citet{schulman2015trust}. Unfortunately, this result trivializes for the average setting and hence can't be used. Instead, we derive a new bound that depends on the worst-case level of ``mixture" of the irreducible Markov chain associated with a policy. Our proposed algorithm, ACPO is based on these theoretical developments. For experimental evaluation, we use several OpenAI Gym environments from \citet{todorov2012mujoco}, train large neural network policies and demonstrate the effectiveness and superior performance of the ACPO algorithm as compared to others.

\noindent\textbf{Main Contributions and Novelty.} \textit{Algorithmic:} We introduce the first practical policy optimization-based RL algorithm for average-constrained MDPs with new and tight performance bounds and violation guarantees. The algorithm draws inspiration from CPO (for discounted-CMDPs) and ATRPO (for average-MDPs) but is not a straightforward extension of either. One may posit that setting the discount factor $\gamma = 1$ in CPO for the discounted setting may suffice but that does not perform well on average-CMDPs even with a large discount factor. Further, constraint violation and policy degradation bounds of CPO do not hold in the average setting and hence we develop novel bounds (in Corollary \ref{cor:thm:avg_constraint_imp}). In fact,  the advantage function estimation routine in our algorithm (line 4 and 6 in Algorithm \ref{alg:practical_acpo}) is also different from that in CPO, since the discounted-setting procedure cannot be used for the average setting (see Appendix \ref{appendix:approx_acpo}): We first approximate the average-reward bias and then use a one-step TD backup to estimate the action-bias function. Furthermore, policy optimization algorithms for the average case \cite{zhang2020average, wan2021learning, liao2022batch} cannot incorporate  constraints. We enable this by introducing sublevel sets of cost constraints. We also introduce an approximate but novel line search procedure that improves the empirical performance of our algorithm, an idea that may help improve performance of other  policy optimization algorithms such as  PPO.\\
\textit{Technical:} Since ACPO is a trust region method, one can expect some overlap in analysis techniques with other similar algorithms.  Nevertheless, our analysis has several novel elements: Lemma \ref{lemma:d_and_pi}, where we use eigenvalues of the transition matrix to relate total variation  of stationary distributions with that of the policies, and in Lemma \ref{lemma:infeasible_kl_bound}, we use the sublevel sets of constraints and projection inequality of Bregman divergence. Furthermore, several important results from CPO and ATRPO papers cannot be applied to the analysis of our algorithm. \\
\textit{Empirical:} We evaluate the empirical performance of ACPO in the OpenAI Gym (Mujoco) environments, a standard benchmark. We find that ACPO outperforms all state-of-the-art Deep RL algorithms  such as CPO in \cite{achiam2017constrained}, PCPO in \cite{yang2020projection}, PPO in \cite{schulman2017proximal}, BVF-PPO in \cite{pmlr-v119-satija20a} and ATRPO in \cite{zhang2021average}. We use a large discount factor if the algorithm is not for the average setting, and a Lagrangian objective if it is not for the constrained setting, and in some cases both.\\
\textit{Significance:} ACPO is the first practical  trust region-style policy optimization algorithm for ACMDPs with excellent empirical performance. ACMDPs are important models because they allow incorporation of long term safety constraints, which are important not only in the context of safe robot learning and control, but also safety-constrained RLHF fine-tuning and inference for LLMs \cite{moskovitz2023confronting} and Diffusion models as well. In the absence of suitable policy optimization algorithms for ACMDPs, researchers have resorted to using adaptations of PPO, etc.

\noindent\textbf{Related Work.}
Learning constraint-satisfaction policies has been explored in the Deep RL literature in \cite{9030307, garcia2015comprehensive}. This can either be  (1) through expert annotations and demonstrations as in \cite{rajeswaran2017learning,gao2018reinforcement} or, (2) by exploration with constraint satisfaction as in \cite{achiam2017constrained,tessler2018reward}. While the former approach is not scalable since it requires human interventions, current state-of-the-art algorithms for the latter are not applicable to the average reward setting. 

Previous work on RL with the average reward criterion has mostly attempted to extend stochastic approximation schemes for the tabular setting, such as Q-learning in \cite{abounadi2001learning,wan2021learning}, to the non-tabular setting with function approximation in \cite{wei2021learning, zhang2020average}. \cite{pmlr-v162-chen22i} deals with online learning in a constrained MDP setting, but their aim is regret minimization or exploration, both in tabular settings. 
We are inspired by the work of \cite{zhang2021average} to develop techniques required to derive the policy degradation and constraint violation bounds in Section \ref{sec:algo}.



The more recent works of \cite{Bhatnagar2012} and \cite{calvo2023state} also fail to address our problem setting as 
the former test on a 2x4 queueing network with maximum state space of 128, while the latter test on a grid of size 10x10 (maximum states of 100). In addition to that, the way they incorporate constraints during training is just via a Lagrangian formulation. In our chapter we show that simply doing this (in the case of PPO and ATRPO for example) leads to much inferior performance to ACPO, which can outperform current state-of-the-art algorithms in state spaces of upto $10^{96}$.

\section{Preliminaries}\label{sec:prelims}

A Markov decision process (MDP) is a tuple, ($S,A,r,P,\mu$), where $S$ is the set of states, $A$ is the set of actions, $r : S \times A \times S \to \Rbb$ is the reward function, $P : S \times A \times S \to [0,1]$ is the transition probability function such that $P(s'|s,a)$ is the probability of transitioning to state $s'$ from state $s$ by taking action $a$, and $\mu : S \to [0,1]$ is the initial state distribution. A stationary policy $\pi : S \to \Delta(A)$ is a mapping from states to probability distributions over the actions, with $\pi(a|s)$ denoting the probability of selecting action $a$ in state $s$, and $\Delta(A)$ is the probability simplex over the action space $A$. We denote the set of all stationary policies by $\Pi$. For the average setting, we will make the standard assumption that the MDP is \textit{ergodic} and is \textit{unichain}.

In reinforcement learning, we aim to select a policy $\pi$ which maximizes a performance measure, $J(\pi)$, which, for continuous control tasks is either the discounted reward criterion or the average reward approach. Below, we briefly discuss both formulations.

\subsection{Discounted criterion}

For a given discount factor $\gamma\in (0,1)$, the discounted reward objective is defined as 

\begin{equation*}
\begin{aligned}
    J_{\gamma}(\pi) &:=  \E{\tau\sim\pi} \left[\sum_{t=0}^{\infty}\gamma^t r(s_t,a_t,s_{t+1}) \right] = \frac{1}{1-\gamma}\E{s\sim\dpid\\ a\sim\pi \\ s'\sim P(\cdot | s,a)}[r(s,a,s')]
\end{aligned} 
\end{equation*}

where $\tau$ refers to a sample trajectory of $(s_0, a_0, s_1, \cdots)$ generated when following a policy $\pi$, that is, $a_t \sim \pi(\cdot | s_t)$ and $s_{t+1} \sim P(\cdot | s_t, a_t) \,$; $d_{\pi, \gamma}$ is the \textit{discounted occupation measure} that is defined by $d_{\pi, \gamma} (s) = (1-\gamma) \sum_{t=0}^{\infty} \gamma^t \underset{\tau\sim\pi}{P}(s_t = s)$, which essentially refers to the discounted fraction of time spent in state $s$ while following policy $\pi$. 


\subsection{Average criterion}

The average-reward objective is given by:
\begin{equation}\label{eq:AvgR_obj}
\begin{split}
    J(\pi) &:= \lim_{N\to\infty}\frac{1}{N}\E{\tau\sim\pi}\left[\sum_{t=0}^{N-1} r(s_t,a_t, s_{t+1})\right] = \E{s\sim\dpi\\ a\sim\pi(\cdot|s) \\ s' \sim P(\cdot | s,a)}[r(s,a,s')],
\end{split}
\end{equation}
where $\dpi(s):=\lim_{N\to\infty}\frac{1}{N} \sum_{t=0}^{N-1} P_{\tau\sim\pi}(s_t=s)$ is the \textit{stationary state distribution} under policy $\pi$. The limits in $J(\pi)$ and $\dpi(s)$ are guaranteed to exist under our ergodic assumption. Since the MDP is aperiodic, it can also be shown that $\dpi(s)=\lim_{t\to\infty}P_{\tau\sim\pi}(s_t=s)$. Since we have $\lim_{\gamma\to 1} \dpid(s)\to\dpi(s), \forall s$, it can be shown that $\lim_{\gamma\to 1} (1-\gamma) J_{\gamma}(\pi) =J(\pi)$.

In the average setting, we seek to keep the estimate of the state value function unbiased and hence, introduce the \textit{average-reward bias function} as
\begin{equation*}
    \vfunc(s):= \E{\tau\sim\pi}\left[\sum_{t=0}^{\infty} (r(s_t,a_t,s_{t+1}) - J(\pi)) \; \bigg| \; s_{0}=s\right]
\end{equation*}
and the \textit{average-reward action-bias function} as
\begin{align*}
\qfunc(s,a):= \E{\tau\sim\pi}\bigg[\sum_{t=0}^{\infty} (r(s_t,a_t,s_{t+1}) - & J(\pi)) \; \bigg| \; s_{0}=s,a_{0}=a \bigg].
\end{align*}
Finally, define the \textit{average-reward advantage function} as $ \adv(s,a) := \qfunc(s,a)-\vfunc(s)$.



\subsection{Constrained MDPs}

A constrained Markov decision process (CMDP) is an MDP augmented with constraints that restrict the set of allowable policies for that MDP.  Specifically, we augment the MDP with a set $C$ of auxiliary cost functions, $C_1, \cdots , C_m$ (with each function $C_i : S \times A \times S \to \Rbb$ mapping transition tuples to costs, just like the reward function), and bounds $l_1, \cdots , l_m$. Similar to the value functions being defined for the average reward criterion, we define the average cost objective with respect to the cost function $C_{i}$ as
\begin{equation}
\begin{split}
    J_{C_{i}}(\pi) &:= \lim_{N\to\infty}\frac{1}{N}\E{\tau\sim\pi}\left[\sum_{t=0}^{N-1} C_{i}(s_t,a_t, s_{t+1})\right] = \E{s\sim\dpi\\ a\sim\pi \\ s' \sim P(\cdot | s,a)}[C_{i}(s,a,s')].
\end{split}
\end{equation}
where $J_{C_i}$ will  be referred to as the \textit{average cost} for constraint $C_{i}$. The set of feasible stationary policies for a CMDP then is given by
$
\Pi_{C} := \left\{\pi \in \Pi \; : \; \jfuncp \leq l_i, \forAll i \in \{1, \cdots , M\} \right\}
$.
The goal is to find a policy $\pi^{\star}$ such that 
$\pi^{\star} \in \arg\max_{ \pi \in \Pi_{C} } J(\pi).$

However, finding an exact $\pi^{\star}$ is infeasible for large-scale problems.
Instead,  we aim to derive an iterative policy improvement algorithm that given a current policy, improves upon it by approximately maximizing the increase in the reward, while not violating the constraints by too much and not being too different from the current policy.



Lastly, 
analogous to $\vfunc$, $\qfunc$, and $\adv$, we define similar  quantities for the cost functions  $C_i(\cdot)$, and denote them by $\vfuncc$, $\qfuncc$, and $\advc$.

\subsection{Policy Improvement for discounted CMDPs}



In many on-policy constrained RL problems, we improve policies iteratively by maximizing a predefined function within a local region of the current best policy as in \cite{tessler2018reward, achiam2017constrained, yang2020projection, song2020v}.
\cite{achiam2017constrained} derived a policy improvement bound for the discounted CMDP setting as:

\begin{equation}\label{eq:cpo_improve}
\begin{aligned}
J_{\gamma}(\pikup) - J_{\gamma}(\pik) \geq \frac{1}{1-\gamma} \E{s \sim d^{\pik} \\ a \sim \pikup} \left[ A_{\gamma}^{\pik} (s,a) - \frac{2\gamma \epsilon^{\pikup}}{1-\gamma}  D_{TV} (\pikup||\pik)[s] \right],
\end{aligned}
\end{equation}
where $A_{\gamma}^{\pik}$ is the discounted version of the advantage function, $\epsilon^{\pikup} := \max_s | \Ebb_{a \sim \pikup} [A_{\gamma}^{\pik} (s,a) ] |$, and  $D_{TV}(\pikup||\pik)[s] = (1/2)\sum_a \left| \pikup(a|s) - \pik(a|s) \right|$ is the total variational divergence between $\pikup$ and $\pik$ at $s$. These results laid the foundations for on-policy constrained RL algorithms as in\cite{wu2017scalable,vuong2019supervised}.

However, Equation \eqref{eq:cpo_improve} does not generalize to the average setting ($\gamma \to 1$) (see Appendix \ref{proof:cpo_bound_trivial}). In the next section, we will derive a policy improvement bound for the average  case  and present an algorithm based on trust region methods, which will generate almost-monotonically improving iterative policies. Proofs of theorems and lemmas, if not already given, are available in Appendix \ref{acpo:appendix}.

\section{ACPO: The Average-Constrained Policy Optimization Algorithm}\label{sec:algo}

In this section, we present the main results of our work. For conciseness, we denote by $\dpi\in\Rbb^{|S|}$ the column vector whose components are $\dpi(s)$ and $P_{\pi}\in\Rbb^{|S|\times|S|}$ to be the state transition probability matrix under policy $\pi$.

\subsection{Policy Improvement for the Average-CMDP}

Let $\pi'$ be the policy obtained via some update rule from the current policy $\pi$. Analogous to the discounted setting of a CMDP, we would like to characterize the performance difference $J(\pi')-J(\pi)$ by an expression which depends on $\pi$ and some divergence metric between the two policies.

\begin{restatable}{lemma}{policydiff} \cite{zhang2020average}
\label{lemma:policy_diff}
Under the unichain assumption of the underlying Markov chain, for any stochastic policies $\pi$ and $\pi'$:
\begin{equation}\label{eq:policy_diff}
    J(\pi') - J(\pi) =  \E{s\sim \dpip \\ a\sim\pi'}\left[\adv(s,a)\right].
\end{equation}
\end{restatable}

Note that this difference depends on the stationary state distribution obtained from the \textit{new} policy, $\dpip$. This is computationally impractical as we do not have access to this $\dpip$. Fortunately, by use of the  following lemma we can show that if $\dpi$ and $\dpip$ are ``close'' with respect to some metric, we can approximate Eq.  \eqref{eq:policy_diff} using samples from $\dpi$.

\begin{restatable}{lemma}{policyimpd}
\label{lemma:policy_impd}
Under the unichain assumption, for any stochastic policies $\pi$ and $\pi'$ we have:
\begin{equation}
\label{eq:policy_diff_tv}
\begin{split}
    \left| J(\pi') - J(\pi) -\E{s\sim \dpi\\ a\sim\pi'}\left[\adv(s,a)\right] \right| \leq 2\epsilon\TV{\dpip}{\dpi}
\end{split}
\end{equation}
\end{restatable}
, where $\epsilon=\max_s\big|\E{a\sim\pi'}[\adv(s,a)]\big|$.
See Appendix \ref{proof:policy_impd_proof} for proof. Lemma \ref{lemma:policy_impd} implies $J(\pi')\approx J(\pi) +\E{} [\adv(s,a)]$ when $\dpi$ and $\dpip$ are ``close''. Now that we have established this approximation, we need to study the relation of how the actual change in policies affects their corresponding stationary state distributions. For this,  we turn to standard analysis of the underlying Markov chain of the CMDP. 

Under the ergodic assumption, we have that $P_{\pi}$ is irreducible and hence its eigenvalues $\{\lambda_{\pi, i}\}_{i=1}^{|S|}$ are such that $\lambda_{\pi, 1}=1$ and $\lambda_{\pi, i \neq 1} < 1$. For our analysis, we define $\sigma^{\pi} = \max_{i\neq 1} \, (1-\lambda_{\pi, i})^{-1/2}$, and from \cite{levene2002kemeny} and \cite{doyle2009kemeny}, we connect $\{\lambda_{\pi, i}\}_{i=1}^{|S|}$ to the sensitivity of the stationary distributions to changes in the policy using the result below. 

\begin{restatable}{lemma}{dandpi}
\label{lemma:d_and_pi}
Under the ergodic assumption, the divergence between the stationary distributions $\dpi$ and $\dpip$ is upper bounded as:
\begin{equation}
    \TV{\dpip}{\dpi} \leq \sigma^{\star} \E{s\sim \dpi}[\TV{\pi'}{\pi}[s]],
\end{equation}
\end{restatable}
, where $\sigma^{\star} = \max_{\pi} \sigma^{\pi}$. See Appendix \ref{proof:d_and_pi} for proof. This bound is tighter and easier to compute than the one given by \cite{zhang2021average}, which replaces $\sigma^{\star}$ by $\kappa^{\star} = \max_{\pi} \kappa^{\pi}$, where $\kappa^{\pi}$ is known as  \textit{Kemeny's constant} from \cite{kemeny1960finite}. It is interpreted as the expected number of steps to get to any goal state, where the expectation is taken with respect to the stationary-distribution of those states. 

Combining the bounds in Lemma \ref{lemma:policy_impd} and Lemma \ref{lemma:d_and_pi} gives us the following result:

\begin{proposition}\label{prop:avg_policy_imp}
Under the ergodic assumption, the following bounds hold for any stochastic policies $\pi$ and $\pi'$:
\begin{equation}\label{eq:avg_policy_imp}
 L_{\pi}^{-}(\pi')\leq J(\pi') - J(\pi)\leq L_{\pi}^{+}(\pi') 
\end{equation}
where
\begin{align*}
    L_{\pi}^{\pm}(\pi') &= \E{s\sim \dpi\\ a\sim\pi'}\left[\adv(s,a)\right] \pm 2 \nu \E{s\sim \dpi}[\TV{\pi'}{\pi}[s]]  \\ \text{and} \quad \nu &= \sigma^{\star} \max_{s} \big| \E{a\sim\pi'}[\adv(s,a)] \big|.
\end{align*}

\end{proposition}

It is interesting to compare the inequalities of Equation \eqref{eq:avg_policy_imp} to Equation \eqref{eq:policy_diff}. The term $\E{}[\adv(s,a)]$ in Prop. \ref{prop:avg_policy_imp} is somewhat of a \textit{surrogate} approximation to $J(\pi') - J(\pi)$, in the sense that it uses $d_{\pi}$ instead of $d_{\pi'}$. As discussed before, we do not have access to $d_{\pi'}$ since the trajectories of the new policy are not available unless the policy itself is updated. This surrogate is a first order approximation to $J(\pi') - J(\pi)$ in the parameters of $\pi'$ in a neighborhood around $\pi$ as in \cite{kakade2002approximately}. Hence, Eq.  \eqref{eq:avg_policy_imp} can be viewed as bounding the worst-case approximation error.

Extending this discussion to the cost function of our CMDP, similar expressions follow immediately.

\begin{corollary}
\label{cor:thm:avg_constraint_imp}
For any policies $\pi', \pi$, and any cost function $C_i$, the following bound holds:

\begin{equation}\label{eq:avg_constraint_imp}
 M_{\pi}^{-}(\pi')\leq J_{C_i} (\pi') - J_{C_i} (\pi) \leq M_{\pi}^{+}(\pi') 
\end{equation}
where
\begin{align*}
    M_{\pi}^{\pm}(\pi') &= \E{s\sim \dpi\\ a\sim\pi'}\left[\advc(s,a)\right] \pm 2 \nu_{C_{i}} \E{s\sim \dpi}[\TV{\pi'}{\pi}[s]] \\ \text{and} \quad \nu_{C_{i}} &= \sigma^{\star} \max_{s} \big| \E{a\sim\pi'}[\advc(s,a)]\big|.
\end{align*}
\end{corollary}

Until now, we have been dealing with bounds given with regards to the TV divergence of the policies. However, in practice, bounds with respect to the KL divergence of policies is more commonly used as in \cite{schulman2015trust, schulman2016high, ma2021average}. 
From Pinsker’s and Jensen's inequalities, we have that 
\begin{equation}\label{eq:tv-kl}
\begin{aligned}
  \E{s\sim\dpi}\big[\TV{\pi'}{\pi}[s]\big] 
  &\leq \sqrt{\E{s\sim\dpi}\big[\KL{\pi'}{\pi}][s]\big]/2}.
\end{aligned}
\end{equation}

We can thus use Eq. \eqref{eq:tv-kl} in the bounds of Proposition  \ref{prop:avg_policy_imp} and Corollary \ref{cor:thm:avg_constraint_imp} to make policy improvement guarantees, i.e., if we find updates such that $\pikup \in \argmax_{\pi} L_{\pik}^{-}(\pi)$, then we will have monotonically increasing policies as, at iteration $k$, $\E{s\sim\dpik,  a\sim\pi}[\advk(s,a)]=0$, $\E{s\sim\dpik}[\KL{\pi}{\pik}[s]]=0$ for $\pi=\pik$, implying that $J(\pikup)-J(\pik)\geq 0$. However, this sequence does not guarantee constraint satisfaction at each iteration, so we now turn to trust region methods to incorporate constraints, do policy improvement and provide safety guarantees.

\subsection{Trust Region Based Approach}

For large or continuous state and action CMDPs, solving for the exact optimal policy is impractical. However, \textit{trust region}-based policy optimization algorithms have proven to be effective for solving such problems as in \cite{schulman2015trust, schulman2016high, schulman2017proximal, achiam2017advanced}.
For these approaches, we usually consider some parameterized policy class $\Pi_{\Theta} = \{\pi_{\theta}: \theta \in \Theta \}$ for tractibility. In addition, for CMDPs, we also require the policy iterates to be feasible, so instead of optimizing just over $\Pi_{\Theta}$, we optimize over $\Pi_{\Theta} \cap \Pi_C$. However, it is much easier to solve the above problem if we introduce hard constraints, rather than limiting the set to $\Pi_{\Theta} \cap \Pi_C$. Therefore, we now introduce the ACPO algorithm, which is inspired by the trust region formulations above as the following optimization problem:
\begin{equation}\label{eq:acpo_trust}
\begin{aligned}
  \underset{\pi\in\Pi_{\Theta}}{\text{maximize}} & \quad
\E{s\sim\dpolk\\ a\sim\pi}[\wb{A}^{\polk}(s,a)] \\
\text{s.t.} &\qquad J_{C_i} (\polk) + \E{s\sim\dpolk\\ a\sim\pi}[\wb{A}_{C_{i}}^{\polk}(s,a)] \leq l_i, \;\;\; \forAll i \\
& \qquad \avKL{\pi}{\polk} \leq \delta
\end{aligned}
\end{equation}

where $\avKL{\pi}{\polk} := \E{s\sim\dpolk}[\KL{\pi}{\polk}[s]]$, $\wb{A}^{\polk}(s,a)$ is the average advantage function defined earlier, and $\delta > 0$ is a step size. We use this form of updates as it is an approximation to the lower bound given in Proposition \ref{prop:avg_policy_imp} and the upper bound given in Corollary \ref{cor:thm:avg_constraint_imp}.

In most cases, the trust region threshold for formulations like Eq. \eqref{eq:acpo_trust} are heuristically motivated. We now show that it is quantitatively motivated and comes with a worst case performance degradation and constraint violation. Proof is in Appendix \ref{proof:trust_proof}.

\begin{restatable}{theorem}{trustdegradationviolation}
\label{th:trust_degradation_violation}
Let $\polkup$ be the optimal solution to Eq.  \eqref{eq:acpo_trust} for some $\polk\in\Pi_{\Theta}$. Then, we have
\begin{align}
J(\polkup)-J(\polk)\geq&-\sqrt{2(\delta+V_{max})}\nu^{\polkup} \\
\text{and} \;\; J_{C_{i}}(\polkup)\leq l_{i} & +\sqrt{2(\delta+V_{max})}\nu^{\polkup}_{C_{i}} \forAll i,
\end{align}
where $\nu^{\polkup}=\sigma^{\polkup}\max_{s}\big|\E{a\sim\polkup}[\wb{A}^{\polk}(s,a)]\big|$, $\nu^{\polkup}_{C_{i}} = \sigma^{\polkup} \max_{i,s}\big|\E{a\sim\polkup}[\wb{A}_{C_{i}}^{\polk}(s,a)]\big|$, $V_{max} = \max_{i} \beta_{i}^{2}$, ~\text{and}~ $\beta_{i} = [J_{C_{i}}(\polk)-l_{i}]_{+}$.
\end{restatable}

 \textbf{Remark 3.7.} Note that if the constraints are ignored (by setting $V_{max}=0$), then this bound is tighter than given in \cite{zhang2021average} for the unconstrained average-reward setting.
 
However, the update rule of Eq.  \eqref{eq:acpo_trust} is difficult to implement in practice as it takes steps that are too small, which degrades convergence. In addition, it requires the exact knowledge of $\wb{A}^{\polk}(s,a)$ which is computationally infeasible for large-scale problems. In the next section, we will introduce a specific sampling-based practical algorithm to alleviate these concerns.

\section{Practical Implementation of ACPO}
\label{sec:acpo_implementation}

In this section, we introduce a practical version of the ACPO algorithm with a principle recovery method. With a small step size $\delta$, we can approximate the reward function and constraints with a first order expansion, and approximate the KL divergence constraint with a second order expansion. This gives us a new optimization problem which can be solved exactly using Lagrangian duality.

\subsection{An Implementation of ACPO}

Since we are working with a parameterized class, we shall now overload notation to use $\theta_{k}$ as the policy at iteration $k$, i.e., $\theta_{k} \equiv \polk$. In addition, we use $g$ to denote the gradient of the advantage function objective, $a_{i}$ to denote the gradient of the advantage function of  the cost $C_{i}$, $H$ as the Hessian of the KL-divergence. Formally,
{\small
\begin{align*}
    g &:= \nabla_\theta\E{\substack{s\sim d_{\theta_k} \\ a\sim \theta}}[\wb{A}^{\theta_k}(s,a)] \Bigr\vert_{\theta=\theta_k}, \\ 
    a_{i} &:= \nabla_\theta\E{\substack{s\sim d_{\theta_k} \\ a\sim \theta}}[\wb{A}^{\theta_k}_{C_{i}}(s,a)] \Bigr\vert_{\theta=\theta_k}, \\ 
    H &:= \nabla^2_\theta \E{s\sim d_{\theta_k}}\big[\KL{\theta}{\theta_k})[s]\big] \Bigr\vert_{\theta=\theta_k}.
\end{align*}
}
In addition, let $c_i := J_{C_i}(\theta_k) - l_i$. The approximation to the problem in Eq.  \eqref{eq:acpo_trust} is: 
\begin{equation}
\label{eq:acpo_approx}
\begin{aligned}
\max_{\theta} \;\;\; &  g^T (\theta - \theta_k) & \\
\text{s.t.} & \quad c_i + a_i^T (\theta - \theta_k) \leq 0, \forAll i \\ \text{and,}& \quad \tfrac{1}{2} (\theta - \theta_k)^T H (\theta - \theta_k) \leq \delta.
\end{aligned}
\end{equation}
This is a convex optimization problem in which strong duality holds, and hence it can be solved using a Lagrangian method. The update rule for the dual problem then takes the form
\begin{equation}
\theta_{k+1} = \theta_{k} + \frac{1}{\lambda^{\star}} H^{-1} \left(g - A \mu^{\star}\right). 
\label{eq:acpo_dual_update}
\end{equation}

where $\lambda^{\star}$ and $\mu^{\star}$ are the Lagrange multipliers satisfying the dual
\begin{equation}
\begin{aligned}
\max_{\begin{subarray}{c} \lambda \geq 0 \\ \mu \succeq 0\end{subarray}} \frac{-1}{2\lambda} \left( g^T H^{-1} g - 2 r^T \mu + \mu^T S \mu\right) + \mu^T c - \frac{\lambda \delta}{2},
\end{aligned} 
\label{eq:acpo_dual}
\end{equation}

with $r := g^T H^{-1} A$, $S := A^T H^{-1} A$, $A := [a_{1} , \cdots , a_{m}]$, and $c := [c_{1}, \cdots , c_{m}]^T$.

\subsection{Feasibility and Recovery}
\label{sec:recovery}

The approximation regime described in Eq.  \eqref{eq:acpo_approx} requires $H$ to be invertible. For large parametric policies, $H$ is computed using the conjugate gradient method as in \cite{schulman2015trust}. However, in practice, using this approximation along with the associated statistical sampling errors, there might be potential violations of the approximate constraints leading to infeasible policies. 

To rectify this, for the case where we only have one constraint, one can recover a feasible policy by applying a recovery step inspired by the TRPO update on the cost surrogate as:
\begin{equation}
\begin{aligned}
\theta_{k+1/2} = \theta_{k} - \sqrt{2\delta} \bigg[t \cdot \frac{H^{-1} a}{\sqrt{a^T H^{-1} a}} + (1-t) \cdot \frac{H^{-1} g}{\sqrt{g^T H^{-1} g}} \bigg]
\end{aligned}
\label{eq:acpo_recovery}
\end{equation}

where $t \in [0,1]$. 
Contrasting with the policy recovery update of \cite{achiam2017constrained} which only uses the cost advantage function gradient $a$, we introduce the reward advantage function gradient $g$ as well. This choice is to ensure recovery while simultaneously balancing the ``regret'' of not choosing the best (in terms of the objective value) policy $\pik$. In other words, we wish to find a  policy $\pi_{k+1/2}$ as close to $\pik$ in terms of their objective function values. We follow up this step with a simple linesearch to find feasible $\pikup$. Based on this, Algorithm \ref{alg:practical_acpo} provides a basic outline of ACPO. For more details of the algorithm, see Appendix \ref{appendix:approx_acpo}.


\begin{algorithm}[!t]
   \caption{Average-Constrained Policy Optimization (ACPO)}
   \label{alg:practical_acpo}
\begin{algorithmic}[1]
   \STATE {\bfseries Input:} Initial random policy $\pi_0 \in \Pi_{\theta}$
	 \FOR{$k = 0,1,2,...,K$} 
	 \STATE Sample a set of trajectories $\Omega$ using $\pi_k = \polk$
	 \STATE Find estimates of $g, a, H, c$ using $\Omega$ 
	 \IF{a feasible solution to Equation \eqref{eq:acpo_approx} exists}
	 	\STATE Solve dual problem in Equation \eqref{eq:acpo_dual} for $\lambda^{\star}_k, \mu^{\star}_k$ 
	 	\STATE Find policy update $\pikup$ with Equation \eqref{eq:acpo_dual_update}
	 \ELSE
	 	\STATE Find recovery policy $\pi_{k+1/2}$ with Equation \eqref{eq:acpo_recovery}
	 	\STATE Obtain $\pikup$ by linesearch till approximate constraint satisfaction of Equation \eqref{eq:acpo_approx}
	 \ENDIF
	\ENDFOR
\end{algorithmic}
\end{algorithm}


\begin{figure*}[t]
 \captionsetup[subfigure]{justification=centering}
    \hspace{0.25cm} Average Rewards: \newline
    {
        \includegraphics[width=0.32\textwidth]{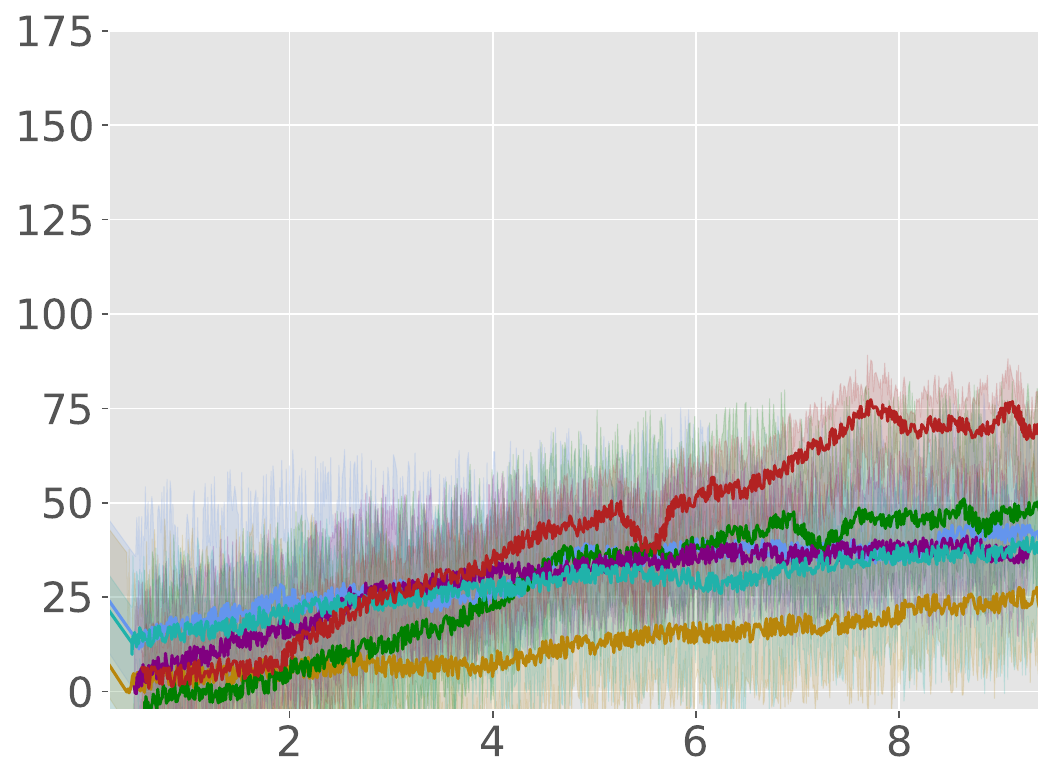}
        \label{fig:ant_gather_rewards}
    }
    {
        \includegraphics[width=0.32\textwidth]{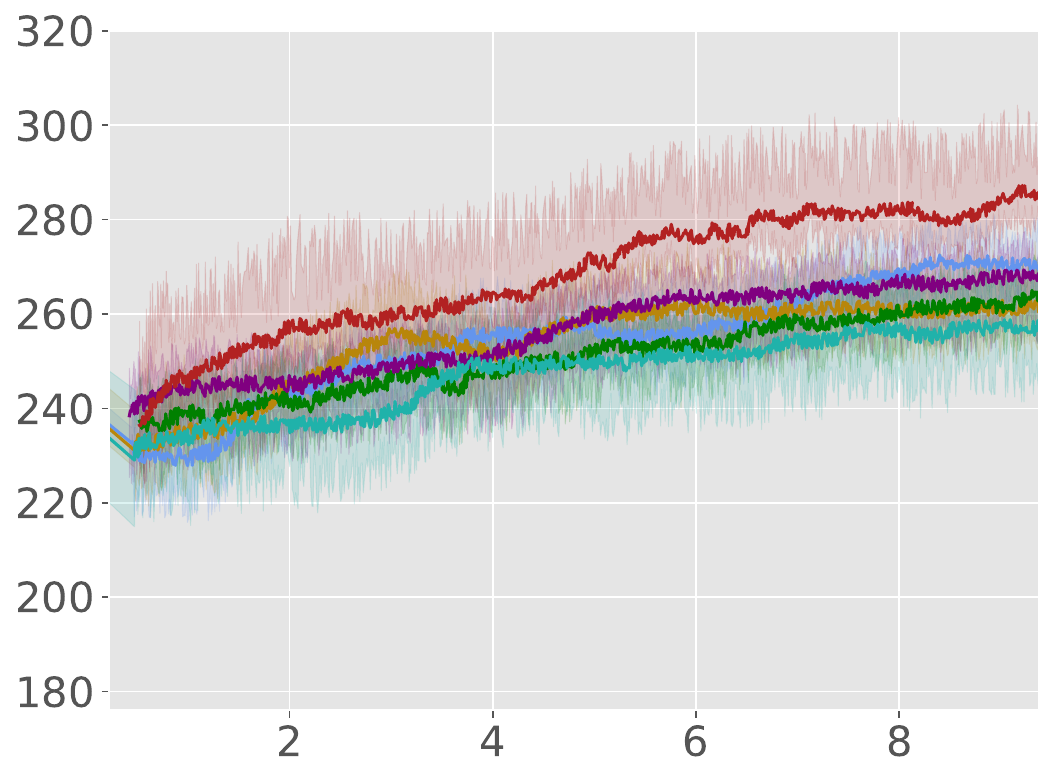}
        \label{fig:bottleneck_rewards}
    }
    {
        \includegraphics[width=0.32\textwidth]{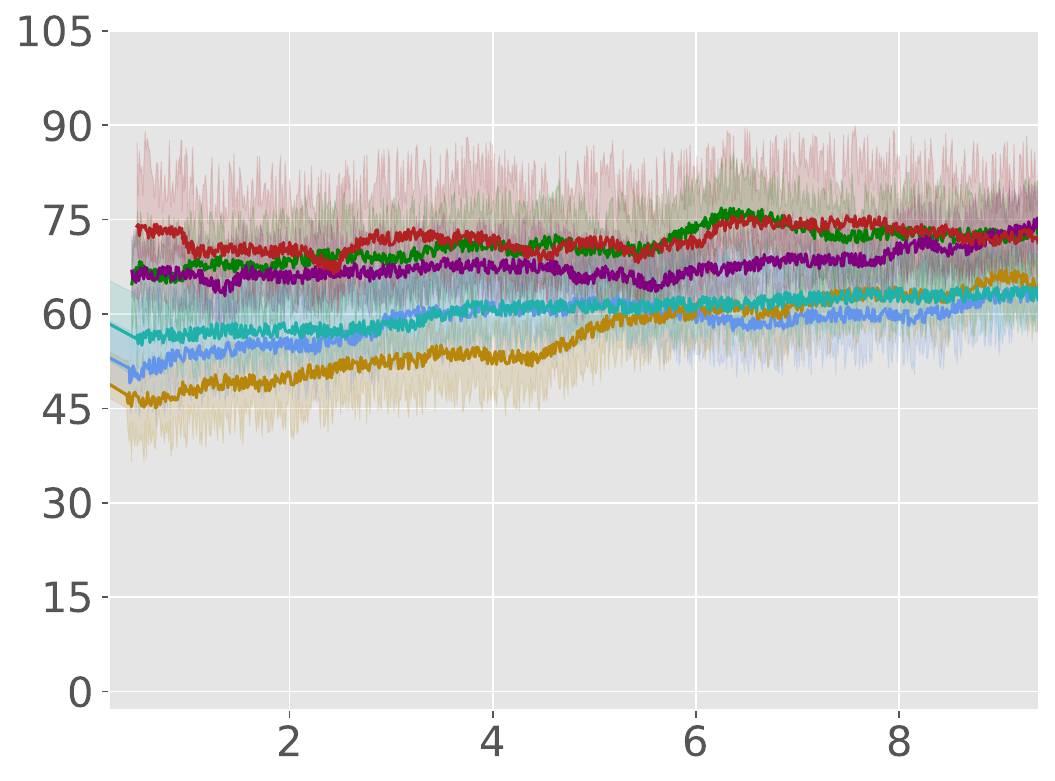}
        \label{fig:grid_rewards}
    }
    \rule{\linewidth}{0.5pt}
    \hspace{2cm} Average Constraint values: \hspace{1.75cm} \includegraphics[width=0.6\textwidth]{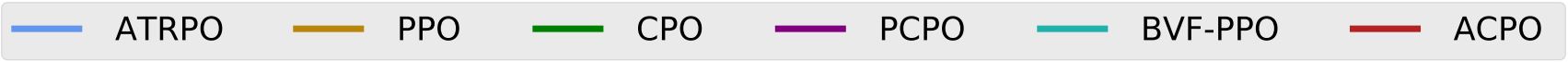} \newline
    \subfloat[Ant Gather]{
        \includegraphics[width=0.32\textwidth]{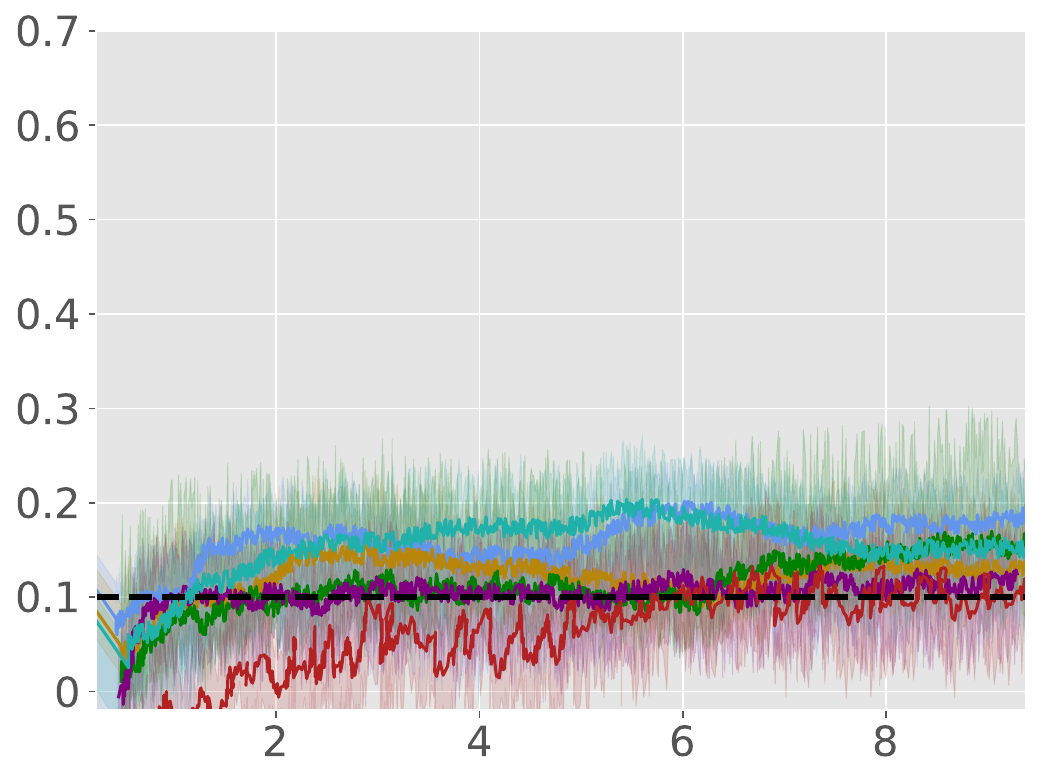}
        \label{fig:ant_gather_costs}
    }
    \subfloat[Bottleneck]{
        \includegraphics[width=0.32\textwidth]{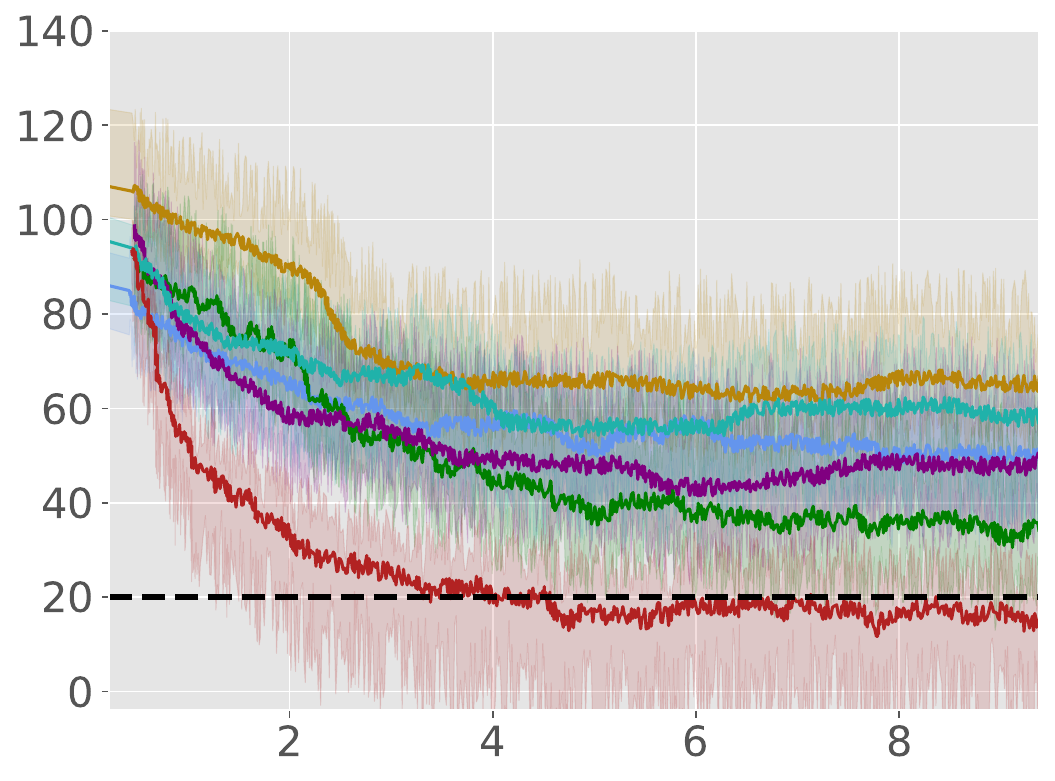}
        \label{fig:bottleneck_costs}
    }
    \subfloat[Grid]{
        \includegraphics[width=0.32\textwidth]{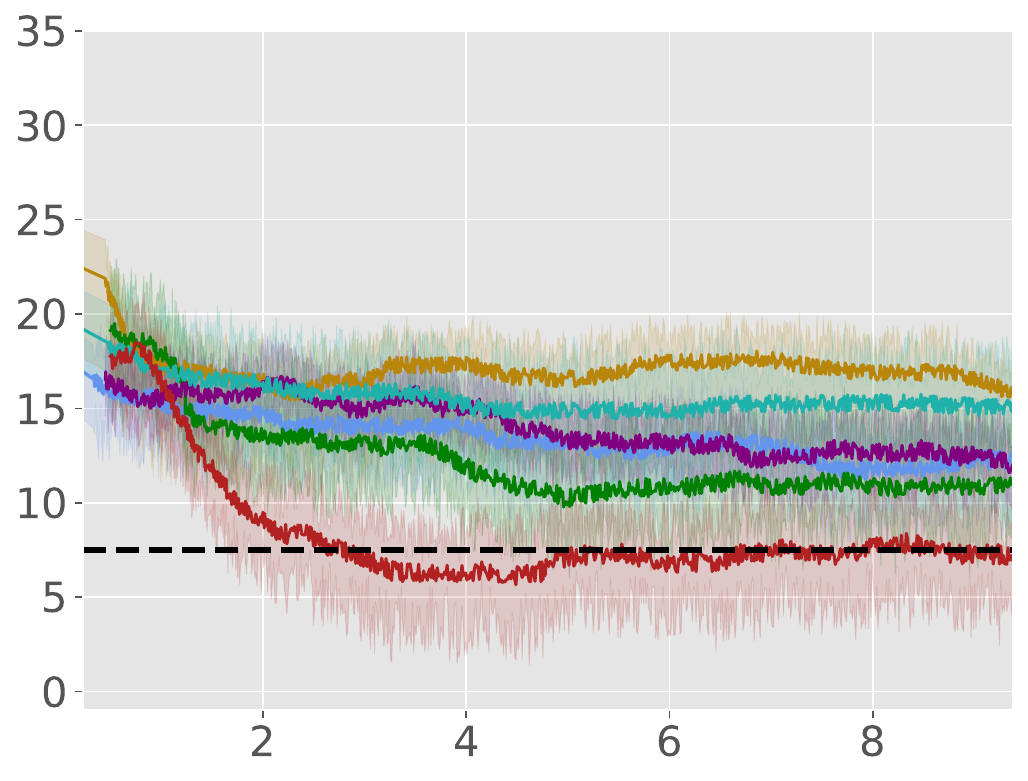}
        \label{fig:grid_costs}
    }
    \caption{The average reward and constraint cost function values vs iterations (in $10^{4}$) learning curves for some algorithm-task pairs. Solid lines in each figure are the empirical means, while the shaded area represents 1 standard deviation, all over 5 runs. The dashed line in constraint plots is the constraint threshold $l$. ATRPO and PPO are tested with constraints, which are included in their Lagrangian formulation. Additional results are available in Appendix \ref{appendix:additional_results}.}
    \label{fig:rewards_costs_comparison}
\end{figure*}


\section{Empirical Results}
\label{sec:empirical}
We conducted a series of experiments to evaluate the relative performance of the ACPO algorithm and  answer the following questions: 
(i) Does ACPO learn a sequence of constraint satisfying policies while maximizing the average reward in the long run?
(ii) How does ACPO compare with the already existing constraint policy optimization algorithms which are applied with a large discount factor?
(iii) What are the factors that affect the performance of ACPO? 

We work with the OpenAI Gym  environments to train the various learning agent on the following tasks - \textit{Gather}, \textit{Circle}, \textit{Grid}, and \textit{Bottleneck} tasks (see Figure \ref{fig:env_overview} in Appendix \ref{appendix:environments} for more details on the environments). For our experiments we only work with a single constraint with policy recovery using Eq.  \eqref{eq:acpo_recovery} (this is only a computational limitation; ACPO in principle can handle multiple constraints). We compare ACPO with the following baseline algorithms: CPO by \cite{achiam2017constrained}, ATRPO by \cite{zhang2021average}, PCPO by \cite{yang2020projection} (a close variant of CPO), BVF-PPO by \cite{pmlr-v119-satija20a} and PPO by \cite{schulman2017proximal}.

Although ATRPO and PPO originally do not incorporate constraints, for fair comparison, we introduce constraints using a Lagrangian. Also, CPO, PCPO and PPO are compared with $\gamma=0.999$. See Appendix \ref{appendix:experimental_details} for more details.

\subsection{Evaluation Details and Protocol}

For the Gather and Circle tasks we test two distinct agents:  a point-mass ($S \subseteq \Rbb^{9}, A \subseteq \Rbb^{2}$), and an ant robot ($S \subseteq \Rbb^{32}, A \subseteq \Rbb^{8}$). The agent in the Bottleneck task in $S \subseteq \Rbb^{71}, A \subseteq \Rbb^{16}$, and for the Grid task is $S \subseteq \Rbb^{96}, A \subseteq \Rbb^{4}$. We use two hidden layer neural networks to represent Gaussian policies for the tasks. For Gather and Circle, size is (64,32) for both layers, and for Grid and Bottleneck the layer sizes are (16,16) and (50,25). We set the step size $\delta$ to $10^{-4}$, and for each task, we conduct 5 runs to get the mean and standard deviation for reward objective and cost constraint values during training. We train CPO, PCPO, and PPO with the discounted objective, however, evaluation and comparison with BVF-PPO, ATRPO and ACPO is done using the average reward objective (this is a standard evaluation scheme  as in \cite{schulman2015trust, wu2017scalable, vuong2019supervised}). 

For each environment, we train an agent for $10^{5}$ steps, and for every $10^{3}$ steps, we instantiate 10 evaluation trajectories with the current (deterministic) policy. For each of these trajectories, we calculate the trajectory average reward for the next $10^{3}$ steps and finally report the total average-reward as the mean of these 10 trajectories. Learning curves for the algorithms are compiled in Figure \ref{fig:rewards_costs_comparison} (for Point-Circle, Point-Gather, and Ant-Circle see Appendix \ref{appendix:additional_results}). 

Since there are two objectives  (rewards in the objective and costs in the constraints), we show the plots which maximize the reward objective while satisfying the cost constraint. See Appendix \ref{appendix:practical_acpo} and \ref{appendix:experimental_details} for more details.


\subsection{Performance Analysis}

From Figure \ref{fig:rewards_costs_comparison}, we can see that ACPO is able to improve the reward objective while having approximate constraint satisfaction on all tasks. In particular, ACPO is the only algorithm that best learns almost-constraint-satisfying maximum average-reward policies across all tasks: in a simple Gather environment, ACPO is able to almost exactly track the cost constraint values to within the given threshold $l$;  however, for the high dimensional Grid and Bottleneck environments we have more constraint violations due to complexity of the policy behavior. Regardless, in these environments, ACPO still outperforms all other baselines. 

\noindent\textbf{ACPO vs. CPO/PCPO.} For the Point-Gather environment (see Figure \ref{fig:appendix_rewards_costs_comparison}),  we see that initially ACPO and CPO/PCPO give relatively similar performance, but eventually ACPO improves over CPO and PCPO by 52.5\% and 36.1\% on average-rewards respectively. This superior performance does not come with more constraint violation. The Ant-Gather environment particularly brings out the effectiveness of ACPO where it shows 41.1\% and 61.5\% improvement over CPO and PCPO respectively, while satisfying the constraint. In the high dimensional Bottleneck and Grid environments, ACPO is particularly quick at optimizing for low constraint violations, while improving over PCPO and CPO in terms of average-reward. 

\noindent\textbf{ACPO vs Lagrangian ATRPO/PPO.} One could suppose to use the state of the art unconstrained policy optimization algorithms with a Lagrangian formulation to solve the average-rewards CMDP problem in consideration, but we see that such an approach, although principled in theory, does not give satisfactory empirical results. This can be particularly seen in the Ant-Circle, Ant-Gather, Bottleneck, and Grid environments, where Lagrangian ATRPO and PPO give the least rewards, while not even satisfying the constraints. If ATRPO and PPO were used with constraints ignored, one would see higher rewards but even worse constraint violations, which are not useful.

\noindent\textbf{ACPO vs BVF-PPO.} BVF-PPO is a whole different formulation than the other baselines, as it translates the cumulative cost constraints into state-based constraints, which results in an almost-safe policy improvement method which maximizes returns at every step. However, we see that this approach fails to satisfy the constraints even in the moderately difficult Ant Gather environment, let alone the high dimensional Bottleneck and Grid environments.



\subsection{Dependence of the Recovery Regime}
\label{subsec:recovery_regime}

\begin{figure}[t]
 \captionsetup[subfigure]{justification=centering}
    \centering
    \subfloat[Rewards]{  \includegraphics[width=0.22\textwidth]{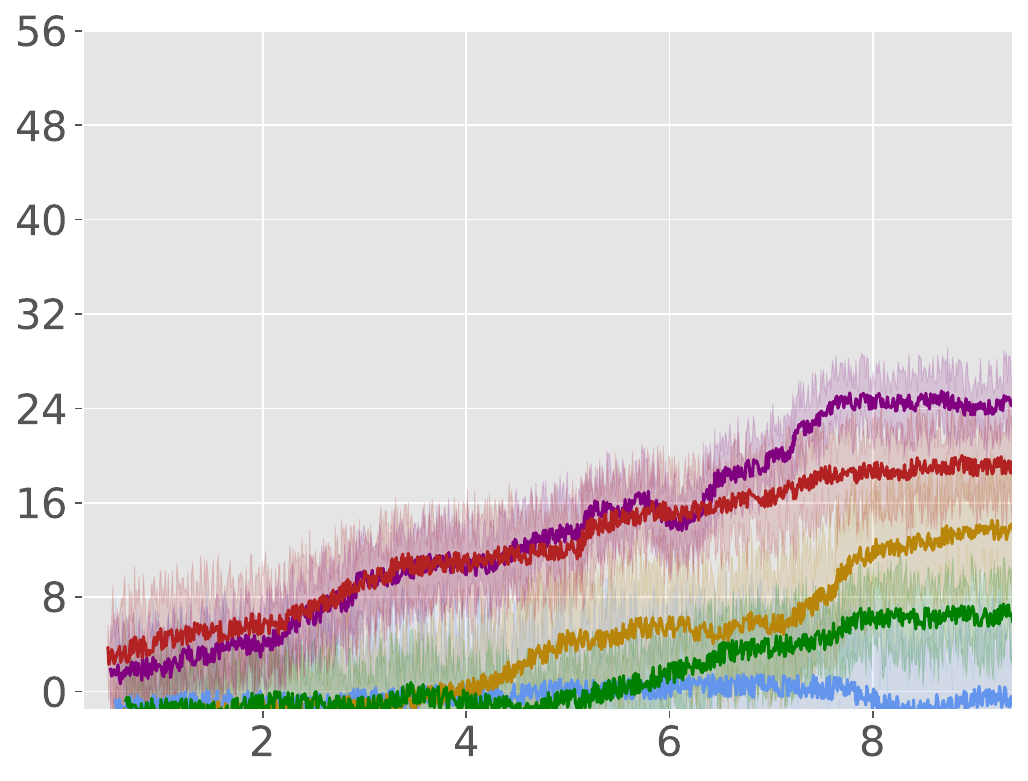}
        \label{fig:point_circle_hyper_rewards}
    }
    \subfloat[Costs]{
        \includegraphics[width=0.22\textwidth]{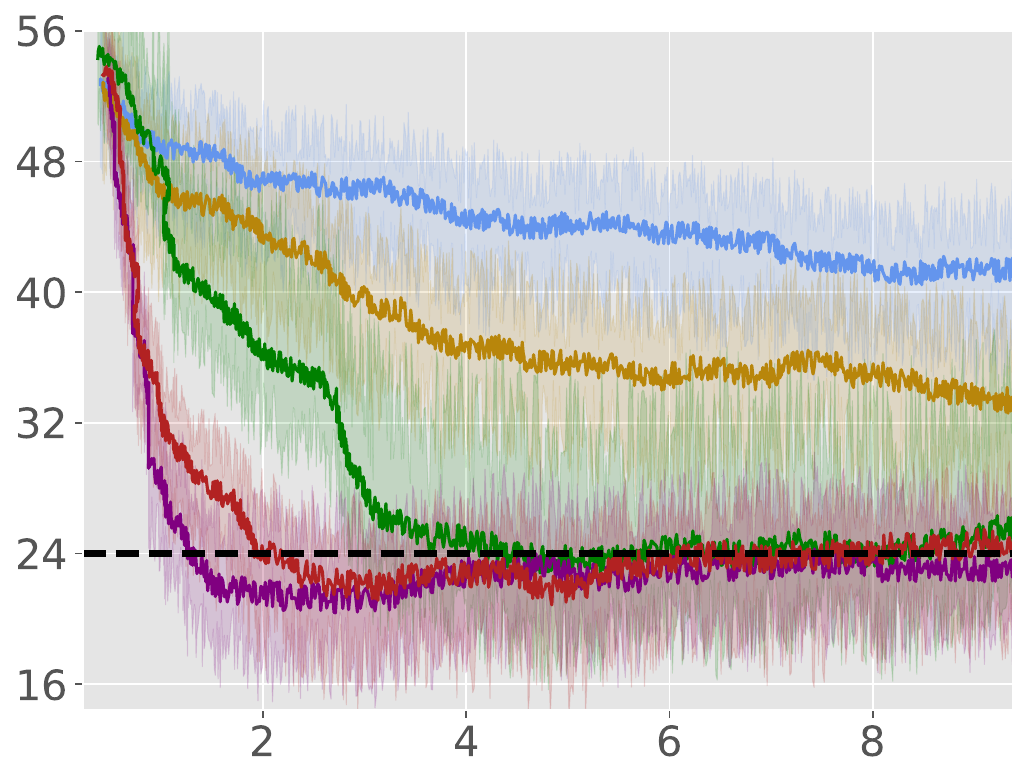}
        \label{fig:point_circle_hyper_costs}
    } 
    
     \includegraphics[width=0.45\textwidth]{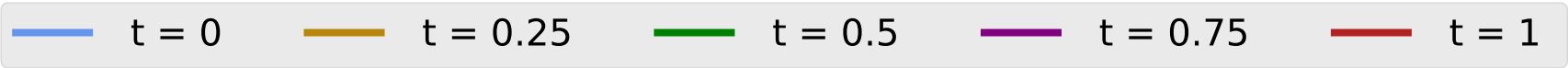}
    \caption[]{Comparison of performance of ACPO with different values of the hyperparameter $t$ in the Point-Circle environment. X-axis is iterations in $10^{4}$. See Appendix \ref{appendix:additional_results} for more details.}
    \label{fig:hyperparam_t_main}
\end{figure}

In Equation \eqref{eq:acpo_recovery} we introduced a hyperparameter $t$, which provides for an intuitive trade-off as follows: either we purely decrease the constraint violations ($t=1$), or we decrease the average-reward ($t=0$), which consequently decreases the constraint violation. The latter formulation is principled in that if we decrease rewards, we are bound to decrease constraints violation due to the nature of the environments. Figure \ref{fig:hyperparam_t_main} shows the experiments we conducted with varying $t$. With $t=1$, we obtain the same recovery scheme as that of \cite{achiam2017constrained}. Our results show that this scheme does not lead to the best performance, and that $t=0.75$ and $t=1$ perform the best across all tasks. See Appendix \ref{appendix:additional_results} for a detailed study.






\clearpage

\newpage

\section{Appendix}
\label{acpo:appendix}

\subsection{Proofs}


\begin{lemma}[Trivialization of Discounted Criterion Bounds]
\label{proof:cpo_bound_trivial}
Consider the policy performance bound of \cite{achiam2017constrained}, which says that for any two stationary policies $\pi$ and $\pi'$:
\begin{equation}\label
{eq:pol_imp_disc}
    J_{\gamma}(\pi')-J_{\gamma}(\pi) \geq \frac{1}{1-\gamma}\left[\E{\substack{s\sim \dpid \\ a\sim\pi'}}[\advd(s,a)] - \frac{2\gamma\epsilon^{\gamma}}{1-\gamma}\E{s\sim\dpid}\TV{\pi'}{\pi}[s]\right]
\end{equation}
where $\epsilon^{\gamma} = \max_s\left|\E{a\sim\pi'}[\advd(s,a)]\right|$. Then, the right hand side times $1-\gamma$ goes to negative infinity as $\gamma\to 1$.
\end{lemma}
\begin{proof}
Since $\dpid$ approaches the stationary distribution $\dpi$ as $\gamma\to 1$, we can multiply the right hand side of \eqref{eq:pol_imp_disc} by $(1-\gamma)$ and take the limit which gives us:
\begin{align*}
    &\lim_{\gamma\to 1}\left(\E{\substack{s\sim \dpid \\ a\sim\pi'}}[\advd(s,a)] \pm \frac{2\gamma\epsilon^{\gamma}}{1-\gamma}\E{s\sim\dpid}\TV{\pi'}{\pi}[s]\right) \\
    =& \E{\substack{s\sim \dpi \\ a\sim\pi'}}[\adv(s,a)] - 2\epsilon \E{s\sim\dpi}\big[\TV{\pi'}{\pi}[s]\big] \cdot \lim_{\gamma\to 1}\frac{\gamma}{1-\gamma} \\
    =& -\infty
\end{align*}
Here $\epsilon = \max_s\left|\E{a\sim\pi'}[\adv(s,a)]\right|$. The first equality is a standard result of $\lim_{\gamma\to 1}\advd(s,a) = \adv(s,a)$.
\end{proof}

\policydiff*
\begin{proof}
\label{proof:policy_diff_proof}
    We directly expand the right-hand side using the definition of the advantage function and  Bellman equation, which gives us:
\begin{align*}
    \E{\substack{s\sim d_{\pi'}\\ a\sim\pi'}}\left[\adv(s,a)\right] 
    &= \E{\substack{s\sim d_{\pi'}\\ a\sim\pi'}}\left[\qfunc(s,a) - \vfunc(s)\right] \\
    &= \E{\substack{s\sim d_{\pi'}\\ a\sim\pi' \\ s'\sim P(\cdot|s,a)}}\left[r(s,a,s') - J(\pi) + \vfunc(s') - \vfunc(s)\right] \\
    &= J(\pi') - J(\pi) + \underbrace{\E{\substack{s\sim d_{\pi'}\\ a\sim\pi'\\ s'\sim P(\cdot|s,a)}}[\vfunc(s')] - \E{s\sim d_{\pi'}}[\vfunc(s)]}_{A}
\end{align*}
Analyzing $A$, since $\dpip(s)$ is a stationary distribution:
\begin{align*}
    \E{\substack{s\sim d_{\pi'}\\ a\sim\pi'\\ s'\sim P(\cdot|s,a)}}[\vfunc(s')] &= \sum_{s}\dpip(s)\sum_{a}\pi'(a|s)\sum_{s'}P(s'|s,a)\vfunc(s') \\ &= \sum_{s}\dpip(s)\sum_{s'}P_{\pi'}(s'|s)\vfunc(s') = \sum_{s'}\dpip(s')\vfunc(s')
\end{align*}
Therefore, $A = \sum_{s'}\dpip(s')\vfunc(s') - \E{s\sim d_{\pi'}}[\vfunc(s)] = 0$. Hence, proved.
\end{proof}

\policyimpd*
\begin{proof}
\label{proof:policy_impd_proof}
\begin{align*}
    \left|J(\pi') - J(\pi) - \E{s\sim \dpi \\ a\sim\pi'}\left[\adv(s,a)\right]\right| &= \left|\E{s\sim d_{\pi'} \\ a\sim\pi'}\left[\adv(s,a)\right] - \E{s\sim \dpi\\ a\sim\pi'}\left[\adv(s,a)\right]\right| \tag{from Lemma \ref{lemma:policy_diff}} \\
    &=\left|\sum_s \E{a\sim\pi'}\left[\adv(s,a)\right]\left(\dpip(s) - \dpi(s)\right)\right| \\
    &\leq \sum_s\left|\E{a\sim\pi'}\left[\adv(s,a)\right]\left(\dpip(s) - \dpi(s)\right)\right| \\
    &\leq \max_s\left|\E{a\sim\pi'}\left[\adv(s,a)\right]\right|\norm{\dpip - \dpi}{1}{} \tag{Holder's inequality} \\
    &= 2\epsilon\TV{d_{\pi'}}{d_{\pi}}
\end{align*}
\end{proof}


\dandpi*
\begin{proof}
\label{proof:d_and_pi}
This proof takes ideas from Markov chain perturbation theory in \cite{cho2001comparison,hunter2005stationary, zhang2020average}. Firstly we state a standard result with $P_{\pi}^{\star} =\textbf{1}\dpi^T$
$$
 (d_{\pi'}-d_{\pi})^T(I-P_{\pi'}+P_{\pi'}^{\star}) = d_{\pi'}^T-d_{\pi}^T - d_{\pi'}^T + d_{\pi}^T P_{\pi'} 
    = d_{\pi}^T P_{\pi'} - d_{\pi}^T
    = d_{\pi}^T(P_{\pi'}-P_{\pi}).
$$

Denoting the fundamental matrix of the Markov chain $\Zpip =( I-P_{\pi'}+P_{\pi'}^{\star})^{-1}$ and the mean first passage time matrix $\Mpip = (I-\Zpip + E \Zpip_{\dg})D^{\pi'}$, and right multiplying the above by $(\Zpip)^{-1}$  we have, 
\begin{align}
\label{eq:d_and_pi_midstep}
d_{\pi'}^T-d_{\pi}^T = d_{\pi}^T(P_{\pi'}-P_{\pi})(I - \Mpip (D^{\pi'})^{-1}) &\implies d_{\pi'}-d_{\pi} = (I - \Mpip (D^{\pi'})^{-1})^T (P_{\pi'}^T-P_{\pi}^T)\dpi \\
\text{i.e.} \qquad \norm{d_{\pi'}-d_{\pi}}{1}{} &\leq \norm{(I - \Mpip (D^{\pi'})^{-1})^T (P_{\pi'}^T-P_{\pi}^T)\dpi}{1}{} \nonumber \tag{submultiplicative property} \\
\norm{d_{\pi'}-d_{\pi}}{1}{} &\leq \underbrace{\norm{(I - \Mpip (D^{\pi'})^{-1})}{\infty}{}}_{T_{1}} \underbrace{\norm{(P_{\pi'}^T-P_{\pi}^T)\dpi}{1}{}}_{T_{2}}    \tag{Holder's inequality}  \nonumber
\end{align}

We know that $\kappa^{\pi} = \text{Tr}(\Zpi)$ and from \cite{hunter2014mathematical}, we can write $T_{1}$ using the eigenvalues $\{\lambda_{\pi, i}\}_{i=1}^{|S|}$ of the underlying $P_{\pi}$ as
$$
T_{1} \leq \frac{1}{|S|} \sum_{i=2}^{|S|} \frac{1}{(1 - \lambda_{\pi, i})^{1/2}} \leq \max_i (1 - \lambda_{\pi, i})^{-1/2} = \sigma^{\pi} \leq \max_{\pi} \sigma^{\pi} = \sigma^{\star}.
$$

For $T_{2}$, we refer to the result by \cite{zhang2020average}, and provide the proof for completeness below.

\begin{align*}
    T_{2} &= \sum_{s'}\left|\sum_s\left(\sum_a P(s'|s,a)\pi'(a|s)-P(s'|s,a)\pi(a|s)\right)d_{\pi}(s)\right| \\
    &\leq \sum_{s',s}\left|\sum_a P(s'|s,a)(\pi'(a|s)-\pi(a|s))\right|d_{\pi}(s) \\
    &\leq \sum_{s,s',a}P(s'|s,a)\left|\pi'(a|s)-\pi(a|s)\right|d_{\pi}(s) \\
    &\leq \sum_{s,a}\left|\pi'(a|s)-\pi(a|s)\right|d_{\pi}(s) = 2 \E{s\sim d_{\pi}}[\TV{\pi'}{\pi}[s]]
\end{align*}
Combining these inequalities of $T_{1}$ and $T_{2}$, we get the desired result.
\end{proof}

\subsection{Performance and Constraint Bounds of Trust Region Approach}

Consider the trust region formulation in Equation \eqref{eq:acpo_trust}. To prove the policy performance bound when the current policy is infeasible (i.e., constraint-violating), we prove the KL divergence between $\pik$ and $\pikup$ for the KL divergence projection, along with other lemmas. We then prove our main theorem for the worst-case performance degradation.

\begin{lemma}
\label{lemma:feasible_kl_bound}
For a closed convex constraint set, if we have a constraint satisfying policy $\pik$ and the KL divergence $\E{s\sim d_{\pik}}\big[\KL{\pi_{k+1/2}}{\pik}[s]\big]$ of the `Improve' step is upper bounded by step size $\delta$, then after KL divergence projection of the `Project' step we have 
\[
\E{s\sim d_{\pik}}\big[\KL{\pikup}{\pik}[s]\big]\leq \delta.
\]
\end{lemma}
\begin{proof}

We make use of the fact that Bregman divergence (hence, KL divergence) projection onto the constraint set ($\in \Rbb^{d} \,, d \in \Nbb$) exists and is unique. Since $\pik$ is safe, we have $\pik$ in the constraint set, and $\pikup$ is the projection of $\pi_{k+\frac{1}{2}}$ onto the constraint set. Using the projection inequality, we have 
\begin{align*}
\E{s\sim d_{\pik}}\big[\KL{\pik}{\pikup}[s]\big] + \E{s\sim d_{\pik}}\big[\KL{\pikup}{\pi_{k+\frac{1}{2}}}[s]\big] \leq \E{s\sim d_{\pik}}\big[\KL{\pik}{\pi_{k+\frac{1}{2}}}[s]\big] & \\
\implies \E{s\sim d_{\pik}}\big[\KL{\pik}{\pikup}[s]\big] \leq \E{s\sim d_{\pik}}\big[\KL{\pik}{\pi_{k+\frac{1}{2}}}[s]\big] \leq \delta \tag{$\KL{\cdot}{\cdot} \geq 0$}.
\end{align*}
Since KL divergence is asymptotically symmetric when updating the policy within a local neighbourhood ($\delta << 1$), we have
$$
\E{s\sim d_{\pik}}\big[\KL{\pikup}{\pik}[s]\big] \leq 
\E{s\sim d_{\pik}}\big[\KL{\pi_{k+\frac{1}{2}}}{\pik}[s]\big] \leq \delta.
$$
\end{proof}

\begin{lemma}
\label{lemma:infeasible_kl_bound}

For a closed convex constraint set, if we have a constraint violating policy $\pik$ and the KL divergence $\E{s\sim d_{\pik}}\big[\KL{\pi_{k+1/2}}{\pik}[s]\big]$ of the first step is upper bounded by step size $\delta$, then after KL divergence projection of the second step we have 
\[
\E{s\sim d_{\pik}}\big[\KL{\pikup}{\pik}[s]\big]\leq \delta+V_{max},
\]
where $V_{max} = \max_{i} \alpha_{i}\beta_{i}^{2}, \,  \beta_{i} = [J_{C_{i}}(\pik)-l_{i}]_{+}$, $\alpha_{i} = \frac{1}{2a_{i}^T H^{-1}a_{i}},$ with
$a_{i}$ as the gradient of the cost advantage function corresponding to constraint $C_{i}$, and $H$ as the Hessian of the KL divergence constraint. \footnote{For any $x \in \Rbb, \, [x]_{+} := \max(0, x)$}.
\end{lemma}
\begin{proof}

Let the sublevel set of cost constraint function for the current infeasible policy $\pik$ be given as:
\[
L_{\pik}=\{\pi~|~J_{C_{i}}(\pi)+ \E{\substack{s\sim d_{\pik} \\ a\sim \pi}}[\wb{A}^{\pik}_{C_{i}}(s,a)] \leq J_{C_{i}}(\pik) \forAll i\}.
\]
This implies that the current policy $\pik$ lies in $L_{\pik}$. The constraint set onto which $\pi_{k+\frac{1}{2}}$ is projected onto is given by: $\{\pi~|~J_{C_{i}}(\pik)+ \E{\substack{s\sim d_{\pik} \\ a\sim \pi}}[\wb{A}^{\pik}_{C_{i}}(s,a)]\leq l_{i} \forAll i \}.$ Let $\pikup^L$ be the projection of $\pi_{k+\frac{1}{2}}$ onto $L_{\pik}.$

Note that the Bregman inequality of Lemma \ref{lemma:feasible_kl_bound} holds for any convex set in $\Rbb^{d} \,, d \in \Nbb$. This implies $\E{s\sim d_{\pik}}\big[\KL{\pikup^L}{\pik}[s]\big] \leq \delta$ since $\pik$ and $\pikup^L$ are both in $L_{\pik}$, which is also convex since the constraint functions are convex. Using the Three-point Lemma \footnote{For any $\phi$, the Bregman divergence identity: $D_\phi(x,y)+D_\phi(y,z)=D_\phi(x,z)+<\nabla \phi(z)-\nabla \phi(y),x-y>$}, for polices $\pik, \pikup,$ and $\pikup^L$, with $\varphi(\textbf{x}):=\sum_i x_i\log x_i$, we have

\begin{align}
\delta \geq  \E{s\sim d_{\pik}}\big[\KL{\pikup^L}{\pik})[s]\big]&=\E{s\sim d_{\pik}}\big[\KL{\pikup}{\pik}[s]\big] \nonumber\\ 
&-\E{s\sim d_{\pik}}\big[\KL{\pikup}{\pikup^L}[s]\big] \nonumber \\
&+\E{s\sim d_{\pik}}\big[(\nabla\varphi(\pik)-\nabla\varphi(\pikup^{L}))^T(\pikup-\pikup^L)[s]\big] \nonumber \\
\Rightarrow \E{s\sim d_{\pik}}\big[\KL{\pikup}{\pik}[s]\big]&\leq \delta + \underbrace{\E{s\sim d_{\pik}}\big[\KL{\pikup}{\pikup^L}[s]\big]}_{T_{1}} \nonumber \\
&- \underbrace{\E{s\sim d_{\pik}}\big[(\nabla\varphi(\pik)-\nabla\varphi(\pikup^L))^T(\pikup-\pikup^L)[s]\big]}_{T_{2}}. 
\end{align}

If the constraint violations of the current policy $\pik$ are small, i.e., $J_{C_{i}}(\pik)-l_{i} = b_{i} $ is small for all $i$, then $T_{1}$ can be approximated by a second order expansion. We analyze $T_{1}$ for any constraint $C_{i}$ and then bound it over all the constraints. As before we overload the notation with $\pik = \pi_{\theta_{k}} = \theta_{k}$ to write. For any constraint $C_{i}$, we can write $T^{i}_{1}$ as the expected KL divergence if projection was onto the constraint set of $C_{i}$ i.e.

\begin{align*}
    T^{i}_{1} \approx \frac{1}{2}(\pikup-\pikup^L)^{T} H(\pikup-\pikup^L) & =\frac{1}{2} \Big(\frac{\beta_{i}}{a_{i}^T H^{-1}a_{i}} H^{-1}a_{i}\Big)^T H\Big(\frac{\beta_{i}}{a_{i}^T H^{-1}a_{i}}H^{-1}a_{i}\Big) \\ &= \frac{\beta_{i}^2}{2a_{i}^T H^{-1}a_{i}} = \alpha_{i} \beta_{i}^2,
\end{align*}

where the second equality is a result of the trust region guarantee (see \cite{schulman2015trust} for more details). Finally we invoke the projection result from \cite{achiam2017advanced} which uses Dykstra's Alternating Projection algorithm from \cite{tibshirani2017dykstra} to bound this projection, i.e., $T_{1} \leq \max_{i} T^{i}_{1} \approx \max_{i} \alpha_{i} \beta_{i}^2$.

And since $\delta$ is small, we have $\nabla\varphi(\pik)-\nabla\varphi(\pikup^{L})\approx 0$ given $s$. Thus, $T_{2} \approx 0$. Combining all of the above, we have $\E{s\sim d_{\pik}}\big[\KL{\pikup}{\pik}[s]\big]\leq \delta+V_{max}.$
\end{proof}

\trustdegradationviolation*
\begin{proof}
\label{proof:trust_proof}

Since $\avKL{\polk}{\polk}=0$, $\polk$ is feasible. The objective value is 0 for $\pol=\polk$. The bound follows from Equation \eqref{eq:avg_policy_imp} and Equation \eqref{eq:tv-kl} where the average KL i.e. $\E{s\sim d_{\pik}}\big[\KL{\pikup}{\pik}[s]\big]$ is bounded by $\delta+V_{max}$ from Lemma \ref{lemma:infeasible_kl_bound}.

Similar to Corollary \ref{cor:thm:avg_constraint_imp}, expressions for the auxiliary cost constraints also follow immediately as the second result.

\remark{Remark} If we look at proof as given by \cite{zhang2020average} in Section 5 of their paper, with the distinction now that $\delta$ is replaced by $\delta + V_{max}$, we have the same result. Our worse bound is due to the constrained nature of our setting, which is intuitive in the sense that for the sake of satisfying constraints, we undergo a worse worst-case performance degradation.

\end{proof}

\subsection{Approximate ACPO}
\label{appendix:approx_acpo}

\subsubsection{Policy Recovery Routine}
As described in Section \ref{sec:recovery}, we need a recovery routine in case the updated policy $\pi_{k+1/2}$ is not approximate constraint satisfying. In this case, the optimization problem is inspired from a simple trust region approach by \cite{schulman2015trust}. Since we only deal with one constraint in the practical implementation of ACPO, the recovery rule is obtained by solving the following problem:

$$
\begin{array}{ll}
\min _{\theta} & c+a^{T}\left(\theta-\theta_{k}\right) \\
\text { s.t. } & \frac{1}{2}\left(\theta-\theta_{k}\right)^{T} H\left(\theta-\theta_{k}\right) \leq \delta .
\end{array}
$$

Let $x=\theta-\theta_{k}$, then the dual function $L(x, \lambda)$ is given by: $L(x, \lambda)=c+a^{T} x+\lambda\left(\frac{1}{2} x^{T} H x-\delta\right)$. Now,

$$
\frac{L}{\partial x}=a+\lambda(H x)=0 \Longrightarrow x=-\frac{1}{\lambda} H^{-1} a .
$$

$x$ obtained above should satisfy the trust-region constraint:

$$
\begin{aligned}
\frac{1}{2}\left(-\frac{1}{\lambda} H^{-1} a\right)^{T} H\left(-\frac{1}{\lambda} H^{-1} a\right) & \leq \delta \\
\Longrightarrow \quad \frac{1}{2} \cdot \frac{1}{\lambda^{2}} \cdot a^{T} H^{-1} a & \leq \delta \\
\Longrightarrow \sqrt{\frac{a^{T} H^{-1} a}{2 \delta}} & \leq \lambda .
\end{aligned}
$$

Therefore, the update rule in case of infeasibility takes the form $ \theta=\theta_{k}-\sqrt{\frac{2 \delta}{a^{T} H^{-1} a}} H^{-1} a$. We augment this rule with the gradient of the reward advantage function as well, so the final recovery is 
$$
\theta_{k+1/2} = \theta_{k} - \sqrt{2\delta} \bigg[t \cdot \frac{H^{-1} a}{\sqrt{a^T H^{-1} a}} + (1-t) \cdot \frac{H^{-1} g}{\sqrt{g^T H^{-1} g}} \bigg] \quad ; \quad t \in [0,1]
$$

\subsubsection{Line Search}
Because of approximation error, the proposed update may not satisfy the constraints in Eq. \eqref{eq:acpo_trust}. Constraint satisfaction is enforced via line search, so the final update is $$ \theta_{k+1} = \theta_k + s^j \left(\theta_{k+1/2} - \theta_k\right),$$ where $s \in (0,1)$ is the backtracking coefficient and $j \in \{0, ..., L\}$ is the smallest integer for which $\pi_{k+1}$ satisfies the constraints in Equation \ref{eq:acpo_trust}. Here, $L$ is a finite backtracking budget; if no proposed policy satisfies the constraints after $L$ backtracking steps, no update occurs.

\subsection{Practical ACPO}
\label{appendix:practical_acpo}

As explained in Section \ref{sec:acpo_implementation}, we use the below problem formulation, which uses first-order Taylor approximation on the objective and second-order approximation on the KL constraint \footnote{The gradient and first-order Taylor approximation of $\avKL{\pol}{\polk}$ at $\theta=\theta_k$ is zero.} around $\theta_k$, given small $\delta$: 
\begin{equation}
\begin{aligned}
\max_{\theta} \;\;\;&  g^T (\theta - \theta_k)  \\
\text{s.t. } \;\;\;& c_i + a_i^T (\theta - \theta_k) \leq 0, \forAll i \qquad ; \qquad  \tfrac{1}{2} (\theta - \theta_k)^T H (\theta - \theta_k) \leq \delta.
\end{aligned}
\end{equation}
where
\begin{align*}
    g &:= \E{\substack{s\sim\dpolk \\ a\sim\polk}}\left[\grad\log\pol(a|s)|_{\theta=\theta_k}\wb{A}^{\polk}(s,a)\right] \qquad &; \qquad  c_i &:= J_{C_i}(\theta_k) - l_i  \forAll i \\
    a_i &:= \E{\substack{s\sim\dpolk \\ a\sim\polk}}\left[\grad\log\pol(a|s)|_{\theta=\theta_k}\wb{A}^{\polk}_{C_{i}}(s,a)\right] \qquad &; \qquad  H &:= \E{\substack{s\sim\dpolk \\ a\sim\polk}}\left[\grad\log\pol(a|s)|_{\theta=\theta_k}\grad\log\pol(a|s)|_{\theta=\theta_k}^T\right] 
\end{align*}

Similar to the work of \cite{achiam2017constrained}, $g$, $a_{i}$, and $H$ can be approximated using samples drawn from the policy $\polk$. The Hessian $H$ is identical to the Hessian $H$ used by \cite{achiam2017constrained} and \cite{zhang2020average}. However, the definitons $g$ and $a_{i}$'s are different since they include the average reward advantage functions, $\wb{A}^{\polk}(s,a) =   \wb{Q}^{\polk}(s,a) -  \wb{V}^{\polk}(s)$.

Since rewards and cost advantage functions can be approximated independently, we use the framework of \cite{zhang2020average} to do so. We describe the process of estimation of rewards advantage function, and the same procedure can be used for the cost advantage functions as well. Specifically, first approximate the average-reward bias $\wb{V}^{\polk}(s)$ and then use a one-step TD backup to estimate the action-bias function. Concretely, using the average reward Bellman equation gives
\begin{equation}\label{eq:bellman_objective}
    \wb{A}^{\polk}(s,a) =  r(s,a) - J(\polk) + \E{s'\sim P(\cdot|s,a)}\left[\wb{V}^{\polk}(s')\right] - \wb{V}^{\polk}(s)
\end{equation}

This expression involves the average-reward bias $\wb{V}^{\polk}(s)$, which we can approximated using the standard critic network $\wb{V}_{\phi_k}(s)$. However, in practice we use the average-reward version of the Generalized Advantage Estimator (GAE) from \cite{schulman2016high}, similar to \cite{zhang2020average}. Hence, we refer the reader to that paper for detailed explanation, but provide an overview below for completeness.

Let $ \hat{J}_{\pi} = \frac{1}{N}\sum_{t=1}^N r_{t} $ denote the estimated average reward. The Monte Carlo target for the average reward value function is $\vtarg_t = \sum_{t'=t}^N (r_{t} - \hat{J}_{\pi})$ and the bootstrapped target is $\vtarg_t = r_{t} - \hat{J}_{\pi} +\vphi(s_{t+1})$.

The Monte Carlo and Bootstrap estimators for the average reward advantage function are:
\begin{align*}
    \hat{A}_{\text{MC}}^{\pi}(s_{t}, a_{t}) = \sum_{t'=t}^N (r_{t} - \hat{J}_{\pi}) -  \vphi(s_{t}) \qquad ; \qquad 
    \hat{A}_{\text{BS}}^{\pi}(s_{t}, a_{t}) = r_{i,t}-\hat{J}_{\pi} + \vphi(s_{t+1}) - \vphi(s_{t})
\end{align*}
We can similarly extend the GAE to the average reward setting:
\begin{equation}\label{eq:avg_GAE}
    \hat{A}_{\text{GAE}}(s_{t},a_{t}) = \sum_{t'=t}^N \lambda^{t'-t}\delta_{t'} \qquad , \qquad \delta_{t'} = r_{t'} - \hat{J}_{\pi}+ \vphi(s_{t'+1}) - \vphi(s_{t'}).
\end{equation}

and set the target for the value function to $ \vtarg_t = r_t - \hat{J}_{\pi}+ \vphi(s_{t+1}) + \sum_{t'=t+1}^N \lambda^{t'-t}\delta_{t'}$.

\subsection{Experimental Details}
\label{appendix:experimental_details}

For detailed explanation of Point-Circle, Point-Gather, Ant-Circle, and Ant-Gather tasks, please refer to \cite{achiam2017constrained}. For detailed explanation of Bottleneck and Grid tasks, please refer to \cite{vinitsky2018benchmarks}. For the simulations in the Gather and Circle tasks, we use neural network baselines with the same architecture and activation functions as the policy networks. For the simulations in the Grid and Bottleneck tasks, we use linear baselines. For all experiments we use Gaussian neural policies whose outputs are the mean vectors and variances are separate parameters to be learned. Seeds used for generating evaluation trajectories are different from those used for training.

For comparison of different algorithms, we make use of CPO, PCPO, ATRPO, and PPO implementations taken from \href{https://github.com/rll/rllab}{\code{rllab}} and \href{https://github.com/openai/safety-starter-agents}{\code{safety-starter-agents}}. Even the hyperparameters are selected so as to showcase the best performance of other algorithms for fair comparison. The choice of the hyperparameters given below is inspired by the original papers since we wanted to understand the performance of the average reward case. 

We use settings which are common in all open-source implementations of the algorithms, such as normalizing the states by the running mean and standard deviation before being fed into the neural network and similarly normalizing the advantage values (for both rewards and constraints) by their batch means and standard deviations before being used for policy updates. Table \ref{tab:hyperparameters} summarizes the hyperparameters below.

\begin{table}[ht]
    \centering
    \caption{Hyperparameter Setup}
    \vskip 0.15in
    \small
    \begin{tabular}{l l l}
    \toprule
       Hyperparameter & PPO/ATRPO & CPO/PCPO/ACPO  \\
        \midrule
    No. of hidden layers & 2 & 2  \\
      Activation & $\tanh$ & $\tanh$ \\
      Initial log std & -0.5 & -1 \\
      Batch size & 2500 & 2500 \\
      GAE parameter (reward) & 0.95 & 0.95  \\ 
      GAE parameter (cost) & N/A & 0.95  \\
      Trust region step size $\delta$ & $10^{-4}$ & $10^{-4}$ \\
      Learning rate for policy & $2\times 10^{-4}$ & $2\times 10^{-4}$ \\
      Learning rate for reward critic net & $2\times 10^{-4}$ & $2\times 10^{-4}$  \\
      Learning rate for cost critic net & N/A & $2\times 10^{-4}$  \\
       Backtracking coeff.  & 0.75 & 0.75  \\
       Max backtracking iterations & 10 & 10 \\
       Max conjugate gradient iterations & 10 & 10 \\
       Recovery regime parameter $t$ & N/A & 0.75 \\
       \bottomrule
    \end{tabular}
    \label{tab:hyperparameters}
\end{table}

For the Lagrangian formulation of ATRPO and PPO, note that the original papers do not provide any blueprint for formulating the Lagrangian, and even CPO and PCPO use \emph{unconstrained} TRPO for benchmarking. However, we feel that this is unfair to these algorithms as they can possibly perform better with a Lagrangian formulation in an average-reward CMDP setting. To this extent, we introduced a Lagrangian parameter $\ell \in [0,1]$ that balances the rewards and constraints in the final objective function. More specifically, Equation \eqref{eq:acpo_approx} for a single constraint now becomes

\begin{equation}
\label{eq:lagrangian_balanced}
\begin{aligned}
\max_{\theta} \;\;\;&  (1-\ell) g^T (\theta - \theta_k) - \ell \big[ \big(c_1 + a_1^T (\theta - \theta_k)\big) + \big( \tfrac{1}{2} (\theta - \theta_k)^T H (\theta - \theta_k) - \delta \big)   \big].
\end{aligned}
\end{equation}

\paragraph{Note}. The authors of the ATRPO and PPO do not suggest any principled approach for finding an optimal $\ell$. Hence, the choice of the Lagrangian parameter $\ell$ is completely empirical and is selected such that these algorithms achieve maximum rewards while satisfying the constraints. Also see in Figure \ref{fig:rewards_costs_comparison}, for Ant-Gather, Bottleneck, and Grid environments, where the constraints cannot be satisfied for \emph{any} value of $\ell$, we include the results for a specific value of $\ell$ for illustrative purposes, as detailed in Table \ref{tab:lagrangian_parameters}.

\begin{table}[ht]
    \centering
    \caption{Lagrangian parameter $\ell$ for ATRPO and PPO}
    \vskip 0.15in
    \small
    \begin{tabular}{l c c c c c}
    \toprule
       Algorithm & Point-Gather & Ant-Circle & Ant-Gather & Bottleneck & Grid  \\
        \midrule
    ATRPO & 0.50 & 0.60 & 0.45 & 0.50 & 0.45 \\
      PPO & 0.55 & 0.50 & 0.50 & 0.50 & 0.60 \\
       \bottomrule
    \end{tabular}
    \label{tab:lagrangian_parameters}
\end{table}

\subsection{Experimental Addendum}
\label{appendix:additional_results}

\subsubsection{Environments}
\label{appendix:environments}

\begin{figure}[ht]
 \captionsetup[subfigure]{justification=centering}
    \centering
    \subfloat[Circle]{
        \includegraphics[width=0.2\textwidth]{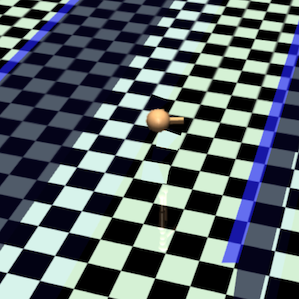}
        \label{fig:circle}
    }
    \subfloat[Gather]{
        \includegraphics[width=0.2\textwidth]{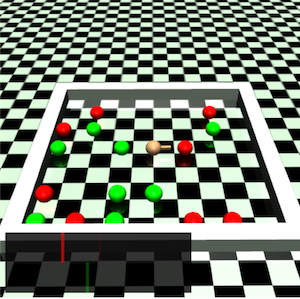}
        \label{fig:gather}
    }
    \subfloat[Grid]{
        \includegraphics[width=0.2\textwidth]{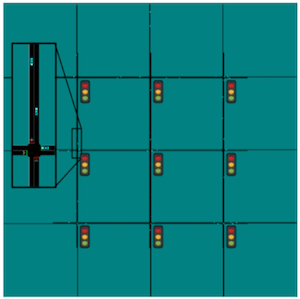}
        \label{fig:grid}
    }
    \subfloat[Bottleneck]{
        \includegraphics[width=0.2\textwidth, angle =90]{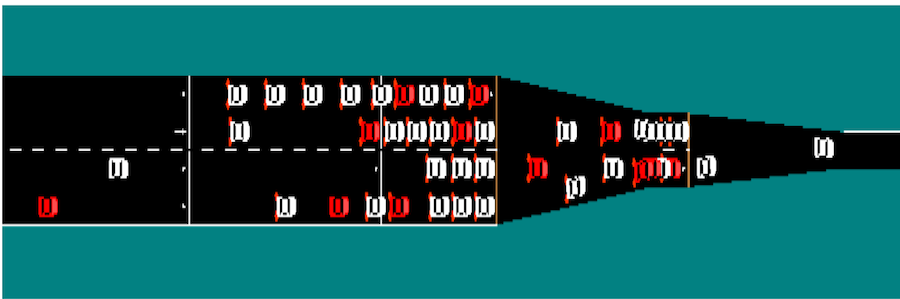}
        \label{fig:bottleneck}
    }
    \caption[]{The Circle, Gather, Grid, and Bottleneck tasks. (a) Circle: The agent is rewarded for moving in a specified circle but is penalized if the diameter of the circle is larger than some value as in \cite{achiam2017constrained}.  (b) Gather: The agent is rewarded for collecting the green balls while penalized to gather red balls as in \cite{achiam2017constrained}. (c) Grid: The agent controls traffic lights in a 3x3 road network and is rewarded for high traffic throughput but is constrained to let lights be red for at most 5 consecutive seconds as in \cite{vinitsky2018benchmarks}. (d) Botteneck: The agent controls vehicles (red) in a merging traffic situation and is rewarded for maximizing the number of vehicles that pass through but is constrained to ensure that white vehicles (not controlled by agent) have ``low'' speed for no more than 10 seconds as in \cite{vinitsky2018benchmarks}.}
    \label{fig:env_overview}
\end{figure}

All environments tested on are illustrated in Figure \ref{fig:env_overview}, along with a detailed description of each.

\subsubsection{Learning Curves}

\begin{figure*}[ht!]
 \captionsetup[subfigure]{justification=centering}
\centering
    \hspace{-6cm} Average Rewards: \newline

    {
        \includegraphics[width=0.3\textwidth]{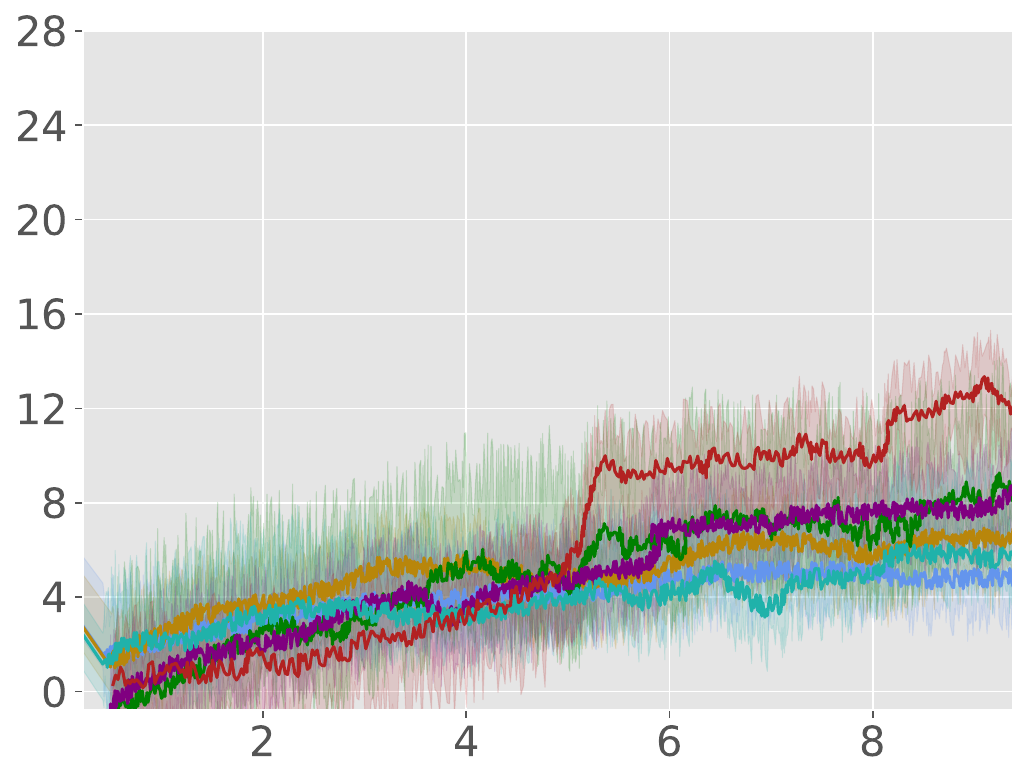}
        \label{fig:point_gather_rewards}
    }
    {
        \includegraphics[width=0.3\textwidth]{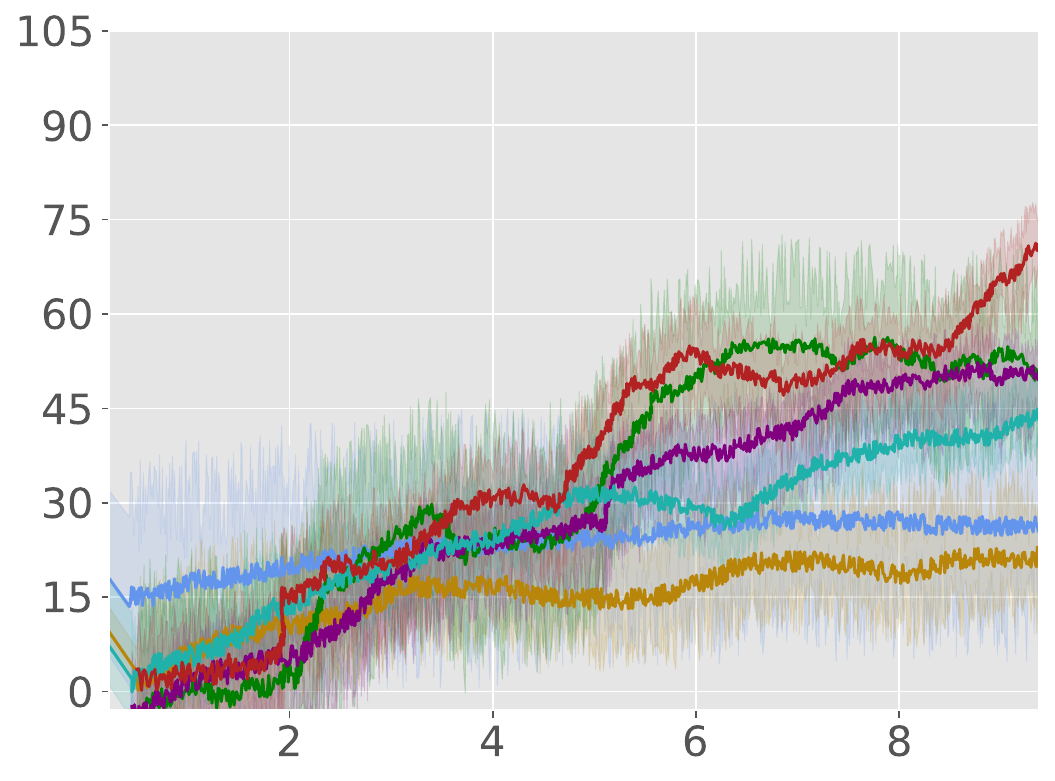}
        \label{fig:ant_circle_rewards}
    }

    \rule{\linewidth}{0.5pt} \newline
    \centering
    \hspace{-2cm} Average Constraint values: \hspace{2cm} \includegraphics[width=0.55\textwidth]{chapter1/images/rewards_costs_legend.JPG} \newline
    \subfloat[Point Gather]{
        \includegraphics[width=0.3\textwidth]{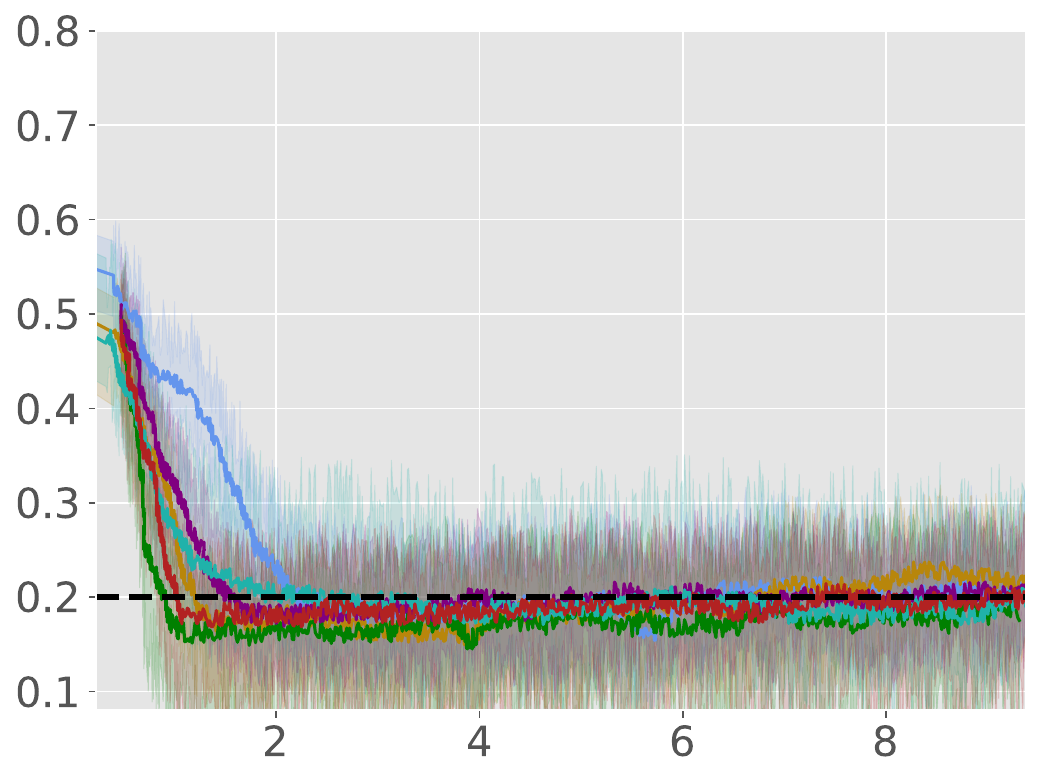}
        \label{fig:point_gather_costs}
    }
    \subfloat[Ant Circle]{
        \includegraphics[width=0.3\textwidth]{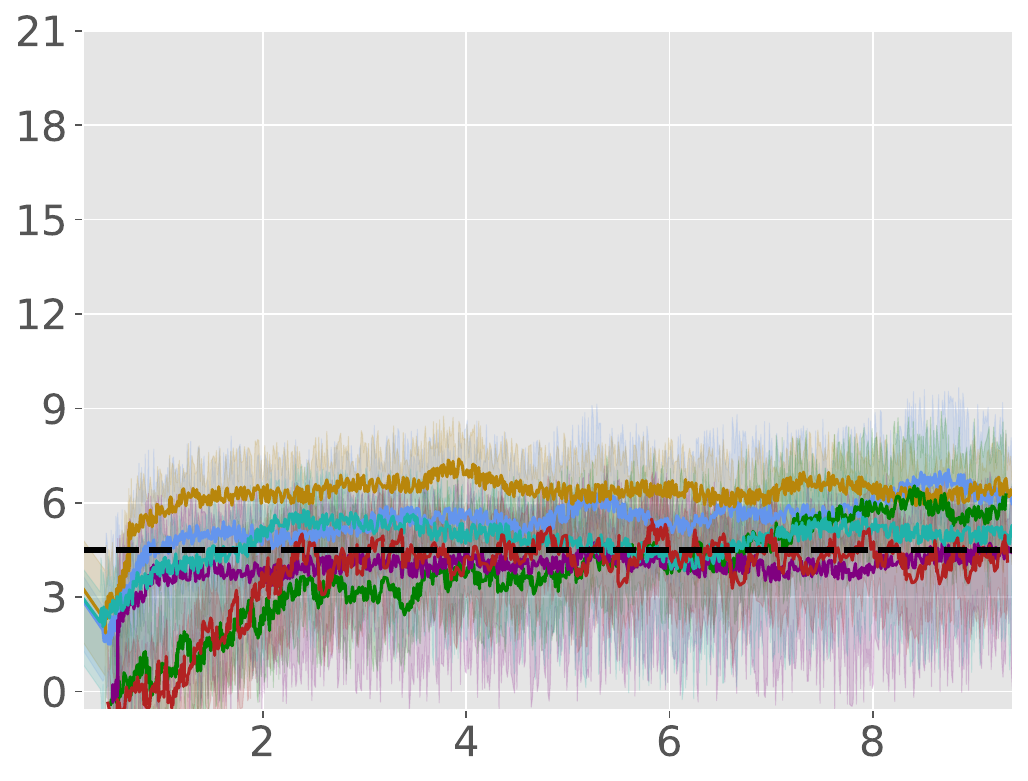}
        \label{fig:ant_circle_costs}
    }

    \caption{The average reward and constraint cost function values vs iterations (in $10^{4}$) learning curves for some algorithm-task pairs. Solid lines in each figure are the empirical means, while the shaded area represents 1 standard deviation, all over 5 runs. The dashed line in constraint plots is the constraint threshold $l$. ATRPO and PPO are tested with constraints, which are included in their Lagrangian formulation.}
    \label{fig:appendix_rewards_costs_comparison}
\end{figure*}

Due to space restrictions, we present the learning curves for the remaining environments in Figure \ref{fig:appendix_rewards_costs_comparison}.

\subsubsection{Recovery Regime Revisited}

\begin{figure*}[ht]
 \captionsetup[subfigure]{justification=centering}
    \hspace{0.25cm} Rewards: \newline
    {
        \includegraphics[width=0.18\textwidth]{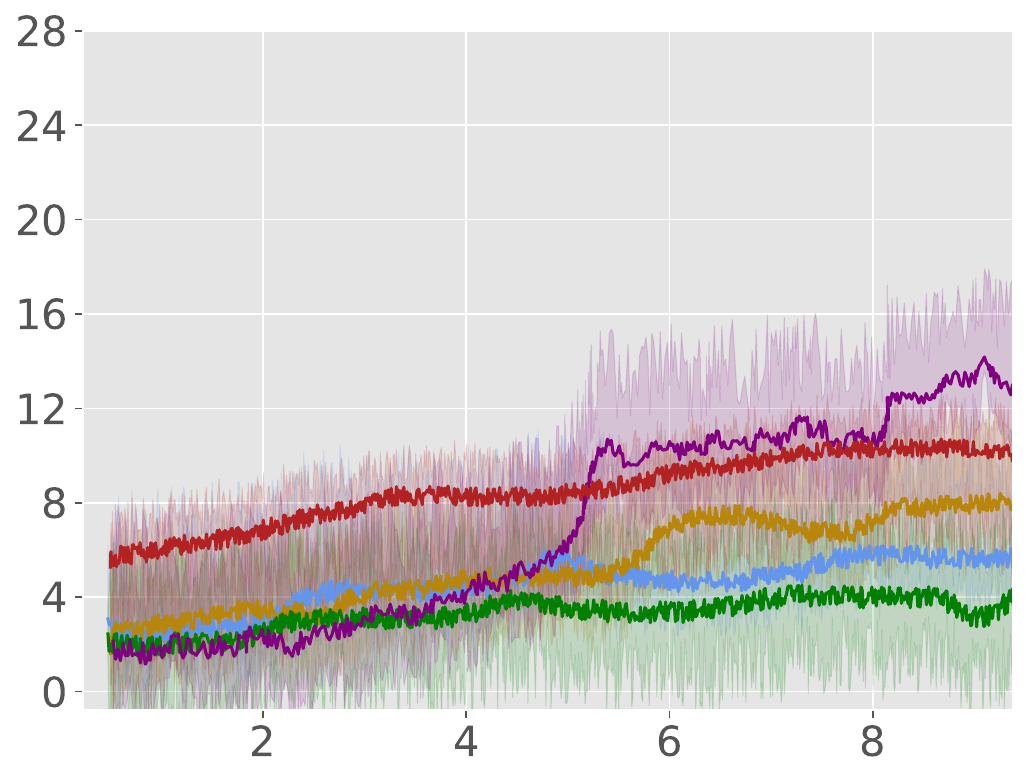}
        \label{fig:point_gather_hyper_rewards}
    }
    {
        \includegraphics[width=0.18\textwidth]{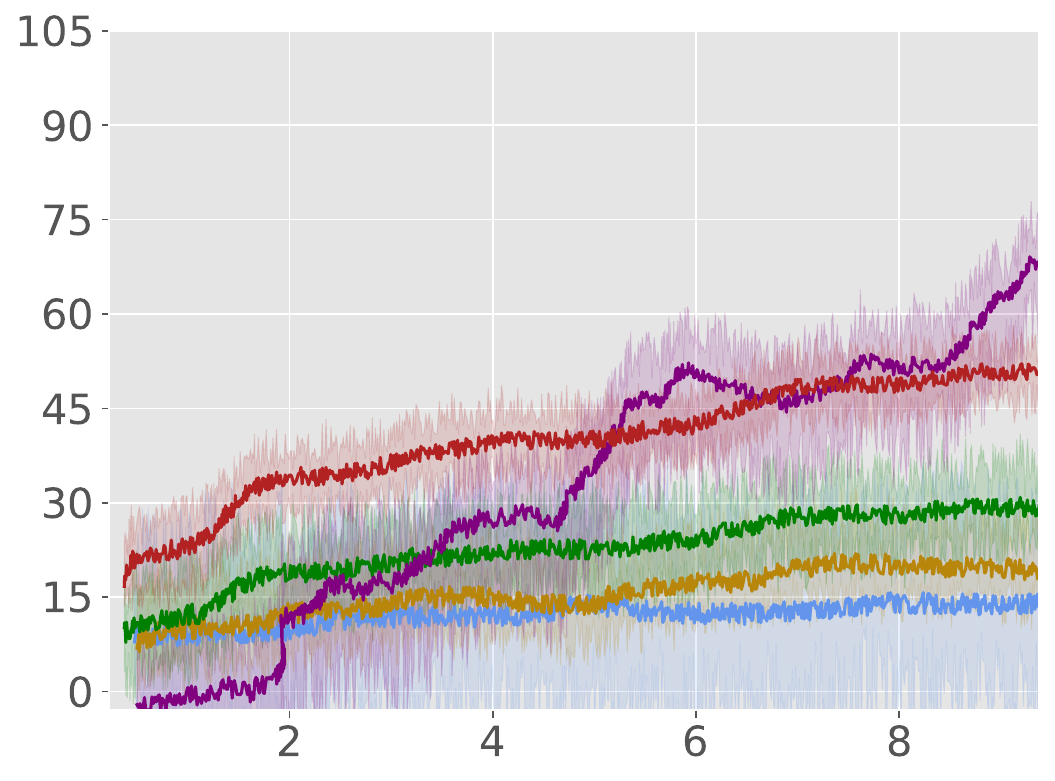}
        \label{fig:ant_circle_hyper_rewards}
    }
    {
        \includegraphics[width=0.185\textwidth]{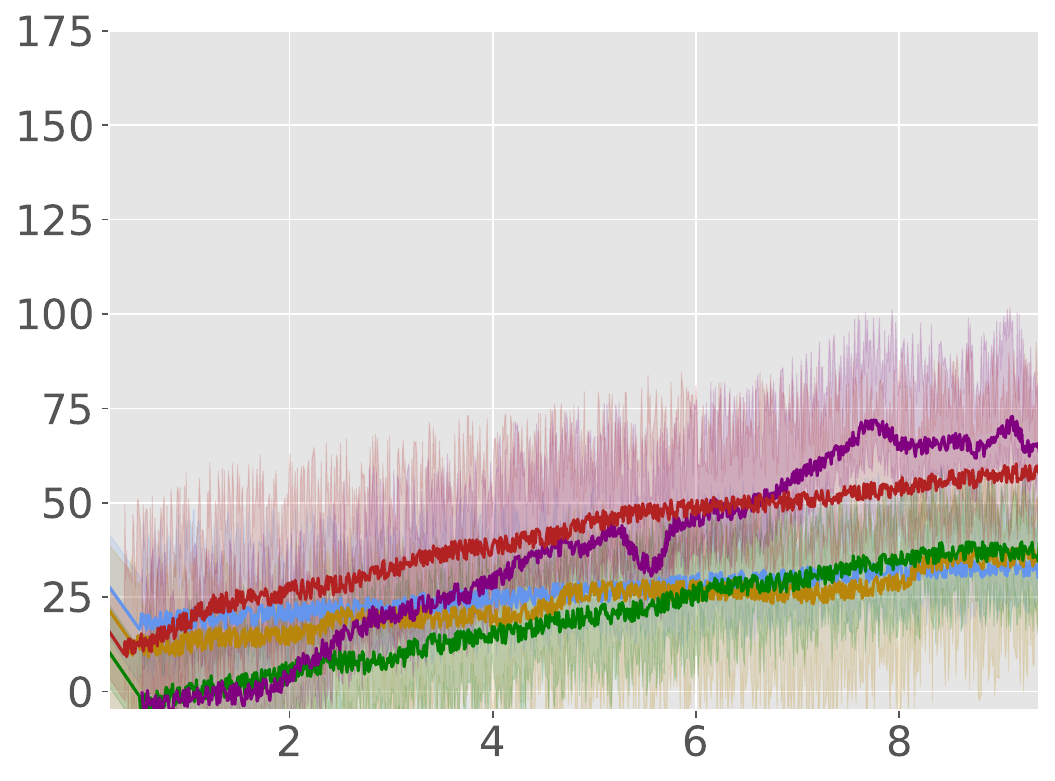}
        \label{fig:ant_gather_hyper_rewards}
    }
    {
        \includegraphics[width=0.185\textwidth]{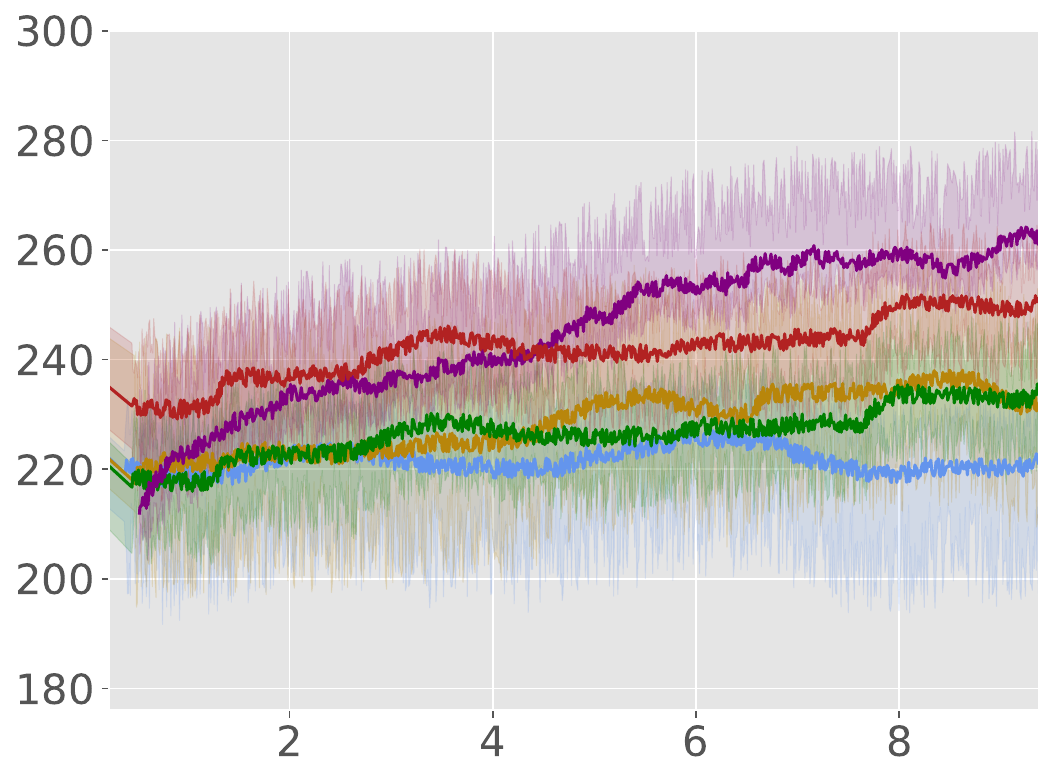}
        \label{fig:bottleneck_hyper_rewards}
    }
    {
        \includegraphics[width=0.185\textwidth]{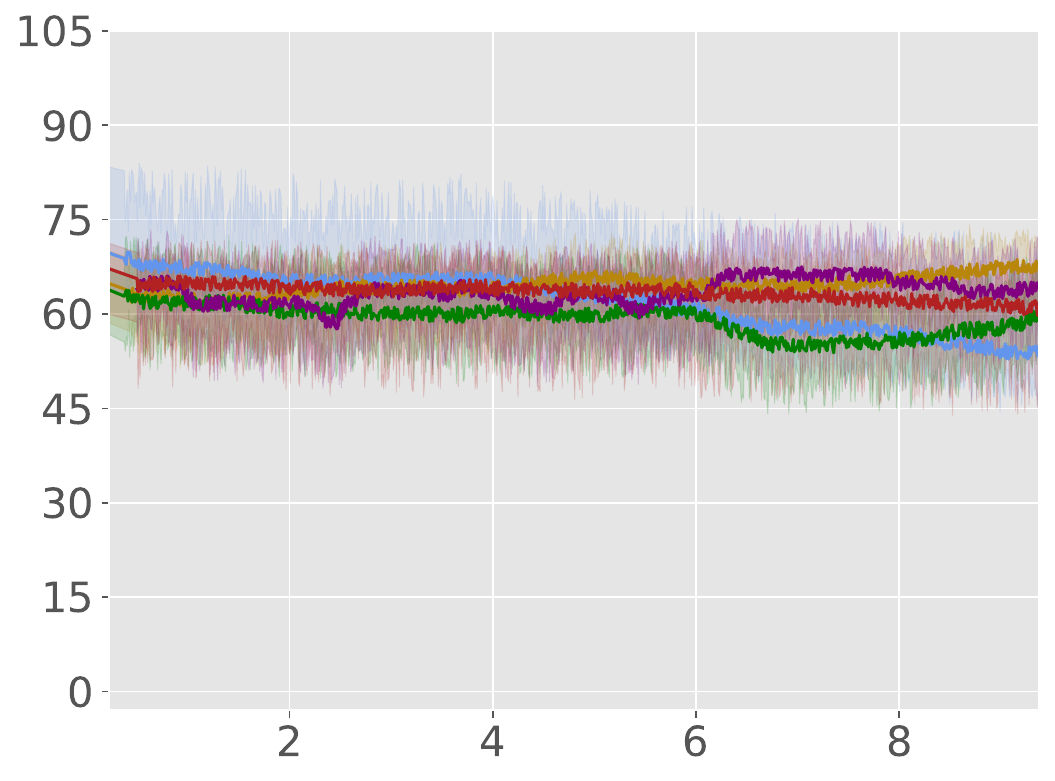}
        \label{fig:grid_hyper_rewards}
    }

    \rule{\linewidth}{0.5pt}
    \hspace{2cm} Constraint values: \hspace{4cm} \includegraphics[width=0.5\textwidth]{chapter1/images/legend_t_hyperparam.JPG} \newline

    \subfloat[Point Gather]{
        \includegraphics[width=0.185\textwidth]{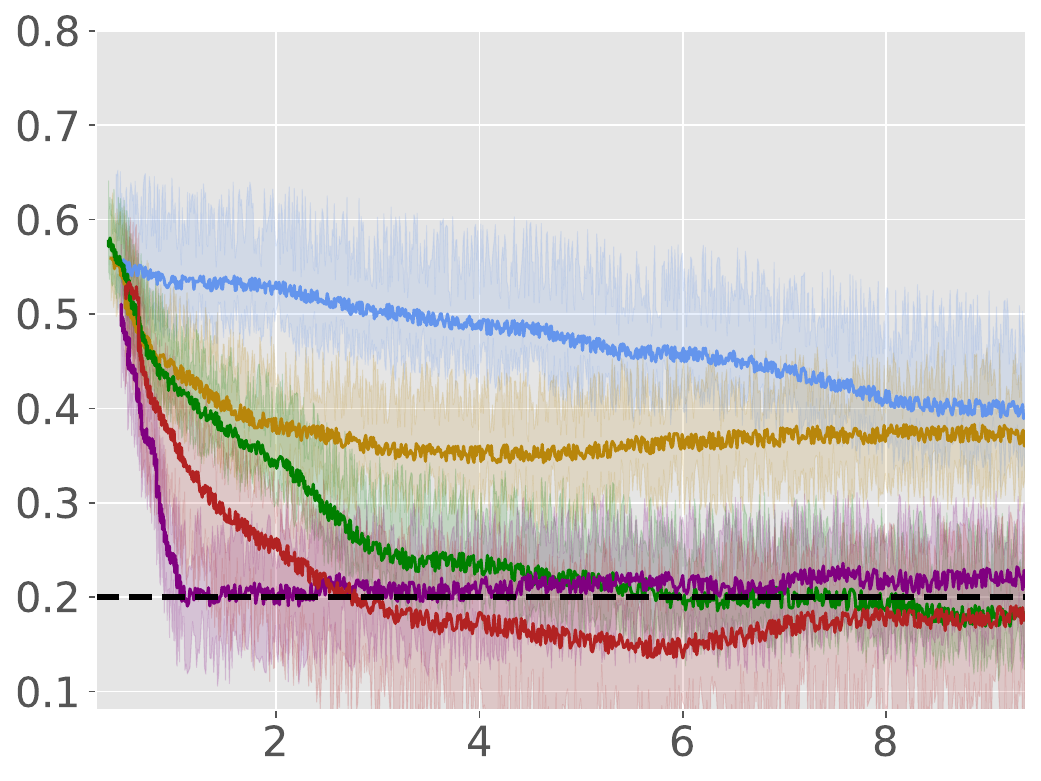}
        \label{fig:point_gather_hyper_costs}
    }
    \subfloat[Ant Circle]{
        \includegraphics[width=0.185\textwidth]{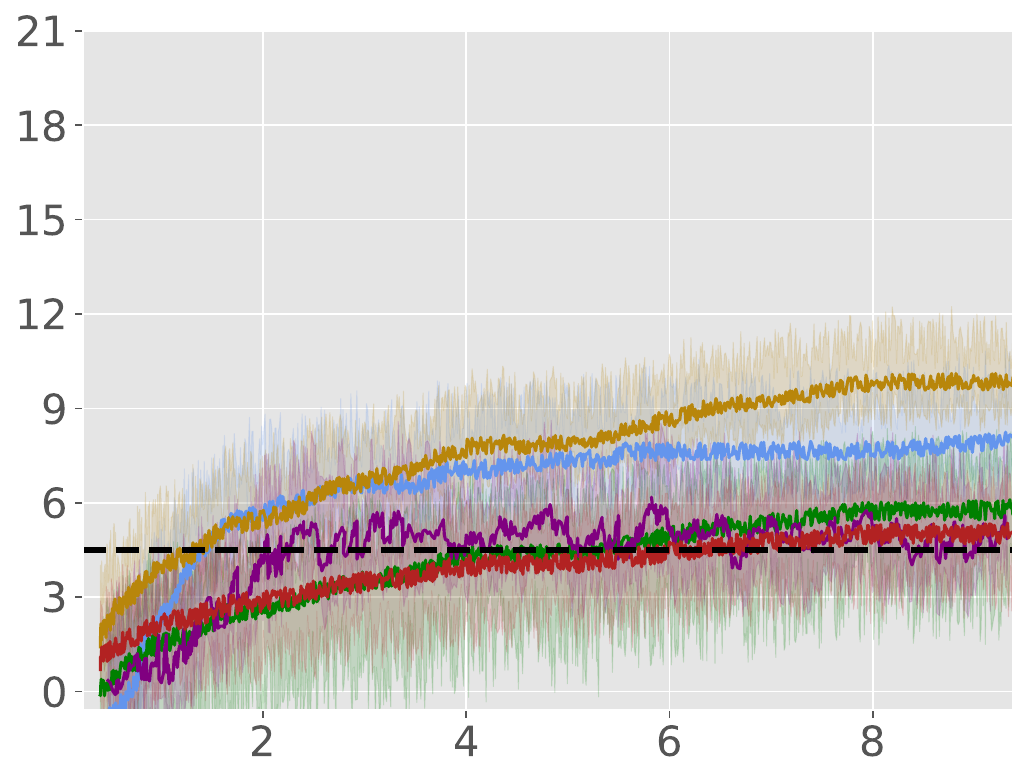}
        \label{fig:ant_circle_hyper_costs}
    }
    \subfloat[Ant Gather]{
        \includegraphics[width=0.185\textwidth]{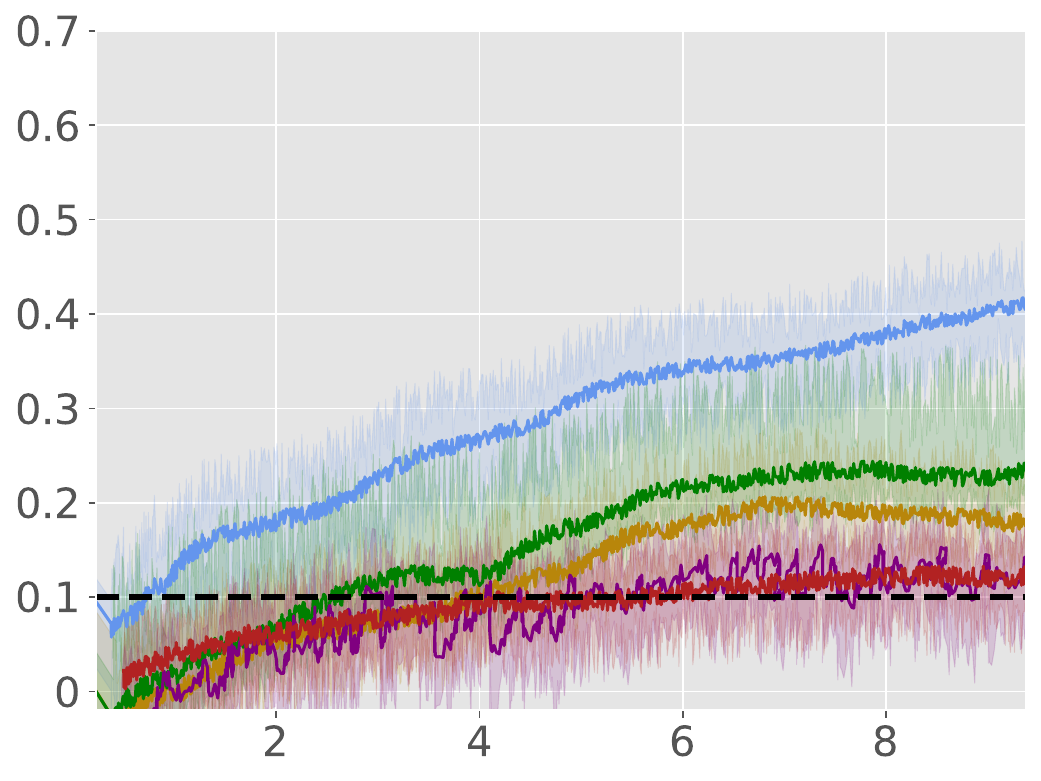}
        \label{fig:ant_gather_hyper_costs}
    }
    \subfloat[Bottleneck]{
        \includegraphics[width=0.185\textwidth]{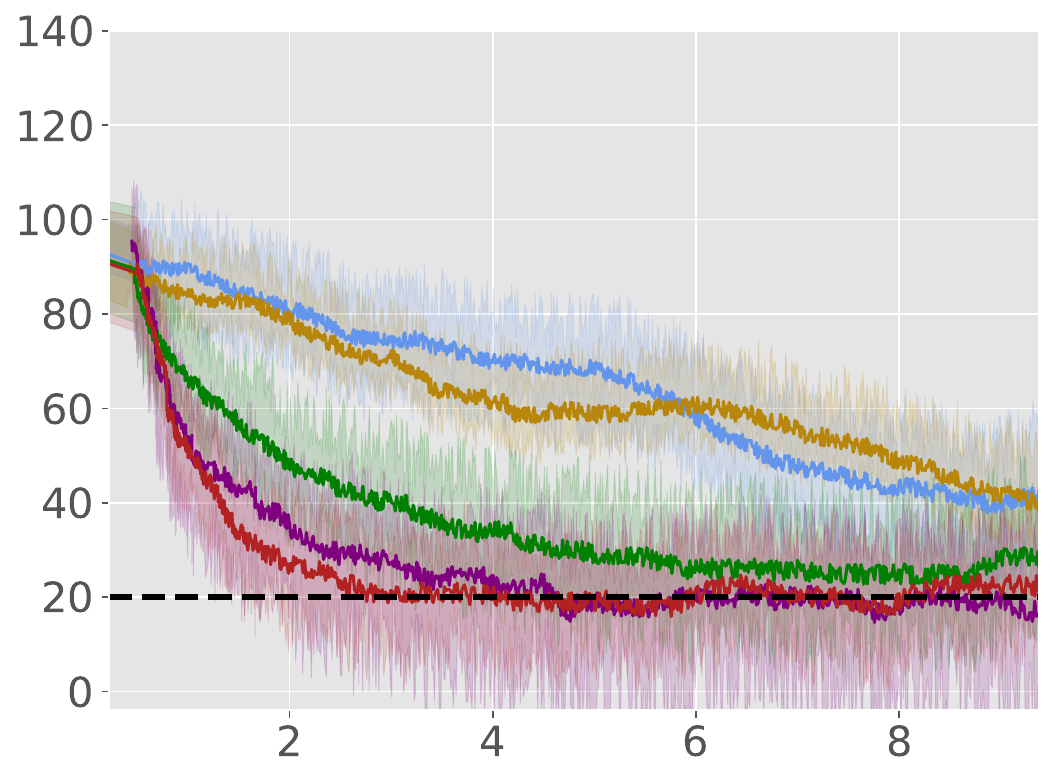}
        \label{fig:bottleneck_hyper_costs}
    }
    \subfloat[Grid]{
        \includegraphics[width=0.185\textwidth]{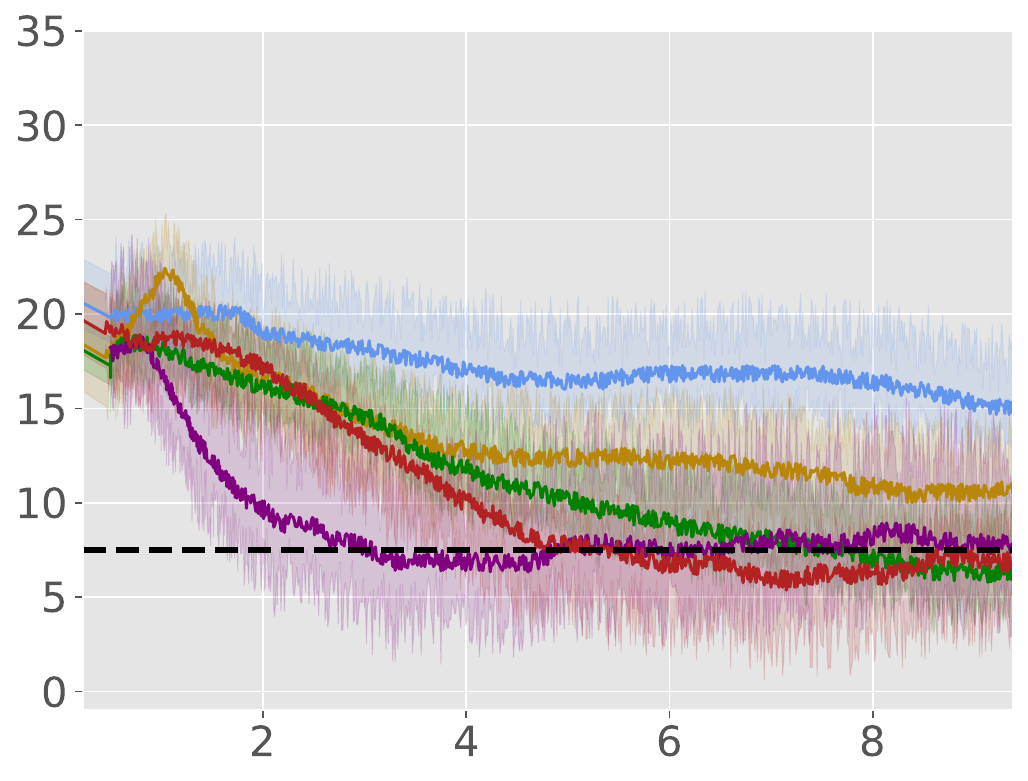}
        \label{fig:grid_hyper_costs}
    }
        \caption[]{Comparison of performance of ACPO with different values of the hyperparameter $t$ in various environment. X-axis is iterations in $10^{4}$.}
    \label{fig:hyperparam_t_appendix}
\end{figure*}

In Subsection \ref{subsec:recovery_regime}, we studied the effect of the hyperparameter $t$ for only one task. Figure \ref{fig:hyperparam_t_appendix} shows the performance of ACPO with different values of $t$ in various environments.

\chapter{Constrained Policy Optimization for Episodic Learning}
\label{chapter:ecop}

\paragraph{Abstract}. In this chapter, we present the \texttt{e-COP} algorithm, the first policy optimization algorithm for constrained Reinforcement Learning (RL) in episodic (finite horizon) settings. Such formulations are applicable when there are separate sets of optimization criteria and constraints on a system's behavior. We approach this problem by first establishing a policy difference lemma for the episodic setting, which provides the theoretical foundation for the algorithm. Then, we propose to combine a set of established and novel solution ideas to yield the \texttt{e-COP} algorithm that is easy to implement and numerically stable, and provide a theoretical guarantee on optimality under certain scaling assumptions. Through extensive empirical analysis using benchmarks in the Safety Gym suite, we show that our algorithm has similar or better performance than SoTA (non-episodic) algorithms adapted for the episodic setting. The scalability of the algorithm  opens the door to its application in safety-constrained Reinforcement Learning from Human Feedback for Large Language or Diffusion Models.

\section{Introduction}
\label{sec:introduction}

RL problems may be formulated in order to satisfy multiple simultaneous objectives. These can include performance objectives that we want to maximize, and physical, operational or other objectives that we wish to constrain rather than maximize. For example, in robotics, we often want to optimize a task completion objective while obeying physical safety constraints. Similarly, in generative AI models, we want to optimize for human preferences while ensuring that the output generations remain safe (expressed perhaps as a threshold on an automatic safety score that penalizes violent or other undesirable content). Scalable policy optimization algorithms such as TRPO \cite{schulman2015trust}, PPO \cite{schulman2017proximal}, etc have been central to the achievements of RL over the last decade \cite{silver2017mastering,vinyals2019grandmaster,akkaya2019solving}. In particular, these have found utility in generative models, e.g., in the training of Large Language Models (LLMs) to be aligned to human preferences through the RL with Human Feedback (RLHF) paradigm \cite{achiam2023gpt}. But these algorithms are designed primarily for the unconstrained infinite-horizon discounted setting: Their use for constrained problems via optimization of the Lagrangian often gives unsatisfactory constraint satisfaction results. This has prompted development of a number of constrained policy optimization algorithms  for the infinite-horizon discounted setting \cite{achiam2017constrained, yang2020projection, zhang2020first, zhang2022penalized, Hu_2022_CVPR, 9030307, agnihotri2024online}, and for the average setting \cite{agnihotri24acpo}.

However, many RL problems are more accurately formulated as episodic, i.e., having a finite time horizon. For instance, in image diffusion models \cite{black2023training,janner2022planning}, the denoising sequence is really a \emph{finite} step trajectory optimization problem, better suited to be solved via RL algorithms for the episodic setting. When existing algorithms for infinite horizon discounted setting are used for such problem, they exhibit sub-optimal performance or fail to satisfy task-specific constraints by prioritizing short-term constraint satisfaction over episodic goals \cite{bojun2020steady, neu2020unifying, greenberg2021detecting}. Furthermore, the episodic setting allows for objective functions to be time-dependent which the infinite-horizon formulations do not. Even when the objective functions are time-invariant, there is a key difference: for non-episodic settings, a stationary policy that is optimal exists whereas for episodic settings, the optimal policy is always non-stationary and time-dependent. This necessitates development of policy optimization algorithms specifically for the episodic constrained setting. We note that such policy optimization algorithms do not exist even for the unconstrained episodic setting. 

In this chapter, we introduce \texttt{e-COP}, a policy optimization algorithm for episodic constrained RL problems. Inspired by PPO, it uses deep learning-based function approximation, a KL divergence-based proximal trust region and gradient clipping, along with several novel ideas specifically for the finite horizon episodic and constrained setting. The algorithm is based on a technical lemma (Lemma \ref{lem:episodicpolicydifference}) on the difference between two policies, which leads to a new loss function for the episodic setting. We also introduce other ideas that obviate the need for matrix inversion thus improving scalability and numerical stability. The resulting algorithm improves performance over state-of-the-art baselines (after adapting them for the episodic setting). In sum, the \texttt{e-COP} algorithm has the following advantages: (i) \textit{Solution equivalence:} We show that the solution set of our \texttt{e-COP} loss function is same as that of the original CMDP problem, leading to precise cost control during training and avoidance of approximation errors (see Theorem \ref{th:relu_solution_same});  (ii) \textit{Stable convergence:} The \texttt{e-COP} algorithm converges tightly to safe optimal policies without the oscillatory behavior seen in other algorithms like PDO \cite{chow2017risk} and FOCOPS \cite{zhang2020first};  (iii) \textit{Easy integration:} \texttt{e-COP} follows the skeleton structure of PPO-style algorithms using clipping instead of steady state distribution approximation, and hence can be easily integrated into existing RL pipelines; and (iv) \textit{Empiricial performance:} the \texttt{e-COP} algorithm demonstrates superior performance on several benchmark problems for constrained optimal control compared to several state-of-the-art baselines. 

\noindent\textbf{Our Contributions and Novelty.} 
We introduce the first policy optimization algorithm for episodic RL problems, both with and without constraints (no constraints is a special case). While some of the policy optimization algorithms can be adapted for the constrained setting via a Lagrangian formulation, as we show, they don't work so well empirically in terms of constraint satisfaction and optimality. The algorithm is based on a policy difference (technical) lemma, which  is novel. We have gotten rid of Hessian matrix inversion, a common feature of policy optimization algorithms (see, for example, PPO \citep{schulman2017proximal}, CPO \citep{achiam2017constrained}, P3O \citep{zhang2022penalized}, etc.) and replaced it with a quadratic penalty term which also improves numerical stability near the edge of the constraint set boundary - a problem unique to constrained RL problems. We provide an instance-dependent hyperparameter tuning routine that generalizes to various testing scenarios. And finally, our extensive empirical results against an extensive suite of baseline algorithms (e.g., adapted PPO \cite{schulman2017proximal}, FOCOPS \cite{zhang2020first}, CPO \cite{achiam2017constrained}, PCPO \cite{yang2020projection}, and P3O \cite{zhang2022penalized}) show that \texttt{e-COP} performs the best or near-best on a range of Safety Gym \citep{brockman2016openai} benchmark problems.

\noindent\textbf{Related Work.} A broad view of planning and model-free RL techniques for Constrained MDPs is provided in \cite{liu2021policy} and \cite{gu2022review}. The development of SOTA policy optimization started with the TRPO algorithm \cite{schulman2015trust}, which used a trust region to update the current policy, and was further improved in PPO by use of proximal ideas \cite{schulman2017proximal}. This led to works like CPO \cite{achiam2017constrained}, RCPO \cite{tessler2018reward}, and PCPO \cite{yang2020projection} for constrained RL problems in the infinite-horizon discounted setting. ACPO \cite{agnihotri24acpo} extended CPO to the infinite-horizon average setting. These methods typically  require inversion of a computationally-expensive Fischer information matrix at each update step, thus limiting scalability. Lagrangian-based algorithms \cite{Ray2019, ray2019benchmarking} showed that you could incorporate constraints but constrained satisfaction remained a concern. Algorithms like PDO \cite{chow2017risk} and RCPO \cite{tessler2018reward} also  use Lagrangian duality but to solve risk-constrained RL problems and suffer from computational overhead. Some other notable algorithms include IPO \cite{liu2020ipo}, P3O \cite{zhang2022penalized},  APPO \cite{Dai2023}, etc. that use penalty terms, and hence do not suffer from computational overhead but have other drawbacks. For example, IPO assumes feasible intermediate iterations, which cannot be fulfilled in practice, P3O requires arbritarily large penalty factors for feasibility which can lead to significant estimation errors. We note that all the above algorithms are for the infinite-horizon discounted (non-episodic) setting (except ACPO \cite{agnihotri24acpo}, which is for the average setting). We are not aware of any policy optimization algorithm for the episodic RL problem, with or without constraints.

\section{Preliminaries}

An episodic, or fixed horizon  Markov decision process (MDP) is a tuple, $\Mcal := (S,A,r,P,\mu, H)$, where $S$ is the set of states, $A$ is the set of actions, $r : S \times A \times S \to \Rbb$ is the reward function, $P : S \times A \times S \to [0,1]$ is the transition probability function such that $P(s'|s,a)$ is the probability of transitioning to state $s'$ from state $s$ by taking action $a$, $\mu : S \to [0,1]$ is the initial state distribution, and $H$ is the time horizon for each episode (characterized by a terminal state $s_{H}$). 

A policy $\pi : S \to \Delta(A)$ is a mapping from states to probability distributions over the actions, with $\pi(a|s)$ denoting the probability of selecting action $a$ in state $s$, and $\Delta(A)$ is the probability simplex over the action space $A$. However, due to the temporal nature of episodic RL, the optimal policies are generally not stationary, and we index the policy at time $h$ by $\pi_{h}$, and denote $\pib_{1:H} = ( \pi_{h} )_{h=1}^{H}$. Then, the total undiscounted reward objective within an episode is defined as 
\begin{align*}
    J(\pib_{1:H}) &:=  \E{\tau \sim \pib_{1:H}} \left[\sum_{h=1}^{H} r(s_{h}, a_{h}, s_{h+1}) \right]
\end{align*} 

where $\tau$ refers to the sample trajectory $(s_1, a_1, s_2, a_2, \dots, s_{H})$ generated when following a policy sequence, i.e., $a_h \sim \pi_{h}(\cdot | s_h)$, ~$s_{h+1} \sim P(\cdot | s_h, a_h),$ and $s_{1} \sim \mu$. 

Let $R_{h:H}(\tau)$ denote the total reward of a trajectory $\tau$ starting from time $h$ until  episode terminal time $H$ generated by following the policy sequence $\pib_{h:H}$. We also define the state-value function of a state $s$ at step $h$ as $V^{\pib}_{h}(s) := \E{\tau \sim \pib}[R_{h:H}(\tau) \given s_h = s]$ and the action-value function as $Q^{\bm{\pi}}_{h}(s,a) := \E{\tau \sim \bm{\pi}} [R_{h:H}(\tau) \given s_h= s, a_h = a]$. The advantage function is $A^{\bm{\pi}}_{h}(s,a) := Q^{\bm{\pi}}_{h}(s,a) - V^{\bm{\pi}}_{h}(s)$. We also define $\Pbb^{\pib}_{h}(s \given s_{1}) = \sum_{a \in A} \Pbb^{\pib}_{h}(s,a \given s_{1})$, where the term $\Pbb^{\pib}_{h}(s,a \given s_{1})$ is the probability of reaching $(s,a)$ at time step $h$ following $\pib$ and starting from $s_{1}.$


\noindent\textbf{Constrained MDPs.}
A constrained Markov decision process (CMDP) is an MDP augmented with constraints that restrict the set of allowable policies for that MDP.  Specifically, we have $m$  cost functions, $C_1, \cdots , C_m$ (with each function $C_i : S \times A \times S \to \Rbb$ mapping transition tuples to costs, similar to the reward function), and bounds $d_1, \cdots , d_m$. And similar to the value function for the reward objective, we define the expected total cost objective for each cost function $C_{i}$ (called cost value for the constraint) as 
\begin{align*}
    J_{C_i}(\pib_{1:H}) &:=  \E{\tau \sim \pib_{1:H}} \left[\sum_{h=1}^{H} C_i(s_{h}, a_{h}, s_{h+1}) \right].
\end{align*} 
The goal then, in each episode, is to find a policy sequence $\pib_{1:H}^{\star}$ such that 
\begin{equation}
J(\pib_{1:H}^{\star})  := \max_{ \pib_{1:H} \in \Pi_{C} } J(\pib_{1:H}), ~~\text{where}~~~\Pi_{C} := \left\{\pib_{1:H} \in \Pi \; : \; J_{C_i}(\pib_{1:H}) \leq d_i, \forAll i \in [m] \right\}
\label{eq:cmdp_problem}
\end{equation}
is the set of feasible policies for a CMDP for some given class of policies $\Pi$.
Lastly, analogous to $\vfunc_{h}$, $\qfunc_{h}$, and $\adv_{h}$, we can also define quantities for the cost functions  $C_i(\cdot)$ by replacing, and denote them by $\vfunc_{C_{i}, h}$, $\qfunc_{C_{i}, h}$, and $\adv_{C_{i}, h}$. Proofs of theorems and statements, if not given, are available in Appendix \ref{ecop:appendix}. 


\noindent\textbf{Notation.} 
$[N]$ denotes $\{1, \dots, N\}$ for some $N \in \Nbb$. $\pi_{h}$ refers to the policy at time step $h$ within an episode. Denote $\pib_{s:t} := ( \pi_{s}, \pi_{s+1}, \dots, \pi_{t})$ for some $s \leq t$ with $s,t \in [H]$. 
We shall only write $\pi_{k,h}$ when it is necessary to differentiate policies from different episodes but at the same time $h$. It then naturally follows to define $\pib_{k,s:t}$ to be the sequence $\pib_{s:t}$ in episode $k$. We will denote $\pib_{k} \equiv \pib_{k,1:H}$, and where not needed drop the index for the episode so that $\pib \equiv \pib_{k}$.

\section{Episodic Constrained Optimization of Policies (e-COP)}
\label{sec:e-COP}


In this section, we propose a constrained policy optimization algorithm for episodic MDPs. Policy optimization algorithms for MDPs have proven remarkably effective given their ability to computationally scale up to high dimensional continuous state and action spaces \cite{schulman2015trust, schulman2016high, schulman2017proximal}. Such algorithms have also been proposed for infinite-horizon constrained MDPs with discounted criterion \cite{achiam2017advanced}  as well as  the average criterion \cite{agnihotri24acpo} but not for the finite horizon (or as it is often called, the episodic) setting. 

We note that finite horizon is not simply a special case of infinite-horizon discounted setting since the reward/cost functions in the former can be time-varying while the latter only allows for time-invariant objectives. Furthermore, even with time-invariant objectives,  the optimal policy is time-dependent, while for the latter setting there an optimal policy that is stationary exists.

\noindent\textbf{A Policy Difference Lemma for Episodic MDPs.} 
Most policy optimization RL algorithms are based on a value or policy difference technical lemma \cite{kakade2002approximately}. Unfortunately, the policy difference lemmas that have been derived previously for the infinite-horizon discounted \cite{schulman2015trust} and average case \cite{agnihotri24acpo} are not applicable here and hence, we derive a new policy difference lemma for the episodic setting. 


\begin{restatable}{lemma}{episodicpolicydifference}
\label{lem:episodicpolicydifference} For an episode of length $H$ and two policies, $\pib$ and $\pib'$, the difference in their performance assuming identical initial state distribution $\mu$ (i.e., $s_{1} \sim \mu$) is given by
\begin{equation}
J(\pib) - J(\pib') = \sum_{h=1}^{H} \E{s_{h}, a_{h} \sim \Pbb^{\pib}_{h}(\cdot, \cdot \given s) \\ s_{1} \sim \mu} \big[  A^{\pib'}_{h}(s_{h}, a_{h}) \big].
\end{equation}
\end{restatable}




The proof can be found in Appendix \ref{proof:episodicpolicydifference}. A key difference to note between the above and similar results for infinite-horizon settings \cite{schulman2015trust, agnihotri24acpo} is that considering stationary policies (and hence corresponding occupation measures) is not enough for the episodic setting since, in general, such a policy may be far from optimal. This explains why Lemma \ref{lem:episodicpolicydifference} looks so different (e.g., see (2) in \cite{schulman2015trust}, and Lemma 3.1 in \cite{agnihotri24acpo}). Indeed, the lemma above indicates that policy updates do not have to recurse backwards from the terminal time as dynamic programming algorithms do for episodic settings, which is somewhat surprising.

\noindent\textbf{A Constrained Policy Optimization Algorithm for Episodic MDPs.} 
Iterative policy optimization algorithms achieve state of the art performance \cite{schulman2017proximal, tessler2018reward, yang2020projection} on RL problems. Most such algorithms maximize the advantage function based on a suitable policy difference lemma, solving an unconstrained RL problem. Some additionally ensure satisfaction of infinite horizon expectation  constraints \cite{achiam2017constrained, agnihotri24acpo}. However, given that our policy lemma for the episodic setting (Lemma \ref{lem:episodicpolicydifference}) is significantly different, we need to re-design the algorithm based on it. A first attempt is presented as Algorithm \ref{alg:iterativepolicyoptimization},  where each iteration corresponds to an update with a full horizon $H$ episode.

\begin{algorithm}[!ht]
   \caption{\textbf{I}terative \textbf{P}olicy \textbf{O}ptimization for \textbf{C}onstrained \textbf{E}pisodic  (IPOCE) RL}
\begin{algorithmic}[1]
   \STATE {\bfseries Input:} Initial policy $\pib_{0}$, number of episodes $K$, episode horizon $H$.
    \FOR{$k = 1,2, \dots ,K$} 
    \STATE Run $\pib_{k-1}$ to collect trajectories $\tau$.
    \STATE Evaluate $A^{\pib_{k-1}}_{h}$ and $A_{C_i, h}^{\pib_{k-1}}$ for $h \in [H]$ from $\tau$.
      \FOR{$t = H, H-1, \dots, 1$} 
	 \STATE \hspace{-0.2cm} \hspace{-1.75cm} $\vcenter{
    \begin{equation}
        \resizebox{0.85\linewidth}{!}{$
            \begin{aligned}
            \pi_{k,t}^{\star} = \argmin_{\pi_{k,t}} \sum_{h=t}^H \E{\substack{s \sim \rho_{\pi_{k,h}}\\ a\sim \pi_{k,h}}}[ -A^{\pib_{k-1}}_{h} (s,a)] 
             \quad \mathrm{s.t.}  \quad J_{C_i}(\pib_{k-1}) +  \sum_{h=t}^H {\E{\substack{s \sim \rho_{\pi_{k,h}}\\ a\sim \pi_{k,h}}}} \big  [A_{C_i, h}^{\pib_{k-1}} (s,a) \big ] \leq d_i,\ \forall i \hspace{-1cm}
            \label{eq:cpo_improvement_vanilla}
            \end{aligned}
        $}
    \end{equation}
    }$
	 \ENDFOR
  \STATE Set $\pib_{k} \leftarrow \left(\pi_{k,1}^{\star}, \pi_{k,2}^{\star}, \, \dots \, , \pi_{k,H}^{\star} \right)$.
    \ENDFOR
\end{algorithmic}
\label{alg:iterativepolicyoptimization}
\end{algorithm}

The iterative constrained policy optimization algorithm introduced above uses the current iterate of the policy $\pib_k$ to collect a trajectory $\tau$, and use them to evaluate $A^{\pib_{k-1}}_{h}$ and $A_{C_i, h}^{\pib_{k-1}}$ for $h \in [H]$. At the end of the episode, we solve $H$ optimization problems (one for each $h \in [H]$) that result in a new sequence of policies $\pib$. As is natural in episodic problems, we do backward iteration in time, i.e., solve the problem in step (6) at $h=H$, and then go backwards towards $h=1$. 

Note that the expectation of advantage functions in equation \eqref{eq:cpo_improvement_vanilla} is with respect to the policy $\pi$ (the optimization variable) and its corresponding \emph{time-dependent} state occupation distribution  $\rho_{\pi_h}$. In the infinite-horizon settings, the expectation is with respect to the steady state stationary distribution, but that does not exist in the episodic setting.  



\noindent\textbf{Using current policy for action selection.} Algorithm \ref{alg:iterativepolicyoptimization} represents an exact principled solution tothe constrained episodic MDP, but the intractable optimization performed in \eqref{eq:cpo_improvement_vanilla} makes it impractical (as in the case of infinite horizon policy optimization algorithms \cite{schulman2015trust, achiam2017constrained, yang2020projection}). We proceed to introduce a sequence of ideas that make the algorithm practical (e.g.,  by avoiding computationally expensive Hessian matrix inversion for use with trust-region methods  \cite{schulman2017proximal, chow2019lyapunov, zhang2022penalized}). However, getting rid of trust regions leads to large updates on policies, but PPO \cite{schulman2017proximal} and P3O \cite{zhang2022penalized} successfully overcome this problem by clipping the advantage function and adding a ReLU-like penalty term to the minimization objective. Motivated by this, we rewrite the optimization problem in \eqref{eq:cpo_improvement_vanilla} as follows by parameterizing the policy  $\pib_{k,t}$ in episode $k$ and time step $t$ by $\theta_{k,t}$:
\begin{equation}
\resizebox{0.99\linewidth}{!}{$
\begin{aligned}
\pi_{k,t} & =  \mathop{\arg\min}_{\pi_{k,t}} \sum_{h=t}^{H} \mathop{\E{\substack{s \sim \rho_{\pi_{k,h}}\\ a\sim \pi_{k-1,h}}}} \big[-{\rho(\theta_{h})}A^{\pib_{k-1}}_{h}(s,a) \big] + \sum_{i}^{m} \lambda_{t,i} \max \bigg\{0, \; \sum_{h=t}^{H} \mathop{{\E{\substack{s \sim \rho_{\pi_{k,h}}\\ a\sim \pi_{k-1,h}}}}} \big  [{\rho(\theta_{h})}A_{C_i, h}^{\pib_{k-1}} (s,a) \big ] + J_{C_i}(\pib_{k-1})-d_i \bigg\}  \, , \hspace{-0.45cm}
\end{aligned}
\label{eq:p3po_objective_fn}
$}
\end{equation}
where ${\rho(\theta_{h})} = \frac{\pi_{\theta_{k,h}}}{\pi_{\theta_{k-1,h}}}$ is the importance sampling ratio, $\lambda_{t,i}$ is a penalty factor for constraint $C_{i}$, and  $\pi_{k,\theta_{h}} \equiv \pi_{k,h} \equiv \theta_{k,h}$. Note that the ReLU-like penalty term above is different from the traditional first-order and second-order gradient approximations that are employed in trust-region methods \cite{achiam2017constrained, yang2020projection}. In essence, the penalty is applied when the agent breaches the associated constraint, while the objective remains consistent with standard policy optimization when all constraints are satisfied. 



\noindent\textbf{Introducing quadratic damping penalty.}
\label{sec:intro_quadratic_term}
It has been noted in such iterative policy optimization  algorithms that the behaviour of the learning agent when it nears the constraint threshold is quite volatile during training \cite{achiam2017constrained, yang2020projection, zhang2021average}. This is because the penalty term is active only when the constraints are violated which results in sharp behavior change for the agent. To alleviate this problem, we introduce an additional quadratic damping term to the objective above, which provides stable cost control to compliment the lagged Lagrangian multipliers. This has proved effective in physics-based control applications \cite{dowling2005feedback, klimesch2012alpha, imayoshi2013oscillatory} resulting in improved convergence since the damping term provides stability,  while keeping the solution set the same as for the original Problem \eqref{eq:cpo_improvement_vanilla} and the adapted Problem \eqref{eq:p3po_objective_fn} (as we prove later). 

For brevity, we denote the constraint term in Problem \eqref{eq:p3po_objective_fn} as
\begin{center}
\resizebox{0.7\linewidth}{!}{$
$$
\Psi_{C_{i}, t}(\pib_{k-1}, \pib_{k}) := \sum_{h=t}^{H} \mathop{{\E{\substack{s \sim \rho_{\pi_{k,h}}\\ a\sim \pi_{k-1,h}}}}} \big  [{\rho(\theta_{h})}A_{C_i, h}^{\pib_{k-1}} (s,a) \big ] + J_{C_i}(\pib_{k-1})-d_i.
$$
$}
\end{center}
\vspace{-0.35cm}
Now introduce a slack variable $x_{t,i} \geq 0$ for each constraint to convert the inequality constraint ($\Psi_{C_{i}, t}(\cdot,\cdot) \leq 0 $) to equality by letting 
$$w_{t,i}(\pib_{k}) := \Psi_{C_{i}, t}(\pib_{k-1}, \pib_{k}) + x_{t,i} = 0.$$ 

With this notation, we restate Problem \eqref{eq:p3po_objective_fn} as:

\begin{center}
\resizebox{0.92\linewidth}{!}{$
$$
 \pi_{k,t}^{\star} = \min_{\pi_{k,t}} \; {\Lcal}_{t}(\pib_{k}, \bm{\lambda}) \; := \;
\sum_{h=t}^{H} {\E{\substack{s \sim \rho_{\pi_{k,h}}\\ a\sim \pi_{k-1,h}}}} \big[-{\rho(\theta_{h})}A^{\pib_{k-1}}_{h}(s,a) \big] + \sum_{i}^{m} \lambda_{t,i} \max \{ 0, \Psi_{C_{i}, t}(\pib_{k-1}, \pib_{k}) \}.
$$
$}   
\end{center}
\vspace{-0.35cm}

Now we introduce the quadratic damping term and the intermediate loss function then takes the form,
\begin{equation}
\resizebox{.78\linewidth}{!}{$
\begin{aligned}
\Lcal_{t}(\pib_{k}, \bm{\lambda}, \bm{x}, \beta) &:=  \sum_{h=t}^{H} {\E{\substack{s \sim \rho_{\pi_{k,h}}\\ a\sim \pi_{k-1,h}}}} \big[-{\rho(\theta_{h})}A^{\pib_{k-1}}_{h}(s,a) \big] + \sum_{i}^{m} \lambda_{t,i} w_{t,i}(\pib_{k}) + \frac{\beta}{2} \sum_{i}^{m} w_{t,i}^{2}(\pib_{k}) \\
\text{Then,} &\quad\quad (\pi_{k,t}^{\star}, \bm{\lambda}_{t}^{\star}, \bm{x}_{t}^{\star}) = \max_{\bm{\lambda} \geq 0} \min_{\pi_{k,t}, \bm{x}} \Lcal_{t}(\pib_{k}, \bm{\lambda}, \bm{x}, \beta) \; , \hspace{1cm}
\label{eq:e-COP_initial_fn}
\end{aligned}
$} 
\end{equation}
where $\beta$ is the damping factor, $\bm{\lambda}_{t} = \left(\lambda_{t,i}\right)_{i=1}^{m}$, and $\bm{x}_{t} = \left(x_{t,i}\right)_{i=1}^{m}$. We can then construct a primal-dual solution to the max-min optimization problem. The need for a slack variable $\bm{x}$ can be obviated by setting the partial derivative of $\Lcal_{t}(\cdot)$ with respect to $\bm{x}$ equal to 0. This leads to a ReLU-like solution: $x_{t,i}^{\star} = \max \big(0, - \Psi_{C_{i}, t}(\pib_{k-1}, \pib_{k}) - \frac{\lambda_{t,i}}{\beta} \big)$. The intermediate problem then takes the form as below.

\begin{restatable}{proposition}{dampedintermediateproblem}
\label{prop:damped_intermediate_problem}
The inner optimization problem in \eqref{eq:e-COP_initial_fn} with respect to $\bm{x}$ is a convex quadratic program with
non-negative constraints, which can be solved to yield the following intermediate problem:
\begin{equation}
\resizebox{.92\linewidth}{!}{$
\begin{aligned}
 (\pi_{k,t}^{\star}, \bm{\lambda}_{t}^{\star}) = \max_{\bm{\lambda} \geq 0} \min_{\pi_{k,t}} & \; \, \Lcal_{t}(\pib_{k}, \bm{\lambda}, \beta), \quad \text{\small where} \\
\Lcal_{t}(\pib_{k}, \bm{\lambda}, \beta) = \sum_{h=t}^{H} {\E{\substack{s \sim \rho_{\pi_{k,h}}\\ a\sim \pi_{k-1,h}}}} \big[-{\rho(\theta_{h})}A^{\pib_{k-1}}_{h}(s,a) \big] &+ \frac{\beta}{2} \sum_{i}^{m} \bigg( \max \bigg\{0, \Psi_{C_{i}, t}(\pib_{k-1}, \pib_{k}) + \frac{\lambda_{t,i}}{\beta} \bigg\}^{2} - \frac{\lambda_{t,i}^{2}}{\beta^{2}}  \bigg). 
\end{aligned}
$}
\label{eq:damped_intermediate_problem}
\end{equation}
\end{restatable}

The proof can be found in Appendix \ref{proof:damped_intermediate_problem}. One can see that the cost penalty is active when $\Psi_{C_{i}, t}(\pib_{k-1}, \pib_{k})$ $\geq - \frac{\lambda_{t,i}}{\beta}$ rather than when $\Psi_{C_{i}, t}(\pib_{k-1}, \pib_{k}) \geq 0$. This allows the agent to act in a constrained manner even before the constraint is violated. Further, as we show next, the introduction of the damping factor and the RELU-like penalty does not change the solution of the problem (under some suitable assumptions):

\begin{restatable}{theorem}{relusolutionsame}
\label{th:relu_solution_same} Let $\pi^{{\eqref{eq:cpo_improvement_vanilla}^{\star}}}$ be a solution to Problem \eqref{eq:cpo_improvement_vanilla}, and let $\big(\pi^{{\eqref{eq:damped_intermediate_problem}^{\star}}}, \bm{\lambda}^{{\eqref{eq:damped_intermediate_problem}^{\star}}} \big)$ be a solution to Problem \eqref{eq:damped_intermediate_problem}. Then, for sufficiently large $\beta > \bar\beta$ and $\lambda_{t,i} > \bar\lambda \forAll i$, $\pi^{{\eqref{eq:cpo_improvement_vanilla}^{\star}}}$ is a solution to Problem \eqref{eq:damped_intermediate_problem}, and $\pi^{{\eqref{eq:damped_intermediate_problem}}^{\star}}$ is a solution to Problem \eqref{eq:cpo_improvement_vanilla}. 
\end{restatable}

We refer the reader to Appendix \ref{proof:relu_solution_same} for the proof. This theorem implies that we can search for the optimal feasible policies of the CMDP Problem \eqref{eq:cmdp_problem} by iteratively solving Problem \eqref{eq:damped_intermediate_problem}. Next, we make some further modifications to Problem \eqref{eq:damped_intermediate_problem} that give us our final tractable algorithm.

\noindent\textbf{Removing Lagrange multiplier dependency.} Problem \eqref{eq:damped_intermediate_problem} requires a primal-dual algorithm that will iteratively solve over the policies and the dual variable $\lambda$. But from the Lagrangian, we can actually take a derivative with respect to $\lambda$, and then solve for it, which yields the following update rule for it:
\begin{equation}
\lambda_{t,i}^{(k)} = \max \big(0, \lambda_{t,i}^{(k-1)} + \beta \Psi_{C_{i}, t}(\pib_{k-1}, \pib_{k-1}) \big).
\label{eq:lagrangeupdaterule}
\end{equation}
This update rule simplifies the optimization problem and updates the Lagrange multipliers in the $k{\textit{th}}$ episode based on the constraint violation in the $(k-1){\textit{th}}$ episode.

\noindent\textbf{Clipping the advantage functions.}
Solving the optimization problem presented in equation \eqref{eq:damped_intermediate_problem} is intractable since we do not know $\rp$ beforehand. Hence, we replace $\rp$ by the empirical distribution observed with the policy of the previous episode, $\pib_{k-1}$, i.e., $\rho_{\pi_{k,h}} \approx \rho_{\pi_{k-1,h}} \forAll h$. Similar to  \cite{schulman2017proximal} for PPO, we also  use \textit{clipped} surrogate objective functions for both the reward and cost advantage functions. Thus, the final problem combining equation \eqref{eq:p3po_objective_fn} and equation \eqref{eq:damped_intermediate_problem} can be constructed as follows.

If we let
\begin{equation*}
\resizebox{.9\linewidth}{!}{$
\begin{aligned}
\mathcal{L}_{t}(\theta) = \sum_{h=t}^{H} \mathop{{\E{\substack{s\sim \rho_{\pi_{k-1,h}} \\a\sim \pi_{k-1,h}}}}}\big [ -\min\big \{  {\rho(\theta_{h})}  A_{h}^{\pib_{k-1}}(s,a), \mathrm{clip} ({\rho(\theta_{h})}, \,  1-\epsilon, \, 1+\epsilon) A_{h}^{\pib_{k-1}} (s,a) \big \} \big ] \quad \text{and,} \\
\mathcal{L}_{C_i, t}(\theta) = \sum_{h=t}^{H} \mathop{{\E{\substack{s\sim \rho_{\pi_{k-1,h}} \\a\sim \pi_{k-1,h}}}}}\big [ -\min\big \{  {\rho(\theta_{h})}  A_{C_{i},h}^{\pib_{k-1}}(s,a), \mathrm{clip} ({\rho(\theta_{h})}, \,  1-\epsilon, \, 1+\epsilon) A_{C_{i},h}^{\pib_{k-1}} (s,a) \big \} \big ]
\end{aligned}
$}
\end{equation*}

then, the final loss function $\tilde{\Lcal}_{t}(\cdot)$ of the final problem is:


\begin{equation}
\resizebox{.93\linewidth}{!}{$
\begin{aligned}
\pi_{k,t}^{\star} = \argmin_{\pi_{k,t}} \tilde{\mathcal{L}}_{t}(\pi_{\theta}, \bm{\lambda}, \beta) := \argmin_{\pi_{k,t}} \;\;  & \mathcal{L}_{t}(\theta) + \sum_{i}^{m} \lambda_{t,i} \max \big\{0,\mathcal{L}_{C_i,t}(\theta) + \big(J_{C_i}(\pib_{k-1})-d_i \big) \big\} \\ & + \frac{\beta}{2} \sum_{i}^{m} \bigg( \max \bigg\{0,\mathcal{L}_{C_i,t}(\theta) + \big(J_{C_i}(\pib_{k-1})-d_i \big) + \frac{\lambda_{t,i}}{\beta} \bigg\}^{2} - \frac{\lambda_{t,i}^{2}}{\beta^{2}}  \bigg)
\end{aligned}
$}
\label{eq:main_e-COP_problem}
\end{equation}

Usually for experiments, Gaussian policies with means and variances predicted from neural networks are used \cite{schulman2015trust, achiam2017constrained, schulman2017proximal, yang2020projection}. We employ the same approach and since we work in the finite horizon setting, the reward and constraint advantage functions can  easily be calculated from any trajectory $\tau \sim \pib$. The surrogate problem in equation \eqref{eq:main_e-COP_problem} then includes the pessimistic bounds on Problem \eqref{eq:damped_intermediate_problem}, which is unclipped. 





\begin{algorithm}[!b]
   \caption{\textbf{E}pisodic \textbf{C}onstrained \textbf{O}ptimization of \textbf{P}olicies (\texttt{e-COP})}
\begin{algorithmic}[1]
   \STATE {\bfseries Input:} Initial policy $\theta_{0} := \pib_{0} := \pib_{\theta_0}$, critic networks $V^{\phi_0}$ and $V^{\psi_0}_{C_i} \forAll i$, penalty factor $\beta$, number of episodes $K$, episode horizon $H$, learning rate $\alpha$, penalty update factor $\kappa$.
    \FOR{$k = 1,2, \dots ,K$} 
    \STATE Collect a set of trajectories $\mathcal{D}_{k-1}$ with policy $\pib_{k-1}$ and update the critic network.
    \STATE Get updated $\lambda^{(k)}$ using Equation \eqref{eq:lagrangeupdaterule}.
      \FOR{$t = H, H-1, \dots, 1$} 
            \STATE Update the policy $\theta_{k,t} \leftarrow \theta_{k,t+1} - \alpha \nabla_{\theta} \tilde{\mathcal{L}}_{t}(\theta, \lambda^{(k)}, \beta) $ using Equation \eqref{eq:main_e-COP_problem}.
        \ENDFOR
      \IF{$\Ccal(\theta_{k}) \geq \zeta_{k}$}
      \STATE $\beta = \min(\beta_{\max}, \kappa \beta)$
      \ENDIF
    \ENDFOR
\end{algorithmic}
\label{alg:practical-ECOP}
\end{algorithm}

\noindent\textbf{Adaptive parameter selection.}
The value of $\beta$ is required to be larger than the unknown $\bar{\beta}$ according to Theorem \ref{th:relu_solution_same}, but we also know that too large a $\beta$ decays the performance (as seen in harmonic oscillator kinetic energy formulations  \cite{klimesch2012alpha, imayoshi2013oscillatory, yuan2018online}). To manage this tradeoff, we provide an instance-dependent adaptive way to adjust the damping factor as a hyperparameter. In each episode $k$, we update the damping parameter whenever a secondary constraint cost value denoted by $\Ccal(\pib_{k})$ is larger than some threshold $\zeta_{k}$. Using Proposition \ref{prop:damped_intermediate_problem}, we provide the following definitions.

$$
\Ccal(\pib_{k}) := \sum_{t=1}^{H} \sum_{i}^{m} \max \bigg\{ J_{C_i}(\pib_{k})-d_i ,  - \frac{\lambda_{t,i}^{(k)}}{\beta} \bigg\} \quad \text{and} \quad \zeta_{k} := \frac{\sqrt{m}}{\beta} \cdot \max_{t \in [H]} \big\| \bm{\lambda}^{(k)}_{t} \big\|_{\infty}
$$

Hence, we increase $\beta$ by a constant factor $\kappa > 1$ after every episode if $\Ccal(\pib_{k}) \geq \zeta_{k}$  until a stopping condition is fulfilled, typically a constant $\beta_{\max}$. This leads to constraint-satisfying iterations that are more stable, and we show that it enables a fixed $\beta$ to generalize well across various tasks. The initial $\beta$ can simply be selected by a quantified line-search to obtain a feasible $\beta > \bar{\beta}$ \cite{achiam2017constrained, agnihotri24acpo}.

Note that the final loss function in Equation \eqref{eq:main_e-COP_problem} is differentiable almost everywhere, so we could easily solve it via any first-order optimizer \cite{kingma2014adam}. The final algorithm, \texttt{e-COP}, is given in Algorithm \ref{alg:practical-ECOP}.

\section{Experiments}
\label{sec:experiments}

We conducted extensive experimental evaluation on the relative empirical performance of the \texttt{e-COP} algorithm to arrive at the following conclusions: 
(i) The \texttt{e-COP} algorithm performs better or nearly as well as all baseline algorithms for infinite-horizon discounted safe RL tasks in maximizing episodic return while satisfying given constraints.
(ii) It is more robust to stochastic and complex environments \cite{ray2019benchmarking}, even where previous methods struggle. 
(iii) It has stable behavior and more accurate cost control as compared to other baselines near the constraint threshold due to the damping term.

\begin{figure}[t]
 \captionsetup[subfigure]{justification=centering}
    \centering
    \subfloat[Humanoid]{
        \includegraphics[height=.12\textwidth, width=0.15\textwidth]{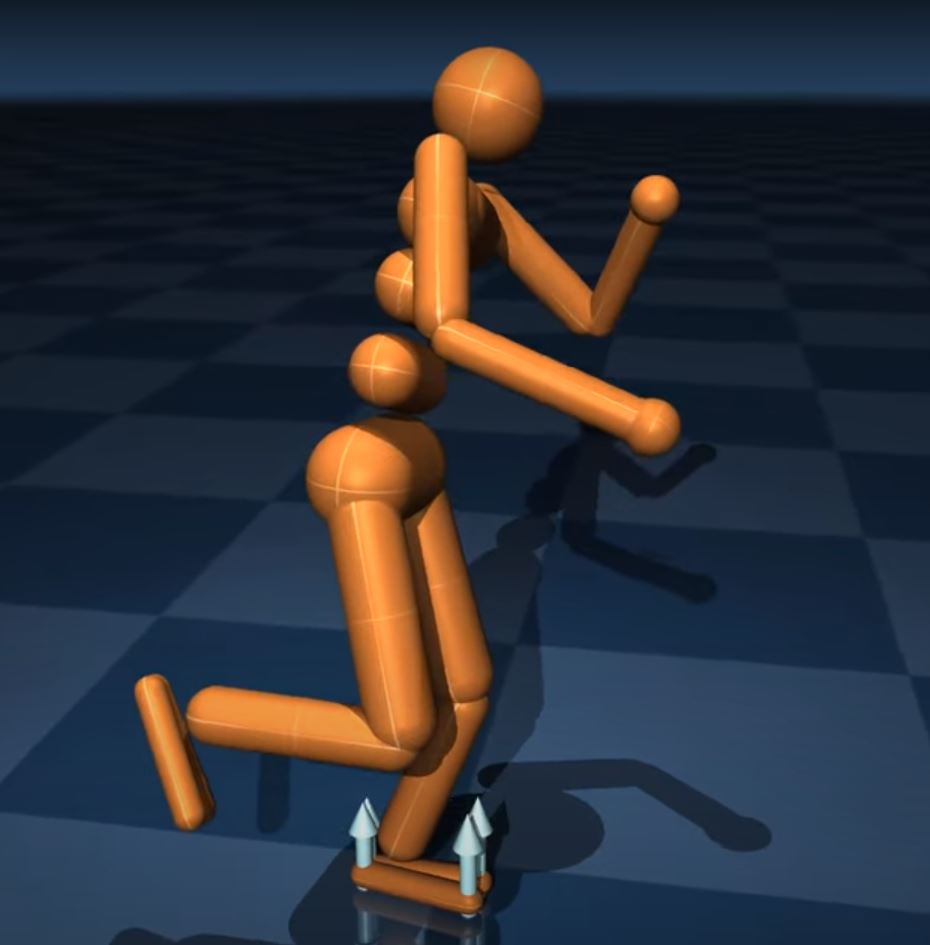}
        \label{fig:task_humanoid}
    }
    \subfloat[Circle]{
        \includegraphics[height=.12\textwidth, width=0.15\textwidth]{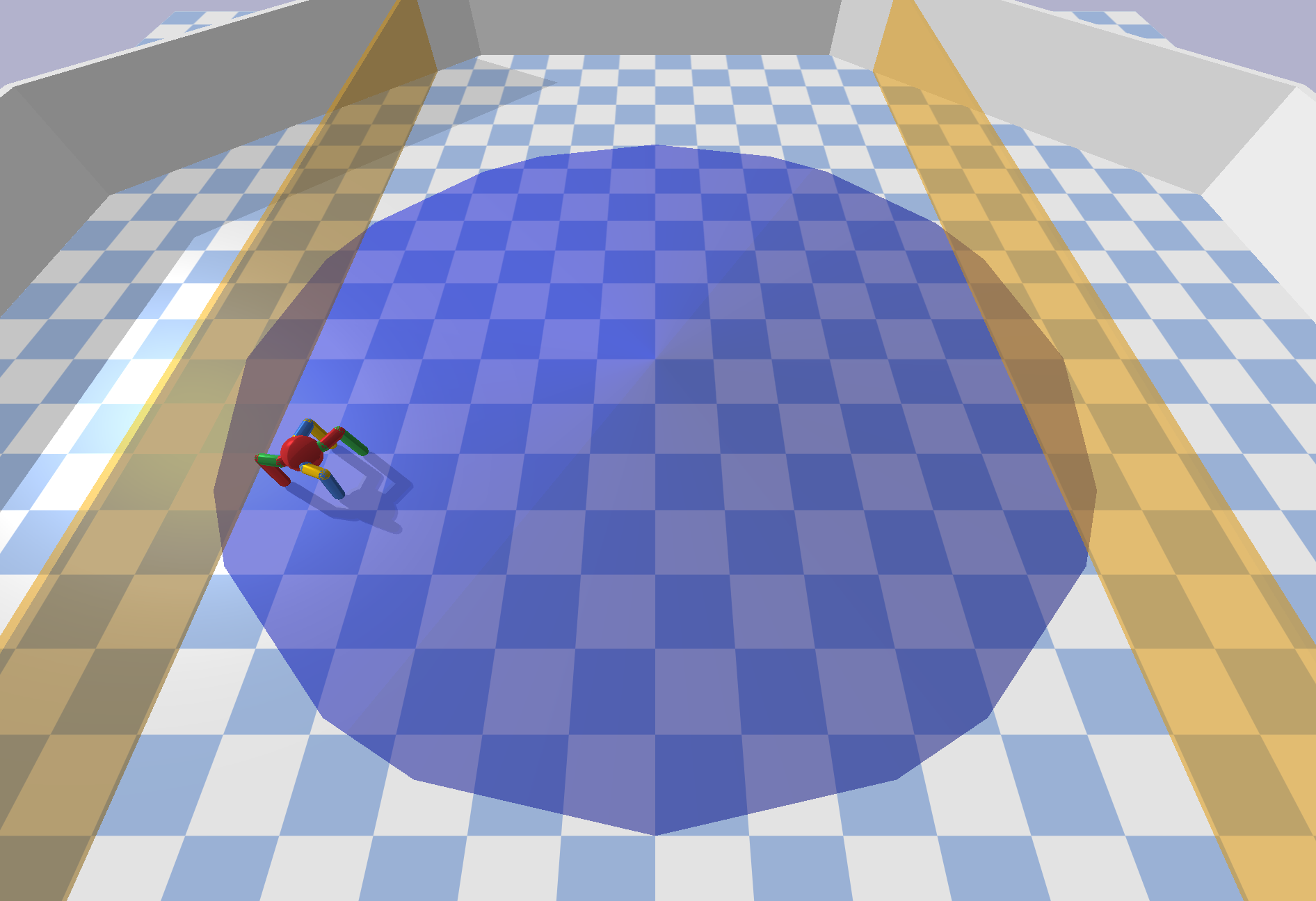}
        \label{fig:task_circle}
    }
    \subfloat[Reach]{
        \includegraphics[height=.12\textwidth, width=0.15\textwidth]{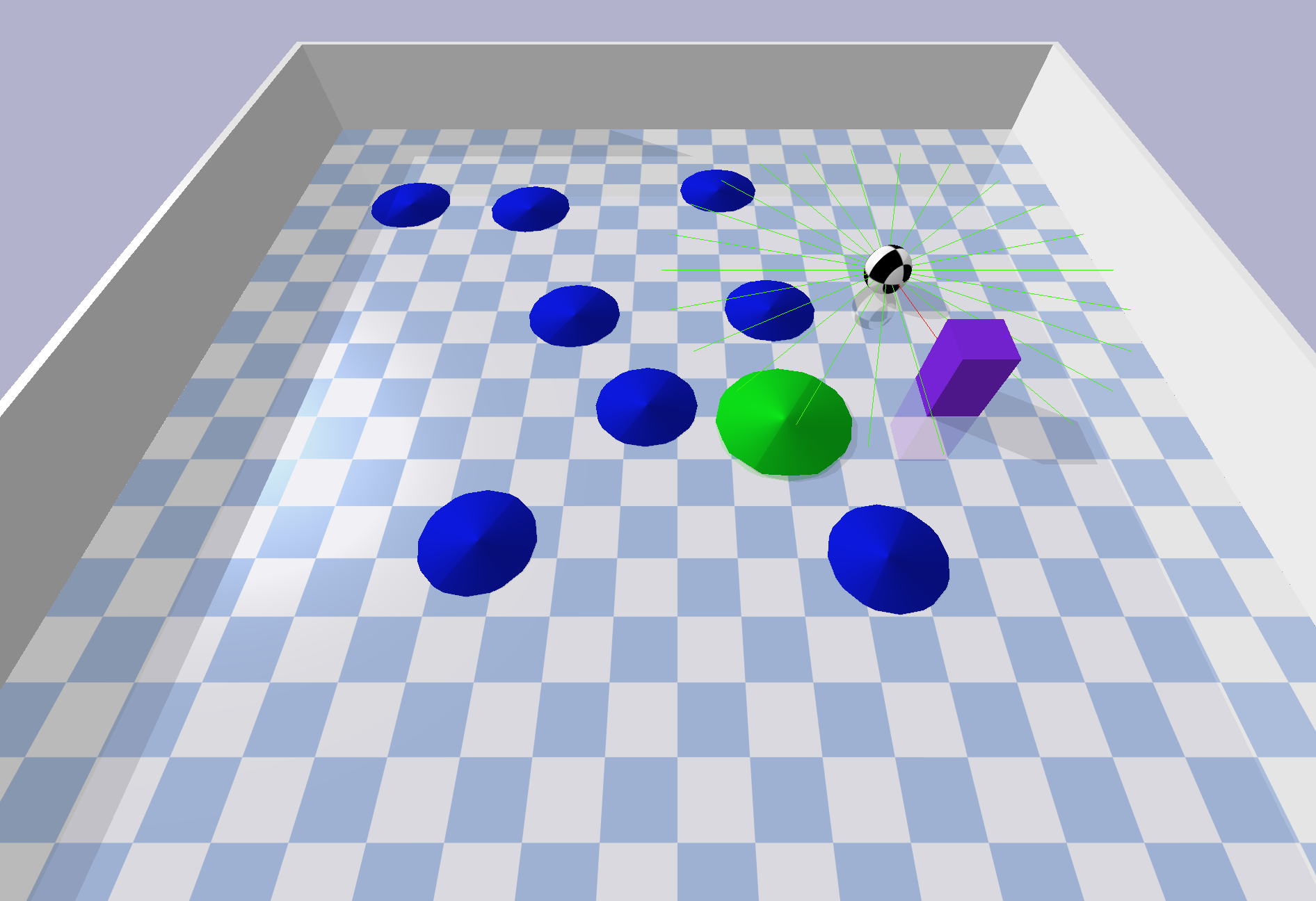}
        \label{fig:task_reach}
    }
    \subfloat[Grid]{
        \includegraphics[height=.12\textwidth, width=0.15\textwidth]{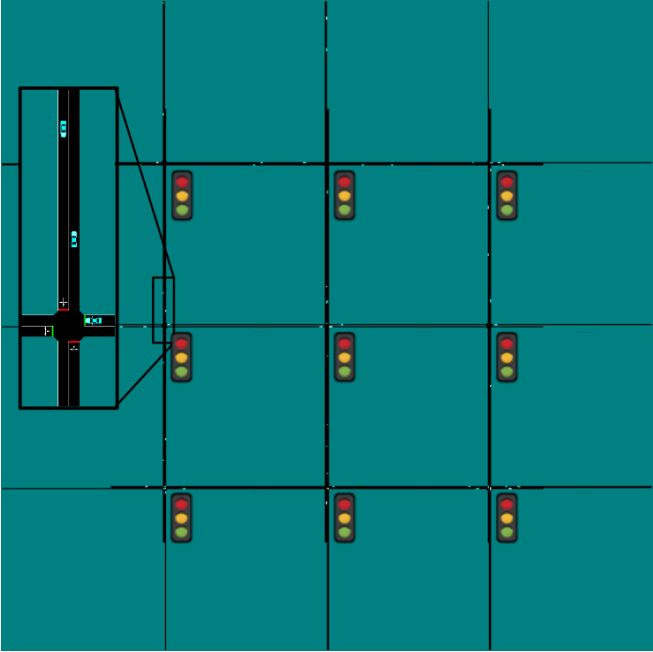}
        \label{fig:task_grid}
    }
    \subfloat[Bottleneck]{
        \includegraphics[height=.12\textwidth, width=0.15\textwidth]{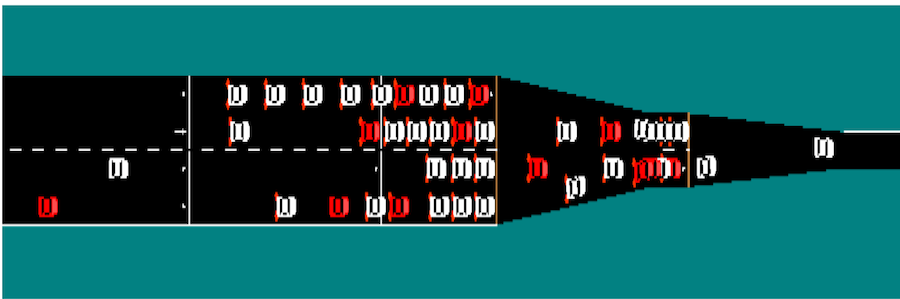}
        \label{fig:task_bottleneck}
    }
    \subfloat[Navigation]{
        \includegraphics[height=.12\textwidth, width=0.15\textwidth]{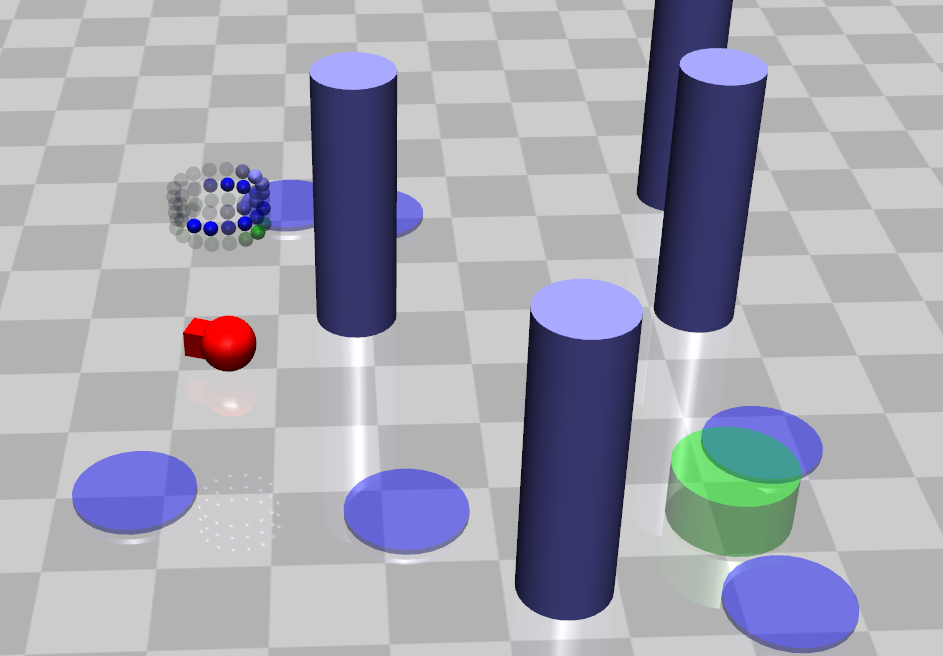}
        \label{fig:task_navigation}
    }
    \caption[]{\small The Humanoid, Circle, Reach, Grid, Bottleneck, and Navigation tasks. 
    See Appendix \ref{appendix:envs} for details.
    }
\label{fig:chapter2_env_overview}
\vspace*{-11pt}
\end{figure}

\noindent\textbf{Environments.}
For a comprehensive empirical evaluation, we selected eight scenarios from well-known safe RL benchmark environments - Safe MuJoCo \cite{zhang2020first} and Safety Gym \cite{ray2019benchmarking}, as well as MuJoCo environments.  These include: \texttt{Humanoid}, \texttt{PointCircle}, \texttt{AntCircle}, \texttt{PointReach}, \texttt{AntReach}, \texttt{Grid}, \texttt{Bottleneck}, and \texttt{Navigation}. See Figure \ref{fig:chapter2_env_overview} for an overview of the tasks and scenarios. Note that \texttt{Navigation} is a multi-constraint task and for the \texttt{Reach} environment, we set the reward as a function of the Euclidean distance between agent's position and goal position. In addition, we make it impossible for the agent to reach the goal before the end of the episode. For more information see Appendix \ref{appendix:envs}.

\noindent\textbf{Baselines.}
We compare our \texttt{e-COP} algorithm with the following baseline algorithms:  CPO \cite{achiam2017constrained}, PCPO \cite{yang2020projection}, FOCOPS \cite{zhang2020first}, PPO with Lagrangian relaxation \cite{schulman2017proximal, stooke2020responsive}, and penalty-based P3O \cite{zhang2022penalized}. Since the above state-of-the-art baseline algorithms are already well understood to outperform other algorithms such as PDO \cite{chow2017risk}, IPO \cite{liu2020ipo}, and CPPO-PID \cite{stooke2020responsive} in prior benchmarking studies, we do not compare against them. Moreover, since PPO does not originally incorporate constraints, for fair comparison, we introduce constraints using a Lagrangian relaxation (called PPO-L). In addition, for each algorithm, we report its performance with the discount factor that achieves the best performance. See Appendix \ref{appendix:chapter2_experimental_details} for more details.

\noindent\textbf{Evaluation Details and Protocol.}  For the Circle task, we use a  a point-mass with $S \subseteq \Rbb^{9}, A \subseteq \Rbb^{2}$ and for the Reach task, an ant robot  with $S \subseteq \Rbb^{16}, A \subseteq \Rbb^{8}$. The Grid task has $S \subseteq \Rbb^{56}, A \subseteq \Rbb^{4}$. We use two hidden layer neural networks to represent Gaussian policies for the tasks. For Circle and Reach, size is (32,32) for both layers, and for Grid and Navigation the layer sizes are (16,16) and (25,25). We set the step size $\delta$ to $10^{-4}$, and for each task, we conduct 5 independent runs of $K=500$ episodes each of horizon $H=200$. Since there are two objectives (rewards in the objective and costs in the constraints), we show the plots which maximize the reward objective while satisfying the cost constraint.

\subsection{Performance Analysis}

{
\renewcommand{\arraystretch}{1.1}
\begin{table*}
\centering
\fontsize{7}{9}\selectfont
\addtolength{\tabcolsep}{-0.65em}
\begin{adjustbox}{max width=\textwidth}
\begin{tabular}{cl|cccccccc}
\hline
 \multicolumn{2}{c|}{Task}  & \texttt{e-COP} & FOCOPS \cite{zhang2020first} & PPO-L \cite{ray2019benchmarking} & PCPO \cite{yang2020projection} & P3O \cite{zhang2022penalized} & CPO \cite{achiam2017constrained} & APPO \cite{Dai2023}  & IPO \cite{liu2020ipo} \\
\hline
\multirow{2}*{Humanoid} & R & \cellcolor{CellColLight}{1652.5 $\pm$ 13.4} & \cellcolor{CellCol}{\textbf{1734.1 $\pm$ 27.4}} & 1431.2 $\pm$ 25.2 & 1602.3 $\pm$ 10.1  &  1669.4 $\pm$ 13.7 & 1465.1 $\pm$ 55.3 & 1488.2 $\pm$ 29.3 & 1578.6 $\pm$ 25.2 \\
 &  C (20.0) & \cellcolor{CellColLight}{17.3 $\pm$ 0.3}  &  \cellcolor{CellCol}{19.7} $\pm$ 0.6 & 18.8 $\pm$ 1.5 & 16.3 $\pm$ 1.4 & 20.1 $\pm$ 3.3 & 18.5 $\pm$ 2.9 & 20.0 $\pm$ 1.3 & 19.1 $\pm$ 2.5 \\ 
 \hline
\multirow{2}*{PointCircle} & R & \cellcolor{CellCol}{\textbf{110.5 $\pm$ 9.3}} & 81.6 $\pm$ 8.4 & 57.2 $\pm$ 9.2 & 68.2 $\pm$ 9.1  &  89.1 $\pm$ 7.1 & 65.3 $\pm$ 5.3 & 91.2 $\pm$ 9.6 & 68.7 $\pm$ 15.2 \\
 &  C (10.0) & \cellcolor{CellCol}{9.8 $\pm$ 0.9}  &  10.0 $\pm$ 0.4 & 9.8 $\pm$ 0.5 & 9.9 $\pm$ 0.4 & 9.9 $\pm$ 0.3 & 9.5 $\pm$ 0.9 & 10.2 $\pm$ 0.6 & 9.3 $\pm$ 0.5 \\ 
\hline
\multirow{2}*{AntCircle} & R & \cellcolor{CellCol}{\textbf{198.6 $\pm$ 7.4}} & 161.9 $\pm$ 22.2  & 134.4 $\pm$ 10.3 & 168.3 $\pm$ 13.3 & 182.6 $\pm$ 18.7 & 127.1 $\pm$ 12.1 & 155.5 $\pm$ 19.4 & 149.3 $\pm$ 33.6 \\
 &  C (10.0) & \cellcolor{CellCol}{9.8 $\pm$ 0.6} & 9.9 $\pm$ 0.5  & 9.6 $\pm$ 1.6  & 9.5 $\pm$ 0.6 & 9.8 $\pm$ 0.2 & 10.1 $\pm$ 0.7 & 10.0 $\pm$ 0.5 & 9.5 $\pm$ 1.0 \\
\hline
\multirow{2}*{PointReach} & R & \cellcolor{CellCol}{\textbf{81.5 $\pm$ 10.2}} & 65.1 $\pm$ 9.6 & 46.1 $\pm$ 14.8 & 73.2 $\pm$ 7.4 & 76.3 $\pm$ 6.4 & 89.2 $\pm$ 8.1  & 74.3 $\pm$ 6.7 & 49.1 $\pm$ 10.6 \\
 &  C (25.0)  & \cellcolor{CellCol}{24.5 $\pm$ 6.1} & 24.8 $\pm$ 7.6 & 25.1 $\pm$ 6.1  & 24.9 $\pm$ 5.6 & 26.3 $\pm$ 6.9 & 33.3 $\pm$ 10.7 & 26.3 $\pm$ 8.1 & 24.7 $\pm$ 11.3 \\
 \hline
\multirow{2}*{AntReach} & R & \cellcolor{CellColLight}{70.8 $\pm$ 14.6} & 48.3 $\pm$ 5.6  & 54.2 $\pm$ 9.5 & 39.4 $\pm$ 5.3  & \cellcolor{CellCol}{\textbf{73.6 $\pm$ 5.1}}  & 102.3 $\pm$ 7.1 & 61.5 $\pm$ 10.4 & 45.2 $\pm$ 13.3 \\
 &  C (25.0) & \cellcolor{CellColLight}{24.2 $\pm$ 8.4} & 25.1 $\pm$ 11.9  & 21.9 $\pm$ 10.7  & 27.9 $\pm$ 12.2 & \cellcolor{CellCol}{24.8 $\pm$ 7.3} & 35.1 $\pm$ 10.9 & 24.5 $\pm$ 6.4 & 24.9 $\pm$ 9.2 \\
 \hline
 & R & \cellcolor{CellColLight}{258.1 $\pm$ 33.1} & 215.4 $\pm$ 45.6 & \cellcolor{CellCol}{\textbf{276.3 $\pm$ 57.9}}  & 226.5 $\pm$ 29.2 & 201.5 $\pm$ 39.2 & 178.1 $\pm$ 23.8 & 184.4 $\pm$ 21.5 & 229.4 $\pm$ 32.8 \\
\multirow{-2}*{Grid} & C (75.0) &  \cellcolor{CellColLight}{71.3 $\pm$ 26.9} & 76.6 $\pm$ 29.8  & \cellcolor{CellCol}{71.8 $\pm$ 25.1} & 72.6 $\pm$ 16.5 & 79.3 $\pm$ 19.3 & 69.3 $\pm$ 19.8 & 79.5 $\pm$ 35.8 & 74.2 $\pm$ 24.6 \\ 
 \hline
 \multirow{2}*{Bottleneck} & R & \cellcolor{CellCol}{\textbf{345.1 $\pm$ 52.6}} & 251.3 $\pm$ 59.1 & 298.3 $\pm$ 71.2  & 264.2 $\pm$ 43.8 & 291.1 $\pm$ 26.7 & 388.1 $\pm$ 36.6 & 220.1 $\pm$ 30.1 & 279.3 $\pm$ 43.8 \\
 &  C (50.0) &  \cellcolor{CellCol}{49.7 $\pm$ 15.1} & 46.6 $\pm$ 19.8  & 41.4 $\pm$ 17.6 & 49.8 $\pm$ 10.5 & 45.3 $\pm$ 8.2 & 54.3 $\pm$ 13.5 & 47.4 $\pm$ 12.3 & 48.2 $\pm$ 14.6 \\
\hline
\multirow{3}*{Navigation} & R & \cellcolor{CellCol}{\textbf{217.6 $\pm$ 11.5}} & & 175.1 $\pm$ 3.7 &  &  153.5 $\pm$ 25.2  & & 135.7 $\pm$ 19.2 & 164.1 $\pm$ 12.8 \\
 &  C1 (10.0) &  \cellcolor{CellCol}{9.6 $\pm$ 1.5} & n/a &  9.9 $\pm$ 1.9 & n/a & 9.9 $\pm$ 1.7 & n/a &  9.9 $\pm$ 2.1 & 10.0 $\pm$ 0.5 \\
  &  C2 (25.0) & \cellcolor{CellCol}{23.7 $\pm$ 4.1}  & & 22.3 $\pm$ 2.1   & & 24.5 $\pm$ 4.1 & & 23.9 $\pm$ 3.8 & 24.6 $\pm$ 3.1 \\
\hline
\end{tabular}
\end{adjustbox}
\caption{\small Mean performance with normal 95\% confidence interval over 5 independent runs on some tasks. 
}
\label{table:rewards_costs_comparison}
\end{table*}
}

\begin{figure*}[t]
\captionsetup[subfigure]{justification=centering}
    \textbf{\small Episodic Rewards:} \vspace{0.2cm} \newline 
    {
        \includegraphics[width=0.23\textwidth]{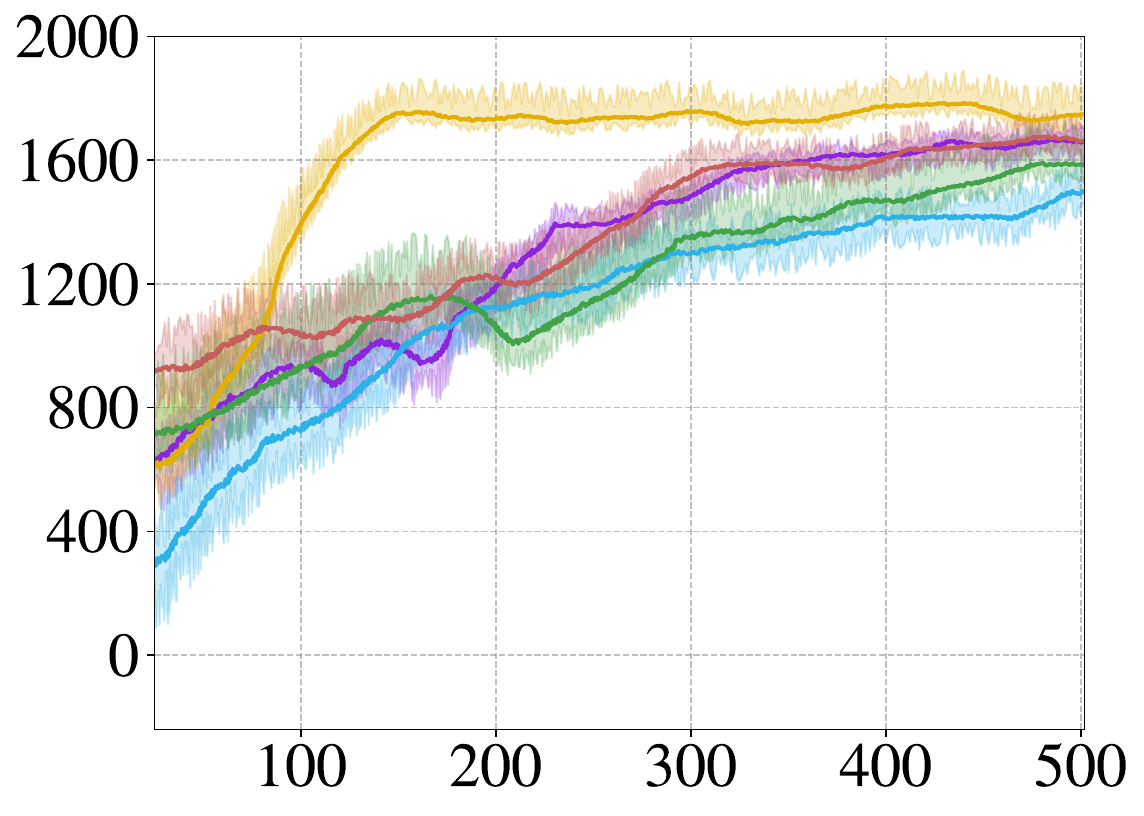}
        \label{fig:humanoid_reward}
    }
    {
        \includegraphics[width=0.23\textwidth]{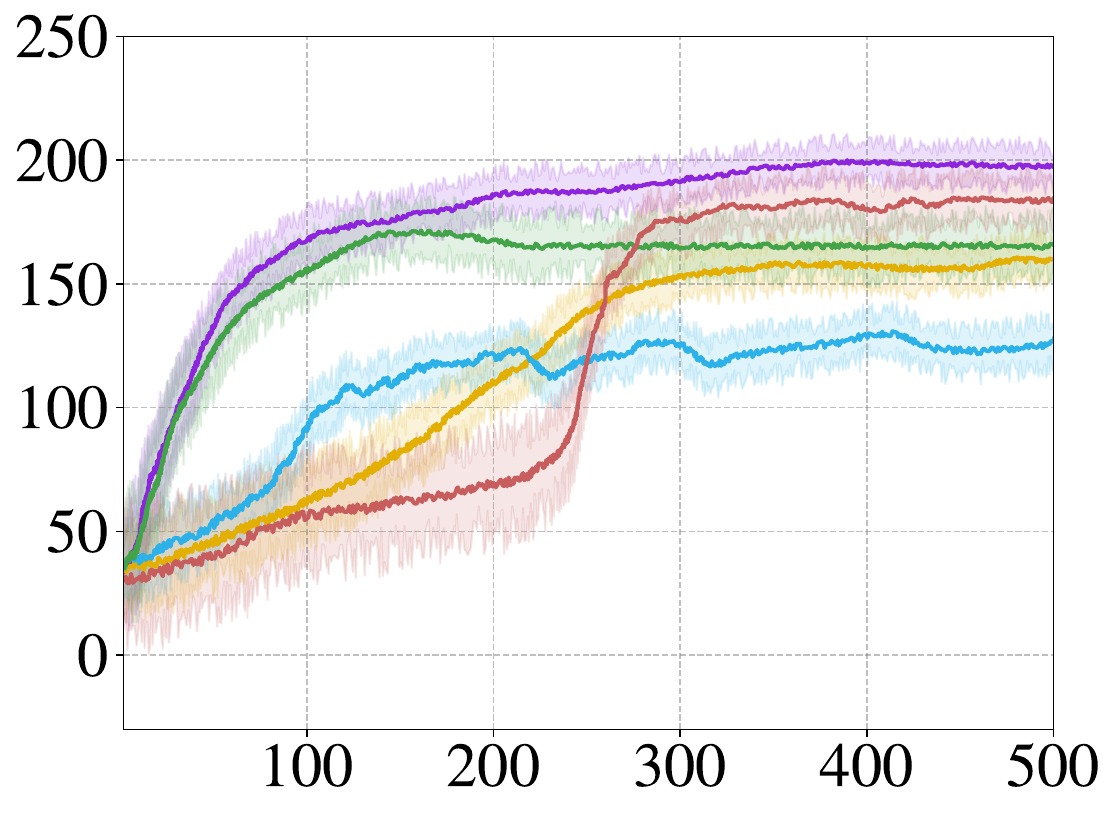}
        \label{fig:antcircle_reward}
    }
    {
        \includegraphics[width=0.23\textwidth]{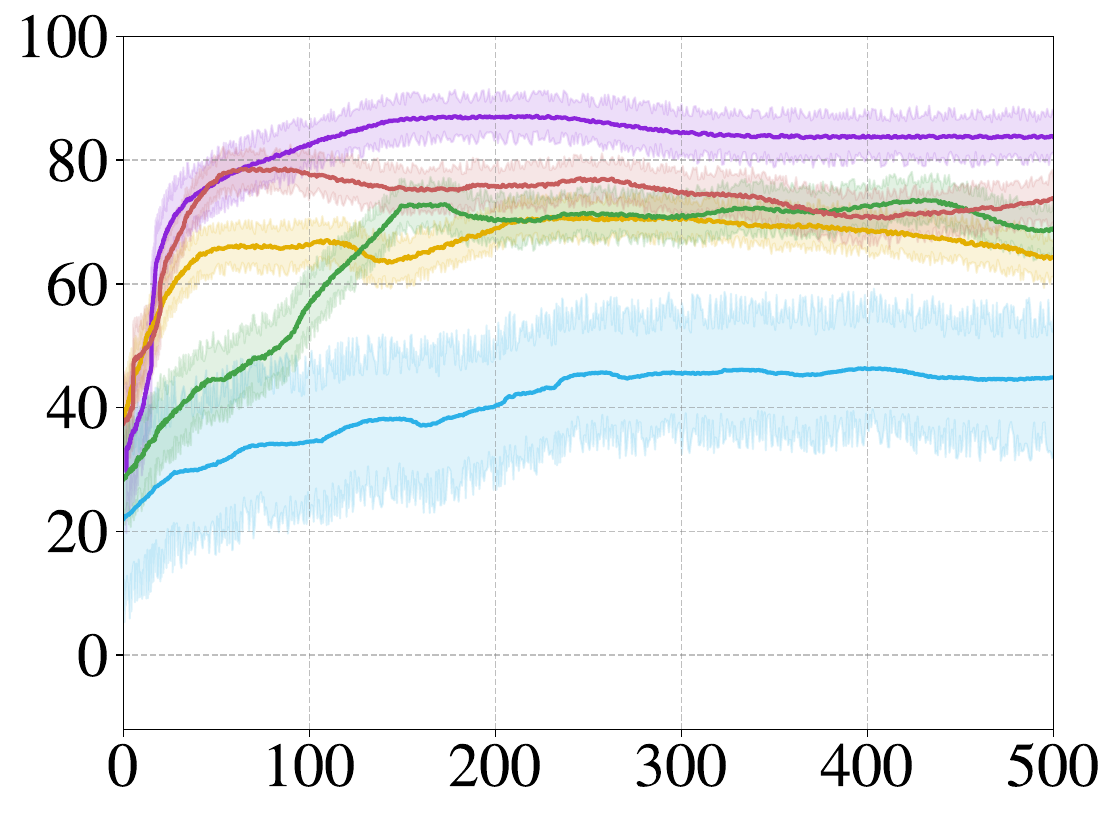}
        \label{fig:pointreach_reward}
    }
    {
        \includegraphics[width=0.23\textwidth]{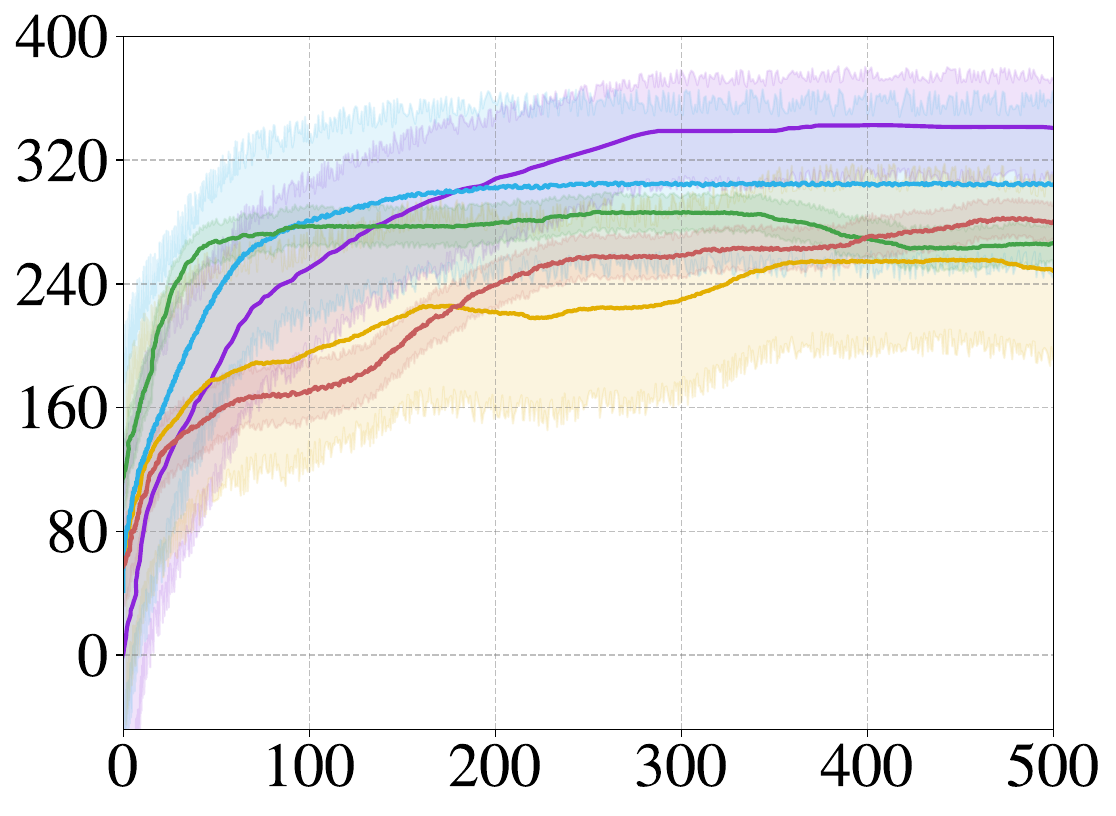}
        \label{fig:bottleneck_reward}
    }
    \newline
    \rule{\linewidth}{0.5pt}
    \hspace{2cm} \textbf{\small Constraint Costs:} 
    \hspace{2.75cm} \includegraphics[width=0.63\textwidth]{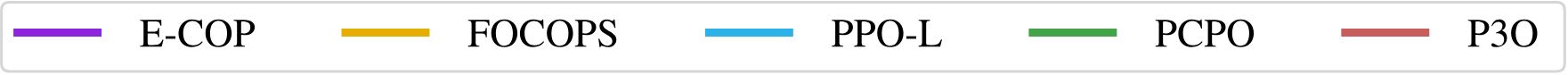} \newline
    \subfloat[Humanoid]{
        \includegraphics[width=0.23\textwidth]{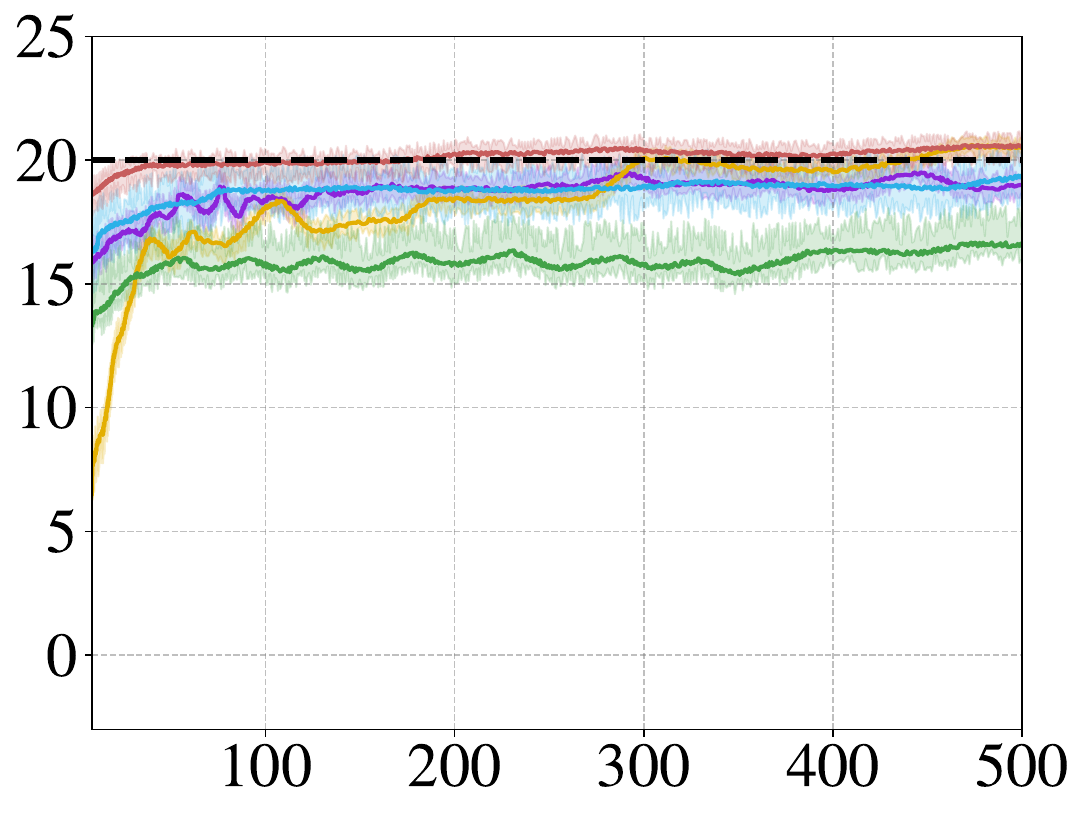}
        \label{fig:humanoid_cost}
    }
    \subfloat[Ant Circle]{
        \includegraphics[width=0.23\textwidth]{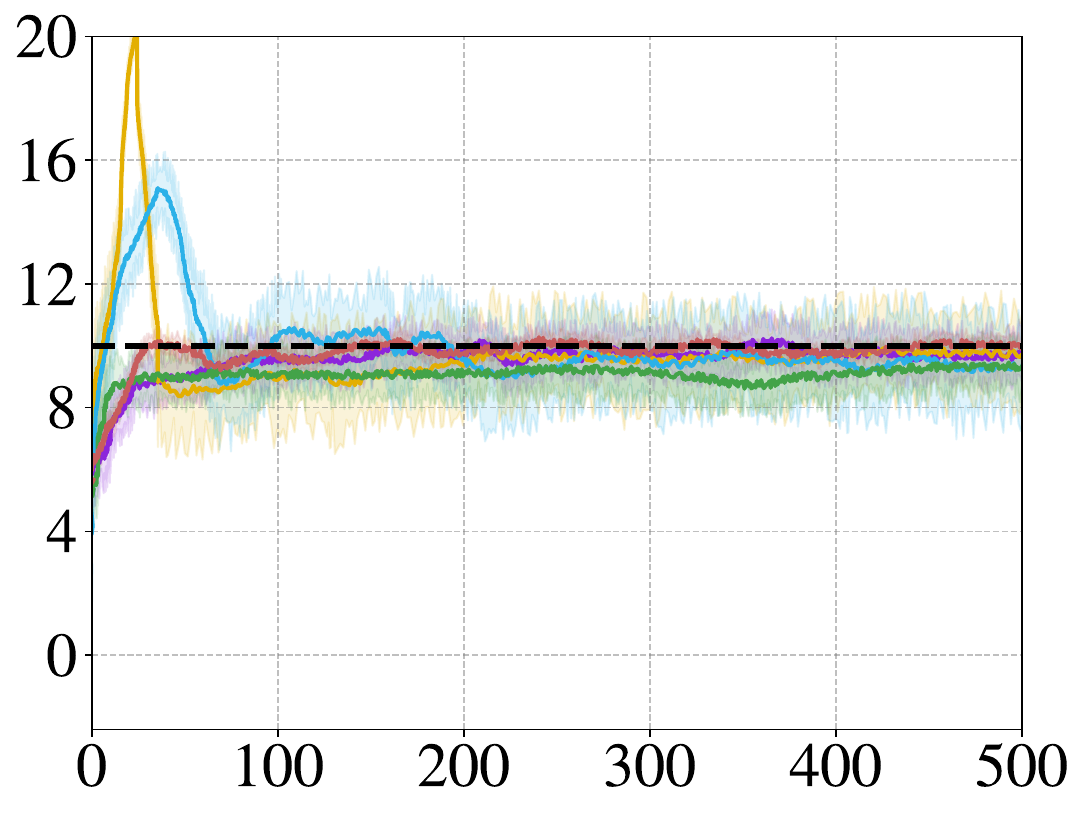}
        \label{fig:antcircle_cost}
    }
    \subfloat[Point Reach]{
        \includegraphics[width=0.24\textwidth]{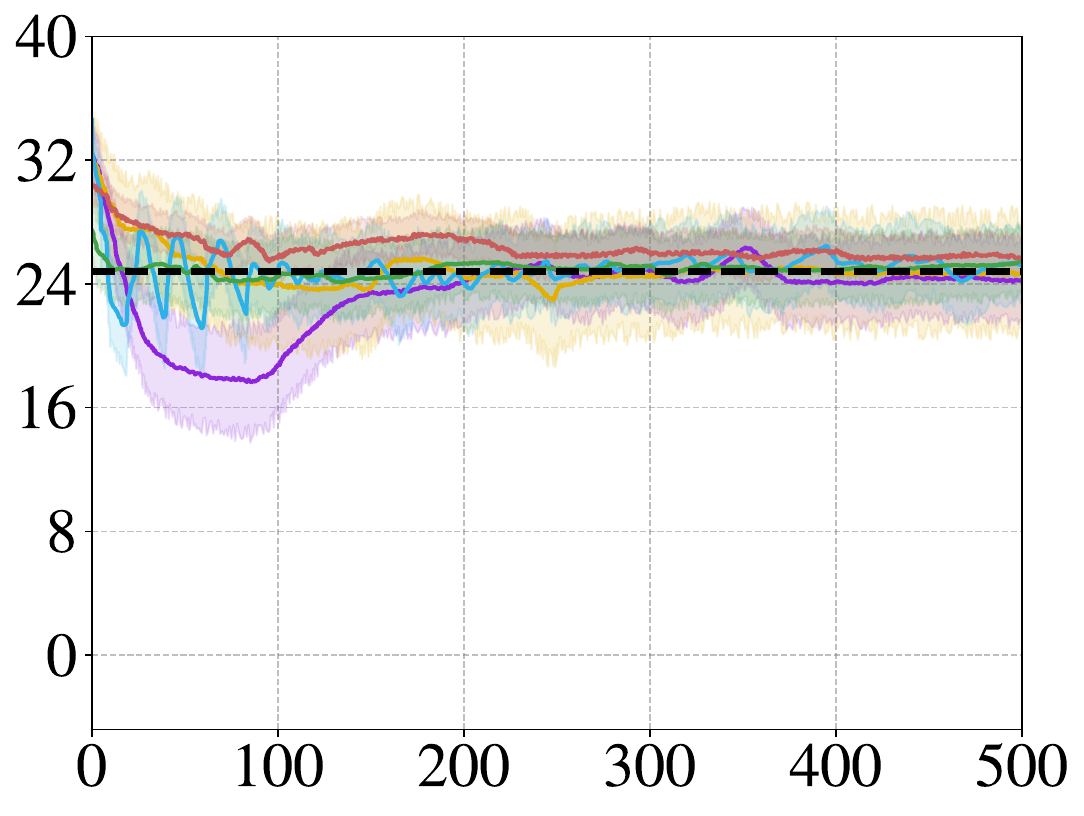}
        \label{fig:pointreach_cost}
    }
    \subfloat[Bottleneck]{
        \includegraphics[width=0.23\textwidth]{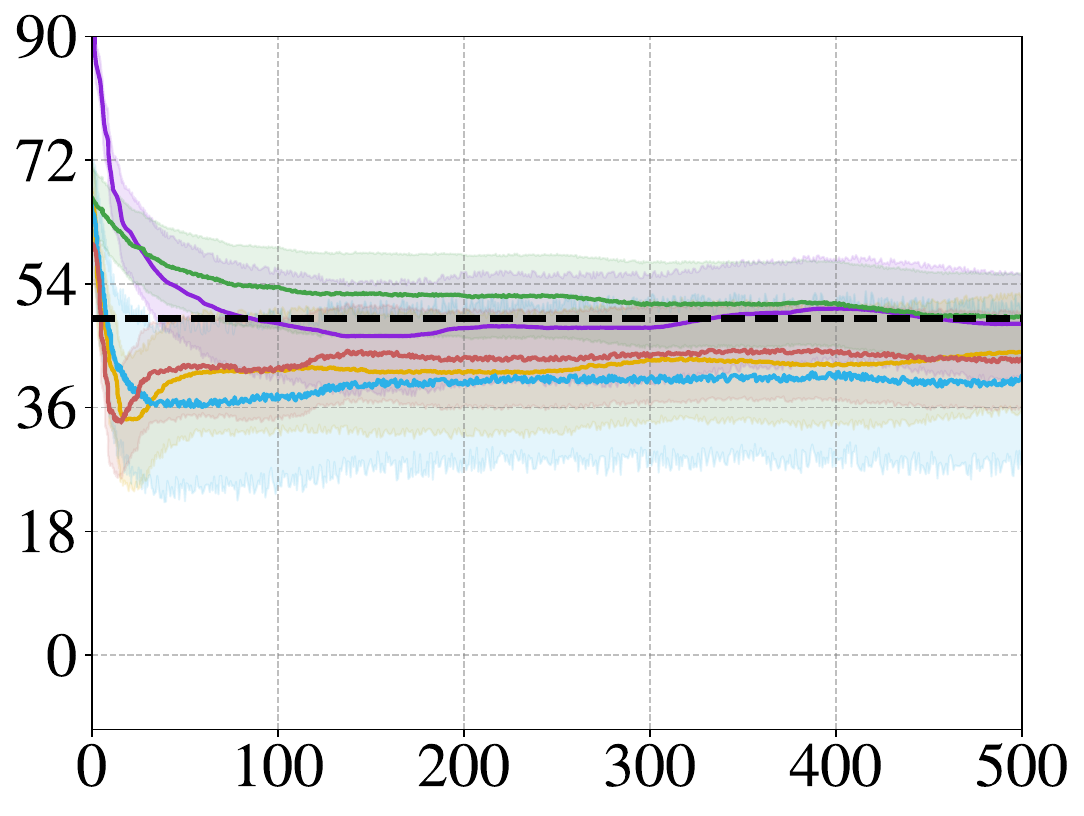}
        \label{fig:bottleneck_cost}
    }
    \caption{The cumulative episodic reward and constraint cost function values vs episode learning curves for some algorithm-task pairs. Solid lines in each figure are the means, while the shaded area represents 1 standard deviation, all over 5 runs. The dashed line in constraint plots is the constraint threshold.}
    \label{fig:chapter2_rewards_costs_comparison}
\end{figure*}

\begin{figure*}[ht]
\captionsetup[subfigure]{justification=centering}
  \centering
  \includegraphics[height=0.03\linewidth, width=0.6\linewidth] {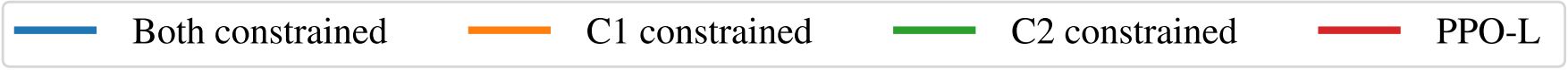} \vspace*{-\baselineskip}
  
    \subfloat[Rewards]{
        \includegraphics[height=0.125\linewidth, width=0.24\linewidth]{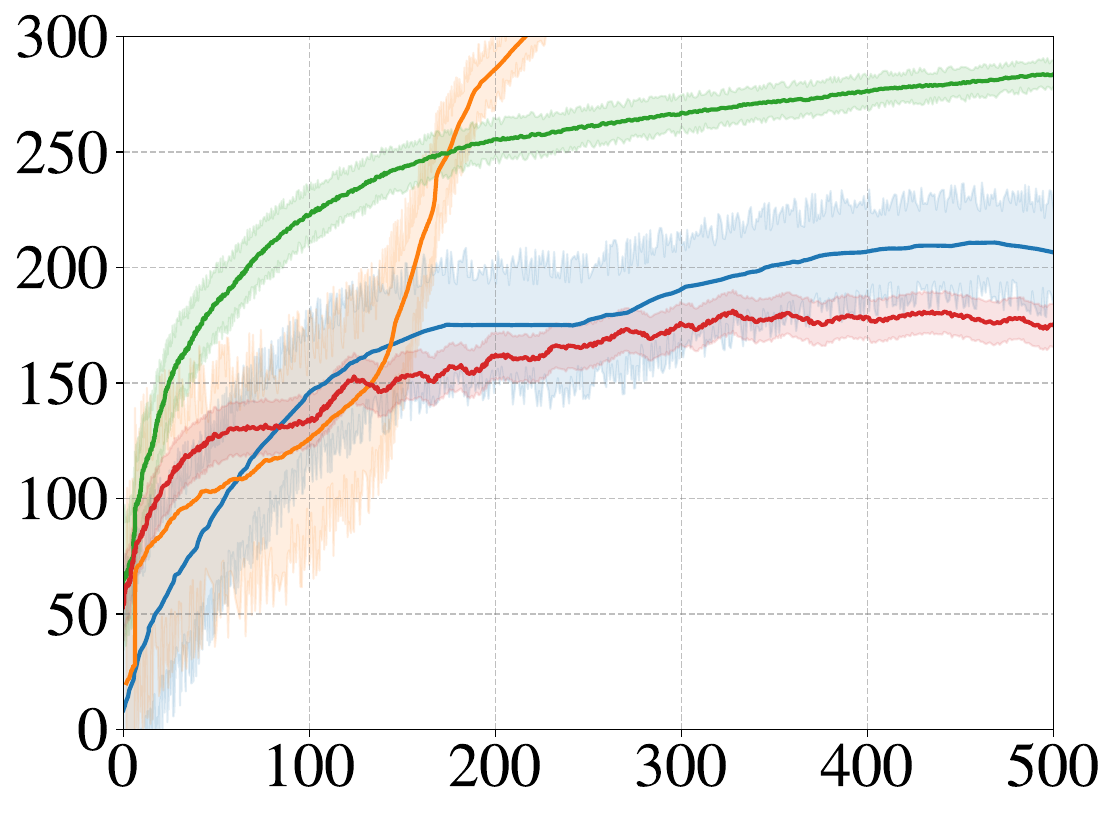}
        \label{fig:navigation_reward}
    }
    \subfloat[Cost1 (hazards)]{
        \includegraphics[height=0.125\linewidth, width=0.24\linewidth]{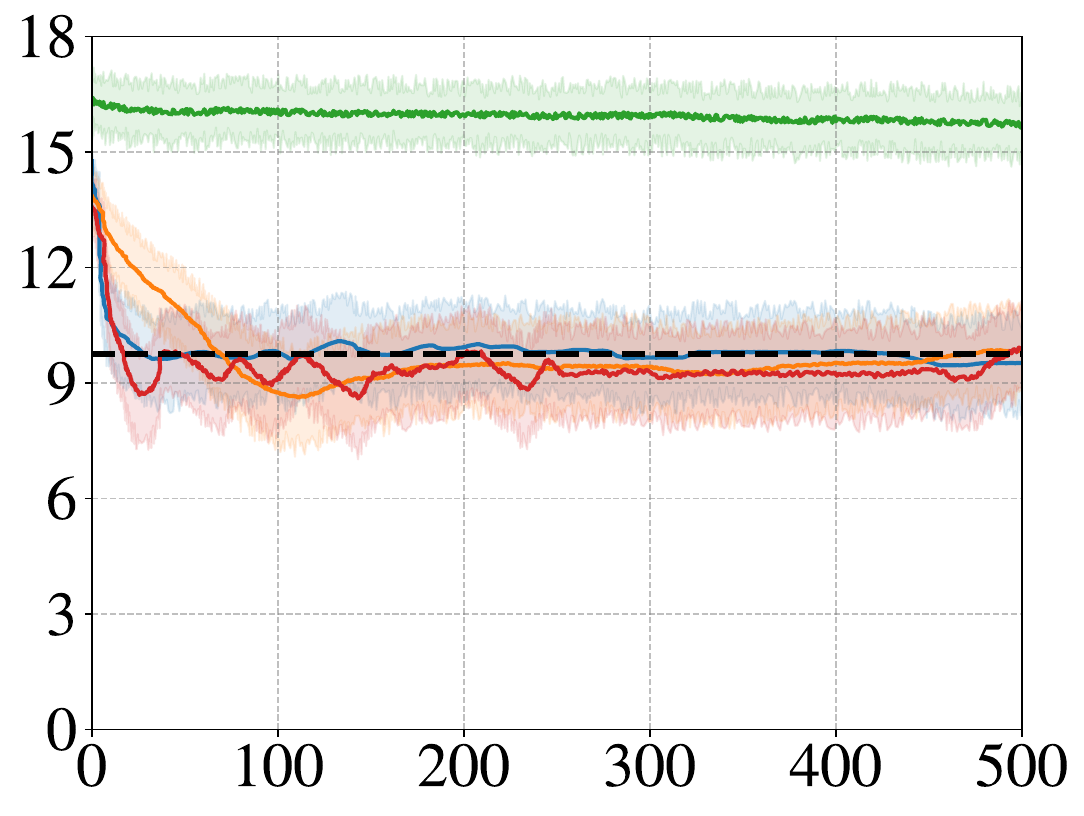}
        \label{fig:navigation_cost1}
    }
    \subfloat[Cost2 (pillars)]{
        \includegraphics[height=0.125\linewidth, width=0.24\linewidth]{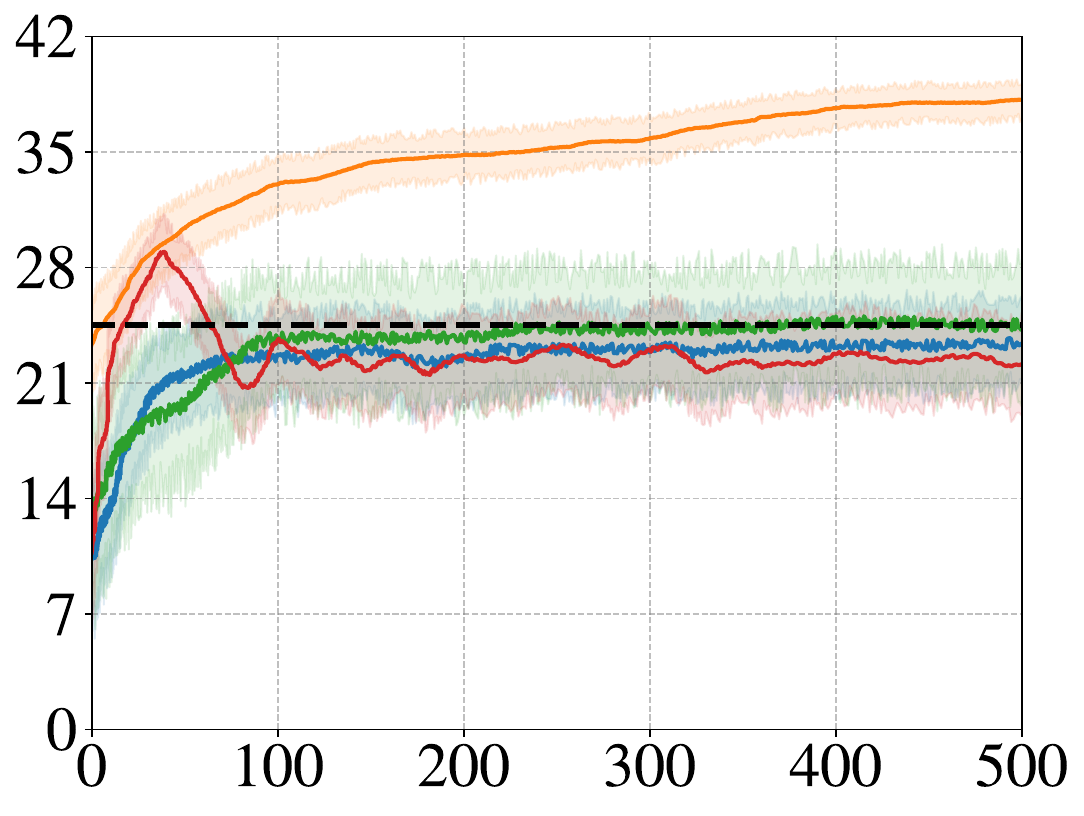}
        \label{fig:navigation_cost2}
    }
  \caption{\small \texttt{Navigation} environment with multiple  constraints: Episodic Rewards (left), Cost1 (center, for hazards) and Cost2 (right, for pillars) of \texttt{e-COP} . The dashed line in the cost plots is the cost threshold (10 for Cost1 and 25 for Cost2). C1/C2 constrained means only taking Cost1/Cost2 into the \texttt{e-COP} loss function and ignoring the other one.}
  \label{fig:navigation_rewards_costs}
  \vspace*{-11pt}
\end{figure*}

Table \ref{table:rewards_costs_comparison} lists the numerical performance of all tested algorithms in seven single constraint scenarios, and one multiple constraint scenario. We find that overall, the \texttt{e-COP} algorithm in most cases outperforms (green) all other baseline algorithms in finding the optimal policy while satisfying the constraints, and in other cases comes a close second (light green). 


From Figure \ref{fig:chapter2_rewards_costs_comparison}, we can see how the \texttt{e-COP} algorithm is able to improve the reward objective over the baselines while having approximate constraint satisfaction. We also see that updates of \texttt{e-COP} are faster and smoother than other baselines due to the added damping penalty, which ensures smoother convergence with only a few constraint-violating behaviors during training. In particular, \texttt{e-COP} is the \emph{only} algorithm that best learns almost-constraint-satisfying maximum reward policies across \emph{all} tasks: in simple \texttt{Humanoid} and \texttt{Circle} environments, \texttt{e-COP} is able to almost exactly track the cost constraint values to within the given threshold. However, for the high dimensional \texttt{Grid} environment we have more constraint violations due to complexity of the policy behavior, leading to higher variance in episodic rewards as compared to other environments. Regardless, overall in these environments, \texttt{e-COP} still outperforms \emph{all} other baselines with the least episodic constraint violation. For the multiple constraint \texttt{Navigation} environment, see Figure \ref{fig:navigation_rewards_costs}.



\subsection{Secondary Evaluation}

In this section, we take a deeper dive into the empirical performance of \texttt{e-COP}. We discuss its dependence on various factors, and try to verify its merits.

\begin{figure*}[ht]
\captionsetup[subfigure]{justification=centering}
  \centering
  \includegraphics[width=0.22\linewidth]{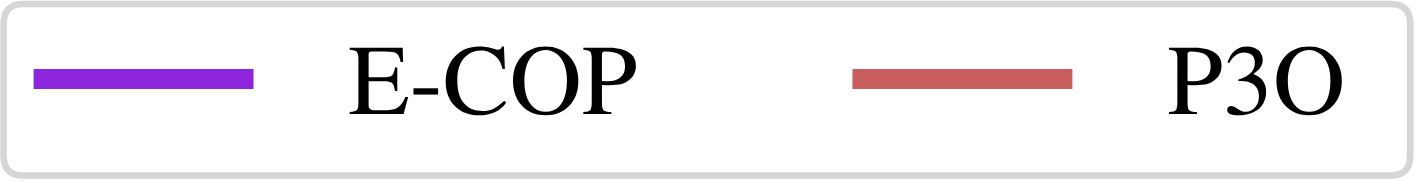}  

    \subfloat[Ant Circle Rewards]{
        \includegraphics[height=0.125\linewidth, width=0.23\linewidth]{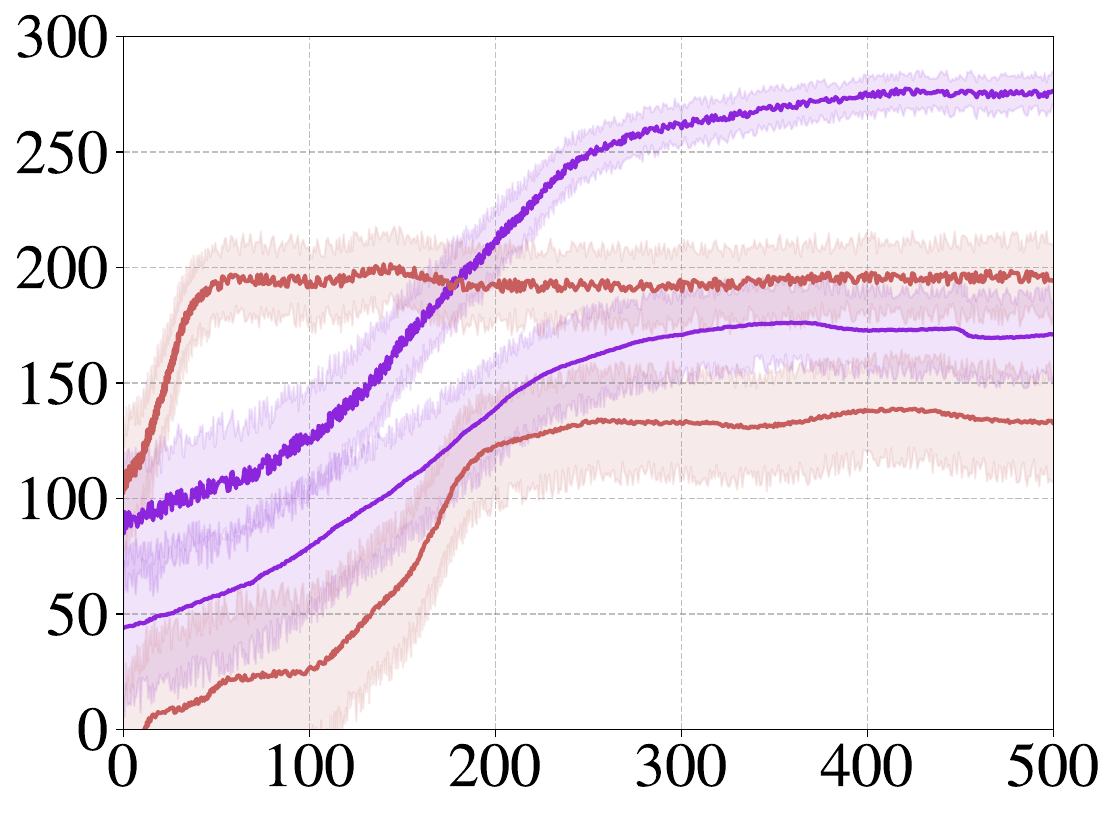}
        \label{fig:general_antcircle_reward}
    }
    \subfloat[Ant Circle Costs]{
        \includegraphics[height=0.125\linewidth, width=0.23\linewidth]{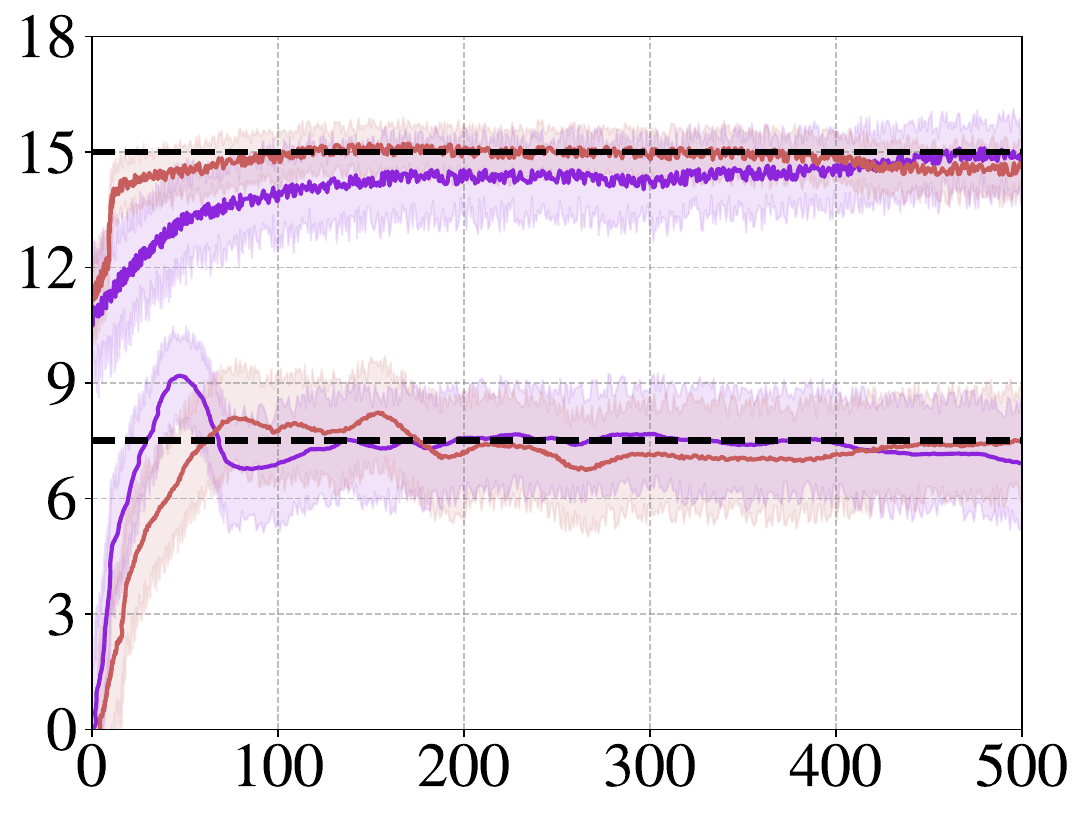}
        \label{fig:general_antcircle_cost}
    }
    \subfloat[Point Reach Rewards]{
        \includegraphics[height=0.125\linewidth, width=0.23\linewidth]{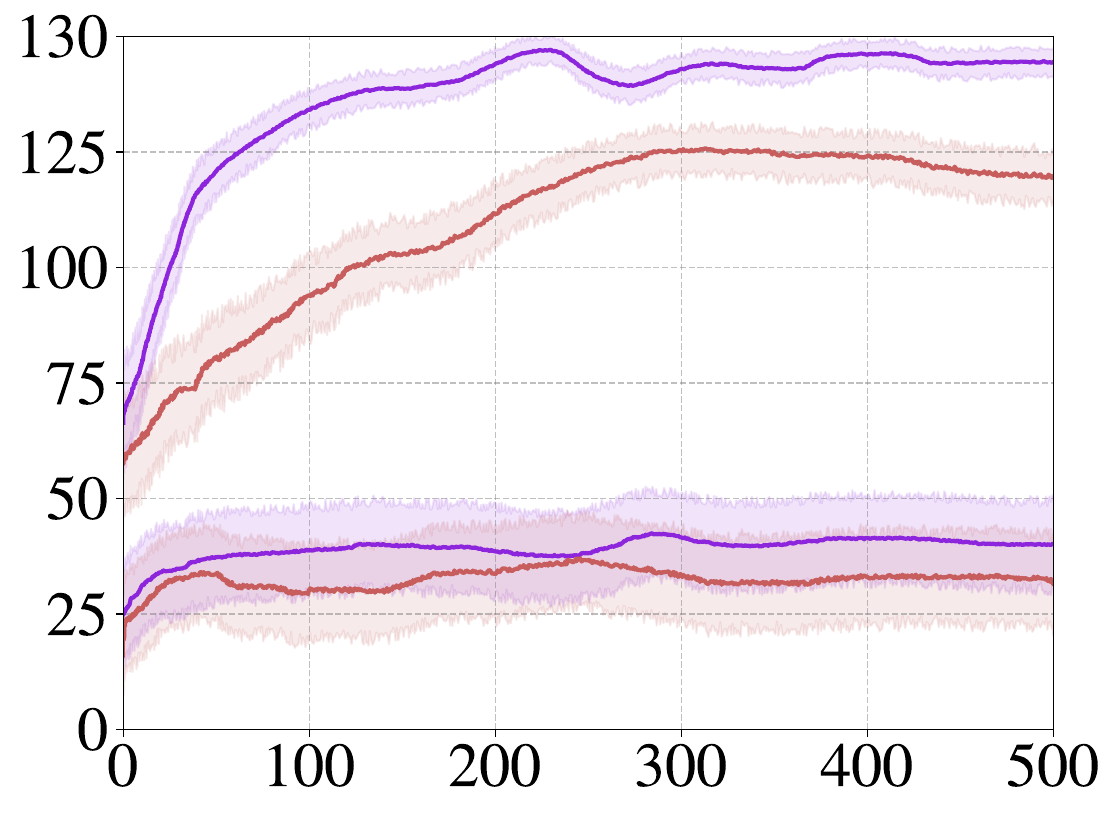}
        \label{fig:general_pointreach_reward}
    }
    \subfloat[Point Reach Costs]{
        \includegraphics[height=0.125\linewidth, width=0.23\linewidth]{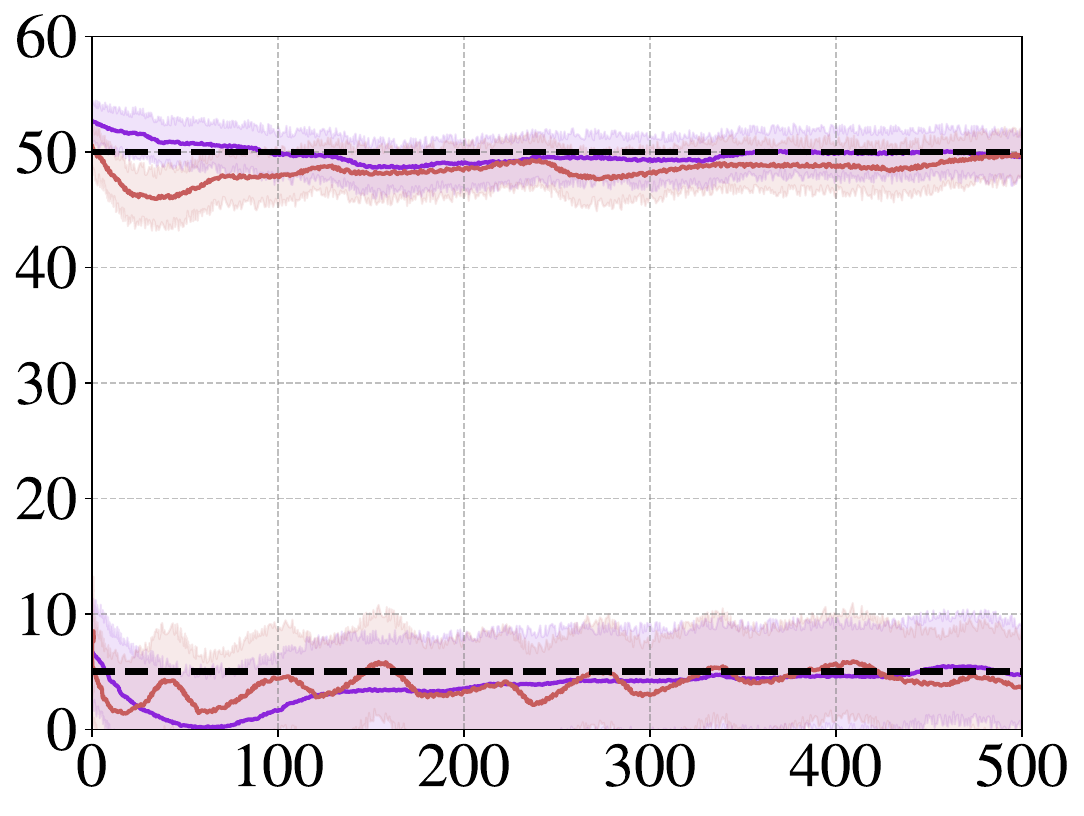}
        \label{fig:general_pointreach_cost}
    }
  \caption{\small Cumulative episodic rewards and costs of baselines in two environments with two constraint thresholds.}
  \label{fig:general_rewards_costs_two}
\end{figure*}

\noindent\textbf{Generalizability.} From the discussion above, it's clear that \texttt{e-COP} demonstrates accurate safety satisfaction across tasks of varying difficulty levels. From Figure \ref{fig:general_rewards_costs_two},  we further see that \texttt{e-COP} satisfies the constraints in all cases and precisely converges to the specified cost limit. Furthermore, the fluctuation observed in the baseline Lagrangian-based algorithms is shown not to be tied to a specific cost limit.


We also conducted a set of experiments wherein we study how \texttt{e-COP} effectively adapts to different cost thresholds. For this, we use the hyperparameters of a pre-trained \texttt{e-COP} agent, which is trained with a particular cost threshold in an environment, for learning on different cost thresholds within the same environment. Figure \ref{fig:general_rewards_costs_three} in Appendix \ref{app:4.2} illustrates the training curves of these pre-trained agents, and we see that while \texttt{e-COP} can generalize well across different cost thresholds, other baseline algorithms may require further tuning to accommodate different constraint thresholds. 

{\renewcommand{\arraystretch}{1.25}
\begin{table*}[!ht]
\centering
\fontsize{7}{8}\selectfont
\vspace{1cm}
\begin{tabular}{cl|cccc|c}
\hline
 \multirow{2}*{Task} &  & \multicolumn{3}{c}{\hspace{2cm} \texttt{e-COP}}   & & \multirow{2}*{P3O \cite{zhang2022penalized}} \\ 
 \cline{3-6}
  &  & $\beta=5$, fixed & $\beta=5$, adaptive & $\beta=10$, fixed & $\beta=10$, adaptive &  \\
\hline
\multirow{2}*{PointCircle} & R & 150.5 $\pm$ 11.1 & \cellcolor{CellCol}{\textbf{168.6 $\pm$ 14.3}} & 145.2 $\pm$ 12.2 & 165.3 $\pm$ 11.4  &  162.4 $\pm$ 14.7 \\
 &  C (20.0) & 17.3 $\pm$ 1.3  &  \cellcolor{CellCol}{19.7 $\pm$ 0.6} & 18.8 $\pm$ 1.5 & 18.3 $\pm$ 1.4 & 19.1 $\pm$ 3.3 \\ 
 \hline
\multirow{2}*{AntReach} & R & 48.2 $\pm$ 3.5 & 58.6 $\pm$ 5.1 & 53.2 $\pm$ 5.3 & \cellcolor{CellCol}{\textbf{65.2 $\pm$ 8.1}}  &  61.1 $\pm$ 5.6 \\
 &  C (20.0) & 19.8 $\pm$ 5.9  &  20.0 $\pm$ 4.4 & 20.6 $\pm$ 4.5 & \cellcolor{CellCol}{19.2 $\pm$ 6.2} & 18.9 $\pm$ 7.3 \\ 
\hline
\end{tabular}
\caption{\small Performance of \texttt{e-COP} for different $\beta$ settings on two tasks. Values are given with normal 95\% confidence interval over 5 independent runs. }
\label{table:ecop-sensitivity}
\end{table*}
}

\noindent\textbf{Sensitivity.}
The effectiveness and performance of \texttt{e-COP} would not be justified if it was not robust to the damping hyperparameter $\beta$, which varies across tasks depending on the values of $\Ccal(\cdot)$ and $\zeta_{k}$. Since this damping penalty enables \texttt{e-COP} to have stable continuous cost control, we update it adaptively as described in Algorithm \ref{alg:practical-ECOP}. As seen in Table \ref{table:ecop-sensitivity}, damping penalty indeed stabilizes the training process and helps in converging to an optimal safe policy. 



\clearpage
\newpage

\section{Appendix}
\label{ecop:appendix}

\subsection{Proofs}

\episodicpolicydifference*
\begin{proof}
\label{proof:episodicpolicydifference}

Let us consider $s_{1} \sim \mu$ and categorize the value function difference between the two policies. Also define $\Pbb^{\pib}_{h}(s \given s_{1}) = \sum_{a \in A} \Pbb^{\pib}_{h}(s,a \given s_{1})$, where the term $\Pbb^{\pib}_{h}(s,a \given s_{1})$ is the probability of reaching $(s,a)$ at time step $h$ following $\pib$ and starting from $s_{1}.$

\begin{align*}
    V^{\pib}_{1}(s_{1}) - V^{\pib'}_{1}(s_{1})  
    &= \E{a_{1},s_2}\big[ r(s_{1},a_{1}) + V^{\pib}_{2}(s_2)|s_1\big] + \E{s_2}\big[ V^{\pib'}_{2}(s_2) - V^{\pib'}_{2}(s_2)|s_1\big] - V^{\pib'}_{1}(s_{1}) \\
    &= \E{s_{2}}\big[ V_{2}^{\pib}(s_{2}) - V_{2}^{\pib'}(s_{2}) |s_1 \big]  + \E{a_{1},s_2}\big[ r(s_{1},a_{1}) + V^{\pib'}_{2}(s_2) - V^{\pib'}_{1}(s_{1}) |s_1\big] \\
    &=  \E{s_{2}}\big[ V_{2}^{\pib}(s_{2}) - V_{2}^{\pib'}(s_{2}) |s_1 \big] + \E{a_{1}}\big[ Q^{\pib'}_{1}(s_{1}, a_{1}) - V^{\pib'}_{1}(s_{1})|s_1 \big] \\
    &=  \E{s_{2}}\big[ V_{2}^{\pib}(s_{2}) - V_{2}^{\pib'}(s_{2}) |s_1 \big] + \E{a_{1}} \big[  A^{\pib'}_{1}(s_{1}, a_{1}) |s_1 \big],
\end{align*}

where $a_1 \sim \pib_{1}(\cdot|s_1)$, $s_2 \sim P(\cdot | s_1, \pib_1(s_1))$ and $s_1 \sim \mu$, the initial state distribution.

Now recursively apply the same procedure to the term $V_{h}^{\pib}(s_{h}) - V_{h}^{\pib'}(s_{h}) \forAll h \in \{2, \dots, H\}$ to obtain the following:
\begin{align*}
    V^{\pib}_{1}(s) - V^{\pib'}_{1}(s) = \sum_{h=1}^{H} \E{s_{h}, a_{h} \sim \Pbb^{\pib}_{h}(\cdot, \cdot \given s)} \big[  A^{\pib'}_{h}(s_{h}, a_{h}) | s \big]
\end{align*}
Now we know that $J(\pib) = \E{s \sim \mu}[ V^{\pib}_{1}(s)]$, this means that we combine this with the above to obtain the final result.
\end{proof}

\dampedintermediateproblem*
\begin{proof}
\label{proof:damped_intermediate_problem}

As in Equation \eqref{eq:p3po_objective_fn}, we have the following equivalent problem:

\resizebox{0.95\linewidth}{!}{$
\begin{aligned}
\pi_{k,t}^{\star} & =  \mathop{\arg\min}_{\pi_{k,t}} \sum_{h=t}^{H} \mathop{\mathbb{E}}_{\substack{s\sim \rho_{\pi_{k,h}} \\a\sim \pi_{k-1,h}}} \big[-{\rho(\theta_{h})}A^{\pib_{k-1}}_{h}(s,a) \big] + \sum_{i}^{m} \lambda_{t,i} \max \bigg\{0, \; \sum_{h=t}^{H} \mathop{\mathbb{E}}_{\substack{s\sim \rho_{\pi_{k,h}} \\a\sim \pi_{k-1,h}}} \big  [{\rho(\theta_{h})}A_{C_i, h}^{\pib_{k-1}} (s,a) \big ] + (J_{C_i}(\pib_{k-1})-d_i) \bigg\}.  
\end{aligned}
$}

Letting $\Psi_{C_{i}, t}(\pib_{k-1}, \pib_{k}) := \sum_{h=t}^{H} \mathop{{\E{\substack{s \sim \rho_{\pi_{k,h}}\\ a\sim \pi_{k-1,h}}}}} \big  [{\rho(\theta_{h})}A_{C_i, h}^{\pib_{k-1}} (s,a) \big ] + (J_{C_i}(\pib_{k-1})-d_i)$, and introducing slack variables $x_{t,i} \geq 0$ and defining $w_{t,i}(\pib_{k}) := \Psi_{C_{i}, t}(\pib_{k-1}, \pib_{k}) + x_{t,i} = 0$, we get the quadratic damped problem same as Equation \eqref{eq:e-COP_initial_fn} below.

\begin{equation}
\begin{aligned}
(\pi_{k,t}^{\star}, \bm{\lambda}_{t}^{\star}, \bm{x}_{t}^{\star}) &= \max_{\bm{\lambda} \geq 0} \min_{\pi_{k,t}, \bm{x}} \Lcal_{t}(\pib_{k}, \bm{\lambda}, \bm{x}, \beta) \\
&=  \max_{\bm{\lambda} \geq 0} \min_{\pi_{k,t}, \bm{x}} \sum_{h=t}^{H} \E{s\sim \rho_{\pi_{k,h}} \\a\sim \pi_{k-1,h}} \big[-{\rho(\theta_{h})}A^{\pib_{k-1}}_{h}(s,a) \big] + \sum_{i}^{m} \lambda_{t,i} w_{t,i}(\pib_{k}) + \frac{\beta}{2} \sum_{i}^{m} w_{t,i}^{2}(\pib_{k})
\end{aligned}
\label{appendix:eq:maxminquadratic}
\end{equation}

Like the Lagrangian method, we can alternately update $\pi, \bm{\lambda}$, and $\bm{x}$ to find the optimal triplet. Consider updating $\pi$ and $\bm{x}$ by minimizing $\Lcal_{t}(\pi, \bm{\lambda}, \bm{x}, \beta)$ at any iteration:

\resizebox{\linewidth}{!}{$
\begin{aligned}
\left(\pi_{k,t}^{\star}, \bm{x}_{t}^{\star} \right)=\argmin_{\pi_{k,t}} \min_{x_{i}>0} \sum_{h=t}^{H} \E{s\sim \rho_{\pi_{k,h}} \\a\sim \pi_{k-1,h}} \big[-{\rho(\theta_{h})}A^{\pib_{k-1}}_{h}(s,a) \big] + \sum_{i=1}^{m} \lambda_{t,i} \big(  \Psi_{C_{i}, t}(\pib_{k-1}, \pib_{k}) + x_{t,i} \big) + \frac{\beta}{2} \sum_{i=1}^{m} \big(  \Psi_{C_{i}, t}(\pib_{k-1}, \pib_{k}) + x_{t,i} \big)^{2}
\end{aligned}
$}

The inner optimization problem with respect to $\bm{x}$ is a convex quadratic programming problem with non-negative constraints,

$$
\bm{x}_{t}^{\star}= \argmin_{x_{i}>0} \sum_{i=1}^{m} \lambda_{t,i} \big(  \Psi_{C_{i}, t}(\pib_{k-1}, \pib_{k}) + x_{t,i} \big) + \frac{\beta}{2} \sum_{i=1}^{m}\big(  \Psi_{C_{i}, t}(\pib_{k-1}, \pib_{k}) + x_{t,i} \big)^{2}
$$

The optimal solution is $\; x_{t, i}^{\star}= \max \left\{0, -\frac{\lambda_{t,i}}{\beta} - \Psi_{C_{i}, t}(\pib_{k-1}, \pib_{k}) \right\}$. Then,

$$
\begin{aligned}
w_{t,i}(\pib_{k}) = \Psi_{C_{i}, t}(\pib_{k-1}, \pib_{k}) &+ x_{t, i}^{\star} =  \Psi_{C_{i}, t}(\pib_{k-1}, \pib_{k}) + \max \left\{0, -\frac{\lambda_{t,i}}{\beta} - \Psi_{C_{i}, t}(\pib_{k-1}, \pib_{k}) \right\} \\
& = \frac{\lambda_{t,i}}{\beta} + \Psi_{C_{i}, t}(\pib_{k-1}, \pib_{k}) + \max \left\{0, -\frac{\lambda_{t,i}}{\beta} - \Psi_{C_{i}, t}(\pib_{k-1}, \pib_{k}) \right\} - \frac{\lambda_{t,i}}{\beta} \\
& =\max \left\{0, \frac{\lambda_{t,i}}{\beta} + \Psi_{C_{i}, t}(\pib_{k-1}, \pib_{k}) \right\} -\frac{\lambda_{t,i}}{\beta}
\end{aligned}
$$

Substituting back into Equation \eqref{appendix:eq:maxminquadratic}, we get

\resizebox{\linewidth}{!}{$
\begin{aligned}
\Lcal_{t}(\pib_{k}, \bm{\lambda}, \bm{x}, \beta) = & \sum_{h=t}^{H} \E{s\sim \rho_{\pi_{k,h}} \\a\sim \pi_{k-1,h}} \big[-{\rho(\theta_{h})}A^{\pib_{k-1}}_{h}(s,a) \big] + \sum_{i}^{m} \lambda_{t,i} w_{t,i}(\pib_{k}) + \frac{\beta}{2} \sum_{i}^{m} w_{t,i}^{2}(\pib_{k}) \\
= & \sum_{h=t}^{H} \E{s\sim \rho_{\pi_{k,h}} \\a\sim \pi_{k-1,h}} \big[-{\rho(\theta_{h})}A^{\pib_{k-1}}_{h}(s,a) \big] + \sum_{i=1}^{m} \lambda_{t,i} \left( \max \left\{0, \frac{\lambda_{t,i}}{\beta} + \Psi_{C_{i}, t}(\pib_{k-1}, \pib_{k}) \right\} - \frac{\lambda_{t,i}}{\beta} \right) \\
& \quad +\frac{\beta}{2} \sum_{i=1}^{m} \left( \max \left\{0, \frac{\lambda_{t,i}}{\beta} + \Psi_{C_{i}, t}(\pib_{k-1}, \pib_{k}) \right\} -\frac{\lambda_{t,i}}{\beta} \right)^{2} \\
= & \sum_{h=t}^{H} \E{s\sim \rho_{\pi_{k,h}} \\a\sim \pi_{k-1,h}} \big[-{\rho(\theta_{h})}A^{\pib_{k-1}}_{h}(s,a) \big] + \frac{\beta}{2} \sum_{i=1}^{m}\left( \max \left\{0, \frac{\lambda_{t,i}}{\beta} + \Psi_{C_{i}, t}(\pib_{k-1}, \pib_{k}) \right\}^{2} -\frac{\lambda_{t,i}^{2}}{\beta^{2}} \right)
\end{aligned}
$}

Hence, we finally get

\resizebox{\linewidth}{!}{$
\begin{aligned}
(\pi_{k,t}^{\star}, \bm{\lambda}_{t}^{\star}) &= \max_{\bm{\lambda} \geq 0} \min_{\pi_{k,t}} \Lcal_{t}(\pib_{k}, \bm{\lambda}, \beta) \\ &= \max_{\bm{\lambda} \geq 0} \min_{\pi_{k,t}} \sum_{h=t}^{H} \E{s\sim \rho_{\pi_{k,h}} \\a\sim \pi_{k-1,h}} \big[-{\rho(\theta_{h})}A^{\pib_{k-1}}_{h}(s,a) \big] + \frac{\beta}{2} \sum_{i=1}^{m} \bigg( \max \bigg\{0, \Psi_{C_{i}, t}(\pib_{k-1}, \pib_{k}) + \frac{\lambda_{t,i}}{\beta} \bigg\}^{2} - \frac{\lambda_{t,i}^{2}}{\beta^{2}}  \bigg)
\end{aligned}
$}

\end{proof}

\begin{restatable}{lemma}{step1}
\label{lemma:withrelusolutionsame1} 
Consider two problems, Problem \eqref{eq:withoutmaxquadratic} and Problem \eqref{eq:maxquadratic} below. 
For sufficiently large $\lambda_{i} > \bar\lambda \forAll i$ and $\beta > \bar\beta$ for some finite $\bar\lambda$ and finite $\bar\beta$, the optimal solution set of Problem \eqref{eq:maxquadratic} (equivalent version of Problem \eqref{eq:damped_intermediate_problem}) is identical to the optimal solution set of Problem \eqref{eq:withoutmaxquadratic}. 

Problem \eqref{eq:withoutmaxquadratic} :
\begin{align*}
\Lcal_{t}^{P}(\pib_{k}, \bm{\lambda}, \bm{x}, \beta) &:=  \sum_{h=t}^{H} \E{s\sim \rho_{\pi_{k,h}} \\a\sim \pi_{k-1,h}} \big[-{\rho(\theta_{h})}A^{\pib_{k-1}}_{h}(s,a) \big] + \sum_{i}^{m} \lambda_{t,i} w_{t,i}(\pib_{k}) + \frac{\beta}{2} \sum_{i}^{m} w_{t,i}^{2}(\pib_{k}) \\
\text{Then,} &\quad\quad (\pi_{k,t}^{\star}, \bm{\lambda}_{t}^{\star}, \bm{x}_{t}^{\star}) = \max_{\bm{\lambda} \geq 0} \min_{\pi_{k,t}, \bm{x}} \Lcal_{t}^{P}(\pib_{k}, \bm{\lambda}, \bm{x}, \beta) \hspace{1cm} \tag{P}
\label{eq:withoutmaxquadratic}
\end{align*}

Problem \eqref{eq:maxquadratic} :

\begin{align*}
\Lcal_{t}^{Q}(\pib_{k}, \bm{\lambda}, \bm{x}, \beta) &:=  \sum_{h=t}^{H} \E{s\sim \rho_{\pi_{k,h}} \\a\sim \pi_{k-1,h}} \big[-{\rho(\theta_{h})}A^{\pib_{k-1}}_{h}(s,a) \big] + \sum_{i}^{m} \lambda_{t,i} \Psi_{C_{i}, t}^{+}(\pib_{k-1}, \pib_{k}, \theta) \\ & \qquad \qquad + \frac{\beta}{2} \sum_{i}^{m} \Psi_{C_{i}, t}^{+}(\pib_{k-1}, \pib_{k}, \theta)^{2} \\
\text{Then,} &\quad\quad (\pi_{k,t}^{\star}, \bm{\lambda}_{t}^{\star}, \bm{x}_{t}^{\star}) = \max_{\bm{\lambda} \geq 0} \min_{\pi_{k,t}, \bm{x}} \Lcal_{t}^{Q}(\pib_{k}, \bm{\lambda}, \bm{x}, \beta) \hspace{1cm} \tag{Q}
\label{eq:maxquadratic}
\end{align*}

, where $x^{+} := \max(0, x)$, and 
$$\Psi_{C_{i}, t}(\pib_{k-1}, \pib_{k}, \theta) := \sum_{h=t}^{H} \mathop{\mathbb{E}}_{\substack{s\sim \rho_{\pi_{k,h}} \\a\sim \pi_{k-1,h}}} \big  [{\rho(\theta_{h})}A_{C_i, h}^{\pib_{k-1}} (s,a) \big ] + (J_{C_i}(\pib_{k-1})-d_i).
$$
\end{restatable}

\begin{proof}
This proof uses some ideas given in \cite{zhang2022penalized} for Part 1 below.

\textbf{Part 1} - Solution of Problem \eqref{eq:withoutmaxquadratic} is solution of Problem \eqref{eq:maxquadratic}. \hfill \break
\vspace{-0.25cm}

Suppose $\bar{\theta}_{t}$ is the optimum of the constrained Problem \eqref{eq:withoutmaxquadratic} augmented with the quadratic penalty. Let $\bar\lambda_{t}$ be the corresponding Lagrange multiplier vector for its dual problem, and $\bar\beta$ be the additive quadratic penalty coefficient. Then for $\lambda_{t,i} \geq ||\bar\lambda||_\infty \forAll i$ and $\beta \geq ||\bar\beta||_\infty$, $\bar{\theta}$ is also a minimizer of its ReLU-penalized optimization Problem \eqref{eq:maxquadratic} as below. Let $\Omega(\theta_{t}) := \sum_{h=t}^{H} \E{s\sim \rho_{\pi_{k,h}} \\a\sim \pi_{k-1,h}} \big[-{\rho(\theta_{h})}A^{\pib_{k-1}}_{h}(s,a) \big]$. Then it follows that:

\resizebox{\linewidth}{!}{$
\begin{aligned}
    \Omega(\theta_{t}) + \sum_{i}^{m} \lambda_{t,i} \Psi_{C_{i}, t}^{+}(\pib_{k-1}, \pib_{k}, \theta) + \frac{\beta}{2} \sum_{i}^{m} \Psi_{C_{i}, t}^{+}(\pib_{k-1}, \pib_{k}, \theta)^{2}  
    &\geq \Omega(\theta_{t}) + \sum_{i}^{m} \bar\lambda_{i} \Psi_{C_{i}, t}^{+}(\pib_{k-1}, \pib_{k}, \theta) + \frac{\bar\beta}{2} \sum_{i}^{m} \Psi_{C_{i}, t}^{+}(\pib_{k-1}, \pib_{k}, \theta)^{2} \\ 
    &\geq \Omega(\theta_{t}) + \sum_{i}^{m} \bar\lambda_{i} \Psi_{C_{i}, t}(\pib_{k-1}, \pib_{k}, \theta) + \frac{\bar\beta}{2} \sum_{i}^{m} \Psi_{C_{i}, t}(\pib_{k-1}, \pib_{k}, \theta)^{2}
\end{aligned}
$}

By assumption, $\bar\theta_{t}$ is a Karush-Kuhn-Tucker point in the constrained Problem \eqref{eq:withoutmaxquadratic}, at which KKT conditions are satisfied with the Lagrange multiplier vector  $\bar{\lambda}$ and $\bar\beta$. We then have:


\resizebox{\linewidth}{!}{$
\begin{aligned}
   \Omega(\theta_{t}) + \sum_{i}^{m} \bar\lambda_{i} \Psi_{C_{i}, t}(\pib_{k-1}, \pib_{k}, \theta) + \frac{\bar\beta}{2} \sum_{i}^{m} \Psi_{C_{i}, t}(\pib_{k-1}, \pib_{k}, \theta)^{2}
    &\geq \Omega(\bar\theta_{t}) + \sum_{i}^{m} \bar\lambda_{i} \Psi_{C_{i}, t}(\pib_{k-1}, \pib_{k}, \bar\theta) + \frac{\bar\beta}{2} \sum_{i}^{m} \Psi_{C_{i}, t}(\pib_{k-1}, \pi, \bar\theta)^{2} \\ 
    &=  \Omega(\bar\theta_{t}) + \sum_{i}^{m} \bar\lambda_{i} \Psi_{C_{i}, t}^{+}(\pib_{k-1}, \pib_{k}, \bar\theta) + \frac{\bar\beta}{2} \sum_{i}^{m} \Psi_{C_{i}, t}^{+}(\pib_{k-1}, \pib_{k}, \bar\theta)^{2} \\ 
     &=  \Omega(\bar\theta_{t}) + \sum_{i}^{m} \lambda_{t,i} \Psi_{C_{i}, t}^{+}(\pib_{k-1}, \pib_{k}, \bar\theta) + \frac{\beta}{2} \sum_{i}^{m} \Psi_{C_{i}, t}^{+}(\pib_{k-1}, \pib_{k}, \bar\theta)^{2}
\end{aligned}
$}

, where the first line holds because $\bar\theta_{t}$ minimizes the Lagrange function, and the second line is derived from the complementary slackness. Thus, we conclude that for the objective function of Problem \eqref{eq:maxquadratic}, call it $\Lcal^{Q}(\theta_{t})$, we have $\mathcal{L}^{Q}(\theta_{t}) \geq \mathcal{L}^{Q}(\bar\theta_{t})$ for all $\theta_{t} \in \Theta$, which means $\bar\theta_{t}$ is a minimizer of  the quadratic damped optimization Problem \eqref{eq:maxquadratic}. 

\vspace{0.5cm}
\textbf{Part 2} - Solution of Problem \eqref{eq:maxquadratic} is solution of Problem \eqref{eq:withoutmaxquadratic}. \hfill \break 
\vspace{-0.25cm}

Let $\widetilde{\theta}_{t}$ be an optimal point of the quadratic damped Problem \eqref{eq:maxquadratic}, with $\bar\theta_{t}$ and $\bar\lambda$ being the same as defined above. Then, if $\widetilde\theta_{t}$ is in the set of feasible solutions $S_{\text{feasible}} = \{\theta \given \Psi_{C_{i}, t}(\pib_{k-1}, \pib_{k}, \theta) \leq 0 \ \forAll i\}$, we have:

\begin{align*}
\Omega(\widetilde\theta_{t}) &= \Omega(\widetilde\theta_{t}) + \sum_{i}^{m} \lambda_{t,i} \Psi_{C_{i}, t}^{+}(\pib_{k-1}, \pib_{k}, \widetilde\theta) + \frac{\beta}{2} \sum_{i}^{m} \Psi_{C_{i}, t}^{+}(\pib_{k-1}, \pib_{k}, \widetilde\theta) \\
&\leq \Omega(\theta_{t}) + \sum_{i}^{m} \lambda_{t,i} \Psi_{C_{i}, t}^{+}(\pib_{k-1}, \pib_{k}, \theta) + \frac{\beta}{2} \sum_{i}^{m} \Psi_{C_{i}, t}^{+}(\pib_{k-1}, \pib_{k}, \theta) 
\\ &= \Omega(\theta_{t})
\end{align*}

The inequality above indicates  $\widetilde{\theta}_{t}$ is also optimal in the constrained Problem \eqref{eq:withoutmaxquadratic}. Now, if  $\widetilde\theta$ is not feasible, we have:

\resizebox{\linewidth}{!}{$
\begin{aligned}
 \Omega(\bar\theta_{t}) + \sum_{i}^{m} \lambda_{t,i} \Psi_{C_{i}, t}^{+}(\pib_{k-1}, \pib_{k}, \bar\theta) + \frac{\beta}{2} \sum_{i}^{m} \Psi_{C_{i}, t}^{+}(\pib_{k-1}, \pib_{k}, \bar\theta)^{2}
    &=  \Omega(\bar\theta_{t}) + \sum_{i}^{m} \bar\lambda_{i} \Psi_{C_{i}, t}^{+}(\pib_{k-1}, \pib_{k}, \bar\theta) + \frac{\bar\beta}{2} \sum_{i}^{m} \Psi_{C_{i}, t}^{+}(\pib_{k-1}, \pib_{k}, \bar\theta)^{2} \\
    & = \Omega(\bar\theta_{t}) + \sum_{i}^{m} \bar\lambda_{i} \Psi_{C_{i}, t}(\pib_{k-1}, \pib_{k}, \bar\theta) + \frac{\bar\beta}{2} \sum_{i}^{m} \Psi_{C_{i}, t}(\pib_{k-1}, \pib_{k}, \bar\theta)^{2} \\
    & \leq  \Omega(\widetilde\theta_{t}) + \sum_{i}^{m} \bar\lambda_{i} \Psi_{C_{i}, t}(\pib_{k-1}, \pib_{k}, \widetilde\theta) + \frac{\bar\beta}{2} \sum_{i}^{m} \Psi_{C_{i}, t}(\pib_{k-1}, \pib_{k}, \widetilde\theta)^{2} \\
   & \leq \Omega(\widetilde\theta_{t}) + \sum_{i}^{m} \bar\lambda_{i} \Psi_{C_{i}, t}^{+}(\pib_{k-1}, \pib_{k}, \widetilde\theta) + \frac{\bar\beta}{2} \sum_{i}^{m} \Psi_{C_{i}, t}^{+}(\pib_{k-1}, \pib_{k}, \widetilde\theta)^{2} \\
   & \leq \Omega(\widetilde\theta_{t}) + \sum_{i}^{m} \lambda_{t,i} \Psi_{C_{i}, t}^{+}(\pib_{k-1}, \pib_{k}, \widetilde\theta) + \frac{\beta}{2} \sum_{i}^{m} \Psi_{C_{i}, t}^{+}(\pib_{k-1}, \pib_{k}, \widetilde\theta)^{2}
\end{aligned}
$}

, which is a contradiction to the assumption that  $\widetilde{\theta}_{t}$ is a minimizer of the penalized optimization Problem \eqref{eq:maxquadratic}. Thus, $\widetilde\theta_{t}$ can only be the feasible optimal solution for Problem \eqref{eq:withoutmaxquadratic}.

\end{proof}


\begin{restatable}{lemma}{step2}
\label{lemma:withrelusolutionsame2} 
Consider two problems, Problem \eqref{eq:withoutmaxquadratic'} and Problem \eqref{eq:simplewithoutmaxquadratic}. 
For sufficiently large $\beta > \bar\beta$ for some finite $\bar\beta$, the feasible optimal solution set of Problem \eqref{eq:simplewithoutmaxquadratic} (equivalent version of Problem \eqref{eq:cpo_improvement_vanilla}) is identical to the solution set of Problem \eqref{eq:withoutmaxquadratic'}. 

Problem \eqref{eq:withoutmaxquadratic'} :
\begin{align*}
\Lcal_{t}^{P'}(\pib_{k}, \bm{\lambda}, \bm{x}, \beta) &:=  \sum_{h=t}^{H} \E{s\sim \rho_{\pi_{k,h}} \\a\sim \pi_{k-1,h}} \big[-{\rho(\theta_{h})}A^{\pib_{k-1}}_{h}(s,a) \big] + \sum_{i}^{m} \lambda_{t,i} w_{t,i}(\pib_{k}) + \frac{\beta}{2} \sum_{i}^{m} w_{t,i}^{2}(\pib_{k}) \\
\text{Then,} &\quad\quad (\pi_{k,t}^{\star}, \bm{\lambda}_{t}^{\star}, \bm{x}_{t}^{\star}) = \max_{\bm{\lambda} \geq 0} \min_{\pi_{k,t}, \bm{x}} \Lcal_{t}^{P'}(\pib_{k}, \bm{\lambda}, \bm{x}, \beta)
\label{eq:withoutmaxquadratic'}
\tag{P'}
\end{align*}

Problem \eqref{eq:simplewithoutmaxquadratic} :
\begin{align*}
\Lcal_{t}^{R}(\pib_{k}, \bm{\lambda}, \bm{x}) &:=  \sum_{h=t}^{H} \E{s\sim \rho_{\pi_{k,h}} \\a\sim \pi_{k-1,h}} \big[-{\rho(\theta_{h})}A^{\pib_{k-1}}_{h}(s,a) \big] + \sum_{i}^{m} \lambda_{t,i} w_{t,i}(\pib_{k}) \\
\text{Then,} &\quad\quad (\pi_{k,t}^{\star}, \bm{\lambda}_{t}^{\star}, \bm{x}_{t}^{\star}) = \max_{\bm{\lambda} \geq 0} \min_{\pi_{k,t}, \bm{x}} \Lcal_{t}^{R}(\pib_{k}, \bm{\lambda}, \bm{x})
\tag{R}
\label{eq:simplewithoutmaxquadratic}
\end{align*}
\end{restatable}

\begin{proof}

Recall that we are using parameterized policies, hence we overload notation as $\theta \equiv \pi$ frequently. For brevity, denote $\Omega_{t}(\pi) := \sum_{h=t}^{H} \E{s\sim \rho_{\pi_{k,h}} \\a\sim \pi_{k-1,h}} \big[{\rho(\theta_{h})}A^{\pib_{k-1}}_{h}(s,a) \big]$. We will also go back and forth between the equivalent problems of Problem \eqref{eq:withoutmaxquadratic'} and Problem \eqref{eq:damped_intermediate_problem} of the main chapter in Section \ref{sec:intro_quadratic_term}.

\textbf{Part 1.} Solution of Problem \eqref{eq:simplewithoutmaxquadratic} is solution of Problem \eqref{eq:withoutmaxquadratic'}.

Suppose that $\pi^{\star}$ is the optimal feasible policy for the primal Problem \eqref{eq:simplewithoutmaxquadratic}, which is a Lagrangian version of Problem \eqref{eq:cpo_improvement_vanilla}. Consider the corresponding Langrangian dual parameter $\bm{\lambda}^{\star}$ of $\pi^{\star}$, which satisfies the KKT conditon,

$$
\nabla_{\pi} \mathcal{L}_{t}^{R} \left(\pi_{k,t}^{\star}, \bm{\lambda}^{\star}, \bm{x}^{\star} \right)=-\nabla_{\pi} \Omega_{t}\left(\pi_{k,t}^{\star} \right) + \sum_{i=1}^{m} \lambda_{t,i}^{\textbf{*}} \nabla_{\pi} w_{t, i}\left(\pib_{k}^{\star} \right) = 0
$$

and the second-order sufficient condition that for all non-zero vectors $\bm{u}$ that satisfy $\bm{u}^{T} \nabla_{\pi} w_{t, i}\left(\pib_{k}^{\star}\right)=$ 0 , we have

\begin{align*}
\bm{u}^{T} \nabla_{\pi}^{2} \mathcal{L}_{t}^{R}\left(\pi_{k,t}^{\star}, \bm{\lambda}^{\star}, \bm{x}^{\star}\right) \bm{u}>0
\tag{A}
\label{eq:secondorder}
\end{align*}

Compare Equation \eqref{eq:simplewithoutmaxquadratic} and Equation \eqref{eq:withoutmaxquadratic'}, we have,

$$
\begin{aligned}
\nabla_{\pi} \mathcal{L}_{t}^{P'}\left(\pi_{k,t}^{\star}, \bm{\lambda}^{\star}, \bm{x}^{\star}, \beta \right) & =-\nabla_{\pi}\Omega_{t}\left( \pi_{k,t}^{\star} \right) + \sum_{i=1}^{m} \lambda_{t,i}^{\star} \nabla_{\pi} w_{t,i} \left(\pib_{k}^{\star}\right) + \beta \sum_{i=1}^{m} w_{t, i} \left( \pib_{k}^{\star} \right) \nabla_{\pi} w_{t, i}\left( \pib_{k}^{\star} \right) \\
& = \nabla_{\pi} \mathcal{L}_{t}^{R} \left(\pi_{k,t}^{\star}, \bm{\lambda}^{\star}, \bm{x}^{\star} \right) + \beta \sum_{i=1}^{m} w_{t, i} \left( \pib_{k}^{\star} \right) \nabla_{\pi} w_{t, i}\left(\pib_{k}^{\star}\right) \\
& =0
\end{aligned}
$$

, where we use $w_{t,i}(\pib_{k}^{\star}) := \Psi_{C_{i}, t}(\pib_{k-1}, \pib^{\star}) + x_{t,i}^{\star} = 0$ with the feasible policy $\pib^{\star}$. Moreover,

$$
\begin{aligned}
\nabla_{\pi}^{2} \mathcal{L}_{t}^{P'}\left(\pi_{k,t}^{\star}, \bm{\lambda}^{\star}, \bm{x}^{\star}, \beta \right) = & -\nabla_{\pi}^{2} \Omega_{t}\left( \pi_{k,t}^{\star} \right) \\
& +\sum_{i=1}^{m} \lambda_{t,i}^{\star} \nabla_{\pi}^{2} w_{t, i}\left(\pib_{k}^{\star}\right) + \beta \nabla_{\pi} \bm{w}_{t}\left(\pib_{k}^{\star}\right) \nabla_{\pi} \bm{w}_{t}\left(\pib_{k}^{\star}\right)^{T} \\
= & \nabla_{\pi}^{2} \mathcal{L}_{t}^{R}\left(\pi_{k,t}^{\star}, \bm{\lambda}^{\star}, \bm{x}^{\star}\right)+\beta \nabla_{\pi} \bm{w}_{t}\left(\pib_{k}^{\star}\right) \nabla_{\pi} \bm{w}_{t}\left(\pib_{k}^{\star}\right)^{T}.
\end{aligned}
$$

To prove that $(\pi_{k,t}^{\star}, \bm{\lambda}^{\star})$ is a strict minimum solution to $\mathcal{L}_{t}^{P'}(\pib_{k}, \bm{\lambda}, \bm{x}, \beta)$, we only need to prove the following is true for sufficiently large $\beta$,

$$
\nabla_{\pi}^{2} \mathcal{L}_{t}^{P'}\left(\pib_{k}^{\star}, \bm{\lambda}^{\star}, \bm{x}^{\star}, \beta \right) \succ 0 .
$$

If the above is not true, then for any large $\beta$, there exists $\bm{u}_{t}$ such that $\left\|\bm{u}_{t}\right\|=1$ and satisfies

\begin{align*}
\bm{u}_{t}^{T} \nabla_{\pi}^{2} \mathcal{L}_{t}^{P'}\left(\pib_{k}^{\star}, \bm{\lambda}^{\star}, \bm{x}^{\star}, \beta \right) \bm{u}_{t} &= \bm{u}_{t}^{T} \nabla_{\pi}^{2} \mathcal{L}_{t}^{R} \left(\pib_{k}^{\star}, \bm{\lambda}^{\star}, \bm{x}^{\star} \right) \bm{u}_{t} + \beta \left\| \nabla_{\pi} \bm{w}_{t}\left(\pib_{k}^{\star}\right)^{T} \bm{u}_{t}\right\|^{2} \leq 0 \\
\implies \left\|\nabla_{\pi} \bm{w}_{t}\left(\pib_{k}^{\star}\right)^{T} \bm{u}_{t}\right\|^{2} & \leq-\frac{1}{\beta} \bm{u}_{t}^{T} \nabla_{\pi}^{2} \mathcal{L}_{t}^{R} \left(\pib_{k}^{\star}, \bm{\lambda}^{\star}, \bm{x}^{\star} \right) \bm{u}_{t} \; \rightarrow \; 0, \; \text{as} \; \beta \rightarrow \infty.
\end{align*}

Therefore, $\left\{\bm{u}_{h}\right\}$ is a bounded sequence and there must be a limit point, denoted by $\ring{\bm{u}}$. Then

$$
\begin{gathered}
\nabla_{\pi} \bm{w}_{t}\left(\pib_{k}^{\star}\right)^{T} \ring{\bm{u}}=0 \\
\ring{\bm{u}}^{T} \nabla_{\pi}^{2} \mathcal{L}_{t}^{R}\left(\pib_{k}^{\star}, \bm{\lambda}^{\star}, \bm{x}^{\star}\right) \ring{\bm{u}} \leq 0 .
\end{gathered}
$$

The above contradicts Equation \eqref{eq:secondorder}, so the conclusion. Hence, $\pi_{k,t}^{\star}$ is also the optimal feasible policy for the primal-dual Problem \eqref{eq:withoutmaxquadratic'}.

\vspace{0.5cm}
\textbf{Part 2.} Solution of Problem \eqref{eq:withoutmaxquadratic'} is solution of Problem \eqref{eq:simplewithoutmaxquadratic}.

This part is straightforward since it is a standard result. Please see Chapter 2 and Chapter 9 of \cite{birgin2014practical}, and Chapter 2 and Chapter 4 of \cite{bertsekas2014constrained} for the proof. For completeness, we provide the result below.

Suppose $\pi_{k,t}^{\star}$ in the feasible optimal solution set of the primal-dual Problem \eqref{eq:withoutmaxquadratic'}. Let $\lambda^{\star}$ be the corresponding dual parameter of $\pi_{k,t}^{\star}$. Consider Problem \eqref{eq:damped_intermediate_problem}, which is an equivalent version of Problem \eqref{eq:withoutmaxquadratic'}. For any feasible $\pib_{k}$, we have

$$
\mathcal{L}_{t}\left(\pib_{k}^{\star}, \bm{\lambda}^{\star}, \beta \right) \leq \mathcal{L}_{t}\left(\pib_{k}, \bm{\lambda}^{\star}, \beta \right) .
$$

Now we have two cases:

\underline{Case 1}. When $\frac{\lambda_{t,i}^{\star}}{\beta} + \Psi_{C_{i}, t}(\pib_{k-1}, \pib_{k}^{\star})>0$, we have

$$
\begin{aligned}
\mathcal{L}_{t}\left(\pib_{k}, \lambda^{\star}, \beta \right) & =-\Omega_{t}(\pi_{k,t}) + \sum_{i=1}^{m} \lambda_{t,i}^{\star} \Psi_{C_{i}, t}(\pib_{k-1}, \pib_{k}) + \frac{\beta}{2} \sum_{i=1}^{m} \Psi_{C_{i}, t}^{2}(\pib_{k-1}, \pib_{k}) \\
& =-\Omega_{t}(\pi_{k,t}) + \sum_{i=1}^{m} \beta \Psi_{C_{i}, t}(\pib_{k-1}, \pib_{k}) \left( \frac{\lambda_{t,i}^{\star}}{\beta} + \Psi_{C_{i}, t}(\pib_{k-1}, \pib_{k}) \right) - \frac{\beta}{2} \sum_{i=1}^{m} \Psi_{C_{i}, t}^{2}(\pib_{k-1}, \pib_{k}) \\
&\leq -\Omega_{t}(\pi_{k,t})
\end{aligned}
$$

where the last step uses $\beta>0,\Psi_{C_{i}, t}(\pib_{k-1}, \pib_{k}) < 0$, and $\frac{\lambda_{t,i}^{\star}}{\beta} + \Psi_{C_{i}, t}(\pib_{k-1}, \pib_{k}) >0$.

\underline{Case 2}. When $\frac{\lambda_{t,i}^{\star}}{\beta} + \Psi_{C_{i}, t}(\pib_{k-1}, \pib_{k}^{\star}) \leq 0$, we have

$$
\mathcal{L}_{t}\left(\pib_{k}, \bm{\lambda}^{\star}, \beta \right) = -\Omega_{t}(\pi_{k,t}) - \frac{1}{2 \beta} \sum_{i=1}^{m} \lambda_{t,i}^{\star 2} \leq -\Omega_{t}(\pi_{k,t}).
$$

Now, combining both cases above, we have $\mathcal{L}_{t}\left(\pib_{k}, \bm{\lambda}^{\star}, \beta \right) \leq -\Omega_{t}(\pi_{k,t}).$ On the other hand, $ \mathcal{L}_{t}\left(\pib_{k}^{\star}, \bm{\lambda}^{\star}, \beta \right)=\mathcal{L}_{t}\left(\pib_{k}^{\star}, \bm{\lambda}^{\star}, \bm{x}^{\star}, \beta \right)= - \Omega_{t}(\pi_{k,t}^{\star}).$ Thus, the combining all of the above we get

$$
-\Omega_{t}(\pi_{k,t}^{\star}) = \mathcal{L}_{t} \left( \pib_{k}^{\star}, \bm{\lambda}^{\star}, \beta \right) \leq \mathcal{L}_{t} \left(\pib_{k}, \bm{\lambda}^{\star}, \beta \right) \leq -\Omega_{t}(\pi_{k,t}).
$$

\end{proof}

\relusolutionsame*
\begin{proof}
\label{proof:relu_solution_same}

We prove this result as a two-step process. 

First, we show that the solution sets of the below problems are identical. See Lemma \ref{lemma:withrelusolutionsame1} for the proof.

Problem \eqref{eq:withoutmaxquadratic}.
\begin{align*}
\Lcal_{t}^{P}(\pib_{k}, \bm{\lambda}, \bm{x}, \beta) &:=  \sum_{h=t}^{H} \E{s\sim \rho_{\pi_{k,h}} \\a\sim \pi_{k-1,h}} \big[-{\rho(\theta_{h})}A^{\pib_{k-1}}_{h}(s,a) \big] + \sum_{i}^{m} \lambda_{t,i} w_{t,i}(\pib_{k}) + \frac{\beta}{2} \sum_{i}^{m} w_{t,i}^{2}(\pib_{k}) \\
\text{Then,} &\quad\quad (\pi_{k,t}^{\star}, \bm{\lambda}_{t}^{\star}, \bm{x}_{t}^{\star}) = \max_{\bm{\lambda} \geq 0} \min_{\pi_{k,t}, \bm{x}} \Lcal_{t}^{P}(\pib_{k}, \bm{\lambda}, \bm{x}, \beta) \hspace{1cm}
\tag{P}
\end{align*}

Problem \eqref{eq:maxquadratic}.
\begin{align*}
\Lcal_{t}^{Q}(\pib_{k}, \bm{\lambda}, \bm{x}, \beta) &:=  \sum_{h=t}^{H} \E{s\sim \rho_{\pi_{k,h}} \\a\sim \pi_{k-1,h}} \big[-{\rho(\theta_{h})}A^{\pib_{k-1}}_{h}(s,a) \big] + \sum_{i}^{m} \lambda_{t,i} \Psi_{C_{i}, t}^{+}(\pib_{k-1}, \pib_{k}, \theta) \\ & \qquad \qquad + \frac{\beta}{2} \sum_{i}^{m} \Psi_{C_{i}, t}^{+}(\pib_{k-1}, \pib_{k}, \theta)^{2} \\
\text{Then,} &\quad\quad (\pi_{k,t}^{\star}, \bm{\lambda}_{t}^{\star}, \bm{x}_{t}^{\star}) = \max_{\bm{\lambda} \geq 0} \min_{\pi_{k,t}, \bm{x}} \Lcal_{t}^{Q}(\pib_{k}, \bm{\lambda}, \bm{x}, \beta)
\tag{Q}
\end{align*}

Second, we show that the solution sets of the below problems are identical. See Lemma \ref{lemma:withrelusolutionsame2} for the proof.

Problem \eqref{eq:withoutmaxquadratic'} :
\begin{align*}
\Lcal_{t}^{P'}(\pib_{k}, \bm{\lambda}, \bm{x}, \beta) &:=  \sum_{h=t}^{H} \E{s\sim \rho_{\pi_{k,h}} \\a\sim \pi_{k-1,h}} \big[-{\rho(\theta_{h})}A^{\pib_{k-1}}_{h}(s,a) \big] + \sum_{i}^{m} \lambda_{t,i} w_{t,i}(\pib_{k}) + \frac{\beta}{2} \sum_{i}^{m} w_{t,i}^{2}(\pib_{k}) \\
\text{Then,} &\quad\quad (\pi_{k,t}^{\star}, \bm{\lambda}_{t}^{\star}, \bm{x}_{t}^{\star}) = \max_{\bm{\lambda} \geq 0} \min_{\pi_{k,t}, \bm{x}} \Lcal_{t}^{P'}(\pib_{k}, \bm{\lambda}, \bm{x}, \beta)
\tag{P'}
\end{align*}

Problem \eqref{eq:simplewithoutmaxquadratic} :
\begin{align*}
\Lcal_{t}^{R}(\pib_{k}, \bm{\lambda}, \bm{x}) &:=  \sum_{h=t}^{H} \E{s\sim \rho_{\pi_{k,h}} \\a\sim \pi_{k-1,h}} \big[-{\rho(\theta_{h})}A^{\pib_{k-1}}_{h}(s,a) \big] + \sum_{i}^{m} \lambda_{t,i} w_{t,i}(\pib_{k}) \\
\text{Then,} &\quad\quad (\pi_{k,t}^{\star}, \bm{\lambda}_{t}^{\star}, \bm{x}_{t}^{\star}) = \max_{\bm{\lambda} \geq 0} \min_{\pi_{k,t}, \bm{x}} \Lcal_{t}^{R}(\pib_{k}, \bm{\lambda}, \bm{x})
\tag{R}
\end{align*}

Now, it follows from equivalency that the optimal solution of  Problem \eqref{eq:maxquadratic} and Problem \eqref{eq:simplewithoutmaxquadratic}, and hence Problem \eqref{eq:damped_intermediate_problem} and Problem \eqref{eq:cpo_improvement_vanilla}, is the same.

\end{proof}

\subsection{Experiments Revisited}
\label{sec:appendix_experiments}

\begin{figure}[t]
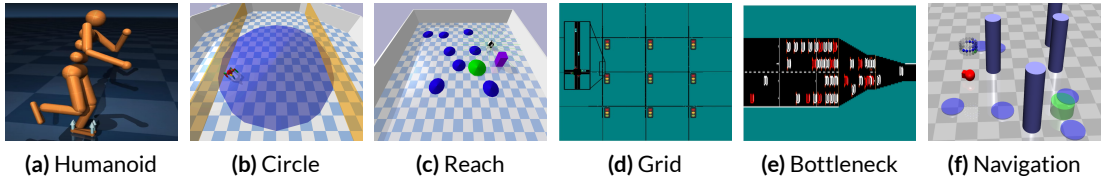

\captionsetup[subfigure]{justification=centering}
    \centering
    \subfloat[Humanoid]{
        \includegraphics[height=.12\textwidth, width=0.15\textwidth]{chapter2/images/task_humanoid.JPG}
    }
    \subfloat[Circle]{
        \includegraphics[height=.12\textwidth, width=0.15\textwidth]{chapter2/images/task_circle.png}
    }
    \subfloat[Reach]{
        \includegraphics[height=.12\textwidth, width=0.15\textwidth]{chapter2/images/task_reach.png}
    }
    \subfloat[Grid]{
        \includegraphics[height=.12\textwidth, width=0.15\textwidth]{chapter2/images/task_grid.JPG}
    }
    \subfloat[Bottleneck]{
        \includegraphics[height=.12\textwidth, width=0.15\textwidth]{chapter2/images/task_bottleneck.png}
    }
    \subfloat[Navigation]{
        \includegraphics[height=.12\textwidth, width=0.15\textwidth]{chapter2/images/task_navigation.png}
    }
    \caption[]{\small The Humanoid, Circle, Reach, Grid, Bottleneck, and Navigation tasks. (a) \textbf{Humanoid}: The agent is to run as fast as possible on a flat surface, while not exceeding a specified speed limit i.e. the cost constraint. (b) \textbf{Circle}: The agent is rewarded for moving in a specified circle but is penalized if the diameter of the circle is larger than some value \cite{achiam2017constrained}.  (c) \textbf{Reach}: The agent is rewarded for reaching a goal while avoiding obstacles (cost constraints) that are placed to hinder the agent \cite{ray2019benchmarking}. (d) \textbf{Grid}: The agent controls traffic lights in a 3x3 road network and is rewarded for high traffic throughput but is constrained to let lights be red for at most 5 consecutive seconds \cite{vinitsky2018benchmarks}. (e) \textbf{Bottleneck}: The agent controls vehicles (red) in a merging traffic situation and is rewarded for maximizing the number of vehicles that pass through but is constrained to ensure that white vehicles (not controlled by agent) have ``low'' speed for no more than 10 seconds \cite{vinitsky2018benchmarks}. (f) \textbf{Navigation}: The agent is rewarded for reaching the target area (green) but is constrained to avoid hazards (light purple) and impassible pillars (dark purple). The cost for hazards and pillars is different \cite{ray2019benchmarking}.}
    \label{fig:appendix:envs}
\end{figure}

Below we detail the experimental attributes that we used in benchmarking. See Figure \ref{fig:appendix:envs} for the environment details. All our experiments are run in the \texttt{omnisafe} module \cite{omnisafe}.

\subsubsection{Environment Details}
\label{appendix:envs}

Comprehensively, our experiments consist of eight tasks ranging from more superficial (Run and Circle tasks) to relatively more stochastic and sophisticated (Bottleneck and Grid tasks), each training different robots. They come from three well-known safe RL benchmark environments, Safe MuJoCo, Bullet-Safety-Gym, and Safety-Gym. For agents maneuvering on a two-dimensional plane, the cost is calculated as $C(s, a)=\sqrt{v_{x}^{2}+v_{y}^{2}}$. For agents moving along a straight line, the cost is calculated as $C(s, a)=\left|v_{x}\right|$, where $v_{x}$ and $v_{y}$ are the velocities of the agent in the $\mathrm{x}$ and $\mathrm{y}$ directions.

\paragraph{Circle}
This environment is inspired by \cite{achiam2017constrained}. Reward is maximized by moving along a circle of radius $d$:
\begin{equation*}
    R = \frac{v^\mathrm{T}[-y,x]}{1 + \big| \sqrt{x^2+y^2} - d\big|},
\end{equation*}
but the safety region $x_{\mathrm{lim}}$ is smaller than the radius $d : C = \mathbf{1}[x > x_{\mathrm{lim}}]$.

\paragraph{Navigation}. 
This environment is inspired by \cite{ray2019benchmarking}. Reward is maximized by getting close to the destination $R = \mathrm{Dist}(target,s_{t-1}) - \mathrm{Dist}(target,s_{t}),$
but it yields a cost of +1 when the agent hits the hazard or the pillar. The two different types of cost functions are returned separately and have different thresholds. In out setting, $d_1 = 25$ for the hazard constraint and $d_2 = 20$ for the pillar constraint.  

Since the main goal in MuJoCo is to train the robot to locomote on the plane, we call it the "Run" task in our article. Our chosen two robots are the relatively complex types in MuJoCo: Ant and Humanoid. \textbf{OpenAI Gym} is open source at https://github.com/openai/gym, and has a documentation at https://www.gymlibrary.ml/. \textbf{Bullet Safety Gym}. The implementation of the Circle task comes from Bullet-safety-Gym (Gronauer 2022), which Stooke, Achiam, and Abbeel (2020) first proposed. The reward is dense and increases by the agent's velocity and the proximity to the boundary of the circle. Costs are received when the agent leaves the safety zone defined by the two yellow boundaries. The environment is open source at https://github.com/SvenGronauer/Bullet-Safety-Gym.

\textbf{Safety Gym}. The remaining two tasks, Goal and Button, are from Safety-Gym (Ray, Achiam, and Amodei 2019). Compared to Run and Circle tasks, they are more stochastic and sophisticated in that agents are challenged to maximize the return while satisfying the constraints.

The environment is open source at https://github.com/openai/safety-gym, and readers can see OpenAI's blog at https://openai.com/blog/safety-gym/ for more details.

\subsection{Agents}
For single-constraint scenarios, Point agent is a 2D mass point($A \subseteq  \mathbb{R}^2$) and Ant is an quadruped robot($A \subseteq  \mathbb{R}^8$). For the multi-constraint scenario which is modified from OpenAI SafetyGym \cite{ray2019benchmarking}, $S \subseteq \mathbb{R}^{28 + 16\cdot m}$  where $m$ is the number of pseudo-radar (one for each type of obstacles and we set two different types of obstacles in the Navigation task) and $A \subseteq \mathbb{R}^2$ for a mass point or a wheeled car.

\subsubsection{Experimental Details}
\label{appendix:chapter2_experimental_details}

To be fair in comparison, the proposed \texttt{e-COP} algorithm and FOCOPS \cite{zhang2020first} are implemented with same rules and tricks on the code-base of \cite{ray2019benchmarking}.

\subsubsection{Hyperparameters}

{
\renewcommand{\arraystretch}{1.15}
\begin{table*}
\centering
\fontsize{7}{11}\selectfont
\begin{tabular}{cccccccccc}
\hline Hyperparameter & APPO & PDO & FOCOPS & CPPO-PID & IPO & P3O & CPO & TRPO-L & PCPO \\
\hline Actor Net layers & $(32,32)$ & $(32,32)$ & $(32,32)$ & $(32,32)$ & $(32,32)$ & $(32,32)$ & $(32,32)$ & $(32,32)$ & $(32,32)$ \\
Critic Net layers & $(32,32)$ & $(32,32)$ & $(32,32)$ & $(32,32)$ & $(32,32)$ & $(32,32)$ & $(32,32)$ & $(32,32)$ & $(32,32)$ \\
Activation & tanh & tanh & tanh & tanh & tanh & tanh & tanh & tanh & tanh \\
Initial log std & 0.5 & 0.5 & 0.5 & 0.5 & 0.5 & 0.5 & 0.5 & 0.5 & 0.5 \\
Discount $\gamma$ & 0.99 & 0.95 & 0.995 & 0.995 & 0.99 & 0.99 & 0.99 & 0.99 & 0.99 \\
Policy lr & $3 \mathrm{e}-4$ & $3 \mathrm{e}-4$ & $3 \mathrm{e}-4$ & $3 \mathrm{e}-4$ & $3 \mathrm{e}-4$ & $3 \mathrm{e}-4$ & $3 \mathrm{e}-4$ & $3 \mathrm{e}-4$ & $3 \mathrm{e}-4$ \\
Critic Net lr & $1 \mathrm{e}-3$ & $1 \mathrm{e}-3$ & $1 \mathrm{e}-3$ & $1 \mathrm{e}-3$ & $1 \mathrm{e}-3$ & $1 \mathrm{e}-3$ & $1 \mathrm{e}-3$ & $1 \mathrm{e}-3$ & $1 \mathrm{e}-3$ \\
No. of episodes & 500 & 500 & 500 & 500 & 500 & 500 & 500 & 500 & 500 \\
Steps per epochs & 300 & 300 & 300 & 300 & 300 & 300 & 300 & 300 & 300 \\
Target KL & 0.01 & 0.01 & 0.01 & 0.01 & 0.01 & 0.01 & 0.01 & 0.01 & 0.01 \\
KL early stop & True & True & True & True & True & True & False & False & False \\
Line Search Times & N/A & N/A & N/A & N/A & N/A & N/A & 25 & 25 & 25 \\
Line Search Decay & N/A & N/A & N/A & N/A & N/A & N/A & 0.8 & 0.8 & 0.8 \\
Proximal clip & 0.2 & 0.2 & 0.2 & 0.2 & 0.2 & 0.2 & N/A & N/A & N/A \\
Max horizon & 200 & 200 & 200 & 200 & 200 & 200 & 200 & 200 & 200 \\
\hline
\end{tabular}
\caption{Hyperparameters used for each baseline.}
\label{table:hyperparams}
\end{table*}
}

Table \ref{table:hyperparams} shows the hyperparameters of baseline algorithms.

\subsubsection{Runtime Environment}

All experiments were implemented in Pytorch 1.7 .0 with CUDA 11.0 and conducted on an Ubuntu 20.04.2 LTS with 8 CPU cores (AMD Ryzen Threadripper PRO 3975WX 8-Coresz), 127G memory and 2 GPU cards (NVIDIA GeForce RTX 4060 Ti Cards).

\subsubsection{Robustness to Cost Thresholds}
\label{app:4.2}

\begin{figure*}[ht]
\captionsetup[subfigure]{justification=centering}
  \centering
  \includegraphics[width=0.35\linewidth] {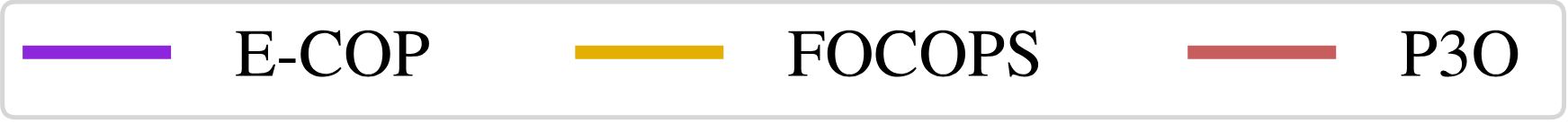}   \vspace*{-\baselineskip}

    \subfloat[Humanoid Rewards]{
        \includegraphics[height=0.125\linewidth, width=0.23\linewidth]{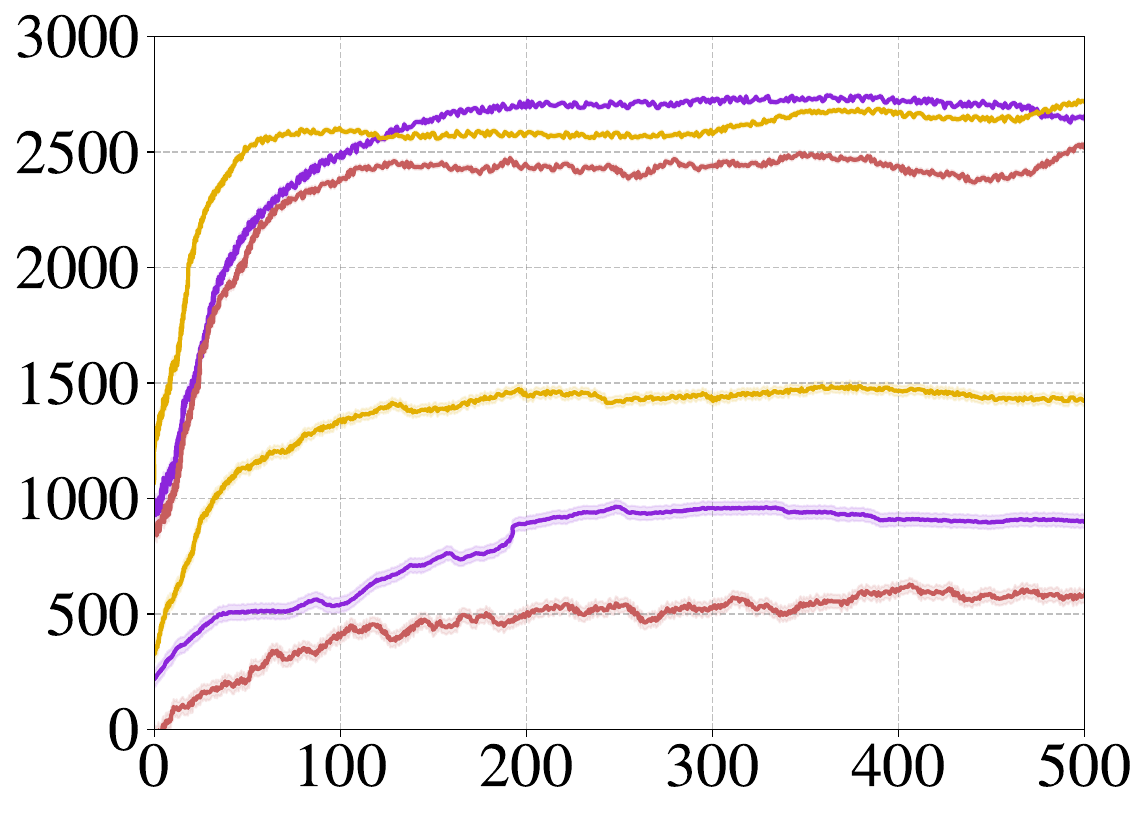}
        \label{fig:general_humanoid_reward}
    }
    \subfloat[Humanoid Costs]{
        \includegraphics[height=0.125\linewidth, width=0.23\linewidth]{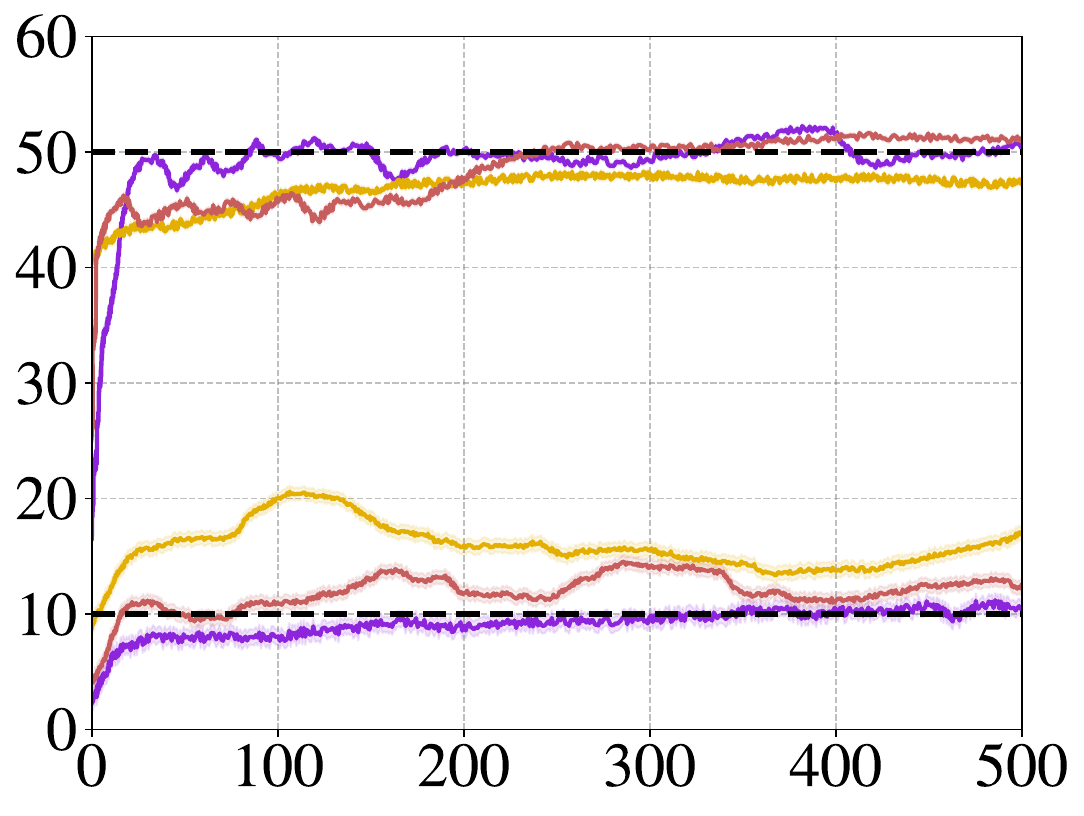}
        \label{fig:general_humanoid_cost}
    }
    \subfloat[Point Circle Rewards]{
        \includegraphics[height=0.125\linewidth, width=0.23\linewidth]{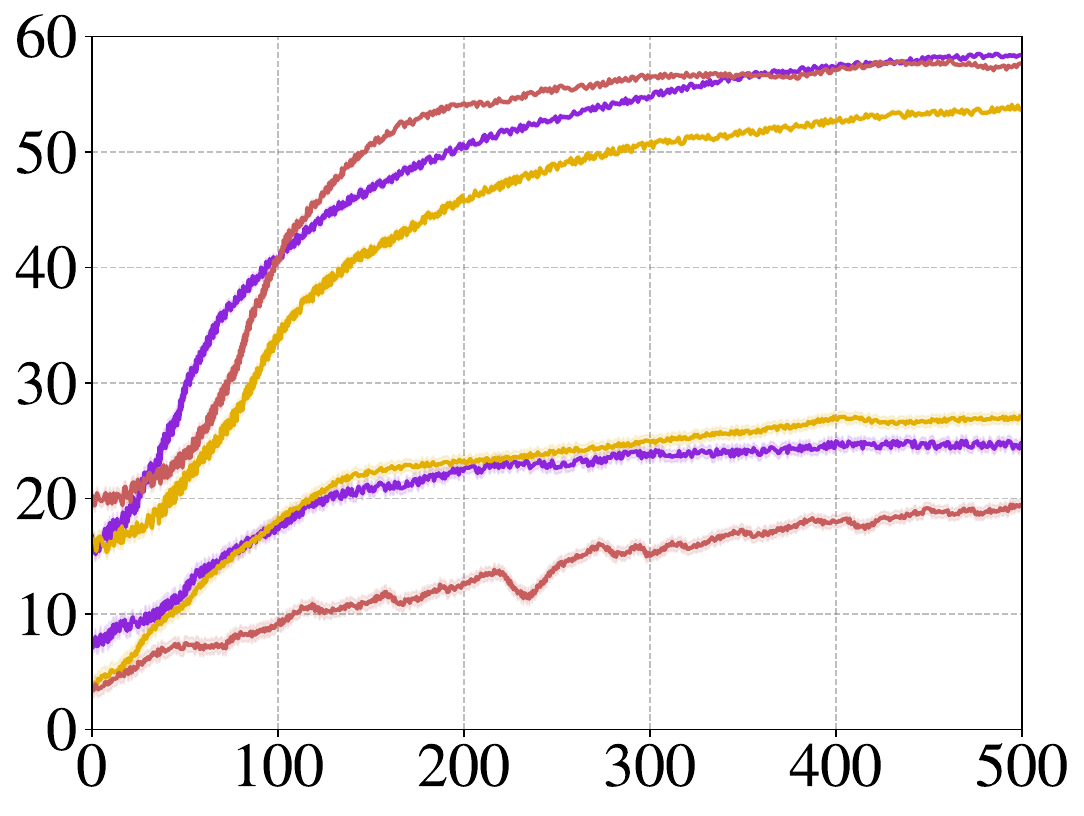}
        \label{fig:general_pointcircle_reward}
    }
    \subfloat[Point Circle Costs]{
        \includegraphics[height=0.125\linewidth, width=0.23\linewidth]{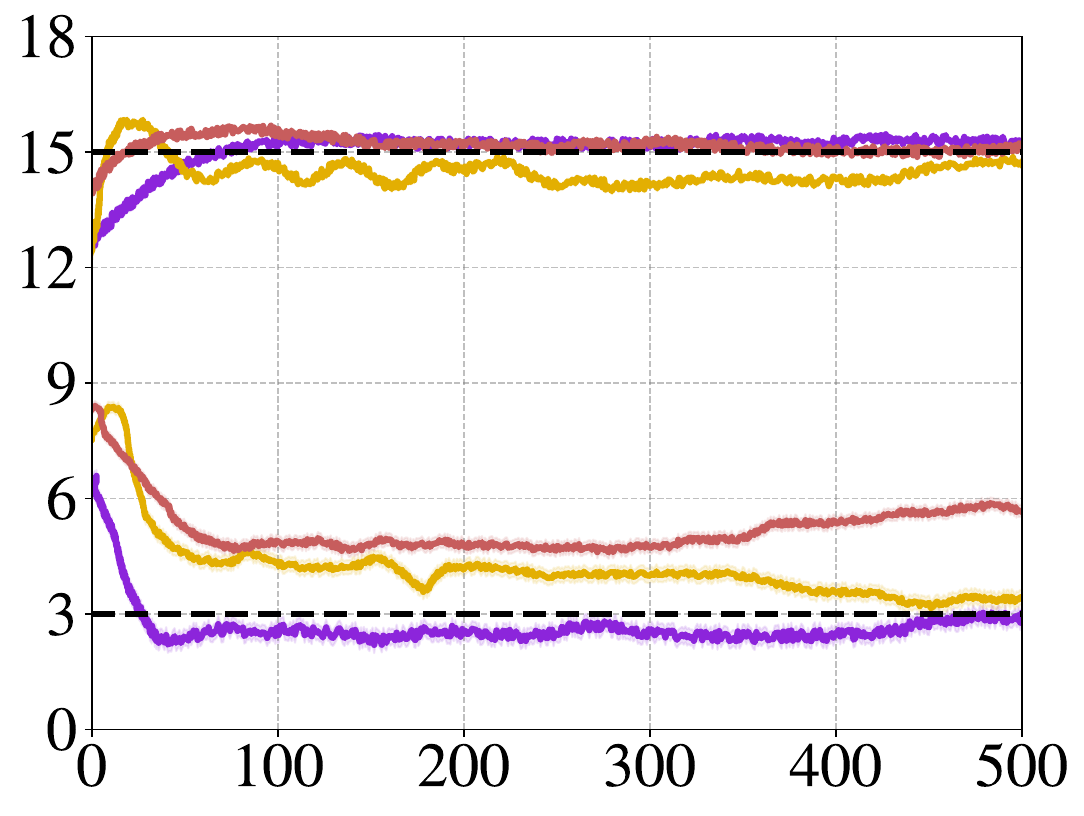}
        \label{fig:general_pointcircle_cost}
    }
  \caption{\small Cumulative episodic rewards and costs of baselines in two environments with two different constraint cost thresholds: 10 and 50 in Humanoid, and 3 and 15 in Point Circle. The hyperparameters are tuned at constraint limit of 20 in Humanoid and 10 in Point Circle.}
  \label{fig:general_rewards_costs_three}
\end{figure*}

We conducted a set of experiments wherein we study how \texttt{e-COP} effectively adapts to different cost thresholds. For this, we use a pre-trained \texttt{e-COP} agent, which is trained with a particular cost threshold in an environment, and test its performance on different cost thresholds within the same environment. Figure \ref{fig:general_rewards_costs_three} illustrates the training curves of these pre-trained agents, and we see that while \texttt{e-COP} can generalize well across different cost thresholds, other baseline algorithms may require further tuning to accommodate different constraint thresholds. 

\chapter{Hybrid Bandit Learning with Preferences}
\label{chapter:warmPref-PS}

\paragraph{Abstract}. Reinforcement Learning with Human Feedback (RLHF) is at the core of fine-tuning methods for generative AI models for language and images. Such feedback is often sought as preference feedback from human raters, as opposed to eliciting scores since the latter tends to be noisy. On the other hand, RL theory and algorithms predominantly assume that a reward feedback is available. In particular, approaches for online learning that can be helpful in adaptive data collection via active learning cannot incorporate offline preference data. In this paper, we adopt a finite-armed linear bandit model as a prototypical model of online learning. We consider an offline preference dataset to be available generated by an rater of unknown `competence'. We propose $\mathsf{warmPref-PS}$, a posterior sampling algorithm for online learning that can be warm-started with an offline dataset with noisy preference feedback. We show that by modeling the `competence' of the rater that generated it, we are able to use such a dataset most effectively. We support our claims with novel theoretical analysis of its Bayesian regret, as well as, extensive empirical evaluation of an approximate loss function that optimizes for infinitely many arms, and performs substantially better than baselines.

\section{Introduction}
\label{sec:chapter3_introduction}

\textcolor{black}{
In the development of generative AI models for language and image generation, it has proven quite effective to first `pretrain' with a very large \emph{offline} dataset followed by \emph{online} reinforcement learning (RL)-based `fine-tuning' with small amounts of high quality Human Feedback (HF) data to improve alignment with human preferences. Although preference based HF data are less noisy and easier to aggregate over multiple raters, absolute score based HF data are generally more informative than relative preferences, and designing mechanisms that find optimal tradeoffs between these different feedback modalities is critical to scaling RLHF. 
}

\textcolor{black}{
In practice today, there is already a lot of \emph{offline} preference data available to the models. These preference data are generated from batches sent to human annotators to provide preferences on. However, for task specific online finetuning, reward models (called `AutoRaters' \citep{anil2023palm}) are used for active learning. This is because it is expensive to do active learning with human raters in an online manner. The reward models are typically trained on these preference datasets in an offline manner, and are used to provide reward feedback in the online phase \citep{achiam2023gpt, anil2023palm}. 
The setting of offline preferences and online numerical rewards is also applicable to the case where a foundational model aligned to general human preferences from an initial \emph{offline} dataset needs to be rapidly personalized to the idiosyncratic \emph{online} preferences of a particular user. We approach the challenge of minimizing \emph{this} online learning. While it is trivial that collecting additional data with online finetuning will improve performance, \emph{how} to effectively combine preference and numerical reward learning is highly nontrivial. Hence, in this setting where the online ratings are absolute scores, we propose a simple Bayesian algorithm for online learning that incorporates learning from an offline preference dataset. We note here that our problem formulation below is motivated by the practical relevance discussed above, and, to the best of our knowledge, no other work that formalizes and analyses this setting exists.
}

\textcolor{black}{
To formalize the practical relevance, we adopt a finite-armed linear bandit model, with arms corresponding to different generated model outputs, with indicated rater preferences available offline before starting the online phase when absolute reward scores from a user become available. To efficiently learn the optimal arm selection strategy, we propose $\mathsf{warmPref-PS}$, a posterior sampling-based Bayesian algorithm that naturally incorporates offline preference data and online reward feedback, and minimizes Bayesian regret. 
}

\begin{wrapfigure}{r}{0.45\textwidth}
\vspace{-0.5cm}
\centering
\includegraphics[height=0.33\textwidth, width=0.44\textwidth]{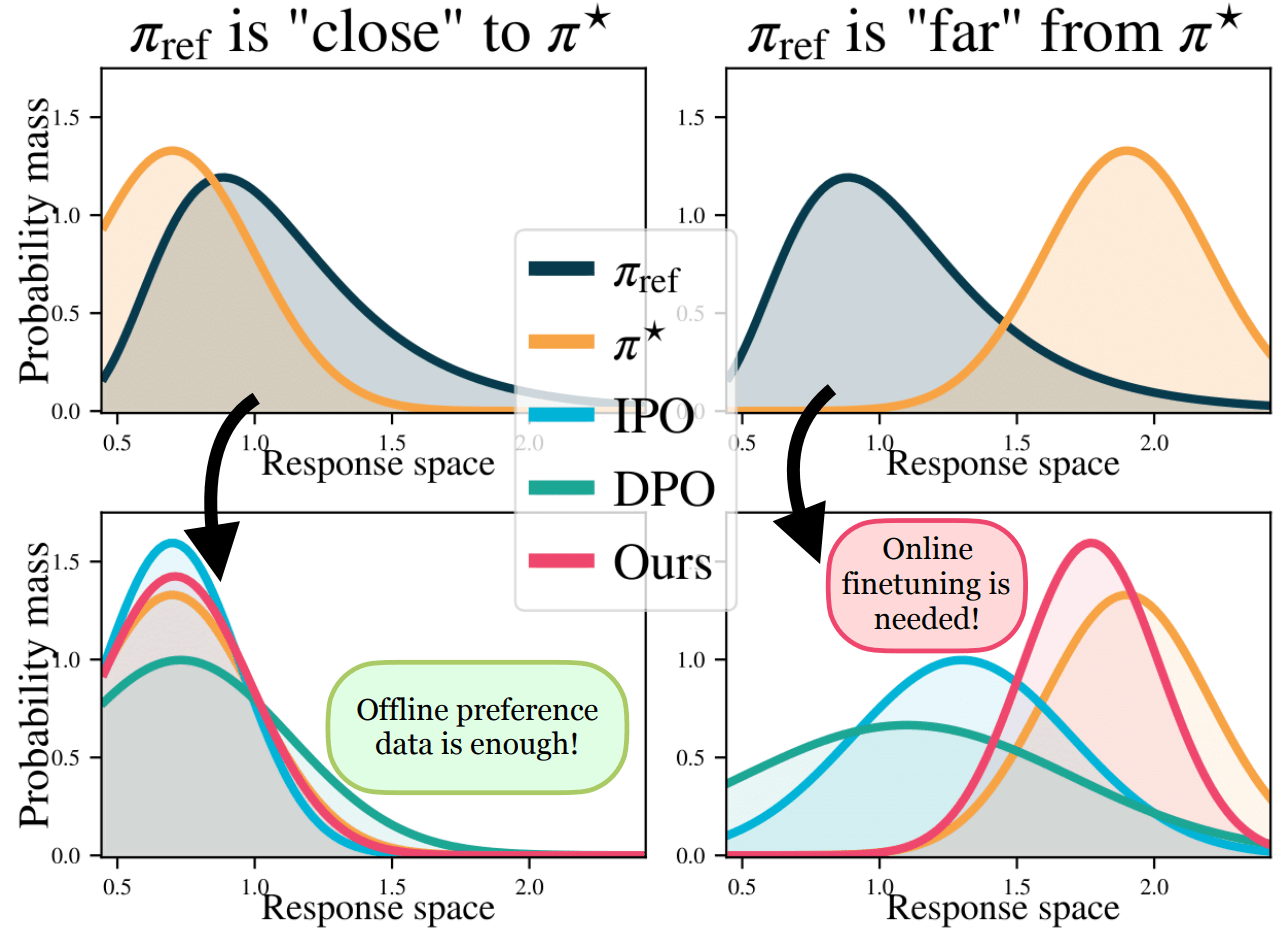}
\caption{$\mathsf{warmPref-PS}$ (ours) comparison with IPO \citep{ipo} and DPO \citep{rafailov2024direct}. When divergence between $\pi_{\mathsf{ref}}$ and $\pi^{\star}$ is large, on-policy online learning is needed for best results.}
\label{fig:rlhf_needed}
\vspace{-0.2cm}
\end{wrapfigure}

\textcolor{black}{
\noindent\textbf{Relevance to RLHF.} Since RLHF can be modeled as a bandit problem with context-output pairs \citep{ipo}, our problem setting sits at the intersection of offline preference learning and online reward-based fine-tuning in RLHF. \cite{tajwar2024preference} shows that when the reward optimum lies in low-probability regions of the reference policy $\pi_{\mathsf{ref}}$ (i.e. KL divergence between optimal policy $\pi^{\star}$ and $\pi_{\mathsf{ref}}$ is large) , on-policy sampling during online fine-tuning becomes crucial. This insight translates directly to our bandit setting, where maintaining a posterior over the reward function and updating via online sampling yields significantly improved performance compared to purely offline approaches, as seen in Figure \ref{fig:rlhf_needed}. Further, \cite{zhang2024reward} introduces a data relabeling scheme that augments offline binary preferences with explicit numerical reward values, avoiding “unlearning” of rejected yet high-quality outputs. This boosts generalization by better utilizing the full response space, a principle shown in our reward-based online phase, which outperforms standard DPO \citep{rafailov2024direct} on multiple benchmarks. \cite{tu2024online, bai2025online} address the scarcity of real-time human feedback in online phases by using LLMs to generate self-augmented preferences. While we take a different route (i.e. direct reward-based online updates), both approaches highlight the same need: augmenting offline preference datasets with online learning.
}

\noindent\textbf{Contributions of this paper.} \textcolor{black}{(i) We present the first online learning algorithm that incorporates \textit{offline preference data} into online learning, even when it comes from a subpar expert. The key is that the algorithm is able to model and learn how `competent' the expert is (with respect to the optimal policy). (ii) While our proposed algorithm is a natural extension of posterior sampling, it requires significantly different theoretical analysis due to the offline preference data (see Lemma \ref{lemma:priordependentinformationset} and Theorem \ref{th:warmtsregretbound}). We provide novel theoretical guarantees on the minimum size of the offline dataset needed for it to allow learning of the optimal action. We also provide an upper bound on the algorithm's Bayesian regret that reveals the dependence of the offline dataset size, and the expert's `competence'. (iii) We propose a practical version of our $\mathsf{warmPref-PS}$ algorithm , called Bootstrapped $\mathsf{warmPref-PS}$, that is computationally tractable for an infinitely armed bandit environment, and establish its superior empirical performance with regard to baselines.}

\noindent\textbf{Related work.} 
There is substantial literature on online learning for bandit models in various settings - finite-armed or linear, stochastic or adversarial, non-contextual or contextual models, etc. \citep{lattimore2020bandit}. There is recent literature on  utilizing offline data to improve learning regret during the online phase but these approaches either do not incorporate offline preference data \citep{shivaswamy12, bouneffouf2019optimal, zhang19b, banerjee2022artificial, agrawal2023optimal}, or solve the best arm identification problem which focuses on pure exploration \citep{agrawal2023optimal}. Furthermore, the quality of the offline data is not accounted for, which usually results in only a marginal regret reduction even while warm-starting with offline data. Ranking, comparison or preference feedback is considered in dueling bandit models \citep{dudik2015contextual,wu2016double,yan2022human,bengs2021preference,szorenyi2015online, agnihotri2023average, 9030307, Hu_2022_CVPR} but it is akin to active learning from preferences \citep{ailon2011active} without incorporating a \emph{given, fixed} offline preference dataset. Another set of works for the contextual bandit setting exist \citep{saha2021optimal, bengs2022stochastic, li2024feel}, however they cannot combine learning from rewards \emph{and} preferences. The importance of offline dataset quality in  imitation learning  was first investigated in \cite{beliaev22a}. Later, \cite{hao2023leveraging} introduced an algorithm which leveraged offline reward feedback to warm start the online phase. While the algorithm uses offline reward feedback data, it cannot incorporate preference feedback as we do in this paper. Since, incorporation of preference feedback is nontrivial, the regret analysis techniques (presented in Appendix \ref{sec:appendix}) are entirely different than for the case when reward feedback is available. \textcolor{black}{To our best knowledge, ours is the first online bandit learning algorithm that can incorporate offline preference data.}

\section{Preliminaries}
\label{sec:preliminaries}

We model unknown quantities as random variables defined on a common probability space $(\Omega, \Fcal, \Pbb)$. Now, consider a stochastic $K$-armed linear bandit problem with a set of actions, $\mathcal{A}=\{a_0, \dots, a_K \} \subseteq \Rbb^d$. The environment is characterized by a random vector $\theta \in \Rbb^d$, with a prior distribution $\nu_0$. At time step $t$, the agent chooses an action $A_t \in \mathcal{A}$ and receives a reward $R_t$: 
\vspace{-0.1cm}
$$
R_t=\left\langle A_t, \theta\right\rangle+\eta_t,
$$

where $\eta_t \sim \mathcal{N}\left(0, \sigma^2\right)$ are i.i.d. sampled at each time step. For RLHF applications, the rewards might correspond to absolute score feedback by an individual rater on outputs (i.e., actions) generated by a foundational model. The agent's objective is to maximize $\sum_{t=1}^{T} \Ebb[R_t]$, the expected total reward over horizon $T$, where the expectation is over the algorithm's decisions and the randomness in the environment. In addition, we also have an initial \emph{offline preference} dataset $\mathcal{D}_0$, which is generated by human raters \emph{with limited competence}. This offline dataset is a sequence of tuples of the form 
$
\mathcal{D}_0=\left((\bar{A}_n^{(0)}, \bar{A}_n^{(1)}, Y_n)\right)_{n \in [N]},
$ 
where $[N]:= [1, 2, \ldots, N]$, $\bar{A}_n^{(0)}, \bar{A}_n^{(1)} \in \mathcal{A}$ are two actions, and $Y_n \in\{0,1\}$ indicates the rater's preference. In particular, $Y_n=0$ if the rater prefers action $\bar{A}_n^{(0)}$ to $\bar{A}_n^{(1)}$, and $Y_n=1$ otherwise. In addition to the dataset size $N$, we characterize the offline dataset by: (i) an action sampling distribution $\mu$, where $\bar{A}_n^{(0)}$ and $\bar{A}_n^{(1)}$ are i.i.d. sampled from $\mu$; and (ii) assuming that given $\bar{A}^{(0)}$ and $\bar{A}^{(1)}$, the rater follows a \textit{noisy} Bradley-Terry model \citep{bradleyterry1952} and chooses $Y=0$ (i.e., ranks action $\bar{A}_n^{(0)}$ above $\bar{A}_n^{(1)}$)  with probability
\small
    \begin{equation}
       \label{eq:rater_preference_dist}
           { P\big(Y=0 \Given \bar{A}^{(0)}, \bar{A}^{(1)} \; ; \; \vartheta \big)} = \frac{\exp \left(\beta\left\langle\bar{A}^{(0)}, \vartheta\right\rangle\right)}{\exp \left(\beta\left\langle\bar{A}^{(0)}, \vartheta\right\rangle\right)+\exp \left(\beta\left\langle\bar{A}^{(1)}, \vartheta\right\rangle\right)}
    \end{equation}
\normalsize
where the parameter $\beta \geq 0$ is a measure of the \emph{deliberateness} of the rater's decision: $\beta=0$ means the rater's decisions are uniformly random, whereas as $\beta \to \infty$, its decisions pick the maximum of the reward under the two actions. The parameter $\vartheta \sim$ $N\left(\theta, \mathbf{I}_{d} / \lambda^2\right)$ ($\mathbf{I}_{d}$ is a $d \times d$ identity matrix) is the rater's estimate of the true reward model and the parameter $\lambda$ is a measure of its \emph{knowledgeability} of it, i.e., as $\lambda \to \infty$, $\vartheta \to \theta$. Alternatively, in the adaptation scenario where the online learning phase is used to align with the desires of a single user, the knowledgeability parameter can be seen as controlling the degree of alignment between the user and the general population from which preferences are aggregated. Denoting the online dataset collected by time $t$ as $\Hcal_t = \{ (A_t, R_t) \}_{s=1}^{t}$, we have $\Dcal_{t} = \Dcal_{0} \cup \Hcal_{t}$, the entire dataset available at time $t$.

\noindent\textbf{Notion of Regret.} 
Given an offline preference dataset $\Dcal_0$ and an arbitrary environment  $\theta$, the Bayesian Regret for $T$ rounds is given by:
\begin{equation}
    \BR_{T}(\pi) := \sum_{t=1}^{T} \Ebb_{\pi, \theta, \Dcal_{0}} \bigg[ \langle A^{\star}, \theta \rangle - R_{t} \bigg],
\label{eq:bayesianregret}
\end{equation}
where expectation is taken over $\left( \pi, \theta, \Dcal_0 \right)$, and $A^{\star}(\theta) = \argmax_{a \in \Acal} \langle a, \theta \rangle$ (the optimal action for environment $\theta$), and $\pi$ is a policy that maps past observations $\Dcal_t$ to a distribution over actions. Here, we assume that the prior distribution over the environment $\theta$ is a Gaussian distribution $\nu_0 = \mathcal{N}(\mu_0, \Sigma_0)$. To distinguish from the ``informed prior" learned from $\mathcal{D}_0$, we call $\nu_0$ as the uninformed prior.
%
%
The goal then is to learn a policy $\pi$ that minimizes the Bayesian regret in Equation \eqref{eq:bayesianregret}.



\section{Introducing the {Preference-Warmed Posterior Sampling} Algorithm}
\label{sec:introducealgo}

The online learning problem for Bayesian regret-minimization (in Equation \eqref{eq:bayesianregret}) that we have introduced in the previous section has two novel elements: an offline dataset to begin with, and such a dataset having only (noisy) preference feedback generated by a human rater with limited capacity, instead of reward feedback. 
We adopt the posterior sampling (PS) approach to designing online bandit learning algorithms since they have a natural structure, and also because they usually offer superior performance as compared to optimism-based algorithms \citep{russo2018tutorial}. Thus, we introduce $\mathsf{warmPref-PS}$ (as Algorithm \ref{alg:Prefwarm-PS}), a (Bayesian) posterior sampling algorithm warm-started with offline preference data. As we will see below, most of the steps are common with any meta-PS algorithm.


\begin{enumerate}[leftmargin=*]
    \item \textbf{Constructing an informed prior.} Using the offline dataset $\Dcal_0$, construct an informed prior $\nu_1$,
\begin{equation}
  \label{eq:informed_prior}
\resizebox{0.85\linewidth}{!}{$
  \begin{aligned}
      \nu_{1}(\theta) := P(\theta \given \Dcal_0) \; \propto \;  P(\Dcal_0 \, | \, \theta) \cdot \nu_0 (\theta) \; \propto \; \bigg[ \prod_{n=1}^{N} P( Y_n \given \Anzero, \Anone, \theta) \cdot P(\Anzero) \cdot P(\Anone) \bigg] \cdot \nu_0 (\theta)
  \end{aligned}
$}
\end{equation}

, where $\nu_0$ is the uninformed prior and  the second step follows from Equation \eqref{eq:rater_preference_dist} and \textcolor{black}{the fact that in the context of RLHF, outputs (actions) are conditionally independent given the prompt}. It is worth emphasizing here that the actions in the offline dataset  carry information about the environment through the term $P(\bar{A}_n^{(\cdot)} \given \theta)$, which incorporates information about the expert's policy, and thus improves the informativeness of the prior distribution.   

\item \textbf{Online decision making.} At time  $t$, get sample $\hat{\theta}_{t} \sim \nu_{t}$, take action $A_t = \argmax_{a \in A} \langle a, \hat{\theta}_t \rangle$, observe reward $R_t$, and update the dataset as $\Dcal_{t} = \Dcal_{t-1} \cup \{(A_t, R_t)\}$.

\item \textbf{Updating knowledge of the environment.}  At time $t$, the environment parameter $\theta$ will have distribution $\nu_t (\theta)$, and we update our posterior as, 
\begin{equation}
  \label{eq:posterior_update_theta}
\resizebox{.92\linewidth}{!}{$
  \begin{aligned}
     \nu_{t+1}(\theta \given \Dcal_{t}) \; \propto \; P(\{(A_t, R_t)\} \given \Dcal_{t-1}, \theta) \cdot \nu_{t}(\theta\given \Dcal_{t-1}) = \; P(R_{t} \given A_{t}, \theta) \cdot P(A_{t} \given \Dcal_{t-1}) \cdot \nu_{t}(\theta\given \Dcal_{t-1}),
  \end{aligned}
$}
\end{equation}
where $P(R_{t} \given A_{t}, \Dcal_{t-1}, \theta) = P(R_{t} \given A_{t}, \theta)$ and $P(A_{t} \given \Dcal_{t-1}, \theta) = P(A_{t} \given \Dcal_{t-1})$. The posterior of $\vartheta$ also changes, and hence, $\vartheta_{t+1} \sim \Ncal (\theta_{t+1}, \Ibf/\lambda^{2})$ with $\theta_{t+1} \sim \nu_{t+1}(\theta).$ We regard $\beta$ to be a known parameter. We relax this in Section \ref{sec:chapter3_empirical}.


      \begin{algorithm}[t]
        \caption{Preference-Warmed Posterior Sampling ($\mathsf{warmPref-PS}$)}
        \begin{algorithmic}[1]
          \STATE {\bfseries Input:} Action set $\Acal$, uninformed prior $\nu_0$ over environment $\theta$, offline preference dataset $\Dcal_{0}$.
          \STATE Construct informed prior ${\nu}_{1}$ from $\Dcal_{0}$ using Equation \eqref{eq:informed_prior}.
          \FOR{$t = 1, 2, \dots, T$}
          \STATE Sample $\hat{\theta}_{t} \sim {\nu}_t$ to take action $A_{t} = \argmax_{a \in \Acal} a^{T} \hat{\theta}_{t}$ and receive reward $R_{t}$.
          \STATE Update dataset $\Dcal_{t}$ and posterior $\nu_{t+1} \leftarrow \Pbb \left( \cdot \given \Dcal_{t} \right)$ using Equation \eqref{eq:posterior_update_theta}.
          \ENDFOR
        \end{algorithmic}
    \label{alg:Prefwarm-PS}
      \end{algorithm}

\end{enumerate}

\begin{remark}
In Equation \eqref{eq:informed_prior}, we construct an informed prior using the offline preference dataset. This step can be intractable. Similarly, the posterior update of Equation \eqref{eq:posterior_update_theta} is also usually intractable, unless the distributions we are working with have a conjugacy property. In which case, we resort to various approximations. In Section \ref{sec:practicalwarmPref-PS}, we present a practical version of this algorithm by introducing a loss function that approximates Steps 2 and 5 of Algorithm \ref{alg:Prefwarm-PS}. This loss function is independent of the size of the action space, and hence, is extendable to infinitely-many armed bandit settings as well. 


\end{remark}


\section{Analysis of warmPref-PS}
\label{sec:analysis}

We now present an analysis of the $\mathsf{warmPref-PS}$ algorithm in two steps. First, in Section~\ref{sec:samplecomplexity}, we present an ``informativeness" analysis of the offline preference data $\Dcal_0$, which establishes a sample complexity result for $\Dcal_0$ to be informative about the optimal action. Then, based on this result, we develop an upper bound on the Bayesian regret for $\mathsf{warmPref-PS}$ in Section~\ref{sec:regretbd}.


\subsection{Informativeness of Offline Preference Data}
\label{sec:samplecomplexity}

We first introduce the notion of \emph{informativeness} of the offline preference data, which characterizes how much information about the optimal action is provided by this offline preference dataset. 
Specifically, for purposes of analysis, we construct an `information' set $\UD \subseteq \Acal$ such that it contains the optimal action with high probability (see Appendix~\ref{appendix:twoactions}, \ref{appendix:multipleactionsinfinitebeta}, and \ref{appendix:multipleactionsfinitebeta} for details). This is useful for the analysis during the online phase; intuitively, during the online phase, $\mathsf{warmPref-PS}$ is expected to sample most actions from $\UD$.
%

\begin{definition}
Consider a random set $\UD \subseteq \Acal$ measurable with 
respect to $\Dcal_{0}$. For any $\epsilon \in [0,1]$, we say $\UD$ as $(1 - \epsilon)$-\textit{informative} if $P (A^{\star} \in \UD) \ge 1 - \epsilon$, i.e., it contains the optimal action with high probability.
\end{definition}

This information set $\UD$ has to be measurable with respect to $\Dcal_0$ (i.e.,  conditionally deterministic given $\Dcal_0$). Intuitively, the offline dataset $\Dcal_0$ is useful in determining the optimal action $A^{\star}$ if there exists a $\UD$ measurable to $\Dcal_0$ such that  (i) $\UD$ is $(1-\epsilon)$-informative, and (ii) $\mathbb{E}[\left| \UD \right|]$ is small. In other words, one can construct a $\UD$ based on $\Dcal_0$ such that $\UD$ has a small expected cardinality and contains $A^{\star}$ with high probability.
%
We first present a sample complexity result (i.e., how large the offline dataset size needs to be) on $\Dcal_0$ such that the set $\UD$ constructed in the appendix is $(1-\epsilon)$-informative. 
We start by studying the special case of the set $\UD$ being a singleton to elucidate its dependence on various parameters, but discussion in the next section will not require this assumption. The result below shows this dependence, i.e., how large does $\Dcal_{0}$ need to be such that $\mathsf{warmPref-PS}$ can infer $A^{\star}$ from it.


\begin{restatable}{theorem}{finalmultipleactionssamplecomplexity}
\label{theorem:finalmultipleactionssamplecomplexity} 
Let the action set $\Acal$ have size $K$ with a sampling distribution $\mu$ such that $0 < \mu_{\text{min}} \leq \mu_{k} \leq \mu_{\text{max}} < 1, \; \forall k \in [K]$. Given some $\epsilon \in (0, 1)$ and finite $\beta < \infty$, let $\lambda \to \infty$. Then, the singleton set $\UD = \{A^{\star}\}$ is $(1-\epsilon)$-informative if
\small
\begin{align}
& N > N_0: = \frac{\ln K + (k_{\text{max}}-1)\ln\ln K}{\mu_{\text{min}}^{2} \epsilon}, \quad  \text{where}\\
\quad k_{\text{max}} &= \max_{i,j \in [K]} \frac{\ln \bigg( \big(\frac{2 K^{2}}{\epsilon}-1 \big) \big(\frac{1}{\Phi(x_{i,j})}-1 \big) \bigg)}{\beta \langle a_{i} - a_{j}, \theta_{0} \rangle} \; , \text{and} \; x_{i,j} = \frac{(a_{i} - a_{j})^T \mu_{0}}{\sqrt{\big(a_{i} - a_{j}\big)^T \Sigma_{0} \big(a_{i} - a_{j}\big)}}, \nonumber
\label{eq:samplecomplexityfinal}
\end{align}
\normalsize
and, $N$ is the size of the preference dataset and $\Phi(\cdot)$ is the standard Normal CDF.
\end{restatable}

The proof can be found in Appendix \ref{appendix:multipleactionsfinitebeta}. The above theorem provides a bound on the size of the offline preference dataset $\Dcal_{0}$ needed to find a singleton information set $\UD$ containing the optimal action. To understand the above result, note that in the special case of $K=2$, we have the following.
\begin{corollary}
For an action set $\Acal = \{a_0, a_1 \}$,
any $\beta \in (0,\infty)$ and $\epsilon \in (0, 1)$, with $\lambda \to \infty$, if 
\begin{equation}
N > N_0:=\frac{\ln \bigg( \big(\frac{1}{\epsilon}-1 \big) \big(\frac{1}{\Phi(x)}-1 \big) \bigg)}{\beta \langle a_{0} - a_{1}, \theta_{0} \rangle}, \label{eq:twoactionssamplecomplexity}
\end{equation}
\text{where}~~$x := \frac{(a_{0} - a_{1})^T \mu_{0}}{\sqrt{(a_{0} - a_{1})^T \Sigma_{0} (a_{0} - a_{1})}}$,
then there exists a singleton $\UD$ that is $(1-\epsilon)$-informative.
\end{corollary}
See Appendix \ref{appendix:twoactions} and \ref{proof:twoactionssamplecomplexity} for proof. 
The above corollary reveals how the offline dataset size $N$ needed to infer the optimal action depends on the deliberateness parameter in the presence of noisy comparisons. We see that as $\beta \to \infty$, we have $N_0 \to 0$, i.e., we only need a single comparison. 

\subsection{Regret Bound}
\label{sec:regretbd}

We now first introduce an information theoretic result from the literature for Bayesian regret for a posterior sampling algorithm. The discussion below is for the general case where the information set $\UD$ is not necessarily a singleton.

\begin{restatable}{theorem}{botaoregretbound}
\label{th:generalizedregretbound} 
(\cite{hao2023leveraging}) For any $(1-\epsilon)$-informative set $\UD \subseteq \Acal$, the Bayesian Regret of any \texttt{PS} algorithm
can be upper bounded as:
$$
    \BR_T(\texttt{PS}) \leq \sqrt{T \Ebb[|\UD|] \ln(\Ebb[|\UD|]) + \epsilon \ln(K / \epsilon)} + C_{1}T\epsilon, 
$$
where $C_{1}$ is the bound on the expected reward range, i.e., $\Ebb[\max a^{T}\theta] - \Ebb[\min a^{T}\theta] \leq C_{1}$.
\end{restatable}



To apply the above theorem, we can construct the set $\UD$ in the following way: it contains all actions that  have been preferred to another action \emph{at least} once in the offline dataset $\Dcal_{0}$ and also includes any actions that do not appear in the dataset $\Dcal_{0}$ \footnote{Note that this construction is an algorithmic choice that integrates well with the posterior sampling style of algorithms. There can be other construction criteria of $\UD$ as well.}. Thus, $\UD$ can contain upto $K$ actions. First, let $\Delta := \ln(T\beta)/\beta \, , \, \alpha_{1}^{\Delta} := K \min(1,\Delta)$, and $\alpha_{2} := \lambda^{-1} \sqrt{2\ln(2d^{1/2}T)}$. Then, denote 

\begin{equation}
\resizebox{.7\linewidth}{!}{$
\begin{aligned}
\tilde{f}_1 &:=  \left(1 - \frac{1}{1 + \exp \left( \beta \big( \min(1, \Delta) + \alpha_{2} - \alpha_{1}^{\Delta} \big) \right)} \right)^{N} + (1 - \mu_{\text{min}})^{2N} \quad , \;  f_1 = \tilde{f}_1 + \frac{1}{T} \;\, , \\
f_{2} &:= \min  \left( \left(\alpha_{1}^{\Delta} \right)^2  + \frac{NK}{T \beta} \left( 1 + \exp \left( -\beta \alpha_{2} + \alpha_{1}^{\Delta} \right) \right)^{-N} + \frac{2}{T}  \; , \, K \right) \; .
\end{aligned}
$}
\end{equation}

The constants $(f_1,f_2)$ characterize the offline preference dataset in terms of its size $N$, and expert's competence parameters $\lambda$ and $\beta$. Then, we have the following guarantee on the informativeness and size of the set ~$\UD$.

\begin{restatable}{lemma}{priordependentinformationset}
\label{lemma:priordependentinformationset} 
If $\mu_{\text{min}} > 0$, then set $\UD$ constructed above is $(1-f_1)$-informative, and $
\mathbb E[|\UD|]\leq f_2\,.
$
\end{restatable}
\begin{proof}[Proof sketch] We upper bound the probability of the optimal action $A^{\star}$ not being in $\UD$ (the `error probability'), and by defining an event $\Ecal_{(n)}:=\{ \langle A^{\star}-a_{n}, \theta \rangle \leq \Delta \}$, where $\left(A^{\star}, a_{n} \right)$ is the action tuple in $\UD$ for some $n \in [N]$ and $\Delta \in \Rbb$. Then, we decompose this error probability based on the event $\Ecal_{(n)}$ and the sub-optimality gap of actions ($\Delta$). To bound the expected cardinality of $\UD$, we again decompose based on $\Ecal_{(n)}$ and use a Poisson approximation to bound the probability. See Appendix \ref{proof:priordependentinformationset} for the complete proof. 
\end{proof}

Here, $\alpha_{1}^{\Delta}$ and $\alpha_{2}$ are representative of the nature of the problem, and take into the consideration the information loss due to finite deliberateness and knowledgeability of the rater respectively. Both $\alpha_{1}^{\Delta}, \alpha_{2}\to 0$, as $\beta$ and $\lambda$ get large. The parameter $f_{1}$ captures the error probability of the optimal action not being in $\UD$, and decays exponentially as the size of the dataset increases. Note that we need $\mu_{\text{min}}$ and $\mu_{\text{max}} \in (0,1)$ to obtain a nonzero sampling probability over all the actions. 

Now using Lemma \ref{lemma:priordependentinformationset} in conjunction with the information theoretic upper bound in Theorem \ref{th:generalizedregretbound}, we obtain the following main result about the Bayesian regret of the $\mathsf{warmPref-PS}$ algorithm:

\begin{restatable}{theorem}{warmtsregretbound}
\label{th:warmtsregretbound} 
The Bayesian regret of the $\mathsf{warmPref-PS}$ algorithm can be bounded as
\begin{equation}
\resizebox{0.75\linewidth}{!}{$
\begin{aligned}
\BR_{T}(\pi_{\mathsf{warmPref-PS}}) \leq \underbrace{\sqrt{Tf_2\left( \ln(f_2) +  f_1\ln\left(K/f_1\right)\right)}}_{\text{main term}} \nonumber  + 2\sqrt{2\ln(K)}T \left(\tilde{f}_1 + \frac{1}{T} \right)
\end{aligned}
$}
\end{equation}
\end{restatable}

The proof can be found in Appendix \ref{appendix:regretanalysis}. Although the bound in Theorem \ref{th:warmtsregretbound} appears to be linear in $T$, in fact $\tilde{f}_1 < 1$, and as the dataset size $N \to \infty$, $\tilde{f}_1 \to 0$, implying that the second term behaves like a constant. Second, for the main term, as the deliberateness $\beta$ and dataset size $N$ increase, the information ratio ($f_{2}$) decreases exponentially and then the entropy part ($\ln(f_2) + f_1 \ln (K/f_1)$) decreases further until $f_2 = 1$.  Finally, note that as the preference dataset parameters $\lambda$ and $\beta$ get large, the main term in the regret bound above converges to 0. Thus, the algorithm has constant regret in the case where the offline dataset is very large and is from a near-optimal expert.

\textcolor{black}{
\begin{remark}
 Note  that $f_2 \leq K$, so $\UD$ cannot grow arbitrarily large. Second, the informativeness of the offline dataset depends on both the dataset size $N$ and its quality (measured by $\beta$). When both $N$ and $\beta$ go to infinity, our regret bound in Theorem \ref{th:warmtsregretbound} reduces to $\mathcal{O}(\sqrt{\ln(K) + \ln(T)})$, which is sublinear. Thus, our result shows that when the offline dataset has both high quality and large size, the per unit regret of the proposed algorithm is negligibly small. Finally, note that in the asymptotic sense this upper bound matches the lower bound in the classical linear bandit setting with no offline data and bandit feedback.
\end{remark}
\begin{remark}
    The current analysis has been done in the finite armed bandit setting for tractability reasons \textcolor{black}{and practical relevance to RLHF with finite vocabulary and context-output pair sizes}, and we leave the analysis of infinite-many armed bandit setting for future work. Nevertheless, in the next section we present an approximate loss function for the $\mathsf{warmPref-PS}$ algorithm that works for the infinite armed setting as well and performs substantially better than available baselines. 
\end{remark}
}

\section{A Practical Approximation of the warmPref-PS Algorithm}
\label{sec:practicalwarmPref-PS}

As mentioned before, the posterior update in Equation \eqref{eq:informed_prior} and Equation \eqref{eq:posterior_update_theta} lacks the conjugacy property due to the $P(A_{t} \given \Dcal_{t-1})$ term, and is hence intractable. However, it inspires us to design a practical algorithm in the manner of  well established Bayesian bootstrapping ideas \citep{osband2019deep} where a surrogate loss function is constructed with added noise, and is then optimized to obtain the Maximum A Posteriori (MAP) estimate. This provides a point estimate of the unknown parameters $(\theta,\vartheta)$, but due to the added noise can be viewed as a sample from an approximation to the posterior distribution.

\noindent \textbf{A surrogate loss function.}
We start with the MAP estimate problem for $(\theta, \vartheta)$ given the offline and online dataset $\Dcal_{t-1}$ at  time $t-1$. We show that this is equivalent to minimizing a particular surrogate loss function as described in the lemma below:


\begin{restatable}{lemma}{mapestimatelemma}
\label{th:mapestimatelemma} 
At time $t$, the MAP estimate of $(\theta, \vartheta)$ can be constructed by solving the following equivalent optimization problem: 
\small
\begin{equation}
\begin{aligned}
(\theta_{opt}, \vartheta_{opt})  &= \underset{\theta, \vartheta}{\argmax} \; P(\theta, \vartheta \, | \, \Dcal_{t-1})  \equiv \underset{\theta, \vartheta}{\argmin} \; \Lcal_{1}(\theta, \vartheta) +  \Lcal_{2}(\theta, \vartheta) +  \Lcal_{3}(\theta, \vartheta) \; , \\
\text{where}, & \qquad \Lcal_{1}(\theta, \vartheta)  := \frac{1}{2} \sum_{s=1}^{t-1} \big(R_s - \langle A_s , \theta \rangle \big)^{2}, \\
& \Lcal_{2}(\theta, \vartheta) := - \sum_{n=1}^{N} \beta \langle \bar{A}_n^{(Y_{n})} , \vartheta \rangle + \ln \bigg(e^{ \beta \langle \bar{A}_n^{(0)}, \vartheta \rangle} + e^{\beta \langle \bar{A}_n^{(1)}, \vartheta \rangle} \bigg),  \\
& \Lcal_{3}(\theta, \vartheta) := \frac{\lambda^2}{2} \norm{\theta - \vartheta}{2}{2} + \frac{1}{2} (\theta - \mu_{0})^{T} \Sigma_{0}^{-1} (\theta - \mu_{0}).
\end{aligned}
\label{eq:mapestimateproblem}
\end{equation}
\normalsize
\end{restatable}

See Appendix \ref{proof:mapestimatelemma} for proof. A close look at Equation \eqref{eq:mapestimateproblem} shows that $\Lcal_{1}$ captures the likelihood of the online rewards, $\Lcal_{2}$ captures the likelihood of preferences from the offline preference dataset $\Dcal_{0}$, and $\Lcal_{3}$ handles the prior distribution of $\theta$ and $\vartheta$. We could also regard $\beta$ to be unknown but that leads to a non-convex loss function. So, we estimate that separately in Section \ref{sec:empirical} and then plug it in Equation \eqref{eq:mapestimateproblem}. Minimizing the above loss function however, only yields a point estimate of $(\theta,\vartheta)$ that is deterministic given the dataset $\Dcal_{t-1}$. 

\begin{algorithm}[!t]
   \caption{{Bootstrapped} $\mathsf{warmPref-PS}$}
   \label{alg:practical_Prefwarm-PS}
\begin{algorithmic}[1]
   \STATE {\bfseries Input:} {\small Horizon $T$, offline preference dataset $\Dcal_0$, action set $\Acal$, knowledgeability $\lambda$, deliberateness $\beta$.}
	 \FOR{$t = 1,2,\dots,T$} 
    	 \STATE Sample a set of perturbations $\Pcal_{t} = \{(\zeta_{s}, \omega_{n}, \theta', \vartheta')\}$.
    	 \STATE Solve Equation \eqref{eq:final_surrogate_perturbed_loss} using this set $\Pcal_{t}$ to find $(\hat{\theta}_{t}, \hat{\vartheta}_{t})$.
         \STATE Take action $A_t = \argmax_{a \in \Acal} \, \langle a, \hat{\theta}_{t} \rangle$, receive reward $R_t$, and update $\Dcal_{t} \leftarrow \Dcal_{t-1} \cup \{(A_{t}, R_{t})\}$.
    \ENDFOR
\end{algorithmic}
\end{algorithm}

\noindent\textbf{Perturbing the loss function.}
As mentioned above, the idea now is to \emph{perturb} the loss function in Equation \eqref{eq:mapestimateproblem} with some noise, so that the MAP point estimates we get from this perturbed surrogate loss function serve as \emph{samples} from a distribution that approximates the true posterior \citep{osband2019deep, NIPS2017_49ad23d1, qin2022analysis, dwaracherla2022ensembles}. To that end, we propose a perturbation of the `online' loss function $\Lcal_1(\cdot)$ by additive Gaussian noise, of the `offline' loss function $\Lcal_2(\cdot)$ by multiplicative random weights, and of the `prior' loss function $\Lcal_3(\cdot)$ by random samples from the prior distribution as follows:
(i) \textit{Online perturbation.} Let $\zeta_{s} \sim \Ncal(0,1)$, all i.i.d. Then, the perturbed $\Lcal_{1}(\cdot)$ becomes $ \Lcal_{1}'(\theta, \vartheta) =  \frac{1}{2} \sum_{s=1}^{t-1} \big(R_s + \zeta_{s} - \langle A_s , \theta \rangle \big)^{2}$, (ii) \textit{Offline perturbation.} Let $\omega_{n} \sim \mathrm{Bern}(0.5)$, all i.i.d. Then, the perturbed $\Lcal_{2}(\cdot)$ becomes $ \Lcal_{2}'(\theta, \vartheta) =  - \sum_{n=1}^{N} \omega_{n} \bigg[ \beta \langle \bar{A}_n^{(Y_{n})} , \vartheta \rangle + \ln \bigg(e^{ \beta \langle \bar{A}_n^{(0)}, \vartheta \rangle} + e^{\beta \langle \bar{A}_n^{(1)}, \vartheta \rangle} \bigg) \bigg]$, and (iii) \textit{Prior perturbation.} Let $\theta' \sim \Ncal(\mu_{0}, \Sigma_{0})$, and $\vartheta' \sim \Ncal (\mu_{0}, \Ibf_{d} /\lambda^{2})$, all i.i.d. Then, the perturbed $\Lcal_{3}(\cdot)$ becomes $ \Lcal_{3}'(\theta, \vartheta)  = \frac{\lambda^2}{2} \norm{\theta - \vartheta + \vartheta'}{2}{2} + \frac{1}{2} (\theta - \mu_{0} - \theta')^{T} \Sigma_{0}^{-1} (\theta - \mu_{0}- \theta')$. Then, at  time $t$, we get the following  MAP point estimate from the perturbed surrogate loss function, 

\vspace{-0.2cm}
\begin{equation}
\label{eq:final_surrogate_perturbed_loss}
\begin{aligned}
    (\hat{\theta}_{t}, \hat{\vartheta}_{t}) =  \underset{\theta, \vartheta}{\argmin} \; \Lcal'(\theta, \vartheta) = \underset{\theta, \vartheta}{\argmin} \; \Lcal_{1}'(\theta, \vartheta) + \Lcal_{2}'(\theta, \vartheta) + \Lcal_{3}'(\theta, \vartheta), 
\end{aligned}
\end{equation}
which are well understood to have a distribution that approximates the actual posterior distribution. \textcolor{black}{Note that the perturbed surrogate loss function is convex, can be optimized easily and is independent of the number of arms, and hence is scalable to infinitely-many armed bandit setting as well. In addition, it can be extended easily to the setting where the offline dataset comes from \emph{multiple} experts with different $(\lambda_{i}, \beta_{i})$ competence tuples. Specifically, for $M$ experts, there will be $M$ similar terms for $\Lcal_{2}'(\cdot)$ and $\Lcal_{3}'(\cdot)$ respectively, while $\Lcal_{1}'(\cdot)$ will remain unchanged. This yields the {Bootstrapped} $\mathsf{warmPref-PS}$ as Algorithm \ref{alg:practical_Prefwarm-PS}. 
}

\begin{remark}
\textcolor{black}{
In the next section, we will show that while a theoretical analysis of the practical approximation proposed above is challenging, it does have excellent empirical performance and can be scaled up to large problems as well.
}
\end{remark}

\begin{figure*}[!t]
    \centering
        {
        \includegraphics[height=0.027\textwidth, width=0.5\textwidth]{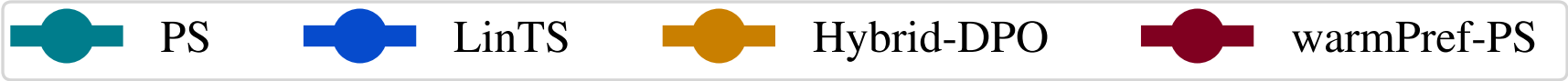}

    }
    {
        \includegraphics[height=0.18\textwidth, width=0.20\textwidth]{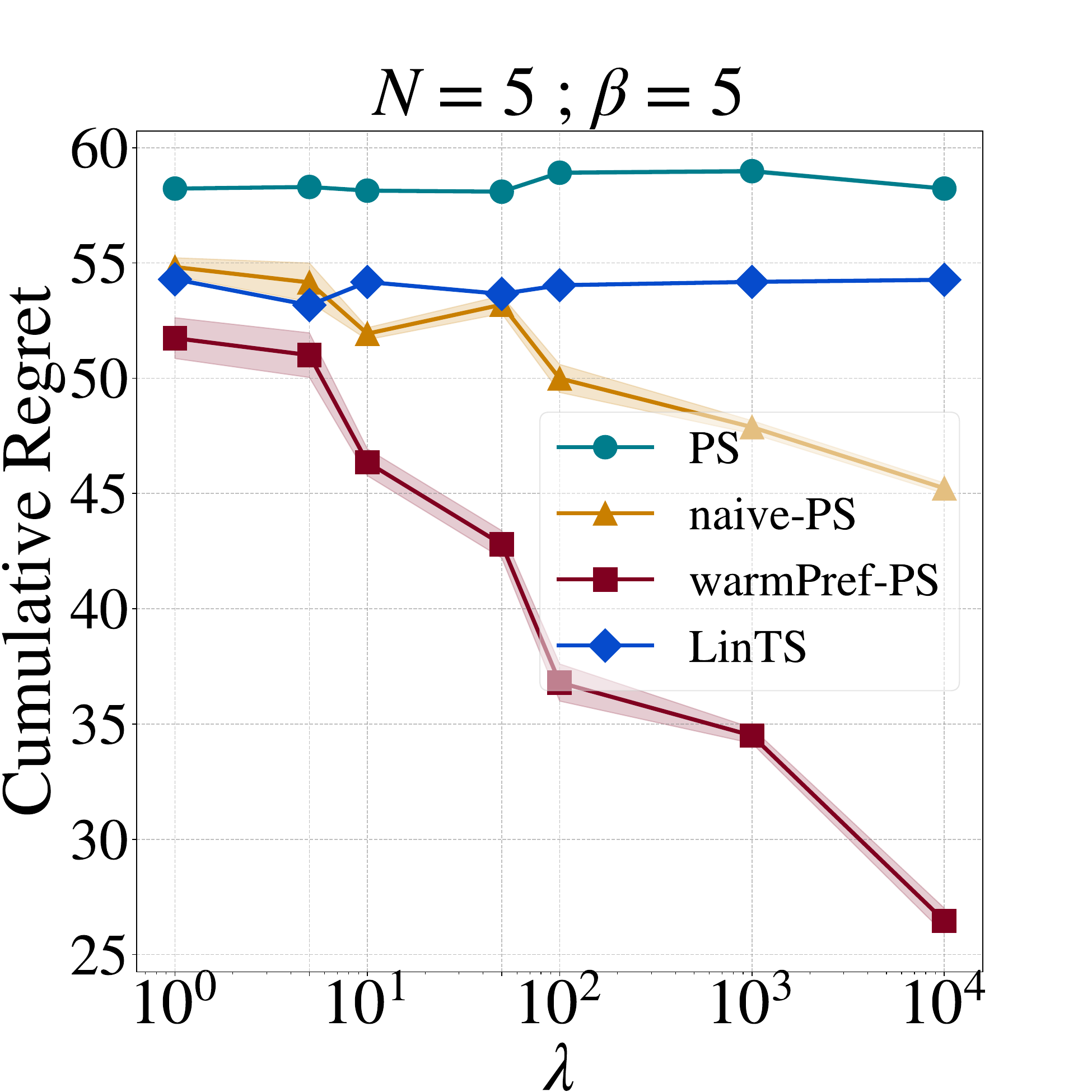}
        \label{fig:lamb-n5-beta5}
    } 
    {
        \includegraphics[height=0.18\textwidth, width=0.20\textwidth]{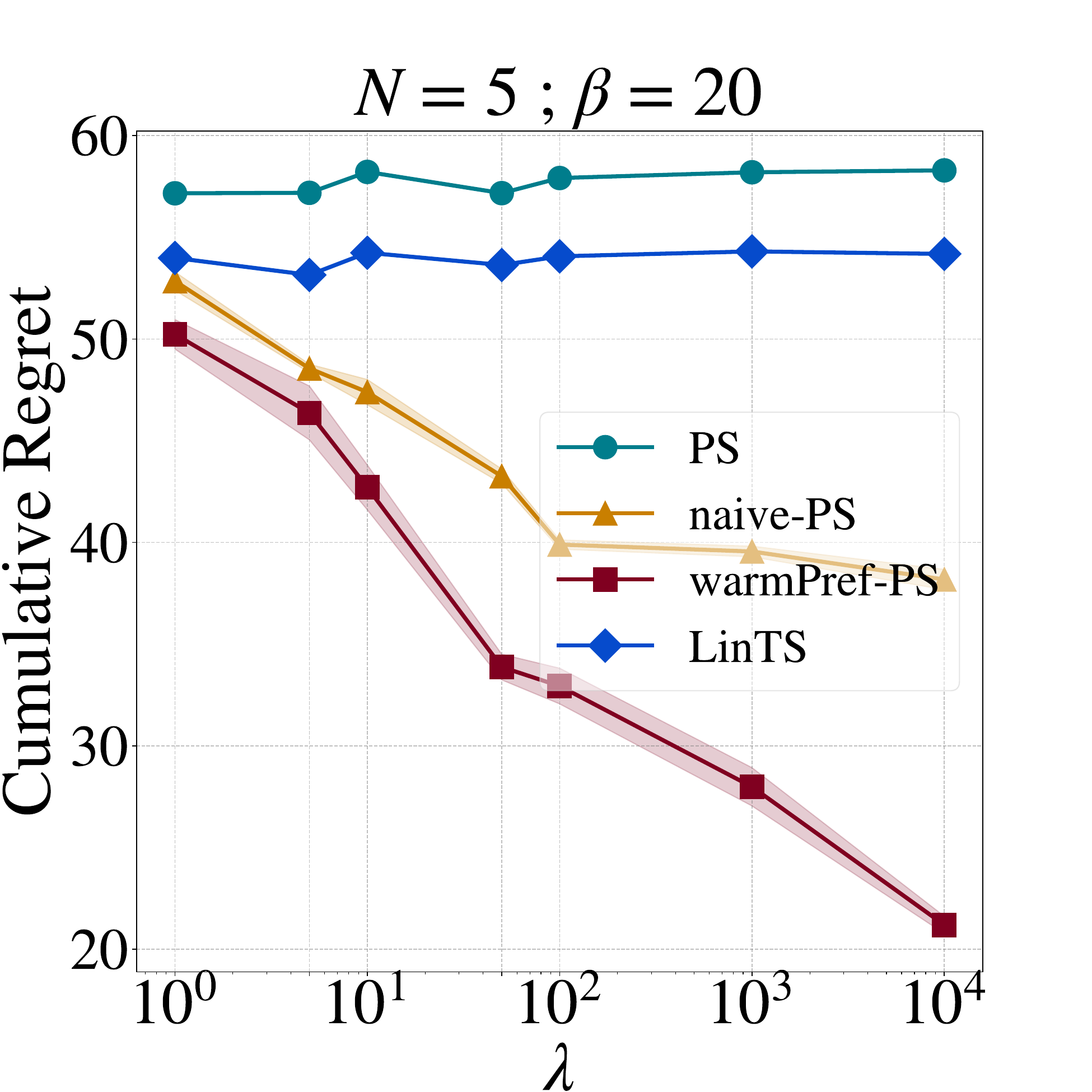}
        \label{fig:lamb-n5-beta20}
    } 
    {
        \includegraphics[height=0.18\textwidth, width=0.20\textwidth]{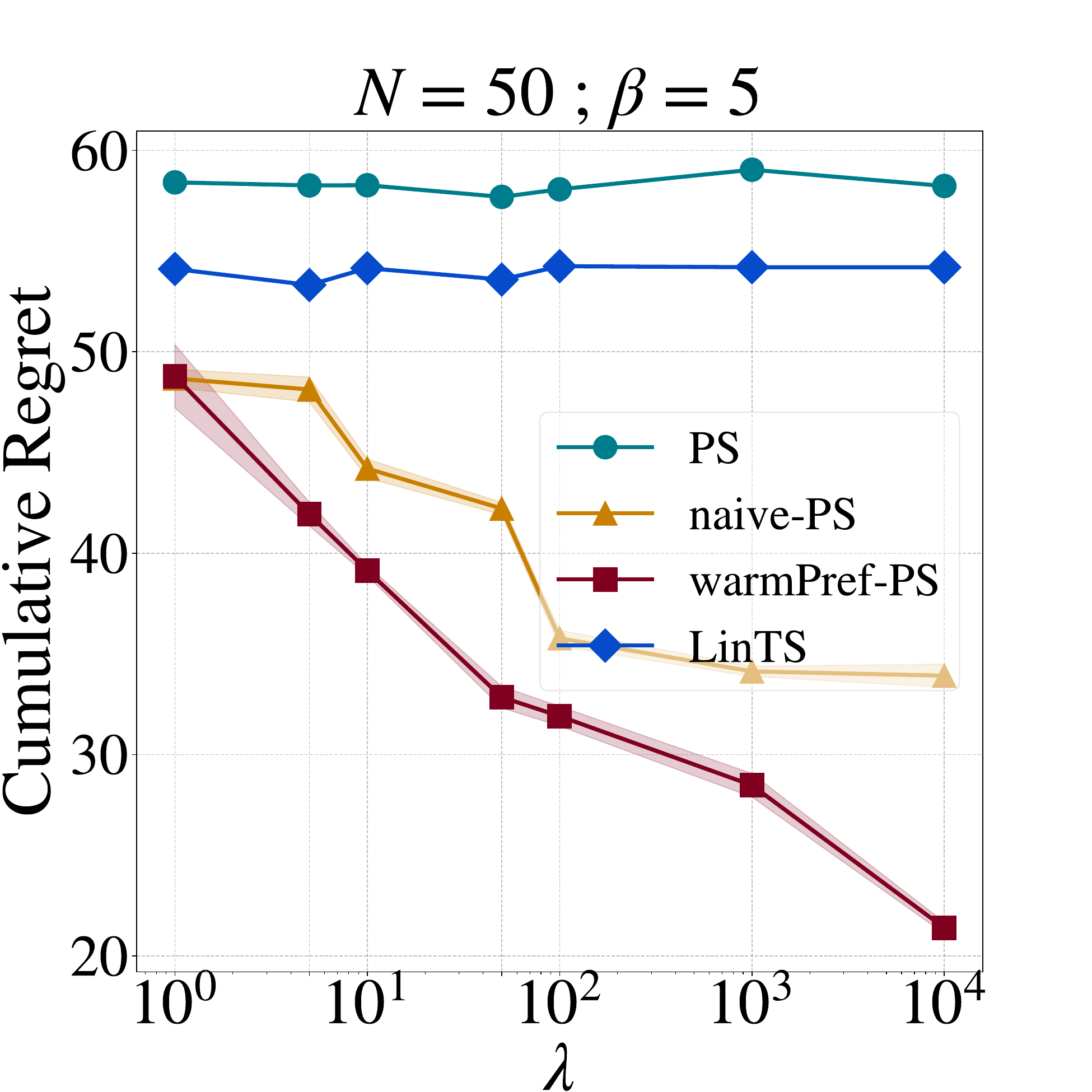}
        \label{fig:lamb-n50-beta5}
    } 
    {
        \includegraphics[height=0.18\textwidth, width=0.20\textwidth]{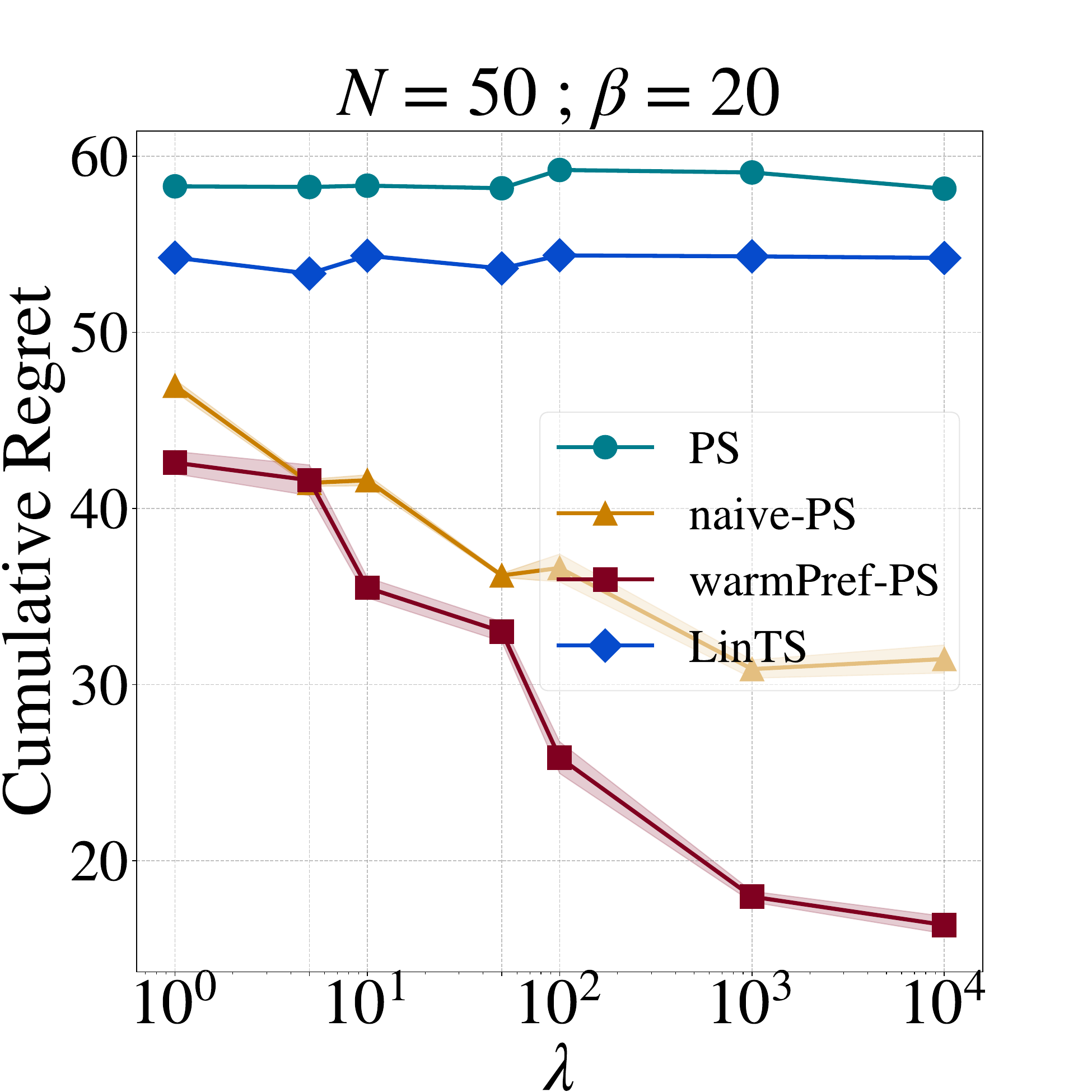}
        \label{fig:lamb-n50-beta20}
    } 
    \newline
    \begin{center}  
    \label{fig:a} \vspace{-0.51cm}
        (a) Varying $\lambda$ for fixed $\beta$ and $N$. 
    \end{center}
    
    {
        \includegraphics[height=0.18\textwidth, width=0.20\textwidth]{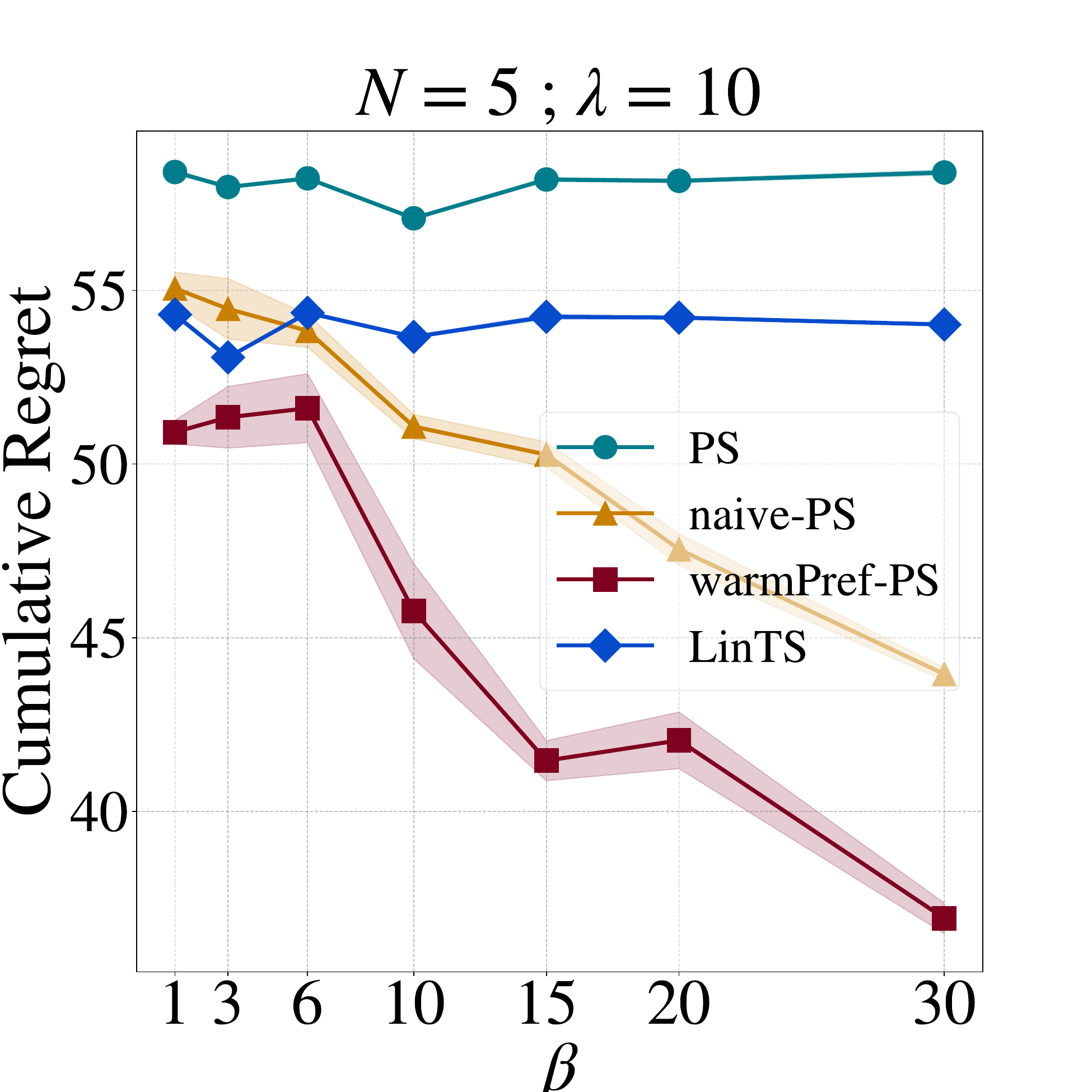}
        \label{fig:beta-n5-lamb10}
    } 
    {
        \includegraphics[height=0.18\textwidth, width=0.20\textwidth]{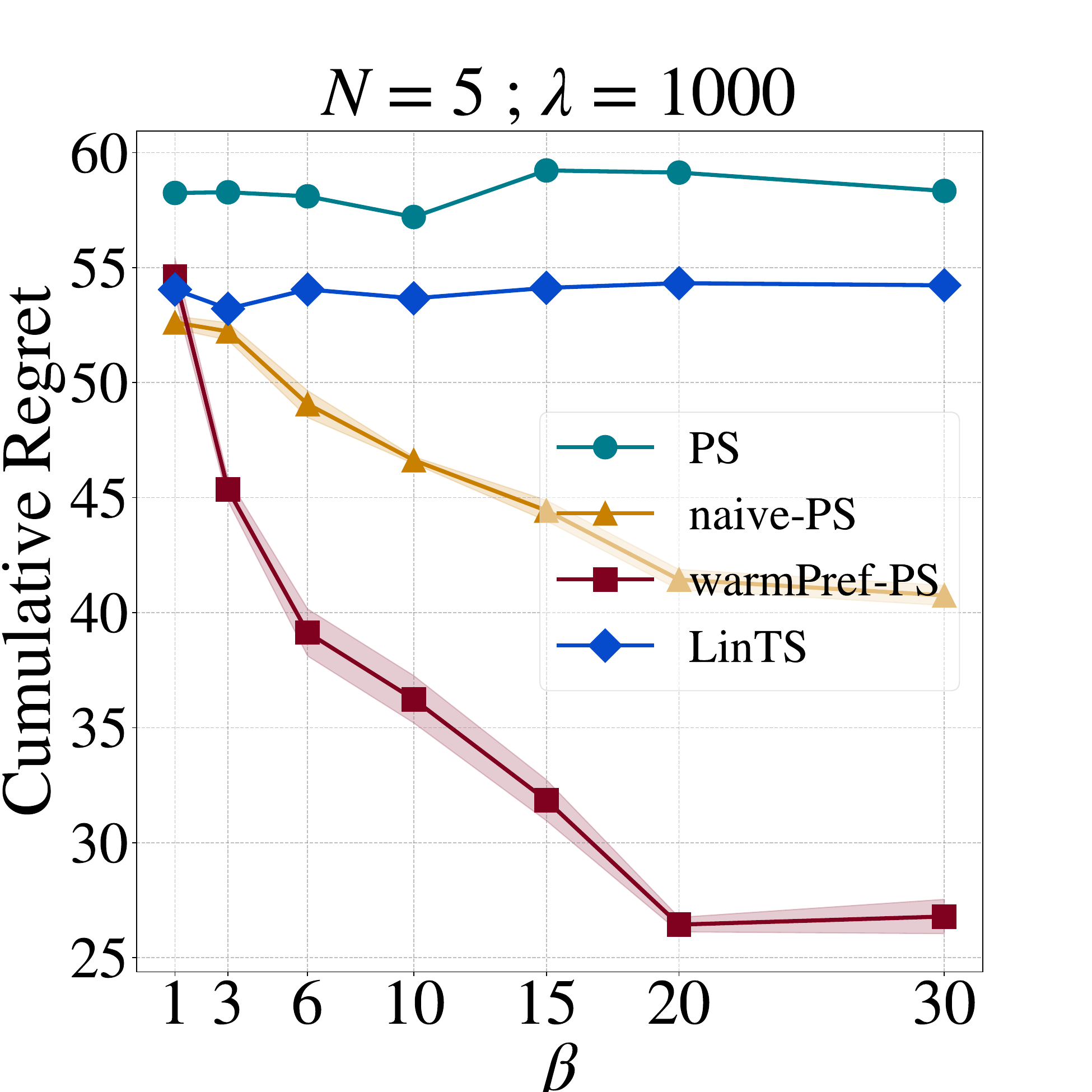}
        \label{fig:beta-n5-lamb1000}
    } 
    {
        \includegraphics[height=0.18\textwidth, width=0.20\textwidth]{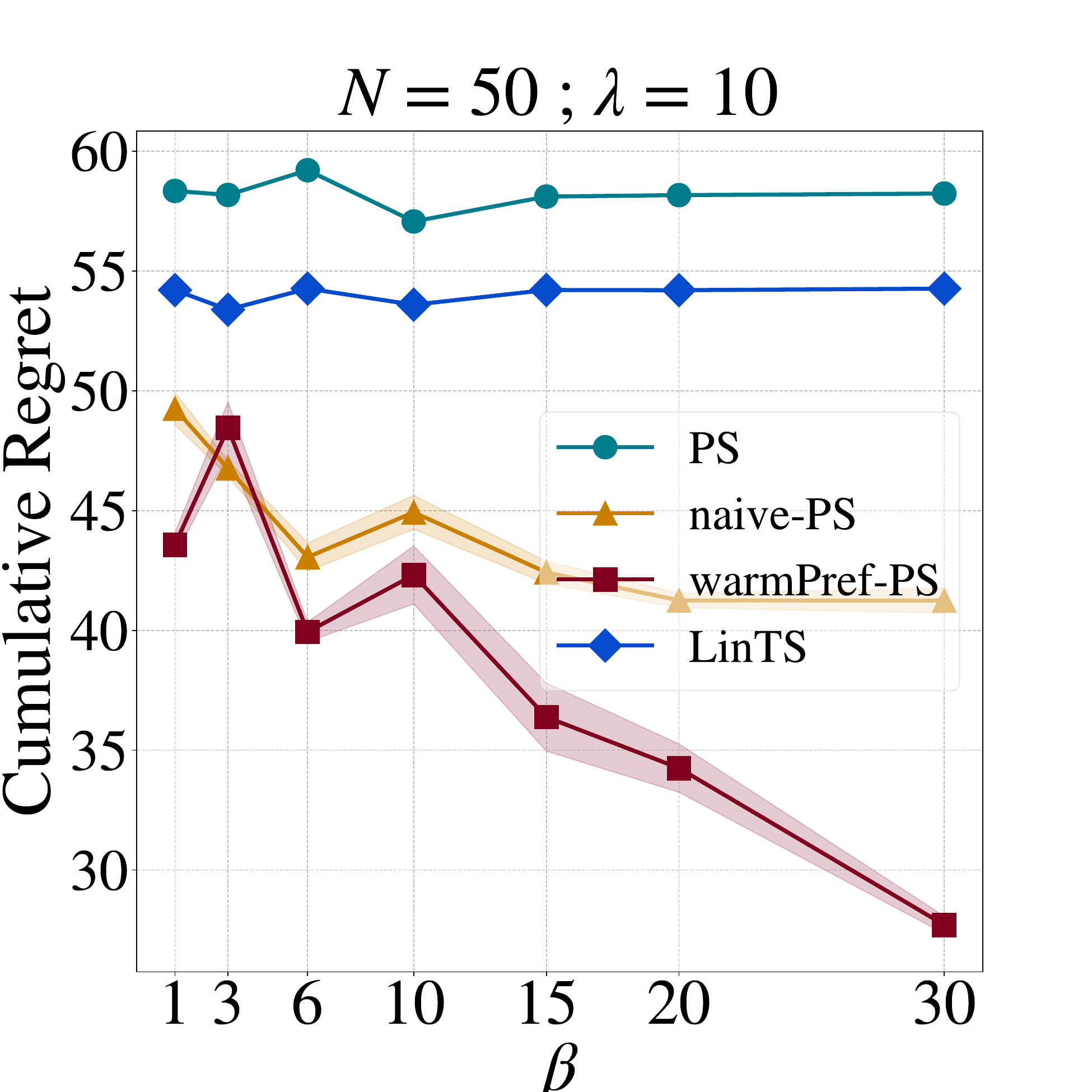}
        \label{fig:beta-n50-lamb10}
    } 
    {
        \includegraphics[height=0.18\textwidth, width=0.20\textwidth]{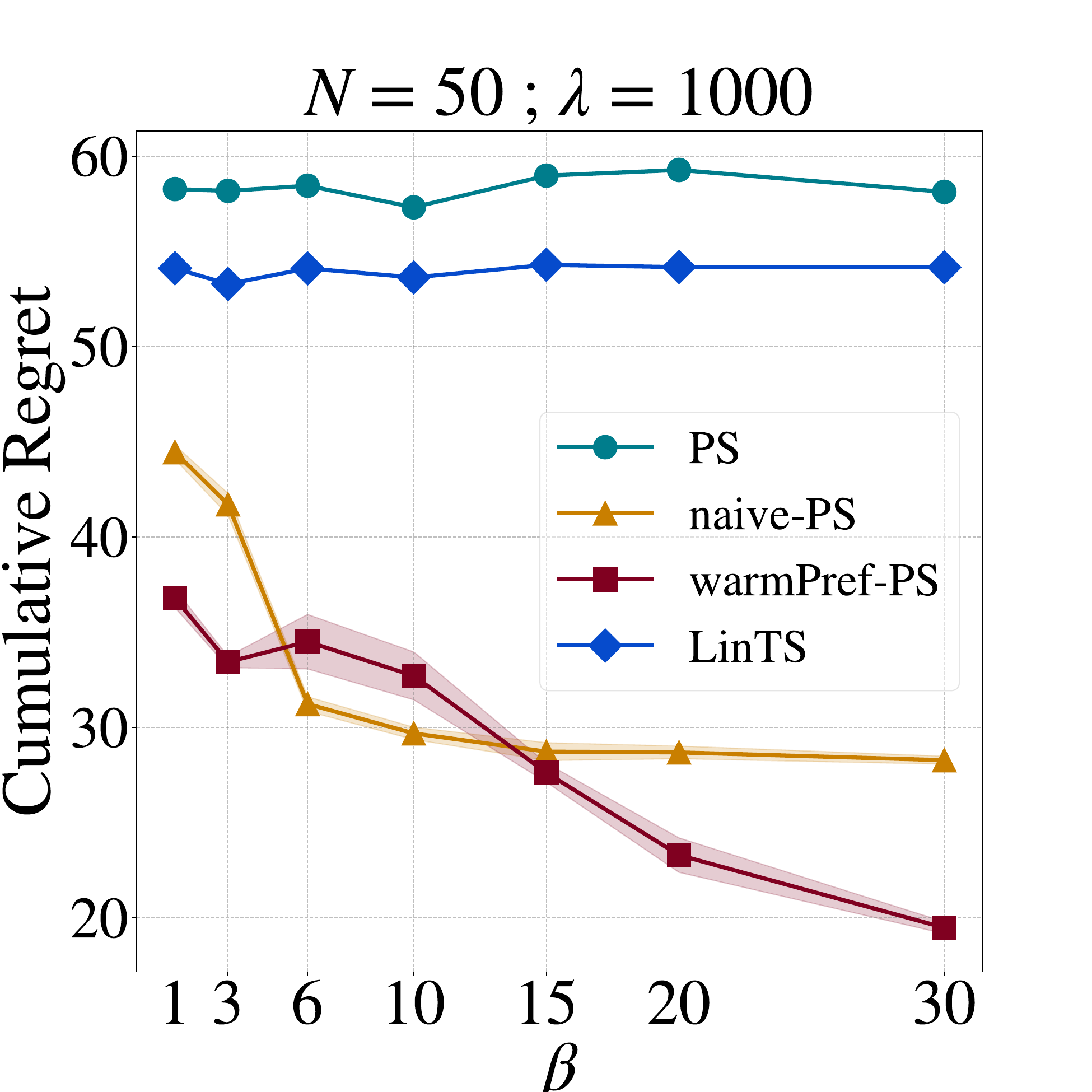}
        \label{fig:beta-n50-lamb1000}
    } 
    \newline
    \begin{center} \vspace{-0.51cm}
        (b) Varying $\beta$ for fixed $\lambda$ and $N$. 
    \end{center}
    {
        \includegraphics[height=0.18\textwidth, width=0.20\textwidth]{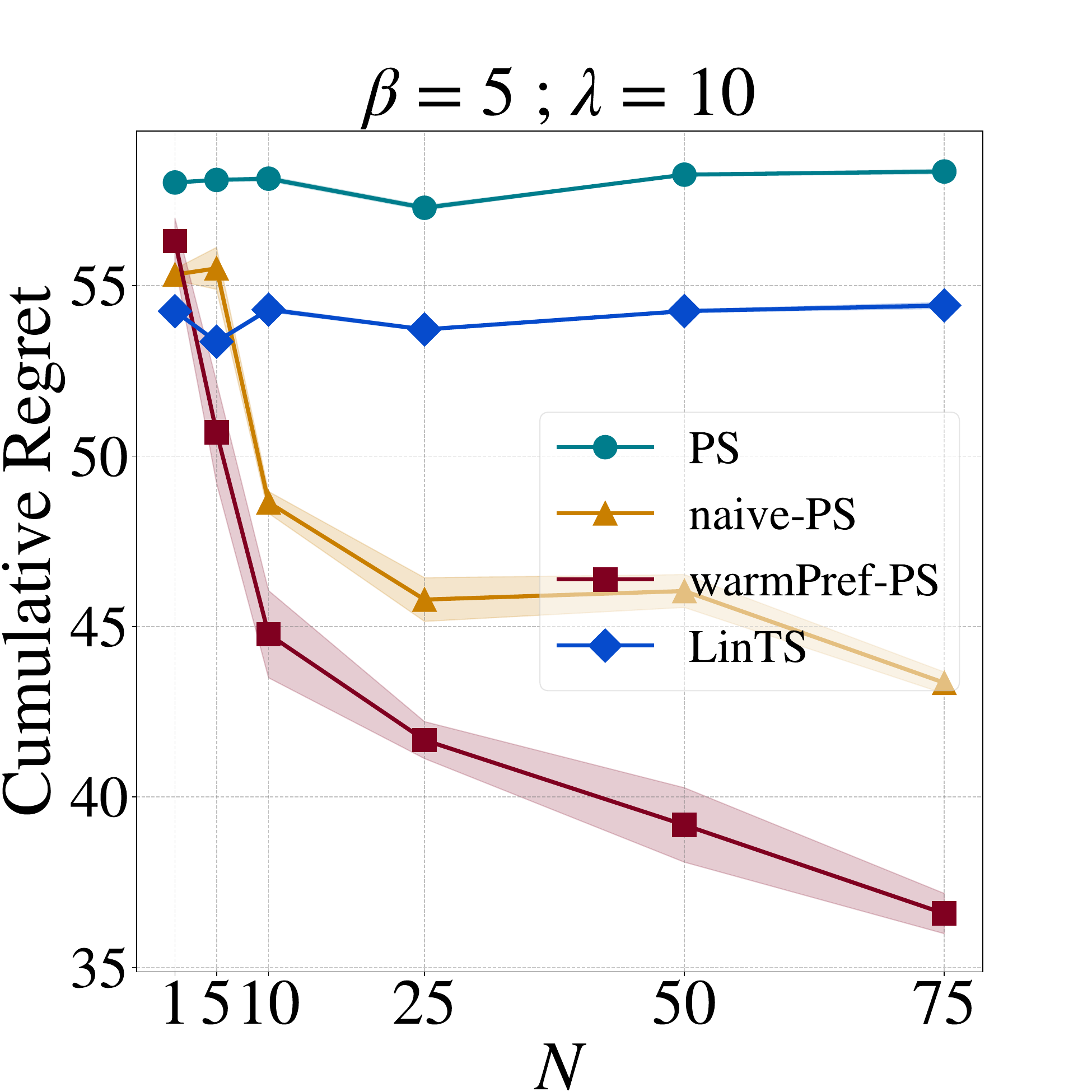}
        \label{fig:n-beta5-lamb10}
    } 
    {
        \includegraphics[height=0.18\textwidth, width=0.20\textwidth]{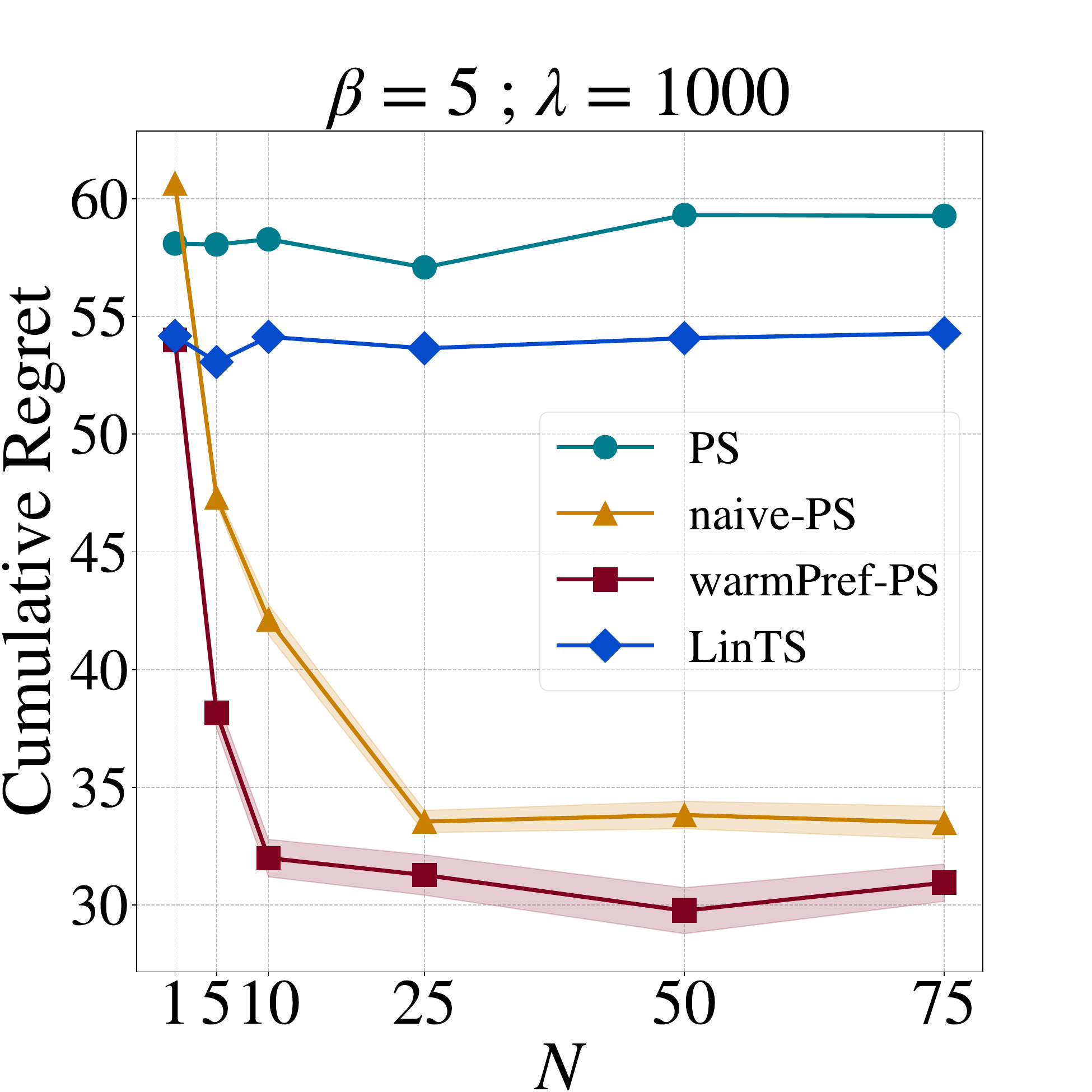}
        \label{fig:n-beta5-lamb1000}
    } 
    {
        \includegraphics[height=0.18\textwidth, width=0.20\textwidth]{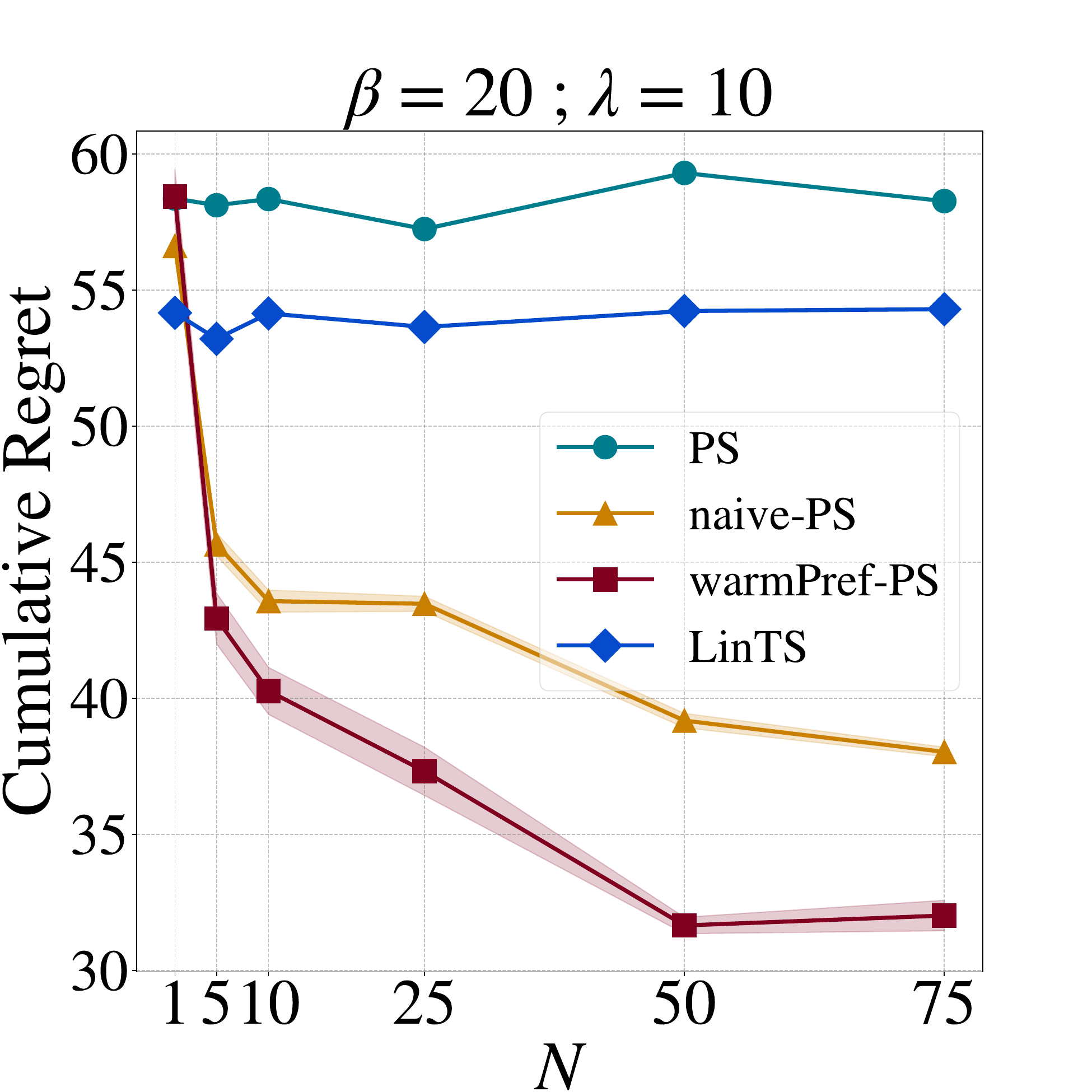}
        \label{fig:n-beta20-lamb10}
    } 
    {
        \includegraphics[height=0.18\textwidth, width=0.20\textwidth]{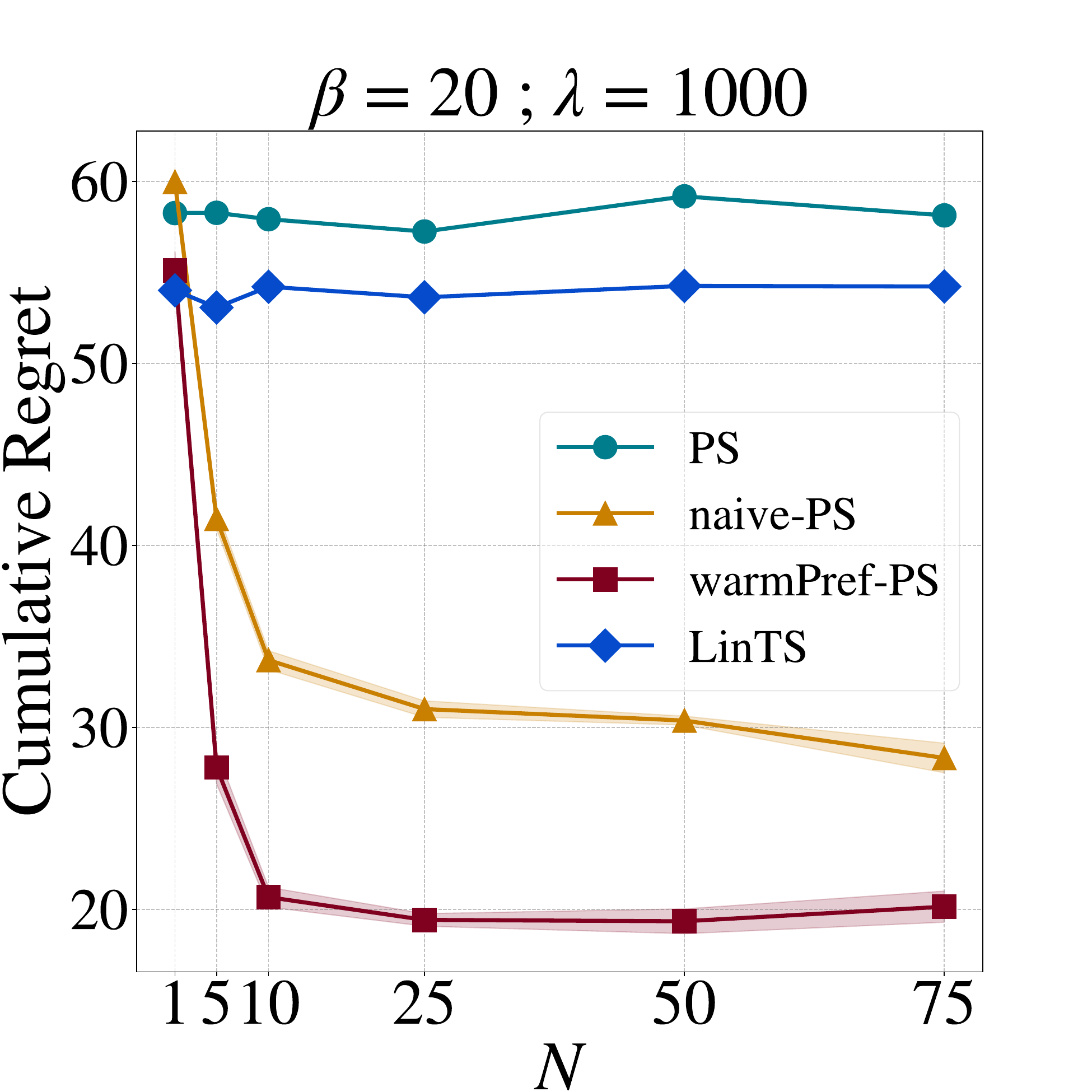}
        \label{fig:n-beta20-lamb1000}
    } 
    \newline
    \begin{center} \vspace{-0.51cm}
        (c) Varying $N$ for fixed $\beta$ and $\lambda$. 
    \end{center}
    \vspace{-0.3cm}
    \caption{Cumulative Regret comparison with varying $N$, $\beta$, and $\lambda$. Shaded region around the mean line represents 1 standard deviation over 5 independent runs.}
    \label{fig:warmTS-ablation}
\end{figure*}

\begin{figure*}[!ht]
\vspace{-0.5cm}
    \centering
    \subfloat[Flawed expert, $\lambda$=10]{\includegraphics[height=0.18\textwidth, width=0.20\textwidth]{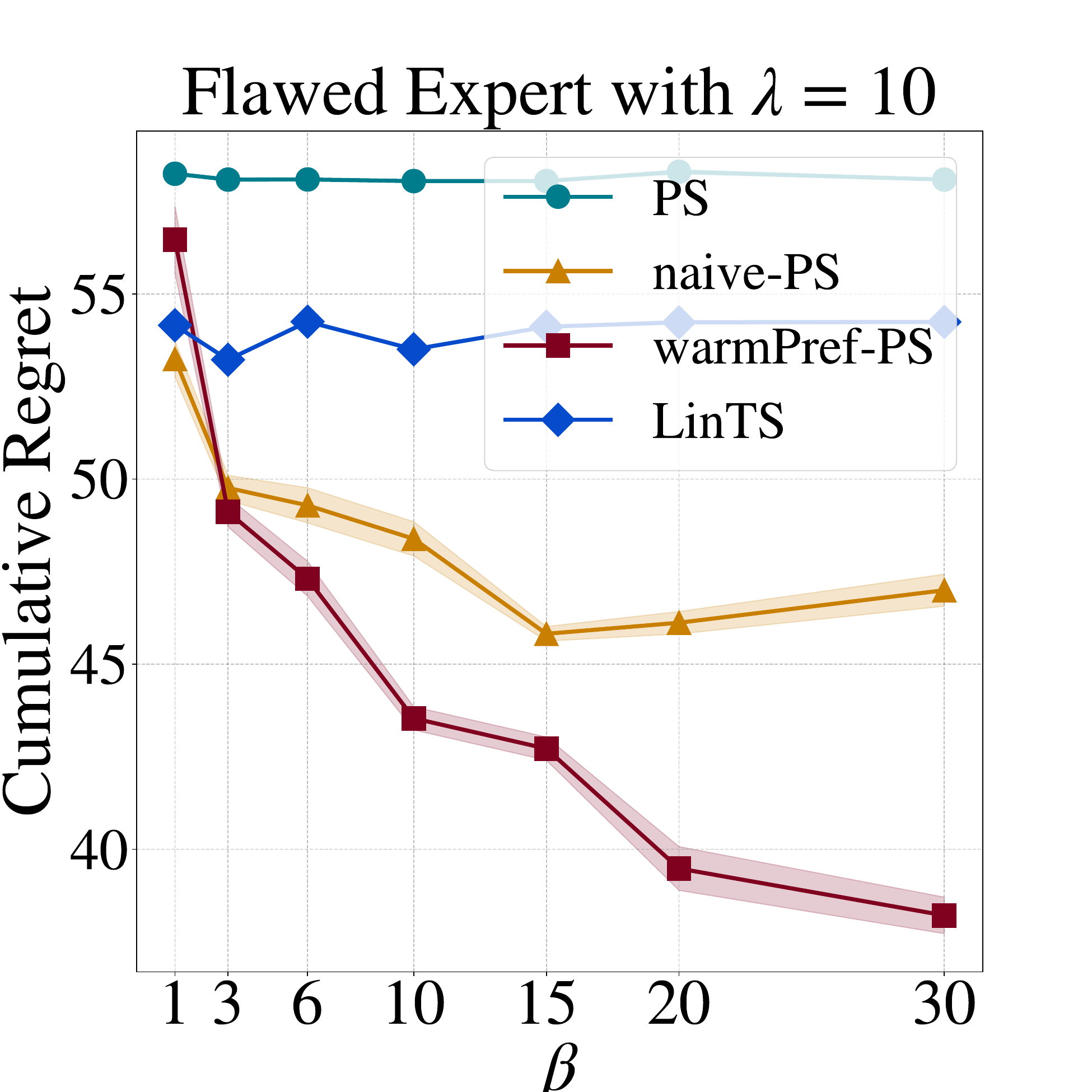}
        \label{fig:policy-lamb-n5-lamb10}
    } 
    \subfloat[Misspecified $\lambda$]{\includegraphics[height=0.18\textwidth, width=0.20\textwidth]{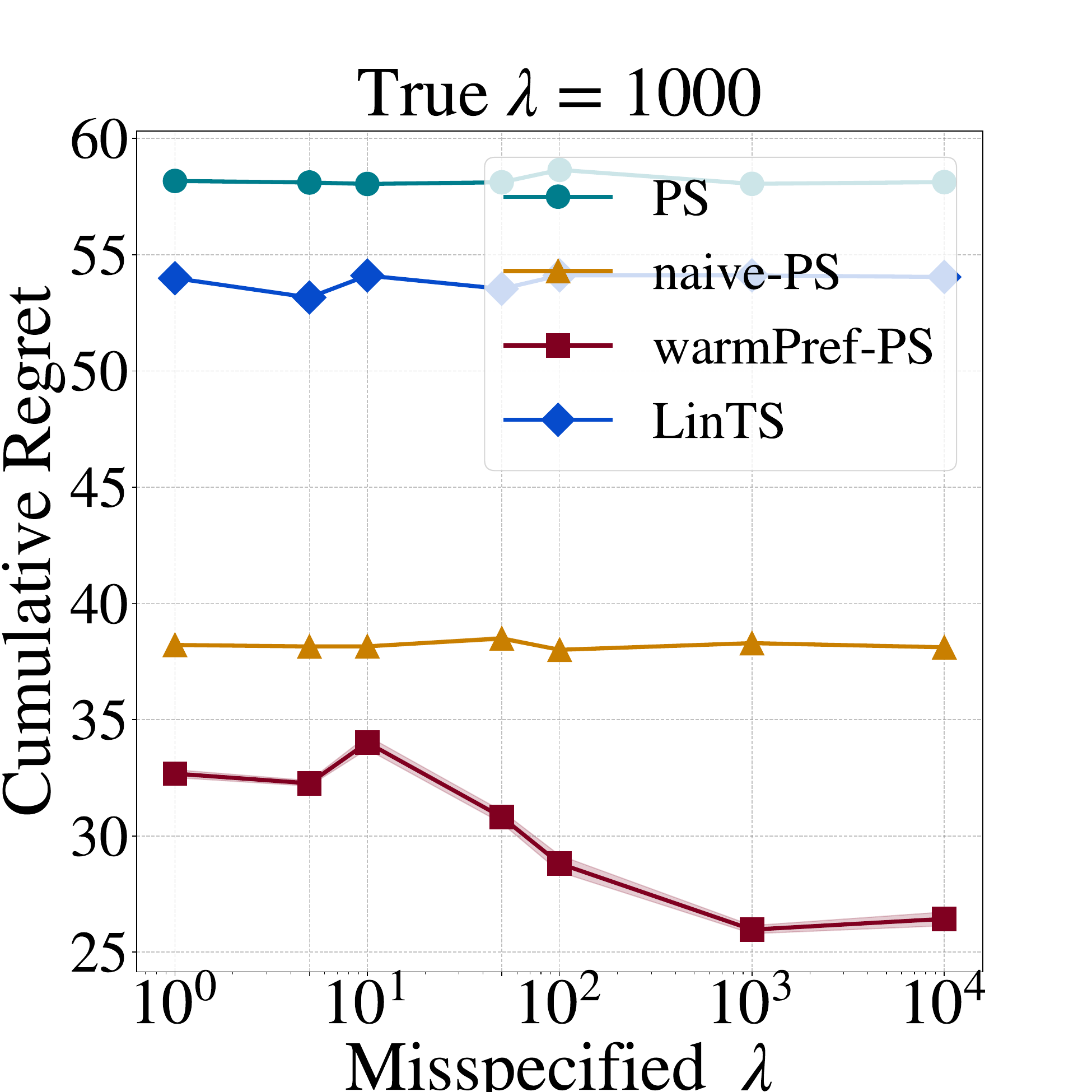}
        \label{fig:competence-lamb}
    } 
    \subfloat[Misspecified $\beta$]
    {\includegraphics[height=0.18\textwidth, width=0.20\textwidth]{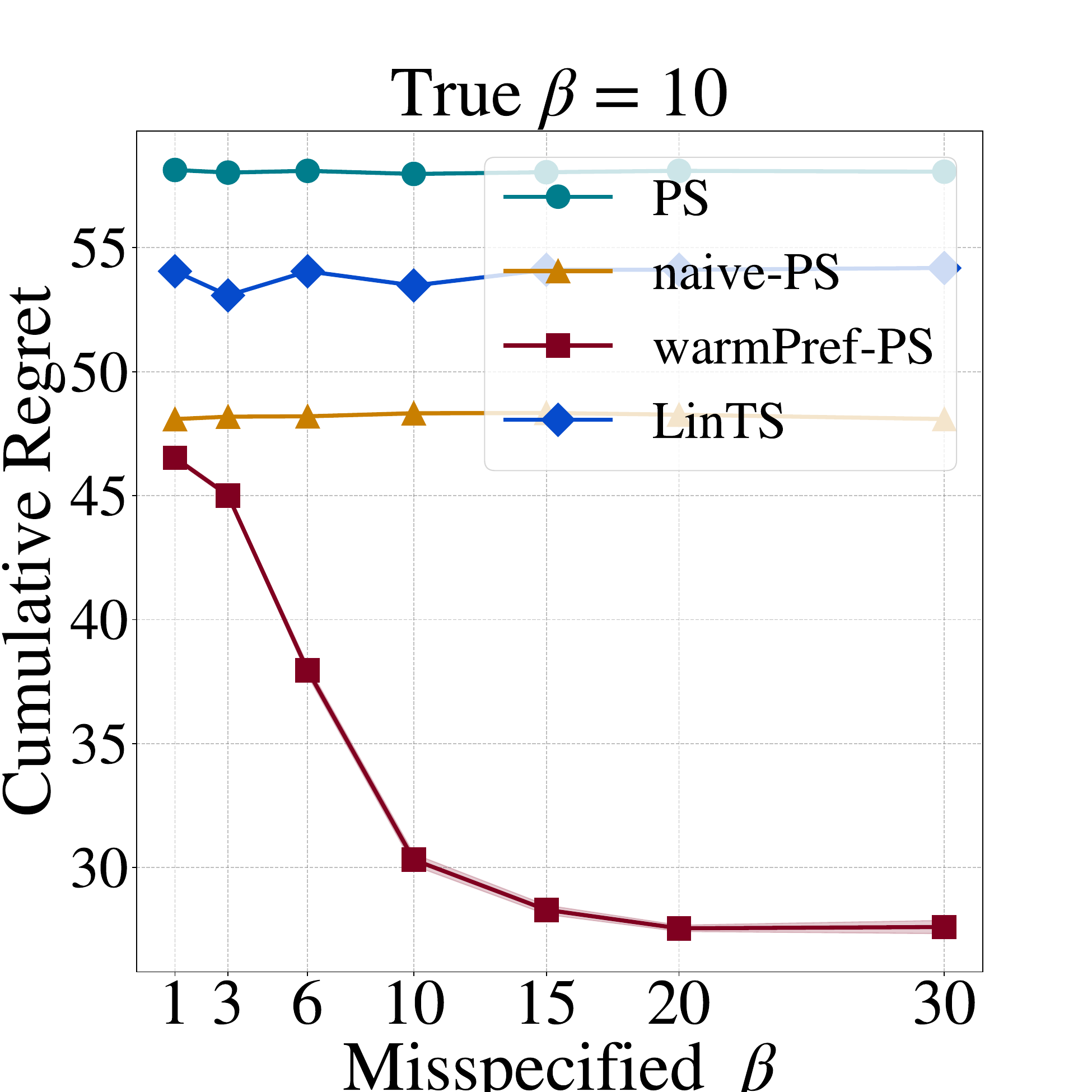}
        \label{fig:competence-beta}
    } 
    \subfloat[Unknown $\beta$]
    {\includegraphics[height=0.18\textwidth, width=0.20\textwidth]{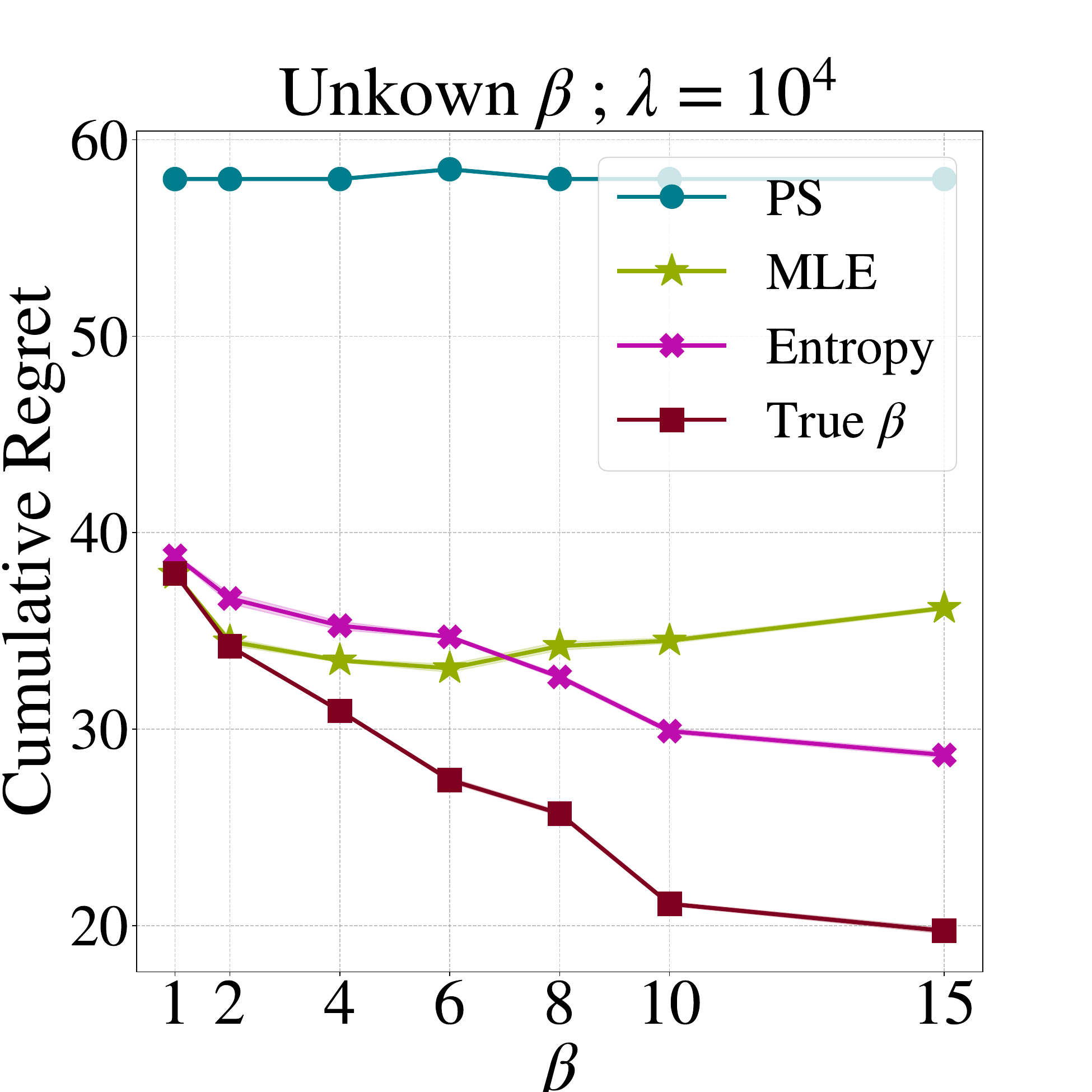}
        \label{fig:unknown-beta}
    }
    \caption{Sensitivity analysis with flawed expert policy, misspecified and unknown competence.}
    \label{fig:misspecification-unknowns-ablation}
\end{figure*}

\section{Empirical Results}
\label{sec:chapter3_empirical}

We now present results on the empirical performance of the {Bootstrapped} $\mathsf{warmPref-PS}$ algorithm introduced in the previous section. We are particularly interested in the following questions: 
(i) How much is the reduction in cumulative Bayesian regret due to warm start with an offline preference dataset? (ii) How much does the competence (in terms of $\lambda$ and $\beta$) of the expert (rater) who generated the offline preference affect regret? (iii) Is $\mathsf{warmPref-PS}$ robust to mis-specification of $\lambda$ and $\beta$?

\noindent\textbf{Baselines.} \textcolor{black}{To evaluate the {Bootstrapped} $\mathsf{warmPref-PS}$ algorithm, we  consider the following baselines:
(i) \texttt{(vanilla) PS}, a PS algorithm that does not use the offline dataset, (ii) \texttt{LinTS} from \cite{li2010contextual} and \cite{pmlr-v28-agrawal13}, and (iii) Direct Preference Optimization (\texttt{DPO}) \citep{rafailov2024direct}. All plots show empirical regret. Another possible baseline can be based on optimism methods to directly learn a $\hat{\theta}$ from $\Dcal_{0}$, and use that to warm-start the online learning. However, such optimism-based algorithms are computationally intractable as they need to construct confidence sets, and then optimize over them. }

\textcolor{black}{
\begin{remark}
\texttt{DPO} cannot be trivially extended to our problem setting, i.e., fixed offline preference dataset with  online numerical reward learning. Comparing \texttt{DPO} trained only on $\Dcal_{0}$ is not fair, and hence, we consider an offline-online variant of \texttt{DPO}, called \texttt{Hybrid-DPO}, with $\epsilon$-greedy online exploration. Please see Appendix \ref{appendix:dpoipo} for more details.
\end{remark}
}

\textcolor{black}{
\begin{remark}
To the best of our knowledge, no other works that formalize learning from offline preference data and online numerical rewards exist, hence, there are no other baselines available in the literature.
`Hybrid bandit' studies cited in Related Work assume numeric rewards in both phases, making direct and fair comparison with $\mathsf{warmPref-PS}$ not possible. A conceivable proxy is to fit a reward model on the offline preference data, convert preferences into pseudo-rewards, and benchmark algorithms on this reward-based offline dataset. However, converting feedback modalities compromises the validity of any fair comparison.
\end{remark}
}

\paragraph{Evaluation protocol}. Unless specified otherwise, for all experiments, we have $K=50$ arms, dimension $d=6$, $\lambda = 100$, $\beta = 10$, dataset size $N=20$, and horizon $T=300$. We averaged over 5 runs (with random seeds). For easy interpretation, we let $\mu \sim \mathrm{Unif}(\cdot)$. \textcolor{black}{Since, to the best of our knowledge, there are no open-sourced datasets available, we work with synthetic datasets.}



\paragraph{Value of Offline Preferences}.
We first aim to understand the impact of the offline preference dataset $\Dcal_{0}$ on the performance of $\mathsf{warmPref-PS}$ as three parameters, $\beta$, $\lambda$ and $N$ vary. Figures \ref{fig:warmTS-ablation}(a) shows that as $\lambda$ increases (the expert has a better estimate of the reward model $\theta$), the regret reduces and this reduction is substantial for the $\mathsf{warmPref-PS}$ algorithm than for the \texttt{naive-PS} algorithm (vanilla-PS and LinTS are unaffected by $\Dcal_{0}$ as they do not use the offline dataset). Figure \ref{fig:warmTS-ablation}(b) shows that (for fixed $\lambda$ and $N$) as $\beta$ increases, the regret reduces substantially. Figure \ref{fig:warmTS-ablation}(c) now fixes $\beta$ and $\lambda$, and shows that as dataset size $N$ increases, even with a `mediocre' expert ($\beta = 5$) the regret reduces substantially, and in fact by 25 to 50\% even with a very small ($N=5$) dataset size. The conclusion from these results is that even by using a small amount of offline data from a mediocre expert, the {Bootstrapped} $\mathsf{warmPref-PS}$ algorithm achieves significant reduction in regret  over the baselines.

\paragraph{Sensitivity to parameter specification errors}.
The  ({Bootstrapped}) $\mathsf{warmPref-PS}$ algorithm in Section \ref{sec:practicalwarmPref-PS} requires a knowledge of expert's parameters $\beta$ and $\lambda$. In Figure \ref{fig:misspecification-unknowns-ablation}, we study the sensitivity of the algorithm's performance to errors in specification of these parameters (as well as of assuming a Bradley-Terry model for the rater). Due to space constraints, further ablation studies are provided in Appendix \ref{appendix:ablations}.
\\
(i) \textbf{Different Preference Generation Expert Policy.} Though the learning agent assumes Equation \eqref{eq:rater_preference_dist} as the expert's generative model, we consider it to actually use a deterministic greedy policy. Actions $\bar{A}_{n}^{(0)}$ and $\bar{A}_{n}^{(1)}$ are sampled, and then choose $Y_{n} = \argmax_{i \in \{0,1\}} \beta \langle \bar{A}_{n}^{(i)} , \vartheta \rangle$, where $\vartheta\sim\Ncal(\theta, \Ibf_d / \lambda^2)$. In Figure \ref{fig:misspecification-unknowns-ablation}(a), we see that even when the learning agent's assumption of the expert policy is \textit{flawed}, $\mathsf{warmPref-PS}$ outperforms the baselines.
\\
(ii) \textbf{Misspecified Competence parameters.} First, we generate the offline data with the true $\lambda=10^{3}$ but the algorithm uses a misspecified $\lambda$. Second, we generate the offline data with the true $\beta=10$ but the algorithm uses a misspecified $\beta$. Figure \ref{fig:misspecification-unknowns-ablation}(b) and \ref{fig:misspecification-unknowns-ablation}(c) show that although the performance of $\mathsf{warmPref-PS}$ decreases as the degree of flawness increases, our algorithm still outperforms the baselines.   
\\
(iii) \textbf{Unknown Competence.} As seen in Section \ref{sec:practicalwarmPref-PS}, Bayesian bootstrapping requires an input for the competence level. In practice, this is not available but, can be estimated from the offline dataset. There are many ways of estimating $\beta$ \citep{beliaev2022imitation, hao2023leveraging}, but the most common methods are: (i) Maximum Likelihood Estimation (MLE) : optimize $\beta$ over the negative log-likelihood of the $\Dcal_{0}$ and, (ii) Entropy : calculate entropy of the empirical distribution of the actions occuring in $\Dcal_{0}$, call it $\Hcal_{\Dcal_{0}}$ and use $\hat{\beta} = c / \Hcal_{\Dcal_{0}}$, where $c>0$ is a hyperparameter. We compare entropy-based method and MLE based method for $\mathsf{warmPref-PS}$ with baselines: (1) use true $\beta$ with $\mathsf{warmPref-PS}$ and, (2) vanilla PS. To isolate the effect of $\beta$, we let $\lambda = 10^{4}$. As shown in Figure \ref{fig:misspecification-unknowns-ablation}(d), although performance degrades due to estimation, $\mathsf{warmPref-PS}$ still beats baselines. 



\newpage
\clearpage

\section{Appendix}
\label{sec:chapter3_appendix}

This appendix is structured as follows.

\vspace{3cm}
{
\renewcommand{\arraystretch}{1.15}
\begin{table*}[ht]
\centering
\fontsize{10}{11}\selectfont
\begin{tabular}{c|p{11.5cm}}
\hline
Section \ref{appendix:twoactions} & First building block of main result Theorem \ref{theorem:finalmultipleactionssamplecomplexity}. Deals with two actions and understanding comparison noises. Contains Lemma \ref{lemma:twoactionssamplecomplexity} and proofs. \\
Section \ref{appendix:multipleactionsinfinitebeta} & Second building block. Deals with multiple actions but no comparison noises i.e. $\beta \to \infty$. Contains Lemma \ref{lemma:multipleactionssamplecomplexity} and proofs. \\
Section \ref{appendix:multipleactionsfinitebeta} & Final building block. Combines the results from Lemmas \ref{lemma:twoactionssamplecomplexity} and \ref{lemma:multipleactionssamplecomplexity}, and gives proof of Theorem \ref{theorem:finalmultipleactionssamplecomplexity}. \\
Section \ref{appendix:regretanalysis} & Concerns regret analysis, and contains proofs of Lemma \ref{lemma:priordependentinformationset} given in Lemmas \ref{lemma:f1informative} and \ref{lemma:cardinalitysmall}. \\
Section \ref{appendix:surrogatelossfunction} & Contains details on Bayesian bootstrapping of $\mathsf{warmPref-PS}$ and proof of Lemma \ref{th:mapestimatelemma}. \\
Section \ref{appendix:warmTSOF} & Gives proof of concept of \texttt{warmTSOF} algorithm and experimental results. \\
Section \ref{appendix:dpoipo} & Gives training details of DPO and IPO. \\
Section \ref{appendix:ablations} & More ablation studies continued from Section \ref{sec:chapter3_empirical}. \\
\hline
\end{tabular}
\label{table:appendixcontents}
\end{table*}
}

\newpage

\subsection{Understanding Two Actions and Finite Deliberateness}
\label{appendix:twoactions}

In the building block towards Theorem \ref{theorem:finalmultipleactionssamplecomplexity}, we consider the case with $K=2$ and hence $\Acal = \{a_0, a_1 \}$, which means $\bar{A}_i^{(0)}=a_0$ and $\bar{A}_i^{(1)}=a_1$ for all $i \in [N]$. Here, we  focus on understanding how the comparison noises, due to finite deliberateness, affect the results. In other words, we see how the effect of deliberateness results in more than one sample being required to determine the optimal action with high probability. 

Essentially, given an offline dataset $\Dcal_0$, we construct a \emph{warm}-posterior of the likelihood that an action is optimal. Based on this posterior over the actions, we can construct an action subset $\UD \subset \Acal$ with $|\UD|=1$.

\subsubsection{Constructing the Algorithm}

In this part, we deal with the question that given an offline dataset $\Dcal_0$, how to develop an algorithm to constructing an action subset $\Ucal \subset \Acal$ with $|\Ucal|=1$.

For this, we need to calculate the posterior distribution of an action being optimal given the offline dataset $\Dcal_{0}$. If $P(a_{0} = A^{\star} \given \Dcal_{0}) > P(a_{1} = A^{\star} \given \Dcal_{0})$, then $\Ucal = \{a_{0}\}$, else $\Ucal = \{ a_{1} \}$. Let $p_{0} := \exp\left(\beta\left\langle a_{0}, \vartheta\right\rangle\right)$ and $p_{1} := \exp\left(\beta\left\langle a_{1}, \vartheta\right\rangle\right)$. So, 

\begin{equation}
\begin{aligned}
    P(a_{0} = A^{\star} \given \Dcal_{0}) &= \frac{P(\Dcal_{0} \given a_{0} = A^{\star}) \cdot P(a_{0} = A^{\star})}{P(\Dcal_{0})} &\\
    &= \frac{\displaystyle\int P(\Dcal_{0} \given a_{0} = A^{\star} \,;\, \vartheta) \, d\vartheta \cdot P(a_{0} = A^{\star})}{\displaystyle\int \displaystyle\int P(\Dcal_{0} \given \beta, \lambda) \, d\lambda d\beta} &\\
    &= \frac{\displaystyle\int \bigg( \frac{p_{0}}{p_{0}+p_{1}} \bigg)^{N} \, d\vartheta \cdot P(a_{0} = A^{\star} \, ; \, \theta_{0})}{\displaystyle\int\displaystyle\int P(\Dcal_{0} \given \beta, \lambda) \, d\lambda d\beta} & \big(\theta_{0} \sim \Ncal(\mu_{0}, \Sigma_{0})\big) \\
    &= \frac{\displaystyle\int \bigg( \frac{p_{0}}{p_{0}+p_{1}} \bigg)^{N} \, d\vartheta \cdot P \big(\langle a_{0}, \theta_{0} \rangle \geq \langle a_{1}, \theta_{0} \rangle \big)}{\displaystyle\int\displaystyle\int P(\Dcal_{0} \given \beta, \lambda) \, d\lambda d\beta}  &\\
    &= \frac{\displaystyle\int \bigg( \frac{p_{0}}{p_{0}+p_{1}} \bigg)^{N} \, d\vartheta}{\displaystyle\int\displaystyle\int P(\Dcal_{0} \given \beta, \lambda) \, d\lambda d\beta} \cdot P \big(\langle a_{0} - a_{1}, \theta_{0} \rangle \geq 0 \big)  &\\
    &= \frac{\displaystyle\int \bigg( \frac{p_{0}}{p_{0}+p_{1}} \bigg)^{N} \, d\vartheta}{\displaystyle\int\displaystyle\int P(\Dcal_{0} \given \beta, \lambda) \, d\lambda d\beta} \cdot \bigg( 1 - \Phi \bigg(-\frac{(a_{0} - a_{1})^T \mu_{0}}{\sqrt{(a_{0} - a_{1})^T \Sigma_{0} (a_{0} - a_{1})}}\bigg) \bigg) &\\
\end{aligned}
\label{eq:parti}
\end{equation}

, where $\Phi$ is the CDF of the standard normal distribution. Similar expression follows for $P(a_{1} = A^{\star} \given \Dcal_{0}) = 1 - P(a_{0} = A^{\star} \given \Dcal_{0})$.

\subsubsection{Limiting Behaviour of the Algorithm for Optimal Expert}

Here, we see that under our specified algorithm, for any offline data size $N \geq 1$, as $\beta, \lambda \rightarrow \infty$, $\Ucal \rightarrow \{ A^*\}$ almost surely. It is easy to see this. As $\lambda \to \infty$, we have $\vartheta \to \theta_{0}$. Then, 
$$
\lim_{\beta \to \infty, \vartheta \to \theta_{0}} \bigg( \frac{p_{0}}{p_{0}+p_{1}} \bigg)^{N} = \lim_{\beta \to \infty, \vartheta \to \theta_{0}} \bigg( \frac{1}{1 + e^{- \beta \langle a_{0} - a_{1}, \vartheta \rangle}} \bigg)^{N}.
$$

Now observe that if $\beta, \lambda \to \infty$, if $a_{1} = A^{\star}$, then $\lim_{\lambda \to \infty} \langle a_{0} - a_{1}, \vartheta \rangle \leq 0 \implies \lim_{\beta, \lambda \to \infty} \bigg( \frac{p_{0}}{p_{0}+p_{1}} \bigg)^{N} \to 0 \implies P(a_{0} = A^{\star} \given \Dcal_{0}) \to 0$. Same holds for when $\beta, \lambda \to \infty$ and if $a_{0}$ is optimal. Which means the specified decision rule above converges with $\Ucal \to \{A^{\star}\}$ almost surely.

\subsubsection{Limiting Behaviour of the Algorithm for Large Datasets}

Here, we show that under our specified algorithm, for any finite $\beta > 0$, as $\lambda, N \rightarrow \infty$, $\Ucal \rightarrow \{ A^*\}$ almost surely.

For this, we just calculate the ratio $\lim_{\lambda,N \to \infty} \frac{P(a_{0} = A^{\star} \given \Dcal_{0})}{P(a_{1} = A^{\star} \given \Dcal_{0})}$ and check whether it tends to zero or infinity. Let $x := \frac{(a_{0} - a_{1})^T \mu_{0}}{\sqrt{(a_{0} - a_{1})^T \Sigma_{0} (a_{0} - a_{1})}}$. So then,

\begin{align*}
    \lim_{\lambda,N \to \infty} \frac{P(a_{0} = A^{\star} \given \Dcal_{0})}{P(a_{1} = A^{\star} \given \Dcal_{0})} &= \lim_{\vartheta \to \theta_{0}, N \to \infty} \frac{\big(\frac{p_{0}}{p_{0}+p_{1}} \big)^{N} \cdot (1-\Phi(-x))}{\big(\frac{p_{1}}{p_{0}+p_{1}} \big)^{N} \cdot (1-\Phi(x))} \\
    &= \lim_{\vartheta \to \theta_{0}, N \to \infty} \bigg(\frac{p_{0}}{p_{1}} \bigg)^{N} \cdot \frac{\Phi(x)}{1-\Phi(x)} \\
    &= \lim_{N \to \infty} [ \exp(\beta \langle a_{0} - a_{1}, \theta_{0} \rangle )]^{N} \cdot \frac{\Phi(x)}{1-\Phi(x)}
\end{align*}

Now, we can apply the same argument of, if $a_{0} = A^{\star}$, then $\langle a_{0} - a_{1}, \theta_{0} \rangle \geq 0$ to see that the above expression tends to positive infinity for any finite $\beta > 0$. Hence, we can construct $\Ucal = \{ A^{\star} \}$ almost surely.

\subsubsection{Sample Complexity for Finite Deliberateness}

In this part, we consider for any finite $\beta > 0$, as $\lambda \to \infty$, and any given $\epsilon \in (0, 1)$, under our specified algorithm, how large does $N$ need to be to ensure $P(\Ucal = \{ A^*\}) \geq 1 - \epsilon$?

\begin{restatable}{lemma}{twoactionssamplecomplexity}
\label{lemma:twoactionssamplecomplexity} 
For an action set $\Acal = \{a_0, a_1 \}$ and any finite $\beta \in (0,\infty)$, with $\lambda \to \infty$ and for some $\epsilon \in (0, 1)$, the size of the offline dataset to ensure $\UD = \{A^{\star}\}$ and hence $(1-\epsilon)$-informative is:
\begin{equation}
N \geq \frac{\ln \bigg( \big(\frac{1}{\epsilon}-1 \big) \big(\frac{1}{\Phi(x)}-1 \big) \bigg)}{\beta \langle a_{0} - a_{1}, \theta_{0} \rangle} \, ,
\label{eq:appendixtwoactionssamplecomplexity}
\end{equation}
where $x := \frac{(a_{0} - a_{1})^T \mu_{0}}{\sqrt{(a_{0} - a_{1})^T \Sigma_{0} (a_{0} - a_{1})}}$, and $\Phi(\cdot)$ is the standard Normal CDF.
\end{restatable}

\begin{proof}
\label{proof:twoactionssamplecomplexity}

Assume $A^{\star} = a_{0}$. Then, we want $P(a_{0} = A^{\star} \given \Dcal_{0}) > 1-\epsilon$ and $P(a_{1} = A^{\star} \given \Dcal_{0}) < \epsilon$. Letting Let $x := \frac{(a_{0} - a_{1})^T \mu_{0}}{\sqrt{(a_{0} - a_{1})^T \Sigma_{0} (a_{0} - a_{1})}}$ same as before and taking the ratio of these as $\lambda \to \infty$ for a finite $\beta, N > 0$, we have

\begin{equation}
\begin{aligned}
\lim_{\lambda \to \infty} \frac{P(a_{0} = A^{\star} \given \Dcal_{0})}{P(a_{1} = A^{\star} \given \Dcal_{0})} &= \lim_{\vartheta \to \theta_{0}} \frac{\big(\frac{p_{0}}{p_{0}+p_{1}} \big)^{N} \cdot (1-\Phi(-x))}{\big(\frac{p_{1}}{p_{0}+p_{1}} \big)^{N} \cdot (1-\Phi(x))} &> \frac{1-\epsilon}{\epsilon} \\
&= [\exp(\beta \langle a_{0} - a_{1}, \theta_{0} \rangle)]^{N} \cdot \frac{\Phi(x)}{1-\Phi(x)} &> \frac{1}{\epsilon} - 1 \\
& \implies N > \frac{\ln \bigg( \big(\frac{1}{\epsilon}-1 \big) \big(\frac{1}{\Phi(x)}-1 \big) \bigg)}{\beta \langle a_{0} - a_{1}, \theta_{0} \rangle}. & 
\end{aligned}
\end{equation}

Without loss of generality, similar argument holds for if $A^{\star} = a_{1}$.
\end{proof}

\subsection{Understanding Multiple Actions and Infinite Deliberateness}
\label{appendix:multipleactionsinfinitebeta}

In this building block towards Theorem \ref{theorem:finalmultipleactionssamplecomplexity}, we focus on the case with $\lambda = \beta = \infty$. In other words, there are no comparison noises. Moreover, as before, the two actions $\bar{A}_n^{(0)}$ and $\bar{A}_n^{(1)}$ are i.i.d. sampled from a distribution $\mu$ over $\Acal$. With this, we understand how this sampling distribution $\mu$ affects the results.

For a finite dataset $\Dcal_{0}$ of size $N$, let $\UD \subset \Acal$ be the set consisting of all unique actions occurring in $\Dcal_{0}$. Then, the informative set $\UD$ can be constructed with two types of actions : (i) actions not appearing in $\UD$ (ii) actions occurring in $\UD$ that have not `lost' the comparison with any another action. We begin by constructing an algorithm for this analysis.

\subsubsection{Developing the Algorithm}
\label{sec:developingalgo}

For a finite dataset $\Dcal_{0}$ of size $N$, let $\Ucal_{N} \subset \Acal$ be the set consisting of all unique actions occurring in $\Dcal_{0}$. Then, $\Ucal$ can be constructed with two types of actions : (i) actions not appearing in $\Ucal_{N}$ (ii) actions occurring in $\Ucal_{N}$ that have not `lost' the comparison with any another action. For this, let $\Ccal_{i}$ be the set of comparisons from $\Dcal_{0}$ involving action $a_{i}$ i.e. $\Ccal_{i} = \big\{ \big(\bar{A}_n^{(0)}, \bar{A}_n^{(1)}, Y_n \big) \; ; \; \bar{A}_n^{(0)} = a_{i} \; \text{or}  \; \bar{A}_n^{(1)} = a_{i} \; , \; n \in [N] \big\}$. Hence, construct $\Ucal = (\Acal \setminus \Ucal_{N}) \cup \Wcal_{\Ucal_{N}}$, where $\Wcal_{\Ucal_{N}} := \{a_{i} \in \Ucal_{N} \; ; \; Y_{j} = a_{i} \forAll \big(\bar{A}_j^{(0)}, \bar{A}_j^{(1)}, Y_j \big) \in \Ccal_{i} \}$.

Note here that the conditions mentioned above are tight conditions, which can be analyzed in the case of uniform action sampling distribution ($\mu =$ \texttt{Uniform}$(\cdot)$). However, in the case of an arbitrary distribution such analysis is intractable. In this case, we then only consider the sufficient condition to obtain complete ordering of actions. The sufficient condition to determine the optimal action with high probability is to sample each pair of actions at least once i.e. sample each of $\binom{K}{2}$ pairs once.

\subsubsection{Finding the Optimal Action Given a Large Dataset}

Here, we show that the construction procedure as described above in Part \ref{sec:developingalgo} yields in finding the optimal action given a large dataset. More formally, we show that if the action sampling distribution $\mu$ is not degenerate i.e. $\lim_{N \to \infty} P(a \in \Ucal_{N}) > 0 \forAll a \in \Acal$, then as $N \rightarrow \infty$, $\Ucal \rightarrow \{ A^*\}$ almost surely.

To see this, if $\mu$ is not degenerate, then $\lim_{N \to \infty} P(a \in \Ucal_{N}) > 0 \forAll a \in \Acal$. Then, $\lim_{N \to \infty} (\Acal \; \setminus \; \Ucal_{N}) = \emptyset$. In addition, as $N \to \infty$, for all possible pairs of actions $(a_{i}, a_{j})$ with $a_{i} \neq a_{j} \in \Acal$, we will have $\bar{A}_n^{(0)} = a_{i}$ and $\bar{A}_n^{(1)} = a_{j}$ for some $\big(\bar{A}_n^{(0)}, \bar{A}_n^{(1)}, Y_n \big) \in \Dcal_{0}$. Due to the construction of $\Wcal_{\Ucal_{N}}$, we will also have $\lim_{N \to \infty} \Wcal_{\Ucal_{N}} = \{ A^{\star} \}$ almost surely. This implies $\lim_{N \to \infty} \Ucal = \{ A^{\star} \}$.

Now that we know that the construction procedure of $\Ucal_{N}$ is principled, we wish to generalize the result for finite size of the offline dataset $\Dcal_{0}$.

\subsubsection{General Sample Complexity Analysis}
\label{appendix:generalsamplecomplexity}

In this section, we present results for the following question : In general, how large $N$ need be to ensure that $P(|\Ucal| = \mathcal{o}(K))$, or even $P(|\Ucal|= 1)$ with high probability? We aim to derive this result for an arbitrary action sampling distribution $\mu$, however we begin by analyzing the case of uniform distribution i.e. $\mu \sim$ \texttt{Uniform$(\cdot)$}.

\paragraph{General analysis of probability of picking all $n$ items in $N$ trials}. 
\label{proof:generalprobcoupon}

\mbox{}\\
\indent Here, we describe the general theoretical framework to bound the probabilities of picking \emph{all} of $n$ given items in $N$ independent trials. We first begin with a uniform distribution over each of these $n$ items and later generalize to an arbitrary distribution $\mu$. Note that we derive a general result for $n$ items, which in our case corresponds to actions ($n=K$) or action pairs ($n = \binom{K}{2}$).

\begin{itemize}
    \item \textbf{Uniform distribution.} We have $n$ items which are equally likely to be selected, so we can invoke the Stirling numbers of the second kind (or Stirling partition number) to get a bound on this probability. Stirling numbers of the second kind give the number of ways to partition a set of $u$ objects into $v$ non-empty subsets and is denoted by $S(u,v)$. For notation, we have $n$ items to be selected, $N$ as the number of trials. 

Now, let $S_i$ be all the outcomes in which an item $i$ is \emph{not} selected. For each $i$, $|S_i|=(n-1)^N$ and there are $\binom{n}{1}$ choices for $i$. For each $ j \neq i$, $|S_j \cap S_i|=(n-2)^N$ and there are $\binom{n}{2}$ choices for $(i,j)$. Continuing in this manner to count the number of outcomes missing at least $1$ number, we get

\begin{align*}
\left| \bigcup_{i=1}^{n} S_i \right|
&=\sum_{i=1}^{n} |S_i| - \sum_{j < i}| S_j \cap S_i| + \sum_{k<j<i} |S_k \cap S_j \cap S_i| - \; \dots \\
&= \binom{n}{1} (n-1)^N - \binom{n}{2}(n-2)^N + \binom{n}{3}(n-3)^N - \; \dots
\end{align*}

Since there are a total of $n^N$ total outcomes, we get the number of desired outcomes in which \emph{all} possible numbers are rolled, denoted by $\#_{\text{desired}}$ as

$$
\#_{\text{desired}} = n^N - \binom{n}{1} (n-1)^N + \binom{n}{2} (n-2)^N -\binom{n}{3} (n-3)^N + \; \dots .
$$

Thus, the probability $p_{n, N}$ of picking all $n$ items in $N$ trials is $\frac{\#_{\text{desired}}}{n^{N}}$. Hence,

\begin{equation}
\begin{aligned}
p_{n,N} &= 1-\binom{n}{1}\left(1-\frac{1}{n} \right)^N+\binom{n}{2}\left(1-\frac{2}{n} \right)^N-\binom{n}{3}\left(1-\frac{3}{n} \right)^N + \dots \\
\implies & p_{n,N} = \sum_{i=0}^{n} (-1)^i\binom{n}{i}\left(1-\frac{i}{n} \right)^N
\end{aligned}
\label{eq:uniformprobpickingall}
\end{equation}

\item \textbf{Arbitrary Distribution.}  Assume now that the actions are sampled from an action sampling distribution $\mu$. Since we are forming action pairs for comparison, denote with $\ring \mu_{k}$, the probability of sampling action pair $k := (i,j) \in [\binom{K}{2}]$ and with $\sum_{k} \ring \mu_{k}=1$, with $(i,j)$ representing the action pair $(a_i, a_j)$. Furthermore, this means that assume that $\mu_{\text{min}}^{2} \leq \ring \mu_{k} \leq \mu_{\text{max}}^{2} \forAll k$ for some arbitrary $ 0 < \mu_{\text{min}} \leq \mu_{\text{max}} < 1$.

For this problem, let $T_{i}$ denote the random number of trials needed to sample item $i$ for the first time. The total number of trials needed can be then denoted by the random variable $T = \max(T_{1}, \dots, T_{n})$.  Note that
$T_{i}$ is a geometric random variable with parameter $\ring \mu_{i}$ because each new item obtained is of type $i$ with probability $\ring \mu_{i}$, but now these variables are no more independent. Since the minimum of $T_{i}$ and $T_{j}$ is the number of trials needed to obtain either item $i$ or item $j$, it follows that for $j \neq i$, $\min(N_{i}, N_{j})$ is a geometric random variable with parameter $\ring \mu_{i} + \ring \mu_{j}$ and the same holds true for the minimum of any finite number of these random variables. Hence, we can write,

\resizebox{0.9\linewidth}{!}{$
\begin{aligned}
    \Ebb[T] &= \Ebb[ \max_{i} T_{i}] \\
    &= \sum_{i} \Ebb[T_{i}] - \sum_{i<j} \Ebb[\min(T_{i}, T_{j})] + \sum_{i<j<k} \Ebb[\min(T_{i}, T_{j}, T_{k})] - \; \dots \; + (-1)^{n+1} \Ebb[\min(T_{1}, \dots, T_{n})] \\
    &= \sum_{i} \frac{1}{\ring \mu_{i}} - \sum_{i<j} \frac{1}{\ring \mu_{i} + \ring \mu_{j}} + \sum_{i<j<k} \frac{1}{\ring \mu_{i} + \ring \mu_{j} + \ring \mu_{k}} - \; \dots \; + (-1)^{n+1} \frac{1}{\ring \mu_{1} + \dots + \ring \mu_{n}}
\end{aligned}
$}

Recall that $ \int_{0}^{\infty} e^{-tx} dx = \frac{1}{t} $. We also know the identity

\begin{align}
1 - \prod_{i=1}^{n}(1 - e^{-t_{i}x}) = \sum_{i} e^{-t_{i}x} - \sum_{i<j} e^{-(t_{i}+t_{j})x} + \; \dots \; + (-1)^{n+1}e^{-(t_{1} + \dots + t_{n})x}
\label{eq:expidentity}
\end{align}

Using the above identity, and integrating it, we get

\begin{equation}
\begin{aligned}
    \Ebb[T] = \int_{0}^{\infty} \bigg( 1 - \prod_{i=1}^{n} \big( 1 - e^{-\ring \mu_{i}x} \big) \bigg) dx
\end{aligned}
\label{eq:arbitraryksample}
\end{equation}

\end{itemize}

\begin{restatable}{lemma}{multipleactionssamplecomplexity}
\label{lemma:multipleactionssamplecomplexity} 
Let the action set be $\Acal = \{a_{0}, \, \dots \, , a_{K} \}$, with a sampling distribution $\mu$ such that $0 < \mu_{\text{min}} \leq \mu_{k} \leq \mu_{\text{max}} < 1 \forAll k \in [K]$. For the case of $\beta, \lambda \to \infty$, with some given $\epsilon \in (0, 1)$, the minimum size of the offline dataset to ensure $\UD = \{A^{\star}\}$ and hence is $(1-\epsilon)$-informative is given by
\begin{equation}
\text{Uniform} \;\, \mu \; : \; N \geq \frac{K^{2} \ln K}{\epsilon} \qquad ; \qquad \text{Arbitrary} \;\, \mu \; : \; N \gtrsim \frac{\ln K}{\mu_{\text{min}}^{2} \epsilon} 
\label{eq:multipleactionssamplecomplexity}
\end{equation}
\end{restatable}

\begin{proof}
\label{proof:multipleactionssamplecomplexity}
We first begin by proving the case of uniform action sampling distribution, and then extend the results to an arbitrary distribution.

\paragraph{Uniform Distribution}. \mbox{} \\

If $|\Ucal| = 1$, then $| \Acal \setminus \Ucal_{N} | = 1 \land |\Wcal_{\Ucal_{N}}| = 0 \;$ \texttt{OR} $\; | \Acal \setminus \Ucal_{N} | = 0 \land |\Wcal_{\Ucal_{N}}| = 1$.

\paragraph{Case 1}. $| \Acal \setminus \Ucal_{N} | = 1$ and $ |\Wcal_{\Ucal_{N}}| = 0$.

For the former, we simply do not want to select the optimal action while making action pairs, and hence the probability is:

$$
P(| \Acal \setminus \Ucal_{N} | = 1) \geq \bigg(1 - \frac{2}{K}\bigg)^{N}
$$

For the latter, we use Equation \eqref{eq:uniformprobpickingall} with $n = \binom{K-1}{2}$ to get

$$
P(|\Wcal_{\Ucal_{N}}| = 0) \geq \sum_{i=0}^{\binom{K-1}{2}} (-1)^i\binom{\binom{K-1}{2}}{i}\left(1-\frac{i}{\binom{K-1}{2}} \right)^N.
$$

\paragraph{Case 2}. $| \Acal \setminus \Ucal_{N} | = 0$ and $ |\Wcal_{\Ucal_{N}}| = 1$.

In this case, if all pairs are sampled at least once, the event $\{ |\Wcal_{\Ucal_{N}}| = 1 \}$ is a sufficient condition for event $\{| \Acal \setminus \Ucal_{N} | = 0 \}$ to occur. Hence, we use Equation \eqref{eq:uniformprobpickingall} with $n = \binom{K}{2}$ to get the probability as:

$$
P(|\Wcal_{\Ucal_{N}}| = 1) \geq \sum_{i=0}^{\binom{K}{2}} (-1)^i\binom{\binom{K}{2}}{i}\left(1-\frac{i}{\binom{K}{2}} \right)^N.
$$

Putting it all together, we need

$$
1 - \epsilon \leq P(| \Acal \setminus \Ucal_{N} | = 1) \cdot P(|\Wcal_{\Ucal_{N}}| = 0) + P(|\Wcal_{\Ucal_{N}}| = 1)
$$

However, the above form is intractable to solve for a closed form solution. Hence, we use the Stirling number approximation for factorials (i.e. $\ln(n!) \approx n\ln(n) - n$) and approximation of the Stirling number of second kind i.e. $S(u,v) \leq n\ln(n) - n\ln(\ln(n)) + n\ln(k)$, where $S(u,v) = \sum_{i=0}^{v} \frac{(-1)^{v-i} i^{u}}{(v-i)! i!}$. In addition, we also Stirling's approximation to the binomial as $\binom{a}{b} \approx \frac{a^{b}}{b!}$ for $a >> b$. Using these, the expression simplifies to $N \geq \frac{K^{2}\ln K}{\epsilon}$.

\paragraph{Arbitrary Distribution}. \mbox{} \\

Similar to the case of uniform distribution, we still need 

$$
1 - \epsilon \leq P(| \Acal \setminus \Ucal_{N} | = 1) \cdot P(|\Wcal_{\Ucal_{N}}| = 0) + P(|\Wcal_{\Ucal_{N}}| = 1)
$$

However, a closed form solution for the above does not exist for the case of multiple actions. Instead, we aim to derive the result based on the sufficient condition for obtaining the optimal action : if all pairs of actions are sampled at least once, we know the optimal action. For this, we just need $P(|\Wcal_{\Ucal_{N}}| = 1) \geq 1-\epsilon$.

Recalling from the analysis given above in Section \ref{proof:generalprobcoupon}, $T_{i}$ denotes the random number of trials needed to sample item $i$ for the first time. The total number of trials needed can be then denoted by the random variable $T = \max(T_{1}, \dots, T_{n})$. 

Now, since $T$ is a random variable denoting the total number of trials needed to obtain all $n$ items \emph{at least} once, it can also be viewed as the \emph{stopping time} for when the agent has collected all items. Hence, we are interested in the probability $P(T \leq N)$ i.e. the probability that this stopping time $T$ is less than the dataset size $N$. This is because the event $\{T \leq N\}$ is the event that by time (or dataset size) $N$, the agent has sampled all $n$ items. 

We then also have $P(T \leq N) = 1 - P(T > N)$. Since $N$ is non-negative, we can bound the $P(T>N)$ using a concentration inequality as below using Equation \eqref{eq:arbitraryksample}.

\begin{align}
P(T > N) &\leq \frac{\Ebb[T]}{N} = \frac{1}{N} \int_{0}^{\infty} \bigg( 1 - \prod_{i=1}^{n} \big( 1 - e^{- \ring \mu_{i}x} \big) \bigg) dx \nonumber \\
\implies P(T \leq N) &\geq 1 - \frac{1}{N} \int_{0}^{\infty} \bigg( 1 - \prod_{i=1}^{n} \big( 1 - e^{- \ring \mu_{i}x} \big) \bigg) dx \\
&\geq 1 - \frac{1}{N} \int_{0}^{\infty} \bigg( \sum_{i} e^{- \ring \mu_{i}x} - \sum_{i<j} e^{-( \ring \mu_{i}+ \ring \mu_{j})x} + \; \dots \; + (-1)^{n+1}e^{-(\ring \mu_{1} + \dots + \ring \mu_{n})x} \bigg) dx \tag{using Identity \eqref{eq:expidentity}} \nonumber \\
&\geq 1 - \frac{1}{N \mu_{\text{min}}^{2}} \bigg( \frac{\binom{n}{1}}{1} - \frac{\binom{n}{2}}{2} + \dots + (-1)^{n+1} \frac{\binom{n}{n}}{n} \bigg) \\
&\geq 1 - \frac{H_{n}}{N \mu_{\text{min}}^{2}}
\label{eq:arbitrarydistprob} 
\end{align}

, where $H_{n}$ is the Harmonic sum of the first $n$ natural numbers. Now, we wish that $P(T \leq N) \geq 1 - \epsilon$. Using the bound above, we find that we need

$$
N \geq \frac{H_{n}}{\mu_{\text{min}}^{2} \epsilon} \gtrsim \frac{\ln K}{\mu_{\text{min}}^{2} \epsilon}.
$$

\end{proof}

\subsection{Understanding Multiple Actions and Finite Deliberateness}
\label{appendix:multipleactionsfinitebeta}

We can break the expected number of samples needed to find an optimal action into parts and then use a generalized version of the Coupon Collection problem, solution of which is known \citep{newman1960double}. The first deals with using Equation \eqref{eq:appendixtwoactionssamplecomplexity} to find the minimum samples needed to determine the more likely optimal action between two actions (one pair) with high probability of $(1-\frac{\epsilon}{2n})$, where $n$ is total number of items. Here, $n$ would be the number of pairs i.e. $n = \binom{K}{2}$. The second part deals with finding the bound on total number of samples needed to determine the more likely optimal action for \emph{every} such pair.

\paragraph{Finding the better action in the $i^{\text{th}}$ item (pair)}. 

The expected number of samples needed to find the better action can be calculated using Equation \eqref{eq:appendixtwoactionssamplecomplexity}. Call this number $k_{i}$. So,

$$
k_{i} \geq \frac{\ln \bigg( \big(\frac{2\binom{K}{2}}{\epsilon}-1 \big) \big(\frac{1}{\Phi(x_{i})}-1 \big) \bigg)}{\beta \langle a_{i}^{(0)} - a_{i}^{(1)}, \theta_{0} \rangle},
$$

where $x_{i} := \frac{(a_{i}^{(0)} - a_{i}^{(1)})^T \mu_{0}}{\sqrt{\big(a_{i}^{(0)} - a_{i}^{(1)}\big)^T \Sigma_{0} \big(a_{i}^{(0)} - a_{i}^{(1)}\big)}}$, $\Phi(\cdot)$ is the CDF of the standard Normal distribution, and $(a_{i}^{(0)}, a_{i}^{(1)})$ are the actions of the $i^{\text{th}}$ pair.

\finalmultipleactionssamplecomplexity*

\begin{proof}
\label{proof:finalmultipleactionssamplecomplexity}

Our sample complexity analysis to achieve no Bayesian regret can be broken down into three main building blocks:
\begin{itemize}[leftmargin=*]
\item Appendix \ref{appendix:twoactions} and Lemma \ref{lemma:twoactionssamplecomplexity} : there are only two actions ($|\Acal| = 2$) but we have finite deliberateness ($\beta < \infty$). 
\item Appendix \ref{appendix:multipleactionsinfinitebeta} and Lemma \ref{lemma:multipleactionssamplecomplexity}: there are many actions ($|\Acal| = K$) but we have very high deliberateness ($\beta \to \infty$).
\item Appendix \ref{appendix:multipleactionsfinitebeta} : there are many actions ($|\Acal| = K$) and finite deliberateness, where we combine the results from the first two cases. In this case, we can break the expected number of samples needed to find an optimal action into two parts. The first deals with using Lemma \ref{lemma:twoactionssamplecomplexity} to find the minimum samples needed to determine the more likely optimal action between two actions (one pair) with high probability of $(1-\frac{\epsilon}{2n})$, where $n$ is total number of items. Here, $n$ would be the number of pairs i.e. $n = \binom{K}{2}$. The second part deals with finding the bound on total number of samples needed to determine the more likely optimal action for \emph{every} such pair.
\end{itemize}

With this in mind, we prove the result below.

\cite{newman1960double} gave a generalization of the coupon collector's problem when $m$ copies of each coupon need to be collected with total coupons being $n$. Let $T_{m}$ be the first time $m$ copies of \emph{each} coupon are collected. We then know that $\Ebb[T_{m}] = n \ln n + (m-1)n \ln(\ln n)$. 

Using similar analysis as before for a general action pair sampling distribution $\mu$ with $\mu_{\text{min}} \leq \mu_{i} \leq \mu_{\text{max}}$ for some $\mu_{\text{min}}, \mu_{\text{max}} \in [0,1)$ for all items (pairs) $i \in [n]$, we can derive the general sampling result. Adapting it to our setting, we need $k_{i}$ samples for $i^{\text{th}}$ pair, and we have $n = \binom{K}{2}$ pairs. Letting $T_{k_{\text{max}}}$ denote the total number of samples needed to obtain $k_{\text{max}}$ number of samples for each item (pair),

$$
\Ebb[T_{k_{\text{max}}}] \leq \frac{1}{\mu_{\text{min}}^{2}} \big[ 2\ln(n) + (k_{\text{max}}-1)\ln(\ln(n)) \big] \quad ; \quad n = \mbinom{K}{2} \, \, , \, \, k_{\text{max}} = \max_{i \in [n]} k_{i}
$$

Denoting $T_{k_{\text{max}}}$ as the random stopping time when at least $k_{\text{max}}$ occurrences of all $n$ items have been collected, we need $P(T_{k_{\text{max}}} > N) \leq \frac{\epsilon}{2}$, where $N$ is the size of the offline dataset $\Dcal_{0}$. Hence, using Markov inequality, we can bound it as:

\begin{align}
N \geq \frac{\ln K + (k_{\text{max}-1})\ln\ln K}{\mu_{\text{min}}^{2} \epsilon} & \quad \text{where}, \\
k_{\text{max}} = \max_{i,j \in [K]} \frac{\ln \bigg( \big(\frac{2 K^{2}}{\epsilon}-1 \big) \big(\frac{1}{\Phi(x_{i,j})}-1 \big) \bigg)}{\beta \langle a_{i} - a_{j}, \theta_{0} \rangle} \; &, \; x_{i,j} = \frac{(a_{i} - a_{j})^T \mu_{0}}{\sqrt{\big(a_{i} - a_{j}\big)^T \Sigma_{0} \big(a_{i} - a_{j}\big)}} \nonumber
\end{align}
    
\end{proof}

\subsection{Regret Analysis Continued}
\label{appendix:regretanalysis}

In this appendix section, we provide the building block proofs that allow us to construct a prior-dependent Bayesian regret bound on the $\mathsf{warmPref-PS}$ algorithm. The heart of these proofs lies in constructing a $(1-\epsilon)$-informative set $\UD$ from the offline dataset $\Dcal_{0}$.

\begin{restatable}{lemma}{f1informative}
\label{lemma:f1informative} 
$\UD$ is $(1-f_{1})-$informative.
\end{restatable}

\begin{proof}

We construct $\UD$ as a \emph{set} of actions that have been preferred to \emph{at least} once in the offline dataset $\Dcal_{0}$ and of actions that do not appear in the $\Dcal_{0}$. Thus, $\UD$ contains at most $K$ actions.

Now, we consider the formulation below. Recall that $\bar{A}_{n}^{(0)}$ and $\bar{A}_{n}^{(1)}$ are i.i.d. sampled from the action set and each datapoint in the dataset $\Dcal_{0}^{i}$, conditioned on $\vartheta, \beta$, is independent of $\Dcal_{0}^{j}$ for $i \neq j$. Now,

\begin{equation}
\begin{aligned}
    P \left(A^{\star} \notin \UD \right)&\leq P(A^{\star} \; \text{has lost all comparisons in} \; \Dcal_{0}) + P(A^{\star} \; \text{is not present in} \; \Dcal_{0}) \\
    &\leq \Ebb \bigg[ \prod_{n=1}^{N} \frac{\exp\big( \beta \langle a_{n}, \vartheta \rangle \big)}{\exp\big( \beta \langle a_{n}, \vartheta \rangle \big) + \exp\big( \beta \langle A^{\star}, \vartheta \rangle \big)} + (1-\mu_{\text{min}})^{2N} \bigg] \\
    &\leq \Ebb \bigg[ \prod_{n=1}^{N} \bigg( 1 - \frac{\exp\big( \beta \langle A^{\star}, \vartheta \rangle \big)}{\exp\big( \beta \langle a_{n}, \vartheta \rangle \big) + \exp\big( \beta \langle A^{\star}, \vartheta \rangle \big)}  \bigg)\bigg]  + (1-\mu_{\text{min}})^{2N} \\
    &\leq \Ebb \bigg[ \prod_{n=1}^{N} \bigg( 1 - \underbrace{\frac{1}{1 + \exp\big( - \beta \langle A^{\star} - a_{n}, \vartheta \rangle \big)}}_{\clubsuit} \bigg) \bigg]  + (1-\mu_{\text{min}})^{2N} 
\label{eq:optimalnotinu}
\end{aligned}
\end{equation} 
, where $A^{\star}$ is a function of $\theta$ and thus a random variable as well. Looking closely at the term $\clubsuit$ above, it can be written as $P(Y_{n} = A^{\star} \given \vartheta)$. We now analyze this term.

\begin{align*}
    P(Y_{n} = A^{\star} \given \vartheta) &= \frac{1}{1 + \exp \big( - \beta \langle A^{\star} - a_{n}, \vartheta \rangle \big)} \\
    &=\big( 1 + \exp\left(\beta \langle A^{\star}-a_{n}, \theta-\vartheta\rangle-\beta \langle A^{\star}-a_{n}, \theta \rangle \right) \big)^{-1} \\
    &\geq \big( 1 + \exp\left(\beta \| A^{\star}-a_{n} \|_{1} \| \theta-\vartheta \|_{\infty} - \beta \langle A^{\star}-a_{n}, \theta \rangle \right) \big)^{-1} \tag{H\"{o}lder's inequality} \\
    &\geq \big( 1 + \exp\left(\beta \| \vartheta-\theta \|_{\infty} - \beta \langle A^{\star}-a_{n}, \theta \rangle \right) \big)^{-1} \tag{$\| A^{\star}-a_{n} \|_{1} \leq 1 \forAll a_{n} \in \Acal$}
\end{align*}

Since $\vartheta-\theta\sim N(0, \Ibf_d/\lambda^2)$, using the Dvoretzky–Kiefer–Wolfowitz inequality bound \citep{massart1990tight, vershynin2010introduction} implies 
\begin{equation*}
     P\left(\|\vartheta-\theta\|_{\infty}\geq t\right)\leq 2 d^{1/2} \exp\left(-\frac{t^2\lambda^2}{2}\right)\,.
\end{equation*}
Set $t=\sqrt{2\ln(2d^{1/2}T)}/\lambda$ and define an event $\Ecal_1:=\{\|\vartheta-\theta\|_{\infty}\leq \sqrt{2\ln(2d^{1/2}T)}/\lambda\}$ such that $P(\Ecal_1^c)\leq 1/T$. We decompose Equation \eqref{eq:optimalnotinu} using Union Bound as:

\begin{equation}
\label{eqn2}
\resizebox{.9\linewidth}{!}{$
\begin{aligned}
     P\left(A^{\star} \notin \UD \right)&\leq \mathbb E\left[ \prod_{n=1}^N\left(1-P\left(Y_n=A^{\star} \Given \theta, \vartheta \right) \right) \Ibb_{\Ecal_1} \right] + P(\Ecal_1^c) + (1-\mu_{\text{min}})^{2N}  \\
     &\leq \mathbb E\left[\prod_{n=1}^N\left(1-\left(1 + \exp\left(\frac{\beta \sqrt{2\ln(2d^{1/2}T)}}{\lambda}\right) \underbrace{\exp\left(-\beta \langle A^{\star}-a_{n}, \theta \rangle \right)}_{\blacktriangle} \right)^{-1} \right) \right]+ \frac{1}{T} + (1-\mu_{\text{min}})^{2N} \,.
\end{aligned}
$}
\end{equation}

Now, we define another event $\Ecal_{(n)}:=\{ \langle A^{\star}-a_{n}, \theta \rangle \leq \Delta \}$. Based on $\Ecal_{(n)}$ we analyze the $\blacktriangle$ term as follows.

\begin{align*}
\exp\left(-\beta \langle A^{\star}-a_{n}, \theta \rangle \right) &= \Ebb \big[ \exp\left(-\beta \langle A^{\star}-a_{n}, \theta \rangle \right) \Ibb_{\Ecal_{(n)}} \big] + \Ebb \big[ \exp\left(-\beta \langle A^{\star}-a_{n}, \theta \rangle \right) \Ibb_{\Ecal_{(n)}^{c}} \big] \\
&\leq \exp\left(0 \right) P(\Ecal_{(n)}) + \exp\left(-\beta \Delta \right) P(\Ecal_{(n)}^{c}) \\
&\leq P(\Ecal_{(n)}) + (1-P(\Ecal_{(n)}))\exp\left(-\beta \Delta \right)
\end{align*}

Plugging this back in Equation \eqref{eqn2} we get,

\begin{equation}
\resizebox{.98\linewidth}{!}{$
\begin{aligned}
 P \left( A^{\star} \notin \UD \right) &\leq \Ebb \left[\prod_{n=1}^N \left(1 - \left(1 + \exp\left(\frac{\beta \sqrt{2\ln(2d^{1/2}T)}}{\lambda}\right) \left( \Ibb_{\Ecal_{(n)}} + (1-\Ibb_{\Ecal_{(n)}}) \exp\left(-\beta \Delta \right) \right) \right)^{-1} \right) \right]+ \frac{1}{T} + (1-\mu_{\text{min}})^{2N} \\
\end{aligned}
$}
\label{eq:eqn3}
\end{equation}

Note that the random variable $\Ibb_{\Ecal_{(n)}}$ depends on the action sampling distribution $\mu$. Denote the probability of sampling this action $a_{n}$ by $\mu_{n}$, and as before we have $\mu$ supported by $[\mu_{\text{min}}, \mu_{\text{max}}]$. We first analyze this for any arbitrary $n \in [N]$ and study the the distribution of $\Ibb_{\Ecal_{(n)}}$ conditionaled on $A^{\star}$. Without loss of generality, we first condition on $A^{\star}= \ring a$ for some $\ring a \in \Acal$. For that, let $\rho(\cdot)$ be the univariate Gaussian distribution and $\theta_{a} = \langle a, \theta \rangle$ for any action $a$.

\begin{equation}
\resizebox{.9\linewidth}{!}{$
\begin{aligned}
    P\left(\Ibb_{\Ecal_{(n)}} = 1 \given A^{\star} = \ring a \right) &= P\left( \Ibb \left(\langle A^{\star} - a_{n}, \theta \rangle \leq \Delta \right) = 1 \Given A^{\star} = \ring a \right) & \\
    &= \frac{1}{P\left(A^{\star} = \ring a \right)}  P\left( \Ibb \left(\langle A^{\star} - a_{n}, \theta \rangle \leq \Delta \right) = 1 \, , \, A^{\star} = \ring a \right) & \\
    &= \frac{1}{ P \left(A^{\star}= \ring a \right)} P \left( \Ibb \left(\theta_{a_{n}} \geq \theta_{\ring a} - \Delta \right) = 1 \, , \, \bigcap_{a \in \Acal} \{\theta_{\ring a} \geq \theta_a \} \right) & \\
    &=\frac{1}{ P\left(A^{\star} = \ring a \right)} \bigintssss_{\Rbb} \left[\int_{\theta_{\ring a} - \Delta}^{\infty} d \rho(\theta) \right] d \rho(\theta_{\ring a}) & \\
    &=\frac{1}{ P\left(A^{\star} = \ring a \right)} \bigintsss_{\Rbb} \left[\int_{\theta_{\ring a} - \Delta}^{\theta_{\ring a}} d \rho(\theta) \right] d \rho(\theta_{\ring a}) \, & (\text{since} \, \theta_{a} \leq \theta_{A^{\star}} \forAll a)
\label{eqn:B_dis}
\end{aligned}
$}
\end{equation}

Noticing that the term inside the integral can be represented as a distribution, we first find a normalizing constant to represent the probabilities. So, define

$$
\Phi(\theta_{\ring a}) =\int_{-\infty}^{\theta_{\ring a}} (2\pi)^{-1/2} \exp(-x^2/2) \; d x \; \qquad ; \qquad g(\theta_{\ring a}) = \frac{1}{\Phi(\theta_{\ring a})} \int_{\theta_{\ring a} - \Delta}^{\theta_{\ring a}} d \rho(\theta) \; .
$$

For fixed $\theta_{\ring a}$, let $X_{\theta_{\ring a}} \sim \text{Bernoulli}(1, g(\theta_{\ring a}))$. With Eq. \eqref{eqn:B_dis} and letting $d \mu(\theta_{\ring a}) = \frac{\Phi(\theta_{\ring a})}{ P(A^{\star} = a_{\ring a})} d \rho(\theta_{\ring a})$ we have,

\begin{equation}
\label{eqn:B_prob}
\resizebox{.75\linewidth}{!}{$
     P\left(\Ibb_{\Ecal_{(n)}} = 1 \given A^{\star} = \ring a \right) = \int_{\Rbb} P(X_{\theta_{\ring a}}=1) \frac{\Phi(\theta_{\ring a})} { P(A^{\star} = \ring a)} d \rho(\theta_{\ring a})=\int_{\Rbb} P(X_{\theta_{\ring a}}=1) \; d \mu(\theta_{\ring a}) \; .
$}
\end{equation}

Plugging this back in Equation \eqref{eq:eqn3} and upper bounding the probabilities we get,

\begin{equation}
\resizebox{.98\linewidth}{!}{$
\begin{aligned}
 P \left( A^{\star} \notin \UD \right) &\leq \sum_{a \in \Acal} 
 \bigintssss_{\Rbb} P(X_{\theta_{a}}=0) d \mu(\theta_a) P\left(A^{\star} = a \right) \left(1 - \left(1 + \exp\left( \beta \left( \lambda^{-1} \sqrt{2\ln(2d^{1/2}T)} - \Delta \right) \right) \right)^{-1} \right)^{N} \cdot \mu_{\text{max}}^{N} \\ & \qquad + \frac{1}{T} + (1-\mu_{\text{min}})^{2N} \\
 &\leq  \sum_{a \in \Acal} \bigintssss_{\Rbb} P(X_{\theta_{a}}=0) \left(1 - \left(1 + \exp\left( \beta \left( \lambda^{-1} \sqrt{2\ln(2d^{1/2}T)} - \Delta \right) \right) \right)^{-1} \right)^{N} \cdot \mu_{\text{max}}^{N} \; d \mu(\theta_a) P\left(A^{\star} = a \right)  \\ & \qquad + \frac{1}{T} + (1-\mu_{\text{min}})^{2N} \\
 &\leq  \bigintssss_{\Rbb} \E{\ring a \, \in \, \Acal} \left[
 \left(1 - \left(1 + \exp\left( \beta \left( \lambda^{-1} \sqrt{2\ln(2d^{1/2}T)} - (1 - X_{\theta_{\ring a}})\Delta \right) \right) \right)^{-1} \right)^{N} \right] \cdot \mu_{\text{max}}^{N} d \mu(\theta_{\ring a})  + \frac{1}{T} + (1-\mu_{\text{min}})^{2N} \, ,
 \label{eq:eq02}
\end{aligned}
$}
\end{equation}

where $\mu_{\text{max}}$ is used to obtain the exponent $N$ by accounting for the sampling distribution $\mu$, and last step follows from the uniformity of each action being optimal. Finally, we need to find the supremum of $g(\theta_{\ring a})$ and hence Equation \eqref{eq:eq02}. Recall that,

$$
g(\theta_{\ring a}) = \frac{1}{\Phi(\theta_{\ring a})} \int_{\theta_{\ring a} - \Delta}^{\theta_{\ring a}} d \rho(\theta) = \frac{\int_{\theta_{\ring a} - \Delta}^{\theta_{\ring a}} d \rho(\theta)}{\int_{-\infty}^{\theta_{\ring a}} d \rho(\theta)} = \frac{\int_{-\infty}^{\theta_{\ring a}} d \rho(\theta) - \int_{-\infty}^{\theta_{\ring a} - \Delta} d \rho(\theta)}{\int_{-\infty}^{\theta_{\ring a}} d \rho(\theta)} = 1 - h_{\Delta}(\ring a) \;
$$

, where $h_{\Delta}(\ring a) := \frac{\int_{-\infty}^{\theta_{\ring a} - \Delta} d \rho(\theta)}{\int_{-\infty}^{\theta_{\ring a}} d \rho(\theta)}$. Setting $\nabla_{\ring a} h_{\Delta}(\ring a) = 0$ and analyzing $\nabla_{\ring a}^{2} h_{\Delta}(\ring a) > 0$, we find that 

\begin{equation}
    g(\theta_{\ring a}) \leq 1 - \Delta \exp \left( - \frac{(2\theta_{\ring a} - \Delta)\Delta}{2} \right) \leq \min(1, \Delta)  \; .
\end{equation}

Finally we, decompose Equation \eqref{eq:eq02} based on the event $\Ecal_{2} := \{ X_{\theta_{\ring a}} = 0 \}$, and upper bound the probability to simplify. Setting $\Delta = \ln(T\beta) / \beta$, we conclude with the following bound:

\begin{equation}
\resizebox{.98\linewidth}{!}{$
\begin{aligned}
 P \left( A^{\star} \notin \UD \right) &\leq  \left(1 - \left(1 + \exp\left( \beta \left( \lambda^{-1} \sqrt{2\ln(2d^{1/2}T)} - (K-1)\min(1, \ln(T\beta) / \beta) \right) \right) \right)^{-1} \right)^{N} + \frac{1}{T} + (1-\mu_{\text{min}})^{2N} 
 \label{eq:eq03}
\end{aligned}
$}
\end{equation}

\end{proof}

\begin{restatable}{lemma}{cardinalitysmall}
\label{lemma:cardinalitysmall} 
$\Ebb[|\UD|] \leq f_{2}$.
\end{restatable}

\begin{proof}

Recall that $\UD$ is a \emph{set} of actions that have been preferred to \emph{at least} once in the offline dataset $\Dcal_{0}$ and of actions that do not appear in the $\Dcal_{0}$. We first see that $\Ebb[|\UD|] = \sum_{k=1}^{K} k \cdot P \left(|\UD| = k \right)$. Define an event $\Ecal_{a} = \{ \langle A^{\star}-a, \theta \rangle \leq \Delta \}$ and analyze as follows,

\begin{equation}
\label{eqn:D_0_1}
\begin{aligned}
    \Ebb [|\UD|] &= \Ebb \left[\sum_{a\in\Acal} \Ibb(a\in \UD)\right] = \sum_{a \in \Acal} \Ebb \left[\Ibb(a \in \UD) \Ibb(\Ecal_{a}) + \Ibb(a \in \UD) \Ibb(\Ecal_{a}^{c}) \right] \\
    &\leq K \min(1,\Delta^{2}/2) + \frac{1}{T} + \Ebb \left[\sum_{a \in \Acal} \Ibb \left(a \in \UD\right)  \Ibb(\Ecal_{a}^{c}) \right] \, ,
\end{aligned}
\end{equation}

where the second step follows from the event $\Ecal_{a}$ and analysis done before : break down the indicator variable conditioning on arbitrary $A^{\star} = \ring a \in \Acal$, and use Poisson approximation to bound the probability. Now, analyze term in expectation above.

\begin{equation}
\label{eqn:D_0_2}
\resizebox{.91\linewidth}{!}{$
\begin{aligned}
     \Ebb \left[ \sum_{a \in \Acal} \Ibb(a \in \UD) \Ibb(\Ecal_{a}^{c}) \right] &= \Ebb \left[ \sum_{n=1}^N \sum_{a \in \Acal} P \left( Y_{n} = a, \langle A^{\star} - a, \theta -\vartheta \rangle + \langle A^{\star}-a, \vartheta \rangle \geq \Delta \Given \theta, \vartheta \right) \Ibb(\Ecal_{a}^{c}) \right] \\
     &\leq \sum_{n=1}^N \Ebb \left[ \sum_{a \in \Acal} P \left(Y_{n} = a \, , \, \langle A^{\star}-a, \vartheta \rangle \geq \Delta -  \sqrt{2\ln(2d^{1/2}T)}/\lambda \Given \vartheta \right)  \Ibb(\Ecal_{a}^{c}) \right] \\
     &\leq N \Ebb \left[ \sum_{a,b \in \Acal \; ; \; \langle A^{\star}-a, \vartheta \rangle \geq \Delta -  \sqrt{2\ln(2d^{1/2}T)}/\lambda} \left(1 + \exp \left( \beta \langle b - a, \vartheta \rangle \right) \right)^{-N} \right] \\
     &\leq N \Ebb \left[ \sum_{a \in \Acal \; ; \; \langle A^{\star}-a, \vartheta \rangle \geq \Delta -  \sqrt{2\ln(2d^{1/2}T)}/\lambda} \left(1 + \exp \left( - \beta \langle A^{\star} - a, \vartheta \rangle \right) \right)^{-N} \right] \\
     &\leq \frac{N(K-1)}{T \beta} \left( 1 + \exp \left( - \beta \left( \lambda^{-1} \sqrt{2\ln(2d^{1/2}T)} + (K-1) \min(1,\Delta) \right) \right) \right)^{-N} \\
     &\leq  \frac{NK}{T \beta} \left( 1 + \exp \left( - \beta \left( \lambda^{-1} \sqrt{2\ln(2d^{1/2}T)} + (K-1) \min(1,\Delta) \right) \right) \right)^{-N}
\end{aligned}
$}
\end{equation}

Putting all of this together, we obtain the bound below.

\begin{equation}
\label{eqn:D_0_3}
\resizebox{.91\linewidth}{!}{$
\begin{aligned}
    \Ebb [|\UD|] & \leq K \min(1,\Delta^{2}/2) + \frac{NK}{T \beta} \left( 1 + \exp \left( - \beta \left( \lambda^{-1} \sqrt{2\ln(2d^{1/2}T)} + (K-1) \min(1,\Delta) \right) \right) \right)^{-N} + \frac{1}{T} \; .
\end{aligned}
$}
\end{equation}

Of course, $|\UD|$ cannot exceed $K$, so we have with the choice of $\Delta = \ln(T\beta)/\beta$,

\begin{equation}
\label{eqn:D_0_4}
\resizebox{.98\linewidth}{!}{$
\begin{aligned}
    \Ebb [|\UD|] &\leq \min \left( K \min \left(1,\frac{\ln^{2}(T\beta)}{2\beta^{2}} \right) + \frac{NK}{T \beta} \left( 1 + \exp \left( - \beta \lambda^{-1} \sqrt{2\ln(2d^{1/2}T)} + (K-1) \min(1,\ln(T\beta)/\beta) \right) \right)^{-N} + \frac{1}{T} , K \right) .
\end{aligned}
$}
\end{equation}

\end{proof}

\priordependentinformationset*
\label{proof:priordependentinformationset}
\begin{proof}
Combining Lemma \ref{lemma:f1informative} and Lemma \ref{lemma:cardinalitysmall} we have the desired result.
\end{proof}

\subsection{Constructing Surrogate Loss Function}
\label{appendix:surrogatelossfunction}

This section contains proofs of construction of the surrogate loss function as described in Section \ref{sec:practicalwarmPref-PS}.

\begin{restatable}{lemma}{appendixmapestimatelemma}
At time $t$, the MAP estimate of $(\theta, \vartheta)$ can be constructed by solving the following equivalent optimization problem: 
\begin{equation}
\begin{aligned}
(\theta_{opt}, \vartheta_{opt})  &= \underset{\theta, \vartheta}{\argmax} \; P(\theta, \vartheta \, | \, \Dcal_{t-1}) \\ & \equiv \underset{\theta, \vartheta}{\argmin} \; \Lcal_{1}(\theta, \vartheta) +  \Lcal_{2}(\theta, \vartheta) +  \Lcal_{3}(\theta, \vartheta) \; , \\
\text{where}, & \qquad \Lcal_{1}(\theta, \vartheta)  := \frac{1}{2} \sum_{s=1}^{t-1} \big(R_s - \langle A_s , \theta \rangle \big)^{2}, \\
& \Lcal_{2}(\theta, \vartheta) := - \sum_{n=1}^{N} \beta \langle \bar{A}_n^{(Y_{n})} , \vartheta \rangle + \ln \bigg(e^{ \beta \langle \bar{A}_n^{(0)}, \vartheta \rangle} + e^{\beta \langle \bar{A}_n^{(1)}, \vartheta \rangle} \bigg),  \\
& \Lcal_{3}(\theta, \vartheta) := \frac{\lambda^2}{2} \norm{\theta - \vartheta}{2}{2} + \frac{1}{2} (\theta - \mu_{0})^{T} \Sigma_{0}^{-1} (\theta - \mu_{0}).
\end{aligned}
\label{eq:appendixmapestimateproblem}
\end{equation}
\end{restatable}

\begin{proof}
\label{proof:mapestimatelemma}

We first analyze the posterior distribution of $\vartheta, \theta$ given the offline dataset $\Dcal_{0}$, optimize it by treating these random variables as parameters.

\begin{equation}
\label{eq:map_surrogate}
\begin{aligned}
    \underset{\theta, \vartheta}{\argmax} \; P(\theta, \vartheta \, | \, \Dcal_{t-1}) &= \underset{\theta, \vartheta}{\argmax} \; P( \Dcal_{t-1} \, | \, \theta, \vartheta) \cdot P(\theta, \vartheta) \\ 
    &= \underset{\theta, \vartheta}{\argmax} \; \ln P(\Dcal_{t-1} \, | \, \theta, \vartheta) + \ln P(\theta, \vartheta) \\ 
    &= \underset{\theta, \vartheta}{\argmax} \underbrace{\ln P(\Hcal_{t-1} \,| \, \Dcal_0, \theta, \vartheta)}_{\Lcal_{1}} + \underbrace{\ln P(\Dcal_0 \, | \, \theta, \vartheta)}_{\Lcal_{2}} +  \underbrace{\ln P(\theta, \vartheta)}_{\Lcal_{3}}
\end{aligned}
\end{equation}

Then,
\begin{equation}
    \begin{aligned}
        \Lcal_{1} &= \sum_{s=1}^{t-1} \underbrace{\ln P(A_{s} \, | \, \Dcal_{s-1}, \theta, \vartheta)}_{\textcolor{gray}{\textbf{indep. of $\theta, \vartheta \implies \textbf{constant}$}}} + \ln P(R_s \, | \, A_s, \theta, \vartheta) \\
        &= \textcolor{gray}{\text{constant}}  \; - \; \frac{t-1}{2} \ln \bigg(\frac{2\pi}{\sigma^2}\bigg) - \frac{1}{2} \sum_{s=1}^{t-1} \big(R_s - \langle A_s , \theta \rangle \big)^{2}. \\
        \Lcal_{2} &= \sum_{n=1}^{N} \ln \bigg( \left(\bar{A}_n^{(0)}, \bar{A}_n^{(1)}, Y_n\right) \, \big| \, \theta, \vartheta \bigg) \\
        &= \sum_{n=1}^{N} \ln \bigg (Y_n \, \big | \, \bar{A}_n^{(0)}, \bar{A}_n^{(1)}, \theta, \vartheta \bigg) + \underbrace{\ln P \bigg( \bar{A}_n^{(0)}, \bar{A}_n^{(1)} \Given \theta, \vartheta \bigg)}_{\textcolor{gray}{\textbf{indep. of $\theta, \vartheta$ \, ; \, depends on $\mu \implies$ constant}}} \\
        &= \sum_{n=1}^{N} \beta \langle \bar{A}_n^{(Y_{n})} , \vartheta \rangle - \ln \bigg(e^{ \beta \langle \bar{A}_n^{(0)}, \vartheta \rangle} + e^{\beta \langle \bar{A}_n^{(1)}, \vartheta \rangle} \bigg) + \textcolor{gray}{\text{\small constant}} \\
        \Lcal_{3} &= \ln P(\vartheta \, | \, \theta) + \ln P(\theta) \\
        &= \frac{d}{2} \ln \bigg(\frac{2\pi}{\lambda^2} \bigg) - \frac{\lambda^2}{2} \norm{\theta - \vartheta}{2}{2} - \frac{1}{2} \ln \big(|2\pi \Sigma_{0}| \big) - \frac{1}{2} (\theta - \mu_{0})^{T} \Sigma_{0}^{-1} (\theta - \mu_{0}).
    \end{aligned}
\end{equation}

Hence, final surrogate loss function is

\begin{equation}
\label{eq:surrogate_loss_function}
\begin{aligned}
    \Lcal(\theta, \vartheta) &= \Lcal_{1}(\theta, \vartheta) +  \Lcal_{2}(\theta, \vartheta) +  \Lcal_{3}(\theta, \vartheta), \qquad \text{where} \\
    \Lcal_{1}(\theta, \vartheta) &= \frac{1}{2} \sum_{s=1}^{t-1} \big(R_s - \langle A_s , \theta \rangle \big)^{2} \\ 
    \Lcal_{2}(\theta, \vartheta) &= - \sum_{n=1}^{N} \beta \langle \bar{A}_n^{(Y_{n})} , \vartheta \rangle + \ln \bigg(e^{ \beta \langle \bar{A}_n^{(0)}, \vartheta \rangle} + e^{\beta \langle \bar{A}_n^{(1)}, \vartheta \rangle} \bigg) \\
    \Lcal_{3}(\theta, \vartheta) &= \frac{\lambda^2}{2} \norm{\theta - \vartheta}{2}{2} + \frac{1}{2} (\theta - \mu_{0})^{T} \Sigma_{0}^{-1} (\theta - \mu_{0}).
\end{aligned}
\end{equation}

Finally the problem in Equation \eqref{eq:map_surrogate} becomes equivalent as follows:
\begin{equation}
\label{eq:surrogate_opt_problem}
(\theta_{opt}, \vartheta_{opt}) = \underset{\theta, \vartheta}{\argmax} \; P(\theta, \vartheta \, | \, \Dcal_{t}) \equiv \underset{\theta, \vartheta}{\argmin} \; \Lcal(\theta, \vartheta)
\end{equation}
    
\end{proof}

\subsection{warmPref-PS with Online Feedback (warmTSOF)}
\label{appendix:warmTSOF}

Here, we present an extension to $\mathsf{warmPref-PS}$, where the agent has the option to ask for feedback during the online phase.
\label{appendix:warmtsof}

\paragraph{Problem Formulation}.

\begin{algorithm}[!t]
   \caption{warm Thompson Sampling with Preference Feedback (\texttt{warmTSOF})}
   \label{alg:prac_warmtsof}
\begin{algorithmic}[1]
   \STATE {\bfseries Input:} Horizon $T$, offline dataset $\Dcal_0$, set of arms $\Acal$, knowledgeability $\lambda$, deliberateness $\beta$, feedback cost $c$.
	 \FOR{$t = 1,2,\dots,T$} 
    	 \STATE Sample a set of perturbations $\Pcal_{t} = \{\zeta_{s}, \omega_{n}, \theta', \vartheta'\}$.
    	 \STATE Solve Equation \eqref{eq:final_surrogate_perturbed_loss} using this set $\Pcal_{t}$ to find $(\hat{\theta}_{t}, \hat{\vartheta}_{t})$.
         \STATE Let $A_{t}^{1} \, , A_{t}^{2}$ be s.t. $\langle A_{t}^{1}, \hat{\theta}_{t} \rangle \geq \langle A_{t}^{2}, \hat{\theta}_{t} \rangle \geq \langle A, \hat{\theta}_{t} \rangle \forAll A \in \Acal \setminus \{A_{t}^{1}, A_{t}^{2}\}$.
         \STATE Compute $\epsilon_{t} = \texttt{get\_epsilon}(c, \Dcal_{t}, t, \lambda, \beta)$.
         \IF{$\big| \langle A_{t}^{1}, \hat{\theta}_{t} \rangle - \langle A_{t}^{2}, \hat{\theta}_{t} \rangle \big| < \epsilon_{t}$}
            \STATE Ask for feedback on $(A_{t}^{1}, A_{t}^{2})$ and receive $Y_{t} \in \{0,1\}$.
            \STATE Update $\Dcal_{t} \leftarrow \Dcal_{t} \cup \{ A_{t}^{1}, A_{t}^{2}, Y_{t}\}$. 
            
            \STATE 
            Update posterior using Equation \eqref{eq:final_surrogate_perturbed_loss} to get new $(\Tilde{\theta_{t}}, \Tilde{\vartheta_{t}})$.
            \STATE Set $A_{t} = \argmax_{a \in \Acal} \langle a, \Tilde{\theta_{t}} \rangle$ and $c_{t} = c$.
        \ELSE
            \STATE Set $A_t = A_{t}^{1}$ and $c_{t} = 0$.
        \ENDIF
         \STATE Take action $A_t$ to receive reward $R_t - c_{t}$.
         \STATE Set $\Dcal_{t+1} = \Dcal_{t} \cup \{A_{t}, R_{t}\}$. 
        
    \ENDFOR
\end{algorithmic}
\end{algorithm}
\begin{figure*}[!ht]
    \centering
    \subfloat[$c=0$]{
        \includegraphics[width=0.35\textwidth]{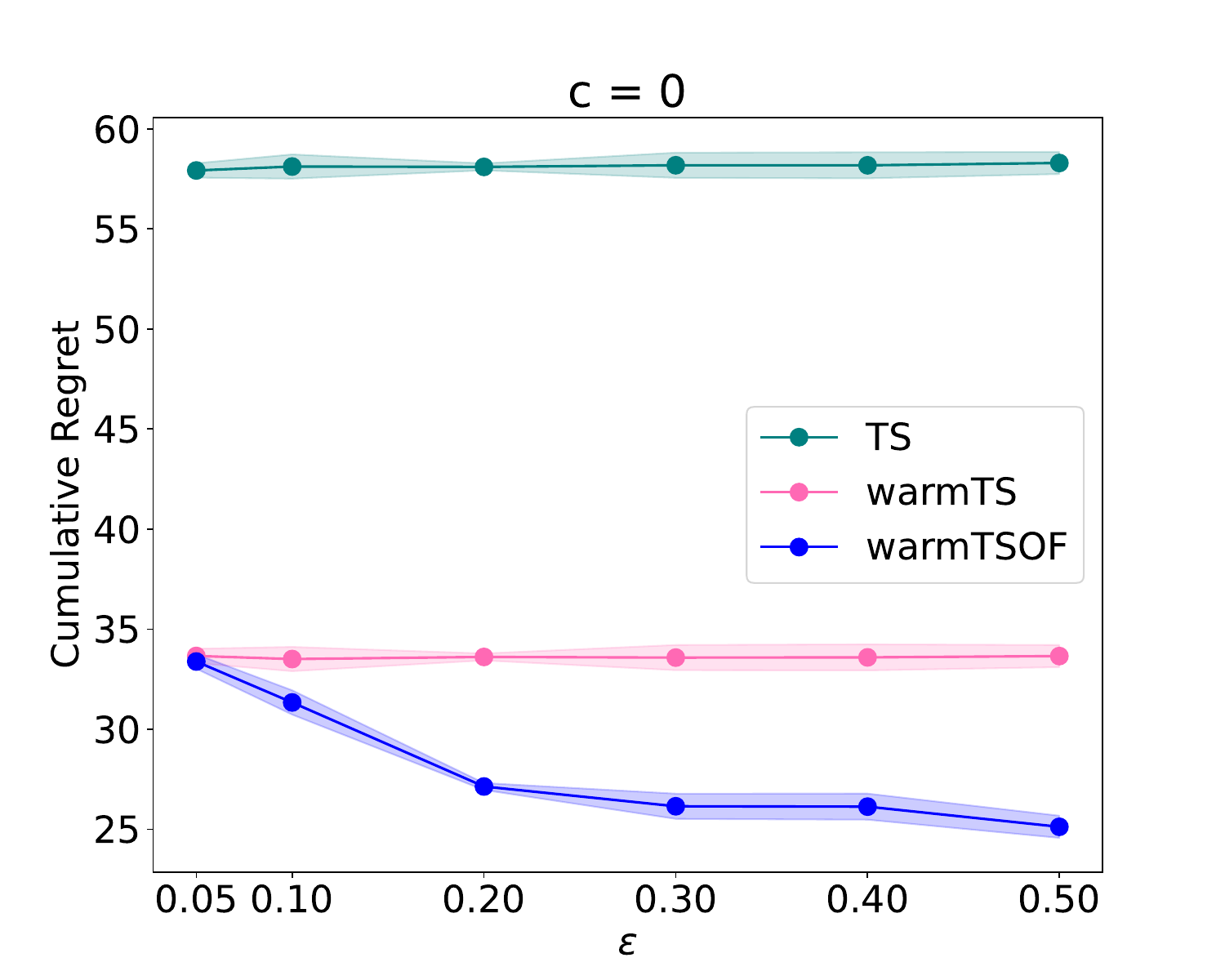}
        \label{fig:wTSOF-Reg-vs-epsilon-c-0}
    }
    \vspace{0.2cm}
    \subfloat[$c=0.1$]{
        \includegraphics[width=0.35\textwidth]{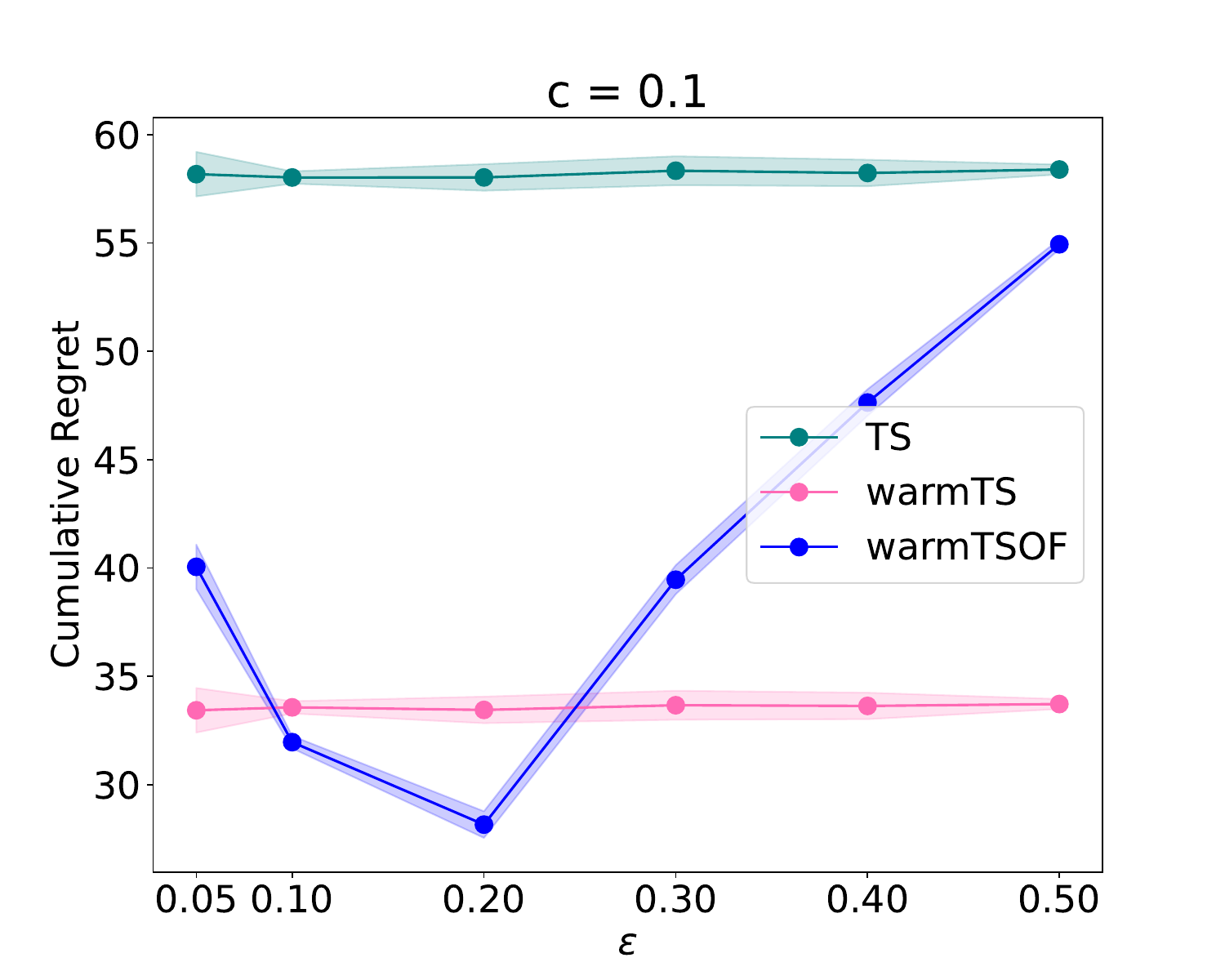}
        \label{fig:wTSOF-Reg-vs-epsilon-c-0.1}
    }
    \caption{Performance of \texttt{warmTSOF} with varying cost of feedback $c$.}
    \label{fig:warmTSOF-ablation}
\end{figure*}

We present the warm Posterior Sampling with Online Feedback (\texttt{warmTSOF}) algorithm wherein the learning agent can ask for preference feedback between two actions at some cost. We present some preliminary empirical results and defer the theoretical analysis to future work. This has applications in active learning \citep{ren2021survey, margatina2021active} and crowd-sourcing data from experts for large language models \citep{mishra2021cross, korbak2023pretraining}. 

Consider, now in addition to Algorithm \ref{alg:practical_Prefwarm-PS} (\texttt{warmTS}), the agent at any time, has the option to ask for \emph{online} preference feedback between two actions. For simplicity, we assume the rater for this feedback is the same rater who generated the offline dataset $\Dcal_{0}$. 

Let the cost incurred for this feedback on actions $A_{t}^{1}$ and $A_{t}^{2}$ be $c_{t} = c \in \Rbb$ if agent asks for feedback, else $c_{t} = 0$. So, this feedback takes the form $\{A_{t}^{1}, A_{t}^{2}, Y_{t} \}$, and $Y_{t} \in \{0,1\}$ and the reward the agent receives then becomes $R_{t}-c_{t}$.

The agent should incorporate the expected current rewards for all actions, cost of feedback, and expert competency into the decision making process. The core idea is to only initiate feedback retrieval process if top-two expected rewards of all actions are `close'. This idea finds its in beginnings in the Top-Two Thompson Sampling procedure \citep{pmlr-v49-russo16}. See \texttt{warmTSOF} (Algorithm \ref{alg:prac_warmtsof}) for exact details. The \texttt{get\_epsilon($\cdot$)} function will be decided through analysis.

\paragraph{Performance}. See Figure \ref{fig:warmTSOF-ablation} for performance comparison. Experiments are run with size of offline dataset $N = 20$, deliberateness $\beta = 10$, and knowledgeability $\lambda = 10$. In addition, we let number of arms $k=10$, dimension of environment $d=4$, and horizon $T=300$, all averaged over 100 runs (random seeds). For baselines, we consider the traditional and warm Thompson Sampling (\texttt{TS} and \texttt{warmTS}).


\subsection{Evaluation using \texttt{DPO} and \texttt{IPO}}
\label{appendix:dpoipo}

DPO \citep{rafailov2024direct} is an alternative approach to the RL paradigm, which avoids the training of a reward model altogether. The loss that DPO optimizes to obtain the optimal policy, given an empirical dataset $\Dcal = \{y_{w}, y_{l}\}$ of the winning (preferred) $y_{w}$ and losing (not preferred) $y_{l}$ outputs (arms in our bandit setting), as a function of the reference policy $\pi_{\text{ref}}$ and regularization strength $\tau \in \Rbb_{+}$, is given by:

\begin{align*}
    \pi^{\star}_{\text{DPO}} = \argmin_{\pi} \quad \Ebb_{(y_{w}, y_{l}) \sim \Dcal} \left[ - \log \sigma \left( \tau \log \left( \frac{\pi(y_{w})}{\pi(y_{l})} \right)  - \tau \log \left( \frac{\pi_{\text{ref}}(y_{w})}{\pi_{\text{ref}}(y_{l})} \right) \right)  \right]
\end{align*}

, where $\sigma(\cdot)$ denotes the sigmoid function.

IPO is an instance of the $\Psi\text{PO}$ algorithm \citep{ipo} . The loss function that IPO optimizes is given by,

\begin{align*}
    \pi^{\star}_{\text{IPO}} = \argmin_{\pi} \quad \Ebb_{(y_{w}, y_{l}) \sim \Dcal} \left[ h_{\pi}(y_{w}, y_{l}) - \frac{1}{2 \tau}  \right]^{2} \; \text{where} \, , \; h_{\pi}(y, y') := \log \left( \frac{\pi(y) \pi_{\text{ref}}(y')}{\pi(y') \pi_{\text{ref}}(y)} \right) \, .
\end{align*}

Note here that DPO and IPO are purely offline learning algorithms that only work with preference datasets. DPO and IPO cannot be trivially extended to our problem setting i.e. fixed offline preference dataset with active online numerical reward learning. Comparing DPO and IPO based solely on offline datasets is not fair. Hence, for a more fair comparison, we consider an offline-online variant of DPO with $\epsilon$-greedy online exploration. The pseudo code is as follows:

\begin{itemize}
\item \textbf{Input:} offline preference data $\Dcal_{0}$, $\epsilon$, $\min_a r^{\star}(a)$, DPO parameters
\item \textbf{Offline learning:} use DPO to learn a policy $\pi$, use $\pi$ to infer a reward model $r$ such that $\min_{a \in \Acal} r(a) = \min_{a \in \Acal} r^{\star}(a)$, where $r^{\star}(\cdot) : \Rbb^{d} \to \Rbb$ is true reward model.
\item \textbf{Online learning:} at each time $t=1, \ldots, T$
\begin{itemize} 
\item with probability $\epsilon$, choose $A_t$ uniformly randomly; otherwise, choose $A_t = \argmax_a r(a)$.
\item observe reward from the environment, which is $r^{\star}(A_t)$ plus noise.
\item update the reward model $r$ based on the received reward.
\end{itemize}
\end{itemize}

For training, mini‑batches are drawn uniformly with replacement from $\Dcal_{0}$ and optimized using \texttt{DPO} and \texttt{IPO} for 20k steps. Policy is encoded simply as $\pi_{\psi}(a_{i}) = \text{softmax}(\bm{\psi})_{i}$ for an action $a_{i} \in \Acal$ using a vector $\bm{\psi} \in \Rbb^{K}$, and is optimized for 20k steps using Adam \citep{kingma2014adam} with a learning rate of 0.015 and mini-batch size 12. Reference policy is $\pi_{\text{ref}}$ is chosen to be uniform over the action space, regularization is set at 0.1, and $\epsilon=0.16$ for best results.

\subsection{Ablation study (cont.)}
\label{appendix:ablations} 

\begin{figure*}[!ht]
    \centering
    \subfloat[Varying $K$]{\includegraphics[height=0.2\textwidth, width=0.23\textwidth]{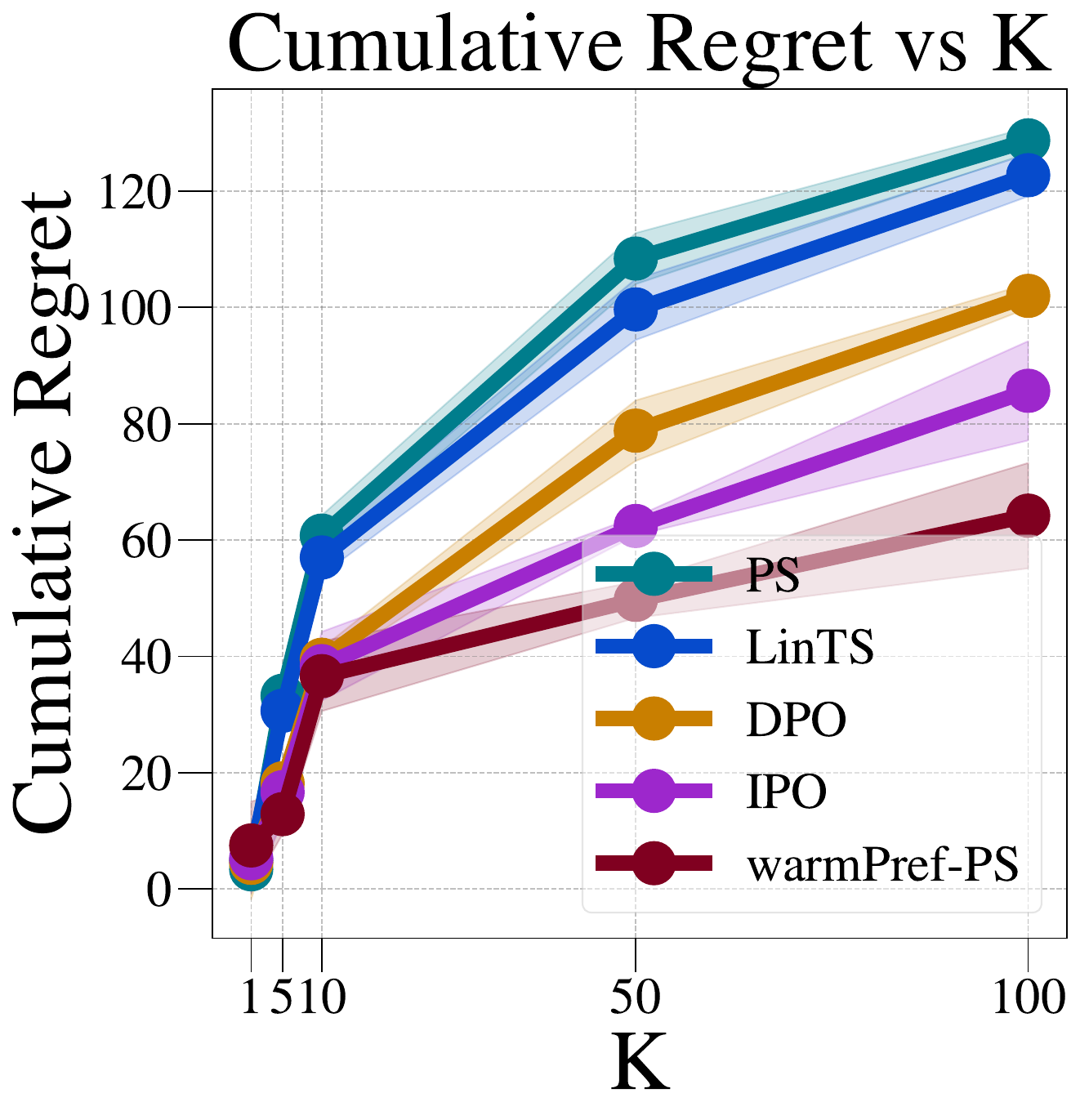}} 
    \subfloat[Varying $T$]{\includegraphics[height=0.2\textwidth, width=0.23\textwidth]{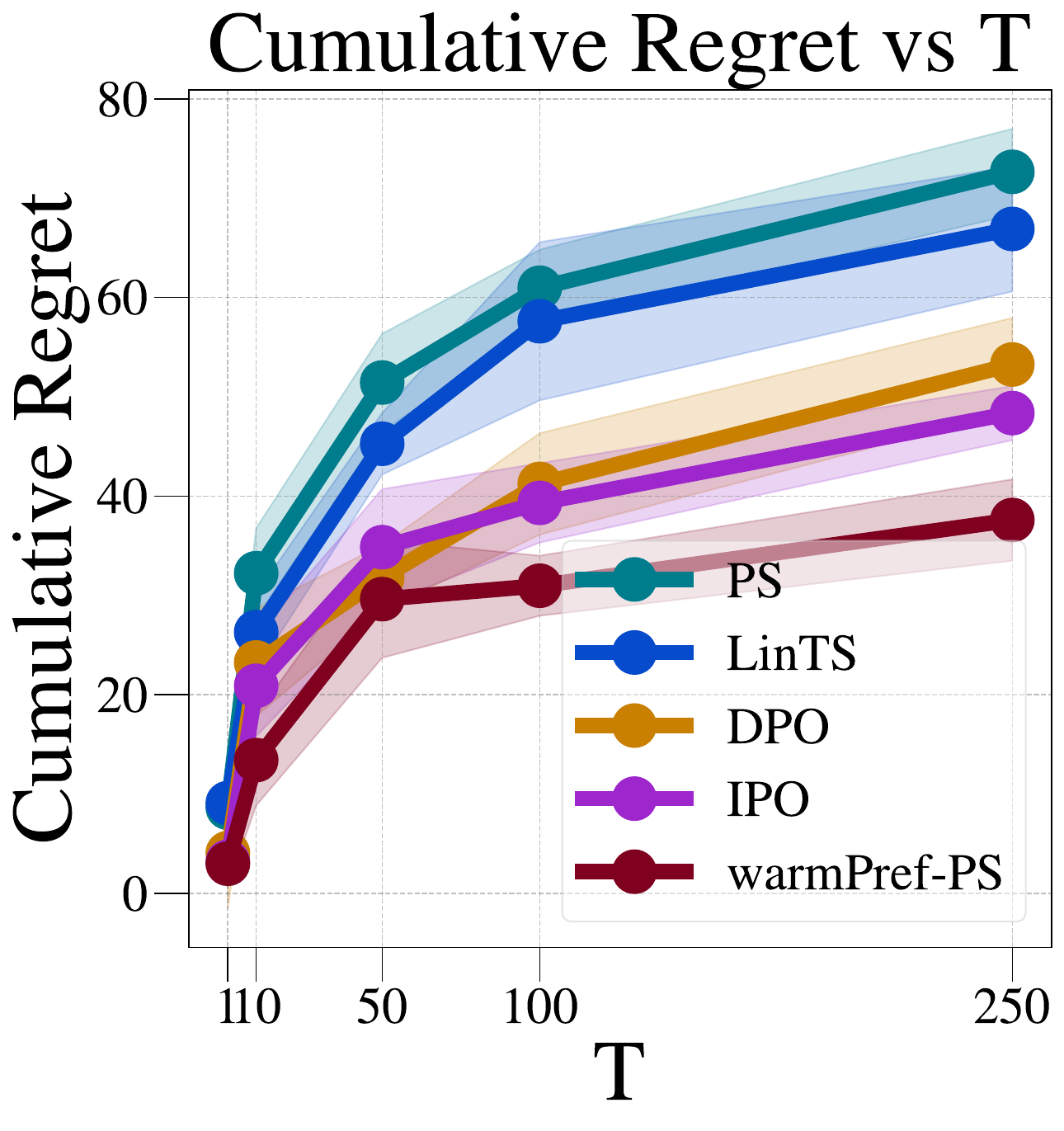}} 
    \caption{Cumulative regret with varying $K$ and $T$.}
    \label{fig:kt_ablation}
\end{figure*}

\paragraph{Effect of number of arms $K$ and online rounds $T$}. Here, we study how cumulative regret scales with $K$ and $T$. For evaluation, we let $d=3, \lambda=100, \beta=8,$ and $N=25$. Empirical cumulative regret values are averaged over 5 runs with independent seeds. See Figure \ref{fig:kt_ablation} for results.

\paragraph{Effect of Action Space Dynamics}. We next study how the dynamics of the action space affect cumulative regret. Specifically, how (i) the relationship between action pairs measured by their correlation ($\rho$), and (ii) the dimensionality of the environment vector $\theta \in \Rbb^{d}$, affect cumulative regret. Table \ref{table:actionspaceregret} shows that the performance of all these posterior sampling methods degrades as dimensionality of the environment and correlation between action increases. However, $\mathsf{warmPref-PS}$ still outperforms the baselines and enjoys a lesser performance degradation than PS as $d$ and $\rho$ increase.

\begin{table}[ht] 
\centering
\fontsize{8}{10}\selectfont
\addtolength{\tabcolsep}{-0.2em}
\caption{Effect of dimensionality and correlation within the action space on cumulative regret. 
}
\begin{tabular}{cccc}
\hline
   & \cellcolor{TS!35}PS & \cellcolor{LinTS!35}\texttt{LinTS} & \cellcolor{warmTS!35}$\mathsf{warmPref-PS}$ \\
\hline
$d=2, \rho=0.1$ & 58.21 $\pm$ 0.45 &  53.23 $\pm$ 0.64 &\cellcolor{cellg}{\textbf{32.65 $\pm$ 1.78}} \\
$d=2, \rho=0.8$ & 61.36 $\pm$ 1.23 &  56.32 $\pm$ 0.97 & \cellcolor{cellg}{\textbf{33.98 $\pm$ 3.07}} \\
$d=5, \rho=0.1$ & 60.42 $\pm$ 0.82 &  55.71 $\pm$ 0.41 & \cellcolor{cellg}{\textbf{34.12 $\pm$ 3.05}} \\
$d=5, \rho=0.8$ & 64.21 $\pm$ 1.57 & 59.55 $\pm$ 1.35 & \cellcolor{cellg}{\textbf{34.77 $\pm$ 2.94}} \\
\hline
\end{tabular}
\label{table:actionspaceregret}
\end{table}

\chapter{Hybrid Reinforcement Learning with Preferences}
\label{chapter:PSPL}

\paragraph{Abstract}.
Reinforcement Learning from Human Feedback (RLHF) has emerged as a powerful approach for aligning generative models, but its reliance on learned reward models makes it vulnerable to mis-specification and reward hacking. Preference-based Reinforcement Learning (PbRL) offers a more robust alternative by directly leveraging noisy binary comparisons over trajectories. We study the best policy identification problem in PbRL, motivated by post-training optimization of generative models, for example, during multi-turn interactions. Learning in this setting combines an offline preference dataset—potentially biased or out-of-distribution and collected from a rater of subpar `competence'—with online pure exploration, making systematic online learning essential. To this end, we propose Posterior Sampling for Preference Learning ($\mathsf{PSPL}$), a novel algorithm inspired by Top-Two Thompson Sampling that maintains posteriors over the reward model and dynamics. We provide the first Bayesian simple regret guarantees for PbRL and introduce an efficient approximation that outperforms existing baselines on simulation and image generation benchmarks.

\section{Introduction}
\label{sec:chapter4introduction}

\textcolor{black}{RLHF has recently become a cornerstone of aligning large generative models with human intent, enabling advances in natural language processing and other domains. However, the standard RLHF pipeline depends on learning a reward model from human annotations, which introduces two critical limitations: misspecification of the reward function and susceptibility to reward hacking \citep{amodei2016concrete, novoseller2020dueling, tucker2020preference}. These issues stem from the inherent difficulty of reducing complex human objectives to a scalar reward, even when learned from data \citep{sadigh2017active, wirth2017survey}.
}

\textcolor{black}{
PbRL provides an appealing alternative by relying on comparative rather than absolute feedback, thereby offering a more direct and robust signal of human intent \citep{christiano2017deep, saha2023dueling, metcalf2024sample}. In many settings, such preferences are expressed over entire trajectories rather than isolated (final) outcomes, which is especially important in human–robot interactions \citep{wirth2013preference, casper2023open, dai2023safe}, experimentation \citep{novoseller2020dueling}, and recommendation systems \citep{bengs2021preference, kaufmann2023survey}. Similarly, for generative systems, trajectory-level feedback captures the multi-turn nature of interactions more faithfully: when users engage with a large language model (LLM), satisfaction often emerges only after a sequence of exchanges \citep{shani2024multi, zhou2024archer}.
}

\textcolor{black}{
Despite its promise, PbRL in practice often begins with offline preference datasets that are collected off-policy. These datasets are prone to biases and out-of-distribution (OOD) limitations, which can degrade generalization and lead to subpar performance of post-trained models \citep{zhang2024policy, ming2024does}. This raises a fundamental question: given an offline dataset of trajectory comparisons, how should one systematically supplement it with online preference data to maximize learning efficiency under a constrained budget? This question is particularly important for fine-tuning generative models with human feedback, where large-scale offline data are typically available but careful online exploration can provide higher-value information \citep{mao2024context, zhai2024fine}.}

\textcolor{black}{
In this work, we propose a systematic framework for leveraging offline datasets to bootstrap online RL algorithms. We demonstrate that, as expected, incorporating offline data consistently improves online learning performance, as reflected in reduced simple regret. More interestingly, when the agent is further informed about the `competence' of the rater generating the offline feedback—equivalently, the behavioral policy underlying the dataset—the resulting informed agent achieves substantially lower simple regret. Finally, we establish that as the rater competence approaches to that of an expert, higher competence levels yield progressively sharper reductions in simple regret compared to baseline methods.
}

\begin{figure}[ht]
\vspace{-0.5cm}
    \centering
    \subfloat[{\fontfamily{lmss}\selectfont MountainCar}  ]{
        \includegraphics[height=0.2\textwidth, width=0.22\textwidth]{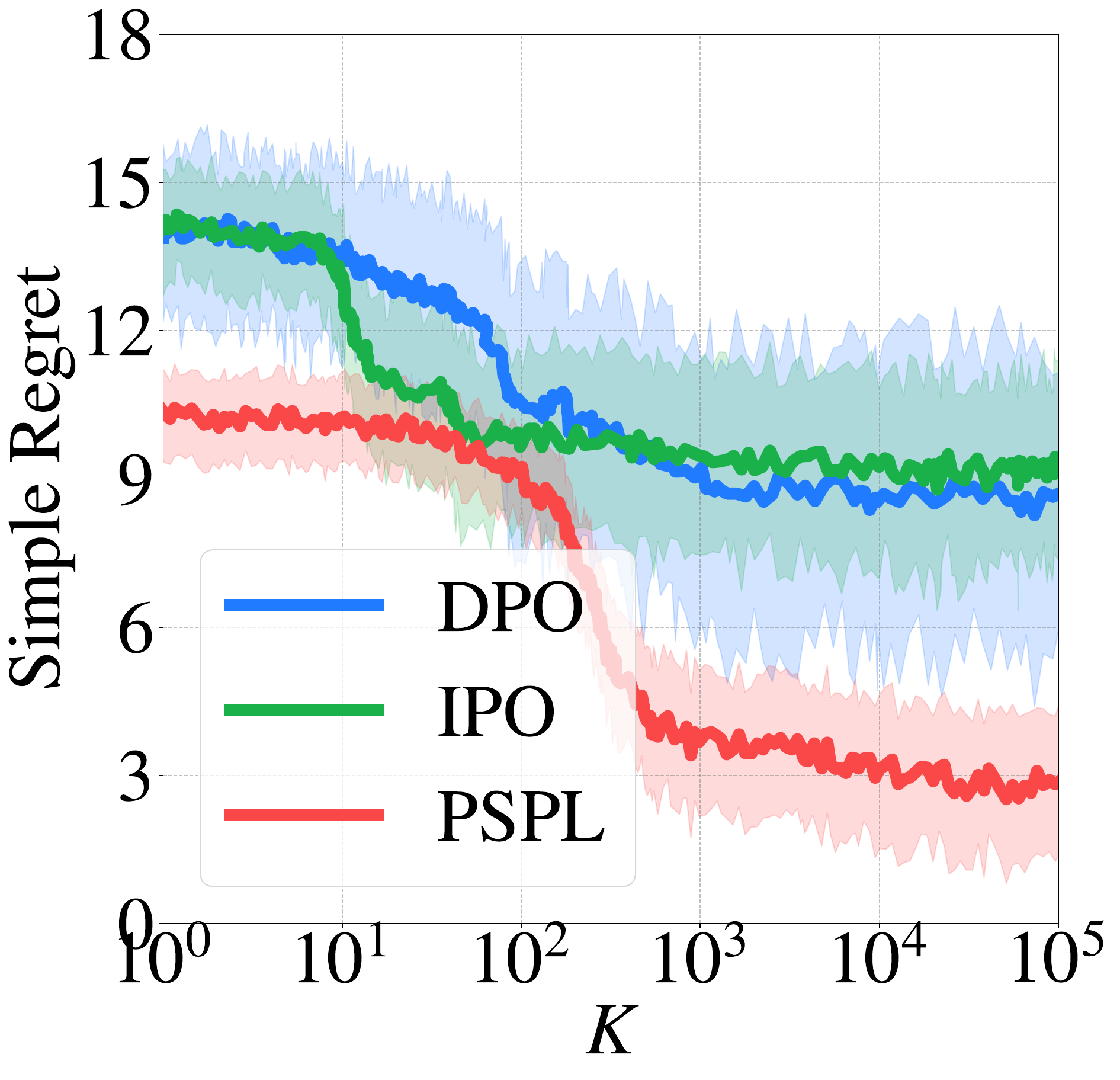}
    } 
    \subfloat[{\fontfamily{lmss}\selectfont RiverSwim}]{
        \includegraphics[height=0.2\textwidth, width=0.22\textwidth]{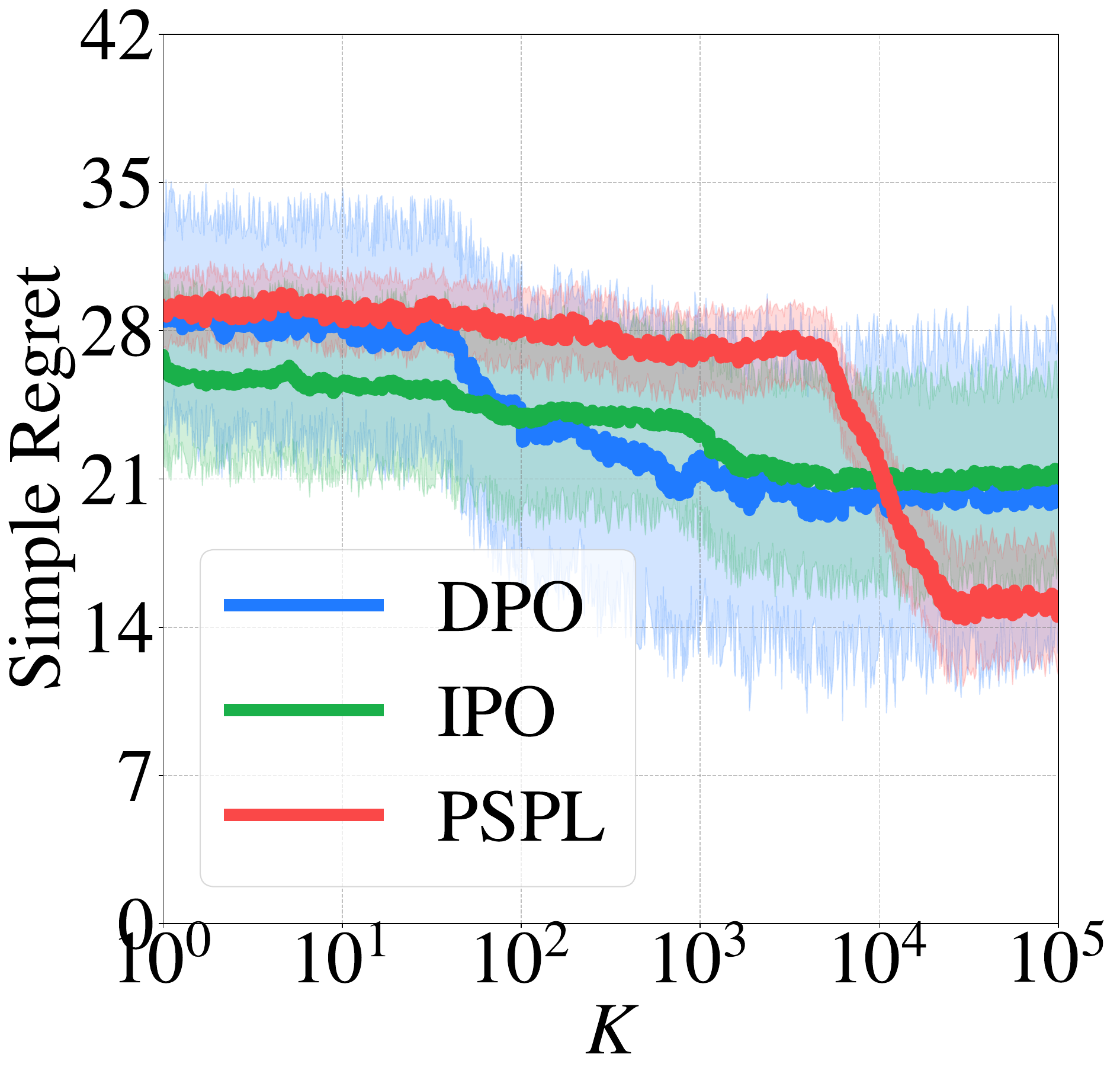}
    } 
\caption{Comparison of $\PSPL$ with current state-of-the-art \emph{offline} finetuning algorithms, DPO and IPO, in two benchmark environments. Online finetuning is necessary for BPI. See Appendix \ref{sec:appendix} for more details.}
\label{fig:intro}
\end{figure}

\textcolor{black}{
\textbf{Relevance to RLHF.} Our problem setting also connects to  LLM alignment, wherein `best policy identification' (BPI) has emerged as a critical objective. Unlike cumulative regret  minimization, which tries to balance exploration and exploitation, BPI ensures that pure exploration leads to the final policy that achieves the highest possible alignment quality \citep{ouyang2022training, chung2024scaling}. Moreover, BPI aligns more naturally with human evaluation processes, which often prioritize assessing final system performance over intermediate learning behaviors. This perspective highlights the importance of optimizing simple regret rather than cumulative regret, which aggregates losses over time \citep{saha2023dueling, ye2024theoretical, zhang2024iterative}. See Figure \ref{fig:intro} for a comparison of $\PSPL$ with Direct Preference Optimization (DPO) \citep{rafailov2024direct} and Identity Preference Optimization (IPO) \citep{ipo}, which shows how on-policy online finetuning helps in BPI.
}

In this paper, we address the problem of BPI for an unknown episodic MDP where both the transition dynamics and reward functions are unknown. We assume that some offline data is available in the form of preferences over pairs of trajectories. And we can then collect additional data by generating pairs of trajectories and seeking contrastive feedback between the two. At the end, we output the best policy we can learn. 

Our main contributions are: (i) we introduce a formal framework for best policy learning using offline and online trajectory-preference feedback, (ii) we present a top-two Posterior Sampling algorithm $\PSPL$, the first best policy learning algorithm for this setting and present theoretical bounds on its simple regret, and (iii) we introduce a computationally practical version of $\PSPL$ and show that it has excellent empirical performance on benchmark environments as compared to baselines.   

\textbf{Related Work.} There has been a fair amount of research for preference-based bandits (contextual and linear) \citep{singh2002optimizing, bengs2021preference, busa2014survey, saha2021optimal, saha2022versatile, scheid2024optimal, cheung2024leveraging}. However, even contrary to classical RL (cumulative regret or best policy identification) with numerical reward feedback, only a few
works consider incorporating preference feedback in the RL and Bayesian optimization framework \citep{swamy2023inverse, talebi2018variance, zhu2023principled, ye2024theoretical, song2022hybrid, gonzalez2017preferential}. Training RL agents from trajectory-level feedback, available only at the end of each episode, is particularly challenging. \cite{novoseller2020dueling} and \cite{saha2023dueling} analyze finite $K$-episode cumulative regret for this PbRL problem, where two independent trajectories are run, and binary preference feedback is received per episode. While \cite{novoseller2020dueling} assumes a weaker Gaussian process regression model, \cite{saha2023dueling} proposes an optimism-based algorithm that is computationally infeasible due to exponential growth in state space variables. \cite{liu2023efficient} presents a regression-based empirical approach but lacks theoretical guarantees and exploration. \cite{xu2024provably, chen2024order} address offline PbRL with fixed state-action preference datasets, without considering dataset quality. \cite{ye2024theoretical, xiong2023gibbs, li2024policy} take a game-theoretic perspective on offline PbRL in RLHF but assume state-action preferences, avoiding credit assignment from trajectory feedback. Hybrid approaches include \cite{agnihotri2024online}, which studies PbRL in linear bandits, and \cite{hao2023bridging}, which incorporates numerical rewards in the online phase. \textit{To our knowledge, ours is the first work on BPI in unknown MDPs using trajectory-level preference feedback from a subpar expert.}

\section{Preliminaries}
\label{sec:preliminaries_pspl}

\textcolor{black}{
Consider a $K$-episode, $H$-horizon Markov Decision Process (MDP) setup $\mathcal{M} := (\Pbb_{\eta}, \Scal, \Acal, H, r_{\theta}, \rho)$, where $\Scal$ is a finite state space, $\Acal$ is a finite action space, $\mathbb \Pbb_{\eta}(\cdot \mid s,a)$ are the fixed MDP transition dynamics parameterized by $\eta$ given a state-action pair $(s,a) \in \Scal \times \Acal$, $H \in \Nbb$ is the length of an episode, $r_{\theta}(\cdot)$ is underlying reward model parameterized by $\theta \in \Rbb^{d}$, and $\rho$ denotes the initial distribution over states. We let $S := |\Scal|$ and $A := |\Acal|$ to be the cardinalities of the state and action spaces respectively. In addition, we denote the learner by $\Upsilon$, and let $[N] := \{1, \dots, N\}$ for $N \in \Nbb$.
}

\textcolor{black}{
Under the above MDP setup, a policy $\pi = \{\pi_{h}\}_{h=1}^{H}$ is a sequence of mappings from the state space $\Scal$ to the the probability simplex $\Delta(\Acal)$ over the actions. Specifically, $\pi_h(s, a)$ denotes the probability of selecting action $a$ in state $s$ at step $h$. In addition, $\Pi$ is an arbitrary policy class against which performance of the learner will be measured. Finally, we denote a trajectory by concatenation of all states and actions visited during $H$ steps $\tau:= (s_1, a_1, \cdots , s_{H}, a_H)$. At the start of each episode, we assume $s_1 \sim \rho$. For any given $(\theta, \eta)$ pair of reward-transition parameters, the Bellman update equation for a policy $\pi = \{\pi_h(\cdot ; \theta, \eta) \}_{h=1}^{H}$ and for $(s,a, h) \in \Scal \times \Acal \times [H]$ takes the form:
\begin{equation}
\begin{aligned}
{Q}_{{\theta}, {\eta}, h}^{\pi}(s, a) = r_{{\theta}}(s,a) + \Ebb_{s' \sim \Pbb_{{\eta}}( \cdot \given s,a)} \left[{V}_{{\theta}, {\eta}, {h+1}}^{\pi} (s') \right],
\end{aligned}
\label{eq:approx_bellman}
\end{equation}
with ${V}_{{\theta}, {\eta}, h}^{\pi}(s) = \Ebb_{a \sim \pi_{h}(\cdot | s)} [{Q}_{{\theta}, {\eta}, h}^{\pi}(s, a)]$ for $h \leq H$ and ${V}_{{\theta}, {\eta}, H+1}^{\pi}(s) = 0 \forAll s \in \Scal$.
}

\textbf{Trajectory embedding.} For any trajectory $\tau$ we assume the existence of a trajectory embedding function $\phi : \Gamma  \rightarrow \mathbb{R}^d$, where $\Gamma$ is the set of all possible trajectories of length $H$ and the map $\phi$ is known to the learner. A special case, which we study in this work, is a decomposed embedding, where $\phi(\tau) =  \sum_{h=1}^H  \phi(s_h, a_h)$ and $\phi: \mathcal{S} \times \mathcal{A} \rightarrow \mathbb{R}^d$ is a mapping from state-action pairs to $\mathbb{R}^d$. For all trajectories $\tau \in \Gamma$, we assume that $\| \phi(\tau) \|_{1} \leq B$ for some constant $B > 0$.

\textbf{Preference Modeling.} The learner has access to a rater of arbitrary `competence'.  When presented with two trajectories $\tau_{0}$ and $\tau_{1}$, the rater provides feedback in terms of a Bernoulli random variable $Y \in \{0,1\}$, where $Y=0$  if the rater prefers $\tau_{0}$ to $\tau_{1}$, else $Y=1$ if $\tau_{1}$ is preferred to $\tau_{0}$. Note here that we are working with the setting of sparse feedback which is awarded on a trajectory-level.

We let $\theta \in \Rbb^{d}$ to be an unknown environment reward vector that the rater has \emph{limited} knowledge of. The rater provides preference feedback based on their `competence' and their knowledge of the environment reward vector $\theta$. The `competence' of the rater is characterized by two factors: (i) $\beta \geq 0$ is a measure of the \emph{deliberateness} or \emph{surety} of the rater's decision, and (ii) $\lambda > 0$ controls the degree of \emph{knowledgeability} of the rater about $\theta$ (see Remark \ref{rem:lambda}) . We define $\vartheta \sim$ $N\left(\theta, \mathbf{I}_{d} / \lambda^2\right)$ ($\mathbf{I}_{d}$ is a $d \times d$ identity matrix) as the rater's estimate of the true environment vector $\theta$. Pairwise comparison between two trajectories $\tau_{0}$ and $\tau_{1}$ from the rater is assumed to follow a Bradley-Terry model \citep{bradleyterry1952}, i.e.,

\small
\begin{align}
\begin{split}
\label{eq:pref_logistic_trajectories}
  \Pr ( Y=0 \given \tau_0, \tau_1 \, ; \, \vartheta )  &= \sigma( \beta \langle \phi(\tau_0) - \phi(\tau_1), \vartheta \rangle ), 
 \end{split}
\end{align}
\normalsize
where $\sigma: \Rbb \mapsto [0,1]$ is the logistic link function, i.e., $\sigma(x) = (1+e^{-x})^{-1}$. This naturally leads to the definition of `rater score' of a trajectory $\tau$ as $g_{\beta, \vartheta}(\tau) := \beta \langle \phi(\tau), \vartheta \rangle $, which is of course dependent on the rater's competence.

\begin{remark}
\label{rem:lambda}
    Intuitively, the parameter $\beta \geq 0$ is a measure of the \emph{deliberateness} of the rater's decision: $\beta=0$ means the rater's decisions are uniformly random, whereas as $\beta \to \infty$ means the rater's decisions are greedy with respect to the trajectory scores. Secondly, $\lambda$ is the rater's estimate of the true environment reward model based on its knowledgeability i.e., as $\lambda \to \infty$, $\vartheta \to \theta$. In the context of RLHF alignment, $\lambda$ can be seen as controlling the degree of alignment between a user and the general population from which preferences are aggregated.
\end{remark}

\textbf{Offline Dataset.} There is an initial \emph{offline preference} dataset $\mathcal{D}_0$, which is generated by the rater. This offline dataset of size $N$ is a sequence of tuples of the form 
$
\mathcal{D}_0=\left((\bar{\tau}_n^{(0)}, \bar{\tau}_n^{(1)}, \bar{Y}_n)\right)_{n \in [N]},
$ 
where $\bar{\tau}_n^{(0)}, \bar{\tau}_n^{(1)} \in \Gamma$ are two sampled trajectories, and $\bar{Y}_n \in\{0,1\}$ indicates the rater's preference.

\textbf{Learning Objective.} \textcolor{black}{Consider an MDP $\mathcal{M}$ with unknown transition model $\Pbb_{\eta}(s' \given s,a ) : \Scal \times \Acal \times \Scal \to [0,1]$, parameterized by $\eta$, and an unknown ground-truth reward function $r_{\theta}(s,a) : \Scal \times \Acal \to [0,1]$, parameterized by $\theta$, with $s,s' \in \Scal$ and $a \in \Acal$. One conceivable idea is to assume $r_{\theta}(\tau)$ as the function $r_{\theta}(\tau):= \langle \phi(\tau), \theta \rangle$, where $\theta \in \Rbb^{d}$ is an unknown reward parameter. 
}

The goal of the learner is to identify the optimal policy that maximizes trajectory rewards. Overload notation to denote trajectory rewards by $r_{\theta}(\tau) := \sum_{h=1}^{H} r_{\theta}(s_{h},a_{h})$. Then denote an optimal policy by $\pi^{\star} \in \argmax_{\pi \in \Pi} \E{\tau \sim \pi}[r_{\theta}(\tau)]$. Consider an optimal policy $\pi^{\star}$ and any arbitrary policy $\pi \in \Pi$. Then, with $\tau^{\star} \sim \pi^{\star}$ and $\tau \sim \pi$, the simple Bayesian regret of $\pi$ after $K$ online episodes is defined as:

\vspace{-0.2cm}
\begin{equation}
    \SR_{K}^{\Upsilon}(\pi, \pi^{\star}) := \E{\tau,\tau^{\star}}[ r_{\theta}(\tau^{\star}) - r_{\theta}(\tau) ]
\label{eq:simple_regret_def}
\end{equation}
\vspace{-0.5cm}

In this paper, the objective of the learner $\Upsilon$ then is to design an exploration based online learning algorithm that is \emph{informed} from the offline preference dataset and that which minimizes simple Bayesian regret in Equation \eqref{eq:simple_regret_def}.

\vspace{-0.5cm}
\textcolor{black}{
\begin{remark}
    One motivation for using pure exploration in the online phase of BPI, rather than active learning approaches that target reward model estimation, is its relevance to RLHF. In practice, online active learning with preference data is infeasible \citep{chung2024scaling, achiam2023gpt, anil2023palm}, since preferences are typically collected in fixed offline batches and used to train reward models that then guide online learning (e.g., the AutoRater pipeline \citep{anil2023palm}). Thus, pure exploration better reflects this setting, where offline batched preference data is available and only minimal online interaction is possible to learn the best policy.
\end{remark}
}
\section{The PSPL Algorithm}
\label{sec:chapter4_algorithm}

For the purpose of pure exploration, the learner $\Upsilon$ is given the opportunity to learn online and generate an \emph{online} dataset of trajectories for which it asks for feedback from the rater. Denote the \emph{online} dataset available to the learner at the beginning of episode $k \in [K]$ as $\mathcal{H}_{k} = \{(\tau_{t}^{(0)}, \tau_{t}^{(1)}, Y_{t})\}_{t=1}^{k-1}$. Then the total available dataset for the learner at the beginning of episode $k$ becomes $\Dcal_{k} = \Dcal_{0} \oplus \Hcal_{k}$, where $\oplus$ denotes concatenation.

Since the parameters $\theta$ and $\eta$ of the reward and transition model are unknown,  
we assume that the prior distribution over the reward vector $\theta$ is a Gaussian
distribution $\nu_{0} \sim \mathcal{N}(\mu_{0}, \Sigma_{0})$ and over the transition model $\eta$ is a Dirichlet distribution $\chi_{0} \sim \mathrm{Dir(\bm{\alpha}_{0}})$ for each state-action pair, where $\bm{\alpha}_{0}$ is a positive real-valued vector of dimension $S$.


Now, note that the offline dataset \emph{informs} the learner $\Upsilon$ of the MDP transition dynamics and the underlying reward model. This is captured in the \emph{informed} prior (before starting the online phase) over the unknown parameters $\eta$ and $\theta$. Denote the probability distributions over the transition parameter and reward parameter by $\chi(\eta)$ and $\nu(\theta)$ respectively. 

\begin{algorithm}[t]
   \caption{Top-two Posterior Sampling for Preference Learning ($\PSPL$)}
\begin{algorithmic}[1]
   \STATE {\bfseries Input:} Initial dataset $\Dcal_{0}$, prior on $\theta$ as $\nu_{0}(\theta)$ and $\eta$ as $\chi_{0}(\eta)$, horizon $H$, episodes $K$.
   \STATE Construct informed prior $\nu_{1}(\theta)$ from Equation \eqref{eq:theta_informed_prior} and $\chi_{1}(\eta)$ from Equation \eqref{eq:eta_informed_prior}.
    \FOR{$k = 1,2, \dots ,K$} 
    \STATE Sample $\hat{\eta}_{k}^{(0)}, \hat{\eta}_{k}^{(1)} \sim \chi_{k}(\eta)$ and $\hat{\theta}_{k}^{(0)}, \hat{\theta}_{k}^{(1)} \sim \nu_{k}(\theta)$.
    \STATE Compute policies $\pi_{k}^{(0)}$ using $(\hat{\eta}_{k}^{(0)}, \hat{\theta}_{k}^{(0)})$ and $\pi_{k}^{(1)}$ using $(\hat{\eta}_{k}^{(1)}, \hat{\theta}_{k}^{(1)})$.
    \STATE Run two trajectories $\tau_{k}^{(0)} \sim \pi_{k}^{(0)}$ and $\tau_{k}^{(1)} \sim \pi_{k}^{(1)}$ for $H$ horizon.
    \STATE Get feedback $Y_{k}$ on $\tau_{k}^{(0)}$ and $\tau_{k}^{(1)}$, and append to dataset as $\Dcal_{k} = \Dcal_{k-1} \oplus \left(\tau_{k}^{(0)}, \tau_{k}^{(1)}, Y_{k} \right)$. 
    \STATE Update posteriors to get $\nu_{k+1}(\theta)$ and $\chi_{k+1}(\eta)$.
    \ENDFOR
    \STATE {\bfseries Output:} \textcolor{black}{Optimal policy $\pi_{K+1}^{\star}$ computed using MAP estimate from $\chi_{K+1}(\eta)$ and $\nu_{K+1}(\theta)$}.
\end{algorithmic}
\label{alg:main_algo_theoretical}
\end{algorithm}

\textbf{Informed prior for \texorpdfstring{$\theta$}{theta}.}
Denoting the \emph{uninformed} prior as $\nu_{0}(\theta)$, we have the \emph{informed} prior $\nu_{1}(\theta)$ as,

\vspace{-0.3cm}
\begin{equation}
\begin{aligned}
     \nu_{1}(\theta) \; :=  \; \nu(\theta \given \Dcal_{0}) \; \propto \; \Pr(\Dcal_{0} \given \theta) \nu_{0}(\theta)  \propto \; \prod_{n=1}^{N} \Pr(\bar{Y}_n \given \bar{\tau}_n^{(0)}, \bar{\tau}_n^{(1)}, \theta) \; \Pr(\bar{\tau}_n^{(0)} \given \theta)   \; \Pr(\bar{\tau}_n^{(1)} \given \theta) \; \nu_{0}(\theta), 
\end{aligned}
\label{eq:theta_informed_prior}
\end{equation}

where the second step follows from Equation \eqref{eq:pref_logistic_trajectories}, and that $\bar{\tau}_n^{(0)}$ and $\bar{\tau}_n^{(1)}$ are assumed independent given $\theta$, \textcolor{black}{as is in the context of RLHF, where outputs (trajectories) are conditionally independent given the prompt}. It is worth emphasizing that the offline dataset carries information about the reward parameter through the terms $\Pr( \cdot \given \theta)$, which incorporates information about the expert’s policy, and thus improves the informativeness of the prior distribution.

\textbf{Informed prior for \texorpdfstring{$\eta$}{eta}.}
Denoting the \emph{uninformed} prior as $\chi_{0}(\eta)$, we have \emph{informed} prior $\chi_{1}(\eta)$ as,

\vspace{-0.3cm}
\begin{equation}
\begin{aligned}
    \chi_{1}(\eta) \; :=  \; \chi(\eta \given \Dcal_{0}) \;  \propto \; \Pr(\Dcal_{0} \given \eta) \, \chi_{0}(\eta) \;  \propto \; \prod_{n=1}^{N} \prod_{j=0}^{1} \prod_{h=1}^{H-1}  \Pbb_{\eta} \left(\bar{s}_{n,h+1}^{(j)} \given \bar{s}_{n,h}^{(j)}, \bar{a}_{n,h}^{(j)} \right) \; \chi_{0}(\eta),
\end{aligned}
\label{eq:eta_informed_prior}
\end{equation}
\vspace{-0.4cm}

where $\bar{\tau}_{n}^{(j)} := \{\bar{s}_{n,1}^{(j)}, \bar{a}_{n,1}^{(j)}, \dots, \bar{s}_{n,H}^{(j)} \}$ for $j \in \{0,1\}$ is an offline trajectory of length $H$. Note here that the proportional sign hides the dependence on the offline dataset generating policy (since we do not make any assumptions on this behavourial policy) and the fact that $\bar{Y}_{n}$ is conditionally independent of dynamics given $\bar{\tau}_n^{(0)}$, $\bar{\tau}_n^{(1)}$.

After constructing informed priors from the offline preference dataset above, the learner begins the online phase for active data collection using pure exploration. The learner maintains posteriors over the true reward and transition kernels, which inherently permit for exploration. In each episode, using samples from these posteriors, the learner computes two policies using value / policy iteration or linear programming, and rolls out two $H$-horizon trajectories. \textcolor{black}{Posteriors are updated based on the trajectory-level preference feedback, and final policy output is constructed from Maximum-A-Posteriori (MAP) estimate from $\chi_{K+1}(\eta)$ and $\nu_{K+1}(\theta)$, as shown in Algorithm \ref{alg:main_algo_theoretical}}.

\begin{remark}
\label{remark:ps_over_optimism}
Although there exist optimism-based algorithms which construct confidence sets around the reward and transition kernels, it is known that posterior sampling based algorithms provide superior empirical performance \citep{ghavamzadeh2015bayesian, ouyang2017learning, liu2023efficient}. In addition, for our problem setting of best policy identification with noisy trajectory level feedback, posterior sampling is a natural method to incorporate beliefs about the environment.
\end{remark}

\section{Theoretical Analysis of PSPL}
\label{sec:chapter4-analysis}

This section focuses on regret analysis of $\PSPL$. The analysis has two main steps: (i) finding a \emph{prior-dependent} upper regret bound in terms of the sub-optimality of any optimal policy estimate $\hat{\pi}^{\star}$ constructed from $\Dcal_{0}$. This part characterizes the online learning phase by upper bounding simple Bayesian regret in terms of the estimate $\hat{\pi}^{\star}$ constructed by $\PSPL$ before the online phase, and  (ii) describing the procedure to construct this $\hat{\pi}^{\star}$ based on the attributes of $\Dcal_{0}$, such as size $N$ and rater competence $(\lambda, \beta)$. Proofs of results, if not given, are provided in Appendix \ref{sec:appendix}.

\subsection{General prior-dependent regret bound}

It is natural to expect some regret reduction if an offline preference dataset is available to warm-start the online learning. However, the degree of improvement must depend on the `quality' of this dataset, for example through its size $N$ or rater competence $(\lambda, \beta)$. Thus, analysis involves obtaining a prior-dependent regret bound, which we obtain next.

\vspace{-0.4cm}
\textcolor{black}{
\begin{restatable}{lemma}{psplerrorregret}
\label{lemma:pspl_error_regret}
For any confidence $\delta_{1} \in (0,\frac{1}{3})$, let $\delta_{2} \in (c,1)$ with $c \in (0,1)$, be the probability that any optimal policy estimate $\hat{\pi}^{\star}$ constructed from the offline preference dataset $\Dcal_{0}$ is $\varepsilon$-optimal with probability at least $(1-\delta_{2})$ i.e., $\Pr \left( \E{s \sim \rho} \left[ V_{\theta,\eta,0}^{\pi^{\star}}(s) -  V_{\theta,\eta,0}^{\hat{\pi}^{\star}}(s) \right] > \varepsilon \right) < \delta_{2}$. Then, the simple Bayesian regret of the learner $\Upsilon$ is upper bounded with probability at least $1-3\delta_{1}$ by, 
\begin{equation}
    \label{eq:pspl_prior_error_bound}
    \begin{aligned}
        \SR_{K}^{\Upsilon}(\pi^{\star}_{K+1}, \pi^{\star}) \leq \sqrt{ \frac{10 \delta_{2}  S^{2}AH^{3} \ln \left( \frac{2KSA}{\delta_{1}}  \right) + 3SAH^{2}\varepsilon^{2}}{2K \left(1 + \ln \frac{SAH}{\delta_{1}} \right) - \ln \frac{SAH}{\delta_{1}}} }
    \end{aligned}
    \end{equation}
\end{restatable}
}
\vspace{-0.2cm}

Please see Appendix \ref{proof:pspl_error_regret} for proof. This lemma tells us that if the learner can construct an $\varepsilon$-optimal policy estimate $\hat{\pi}^{\star}$ with a probability of $(1-\delta_{2})$, then the learner's simple regret decreases as $\delta_{2}$ decreases. Next, we describe how to incorporate information from $\Dcal_{0}$ before the online phase.

\subsection{Incorporating offline preferences for regret analysis}

Before we begin to construct $\hat{\pi}^{\star}$, we define the state visitation probability $p_{h}^{\pi}(s)$ for any given policy $\pi$. Construction of $\hat{\pi}^{\star}$ will revolve around classification of states in a planning step based on $p_{h}^{\pi}(s)$, which will lead to a reward free exploration strategy to enable learning the optimal policy.


\begin{definition}[\textcolor{black}{State Visitation Probability}]
\label{def:visitation}
    Given $(h,s)\in [H]\times \mathcal{S}$, the state (occupancy measure) and state-action visitation probabilities of a policy $\pi$ is defined for all $h' \in [h-1]$ as follows:
    \begin{equation}
    \begin{aligned}
        p_{h}^{\pi}(s) &= \Pr \left( s_h = s \given s_1 \sim \rho, a_{h'} \sim \pi(s_{h'})\right) \\
        p_{h}^{\pi}(s,a) &= \Pr \left( s_h = s, a_h = a \given s_1 \sim \rho, a_{h'} \sim \pi(s_{h'}) \right) \, .
    \end{aligned}
    \end{equation}
\end{definition}

Let $p_{\min} = \min_{h,s} \max_{\pi}  p_{h}^{\pi}(s)$, and we assume it is positive. Further, define the infimum probability of any reachable state under $\pi^{\star}$ as $p_{\min}^{\star} := \min_{h,s} p_{h}^{\pi^{\star}}(s)$ and we assume it positive as well.



\textcolor{black}{
We now describe a procedure to construct an estimator $\hat{\pi}^{\star}$ of the optimal policy  from $\mathcal{D}_0$ such that it is $\varepsilon$-optimal with probability $\delta_{2}$. For each ($\theta, \eta$), define a deterministic Markov policy $\pi^{{\star}}(\theta, \eta) = \{\pi_h^{{\star}}(\cdot ; \theta, \eta) \}_{h=1}^{H}$, 
\begin{equation}
    \begin{aligned}
    \pi_h^{{\star}}(s ; \theta, \eta) =    \begin{cases}
        \arg\max_a Q_{\theta, \eta, h}(s, a), & \text{if } p_{h}^{\pi^{\star}}(s) > 0 \\
        \mathring{a} \sim \mathrm{Unif}(\Acal), &\text{if } p_{h}^{\pi^{\star}}(s) = 0,
    \end{cases}
    \end{aligned}
\end{equation}
}
\textcolor{black}{where $Q_{\theta, \eta, h}(\cdot, \cdot)$ is the Q-value function in a MDP with reward-transition parameters $(\theta, \eta)$, and the tiebreaker for the argmax operation is based on any fixed order on actions}. It is clear from construction that $\{\pi_{h}^{\star}(\cdot ; \theta, \eta)\}_{h=1}^{H}$ is an optimal policy for the MDP with parameters ($\theta, \eta$). Furthermore, for those states that are impossible to be visited, we choose to take an action uniformly sampled from $\Acal$. 

\textbf{Construction of $\hat{\pi}^{\star}$.} To attribute preference of one trajectory to another, we build `winning' ($U_{h}^{W}$), and `undecided' ($U_{h}^{U}$) action (sub)sets for each state in the state space for each time step $h \in [H]$. To achieve this, we define the net count of each state-action pair from the offline dataset $\Dcal_{0}$. Recall that any offline trajectory $\bar{\tau}_{n}^{(\cdot)}$ is composed of $(\bar{s}_{n,1}^{(\cdot)}, \bar{a}_{n,1}^{(\cdot)}, \dots, \bar{s}_{n,H}^{(\cdot)})$. Then if the `winning' counts are defined as $w_{h}(s,a) = \sum_{n=1}^{N} \Ibf \{\bar{s}_{n,h}^{(\bar{Y_{n}})} = s , \bar{a}_{n,h}^{(\bar{Y_{n}})} = a \}$, and `losing' counts are defined as $l_{h}(s,a) = \sum_{n=1}^{N} \Ibf \{\bar{s}_{n,h}^{(1-\bar{Y_{n}})} = s , \bar{a}_{n,h}^{(1-\bar{Y_{n}})} = a \}$, then the `net' counts are defined as $c_{h}(s,a) = w_{h}(s,a) - l_{h}(s,a)$. The action sets are then constructed as,
$$
U_{h}^{W}(s) = \{a ; c_{h}(s,a) > 0 \} \,,  U_{h}^{U}(s) = \{a ; c_{h}(s,a) \leq 0 \}
$$

By construction, if any state-action pair does not appear in $\Dcal_{0}$, it is attributed to the undecided set $U_{h}^{U}$. Finally, for some fixed $\delta \in (0,1)$, construct the policy estimate $\hat{\pi}^{\star} := \{\hat{\pi}_{h}^{\star}\}_{h=1}^{H}$ as 

\vspace{-0.3cm}
\begin{equation}
\begin{aligned}
    \hat{\pi}_{h}^{\star}(s) := 
    \begin{cases}
        \argmax_{a \in U_{h}^{W}(s)} c_{h}(s,a) & \text{if } \sum_{a'} c_{h}(s,a') \geq \delta N \\
        \mathring{a} \sim \mathrm{Unif}(U_{h}^{U}) & \text{if } \sum_{a'} c_{h}(s,a') < \delta N \\
    \end{cases} \nonumber
\end{aligned}
\end{equation}
\vspace{-0.3cm}

To ensure that for any state-time pair $(s,h)$ with $p_{h}^{\pi^{\star}}(s) > 0$, the optimal action $\pi_{h}^{\star}(s ; \cdot) \in U_{h}^{W}(s)$ with high probability, we first need an upper bound on the `error' probability of the rater w.r.t any trajectory pair in the offline dataset $\Dcal_{0}$. Here, the `error' probability refers to the probability of the event in which the rater prefers the sub-optimal trajectory (w.r.t. the trajectory score $g(\cdot)$). Hence, for $n \in [N]$, define an event $\Ecal_{n} := \left\{\bar{Y}_{n} \neq \argmax_{i \in \{0,1\}} g_{\beta, \vartheta}\left (\bar{\tau}_{n}^{(i)} \right) \right\}$, i.e., at the $n$-th index of the offline preference dataset, the rater preferred the suboptimal trajectory. Then,

\begin{lemma}
\label{lemma:rater_error_upper_bound}
Given a rater with competence $(\lambda, \beta)$ such that $\beta > \frac{2\ln \left(2 d^{1/2} \right)}{\left| B\lambda^{2} - 2 \Delta_{\min}  \right|}$ and with $\Delta_{\min}$ such that $\Delta_{\min} = \min_{n \in [N]} \left| r_{\theta}(\bar{\tau}_{n}^{(0)}) - r_{\theta}(\bar{\tau}_{n}^{(1)}) \right|$, and an offline preference dataset $\Dcal_{0}$ of size $N>2$, we have
\begin{equation}
\begin{aligned}
    \Pr (\Ecal_{n} \given \beta, \vartheta) \leq \exp\left(- \beta B \sqrt{2\ln(2d^{1/2}N)}/\lambda - \beta \Delta_{\min} \right)  + \frac{1}{N} \; := \; \gamma_{\beta, \lambda, N} \nonumber
\end{aligned}
\end{equation}
\end{lemma}
\vspace{-0.3cm}

See Appendix \ref{proof:pspl_final_regret} for proof. This lemma establishes the relationship between the competence of a rater (in terms of $(\lambda, \beta)$) and the probability with which it is sub-optimal in its preference. Observe that as the rater tends to an expert i.e., $\lambda, \beta \to \infty$ and we have increasing offline dataset size $N$, we get $\Pr (\Ecal_{n} \given \beta, \vartheta) \to 0$ i.e., the rater \emph{always} prefers the trajectory with higher rewards in an episode, which is intuitive to understand. Using this Lemma \ref{lemma:rater_error_upper_bound}, we can upper bound the simple Bayesian regret of $\PSPL$ as below.

\begin{restatable}{theorem}{psplfinalregretbound}
\label{lemma:pspl_final_simple_regret}
For any confidence $\delta_{1} \in (0,\frac{1}{3})$ and offline preference dataset size $N>2$, the simple Bayesian regret of the learner $\Upsilon$ is upper bounded with probability of at least $1-3\delta_{1}$ by, 
    \begin{equation}
    \begin{aligned}
        &\SR_{K}^{\Upsilon}(\pi^{\star}_{K+1}, \pi^{\star}) \leq \sqrt{ \frac{20 \delta_{2}  S^{2}AH^{3} \ln \left( \frac{2KSA}{\delta_{1}}  \right)}{2K \left(1 + \ln \frac{SAH}{\delta_{1}} \right) - \ln \frac{SAH}{\delta_{1}}} } \;\; \\ ,\text{with} & \; \delta_{2} = 2 \exp \left( - N \left( 1 + \gamma_{\beta, \lambda, N} \right)^{2} \right) + \exp \left( - \frac{N}{4} (1 - \gamma_{\beta, \lambda, N})^{3} \right). \nonumber
    \end{aligned}
    \end{equation}
\end{restatable}

See Appendix \ref{proof:pspl_final_regret} for the proof. For a fixed $N>2$, and large $S$ and $A$, the simple regret bound is $\widetilde{\Ocal}\left( \sqrt{S^{2}AH^{3}K^{-1}} \right)$. Note that this bound converges to zero exponentially fast as $N \to \infty$ and as the rater tends to an expert (large $\beta, \lambda$). In addition, as the number of episodes $K$ gets large, $\PSPL$ is able to identify the best policy with probability at least $(1-3\delta_{1})$.

\vspace{-0.3cm}
\textcolor{black}{
\begin{remark}
(i) This paper is different from existing hybrid RL works which usually consider some notion of `coverability' in the offline dataset $\Dcal_{0}$ \citep{wagenmaker2022instance, song2022hybrid}. We instead deal directly with the estimation of the optimal policy, which depends on the visits to each state-action pair in $\Dcal_{0}$. Rather than dealing with concentratability coefficients, we deal with counts of visits, which provides a simple yet effective way to incorporate coverability of the offline phase into the online phase. If the coverability (counts of visits) is low in $\Dcal_{0}$, the failure event in Lemma \ref{lemma:rater_error_upper_bound} has a higher upper bound, which directly impacts the simple regret in Theorem \ref{lemma:pspl_final_simple_regret} negatively. (ii) \textcolor{black}{Since bandit models are a special case of our setting, results specific for them are presented in Appendix \ref{sec:appendixbandits}.}
\end{remark}
}


\section{A Practical Approximation}
\label{sec:practical_approx_pspl}

The $\PSPL$ algorithm introduced above assumes that posterior updates can be solved in closed-form. In practice, the posterior update in equations \eqref{eq:theta_informed_prior} and \eqref{eq:eta_informed_prior} is challenging due to loss of conjugacy. Hence, we propose a novel approach using Bayesian bootstrapping to obtain approximate posterior samples. The idea is to perturb the loss function for the maximum a posteriori (MAP) estimate and treat the resulting point estimate as a proxy for an exact posterior sample.

We start with the MAP estimate problem for $(\theta, \vartheta, \eta)$ given the offline and online dataset $\Dcal_{k}$ at the beginning of episode time $k$. We show that this is equivalent to minimizing a particular surrogate loss function described below. For any parity of trajectory $j \in \{0,1\}$ during an episode $k$, we denote the estimated transition from state-action $s^{(j)}_{k,h}, a^{(j)}_{k,h}$ to state $s^{(j)}_{k,h+1}$ by $\Pr_{\eta} \left({s}_{t,h+1}^{(j)} \given {s}_{t,h}^{(j)}, {a}_{t,h}^{(j)} \right)$, where $\Pr_{\eta}(\cdot)$ is updated based on counts of visits. Note here that as transition dynamics are independent of trajectory parity, we update these counts based on all trajectories in $\Dcal_{k}$. See Appendix \ref{proof:chapter4mapestimatelemma} for proof. 

\begin{restatable}{lemma}{chapter4mapestimatelemma}
\label{th:chapter4mapestimatelemma} 
At episode $k$, the MAP estimate of $(\theta, \vartheta, \eta)$ can be constructed by solving the following equivalent optimization problem: 
\begin{equation}
\resizebox{\linewidth}{!}{$
\begin{aligned}
(\theta_{opt}, \vartheta_{opt}, \eta_{opt})  &= \underset{\theta, \vartheta, \eta}{\argmax} \; \Pr(\theta, \vartheta, \eta \, | \, \Dcal_{k}) \\ & \equiv \underset{\theta, \vartheta, \eta}{\argmin} \; \Lcal_{1}(\theta, \vartheta, \eta) +  \Lcal_{2}(\theta, \vartheta, \eta) +  \Lcal_{3}(\theta, \vartheta, \eta) \; , \, \text{where},  \\
 \Lcal_{1}(\theta, \vartheta, \eta)  &:= - \sum_{t=1}^{k-1} \vphantom{\int_1^2} \left[ \beta \langle {\tau}_t^{(Y_{t})} , \vartheta \rangle - \ln \bigg(e^{ \beta \langle {\tau}_t^{(0)}, \vartheta \rangle} + e^{\beta \langle {\tau}_t^{(1)}, \vartheta \rangle} \bigg) + \sum_{j=0}^{1} \sum_{h=1}^{H-1} \ln \Pr_{\eta}\left({s}_{t,h+1}^{(j)} \given {s}_{t,h}^{(j)}, {a}_{t,h}^{(j)} \right) \vphantom{\int_1^2} \right], \\
 \Lcal_{2}(\theta, \vartheta, \eta) &:= - \sum_{n=1}^{N} \left[ \beta \langle \bar{\tau}_n^{(\bar{Y}_{n})} , \vartheta \rangle - \ln \bigg(e^{ \beta \langle \bar{\tau}_n^{(0)}, \vartheta \rangle} + e^{\beta \langle \bar{\tau}_n^{(1)}, \vartheta \rangle} \bigg) \right],  \\
 \Lcal_{3}(\theta, \vartheta, \eta) &:= \frac{\lambda^2}{2} \norm{\theta - \vartheta}{2}{2} - SA \sum_{i=1}^{S} (\bm{\alpha}_{0,i} - 1) \ln \eta_{i} + \frac{1}{2} (\theta - \mu_{0})^{T} \Sigma_{0}^{-1} (\theta - \mu_{0}). \nonumber
\end{aligned}
$}
\label{eq:chapter4mapestimateproblem}
\end{equation}
\end{restatable}
\vspace{-0.3cm}

\textbf{Perturbation of MAP estimate.} As mentioned above, the idea now is to \emph{perturb} the loss function in Equation \eqref{eq:mapestimateproblem} with some noise, so that the MAP point estimates we get from this perturbed surrogate loss function serve as \emph{samples} from a distribution that approximates the true posterior \citep{osband2019deep, NIPS2017_49ad23d1, qin2022analysis, dwaracherla2022ensembles}. To that end, we use a perturbation of the `online' loss function $\Lcal_1(\cdot)$ and of the `offline' loss function $\Lcal_2(\cdot)$ by multiplicative random weights, and of the `prior' loss function $\Lcal_3(\cdot)$ by random samples from the prior distribution \footnote{$\mathrm{Bern}(\cdot)$ parameters that maximize performance are provided.}:

\noindent(i) \textit{Online perturbation.} Let $\zeta_{t} \sim \mathrm{Bern}(0.75)$, all i.i.d. Then, the perturbed $\Lcal_{1}(\cdot)$ becomes,
\vspace{-0.25cm}
\begin{equation}
\begin{aligned}
 \Lcal_{1}'(\theta, \vartheta, \eta) = - \sum_{t=1}^{k-1} \zeta_{t}  \left[ \beta \langle {\tau}_t^{(Y_{t})} , \vartheta \rangle - \ln \bigg(e^{ \beta \langle {\tau}_t^{(0)}, \vartheta \rangle} + e^{\beta \langle {\tau}_t^{(1)}, \vartheta \rangle} \bigg) + \sum_{j=0}^{1} \sum_{h=1}^{H-1} \ln P_{\eta} \left({s}_{t,h+1}^{(j)} \given {s}_{t,h}^{(j)}{a}_{t,h}^{(j)} \right) \right] \nonumber
\end{aligned}
\end{equation}
\vspace{-0.5cm}

\noindent(ii) \textit{Offline perturbation.} Let $\omega_{n} \sim \mathrm{Bern}(0.6)$, all i.i.d. Then, the perturbed $\Lcal_{2}(\cdot)$ becomes,
\vspace{-0.3cm}
    \begin{equation}
    \begin{aligned}
     \Lcal_{2}'(\theta, \vartheta, \eta) = - \sum_{n=1}^{N} \omega_{n} \left[ \beta \langle \bar{\tau}_n^{(\bar{Y}_{n})} , \vartheta \rangle - \ln \bigg(e^{ \beta \langle \bar{\tau}_n^{(0)}, \vartheta \rangle} + e^{\beta \langle \bar{\tau}_n^{(1)}, \vartheta \rangle} \bigg) \right] \nonumber
    \end{aligned}
    \end{equation}
    
\noindent(iii) \textit{Prior perturbation.} Let $\theta' \sim \Ncal(\mu_{0}, \Sigma_{0})$, $\vartheta' \sim \Ncal (\mu_{0}, \Ibf_{d} /\lambda^{2})$, all i.i.d. The perturbed $\Lcal_{3}(\cdot)$ is,
\vspace{-0.6cm}
    \begin{equation}
    \begin{aligned}
       \Lcal_{3}'(\theta, \vartheta, \eta)  &= \frac{\lambda^2}{2} \norm{\theta - \vartheta + \vartheta'}{2}{2} - SA \sum_{i=1}^{S} (\bm{\alpha}_{0,i} - 1) \ln \eta_{i} + \frac{1}{2} (\theta - \mu_{0} - \theta')^{T} \Sigma_{0}^{-1} (\theta - \mu_{0}- \theta') \nonumber
    \end{aligned}
    \end{equation}
\vspace{-0.55cm}
    
Then, for the $k^{th}$ episode, we get the following MAP point estimate from the perturbed loss function, 

\vspace{-0.4cm}
\begin{equation}
\label{eq:surrogate_perturbed_loss}
\begin{aligned}
    (\hat{\theta}_{k}, \hat{\vartheta}_{k}, \hat{\eta}_{k}) = \underset{\theta, \vartheta, \eta}{\argmin} \; \Lcal_{1}'(\theta, \vartheta, \eta) + \Lcal_{2}'(\theta, \vartheta, \eta) + \Lcal_{3}'(\theta, \vartheta, \eta), 
\end{aligned}
\end{equation}
\vspace{-0.8cm}

which are understood to have a distribution that approximates the actual posterior distribution. Note that this loss function can be extended easily to the setting where the offline dataset comes from \emph{multiple} raters with different $(\lambda_{i}, \beta_{i})$ competencies. Specifically, for $M$ raters, there will be $M$ similar terms for $\Lcal_{1}'(\cdot)$, $\Lcal_{2}'(\cdot)$, and $\Lcal_{3}'(\cdot)$. The final algorithm is given as Algorithm \ref{alg:main_algo_empirical}.

\begin{figure}[t]
\vspace{-1em}
\begin{algorithm}[H]
   \caption{Bootstrapped Posterior Sampling for Preference Learning}
\begin{algorithmic}[1]
   \STATE {\bfseries Input:} Initial dataset $\Dcal_{0}$, priors $\nu_{0}(\theta)$ and $\chi_{0}(\eta)$.
    \FOR{$k = 1,2, \dots ,K+1$}
        \FOR{$i = 0,1$}
    	 \STATE Sample a set of perturbations $\Pcal_{k}^{(i)} = \{(\zeta_{t}^{(i)}, \omega_{n}^{(i)}, \theta'^{(i)}, \vartheta'^{(i)})\}$.
    	 \STATE Solve Equation \eqref{eq:surrogate_perturbed_loss} using $\Pcal_{k}^{(i)}$ to find $(\hat{\theta}_{k}^{(i)}, \hat{\vartheta}_{k}^{(i)}, \hat{\eta}_{k}^{(i)})$.
         \ENDFOR
         \STATE Compute optimal policies $\pi_{k}^{(0)}$ using $(\hat{\eta}_{k}^{(0)}, \hat{\theta}_{k}^{(0)})$ and $\pi_{k}^{(1)}$ using $(\hat{\eta}_{k}^{(1)}, \hat{\theta}_{k}^{(1)})$.
    \STATE Run two trajectories $\tau_{k}^{(0)} \sim \pi_{k}^{(0)}$ and $\tau_{k}^{(1)} \sim \pi_{k}^{(1)}$ for $H$ horizon.
    \STATE Get feedback $Y_{k}$ on $\tau_{k}^{(0)}$ and $\tau_{k}^{(1)}$, and append to dataset as $\Dcal_{k} = \Dcal_{k-1} \oplus \left(\tau_{k}^{(0)}, \tau_{k}^{(1)}, Y_{k} \right)$.
    \ENDFOR
    \STATE {\bfseries Output:} Optimal policy $\pi_{K+1}^{\star} \equiv \pi_{K+1}^{(0)} $.
\end{algorithmic}
\label{alg:main_algo_empirical}
\end{algorithm}
\vspace{-2em}
\end{figure}

\section{Empirical results}
\label{sec:chapter4experiments}

We now present results on the empirical performance of Bootstrapped $\PSPL$ algorithm. We first demonstrate the effectiveness of PSPL on synthetic simulation benchmarks and then on realworld datasets. We study: 
(i) How much is the reduction in simple Bayesian regret after warm starting with an offline dataset? (ii) How much does the competence ($\lambda$ and $\beta$) of the rater who generated the offline preferences affect simple regret? (iii) Is $\PSPL$ robust to mis-specification of $\lambda$ and $\beta$? 

\begin{figure}[ht]
\centering
\begin{tcolorbox}[width=.45\textwidth, nobeforeafter, coltitle = black, fonttitle=\fontfamily{lmss}\selectfont, title= MountainCar, halign title=flush center, colback=backg_blue!5, colframe=lightorange!40, boxrule=2pt, grow to left by=-0.5mm, left=-15pt, right=-15pt]
    \centering
    {
        \includegraphics[height=0.32\textwidth, width=0.38\textwidth]{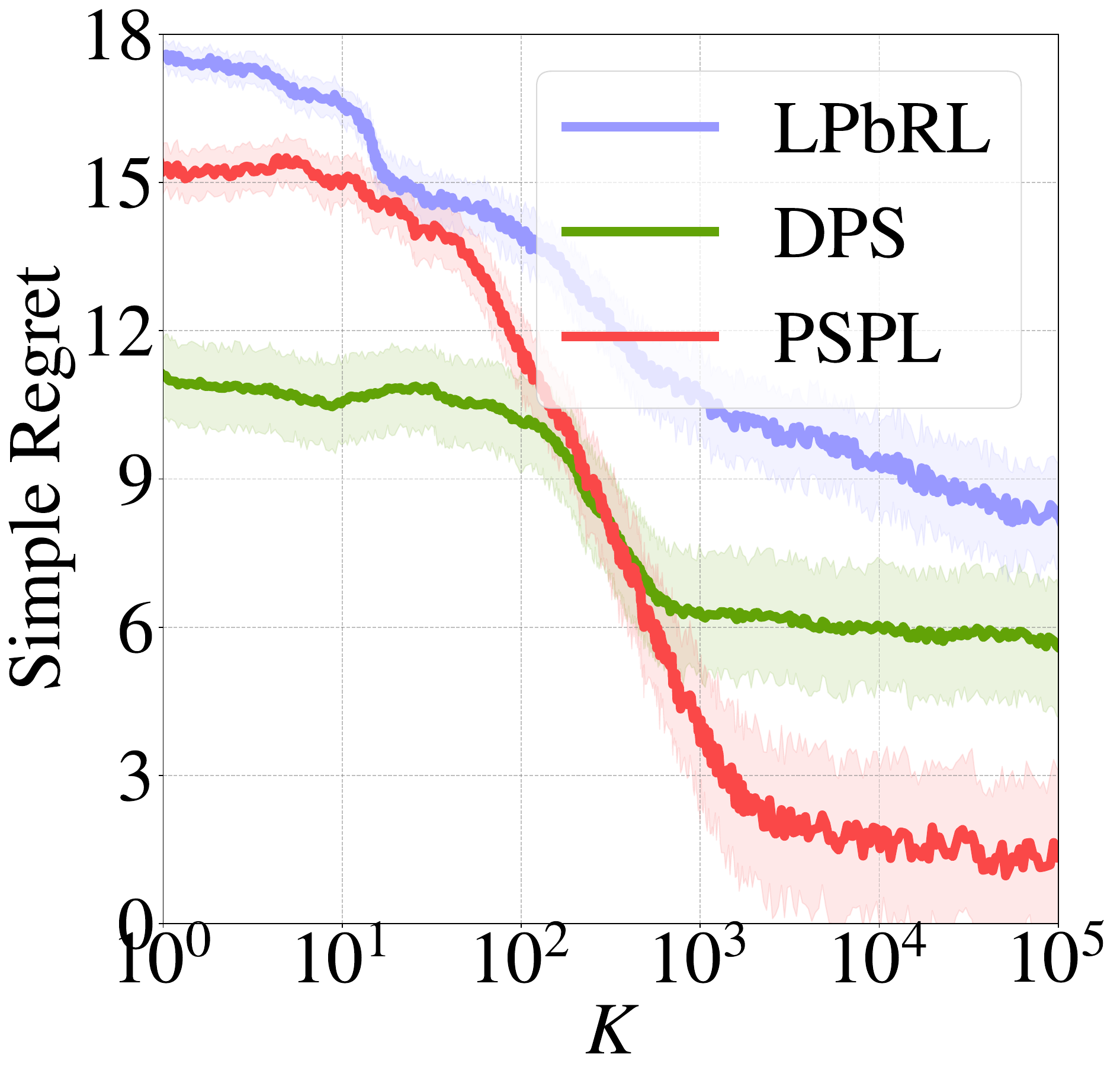}
    } 
    {
        \includegraphics[height=0.32\textwidth, width=0.38\textwidth]{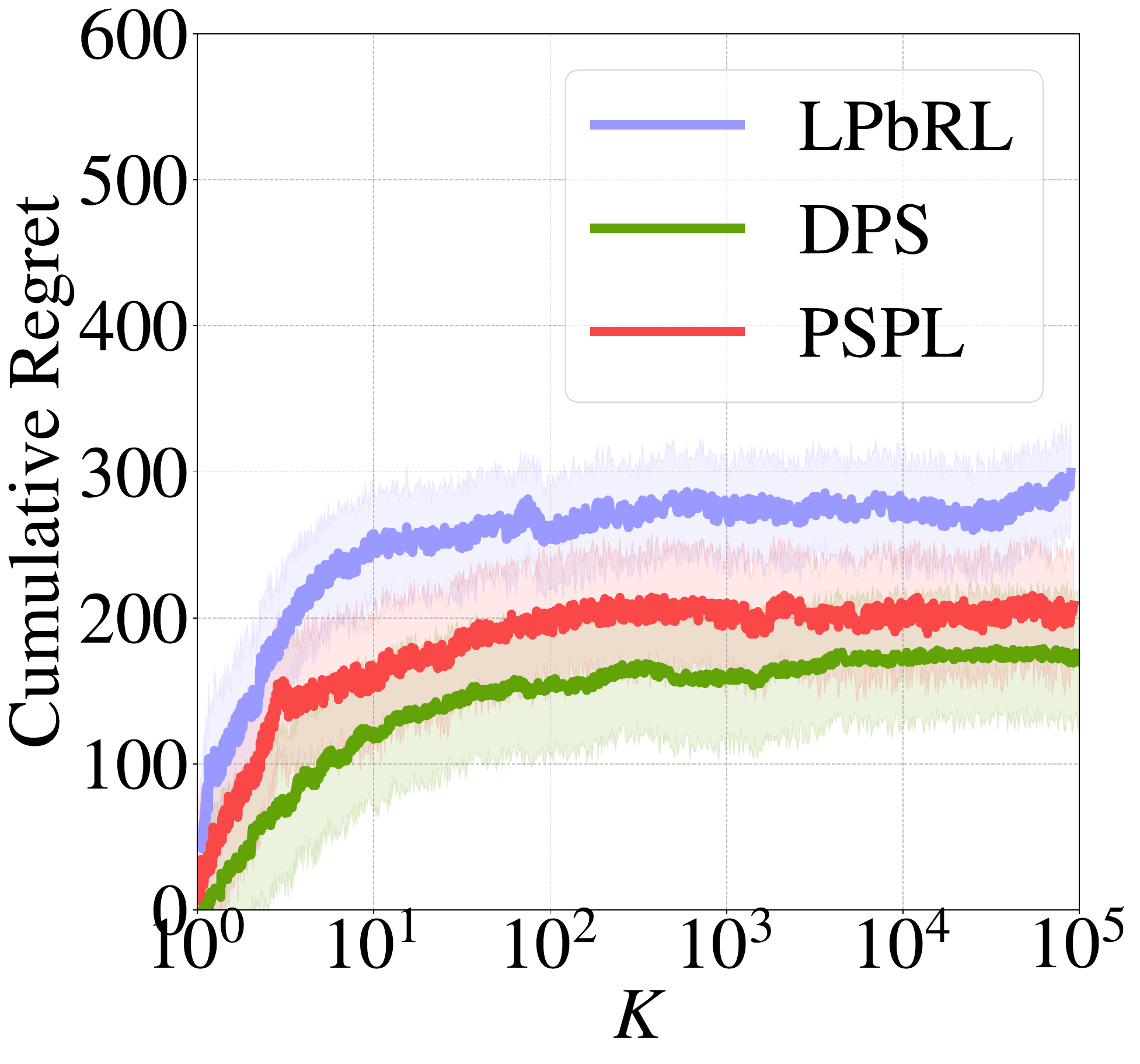}
    } 
\end{tcolorbox} 
\begin{tcolorbox}[width=.45\textwidth, nobeforeafter, coltitle = black, fonttitle=\fontfamily{lmss}\selectfont, title= RiverSwim, halign title=flush center, colback=backg_blue!5, colframe=brightblue!25, boxrule=2pt, grow to left by=-0.5mm, left=-15pt, right=-15pt]
    \centering
    {
        \includegraphics[height=0.32\textwidth, width=0.38\textwidth]{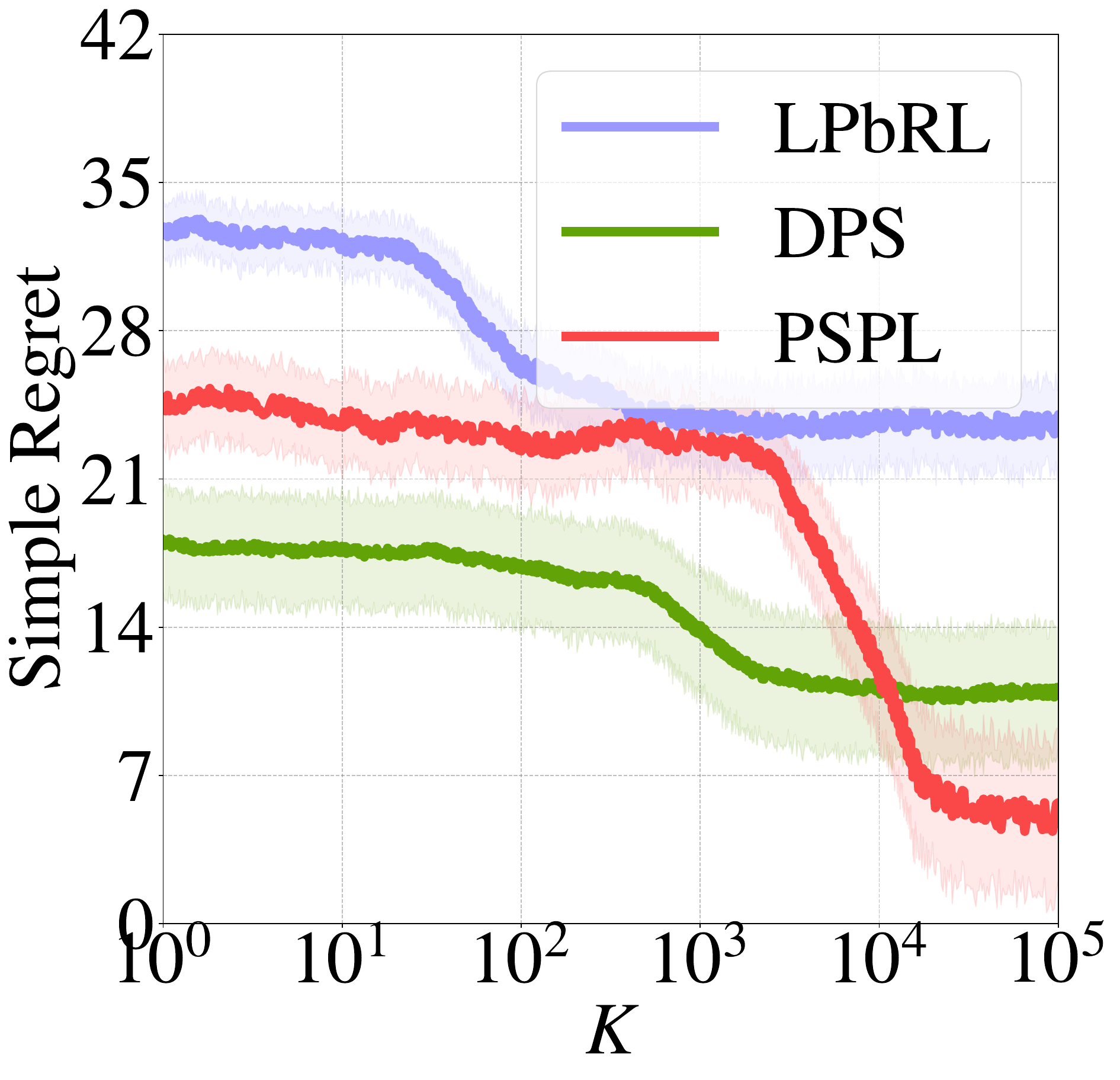}
    } 
    {
        \includegraphics[height=0.32\textwidth, width=0.38\textwidth]{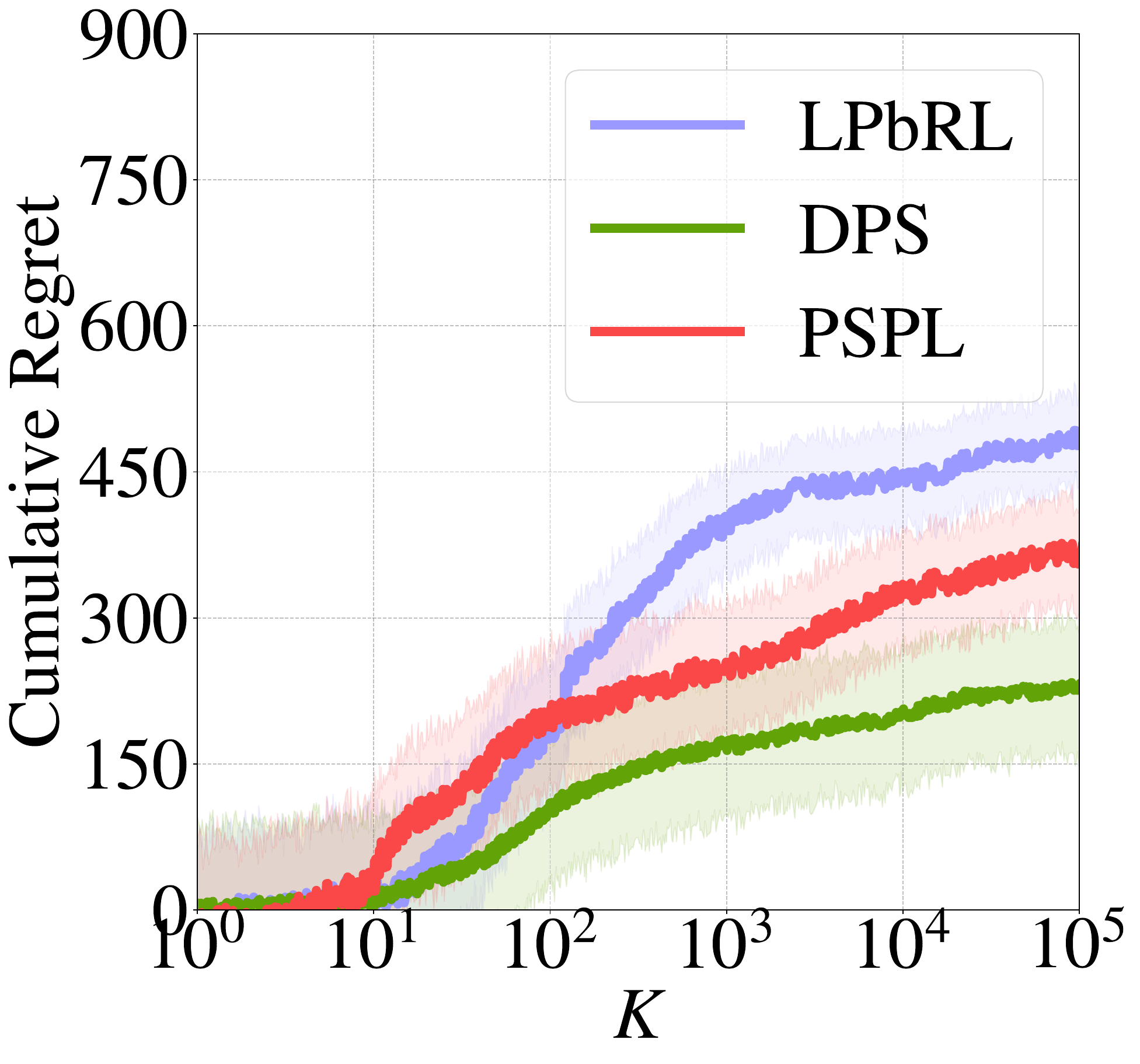}
    } 
\end{tcolorbox} 
\vspace{-0.2cm}
\caption{\small Simple and Cumulative Regret ($\div 10^{3}$) vs $K$ plots. $\PSPL$ is run with $\lambda=50,\beta=10,N=10^{3}$.}
\label{fig:simple-cum-regrets}
\end{figure}

\noindent\textbf{Baselines.} To evaluate the Bootstrapped $\PSPL$ algorithm, we  consider the following baselines:
(i) Logistic Preference based Reinforcement Learning (LPbRL) \citep{saha2023dueling}, 
and (ii) Dueling Posterior Sampling (DPS) \citep{novoseller2020dueling}. LPbRL does not specify how to incorporate prior offline data in to the optimization problem, and hence, no data has been used to warm start LPbRL, but, we initialize the transition and reward models in DPS using $\Dcal_{0}$. We run and validate the performance of all algorithms in the {\fontfamily{lmss}\selectfont RiverSwim} \citep{STREHL20081309} and {\fontfamily{lmss}\selectfont MountainCar} \citep{moore1990efficient} environments. See Figure \ref{fig:simple-cum-regrets} for empirical Bayesian regret comparison.

\begin{figure*}[ht]
\noindent
\begin{minipage}{\textwidth}
\begin{tcolorbox}[width=.495\textwidth, nobeforeafter, coltitle = black, fonttitle=\fontfamily{lmss}\selectfont, title=MountainCar, halign title=flush center, colback=backg_blue!5, colframe=brown!25, boxrule=2pt, grow to left by=-0.5mm, left=-1pt, right=-1pt]
    \centering
    {   
        \includegraphics[height=0.33\textwidth, width=0.42\textwidth]{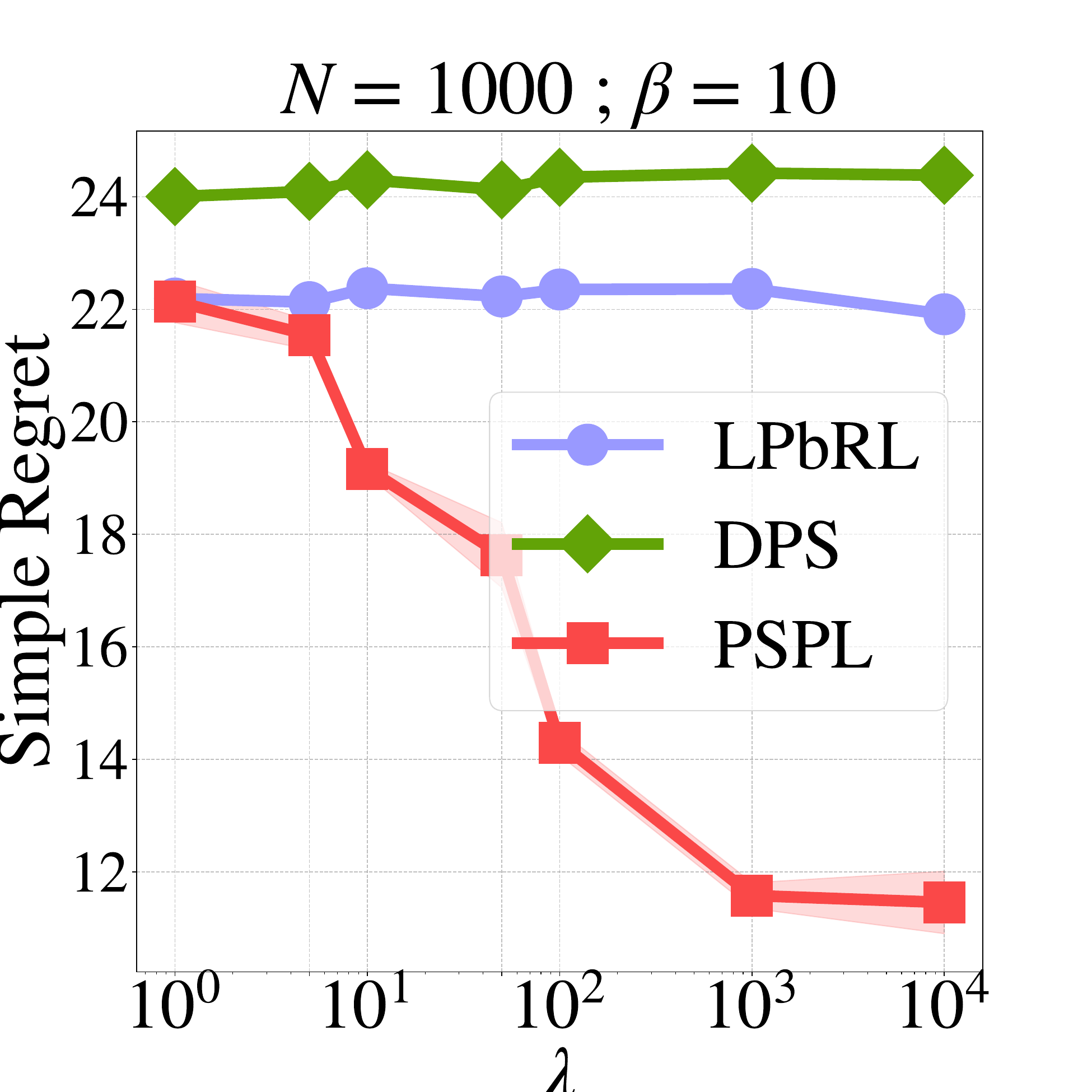}
    } 
    {
        \includegraphics[height=0.33\textwidth, width=0.42\textwidth]{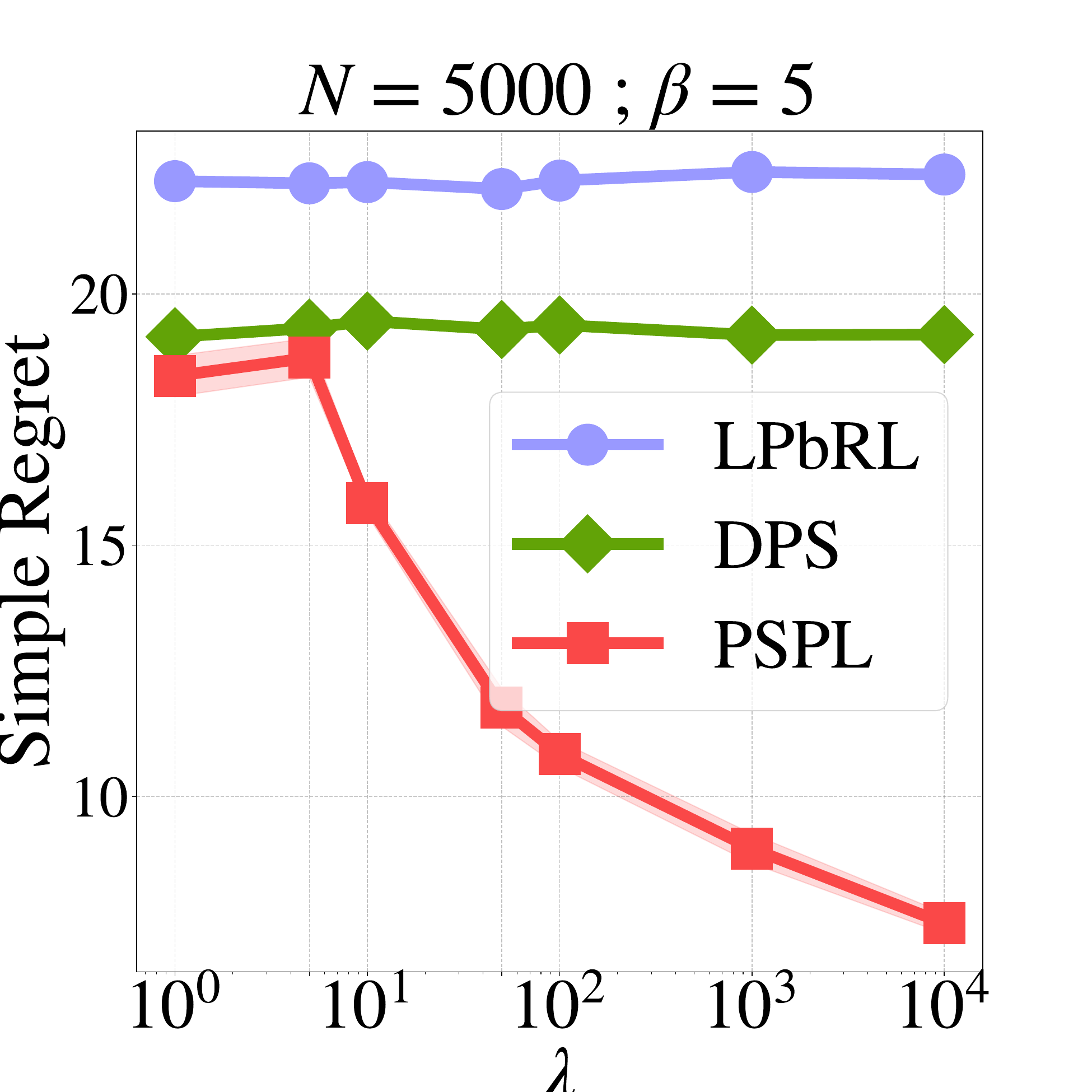}
    } 
    \newline
    \begin{center} \vspace{-0.60cm} 
        (a) Varying $\lambda$ for fixed $\beta$ and $N$. 
    \end{center}
    {
        \includegraphics[height=0.33\textwidth, width=0.42\textwidth]{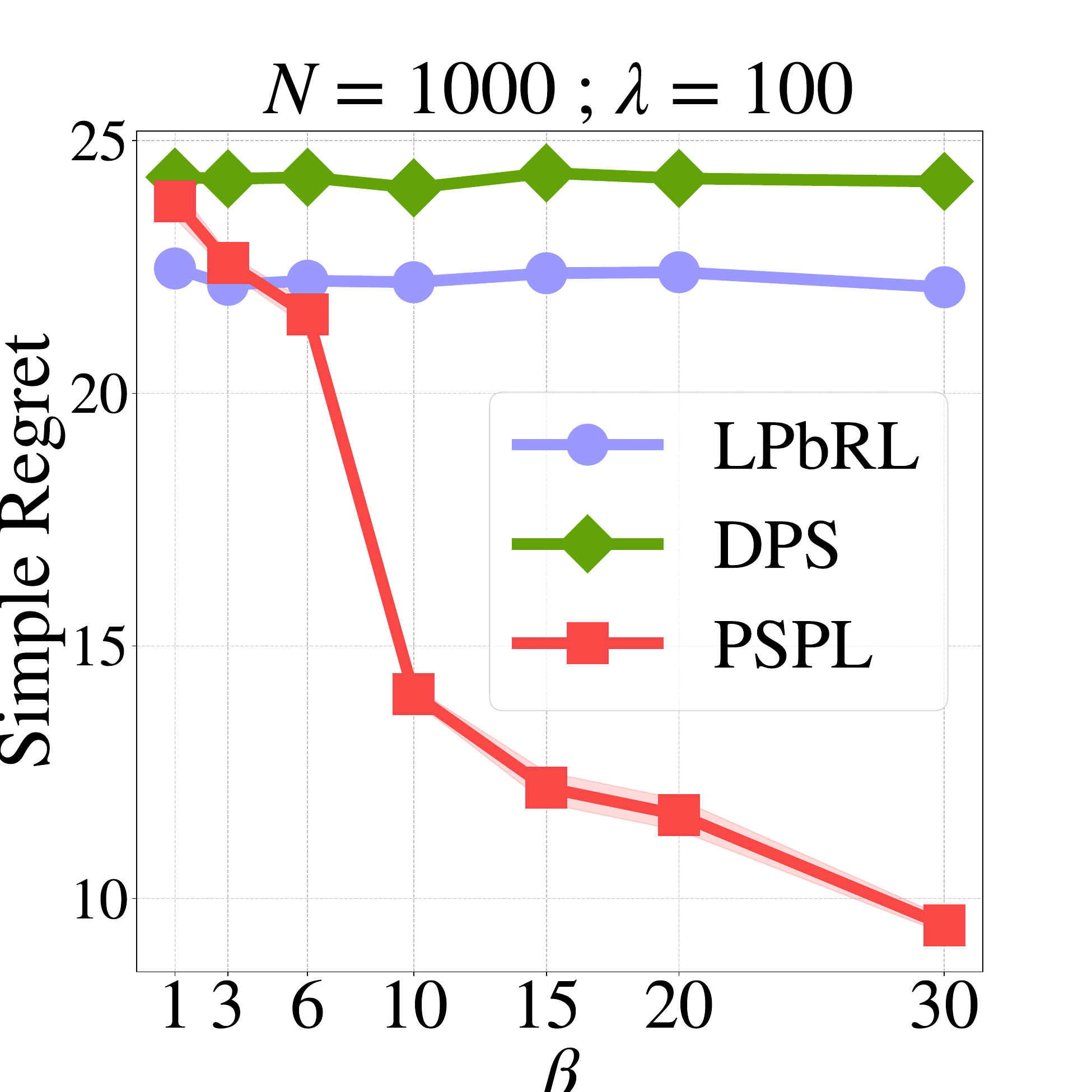}
    } 
    {
        \includegraphics[height=0.33\textwidth, width=0.42\textwidth]{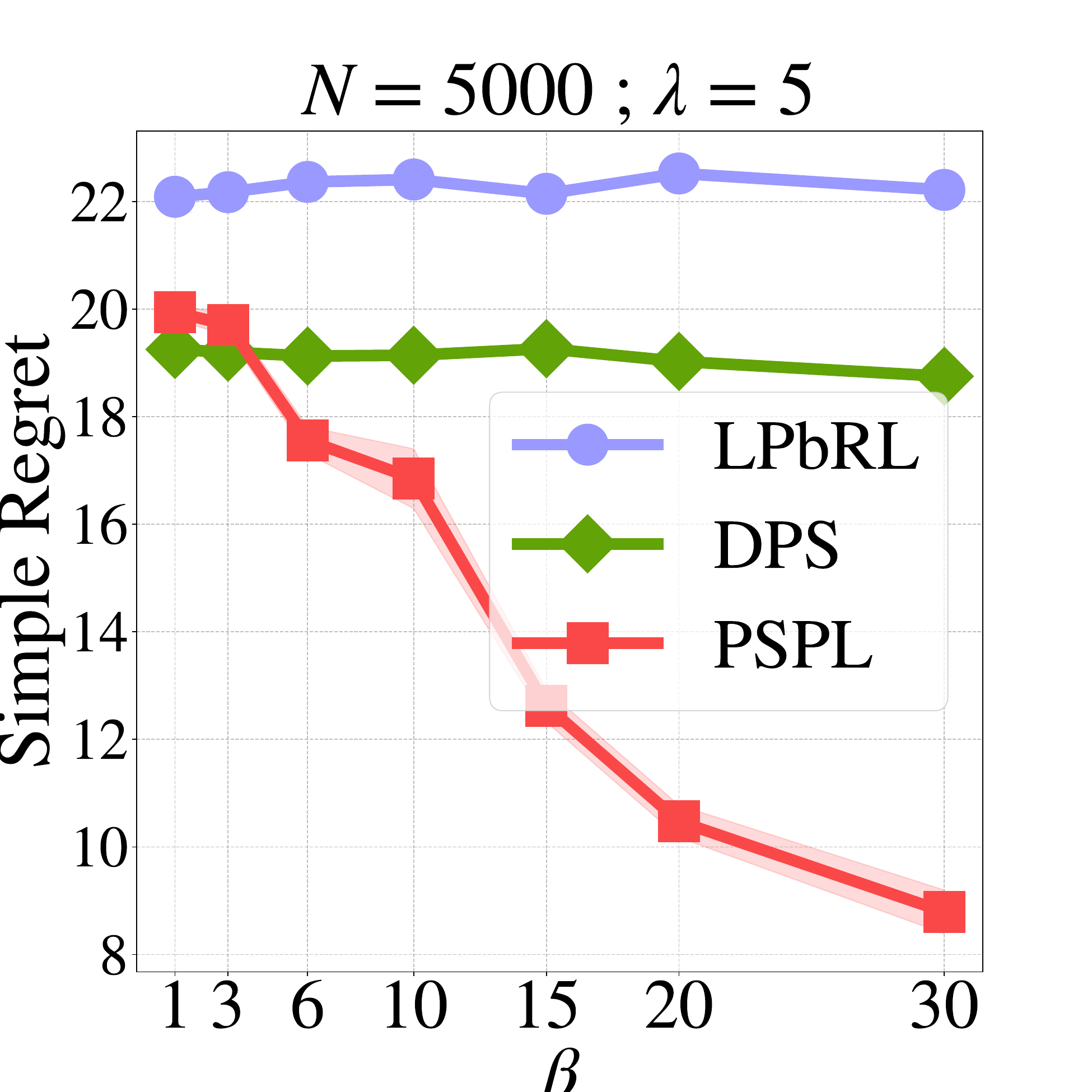}
    } 
    \newline
    \begin{center} \vspace{-0.60cm}
        (b) Varying $\beta$ for fixed $\lambda$ and $N$. 
    \end{center}
    {
        \includegraphics[height=0.33\textwidth, width=0.42\textwidth]{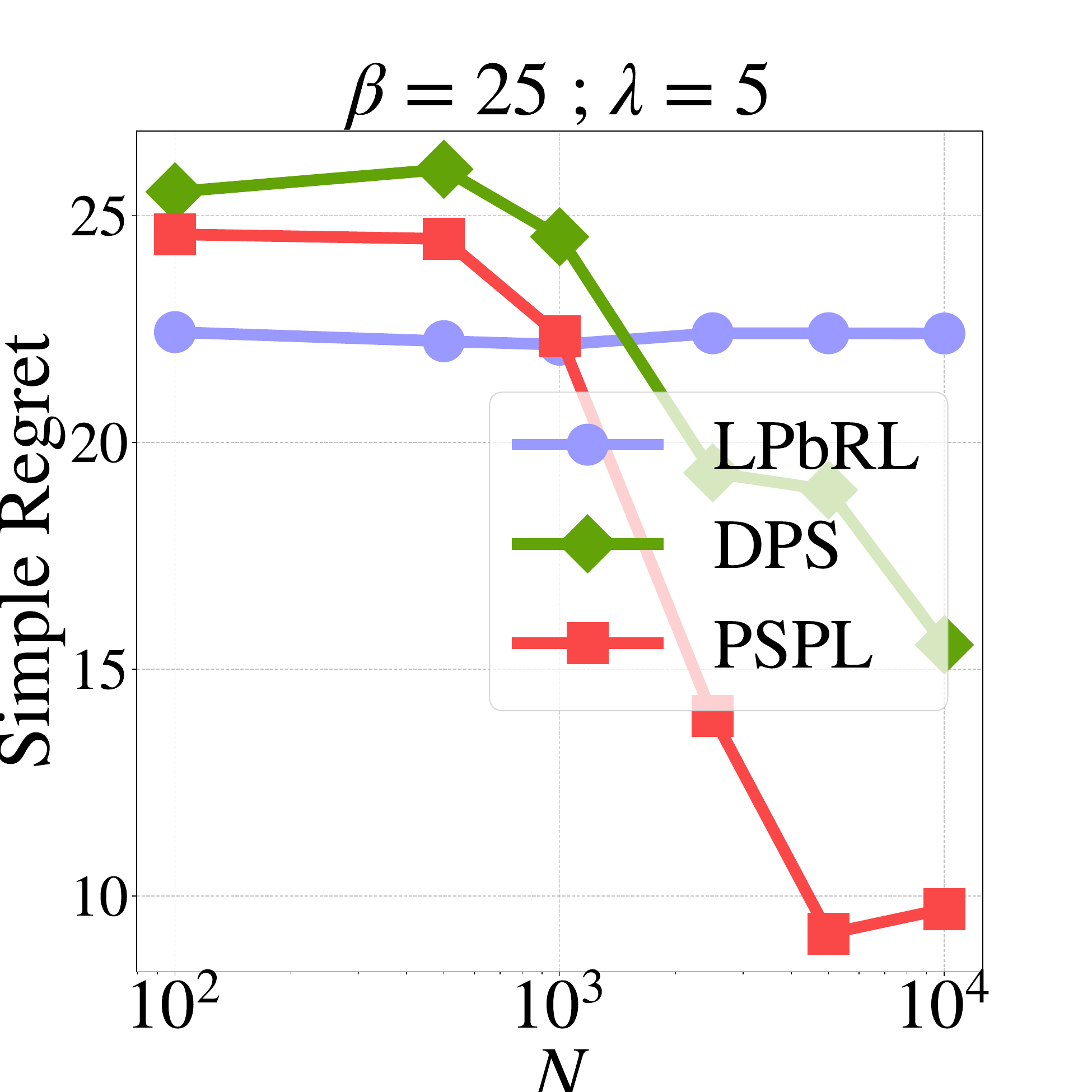}
    } 
    {
        \includegraphics[height=0.33\textwidth, width=0.42\textwidth]{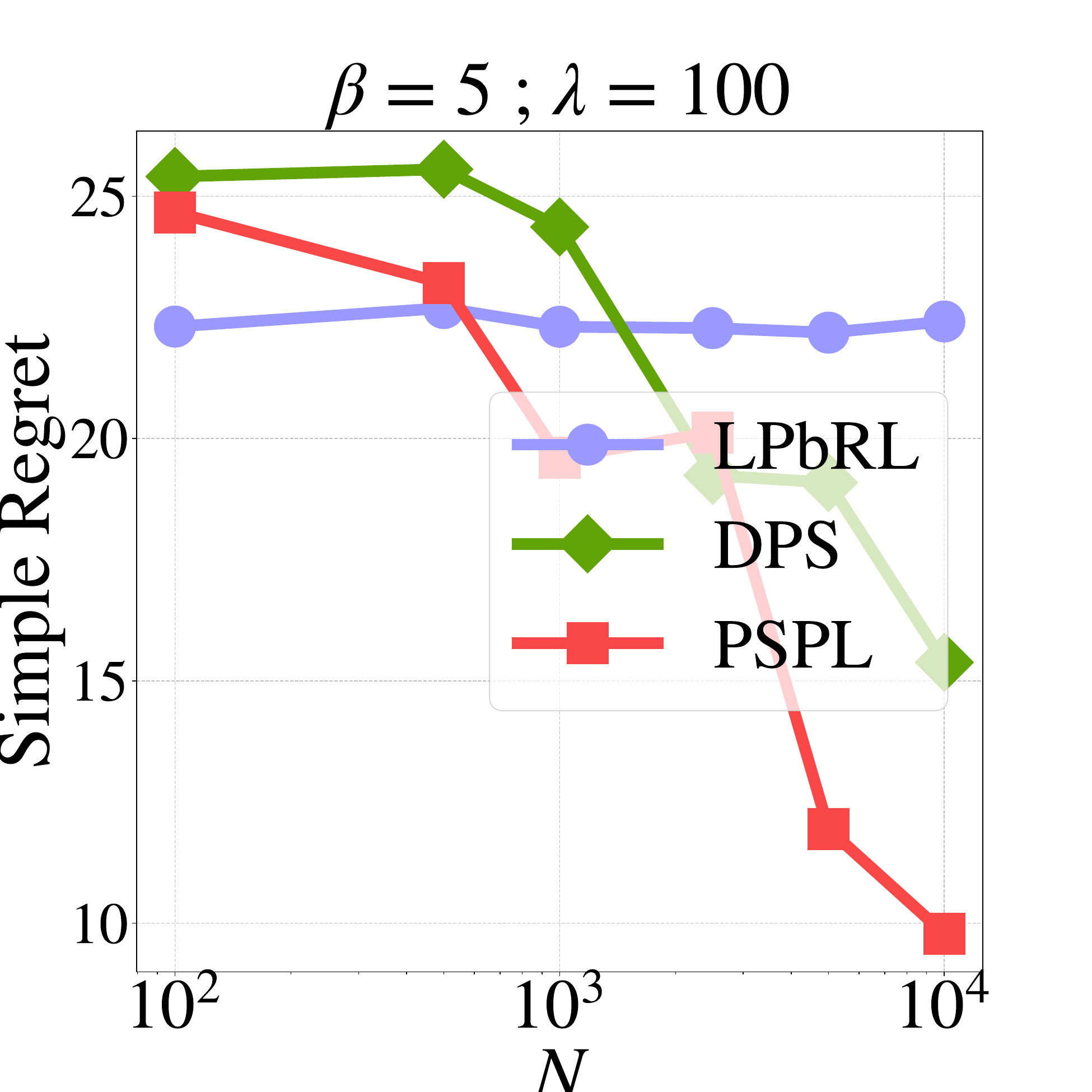}
    } 
    \newline
    \begin{center} \vspace{-0.60cm}
        (c) Varying $N$ for fixed $\beta$ and $\lambda$. 
    \end{center}
\end{tcolorbox} \hfill
\begin{tcolorbox}[width=.495\textwidth, nobeforeafter, fonttitle=\fontfamily{lmss}\selectfont, coltitle = black, title=RiverSwim, halign title=flush center,  colback=backg_blue!5, colframe=teal!15, boxrule=2pt, grow to left by=-0.5mm, grow to left by=-0.5mm, left=-1pt, right=-1pt]
\centering
    {   
        \includegraphics[height=0.33\textwidth, width=0.42\textwidth]{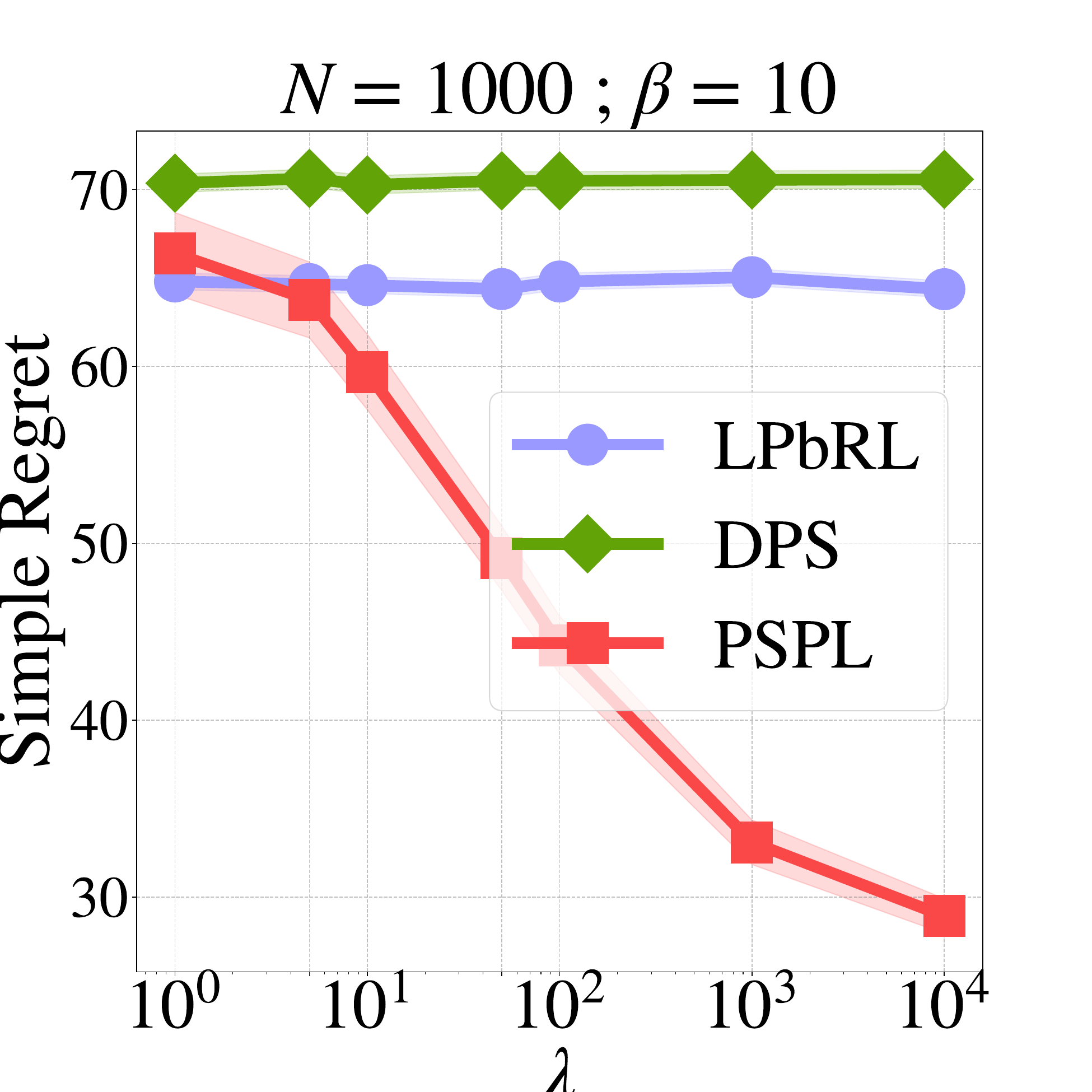}
    } 
    {
        \includegraphics[height=0.33\textwidth, width=0.42\textwidth]{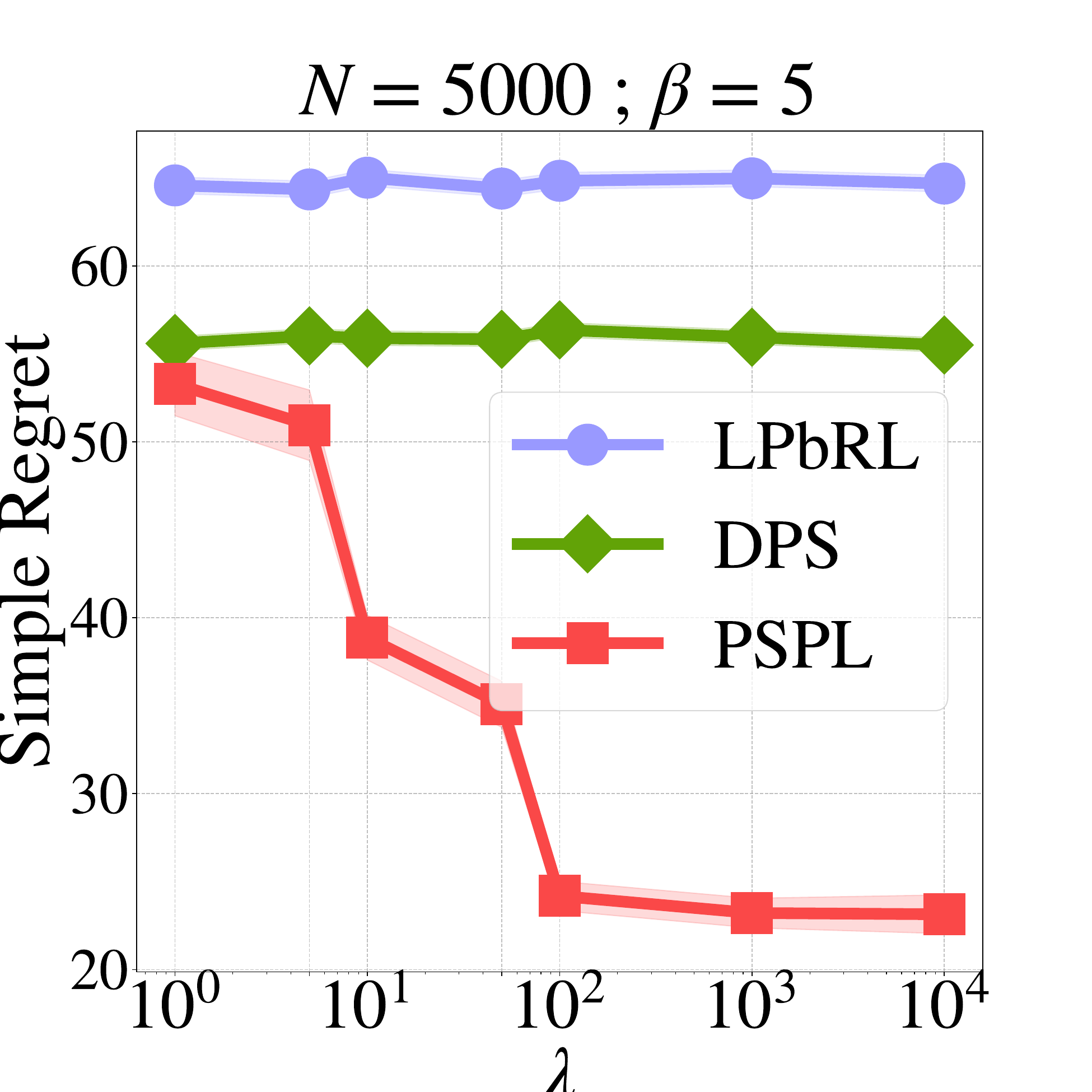}
    } 
    \newline
    \begin{center} \vspace{-0.60cm} 
        (a) Varying $\lambda$ for fixed $\beta$ and $N$. 
    \end{center}
    {
        \includegraphics[height=0.33\textwidth, width=0.42\textwidth]{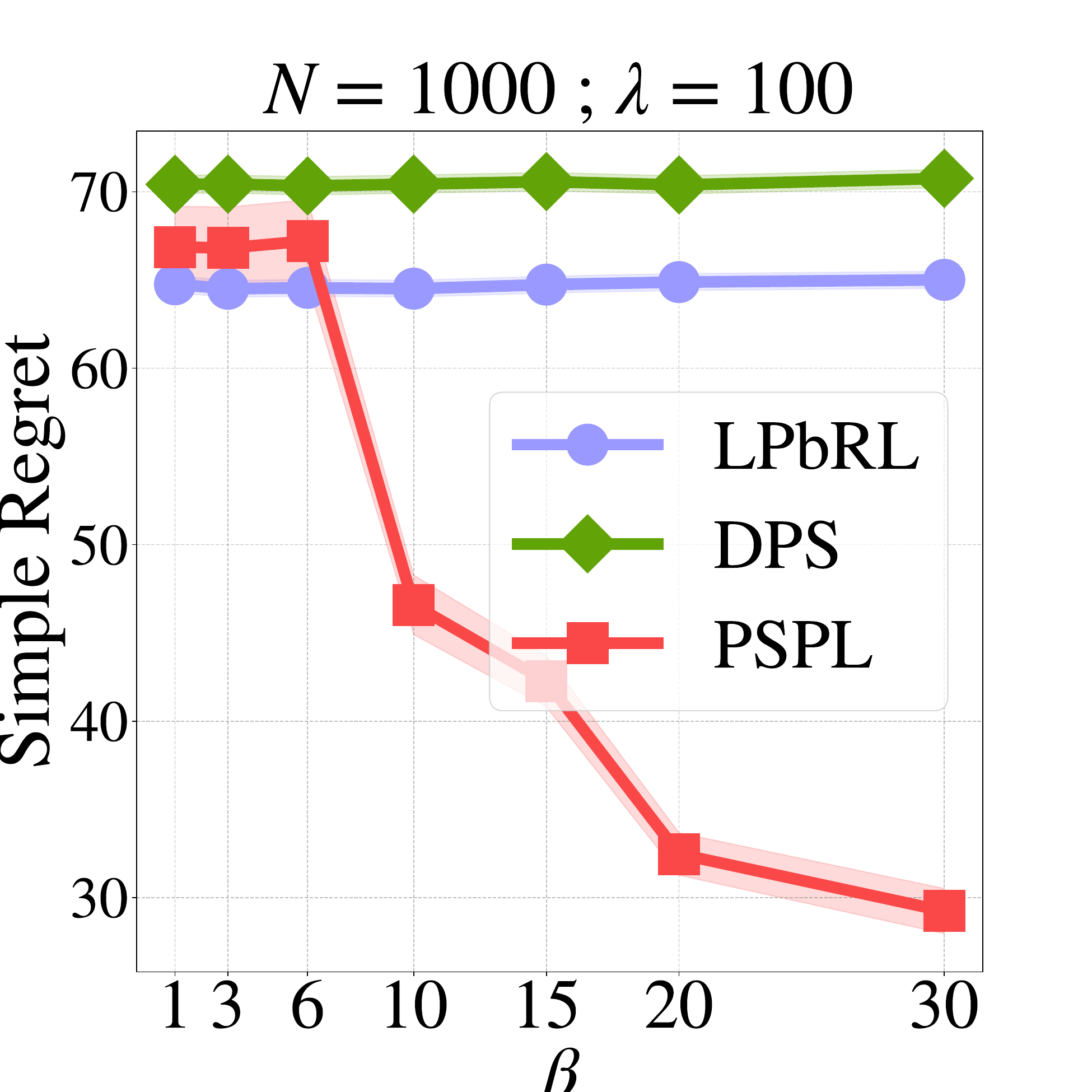}
    } 
    {
        \includegraphics[height=0.33\textwidth, width=0.42\textwidth]{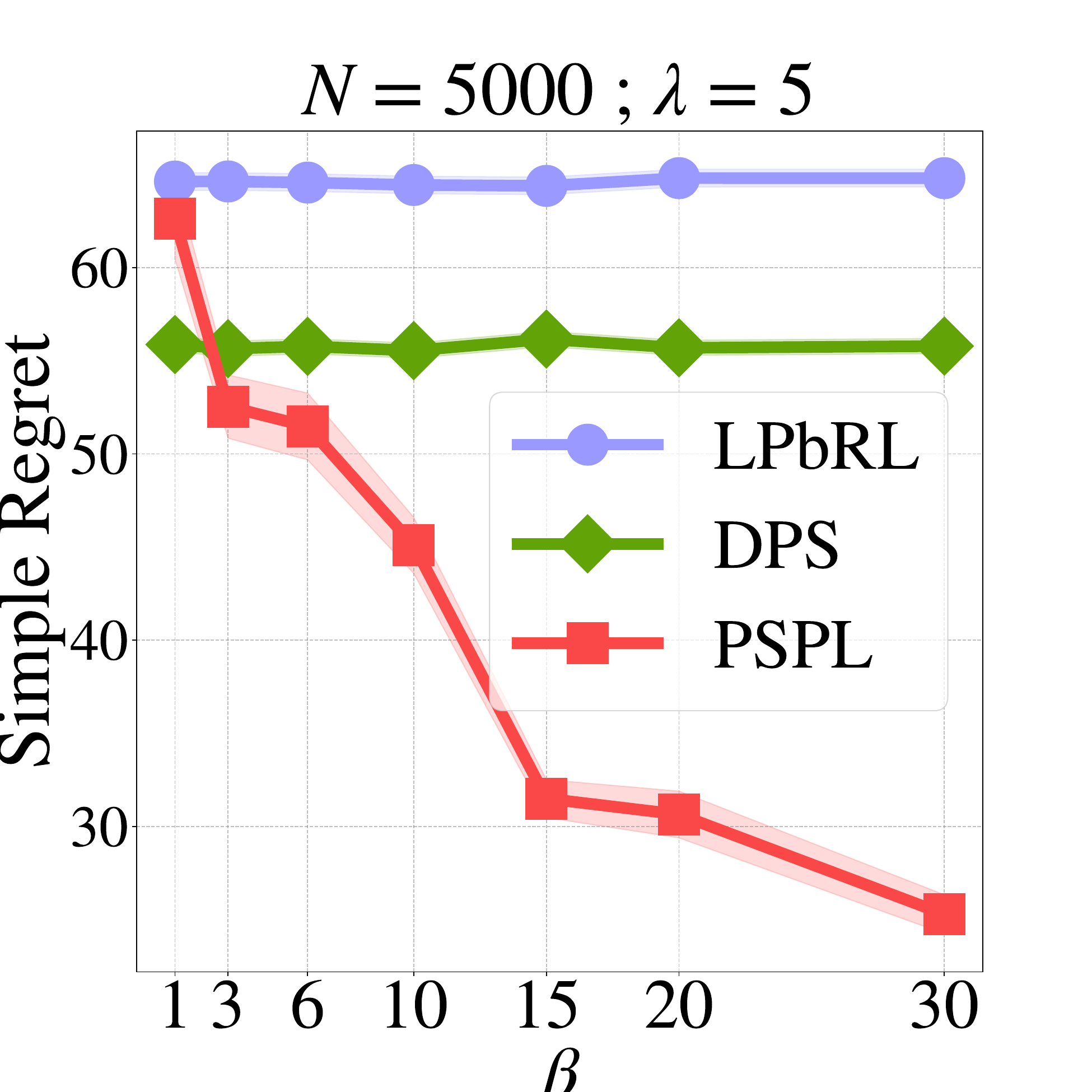}
    } 
    \newline
    \begin{center} \vspace{-0.60cm}
        (b) Varying $\beta$ for fixed $\lambda$ and $N$. 
    \end{center}
    {
        \includegraphics[height=0.33\textwidth, width=0.42\textwidth]{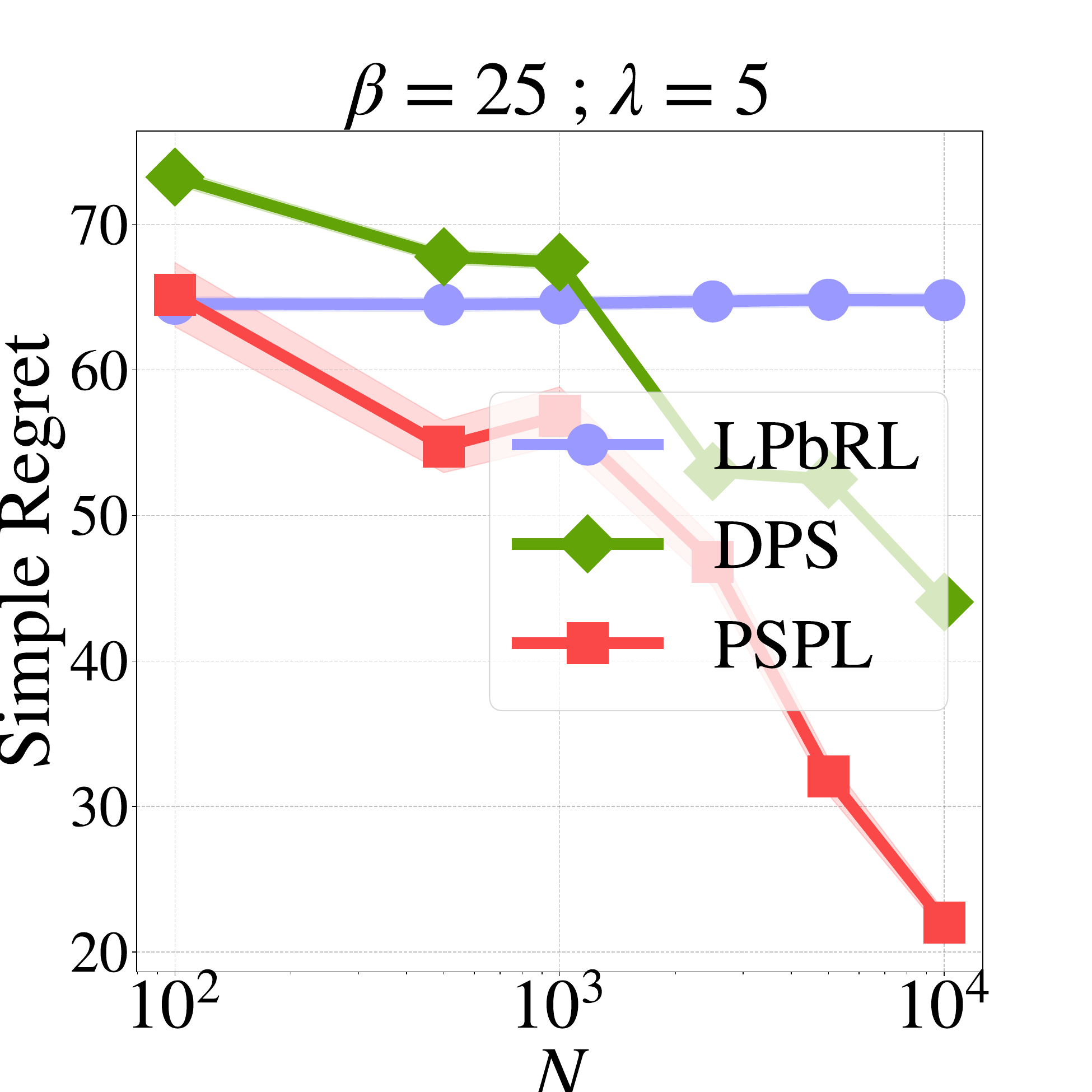}
    } 
    {
        \includegraphics[height=0.33\textwidth, width=0.42\textwidth]{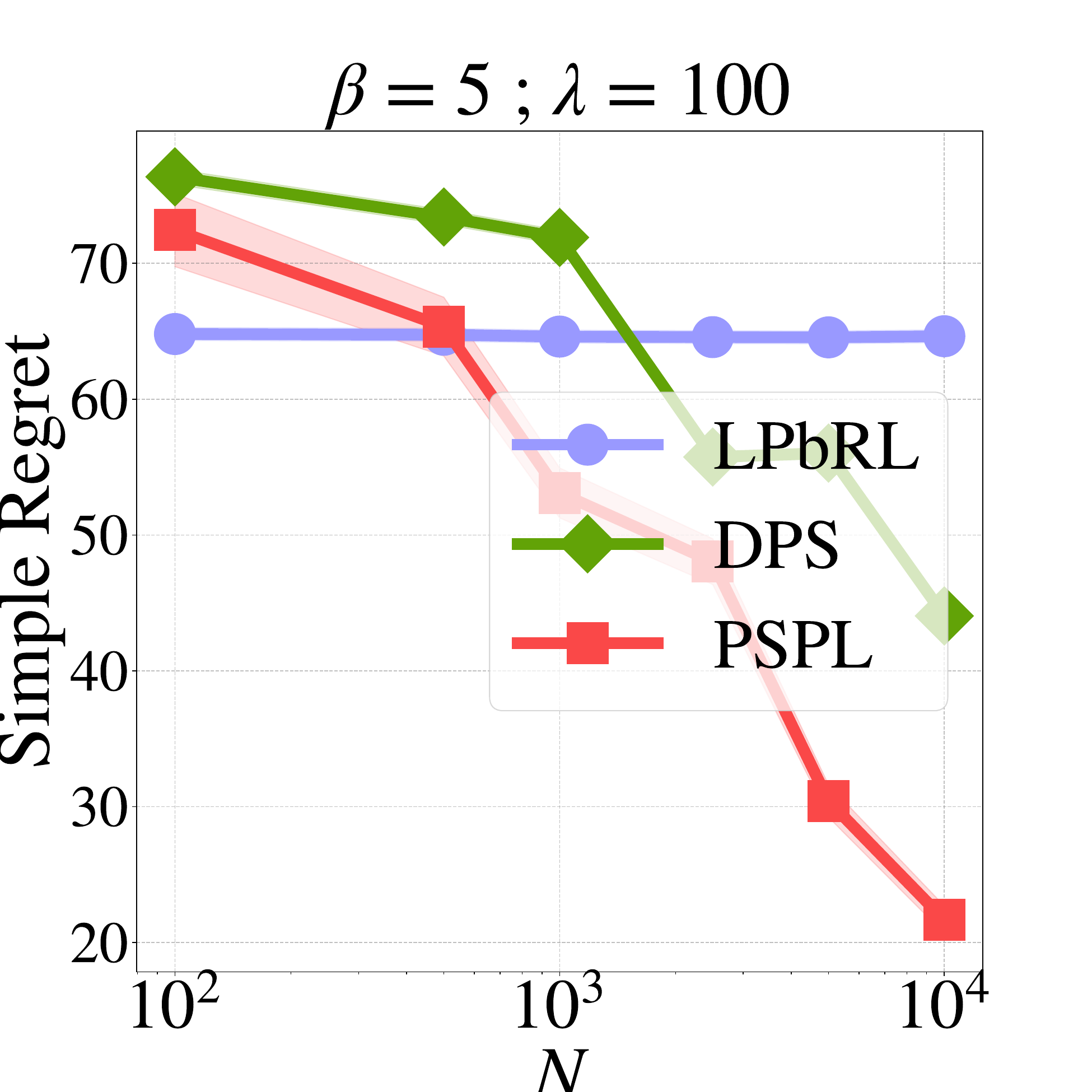}
    } 
    \newline
    \begin{center} \vspace{-0.60cm}
        (c) Varying $N$ for fixed $\beta$ and $\lambda$. 
    \end{center}
\end{tcolorbox}
\end{minipage}
\vspace{-0.3cm}
\caption{$\PSPL$ with varying $N$, $\beta$, and $\lambda$ in benchmark environments. Shaded region around mean line represents 1 standard deviation over 5 independent runs.}
\label{fig:main_ablation}
\end{figure*}

\textbf{Value of Offline Preferences.} We first understand the impact of $\Dcal_{0}$ on the performance of $\PSPL$ as the parameters ($\beta$, $\lambda$ and $N$) vary. Figure \ref{fig:main_ablation} shows that as $\lambda$ increases (the rater has a better estimate of the reward model $\theta$), the regret reduces substantially. Also notice that (for fixed $\lambda$ and $N$) as $\beta$ increases, the regret reduces substantially. Lastly, for fixed $\beta$ and $\lambda$, as dataset size $N$ increases, even with a `mediocre' expert ($\beta = 5$) the regret reduces substantially. \textcolor{black}{Overall, notice that $\PSPL$ correctly incorporates rater competence, which is intuitive to understand since in practice, rater feedback is imperfect, so weighting by competence is crucial.}

\textbf{Sensitivity to specification errors.}
The  Bootstrapped $\PSPL$ algorithm in Section \ref{sec:practical_approx_pspl} requires knowledge of rater's parameters $\lambda, \beta$. We study the sensitivity of $\PSPL$'s performance to mis-specification of these parameters. See Appendix \ref{sec:appendix_practical_pspl} for procedure to estimate $\lambda, \beta$ in practice, and Appendix \ref{sec:appendix_ablation} for ablation studies on robustness of PSPL to mis-specified parameters and mis-specified preference generation expert policies.

\begin{figure}[ht]
\centering
\begin{tcolorbox}[width=.4\textwidth, nobeforeafter, coltitle = black, fonttitle=\fontfamily{lmss}\selectfont, title= Image Generation Tasks, halign title=flush center, colback=backg_blue!5, colframe=violet!25, boxrule=2pt, grow to left by=-0.2mm, left=-15pt, right=-15pt]
    \centering
    {
        \includegraphics[height=0.85\textwidth, width=0.85\textwidth]{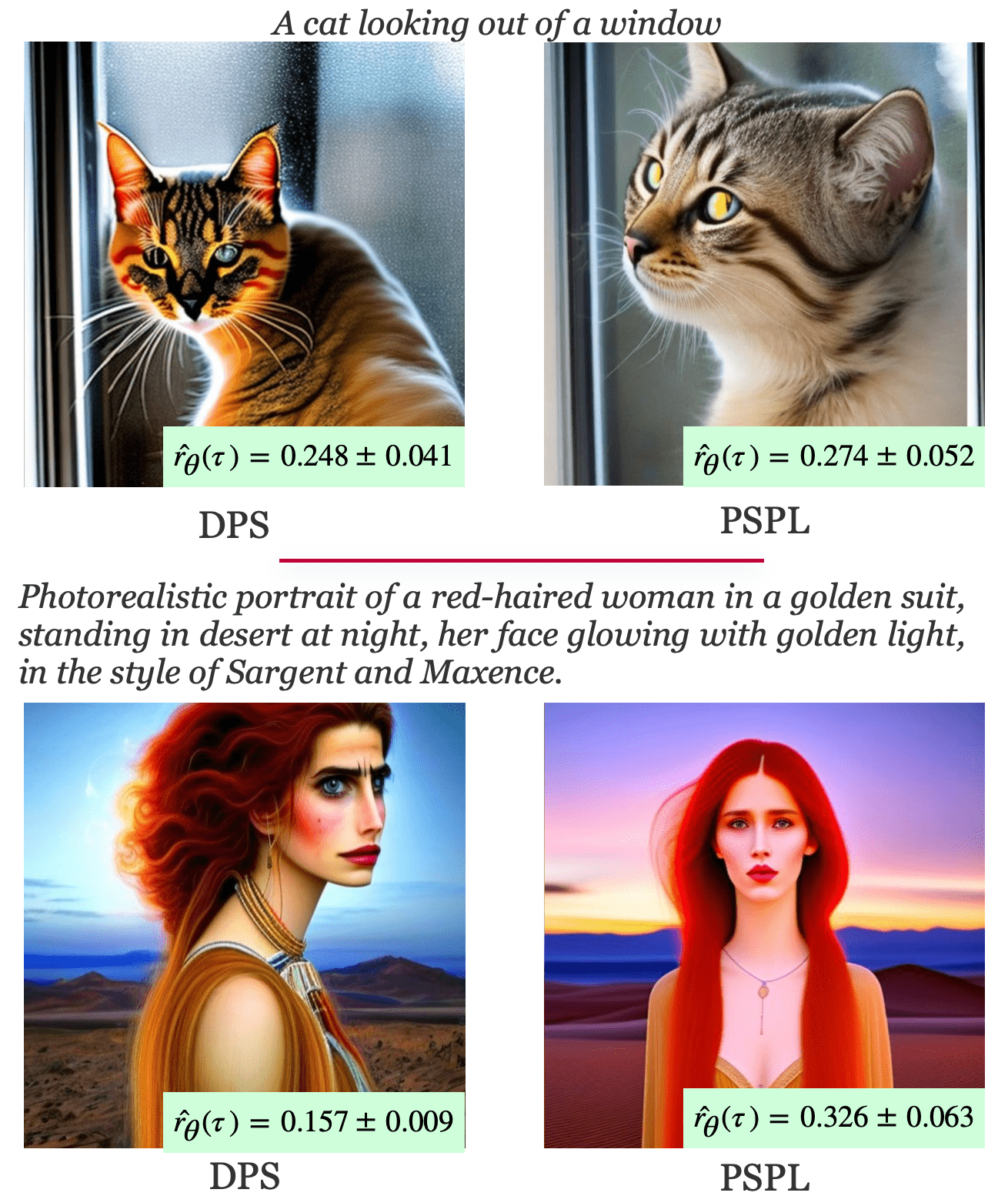}
    } 
\end{tcolorbox} 
\caption{\small Sample image generations along with final image reward $\hat{r}_{\theta}(\cdot)$ over 5 independent runs.}
\label{fig:pickapic_example}
\end{figure}

\textcolor{black}{
\textbf{Results on Image Generation Tasks.}
We instantiate our framework on the Pick-a-Pic dataset of human preferences for text–to–image generation \citep{kirstain2023pick}. The dataset contains over 500,000 examples and 35,000 distinct prompts. Each example contains a prompt, sequence generations of two images, and a corresponding preference label. We let each generation be a trajectory, so the dataset contains trajectory preferences $\mathcal D_{0} = { (\tau_{i}^{+},\tau_{i}^{-},y_{i}) }_{i=1}^{N}$ with $y_i = 1$ iff $\tau_{i}^{+} \succ \tau_{i}^{-}$. Each trajectory $\tau =(p,z_{0 : T})$ is the entire latent denoising chain $z_{0 : T}$ of length $T$ for prompt $p$ sampled from some prompt distribution. See Appendix \ref{appendix:pickapic} for the full experimental setup. As LPbRL involves optimization over confidence sets that grow exponentially in state-action space, we only present results of DPS and PSPL. See Figure \ref{fig:pickapic_example} for example generations given some prompts, where we observe higher final reward (hence, lower simple regret) of PSPL as compared to DPS.
}

\clearpage
\newpage

\section{Appendix}
\label{sec:appendix}



\paragraph{Network Architecture for DPO \cite{rafailov2024direct} and IPO \cite{ipo}}. We use a lightweight shared‐trunk MLP with two hidden layers of 64 ReLU units each (Xavier initialization \cite{pmlr-v9-glorot10a}), followed by a policy action head, outputting action logits for {\fontfamily{lmss}\selectfont MountainCar} \citep{moore1990efficient} and {\fontfamily{lmss}\selectfont RiverSwim} \citep{STREHL20081309} environments. We train with Adam (learning rate \(3\times10^{-4}\) \cite{kingma2014adam}), and apply dropout (0.1) on the trunk to regularize. This architecture balances expressivity and sample efficiency on these control tasks.

DPO \cite{rafailov2024direct} is an alternative approach to the RL paradigm, which avoids the training of a reward model altogether. The loss that DPO optimizes to obtain the optimal policy, given an empirical dataset $\Dcal = \{y_{w}, y_{l}\}$ of the winning (preferred) $y_{w}$ and losing (not preferred) $y_{l}$ trajectories, as a function of the reference policy $\pi_{\text{ref}}$ and regularization strength $\tau \in \Rbb$, is given by:

\begin{align*}
    \pi^{\star}_{\text{DPO}} = \argmin_{\pi} \quad \Ebb_{(y_{w}, y_{l}) \sim \Dcal} \left[ - \log \sigma \left( \tau \log \left( \frac{\pi(y_{w})}{\pi(y_{l})} \right)  - \tau \log \left( \frac{\pi_{\text{ref}}(y_{w})}{\pi_{\text{ref}}(y_{l})} \right) \right)  \right]
\end{align*}

, where $\sigma(\cdot)$ denotes the sigmoid function.

IPO is an instance of the $\Psi\text{PO}$ algorithm \cite{ipo} . The loss function that IPO optimizes is given by,

\begin{align*}
    \pi^{\star}_{\text{IPO}} = \argmin_{\pi} \quad \Ebb_{(y_{w}, y_{l}) \sim \Dcal} \left[ h_{\pi}(y_{w}, y_{l}) - \frac{1}{2 \tau}  \right]^{2} \; \text{where} \, , \; h_{\pi}(y, y') := \log \left( \frac{\pi(y) \pi_{\text{ref}}(y')}{\pi(y') \pi_{\text{ref}}(y)} \right) \, .
\end{align*}

\subsection{Prior-dependent analysis}
\label{appendix:prior_dependent_analysis}

\begin{lemma}[Monotone Contraction]
\label{lemma:contractionPS}
$\PSPL$ is monotone with respect to optimality of candidate policies $\pi_{k}^{(0)}$ and $\pi_{k}^{(1)}$ at any episode $k \in [K]$ i.e. $\Pr(\pi_{k}^{(i)} \neq \pi^{\star}) \leq \Pr(\pi_{1}^{(i)} \neq \pi^{\star})$ for $i \in \{0,1\}$ $\forAll k \geq 1$.
\end{lemma}
\begin{proof}
    Define $f(x) = x(1-x)$. $f$ is a concave function. We have for any $i \in \{0,1\}$, 
    \begin{align*}
        \Pr(\pi_{k}^{(i)} \neq \pi^{\star} ) &= \Ebb \left[\Pr \left(\pi_{k}^{(i)} \neq \pi^{\star} \Given \Dcal_{k} \right) \right]= \Ebb \left[ \sum_{\pi \in \Pi} \Pr(\pi_{k}^{(i)} = \pi, \pi^{\star} \neq \pi \Given \Dcal_{k})  \right]\\
        &= \Ebb \left[ \sum_{\pi\in\Pi} f \left(\Pr(\pi^{\star} = \pi \Given \Dcal_{k}) \right)  \right] \\
        &=  \Ebb \left[\sum_{\pi\in\Pi} \E{\Dcal_{1}} \left[f \left(\Pr \left(\pi^{\star} = \pi \given \Dcal_{k} \right) \right) \right] \right] \leq \Ebb \left[ \sum_{\pi\in\Pi} f \left(\E{\Dcal_{1}} \left[\Pr \left(\pi^{\star} = \pi \given \Dcal_{k} \right) \right] \right)\right] \\
        &= \Ebb\left[\sum_{\pi\in\Pi} f \left( \Pr \left(\pi^{\star} = \pi \given \Dcal_{1} \right) \right) \right]= \Ebb\left[ \sum_{\pi\in\Pi} \Pr(\pi_{1}^{(i)} = \pi, \pi^{\star} \neq \pi \given \Dcal_{1})  \right]\\
        &= \Pr(\pi_{1}^{(i)} \neq \pi^{\star}) \\
        \implies &  \Pr(\pi_{k}^{(i)} \neq \pi^{\star} ) \leq \Pr(\pi_{1}^{(i)} \neq \pi^{\star})
    \end{align*}
\end{proof}

\begin{lemma}
\label{lemma:empiricalcountsbound}
For any policy $\pi$, let an event $ E_{1} := \left\{  k, s, a: \, n_{k}(s, a) < \frac{1}{2} \sum_{j<k} w_{j} (s, a) - H\ln \left( \frac{SAH}{\delta'} \right) \right\}$ for some $\delta' \in (0,1)$, where $w_{h,j} (s, a)$ is the probability of visiting the $(s, a)$ pair at time $h$ of episode $j$ under the chosen policy, and $w_{j}(s,a) = \sum_{h} w_{h,j} (s, a)$ is the sum of the probabilities of visiting the $(s, a)$ pair in episode $j$. Then, $\Pr(E_{1}) \leq \delta' SAH = \delta$, where we set $\delta' = \frac{\delta}{SAH}$.
\end{lemma}
\begin{proof}

Consider a fixed $s \in \Scal, a \in \Acal, h \in [H]$, and denote the state and action visited in the $k^{\text{th}}$ episode at step $h$ as $s_{k,h}$ and $a_{k,h}$ respectively. We define $\mathcal F_k$ to be the sigma-field induced by the first $k-1$ episodes. Let $X_k$ be the indicator whether $s, a$ was observed in episode $k$ at time $h$. The probability $\Pr(s=s_{k,h}, a=a_{k,h} \given s_{k,1}, \pi_{k}^{(i)})$ (for $i \in \{0,1\}$) of whether $X_k = 1$ is $\mathcal F_k$-measurable and hence we can apply Lemma F.4 from \cite{dann2017unifying} with $W = \ln \frac{SAH}{\delta'}$ and obtain that $\Pr(E_{1}) \leq SAH \delta'$ after summing over all statements for $h \in [H]$ and applying the union bound over $s, a, h$.
\end{proof}

\begin{lemma}
\label{lemma:transition_est_error}
For all estimates $\hat{\theta}_{k}^{(i)}, \hat{\eta}_{k}^{(i)}$  ($i \in \{0,1\}$) of $\; \theta, \eta \;$ in any fixed episode $k \in [K]$, and for all $(s, a) \in \Scal \times \Acal$, with probability at least $1 - 2 \delta$ we have, 
$$
\sum_{s' \in \Scal} \left| \Pbb_{\theta, \eta}(s' \given s,a) - \Pbb_{\hat{\theta}_{k}^{(i)}, \hat{\eta}_{k}^{(i)}} (s' \given s,a) \right| \leq \sqrt{\frac{4S \ln \left( \frac{2KSA}{\delta}  \right)}{\sum_{j<k} w_{j} (s, a) - 2H\ln \left( \frac{SAH}{\delta} \right)}}
$$
\end{lemma}
\begin{proof}

\cite{weissman2003inequalities} gives the following high probability bound on the one norm of the maximum likelihood estimate using posterior sampling. In particular, with probability at least $1-\delta$, it holds that:
$$
\sum_{s' \in \Scal} \left| \Pbb_{\theta, \eta}(s' \given s,a) - \Pbb_{\hat{\theta}_{k}^{(i)}, \hat{\eta}_{k}^{(i)}} (s' \given s,a) \right| \leq \sqrt{2S \ln \left( \frac{2KSA}{\delta}  \right) /  \, n_{k}(s,a)} \quad .
$$

Then, using Lemma \ref{lemma:empiricalcountsbound}, the proof is complete.
\end{proof}

\begin{lemma}
\label{lemma:all_states_actions_visited}
Under Algorithm \ref{alg:main_algo_theoretical} for large enough K, every reachable state-action pair is visited infinitely-often.
\end{lemma}

\begin{proof}
The proof proceeds by assuming that there exists a state-action pair that is visited only finitely-many times.  This assumption will lead to a contradiction \footnote{Note that in finite-horizon MDPs, the concept of visiting a state finitely-many times is not the same as that of a transient state in an infinite Markov chain, because: 1)  due to a finite horizon, the state is resampled from the initial state distribution $\rho$ every $H$ timesteps, and 2) the policy---which determines which state-action pairs can be reached in an episode---is also resampled every $H$ timesteps.}: once this state-action pair is no longer visited, the reward model posterior is no longer updated with respect to it. Then, Algorithm \ref{alg:main_algo_theoretical}~is guaranteed to eventually sample a high enough reward for this state-action that the resultant policy will prioritize visiting it.   

First we note that Algorithm \ref{alg:main_algo_theoretical} is guaranteed to reach at least one state-action pair infinitely often: given our problem's finite state and action spaces, at least one state-action pair must be visited infinitely-often during execution of Algorithm \ref{alg:main_algo_theoretical}. If all state-actions are \textit{not} visited infinitely-often, there must exist a state-action pair $(s, a)$ such that $s$ is visited infinitely-often, while $(s, a)$ is not. Otherwise, if all actions are selected infinitely-often in all infinitely-visited states, the finitely-visited states are unreachable (in which case these states are irrelevant to the learning process and simple regret minimization, and can be ignored). Without loss of generality, we label this state-action pair $(s, a)$ as $\tilde{s}_1$. To reach a contradiction, it suffices to show that $\tilde{s}_1$ is visited infinitely-often.

Let $\bm{r}_1$ be the reward vector with a reward of $1$ in state-action pair $\tilde{s}_1$ and rewards of zero elsewhere. Let $\pi_{pi}(\tilde{\eta}, \bm{r}_1)$ be the policy that maximizes the expected number of visits to $\tilde{s}_1$ under dynamics $\tilde{\eta}$ and reward vector $\bm{r}_1$:
\begin{equation*}
\resizebox{0.85\linewidth}{!}{$
\begin{aligned}
        \pi_{pi}(\tilde{\eta}, \bm{r}_1) = \text{argmax}_\pi V(\tilde{\eta}, \bm{r}_1, \pi), \quad \text{where}, \quad V(\tilde{\eta}, \bm{r}, \pi) = \E{s_{1} \sim \rho} \left[ \sum_{t = 1}^H \overline{r}(s_t, \pi(s_t, t)) \,\Big|\, s_{t+1} \sim  \Pbb_{\tilde{\eta}}, \bm{\overline{r}} = \bm{r} \right].
\end{aligned}
$}
\end{equation*}
where $V(\tilde{\eta}, \bm{r}_1, \pi)$ is the expected total reward of a length-$H$ trajectory under $\tilde{\eta}, \bm{r}_1$, and $\pi$, or equivalently (by definition of $\bm{r}_1$), the expected number of visits to state-action $\tilde{s}_1$.

We next show that there exists a $\kappa > 0$ such that $\Pr(\pi = \pi_{pi}(\tilde{\eta}, \bm{r}_1)) > \kappa$ for all possible values of $\tilde{\eta}$. That is, for any sampled parameters $\tilde{\eta}$, the probability of selecting policy $\pi_{pi}(\tilde{\eta}, \bm{r}_1)$ is uniformly lower-bounded, implying that Algorithm \ref{alg:main_algo_theoretical} ~must eventually select $\pi_{pi}(\tilde{\eta}, \bm{r}_1)$.

Let $\tilde{r}_j$ be the sampled reward associated with state-action pair $\tilde{s}_j$ in a particular Algorithm \ref{alg:main_algo_theoretical}~episode, for each state-action $j \in \{1, \ldots, d\}$, with $d=SA$. We show that conditioned on $\tilde{\eta}$, there exists $v > 0$ such that if $\tilde{r}_1$ exceeds $\text{max}\{v \tilde{r}_2, v \tilde{r}_3, \ldots, v \tilde{r}_d\}$, then policy iteration returns the policy $\pi_{pi}(\tilde{\eta}, \bm{r}_1)$, which is the policy maximizing the expected amount of time spent in $\tilde{s}_1$. This can be seen by setting $v := \frac{H}{\kappa_1}$, where $\kappa_1$ is the expected number of visits to $\tilde{s}_1$ under $\pi_{pi}(\tilde{\eta}, \bm{r}_1)$. Under this definition of $v$, the event $\left\{\tilde{r}_{1} \ge \text{max}\{v \tilde{r}_2, v \tilde{r}_3, \ldots, v \tilde{r}_d\}\right\}$ is equivalent to $\{\tilde{r}_{1}\kappa_1 \ge h \, \text{max}\{\tilde{r}_2, \tilde{r}_3, \ldots, \tilde{r}_d\}\}$; the latter inequality implies that given $\tilde{\eta}$ and $\bm{\tilde{r}}$, the expected reward accumulated solely in state-action $\tilde{s}_1$ exceeds the reward gained by repeatedly (during all $H$ time-steps) visiting the state-action pair in the set $\{\tilde{s}_2, \ldots, \tilde{s}_d\}$ having the highest sampled reward. Clearly, in this situation, policy iteration results in the policy $\pi_{pi}(\tilde{\eta}, \bm{r}_1)$.

Next we show that $v = \frac{h}{\kappa_1}$ is continuous in the sampled dynamics $\tilde{\eta}$ by showing that $\kappa_1$ is continuous in $\tilde{\eta}$. Recall that $\kappa_1$ is defined as expected number of visits to $\tilde{s}_1$ under $\pi_{pi}(\tilde{\eta}, \bm{r}_1)$. This is equivalent to the expected reward for following $\pi_{pi}(\tilde{\eta}, \bm{r}_1)$ under dynamics $\tilde{\eta}$ and rewards $\bm{r}_1$:
\begin{flalign}\label{eqn:defn_p_1}
\kappa_1 = V(\tilde{\eta}, \bm{r}_1, \pi_{pi}(\tilde{\eta}, \bm{r}_1)) = \max_\pi V(\tilde{\eta}, \bm{r}_1, \pi).
\end{flalign}
The value of any policy $\pi$ is continuous in the transition dynamics parameters, and so $V(\tilde{\eta}, \bm{r}_1, \pi)$ is continuous in $\tilde{\eta}$. The maximum in \eqref{eqn:defn_p_1} is taken over the finite set of deterministic policies; because a maximum over a finite number of continuous functions is also continuous, $\kappa_1$ is continuous in $\tilde{\eta}$.

Next, recall that a continuous function on a compact set achieves its maximum and minimum values on that set. The set of all possible dynamics parameters $\tilde{\eta}$ is such that for each state-action pair $j$, $\sum_{k = 1}^S p_{jk} = 1$ and $p_{jk} \ge 0 \, \forall \, k$; the set of all possible vectors $\tilde{\eta}$ is clearly closed and bounded, and hence compact. Therefore, $v$ achieves its maximum and minimum values on this set, and so for any $\tilde{\eta}$, $v \in [v_{\text{min}}, v_{\text{max}}]$, where $v_{\text{min}} > 0$ ($v$ is nonnegative by definition, and $v = 0$ is impossible, as it would imply that $\tilde{s}_1$ is unreachable).

Then, $\Pr(\pi = \pi_{pi}(\tilde{\eta}, \bm{r}_1))$ can then be expressed in terms of $v$ and the parameters of the reward posterior. Firstly,
\small
\begin{align*}
\Pr(\pi = \pi_{pi}(\tilde{\eta}, \bm{r}_1)) \ge \Pr(\tilde{r}_1 > \text{max}\{v \tilde{r}_2, \ldots, v \tilde{r}_d\}) \ge \prod_{j = 2}^d \Pr(\tilde{r}_1 > v \tilde{r}_j) = \prod_{j = 2}^d [1 - \Pr(\tilde{r}_1 - v \tilde{r}_j \le 0)]
\end{align*}
\normalsize

The posterior updates for the reward model are given by Equation \eqref{eq:theta_informed_prior}, which is intractable to compute in closed form. Since we have $\theta_{0} \sim \Ncal(\mu_{0}, \Sigma_{0})$, we can use the result of Lemma 3 in Appendix of \cite{novoseller2020dueling}. The remaining proof thereby follows, and as a result, there exists some $\kappa > 0$ such that $\Pr(\pi = \pi_{pi}(\tilde{\eta}, \bm{r}_1)) \ge \kappa > 0$.

In consequence, Algorithm \ref{alg:main_algo_theoretical}~is guaranteed to infinitely-often sample pairs $(\tilde{\eta}, \pi)$ such that $\pi = \pi_{pi}(\tilde{\eta}, \bm{r}_1)$. As a result, Algorithm \ref{alg:main_algo_theoretical}~infinitely-often samples policies that prioritize reaching $\tilde{s}_1$ as quickly as possible. Such a policy always takes action $a$ in state $s$. Furthermore, because $s$ is visited infinitely-often, either a) $p_0(s) > 0$ or b) the infinitely-visited state-action pairs include a path with a nonzero probability of reaching $s$. In case a), since the initial state distribution is fixed, the MDP will infinitely-often begin in state $s$ under the policy $\pi = \pi_{pi}(\tilde{\eta}, \bm{r}_1)$, and so $\tilde{s}_1$ will be visited infinitely-often. In case b), due to Lemma 1 in Appendix of \cite{novoseller2020dueling}, the transition dynamics parameters for state-actions along the path to $s$ converge to their true values (intuitively, the algorithm knows how to reach $s$). In episodes with the policy $\pi = \pi_{pi}(\tilde{\eta}, \bm{r}_1)$, Algorithm \ref{alg:main_algo_theoretical}~is thus guaranteed to reach $\tilde{s}_1$ infinitely-often. Since Algorithm \ref{alg:main_algo_theoretical}~selects $\pi_{pi}(\tilde{\eta}, \bm{r}_1)$ infinitely-often, it must reach $\tilde{s}_1$ infinitely-often. This presents a contradiction, and so every state-action pair must be visited infinitely-often as the number of episodes tend to infinity.  
\end{proof}

\begin{restatable}{lemma}{psplerrorregretappendix}
\label{lemma:pspl_error_regret_appendix}
For any confidence $\delta_{1} \in (0,\frac{1}{3})$, let $\delta_{2} \in (c,1)$ with $c \in (0,1)$, be the probability that any optimal policy estimate $\hat{\pi}^{\star}$ constructed from the offline preference dataset $\Dcal_{0}$ is $\varepsilon$-optimal with probability at least $(1-\delta_{2})$ i.e. $\Pr \left( \E{s \sim \rho} \left[ V_{\theta,\eta,0}^{\pi^{\star}}(s) -  V_{\theta,\eta,0}^{\hat{\pi}^{\star}}(s) \right] > \varepsilon \right) < \delta_{2}$. Then, the simple Bayesian regret of the learner $\Upsilon$ is upper bounded with probability of at least $1-3\delta_{1}$ by, 
    \begin{equation}
    \label{eq:pspl_prior_error_bound_appendix}
    \begin{aligned}
        \SR_{K}^{\Upsilon}(\pi^{\star}_{K+1}, \pi^{\star}) \leq \sqrt{ \frac{10 \delta_{2}  S^{2}AH^{3} \ln \left( \frac{2KSA}{\delta_{1}}  \right) + 3\delta_{2}SAH^{2}\varepsilon^{2}}{2K \left(1 + \ln \frac{SAH}{\delta_{1}} \right) - \ln \frac{SAH}{\delta_{1}}} }
    \end{aligned}
    \end{equation}
\end{restatable}

\begin{proof}
\label{proof:pspl_error_regret}

Let $\Theta := (\theta, \eta)$ denote the unknown true parameters of the MDP $\Mcal$, and let $\hat{\Theta}_{k}^{(i)} := \left(\hat{\theta}_{k}^{(i)}, \hat{\eta}_{k}^{(i)} \right)$ denote the sampled reward and transition parameters at episode $k$, which are used to compute policy $\pi_{k}^{(i)}$ for $i \in \{0,1\}$. Let $J_{\pi}^{\tilde{\Theta}} := \E{\tilde{\Theta}, \tau \sim \pi}[r(\tau)]$ denote the expected total reward for a trajectory sampled from policy $\pi$ under some environment $\tilde{\Theta} := \left(\tilde{\theta}, \tilde{\eta} \right)$. Then define for each $i \in \{0,1\}$,
    $$
        Z_{k}(i) := J_{\pi^{\star}}^{\Theta} - J_{{\pi}_{k}^{(i)}}^{\Theta} - \varepsilon \quad ; \quad \Tilde{Z}_{k}(i) := J_{{\pi}_{k}^{(i)}}^{\hat{\Theta}_{k}^{(i)}} - J_{{\pi}_{k}^{(i)}}^{\Theta} - \varepsilon \quad ; \quad   I_{k}(i) := \Ibf \left\{ J^{\Theta}_{\pi_{k}^{(i)}} \neq J^{\Theta}_{\pi^{\star}} - \varepsilon \right\}
   $$

    First, note that $Z_{k}(i) = Z_{k}(i) I_{k}(i)$ with probability 1. Then compute,
    \begin{align*}
        \E{\Dcal_{k}}[Z_{k}(i) - \Tilde{Z}_{k}(i) I_{k}(i) ] &= \E{\Dcal_{k}} \left[ \left(Z_{k}(i) - \Tilde{Z}_{k}(i) \right) I_{k}(i) \right] = \E{\Dcal_{k}} \left[ \left(J_{\pi^{\star}}^{\Theta} - J_{{\pi}_{k}^{(i)}}^{\hat{\Theta}_{k}^{(i)}} \right) I_{k}(i) \right] \\
        &= \E{\Dcal_{k}} \left[J_{\pi^{\star}}^{\Theta} \Ibf \left\{ J^{\Theta}_{\pi_{k}^{(i)}} \neq J^{\Theta}_{\pi^{\star}} - \varepsilon \right\}  \right] - \E{\Dcal_{k}} \left[J_{{\pi}_{k}^{(i)}}^{\hat{\Theta}_{k}^{(i)}} \Ibf \left\{ J^{\Theta}_{\pi_{k}^{(i)}} \neq J^{\Theta}_{\pi^{\star}} - \varepsilon \right\} \right]= 0,
    \end{align*}
    where the last equality is true since $\Theta$ and $\hat{\Theta}_{k}^{(i)}$ are independently identically distributed given $\mathcal{D}_k$. Therefore, we can write the simple Bayesian regret upper bounded by $ \Ebb[\Tilde{Z}_{K+1}(0) I_{K+1}(0)]$. By Cauchy-Schwartz inequality, we have
    \begin{align*}
    \Ebb[\Tilde{Z}_{K+1}(0) I_{K+1}(0)] \leq \sqrt{\left(\Ebb[I_{K+1}(0)^2] \right) \left(  \Ebb[\Tilde{Z}_{K+1}(0)^2]\right) } 
    \end{align*}

    Since $\pi^{\star} := \argmax_{\pi} J^{\Theta}_{\pi}$, using Lemma \ref{lemma:contractionPS} in conjunction with Appendix B of \cite{hao2023bridging}, the first part can be bounded by
    \begin{align*} 
        \Ebb[I_{K+1}(0)^2] &\leq \Pr \left( J^{\Theta}_{\pi_{K+1}^{(0)}} < J^{\Theta}_{\pi^{\star}} - \varepsilon \right) \leq \max_{i \in \{0,1\}} \Pr \left( J^{\Theta}_{\pi_{1}^{(i)}} < J^{\Theta}_{\pi^{\star}} - \varepsilon \right) \leq \delta_{2} . \label{eq:sumik}
    \end{align*}

    Let $\mathcal{T}_{\pi_h}^\Theta$ be the Bellman operator at time $h$ defined by $\mathcal{T}_{\pi_h}^\Theta V_{h+1}(s) := r_{\theta}(s_{h}, a_{h}) + \sum_{s'\in\mathcal{S}} V_{h+1}(s') \Pbb_{\eta}(s'|s_{h}, \pi(s_{h}))$ and $\mathcal{T}_{\pi_{H}}^\Theta V_{H+1}(s) = 0$ for all $s \in \Scal$. Using Equation (6) of \cite{osband2013more} (also see Lemma A.14 of \cite{zhang2024policy}), we have
    \begin{align*}
        \Tilde{Z}_{K+1}(0) = \E{\Theta, \hat{\Theta}_{K+1}^{(0)}} \left[\sum_{h=1}^{H} \left[ \mathcal{T}_{{\pi}_h^{K+1}}^{\hat{\Theta}_{K+1}^{(0)}} V_{h+1}^{\hat{\Theta}_{K+1}^{(0)}} (s_{K+1,h}) - \mathcal{T}_{{\pi}_h^{K+1}}^{\Theta} V_{h+1}^{\hat{\Theta}_{K+1}^{(0)}} (s_{K+1,h}) \right]  \right]
    \end{align*}
    
    Recall that the instantaneous reward satisfies $r_{\theta}(\cdot, \cdot)\in [0, 1]$. So we have with $a_{K+1,h} := \pi_{K+1}(s_{K+1,h})$,

    \begin{equation*}
    \resizebox{\linewidth}{!}{$
    \begin{aligned}
       \left| \mathcal{T}_{{\pi}_h^{K+1}}^{\hat{\Theta}_{K+1}^{(0)}} V_{h+1}^{\hat{\Theta}_{K+1}^{(0)}} (s_{K+1,h}) - \mathcal{T}_{{\pi}_h^{K+1}}^{\Theta} V_{h+1}^{\hat{\Theta}_{K+1}^{(0)}} (s_{K+1,h}) \right| \leq H \norm{{\Pbb}_{\hat{\eta}_{K+1}^{(0)}}(\cdot \given s_{K+1,h}, a_{K+1,h}) - \Pbb_{\eta}(\cdot \given s_{K+1,h}, a_{K+1,h})}{1}{1} 
    \end{aligned}
    $}
    \end{equation*}

    Therefore,
    \begin{align*}
        \Ebb[\Tilde{Z}_{K+1}(0)^2] &\leq H \Ebb\left[\sum_{h=1}^{H} \left[ \mathcal{T}_{{\pi}_h^{K+1}}^{\hat{\Theta}_{K+1}^{(0)}} V_{h+1}^{\hat{\Theta}_{K+1}^{(0)}} (s_{K+1,h}) - \mathcal{T}_{{\pi}_h^{K+1}}^{\Theta} V_{h+1}^{\hat{\Theta}_{K+1}^{(0)}} (s_{K+1,h}) \right]^2 \right] \tag{Cauchy-Schwartz} \\
        &\leq H^3 \Ebb\left[\sum_{h=1}^{H} \norm{{\Pbb}_{\hat{\eta}_{K+1}^{(0)}}(\cdot \given s_{K+1,h}, a_{K+1,h}) - \Pbb_{\eta}(\cdot \given s_{K+1,h}, a_{K+1,h})}{1}{2} \right] \\
        &\leq 5H^{3} \Ebb \left[ \sum_{h=1}^{H} \frac{4S \ln \left( \frac{2KSA}{\delta_{1}}  \right)}{\sum_{j<K+1} w_{j} (s_{K+1,h}, a_{K+1,h}) - 2H\ln \frac{SAH}{\delta_{1}}} \right]
    \end{align*}

, where the last line holds with probability $(1-2\delta_{1})$ from Lemma \ref{lemma:transition_est_error} and Section H.3 in \cite{zanette2019tighter}. Now it remains to bound $\sum_{j<K+1} w_{j} (s_{K+1,h}, a_{K+1,h})$, which is essentially the sum of probabilities of visiting the pair $(s_{K+1,h}, a_{K+1,h})$ before the $(K+1)^{\text{th}}$ episode. We know from Appendix G of \cite{zanette2019tighter} (also see Appendix B of \cite{jin2019learning}) that with probability at least $(1-\delta')$, for any state-action $(s,a)$ in any episode $e$, we have $ \frac{1}{4} w_{e}(s, a) \geq H \ln \frac{SAH}{\delta'} + H \; .$ Using this, we have with probability at least $(1-3\delta_{1})$,


\begin{align*}
    \Ebb[\Tilde{Z}_{K+1}(0)^2] &\leq 5H^{3} \Ebb \left[ \frac{4SH \ln \left( \frac{2KSA}{\delta_{1}} \right) + \varepsilon^{2}  }{4HK \left(1 + \ln \frac{SAH}{\delta_{1}} \right) - 2H\ln \frac{SAH}{\delta_{1}}} \right] \\    
    &\leq \frac{20S^{2}AH^{4} \ln \left( \frac{2KSA}{\delta_{1}}  \right) + 5SAH^{3}\varepsilon^{2}  }{4HK \left(1 + \ln \frac{SAH}{\delta_{1}} \right) - 2H\ln \frac{SAH}{\delta_{1}}}  \\
    &\leq \frac{10S^{2}AH^{3} \ln \left( \frac{2KSA}{\delta_{1}}  \right) + 3SAH^{2}\varepsilon^{2} }{2K \left(1 + \ln \frac{SAH}{\delta_{1}} \right) - \ln \frac{SAH}{\delta_{1}}} \; .
\end{align*}
 
Putting it all together, we have with probability at least $(1-3\delta_{1})$, 

$$
\SR_{K}^{\Upsilon}(\pi^{\star}_{K+1}, \pi^{\star}) \leq \sqrt{ \frac{10 \delta_{2}  S^{2}AH^{3} \ln \left( \frac{2KSA}{\delta_{1}}  \right) + 3\delta_{2}SAH^{2}\varepsilon^{2}}{2K \left(1 + \ln \frac{SAH}{\delta_{1}} \right) - \ln \frac{SAH}{\delta}} }
$$

\end{proof}

\begin{restatable}{theorem}{psplfinalregretboundappendix}
\label{lemma:pspl_final_simple_regretappendix}
For any confidence $\delta_{1} \in (0,\frac{1}{3})$ and offline preference dataset size $N>2$, the simple Bayesian regret of the learner $\Upsilon$ is upper bounded with probability of at least $1-3\delta_{1}$ by, 
    \begin{equation}
    \begin{aligned}
        \SR_{K}^{\Upsilon}(\pi^{\star}_{K+1}, \pi^{\star}) \leq \sqrt{ \frac{20 \delta_{2}  S^{2}AH^{3} \ln \left( \frac{2KSA}{\delta_{1}}  \right)}{2K \left(1 + \ln \frac{SAH}{\delta_{1}} \right) - \ln \frac{SAH}{\delta_{1}}} } \;\; ,\text{where}\, \\ \delta_{2} = 2 \exp \left( - N \left( 1 + \gamma_{\beta, \lambda, N} \right)^{2} \right) + \exp \left( - \frac{N}{4} (1 - \gamma_{\beta, \lambda, N})^{3} \right)
    \end{aligned}
    \end{equation}
\end{restatable}

\begin{proof}
\label{proof:pspl_final_regret}

Define an event $\Ecal_{n} = \left\{\bar{Y}_{n} \neq \argmax_{i \in \{0,1\}} g_{\beta, \vartheta}\left (\bar{\tau}_{n}^{(i)} \right) \right\}$, i.e. at the $n$-th index of the offline preference dataset, the rater preferred the suboptimal trajectory (wrt trajectory score $g(\cdot)$). Given the optimal trajectory parity at index $n$ as $\bar{Y}_{n}^{\star} = \argmax_{i \in \{0,1\}} g_{\beta, \vartheta}(\bar{\tau}_{n}^{(i)})$, we have,

\begin{equation}
\resizebox{0.85\linewidth}{!}{$
\begin{aligned}
    \Pr (\Ecal_{n} \given \beta, \vartheta) &= 1 - \Pr (\Ecal_{n}^{c} \given \beta, \vartheta) \\
    &= 1 - \frac{g_{\beta, \vartheta} \left(\bar{\tau}_{n}^{\bar{Y}_{n}^{\star}} \right)}{g_{\beta, \vartheta} \left(\bar{\tau}_{n}^{\bar{Y}_{n}^{\star}} \right) + g_{\beta, \vartheta}\left(\bar{\tau}_{n}^{(1-\bar{Y}_{n}^{\star})} \right) } \\ 
    &= 1 - \frac{1}{1 + \exp \left( \beta \left\langle \phi \left(\bar{\tau}_{n}^{(\bar{Y}_{n}^{\star})}\right) - \phi \left(\bar{\tau}_{n}^{(1-\bar{Y}_{n}^{\star})} \right) ,  - \vartheta \right\rangle \right)} \\
    &= 1 - \frac{1}{1 + \exp \left( \beta \left\langle \phi \left(\bar{\tau}_{n}^{(\bar{Y}_{n}^{\star})}\right) - \phi \left(\bar{\tau}_{n}^{(1-\bar{Y}_{n}^{\star})} \right) , \theta - \vartheta \right\rangle - \beta \left\langle \phi \left(\bar{\tau}_{n}^{(\bar{Y}_{n}^{\star})}\right) - \phi \left(\bar{\tau}_{n}^{(1-\bar{Y}_{n}^{\star})} \right) , \theta \right\rangle \right)} \\ 
    &\leq 1 - \frac{1}{1 + \exp \left( \beta \left\| \phi \left(\bar{\tau}_{n}^{(\bar{Y}_{n}^{\star})}\right) - \phi \left(\bar{\tau}_{n}^{(1-\bar{Y}_{n}^{\star})} \right) \right\|_{1} \|\theta-\vartheta\|_{\infty} - \beta \left\langle \phi \left(\bar{\tau}_{n}^{(\bar{Y}_{n}^{\star})}\right) - \phi \left(\bar{\tau}_{n}^{(1-\bar{Y}_{n}^{\star})} \right) , \theta \right\rangle \right)} \\
    &\leq 1 - \frac{1}{1 + \underbrace{\exp\left( \beta B \|\vartheta-\theta\|_{\infty} - \beta \left\langle \phi \left(\bar{\tau}_{n}^{(\bar{Y}_{n}^{\star})}\right) - \phi \left(\bar{\tau}_{n}^{(1-\bar{Y}_{n}^{\star})} \right) , \theta \right\rangle   \right)}_{\clubsuit}} \nonumber
\end{aligned}
$}
\end{equation}

, where the last two lines use H\"{o}lder's inequality, and bounded trajectory map assumption respectively. Since $\vartheta-\theta\sim \Ncal(0, \Ibf_d/\lambda^2)$, using the Dvoretzky–Kiefer–Wolfowitz inequality bound \citep{massart1990tight, vershynin2010introduction} implies 
\begin{equation*}
     \Pr \left(\|\vartheta-\theta\|_{\infty}\geq t\right)\leq 2 d^{1/2} \exp\left(-\frac{t^2\lambda^2}{2}\right)\,.
\end{equation*}
Set $t=\sqrt{2\ln(2d^{1/2}N)}/\lambda$ and define an event $\Ecal_{\theta, \vartheta}:=\{\|\vartheta-\theta\|_{\infty}\leq \sqrt{2\ln(2d^{1/2}N)}/\lambda\}$ such that $P(\Ecal_{\theta, \vartheta}^{c})\leq 1/N$. We apply Union Bound on the $\clubsuit$ term, and decompose the entire right hand side term as:

\begin{equation}
\label{eq:rater_pref_error}
\resizebox{0.55\linewidth}{!}{$
\begin{aligned}
     \Pr (\Ecal_{n} \given \beta, \vartheta) &\leq \frac{1}{1 + \exp \left( \beta B \sqrt{2\ln(2d^{1/2}N)}/\lambda + \beta \Delta_{\min} \right)} + \frac{1}{N} \\
     &\leq \exp\left(- \beta B \sqrt{2\ln(2d^{1/2}N)}/\lambda - \beta \Delta_{\min} \right) + \frac{1}{N} := \gamma_{\beta, \lambda, N} \; .
\end{aligned}
$}
\end{equation}

Now we need to provide conditions on the rater's competence, in terms of $(\lambda, \beta)$ to be a valid expert i.e. for $\gamma_{\beta, \lambda, N} \in (0,1)$. Let $k_{1} = \beta B, k_{2} = \frac{2 \ln \left(2 d^{1/2}\right)}{\lambda^{2}}, k_{3} = 2/\lambda^{2}$, and $k_{4} = \beta \Delta_{\min}$. We then have,

\begin{align*}
    \exp\left(- k_{1} \sqrt{k_{2} + k_{3} \ln N}  - k_{4} \right) + \frac{1}{N} &< 1 \\
    -k_{1} \sqrt{k_{2} +  k_{3} \ln N } - k_{4}< \ln N \\
    \left( \ln N \right)^{2} + (2k_{4} - k_{1}^{2}k_{3}) \ln N + (k_{4}^{2} - k_{1}^{2}k_{2}) > 0
\end{align*}

The above is a quadratic inequality that holds for all $N>1$ if $\beta$ is large enough i.e.

$$
\text{If} \; \beta > \frac{2\ln \left(2 d^{1/2} \right)}{\left| B\lambda^{2} - 2 \Delta_{\min}  \right|}, \text{then,} \; \gamma_{\beta, \lambda, N} \in (0,1) \, .
$$

Now, since we have Lemma \ref{lemma:all_states_actions_visited}, the remaining argument shows to find a separation of probability between two types of states and time index pairs under the rater's preference, parameterized by $\beta$ and $\lambda$, and the offline dataset, characterized by its size $N$: the ones that are probable under the optimal policy $\pi^{\star}$, and the ones that are not. We have two cases:

\textbf{Case I.} $p^{\pi^{\star}}_{h}(s) > 0$

In this case, we want the rater to prefer trajectories that are most likely to occur under the optimal policy $\pi^{\star}$. Given $p^{\pi^{\star}}_{h}(s) > 0$, we now lower bound the probability of the state $s$ having $c_{h}(s) \equiv \sum_{a \in \Acal} c_{h}(s,a) > 0$ i.e.

\begin{equation}
\label{eq:lower_bound_chs}
\begin{aligned}
     \Pr (c_{h}(s) > 0) &= \Pr \left( \sum_{a} w_{h}(s,a) > \sum_{a} l_{h}(s,a) \right) = 1 - \Pr \left( \sum_{a} w_{h}(s,a) \leq \sum_{a} l_{h}(s,a) \right) \\
     &= 1 - \sum_{t=0}^{H} \Pr \left( \sum_{a} w_{h}(s,a) \leq t \right) \cdot \Pr \left( \sum_{a} l_{h}(s,a) = t \right) \\
     &\geq 1 - \sum_{t=1}^{H+1} \Pr \left( \sum_{a} w_{h}(s,a) < t \right) \\
     &\geq 1 - \gamma_{\beta, \lambda, N} / (1 - \gamma_{\beta, \lambda, N})
\end{aligned}
\end{equation}

, where the last step uses Equation \eqref{eq:rater_pref_error}.

\textbf{Case II.} $p^{\pi^{\star}}_{h}(s) = 0$

In this case, we wish to upper bound the probability of the rater preferring trajectories (and hence, states) which are unlikely to occur under the optimal policy $\pi^{\star}$. If for some state $s$ we have $p^{\pi^{\star}}_{h}(s) = 0$ but also this state $s \in \bar{\tau}_{n}^{(1 - \bar{Y}_{n}^{\star})}$, this means the rater preferred the suboptimal trajectory. Similar to the proof above we conclude that,

\begin{equation}
\label{eq:upper_bound_chs}
\Pr (c_{h}(s) > 0) = \Pr \left(\bigcup_{n=1}^{N} \{ \Ecal_{n} \} \given \beta, \vartheta \right) \leq \sum_{n=1}^{N} \Pr_\theta(\Ecal_{n} \given \beta, \vartheta) \leq \gamma_{\beta, \lambda, N} / (1 - \gamma_{\beta, \lambda, N})
\end{equation}

The above argument shows that there's a probability gap between parity of states and time index pairs under the rater's preference model: the ones that are probable under the optimal policy $\pi^{\star}$ and the ones that are not i.e. we have shown that the states which are \emph{more} likely to be visited under $\pi^{\star}$ have a lower bound on the probability of being part of preferred trajectories, and also the states which are \emph{less} likely to be visited by $\pi^{\star}$ have an upper bound on the probability of being part of preferred trajectories by the rater. Using this decomposition, we will show that when the rater tends to an expert ($\beta \to \infty, \lambda \to \infty$) and size $N$ of the offline preference dataset $\Dcal_{0}$ is large, we can distinguish the two types of state and time index pairs through their net counts in $\Dcal_0$. This will allow us to construct an $\varepsilon$-optimal estimate of $\pi^{\star}$ with probability at least $(1-\delta_{2})$. Noticing the structure of Equation \eqref{eq:pspl_prior_error_bound}, we see that the upper bound is minimized when $\varepsilon \to 0$, i.e. we construct an \emph{optimal} estimate of the optimal policy from the offline preference dataset. We now upper bound the probability that the estimated optimal policy $\hat{\pi}^{\star}$ is \emph{not} the optimal policy.

Let $\hat{\pi}^{\star}$ be the optimal estimator of $\pi^{\star}$ constructed with probability at least $(1-\delta_{2})$. Based on the separability of states and time index pairs, we have four possible events for each $(s,h) \in \Scal \times [H]$ pair. For $\delta = (1-\gamma_{\beta, \lambda, N})/2$, we have,

\begin{enumerate}
    \item $ E_{1} := \{ p_{h}^{\pi^{\star}}(s) > 0$ and $c_h(s) < \delta N \}$;
    \item $E_{2} := \{p_{h}^{\pi^{\star}}(s) > 0$ and $c_h(s) \geq \delta N$, but $\pi_{h}^{\star}(s) = a_{h}^{\star}(s) \neq \argmax_{a} c_h(s, a) = \hat{\pi}_{h}^{\star}(s) \}$.
    \item $E_{3} := \{p_{h}^{\pi^{\star}}(s) = 0$ and $c_h(s) \geq \delta N \}$;
    \item $E_{4} := \{p_{h}^{\pi^{\star}}(s) = 0$ and $c_h(s) < \delta N\}$;
\end{enumerate}

Denoting the the event to occur with high probability as $\Tcal$ i.e. $\Tcal := \{ r_{\theta}(s_{h}, \pi_{h}^{\star}(s_{h})) - r_{\theta}(s_{h}, \hat{\pi}^{\star}_{h}(s_{h})) \leq 0 \}$.
If we can show that $\Pr(\Tcal \given E_{i}) > 1 - \delta_{2}$ for $i \in \{1,2,3,4\}$, then union bound implies that $\hat{\pi}^{\star}$ is not optimal with probability at most $\delta_{2}$. 


\begin{enumerate}
    \item \textbf{Under the event $E_{1}$.}  Let $b \sim \mathrm{Bin}(T, q)$ denote a binomial random variable with parameters $T \in \Nbb$ and $q \in [0, 1]$. Notice that the each $c_{h}(s)$ is the difference of two binomial random variables $b_{1} \sim \mathrm{Bin}(N, 1 - \gamma_{\beta, \lambda, N})$ and $b_{2} \sim \mathrm{Bin}(N, \gamma_{\beta, \lambda, N})$. This implies that $c_{h}(s) + N \sim \mathrm{Bin}(2N, 1 - \gamma_{\beta, \lambda, N})$. We then have,

    \begin{align*}
        \Pr(\Tcal^{c} \given E_{1}) &\leq \Pr(c_{h}(s) < \delta N)  
        \leq \Pr \left( \mathrm{Bin}(2N, 1 - \gamma_{\beta, \lambda, N}) < (1 + \delta) N \right) \\
        &\leq \exp \left( - 4N \left( \delta + \gamma_{\beta, \lambda, N} \right)^{2}  \right) \leq \exp \left( - N \left( 1 + \gamma_{\beta, \lambda, N} \right)^{2} \right)
    \end{align*}

    \item \textbf{Under the event $E_{2}$.} Given Equation \eqref{eq:lower_bound_chs}, we have that,

    \begin{align*}
        \Pr(\Tcal^{c} \given E_{2}) &\leq   \Pr \left( \argmax_{a \in U^{W}_{h}(s)} Q^{\pi^{\star}}_{\theta, \eta, h}(s, a) \neq \argmax_{a \in U^{W}_{h}(s)} Q^{\pi^{\star}}_{\theta, \eta, h}(s, a) \given E_{2} \right) \\
        &\leq   \Pr \left(c_{h} \left(s, \argmax_{a \in U^{W}_{h}(s)} Q^{\pi^{\star}}_{\theta, \eta, h}(s, a) \right) \leq c_{h}(s)/2 \given E_{2} \right) \\ 
        &\leq   \Pr \left( \mathrm{Bin}(c_{h}(s), 1 - \gamma_{\beta, \lambda, N}) \geq c_{h}(s)/2 \given E_{2} \right) \\
        &\leq   \left[ \exp \left( -2c_{h}(s) (1 - \gamma_{\beta, \lambda, N} - c_{h}(s)/2)^{2} \right) \right] \rvert_{c_{h}(s) = \delta N} \\
        &\leq   \exp \left( - \frac{N}{4} (1 - \gamma_{\beta, \lambda, N})^{3} \right) 
    \end{align*}

    \item \textbf{Under the event $E_{3}$.} Similar to event $E_{1}$, we have, 

    \begin{align*}
        \Pr(\Tcal^{c} \given E_{3}) &\leq \Pr \left( \mathrm{Bin}(2N, 1 - \gamma_{\beta, \lambda, N}) > (1+\delta) N \right) \\ 
        &\leq \exp \left( - 4N \left( \delta + \gamma_{\beta, \lambda, N} \right)^{2}  \right) \leq \exp \left( - N \left( 1 + \gamma_{\beta, \lambda, N} \right)^{2} \right)
    \end{align*}

    \item \textbf{Under the event $E_{4}$.} Under this event, notice that conditioned on $\theta$, we have 

    \begin{align*}
        r_{\theta}(s_{h}, \pi_{h}^{\star}(s_{h})) - r_{\theta}(s_{h}, \hat{\pi}^{\star}_{h}(s_{h})) &= \Ebb \left[ \E{\theta} \left[ r_{\theta}(s_{h}, \pi_{h}^{\star}(s_{h})) - r_{\theta}(s_{h}, \hat{\pi}^{\star}_{h}(s))  \right]  \right] \\
        &= \E{\mathring{a} \sim \Acal} \left[ \E{\theta} \left[ r_{\theta}(s_{h}, \mathring{a}) - r_{\theta}(s_{h}, \mathring{a})  \right]  \right] = 0
    \end{align*}

    This means that $\Pr(\Tcal \given E_{4})$ is a non-failure event that occurs with probability 1 i.e. $\Pr(\Tcal^{c} \given E_{4}) = 0$.

\end{enumerate}

Combining all of the above, we have for $N > 2$, 

\begin{align*}
\Pr(\Tcal) &\geq  1 - \left(2 \exp \left( - N \left( 1 + \gamma_{\beta, \lambda, N} \right)^{2} \right) + \exp \left( - \frac{N}{4} (1 - \gamma_{\beta, \lambda, N})^{3} \right) \right) \\
&\geq 1 - \delta_{2}
\end{align*}

Using Lemma \ref{lemma:pspl_error_regret}, the proof is complete.

\end{proof}

\subsection{Results for Bandits}
\label{sec:appendixbandits}

For the bandit setting we let the action set be $\Acal \subseteq \Rbb^{d} $ with number of arms be $A=|\Acal|$, online episodes (rounds) be $K$, and offline dataset $\Dcal_{0} = \left\{ \left( \bar{a}^{(0)}_{n}, \bar{a}^{(1)}_{n}, Y_{n} \right) \right\}_{n=1}^{N}$ of size $N$, where $\bar{a}^{(0)}_{n}, \bar{a}^{(1)}_{n} \in \Acal$. Also let $\mu_{\text{min}}(\cdot) \in (0,1)$ be the minimum action sampling distribution during construction of $\Dcal_0$.

Letting $\Delta_{\text{min}} = \min_{n \in [N]} \left| r_{\theta}(\bar{a}^{(0)}_{n}) - r_{\theta}(\bar{a}^{(1)}_{n}) \right|$, where $\theta$ is the underlying reward model of the environment with $r_{\theta}(\cdot) : \Rbb^{d} \to \Rbb$, and $\gamma_{\beta, \lambda, N}$ to be error upper bound of the rater's preference (similar to Lemma \ref{lemma:rater_error_upper_bound}), we have the following result for simple regret of the learner $\Upsilon$, where simple regret is defined as $\SR_{K}^{\Upsilon}(\pi_{K+1}^{\star}, \pi^{\star}) = r_{\theta}(a^{\star}) - r_{\theta}(a^{\star}_{K+1})$, where $\pi^{\star}$ is the optimal policy that picks the optimal action $a^{\star} = \argmax_{a \in \Acal} r_{\theta}(a)$, and $\pi_{K+1}^{\star}$ is the policy of the learner after $K$ online rounds. We shall use $A^{\star}$ and $a^{\star}$ interchangebly to refer to the optimal action.

Analogous to the winning and undecided subsets of Section \ref{sec:analysis}, we construct an \emph{information} subset of $\Acal$, call it $\Ucal_{\Dcal_{0}}$ such that $\Pr(a^{\star} \in \Ucal_{\Dcal_{0}}) \geq 1 - \epsilon$, where $\epsilon \in (0,1)$ is the \emph{error} probability. As an algorithmic choice, we let $\Ucal_{\Dcal_{0}}$ consist of actions that have been preferred to \emph{at least} once in the offline preference dataset and of actions that do not appear in $\Dcal_{0}$. Given this construction, we have the following result.

\begin{restatable}{theorem}{banditregretbound}
\label{th:bandit_final_simple_regret}
For any confidence $\delta_{1} \in (0,\frac{1}{3})$ and offline preference dataset size $N>2$, the simple Bayesian regret of the learner $\Upsilon$ is upper bounded with probability of at least $1-3\delta_{1}$ by, 
    \begin{equation}
    \begin{aligned}
        \SR_{K}^{\Upsilon}(\pi^{\star}_{K+1}, \pi^{\star}) & \leq \sqrt{ \frac{20 \delta_{2}  A \ln \left( \frac{2KA}{\delta_{1}}  \right)}{2K \left(1 + \ln \frac{A}{\delta_{1}} \right) - \ln \frac{A}{\delta_{1}}} } \quad \text{with} , \\
        \; \gamma_{\beta, \lambda, N} &= \exp\left(- \beta B \sqrt{2\ln(2d^{1/2}N)}/\lambda - \beta \Delta_{\min} \right)  + \frac{1}{N}  \;\; \text{, and} \\ 
        \; \delta_{2} &= 2 \exp \left( - N \left( 1 + \gamma_{\beta, \lambda, N} \right)^{2} \right) + \exp \left( - \frac{N}{4} (1 - \gamma_{\beta, \lambda, N})^{3} \right).
    \end{aligned}
    \end{equation}
\end{restatable}

For a fixed $N>2$, and large $A$, the simple regret bound is $\widetilde{\Ocal}\left( \sqrt{AK^{-1}} \right)$. Note that this bound converges to zero exponentially fast as $N \to \infty$ and as the rater tends to an expert (large $\beta, \lambda$). In addition, as the number of online episodes $K$ gets large, $\PSPL$ is able to identify the best policy (arm) with probability at least $(1-3\delta_{1})$.

\subsection{Constructing Surrogate Loss Function}
\label{appendix:chapter4surrogatelossfunction}

\begin{restatable}{lemma}{mapestimatelemmaappendix}
\label{th:mapestimatelemmaappendix} 
At episode $k$, the MAP estimate of $(\theta, \vartheta, \eta)$ can be constructed by solving the following equivalent optimization problem: 
\begin{equation}
\begin{aligned}
(\theta_{opt}, \vartheta_{opt}, \eta_{opt})  &= \underset{\theta, \vartheta, \eta}{\argmax} \; \Pr(\theta, \vartheta, \eta \, | \, \Dcal_{k}) \\ & \equiv \underset{\theta, \vartheta, \eta}{\argmin} \; \Lcal_{1}(\theta, \vartheta, \eta) +  \Lcal_{2}(\theta, \vartheta, \eta) +  \Lcal_{3}(\theta, \vartheta, \eta) \; , \text{where},  \\
 \Lcal_{1}(\theta, \vartheta, \eta)  &:= - \sum_{t=1}^{k-1} \vphantom{\int_1^2} \left[ \beta \langle {\tau}_t^{(Y_{t})} , \vartheta \rangle - \ln \bigg(e^{ \beta \langle {\tau}_t^{(0)}, \vartheta \rangle} + e^{\beta \langle {\tau}_t^{(1)}, \vartheta \rangle} \bigg) \right. \\ & \left. \qquad + \sum_{j=0}^{1} \sum_{h=1}^{H-1} \ln \Pr_{\eta}\left({s}_{t,h+1}^{(j)} \given {s}_{t,h}^{(j)}, {a}_{t,h}^{(j)} \right) \vphantom{\int_1^2} \right], \\
 \Lcal_{2}(\theta, \vartheta, \eta) &:= - \sum_{n=1}^{N} \left[ \beta \langle \bar{\tau}_n^{(\bar{Y}_{n})} , \vartheta \rangle - \ln \bigg(e^{ \beta \langle \bar{\tau}_n^{(0)}, \vartheta \rangle} + e^{\beta \langle \bar{\tau}_n^{(1)}, \vartheta \rangle} \bigg) \right],  \\
 \Lcal_{3}(\theta, \vartheta, \eta) &:= \frac{\lambda^2}{2} \norm{\theta - \vartheta}{2}{2} - SA \sum_{i=1}^{S} (\bm{\alpha}_{0,i} - 1) \ln \eta_{i} \\ & \qquad + \frac{1}{2} (\theta - \mu_{0})^{T} \Sigma_{0}^{-1} (\theta - \mu_{0}).
\end{aligned}
\label{eq:mapestimateproblemappendix}
\end{equation}
\end{restatable}
\begin{proof}
\label{proof:chapter4mapestimatelemma}

We first analyze the posterior distribution of $\vartheta, \theta, \eta$ given the dataset $\Dcal_{k}$ at the beginning of episode $k$, and then optimize it by treating these random variables as parameters.

\begin{equation}
\label{eq:chapter4map_surrogate}
\begin{aligned}
    \underset{\theta, \vartheta, \eta}{\argmax} \; \Pr(\theta, \vartheta, \eta \, | \, \Dcal_{k}) &= \underset{\theta, \vartheta, \eta}{\argmax} \; \Pr( \Dcal_{k} \, | \, \theta, \vartheta, \eta) \cdot \Pr(\theta, \vartheta, \eta) \\ 
    &= \underset{\theta, \vartheta, \eta}{\argmax} \; \ln \Pr(\Dcal_{k} \, | \, \theta, \vartheta, \eta) + \ln \Pr(\theta, \vartheta, \eta) \\ 
    &= \underset{\theta, \vartheta, \eta}{\argmax} \underbrace{\ln \Pr(\Hcal_{k} \,| \, \Dcal_{0}, \theta, \vartheta, \eta)}_{\Lcal_{1}} + \underbrace{\ln \Pr(\Dcal_{0} \, | \, \theta, \vartheta, \eta)}_{\Lcal_{2}} +  \underbrace{\ln \Pr(\theta, \vartheta, \eta)}_{\Lcal_{3}}
\end{aligned}
\end{equation}

Then,
\begin{equation}
    \begin{aligned}
        \Lcal_{1} &= \sum_{t=1}^{k-1} \ln \Pr \bigg( \left({\tau}_t^{(0)}, {\tau}_t^{(1)}, Y_t\right) \, \big| \, \Dcal_{t}, \theta, \vartheta, \eta \bigg) \\
        &= \sum_{t=1}^{k-1} \ln \Pr \bigg (Y_t \, \big | \, {\tau}_t^{(0)}, {\tau}_t^{(1)}, \theta, \vartheta, \eta \bigg) + \ln \Pr \bigg( {\tau}_t^{(0)}, {\tau}_t^{(1)} \Given \Dcal_{t}, \theta, \vartheta, \eta \bigg) \\
        &= \sum_{t=1}^{k-1} \left[ \beta \langle {\tau}_t^{(Y_{t})} , \vartheta \rangle - \ln \bigg(e^{ \beta \langle {\tau}_t^{(0)}, \vartheta \rangle} + e^{\beta \langle {\tau}_t^{(1)}, \vartheta \rangle} \bigg) + \sum_{j=0}^{1} \sum_{h=1}^{H-1} \ln \Pr_{\eta} \left({s}_{t,h+1}^{(j)} \given {s}_{t,h}^{(j)}, {a}_{t,h}^{(j)} \right) \right] \\
        \Lcal_{2} &= \sum_{n=1}^{N} \ln \Pr \bigg( \left(\bar{\tau}_n^{(0)}, \bar{\tau}_n^{(1)}, \bar{Y}_n\right) \, \big| \, \theta, \vartheta, \eta \bigg) \\
        &= \sum_{n=1}^{N} \ln \Pr \bigg (\bar{Y}_n \, \big | \, \bar{\tau}_n^{(0)}, \bar{\tau}_n^{(1)}, \theta, \vartheta, \eta \bigg) + \underbrace{\ln \Pr \bigg( \bar{\tau}_n^{(0)}, \bar{\tau}_n^{(1)} \Given \theta, \vartheta, \eta \bigg)}_{\textcolor{gray}{\textbf{indep. of $\theta, \vartheta, \eta \implies$ constant}}} \\
        &= \sum_{n=1}^{N} \beta \langle \bar{\tau}_n^{(\bar{Y}_{n})} , \vartheta \rangle - \ln \bigg(e^{ \beta \langle \bar{\tau}_n^{(0)}, \vartheta \rangle} + e^{\beta \langle \bar{\tau}_n^{(1)}, \vartheta \rangle} \bigg) + \textcolor{gray}{\text{constant}} \\
        \Lcal_{3} &= \ln \Pr(\vartheta \, | \, \theta) + \ln \Pr(\theta) + \ln \Pr(\eta) \\
        &= \frac{d}{2} \ln \bigg(\frac{2\pi}{\lambda^2} \bigg) - \frac{\lambda^2}{2} \norm{\theta - \vartheta}{2}{2} - \frac{1}{2} \ln \big(|2\pi \Sigma_{0}| \big) - \frac{1}{2} (\theta - \mu_{0})^{T} \Sigma_{0}^{-1} (\theta - \mu_{0}) + SA \sum_{i=1}^{S} (\bm{\alpha}_{0,i} - 1) \ln \eta_{i}.
    \end{aligned}
\end{equation}

Hence, final surrogate loss function is

\begin{equation}
\label{eq:chapter4surrogate_loss_function}
\begin{aligned}
    \Lcal(\theta, \vartheta, \eta) &= \Lcal_{1}(\theta, \vartheta, \eta) +  \Lcal_{2}(\theta, \vartheta, \eta) +  \Lcal_{3}(\theta, \vartheta, \eta), \qquad \text{where} \\
    \Lcal_{1}(\theta, \vartheta, \eta) &= - \sum_{t=1}^{k-1} \left[ \beta \langle {\tau}_t^{(Y_{t})} , \vartheta \rangle - \ln \bigg(e^{ \beta \langle {\tau}_t^{(0)}, \vartheta \rangle} + e^{\beta \langle {\tau}_t^{(1)}, \vartheta \rangle} \bigg) + \sum_{j=0}^{1} \sum_{h=1}^{H-1} \ln \Pr_{\eta} \left({s}_{t,h+1}^{(j)} \given {s}_{t,h}^{(j)}, {a}_{t,h}^{(j)} \right) \right] \, , \\ 
    \Lcal_{2}(\theta, \vartheta, \eta) &= - \sum_{n=1}^{N} \left[ \beta \langle \bar{\tau}_n^{(\bar{Y}_{n})} , \vartheta \rangle - \ln \bigg(e^{ \beta \langle \bar{\tau}_n^{(0)}, \vartheta \rangle} + e^{\beta \langle \bar{\tau}_n^{(1)}, \vartheta \rangle} \bigg) \right] \, ,\\
    \Lcal_{3}(\theta, \vartheta, \eta) &= \frac{\lambda^2}{2} \norm{\theta - \vartheta}{2}{2} + \frac{1}{2} (\theta - \mu_{0})^{T} \Sigma_{0}^{-1} (\theta - \mu_{0}) - SA \sum_{i=1}^{S} (\bm{\alpha}_{0,i} - 1) \ln \eta_{i}.
\end{aligned}
\end{equation}

Finally the problem becomes equivalent as follows:
\begin{equation}
\label{eq:chapter4surrogate_opt_problem}
(\theta_{opt}, \vartheta_{opt}, \eta_{opt}) = \underset{\theta, \vartheta, \eta}{\argmax} \; \Pr(\theta, \vartheta, \eta \, | \, \Dcal_{k}) \equiv \underset{\theta, \vartheta, \eta}{\argmin} \; \Lcal(\theta, \vartheta, \eta)
\end{equation}

\end{proof}

\subsection{Practical PSPL (cont.)}

\subsubsection{Estimating Rater Competence in Practice}
\label{sec:appendix_practical_pspl}

There are two main methods of estimating rater competence in practice:
\begin{enumerate}
    \item Based on maximum likelihood estimation (MLE). Similar idea has been proposed to estimate the expertise level in imitation learning \cite{beliaev2022imitation, beliaev2025inverse}.
    \item The second method is to simply look at the entropy of the empirical distribution of the action in the offline dataset. Suppose the empirical distribution of $\zeta$. Then we use $c/\mathcal{H}(\zeta)$ as an estimation for $\beta$, where $c > 0$ is a hyperparameter. The intuition is that for smaller $\beta$, the net state-action pair visit counts tend to be more uniform and thus the entropy will be larger. This is an unsupervised approach and agnostic to specific offline data generation processes. The knowledgeability $\lambda$ is not quite ‘estimable’ because for a single environment, even though we know the true environment $\theta$ and the expert’s knowledge $\vartheta$, we only have one pair of observations generated with the same $\vartheta$. Thus, the variance of the estimation for $\lambda$ could be infinite. However, exact estimation of $\lambda$ is not necessary as we show that the algorithm is robust to misspecified $\lambda$ through experiments in Section \ref{sec:experiments}.
\end{enumerate}

\subsubsection{Ablation Studies}
\label{sec:appendix_ablation}

The  Bootstrapped $\PSPL$ algorithm in Section \ref{sec:practical_approx_pspl} requires a knowledge of rater's parameters $\lambda, \beta$. We study the sensitivity of the algorithm's performance to mis-specification of these parameters. 

\begin{figure*}
\vspace{-0.2cm}
\begin{minipage}{\textwidth}
\begin{tcolorbox}[width=.495\textwidth, nobeforeafter, coltitle = black, fonttitle=\fontfamily{lmss}\selectfont, title=RiverSwim, halign title=flush center, colback=backg_blue!5, colframe=darkgreen!15, boxrule=2pt, grow to left by=-0.5mm, left=-15pt, right=-15pt]
    \centering
    {
        \includegraphics[height=0.35\textwidth, width=0.4\textwidth]{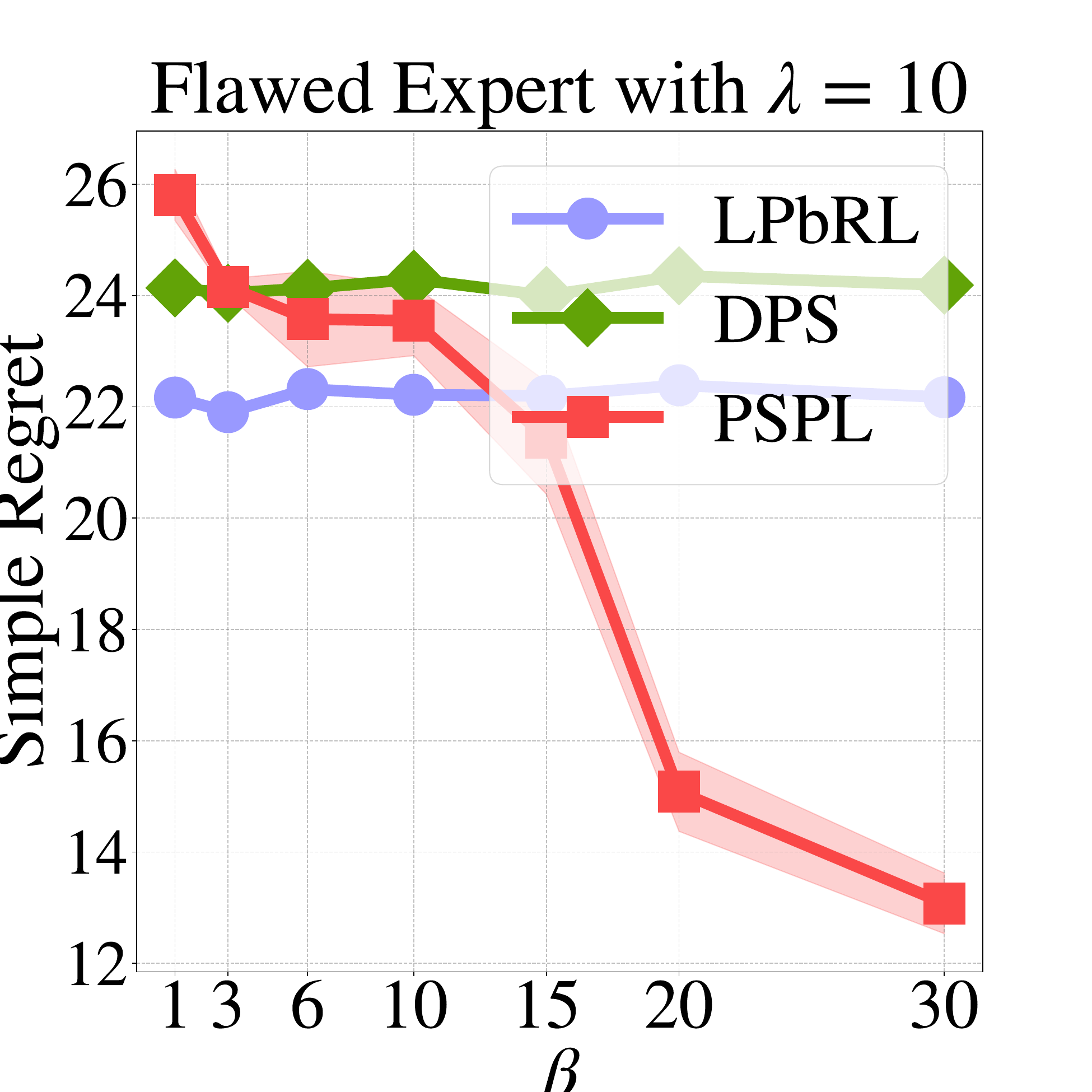}
    }
    {
        \includegraphics[height=0.35\textwidth, width=0.4\textwidth]{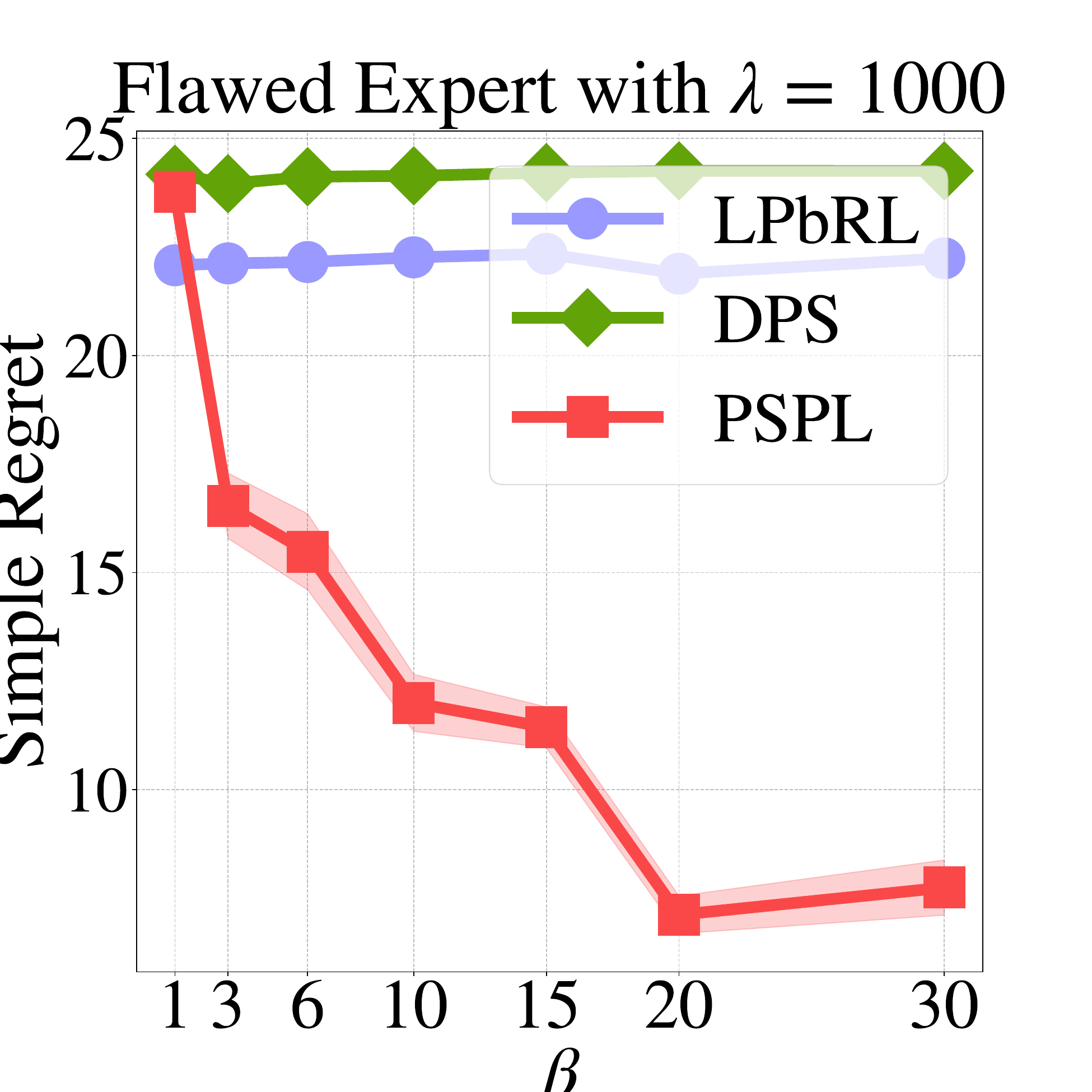}
    } 
\end{tcolorbox}  \hfill
\begin{tcolorbox}[width=.495\textwidth, nobeforeafter, coltitle = black, fonttitle=\fontfamily{lmss}\selectfont, title=RiverSwim, halign title=flush center, colback=backg_blue!5, colframe=purple!10, boxrule=2pt, grow to left by=-0.5mm, left=-15pt, right=-15pt]
    \centering
    {
        \includegraphics[height=0.35\textwidth, width=0.4\textwidth]{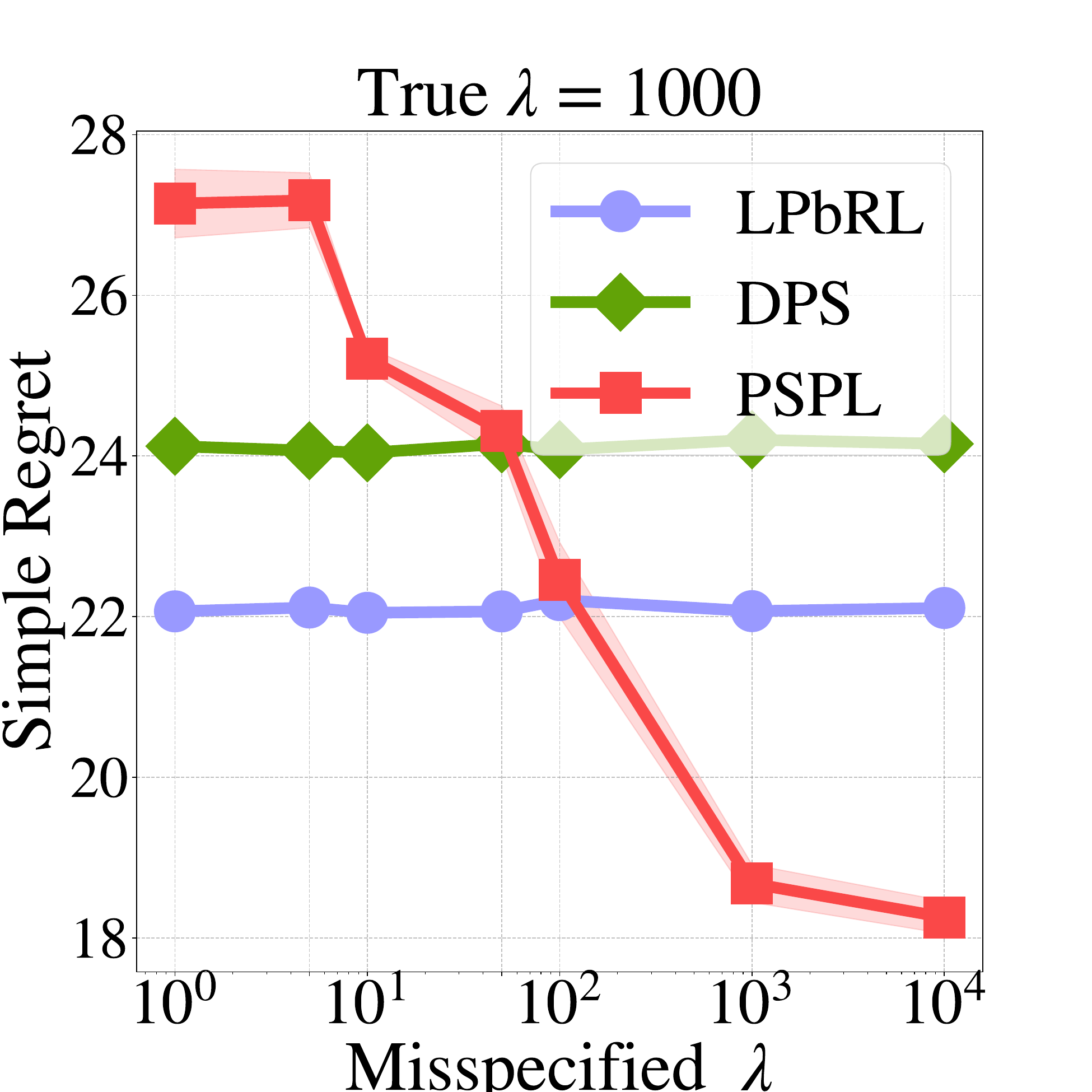}
    } 
    {
        \includegraphics[height=0.35\textwidth, width=0.4\textwidth]{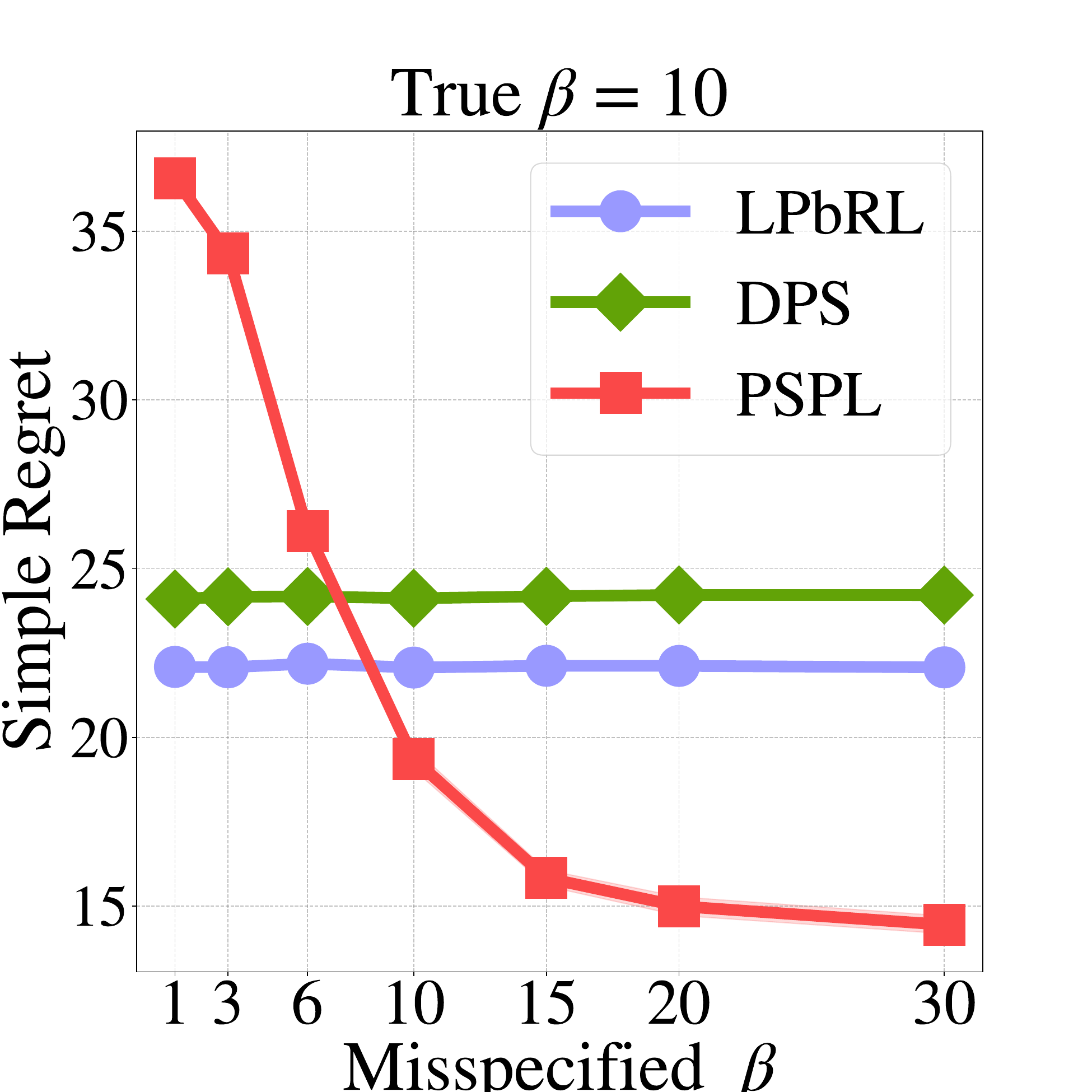}
    } 
\end{tcolorbox} 
\caption{Sensitivity to flawed expert policy with $\lambda = \{10, 10^{3}\}$, and misspecified 
 competence.}
\vspace{-0.5cm}
\label{fig:more_ablation}
\end{minipage}
\end{figure*}

(i) \textbf{Different Preference Generation Expert Policy.} Though the learning agent assumes Equation \eqref{eq:pref_logistic_trajectories} as the expert's generative model, we consider it to use a deterministic greedy policy. Trajectories $\bar{\tau}_{n}^{(0)}$ and $\bar{\tau}_{n}^{(1)}$ are sampled, and then choose $\bar{Y}_{n} = \argmax_{i \in \{0,1\}} \beta \langle \bar{\tau}_{n}^{(i)} , \vartheta \rangle$, where $\vartheta\sim\Ncal(\theta, \Ibf_d / \lambda^2)$. In Figure \ref{fig:more_ablation}, we see that even when the learning agent's assumption of the rater policy is \textit{flawed}, $\PSPL$ significantly outperforms the baselines.
\\
(ii) \textbf{Misspecified Competence parameters.} First, we generate offline data with true $\lambda=10^{3}$ but $\PSPL$ uses a misspecified $\lambda$. Second, we generate offline data with true $\beta=10$ but $\PSPL$ uses a misspecified $\beta$. Figure \ref{fig:more_ablation} shows that although the performance of $\PSPL$ decreases as the degree of flawness increases, it still outperforms the baselines.      

\subsubsection{Experiments on Image Generation Tasks (cont.)}
\label{appendix:pickapic}

We instantiate our framework on the Pick-a-Pic dataset of human preferences for text–to–image generation \cite{kirstain2023pick}. Overall, the dataset contains over 500,000 examples and 35,000 distinct prompts. Each example contains a prompt, sequence generations of two images, and a label for which image is preferred. We let each generation be a trajectory, so the dataset contains trajectory preferences $\mathcal D_{0} = { (\tau_{i}^{+},\tau_{i}^{-},y_{i}) }_{i=1}^{N}$ with $y_i = 1$ iff $\tau_{i}^{+} \succ \tau_{i}^{-}$. Each trajectory $\tau =(p,z_{0 : T})$ is the entire latent denoising chain of length $T$ for prompt $p$ sampled from some prompt distribution.

Following \cite{blacktraining, zhang2024flow}, text--to--image generation is a finite--horizon MDP $\mathcal M=(\mathcal S,\mathcal A,P)$: $s_t=(p,z_t),a_t=\epsilon_t,z_{t+1}=f_{\rho}(z_t,\epsilon_t),$ where $z_t \in \mathbb R^{d}$ is the latent, $a_t$ is the noise $\epsilon_{t}$ predicted by the policy $\pi_\theta(a_t \mid s_t)$, and $f_{\boldsymbol\alpha}$ is the deterministic DDPM transition with frozen scheduler $\rho$. Please see \cite{blacktraining} for a comprehensive and detailed discussion on modeling diffusion as a MDP; we follow the same approach. The episode horizon is $H=T=50$. For each trajectory, we adopt the additive embedding as discussed in the paper: $ \phi(\tau)=\sum_{t=1}^T \phi(s_t,a_t)$ with $\phi(s_t,a_t)=\bigl[ \text{CLIP}(p,\text{Dec}(z_T)), \lVert\epsilon_t\rVert_2 , t/T \bigr],$ where $\text{Dec}(\cdot)$ is the standard VAE decoder shipped with Stable‐Diffusion \cite{rombach2022high}. As in the paper, the model assumes that rater of competence $\lambda, \beta$ follows the Bradley-Terry model i.e. $\Pr(Y = 1\mid\tau^+,\tau^- , \vartheta,\beta) =\sigma \bigl(\beta\langle\vartheta,\phi(\tau^+) - \phi(\tau^-)\rangle\bigr)$, where $\sigma(\cdot)$ is the sigmoid link function.

To model the reward parameters, we let the uninformed prior be $\nu_0=\mathcal N(\mu_0,\Sigma_0)$, $\mu_0=0$, and diagonal $\Sigma_0= \mathrm{diag}(\sigma_{\text{clip}}^2,\sigma_{\text{noise}}^2, \sigma_{\text{time}}^2)$, where $\sigma_{\text{clip}}^2,\sigma_{\text{noise}}^2$ are empirical variances from $10$k random chains and $\sigma_{\text{time}}^2 =1/12$. Regarding modeling of transitions, the physical scheduler $f_{\rho}$ is known; uncertainty remains only in the $\ell_2$-norm of $\epsilon_t$. We discretize this norm into $C{=}10$ bins as $b_i$ for $i \in [C]$ and model $P(b_i\mid s_t) \sim \mathrm{Dir}(\boldsymbol\alpha_0), \;\; \boldsymbol\alpha_0=\mathbf 1_C.$ The resulting Dirichlet counts are updated from both $\mathcal{D}_0$ and online episodes.

Since we do not know the optimal generation sequence, minimizing simple regret is equivalent to maximizing expected reward in the final online episode. To evaluate this final generation, we conduct $L=10$ rollouts and compute the average reward with a weighted ensemble of automatic quality metrics, similar to \cite{clark2023directly, xu2023imagereward} i.e. $r_{\theta}(\cdot)= 0.7*\text{ImageReward--v2} + 0.3*\text{Aesthetic--LAION}$ where ImageReward--v2 is a reward model from \cite{xu2023imagereward}, and Aesthetic--LAION is a reward model hosted on HuggingFace \cite{laion_ev_2025}. For evaluation, we sweep $N_{\text{off}} = 50$k preference triplets for the prior, reserve $N_{\text{val}} = 15$k examples for evaluation, and create an exploration pool of $10$k unseen examples. For implementation, we use a 2 layer MLP, 512 GELU units, and outputs $(\mu_\theta)_{t}$ and $\log ((\sigma_\theta)_{t})$, where $(\mu_\theta)_{t}$ is the predicted noise vector the agent believes will best denoise $z_{t}$ and $(\sigma_\theta)_{t}$ controls exploratory perturbations around that prediction, enabling posterior sampling for PSPL. This is analogous to solving the unconstrained optimization Problem \eqref{eq:surrogate_perturbed_loss} in the function approximation setting. 

For the online phase, we use the above reward model $r_{\theta}(\cdot)$ for the BT preference model (see Equation \eqref{eq:pref_logistic_trajectories}), with an expert rater (i.e. $\lambda,\beta \to \infty$), similar to \cite{kirstain2023pick}. Since, we show that PSPL is robust to mis-specifications in rater competence (please see Appendix \ref{sec:appendix_ablation}), we use an expert rater for ease of comparison.

\begin{figure}[ht]
\centering
\begin{tcolorbox}[width=.75\textwidth, nobeforeafter, coltitle = black, fonttitle=\fontfamily{lmss}\selectfont, title= Image Generation Tasks, halign title=flush center, colback=backg_blue!5, colframe=Pink!10, boxrule=2pt, grow to left by=-0.2mm, left=-15pt, right=-15pt, bottom=-1.5pt, top=-1pt]
    \centering
    {
        \includegraphics[scale=0.3]{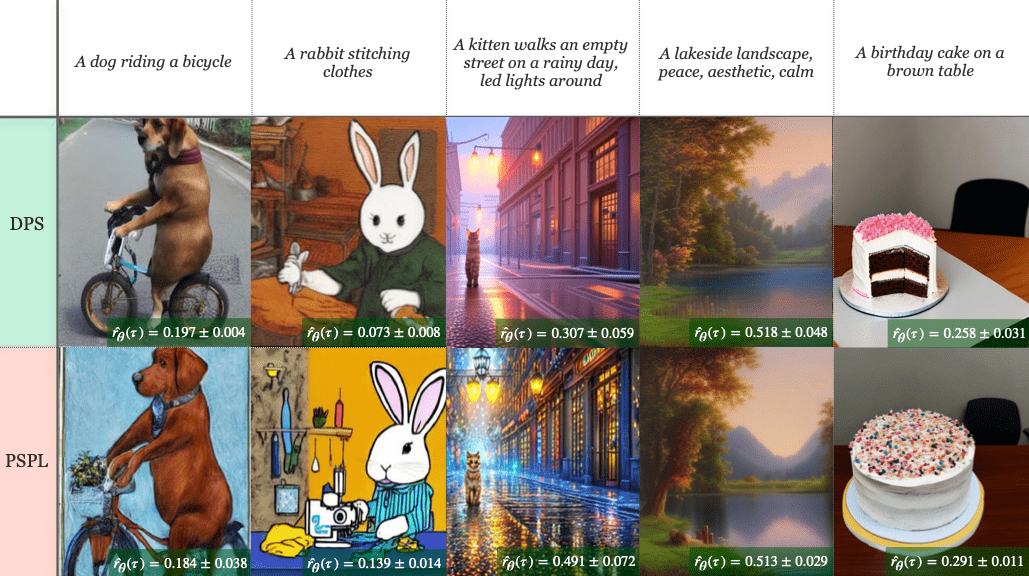}
    } 
\end{tcolorbox} 
\caption{\small Sample image generations with final image reward $\hat{r}_{\theta}(\cdot)$ over 5 independent runs. Images are enlarged for clarity.}
\label{fig:more_image_gen}
\end{figure}

\textbf{Training details.} Following \cite{zhang2024flow}, we first cluster the prompts in the dataset to obtain a mapping from $\text{cluster}_{j} \to (\Dcal_{0})_{j}$, where $j \in [J]$ is the cluster index out of J clusters, and $(\Dcal_{0})_{j}$ is the dataset of trajectories and corresponding preference labels for prompts in prompt $\text{cluster}_{j}$ i.e. $(\Dcal_{0})_{j} = { (\tau_{j,i}^{+},\tau_{j,i}^{-},y_{j,i}) }_{i=1}^{N_{j}}$, where $\tau_{j,\cdot}^{+}$ and $\tau_{j,\cdot}^{-}$ are the \emph{winning} and \emph{losing} trajectory generations given a prompt from $\text{cluster}_{j}$ for all $j \in [J]$. For tractability, we compress the training images to be 128$\times$128 pixels, and optimize for $K=100k$ episodes for each cluster $j \in [J]$. Future direction of this work will consider incorporating prompt information as a prior to the MLP, resulting in prompt conditioned inference. However, that is beyond the current scope of the paper. Finally, all experiments are run on NVIDIA GeForce RTX 5080, GPU 16GB, and Memory DDR5 64 GB. Training times for all algorithms are given in Table \ref{tab:training_times} and comprehensive validation results are shown in Figure \ref{fig:more_image_gen}.

\begin{table}[ht]
\caption{Average training times of baselines over 5 independent runs.}
\label{tab:training_times}
\centering
\small
\begin{tabular}{ccc}
\toprule
                  & DPS             & PSPL            \\ 
                  \midrule 
Training Time (h) & $3.52 \pm 0.11$ & $3.73 \pm 0.09$ \\
\bottomrule
\end{tabular}
\end{table}

\section{Auxiliary Definitions and Lemmas}

\begin{lemma}
\label{lem:binomial}
    Let $X$ be the sum of $L$ i.i.d. Bernoulli random variables with mean $p\in (0, 1)$. Let $q\in (0, 1)$, then 
    \begin{align*}
        \Pr(X \leq qL) &\leq \exp\left(-2L(q-p)^2\right),\qquad\text{if }q < p,\\
        \Pr(X \geq qL) &\leq \exp\left(-2L(q-p)^2\right),\qquad\text{if }q > p.
    \end{align*}
\end{lemma}

\begin{proof}
    Both inequalities can be obtained by applying Hoeffding's Inequality (see \ref{Hoeffding's inequality}).
\end{proof}

\begin{definition}
\textbf{ $\alpha$-dependence in \cite{russo2013eluder}}. For $\alpha>0$ and function class $\mathcal{Z}$ whose elements are with domain $\mathcal{X}$, an element $x \in \mathcal{X}$ is $\alpha$-dependent on the set $\mathcal{X}_n:=\left\{x_1, x_2, \cdots, x_n\right\} \subset \mathcal{X}$ with respect to $\mathcal{Z}$, if any pair of functions $z, z^{\prime} \in \mathcal{Z}$ with $\sqrt{\sum_{i=1}^n\left(z\left(x_i\right)-z^{\prime}\left(x_i\right)\right)^2} \leqslant$ $\alpha$ satisfies $z(x)-z^{\prime}(x) \leqslant \alpha$. Otherwise, $x$ is $\alpha$-independent on $\mathcal{X}_n$ if it does not satisfy the condition.
\end{definition}

\begin{definition}
\textbf{ Eluder dimension  in \cite{russo2013eluder}}. For $\alpha>0$ and function class $\mathcal{Z}$ whose elements are with domain $\mathcal{X}$, the Eluder dimension $\operatorname{dim}_E(\mathcal{Z}, \alpha)$, is defined as the length of the longest possible sequence of elements in $\mathcal{X}$ such that for some $\alpha^{\prime} \geqslant \alpha$, every element is $\alpha^{\prime}$ independent of its predecessors.
\end{definition}

\begin{definition}
    \textbf{Covering number.} Given two functions $l$ and $u$, the bracket $[l, u]$ is the set of all functions $f$ satisfying $l \leq f \leq u$. An $\alpha$-bracket is a bracket $[l, u]$ with $\|u-l\|<$ $\alpha$. The covering number $N_{[\cdot]}(\mathcal{F}, \alpha,\|\cdot\|)$ is the minimum number of $\alpha$-brackets needed to cover $\mathcal{F}$.
\end{definition}

\begin{lemma}
   \label{linear-dimension-covering}
   \textbf{(Linear Preference Models Eluder dimension and Covering number).} For the case of $d$-dimensional generalized trajectory linear feature models  $r_{\xi}\left(\xi_H\right):=\left\langle\phi\left(\xi_H\right), \mathbf{w}_r\right\rangle$, where $\phi:$ Traj  $\rightarrow \mathbb{R}^{dim_{\mathbb{T}}}$ is a known $dim_{\mathbb{T}}$ dimension feature map satisfying $\left\|\psi\left(\xi_H\right)\right\|_2 \leq B$ and $\theta \in \mathbb{R}^d$ is an unknown parameter with $\|\mathbf{w}_r\|_2 \leq \kappa_w$. Then the $\alpha$-Eluder dimension of $r_{\xi}(\xi_H)$ is at most $\mathcal{O}(dim_{\mathbb{T}} \log (B \kappa_w / \alpha))$. The $\alpha$ - covering number  is upper bounded by $\left(\frac{1+2 B\kappa_w}{\alpha}\right)^{dim_{\mathbb{T}}}$.
\end{lemma}

Let $\left(X_p, Y_p\right)_{p=1,2, \ldots}$ be a sequence of random elements, $X_p \in X$ for some measurable set $X$ and $Y_p \in \mathbb{R}$. Let $\mathcal{F}$ be a subset of the set of real-valued measurable functions with domain $X$. Let $\mathbb{F}=\left(\mathbb{F}_p\right)_{p=0,1, \cdots}$ be a filtration such that for all $p \geq 1,\left(X_1, Y_1, \cdots, X_{p-1}, Y_{p-1}, X_p\right)$ is $\mathbb{F}_{p-1}$ measurable and such that there exists some function $f_{\star} \in \mathcal{F}$ such that $\mathbb{E}\left[Y_p \mid \mathbb{F}_{p-1}\right]=f_*\left(X_p\right)$ holds for all $p \geq 1$. The (nonlinear) least square predictor given $\left(X_1, Y_1, \cdots, X_t, Y_t\right)$ is $\hat{f}_t=\operatorname{argmin}_{f \in \mathcal{F}} \sum_{p=1}^t\left(f\left(X_p\right)-Y_p\right)^2$. We say that $Z$ is conditionally $\kappa$-subgaussion given the $\sigma$-algebra $\mathbb{F}$ is for all $\lambda \in \mathbb{R}, \log \mathbb{E}[\exp (\lambda Z) \mid \mathbb{F}] \leq \frac{1}{2} \lambda^2 \kappa^2$. For $\alpha>0$, let $N_\alpha$ be the $\|\cdot\|_{\infty}$-covering number of $\mathcal{F}$ at scale $\alpha$. For $\beta > 0$, define
\begin{equation}
\mathcal{F}_t(\beta)=\left\{f \in \mathcal{F}: \sum_{p=1}^t\left(f\left(X_p\right)-\hat{f}_t\left(X_p\right)\right)^2 \leq \beta\right\} .
\end{equation}
\begin{lemma}
    \label{transition limit}
    (Theorem 5 of \cite{ayoub2020model}).
    Let $\mathbb{F}$ be the filtration defined above and assume that the functions in $\mathcal{F}$ are bounded by the positive constant $C>0$. Assume that for each $s \geq 1,\left(Y_p-f_*\left(X_p\right)\right)$ is conditionally $\sigma$-subgaussian given $\mathbb{F}_{p-1}$. Then, for any $\alpha>0$, with probability $1-\delta$, for all $t \geq 1, f_* \in \mathcal{F}_t\left(\beta_t(\delta, \alpha)\right)$, where
$$
\beta_t(\delta, \alpha)=8 \sigma^2 \log \left(2 N_\alpha / \delta\right)+4 t \alpha\left(C+\sqrt{\sigma^2 \log (4 t(t+1) / \delta)}\right) .
$$
\end{lemma}

\begin{lemma}

    (Lemma 5 of \cite{russo2013eluder}). Let $\mathcal{F} \in B_{\infty}(X, C)$ be a set of functions bounded by $C>0$, $\left(\mathcal{F}_t\right)_{t \geq 1}$ and $\left(x_t\right)_{t \geq 1}$ be sequences such that $\mathcal{F}_t \subset \mathcal{F}$ and $x_t \in \mathcal{X}$ hold for $t \geq 1$. Let $\left.\mathcal{F}\right|_{x_{1: t}}=\left\{\left(f\left(x_1\right), \ldots, f\left(x_t\right)\right): f \in \mathcal{F}\right\}\left(\subset \mathbb{R}^t\right)$ and for $S \subset \mathbb{R}^t$, let $\operatorname{diam}(S)=\sup _{u, v \in S}\|u-v\|_2$ be the diameter of $S$. Then, for any $T \geq 1$ and $\alpha>0$ it, holds that
$$
\sum_{t=1}^T \operatorname{diam}\left(\left.\mathcal{F}_t\right|_{x_t}\right) \leq \alpha+C(d \wedge T)+2 \delta_T \sqrt{d T},
$$
where $\delta_T=\max _{1 \leq t \leq T} \operatorname{diam}\left(\left.\mathcal{F}_t\right|_{x_{1: t}}\right)$ and $d=\operatorname{dim}_{\mathcal{E}}(\mathcal{F}, \alpha)$.
\label{upper bound}
\end{lemma}

\begin{lemma}
    If $\left(\beta_t \geq 0 \mid t \in \mathbb{N}\right)$ is a nondecreasing sequence and $\mathcal{F}_t:=\left\{f \in \mathcal{F}:\left\|f-\hat{f}_t^{L S}\right\|_{2, E_t} \leq \sqrt{\beta_t}\right\}$, where $\hat{f}_t^{L S} \in \arg \min _{f \in \mathcal{F}} L_{2, t}(f)$ and $L_{2, t}(f)=\sum_1^{t-1}\left(f\left(A_t\right)-R_t\right)^2$, then for all $T \in \mathbb{N}$ and $\epsilon>0$,
$$
\sum_{t=1}^T \mathbf{1}\left(w_{\mathcal{F}_t}\left(A_t\right)>\epsilon\right) \leq\left(\frac{4 \beta_T}{\epsilon^2}+1\right) \operatorname{dim}_E(\mathcal{F}, \epsilon)
$$
 where $w_{\mathcal{F}}(a):=\sup _{f \in \mathcal{F}} f(a)-\inf _{f \in \mathcal{F}} f(a)$ denotes confidence interval widths.
\end{lemma}

\begin{theorem}
\label{Hoeffding's inequality}
\textbf{Hoeffding's inequality}\citep{hoeffding1994probability}.
Let \(X_1, X_2, \ldots, X_n\) be independent random variables that are sub-Gaussian with parameter \( \sigma \). Define \( S_n = \sum_{i=1}^n X_i \). Then, for any \( t > 0 \), Hoeffding's inequality provides an upper bound on the tail probabilities of \( S_n \), which is given by:
\[
\Pr\left( |S_n - \mathbb{E}[S_n]| \geq t \right) \leq 2 \exp\left(-\frac{t^2}{2n\sigma^2}\right).
\]
This result emphasizes the robustness of the sum \( S_n \) against deviations from its expected value, particularly useful in applications requiring high confidence in estimations from independent sub-Gaussian observations.
\end{theorem}

\begin{lemma}
\label{visitation}
    (Lemma F.4. in \cite{dann2017unifying}) Let $\mathcal{F}_i$ for $i=1 \ldots$ be a filtration and $X_1, \ldots X_n$ be a sequence of Bernoulli random variables with $\mathbb{P}\left(X_i=1 \mid \mathcal{F}_{i-1}\right)=P_i$ with $P_i$ being $\mathcal{F}_{i-1}$-measurable and $X_i$ being $\mathcal{F}_i$ measurable. It holds that
$$
\mathbb{P}\left(\exists n: \sum_{t=1}^n X_t<\sum_{t=1}^n P_t / 2-W\right) \leq e^{-W}
$$
\end{lemma}

\chapter{Multi-Objective Alignment of Language Models}
\label{chapter:mopo}

\paragraph{Abstract}.
Post-training LLMs with RLHF and preference optimization methods (e.g., DPO, IPO) has greatly improved alignment, yet these approaches assume a single objective. In reality, humans express multiple, often conflicting objectives—such as helpfulness and harmlessness—with no natural scalarization. We study the multi-objective preference alignment problem, where a policy must balance several objectives simultaneously. We propose Multi-Objective Preference Optimization (\texttt{MOPO}), a constrained KL-regularized framework that maximizes a primary objective while enforcing lower bounds on secondary objectives via tunable safety thresholds. \texttt{MOPO} operates directly on pairwise preferences without point-wise rewards, and admits simple closed-form iterative updates. Empirically, \texttt{MOPO} recovers Pareto-optimal policies on synthetic benchmarks and, when fine-tuned on human-preference data, yields 1.3B-parameter models that achieve higher rewards and Pareto-dominate baselines, with stable and robust optimization dynamics.

\section{Introduction}
\label{sec:chapter5-introduction}

Aligning Large Language Models (LLMs) and other generative models with human preferences \cite{ouyang2022training, rafailov2023direct, ipo} has evolved from single-objective to multi-objective \cite{rame2023rewarded, yang2024rewards, zhong2024panacea}, aiming to comprehensively capture the inherent heterogeneity of human preferences. Multi-objective alignment jointly considers multiple human preference objectives, such as safety, helpfulness, factuality, and diversity, to optimize the LLM. However, human preferences are neither one dimensional nor fixed: a single conversation may demand an answer that is simultaneously helpful, harmless, concise, and imaginative. Optimizing an LLM for just one of these axes often degrades the others, as seen between helpfulness and safety tasks \cite{bai2022training, rame2023rewarded, yang2024rewards}. Consequently, we seek multi‑objective alignment, where the model is judged by reward vector performance on possibly conflicting objectives, rather than a single scalarized score. 

From an optimization standpoint this setting is naturally cast as multi‑objective optimization (MOO) \cite{miettinen1999nonlinear, keeney1993decisions}, where desirable solutions form a Pareto front. Existing alignment pipelines largely collapse this vector into a weighted sum and run reinforcement learning from human feedback (RLHF) on the resulting scalar reward \cite{ouyang2022training, rafailov2023direct}, but linear or non-linear scalarization hides important corner cases, and requires retraining for each weight choice. To overcome this, there is some work on decoding at inference time, but it still aims to cast user preferences as vector inputs to the model \cite{shi2024decoding, wang2024arithmetic}. However, at deployment time, users seldom articulate explicit weights; they expect models to adapt interactively and at low latency. These observations motivate algorithms that (i) approximate the Pareto front \emph{offline}, and (ii) provide inference without probing users for preference vector inputs. Motivated by these insights, we introduce \MOPO{MOPO}, an offline constrained‑optimization framework that unifies these desiderata and delivers multi-objective alignment with a single multi-head policy.

\textcolor{black}{
We begin by formulating the problem as a concave constrained optimization problem where preferences along the `primary' objective are maximized while preferences along the `secondary' objectives are constrained above a tunable threshold. We then motivate bounding the \emph{lower bound} of preferences (instead of the naively constraining these secondary objectives), followed by a behaviour cloning approach to extract the optimal policy from the resulting optimal importance sampling ratio. Overall, this procedure results in iterative updates of the underlying optimization variables, which is scalable and robust to the hyperparameters. Our contributions are as follows:
(i) We propose \MOPO{MOPO}, an offline constrained optimization based preference-only learning algorithm that optimizes for multiple objectives and achieves Pareto optimality provably.
(ii) We empirically validate the correctness on a variety of canonical preference dataset types that show how \MOPO{MOPO} approximates the Pareto front when it is known.
(iii) We conduct extensive LLM experiments on real world data to validate the effectiveness of \MOPO{MOPO}, and show optimization stability through ablation studies.
}

\textbf{Related works.} \rlhf \citep{christiano2017deep, ziegler2019fine} has become the de-facto paradigm for aligning LLMs such as GPT‑4 \citep{achiam2023gpt} and LLaMA‑3 \citep{grattafiori2024llama}. Most \rlhf pipelines fit a reward model to pairwise preferences and then fine‑tune the policy with PPO \citep{schulman2017proximal, ouyang2022training}. Instability and sample inefficiency have motivated alternatives that still target a scalar reference‑regularized objective, including RAFT \citep{zhang2024raft}, RRHF \citep{yuan2023rrhf}, DPO \citep{rafailov2023direct}, $\Psi$PO \cite{ipo}, and Nash‑\rlhf \citep{munos2023nash}. These methods (except $\Psi$PO) inherit a fundamental limitation: all preferences are collapsed into a single reward signal, obscuring trade‑offs between objective(s). Recent works attempt to optimize multiple objectives by learning scalarization functions or prompt contexts \citep{hu2023aligning,zhong2024panacea,guo2024controllable,wu2023fine,li2025self,lee2025calibrated}. Although effective in specific domains, such approaches seldom achieve Pareto-optimal solutions even when the Pareto front is known \cite{yang2024rewards, mukherjee2024multi}. While \cite{rame2023rewarded} mitigate tuning via `Rewarded Soups', and MORLHF \citep{li2020deep} and MODPO \citep{zhou2023beyond} borrow ideas from multi‑objective RL, they still learn with respect to a single functional combination of rewards. RiC \citep{yang2024rewards}, HaM \cite{mukherjee2024multi}, MOD \cite{shi2024decoding}, and DPA \cite{wang2024arithmetic} move beyond heuristic scalarization by conditioning on multiple rewards at inference time. While these algorithms improve controllability, they still rely on inference-time user preference input to optimize multiple objectives, which can misrepresent complex preference structures and are hard to quantify practically (for instance, what does "0.6 helpful, 0.4 safe" imply?). PARM \cite{lin2025parm} trains a \emph{single}, preference conditioned policy across multiple objectives, however, it still requires user preference inputs. A contemporary work to ours is L3Ms \cite{dhillonl3ms}, which uses log barrier functions to incorporate constraints into the primary alignment problem. Classical multi-objective RL (MORL) focuses on discovering Pareto‑efficient policies under vector rewards \citep{roijers2013survey,van2014multi,hayes2022practical}. Constrained MORL \citep{pmlr-v164-huang22a, liu2024c, pmlr-v235-agnihotri24a, NEURIPS2024_c81d09a6} methods maximize a primary objective while enforcing lower bounds on the other objectives, a strategy that inspires our formulation. However, constrained MORL assumes point‑wise rewards and cannot be applied directly to preference data. 


\section{Preliminaries}
\label{sec:chapter5-preliminaries}

\begin{wrapfigure}{r}{0.5\textwidth}
\vspace{-1.3cm}
\centering
\subfloat[\COP{COP} comparison with traditional approaches]{{
    \includegraphics[height=0.2\textwidth, width=0.23\textwidth]{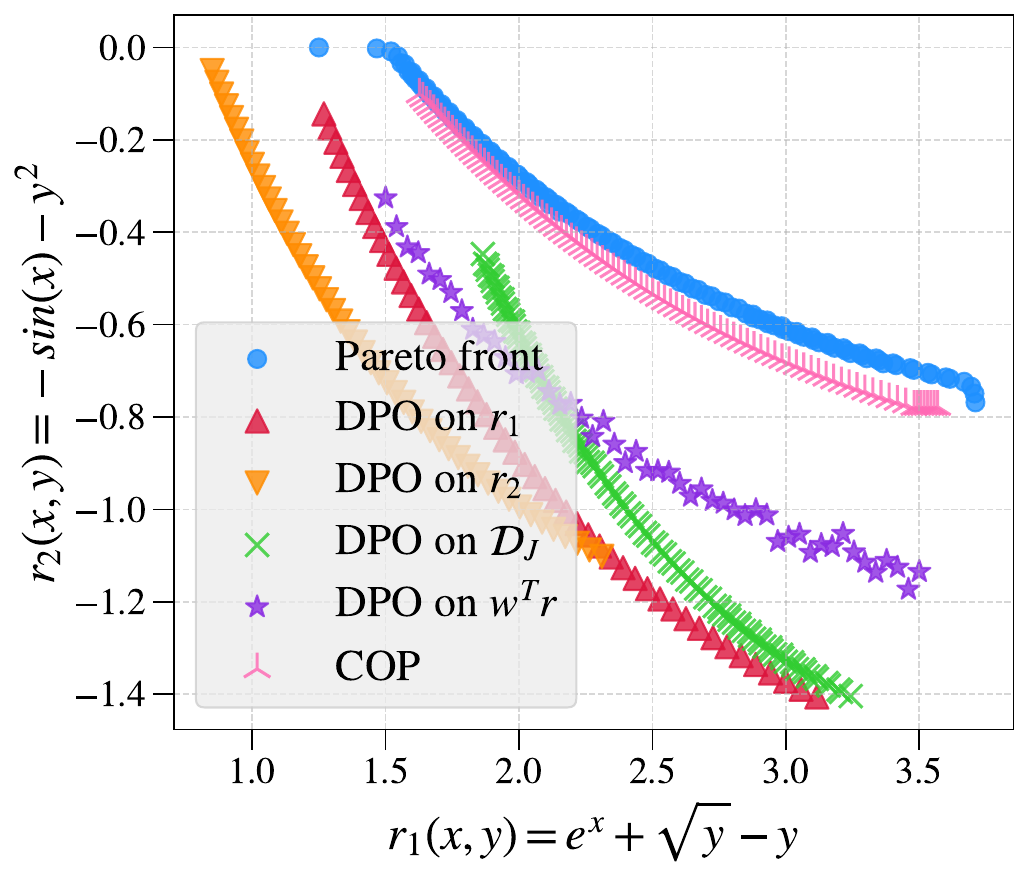}
}
{
    \includegraphics[height=0.2\textwidth, width=0.23\textwidth]{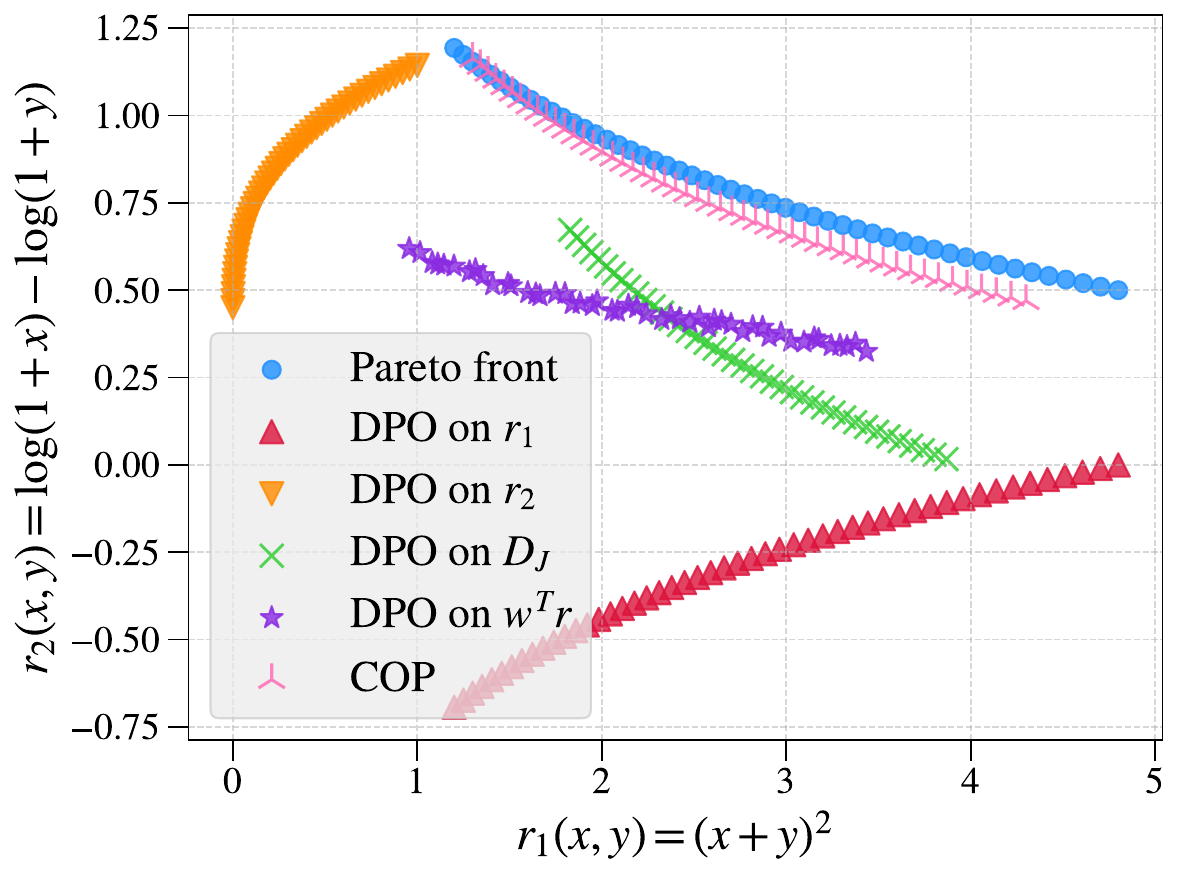}
} 
} \newline
\subfloat[\MOPO{MOPO} approximates the Pareto front]{
{
    \includegraphics[height=0.2\textwidth, width=0.23\textwidth]{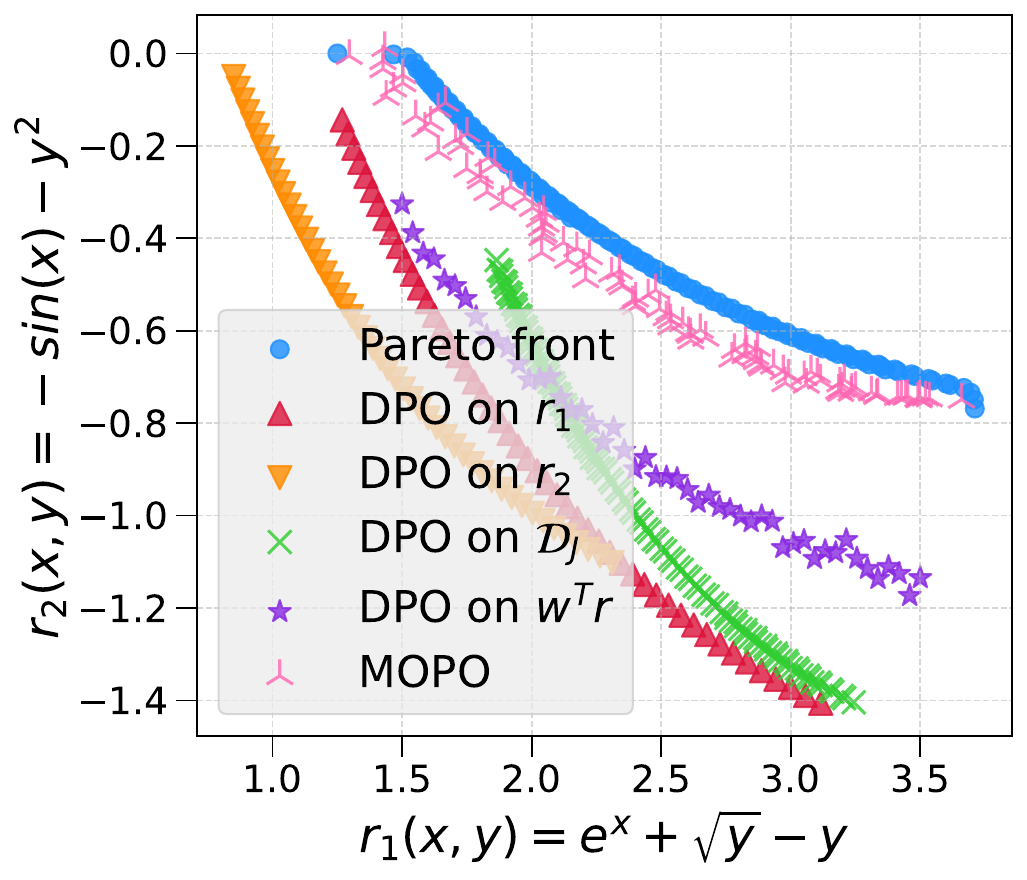}
}
{
    \includegraphics[height=0.2\textwidth, width=0.23\textwidth]{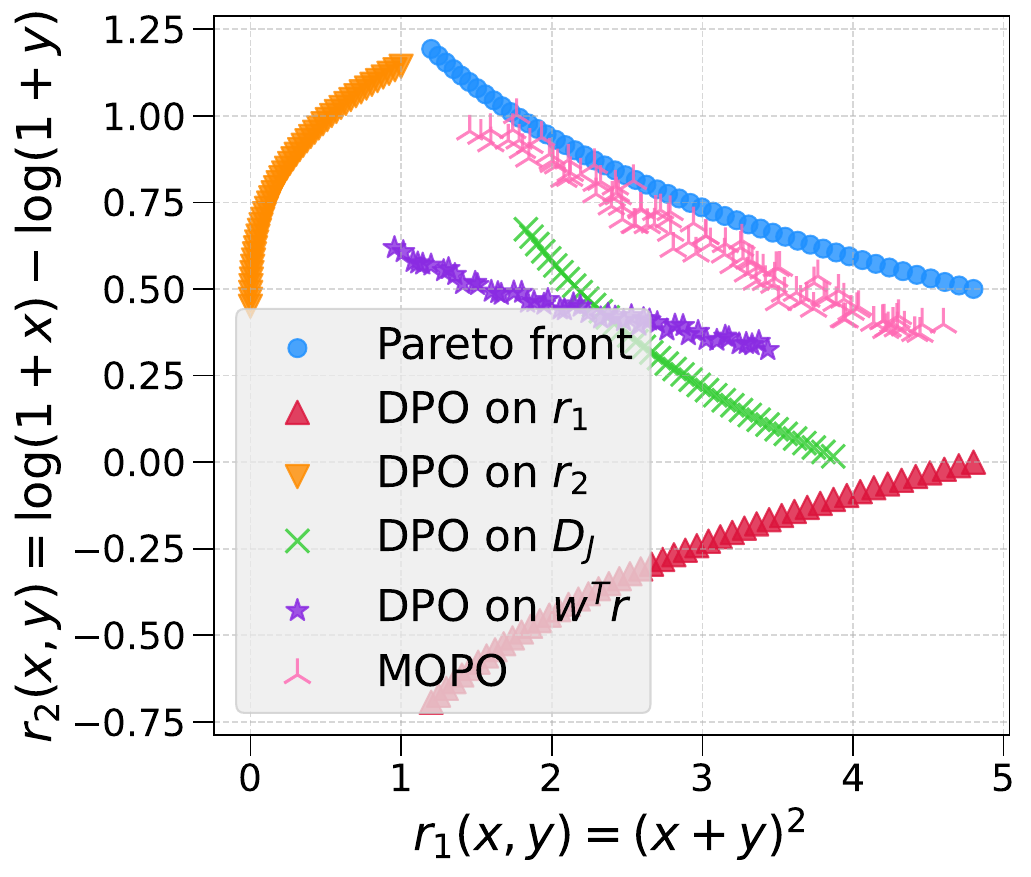}
}
}
\vspace{-0.1cm}
\caption{Illustration of how a \COP{COP} approach, and hence \MOPO{MOPO}, achieves Pareto-optimal alignment.}
\label{fig:dpo_comparison}
\vspace{-0.3cm}
\end{wrapfigure}

We first begin with a motivating example that inspires development of \MOPO{MOPO} as an offline constrained optimization algorithm learning from preference data. Following this, we describe the problem setting with a formal problem statement. Throughout the main text, we keep notation light and refer the reader to Appendix \ref{sec:appendix} for a complete discussion, where we prove that current literature as discussed above fails to achieve Pareto-optimality.

\textbf{A Motivating Example.} In this example, we empirically demonstrate the necessity of \emph{principled} multi-objective optimization methods that account for multi-dimensional preferences. We benchmark various approaches for multi-objective alignment and show that existing state-of-the-art techniques consistently fail to reach the Pareto front. To ensure clarity, we conduct experiments on synthetic datasets where true Pareto front is known, allowing for precise evaluation of alignment quality. Due to space constraints we keep discussion concise, and refer the reader to Appendix \ref{appendix:motivating_example} for completeness. 

Consider this toy preference example with input and action spaces $\Xcal=\Ycal=[0,1]$.  For any triplet $(x,y,y')$ we draw a preference label $z\in\{y,y'\}$ from the Bradley-Terry model $\text{BT}(r)$ with $\Pr[z=y \given x, y , y'] = \exp(r(x,y)) / (\exp(r(x,y))+\exp(r(x,y')))$, where $r(\cdot)$ is the underlying reward model \citep{bradleyterry}. We study two bi‑objective settings: \emph{Set A} with $r_{1} = e^{x} + \sqrt{y} - y$ and $r_{2} = -\sin x - y^{2}$, and \emph{Set B} with $r_{1} = (x+y)^2$ and $r_{2}=\log ((1+x)/(1+y))$. From i.i.d. samples $(x,y,y') \sim \Ucal([0,1]^3)$ we construct four datasets: (i) \Done{$\Dcal_{1}$}, labeled by $r_1$ only; (ii) \Dtwo{$\Dcal_{2}$}, by $r_2$ only; (iii) \Djoint{$\Dcal_{J}$}, the \textbf{j}oint dataset retaining samples where the two labels coincide; and (iv) \Dcombined{$\Dcal_{C}$}, the \textbf{c}ombined dataset labeled by $\text{BT}(w r_{1} + (1-w) r_{2})$ for $w \in [0,1]$. As seen in Figure \ref{fig:dpo_comparison}(a), policies trained with DPO on \Done{$\Dcal_{1}$} or \Dtwo{$\Dcal_{2}$} alone ignore one objective, those trained on \Djoint{$\Dcal_{J}$} see only non‑conflicting pairs, and those trained on \Dcombined{$\Dcal_{C}$} are biased toward a single scalarization of reward functions, so all three miss large portions of the Pareto front.  A constrained optimization baseline $\pi_{\text{\COP{\small COP}}}(x) = \argmax_{y} r_1(x,y) \suchthat  r_2(x,y) \geq b$ for some $b \in \Rbb$ approaches the Pareto front. These limitations motivate an algorithm that optimizes all objectives \emph{jointly}. Hence, we develop \MOPO{MOPO} as an offline constrained optimization algorithm that recovers policies which lie near the true Pareto front as in Figure \ref{fig:dpo_comparison}(b). Given this motivation, we now turn our attention to introducing notations and providing a formal problem statement.

\textbf{Problem Setting.} We define a finite set of contexts $\mathcal{X}$ and a finite action space $\mathcal{Y}$. A policy $\pi \in \Delta_{\mathcal{Y}}^{\mathcal{X}}$ defines a probability distribution over actions given a context, where $\Delta_{\mathcal{Y}}$ is the probability simplex over $\mathcal{Y}$. The policy is learned from human preferences, which are provided in a pairwise manner over actions. For each context $x \in \mathcal{X}$, two actions $y, y' \sim \mu(\cdot  \given  x)$ are sampled from a behavior policy $\mu$, and a human annotator provides a preference signal indicating which action is preferred. We also let the contexts $x$ be sampled from a context distribution $\nu$, denote a vector by $\bm{v}$, let $\bm{v}_{j}$ to be the element at the $j^{th}$ dimension of $\bm{v}$, and let $[N]$ denote the set $\{1,\dots,N\}$ for some $N \in \Nbb$.

Typically, in single objective preference optimization, the preference for one generation  over another is denoted as $y_w\succ y_l$, where $y_w$ and $y_l$ denote the preferred and dis-preferred actions amongst $\{ y, y' \}$ respectively. This true human preference takes the form $p(y \succ y' \given x)$, the probability of $y$ being preferred to $y'$ knowing the context $x$. In our multi-objective preference setting, we extend this notation to $K$ objectives, wherein $p_{k}(y \succ y' \given x)$ denotes the preferred and dis-preferred actions amongst $\{ y, y' \}$ for $k^{th}$ objective with $k \in [K]$. Moreover, we also set the expected preference of a generation $y$ \emph{over} a distribution $\mu$ knowing $x$ for the $k^{th}$ objective as $p_{k}(y \succ \mu \given x) = \Ebb_{y' \sim \mu( \cdot \given x)} \left[p_{k}(y \succ y' \given x) \right]$. We also let for any two policies  $\pi, \mu \in \Delta_\mathcal{Y}^\mathcal{X}$ and a context distribution $\nu$ the total preference of policy $\pi$ to $\mu$ w.r.t. $k^{th}$ objective as $ p_{k}^{\nu}(\pi \succ \mu) = \Ebb_{x\sim\nu , y \sim \pi(.|x)}[p_{k}(y \succ \mu \given x)]$. Without loss of generality and clarity of notation, we let $p_{K}(y \succ y' \given x) \equiv p(y \succ y' \given x)$ denote the preference for the $K^{th}$ (the primary) objective, and $\bm{q}(y \succ y' \given x) \in [0,1]^{K-1}$ denotes the vector of preferences for the $K-1$ (the secondary) remaining objectives, wherein the preferences are applied objective-wise i.e. $\bm{q}_{k}(y \succ y' \given x) = p_{k}(y \succ y' \given x)$ for $k \in [K-1]$. Following this notation, we also have the following definitions.

\small
\begin{equation*}
\begin{aligned}
    \bm{q}(y \succ \mu \given x) &= \E{y' \sim \mu( \cdot \given x)}[\bm{q}(y \succ y' \given x)] \quad \text{and} \quad \bm{q}^{\nu}(\pi \succ \mu) &= \E{x \sim \nu \\ y \sim \pi(\cdot \given x)}[\bm{q}(y \succ \mu \given x)] \, .
\end{aligned}
\end{equation*}
\normalsize

\textbf{Pareto optimality.} In multi-objective preference optimization (\MOPO{MOPO}), a policy that simultaneously optimizes all objectives does not exist. Thus, a set of non-dominated solutions is desired. We say policy $\pi$ is dominated by policy $\pi'$ when there is no objective under which $\pi'$ is \emph{worse} than $\pi$, i.e., $p_{k}^{\nu}(\pi \succ \mu \given x) \leq p_{k}^{\nu}(\pi' \succ \mu \given x)$ for $\forAll k \in [K]$. A policy $\pi$ is Pareto-optimal if and only if it is not dominated by any other policy. The Pareto set is composed of non-dominated solutions, denoted as $\paretoset$. Overall, the goal of \MOPO{MOPO} is to obtain an optimal policy in $\paretoset$.

\textbf{Problem statement.} The goal is to propose a general solution for \rlhf with multiple objectives, based on constrained optimization of a function of preferences. We propose this constrained optimization problem as maximizing a primary objective, and constraining the remaining objective values. To this end, we consider a reference policy $\piref \in \Delta^{\mathcal{X}}_{\mathcal{Y}}$, a real positive regularization parameter $\tau \in \mathbb{R}_{+}$, and let $\bm{b} \in [0,1]^{K-1}$. The concave constrained optimization problem (\COP{\small COP}) for \MOPO{MOPO} becomes,

\begin{align}
\label{eq:first-obj}
    \max_{\pi} \, \E{x \sim \nu \\ y \sim \pi( \cdot \given x) \\ y' \sim \mu( \cdot \given x)}[p(y \succ y'  \given  x)] - \tau \KL(\pi \;  || \; \piref) \,  \suchthat \E{x \sim \nu \\ y \sim \pi( \cdot \given x) \\ y' \sim \mu( \cdot \given x)}[\bm{q}(y \succ y' \given  x)] \geq \bm{b}.
\end{align}

See Definition \ref{def:kl} for the definition of $\KL$ divergence. We now focus our attention on designing a  \MOPO{MOPO} algorithm to solve the \COP{\small COP} problem above.

\section{The MOPO Algorithm}
\label{sec:algorithm}

\begin{wrapfigure}{r}{0.3\textwidth}
\centering
{
    \includegraphics[height=0.17\textwidth, width=0.3\textwidth]{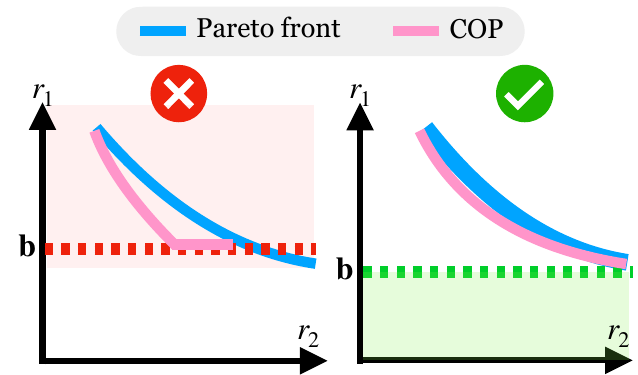}
}
\vspace{-0.7cm}
\caption{Illustration of how constraint threshold initialization affects \COP{\small COP} solutions.}
\vspace{-0.4cm}
\label{fig:correctb}
\end{wrapfigure}

First, to find an optimal policy that lies in $\paretoset$, it is crucial to set proper constraint values $\bm{b}$ such that the solution of Problem \eqref{eq:first-obj} contributes to the Pareto front. See Appendix \ref{sec:suboptimalbaselines} and \ref{appendix:constraintspec} for the theoretical discussion, and Figure \ref{fig:correctb}, which illustrates that if $\bm{b}$ is correctly initialized, then solving the \COP{\small COP} problem yields solutions on the Pareto-front. In Section \ref{sec:function_approx} we will propose a more practical method of specifying constraint values. For now, we focus on solving Problem \eqref{eq:first-obj}. To deal with the optimization variable in expectation, we let the importance sampling ratio be $\rho(y) = \frac{\pi(y)}{\piref(y)}$. For this we assume that \texttt{Supp}$(\pi)$ = \texttt{Supp}($\piref$). In addition, for clarity, we shall omit the dependency on context $x$ as the all results hold true for all $x \in \texttt{Supp}(\nu)$. Then the final \MOPO{MOPO} problem takes the form,

\small
\begin{align}
\label{eq:main-obj}
    \max_{\rho} \underbrace{\E{y \sim \piref \\ y' \sim \mu}[ \rho(y) \, p(y \succ y')] - \tau \E{y \sim \piref}[ \rho(y) \, \ln(\rho(y))]}_{\Fcal(\rho)} \; \suchthat \underbrace{\E{y \sim \piref \\ y' \sim \mu}[ \rho(y) \, \bm{q}(y \succ y')]}_{\bm{\Gcal}(\rho)} \geq \bm{b} \, ,
\end{align}
\normalsize

which is a strictly concave optimization problem w.r.t. $\rho$. We then formulate the Lagrangian of the above \MOPO{MOPO} problem. For some $\bm{\lambda} := \{\lambda_{k}\}_{k=1}^{K-1} \geq \bm{0}$, we have the Lagrangian as $\Lcal(\rho, \bm{\lambda}) = \Fcal(\rho) - \bm{\lambda}^{T} \left(\bm{b} - \bm{\Gcal}(\rho) \right)$. This leads to the following proposition.

\begin{restatable}{proposition}{maindual}
\label{prop:main_dual}
The dual formulation of Problem \eqref{eq:main-obj} is given by,
\small
\begin{align}
\label{eq:dual}
    \text{\dual} & \triangleq \min_{\bm{\lambda} \geq \bm{0}} \max_{\rho} \; \Lcal(\rho, \bm{\lambda}) = \min_{\bm{\lambda} \geq \bm{0}} \Lcal(\rho^{\star}_{\bm{\lambda}}, \bm{\lambda}) = \min_{\bm{\lambda} \geq 0} \Fcal(\rho_{\bm{\lambda}}^{\star}) - \bm{\lambda}^{T}(\bm{b} - \bm{\Gcal}(\rho_{\bm{\lambda}}^{\star})) \nonumber \\
    \text{where}, & \;\, \rho^{\star}_{\bm{\lambda}}(y) = \exp\left( \tau^{-1} \E{y' \sim \mu}[p(y \succ y') + \bm{\lambda}^{T} \bm{q}(y \succ y')]  - 1 \right)  \forAll y \in \Ycal.
\end{align}
\normalsize
\end{restatable}

See Appendix \ref{proof:main_dual} for proof. The inner maximization in Equation \eqref{eq:dual} corresponds to computing an optimal policy (importance sampling ratio $\rho$) that maximizes scalarized preferences for the $K^{th}$ objective, while the outer minimization corresponds to balancing the penalty of suboptimal policy w.r.t. the other $(K-1)$ objectives: if the current policy ($\rho$) is under performing w.r.t. the $k^{th}$ objective, ${\lambda}_{k}$ increases so that the under performance is penalized more, and vice versa.

\begin{remark}
\label{remark:logbarrier}
    Formulation in Problem \eqref{eq:dual} also connects to the use of barrier functions in optimization literature. For some $\sigma, s > 0$, consider the following log barrier function for all $z \in \Rbb$,
    \small
    \begin{equation*}
    \label{eq:logbarrierdef}
    \Bcal_{\sigma,s}(z) = \begin{cases}
        - \sigma \log(-z) \, , & z \leq -s \\
        \frac{\sigma}{s} z + (1-\log(s))\sigma \, , & z > -s
    \end{cases} \;\; , \; \;  \text{with}  \; \; \partial_{z}  \Bcal_{\sigma,s}(z) = \frac{\sigma}{\max(-z,s)} \, .
    \end{equation*}
    \normalsize
    This relaxed log-barrier function can be used to construct an unconstrained Lagrangian as,
    \small
    \begin{equation}
    \label{eq:logbarrier}
        \Lcal_{\text{LB}}(\rho, \bm{\sigma}) = \Fcal(\rho) - \sum_{k=1}^{K-1} \Bcal_{\bm{\sigma}_{k},\bm{\sigma}_{k}^{2}}(\bm{b}_{k} - \bm{\Gcal}_{k}(\rho)) \quad , \; \text{with} \quad \bm{\sigma} = \{\bm{\sigma}_{k}\}_{k=1}^{K-1} > \bm{0} \, .
    \end{equation}
    \normalsize
    See Appendix \ref{appendix:logbarrier} for more details. Although comparable theoretically, we will see in Section \ref{sec:experiments} how formulation of Problem \eqref{eq:dual} is empirically superior to that of Problem \eqref{eq:logbarrier}.
\end{remark}

\begin{wrapfigure}{r}{0.5\textwidth}
\vspace{-0.1cm}
\centering
{
    \includegraphics[height=0.25\textwidth, width=0.35\textwidth]{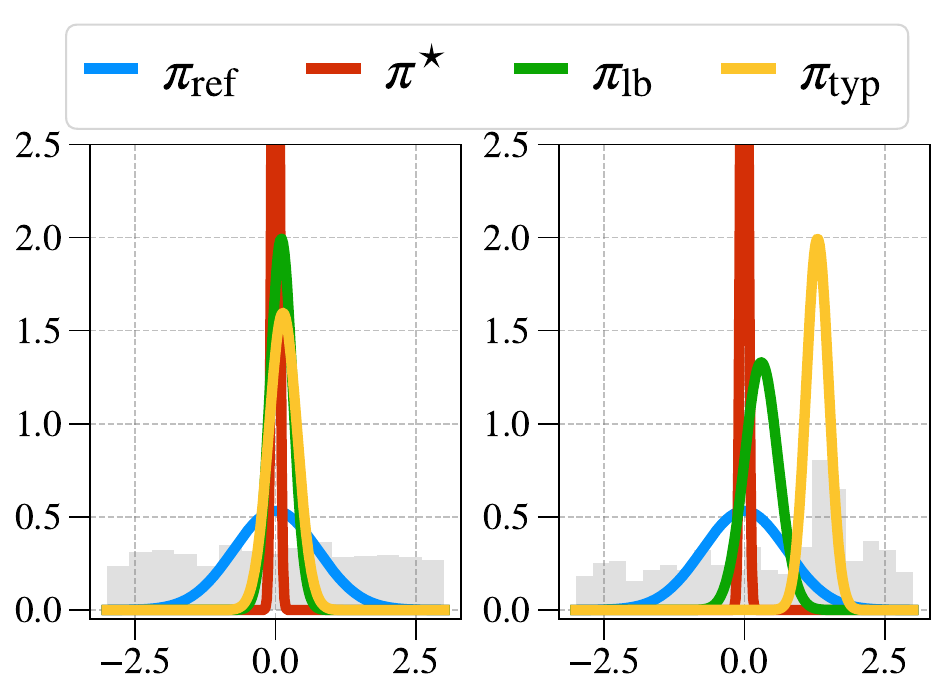}
}
\caption{\MOPO{MOPO}: Impact of preference distribution (gray) over output space with lower-magnitude outputs being preferred with probability 1. $\pi_{\text{lb}}$ constrains lower bound of $\bm{\Gcal}(\rho)$, while $\pi_{\text{typ}}$ constrains $\bm{\Gcal}(\rho)$ directly (typically).}
\label{fig:lowerbound}
\end{wrapfigure}

Returning to our discussion of Problem \eqref{eq:dual}, we find that constraining the preference vector $\bm{q}(\cdot)$ naively can result in constraint violation when deployed to the real environment. This is due to the fact that empirical importance sampling weighted preferences $\hat{\bm{q}}(\cdot)$ collected from a finite dataset inevitably have estimation error (see Figure \ref{fig:lowerbound}). For the $K^{th}$ objective, preference estimation error may be tolerated as long as those estimates are useful as policy improvement signals, i.e., it is sufficient to maintain the relative order of preferences. For the remaining $(K-1)$ constrained objectives, Equation \eqref{eq:dual} instead relies on the estimated values directly. Hence, to make a policy robust against these estimation errors, we consider a constrained policy optimization scheme that instead constrains the \emph{lower bound} of the preference estimates $\hat{\bm{q}}(\cdot)$,  i.e.,
\small
\begin{align*}
\max_{\rho} \Fcal(\rho) \; \suchthat \; \texttt{LowerBound}(\bm{\Gcal}(\rho)) \geq \bm{b} \, .
\end{align*}
\normalsize
\vspace{-0.2cm}

Then, the key question is how to estimate the lower bound of the $\rho$-weighted preference vector $\bm{q}(\cdot)$. One natural way is to exploit bootstrap confidence interval estimation \cite{efron1994introduction}, by sampling bootstrap datasets $\Dcal_{i}$ from $\Dcal$ and constructing population statistics for confidence interval estimation. However, this procedure
is computationally expensive. Instead, we take a different, computationally efficient approach. Specifically, we solve a constrained optimization problem for each objective $k \in [K-1]$. For a policy $\pi_{k} \in \Delta_{\Ycal}^{\Xcal}$ and some $\epsilon \in \Rbb_{+}$, the lower bound optimization problem of $\bm{\Gcal}(\rho)$ becomes:

\vspace{-0.3cm}
\begin{align}
\label{eq:lbop}
\min_{\pi_{k}} \E{y \sim \pi_{k} \\ y' \sim \mu}[\rho(y) \bm{q}_{k}(y \succ y')] \; \suchthat \; \KL(\pi_{k} \;  || \;  \piref) \leq \epsilon \quad \text{and}\, , \quad \sum_{y \in \Ycal} \pi_{k}(y) = 1
\end{align}
\vspace{-0.4cm}

In essence, we want to adversarially optimize a distribution $\pi_{k}$ so that it underestimates the preference objective $k$, and simultaneously, we enforce that this $\pi_{k}$ should not be perturbed too much from $\piref$. As $\epsilon$ increases, the degree of underestimation of preference probabilities also increases. Now we simplify the constrained optimization problem into a single unconstrained problem as follows. 

\begin{restatable}{proposition}{lowerbounddual}
\label{prop:lowerbound_dual}
The optimal solution to Problem \eqref{eq:lbop} can be obtained by solving the following optimization problem.
\small
\begin{align}
\label{eq:lbop_lagrangian}
\chi_{k}^{\star} & = \argmax_{\chi_{k} \geq 0} \; \Lcal_{k}(\chi_{k} \, ; \, \rho) := - \chi_{k} \ln \left( \Ebb_{y \sim \piref , y' \sim \mu} \left[ 
 \exp \left( \chi_{k}^{-1} \rho(y) \bm{q}_{k}(y \succ y')  \right) \right] \right) - \chi_{k} \epsilon \nonumber \\ 
& \text{with}, \;\, \pi_{k}^{\star}(y) \propto \piref(y) \underbrace{\exp \left(  (\chi_{k}^{\star})^{-1} \E{y' \sim \mu}[\rho(y) \bm{q}_{k}(y \succ y')] \right)}_{w(y) \; (\text{unnormalized weight})}.
\end{align}
\normalsize
\end{restatable}
\vspace{-0.4cm}

See Appendix \ref{proof:lowerbound_dual} for proof. Note that each term in Equation \eqref{eq:lbop_lagrangian} can be estimated using samples from the offline dataset $\Dcal$, thus it can be optimized in a fully offline manner. This procedure can be understood as computing the weights for each sample while adopting reweighting, that is, $\texttt{LowerBound}(\bm{\Gcal}(\rho)) = \Ebb_{y \sim \piref, y' \sim \mu}[\tilde{w}(y) \rho(y) \bm{q}_{k}(y \succ y')]$, where $\tilde{w}(y)$ is normalized $w(y)$. Solving this unconstrained optimization problem and plugging it in the main \dual $\,$ Problem \eqref{eq:dual} corresponds to the following iterative updates. 

\vspace{-0.35cm}
\small
\begin{align}
\label{eq:theoretical_chi_lambda_update}
    \chi^{\star} \leftarrow \argmax_{\chi \geq 0} \sum_{k=1}^{K-1} \Lcal_{k}(\chi_{k} ; \rho_{\bm{\lambda}}) \quad \text{and} \, , \quad \bm{\lambda}^{\star} \leftarrow \argmin_{\bm{\lambda} \geq 0} \; \Fcal(\rho_{\bm{\lambda}}) -  \bm{\lambda}^{T} ( \bm{b} - \underbrace{\bm{\Lcal}(\chi^{\star} ; \rho_{\bm{\lambda}})}_{\texttt{LowerBound}(\bm{\Gcal}(\rho))} ),
\end{align}
\normalsize
\vspace{-0.2cm}

where $\bm{\Lcal}(\chi^{\star} ; \rho_{\bm{\lambda}}) = (\Lcal_{1}(\chi^{\star}, \rho_{\bm{\lambda}}), \dots, \Lcal_{K-1}(\chi^{\star}, \rho_{\bm{\lambda}}))^{T}$. Compared to the original dual Problem \eqref{eq:dual}, the additional maximization for $\chi$ is introduced to estimate the lower bound of preference probabilities for the constrained objectives. 
Once the optimal solution $\bm{\lambda}^{\star}$ is computed, $\rho_{\bm{\lambda}^{\star}}^{\star}(y) \equiv \rho^{\star}(y) = \pi^{\star}(y) / \piref(y)$ is also derived from Equation \eqref{eq:dual}.

\textbf{Policy Extraction.}  The current procedure estimates the importance sampling ratio $\rho^{\star}(y)$ of the optimal policy, rather than directly obtaining the policy itself. Since the importance sampling ratio does not provide a direct way to sample an action, we need to extract the optimal policy $\pi^{\star}$ from $\rho^{\star}$ in order to select actions when deployed. For tabular cases, it is straightforward to obtain $\pi^{\star}(y) = \left( \piref(y) \rho^{\star}(y) \right) / \left(\sum_{y' \in \Ycal} \piref(y') \rho^{\star}(y') \right)$. However, the same method cannot directly be applied to large scale optimization problems due to the intractability of computing the normalization constant. For such cases, we instead extract the policy using importance-weighted behavioral cloning by solving the following problem: 

\vspace{-0.3cm}
\small
\begin{align}
\label{eq:policy-extraction}
    \max_{\pi} \E{y \sim \pi^{\star}}[ \log(\pi(y))] = \max_{\pi} \E{y \sim \piref}[ \rho^{\star}(y) \log(\pi(y))],
\end{align}
\normalsize
\vspace{-0.5cm}

which maximizes the log-likelihood of actions to be selected by the optimal policy $\pi^{\star}$. Based on the above discussion, we now turn our attention to the practical implementation of \MOPO{MOPO}.

\subsection{Practical Algorithm with Function Approximation}
\label{sec:function_approx}

For this section we consider the practical implementation of \MOPO{MOPO}, using a given offline dataset of preferences $\Dcal$. First, we discuss the function approximations used to parameterize the optimization variables, and then we discuss how to implement the procedure discussed above.

\textbf{Function Approximations.}  We let the optimization variables $\bm{\lambda}, \bm{\chi} \in \Rbb_{+}^{K-1}$, and the policy $\pi_{\psi}$ to be parameterized by $\psi$. The parameter $\bm{\chi} = \{\chi_{k}\}_{k=1}^{K-1}$ is trained by minimizing the following loss:

\vspace{-0.3cm}
\begin{align}
\min_{\bm{\chi} \geq \bm{0}} \;  \sum_{k=1}^{K-1} \left[ \chi_{k} \ln \left( \Ebb_{y \sim \piref, y' \sim \mu} \left[ 
 \exp \left(  \chi_{k}^{-1} \rho(y) \bm{q}_{k}(y \succ y')  \right) \right] \right) + \chi_{k} \epsilon \right] \; \nonumber.
\end{align}
\normalsize
\vspace{-0.4cm}

Since this involves a logarithm outside of the expectation, to overcome bias we use mini-batch approximation for computational efficiency. The empirical form is then given by:

\vspace{-0.3cm}
\small
\begin{align}
\min_{\bm{\chi} \geq \bm{0}} J(\bm{\chi} \,; \rho) := \E{\text{batch}(\Dcal) \sim \Dcal} \left[ \sum_{k=1}^{K-1} \left[ \chi_{k} \ln \left( \E{y, y' \sim \text{batch}(\Dcal)} \left[ 
 \exp \left(  \chi_{k}^{-1} \rho(y) \bm{q}_{k}(y \succ y')  \right) \right] \right) + \chi_{k} \epsilon \right] \right]  \, . 
\label{eq:chi_update}
\end{align}
\normalsize
\vspace{-0.3cm}

Finally, following the discussion before, $\bm{\lambda}$ and the policy parameterized by $\psi$ are optimized by:

\vspace{-0.3cm}
\small
\begin{align}
\label{eq:lambda_policy_update}
    \min_{\bm{\lambda} \geq \bm{0}} J(\bm{\lambda} \,; \bm{\chi}) := \Fcal(\rho)  - \bm{\lambda}^{T}( \bm{b} - J(\bm{\chi} \,; \rho)) \quad \text{and}, \quad 
    \min_{\psi} J_{\rho}(\pi_{\psi}) := - \E{y \sim \piref}[\rho(y) \log(\pi_{\psi}(y))],
\end{align}
\normalsize
where all variables are optimized jointly. For the empirical derivation of \MOPO{MOPO} given a \emph{fixed} offline dataset of preferences, please see Appendix \ref{appendix:samplingbasedcop}, where we discuss its practical implementation. However, two caveats still remain. 

\textbf{Lagged reference policy.}  The $\KL$ regularizer in Equation \eqref{eq:main-obj} keeps $\pi_{\psi}$ close to a fixed reference $\piref$.  
Because successive iterates move toward the Pareto front, it is advantageous to regularize against a \emph{stronger} policy than the initial prior.
Analogous to target networks in Q‑learning \cite{mnih2013playing} and recent self‑improvement loops for LLMs \cite{chen2024self, pang2024iterative}, we update the reference every $t_0$ steps:
$\piref \leftarrow \pi_{\psi}^{(t-t_0)}$. All expectations in Equation \eqref{eq:empirical-cop} are then reweighted by the ratio
$\rho_{\text{lag,ref}}(y) = \pi_{\psi}^{(t-t_0)}(y) / \pi_{\mathrm{ref}}(y)$ for all $y \in \Ycal$, requiring no additional data collection. We find that this leads to more stable optimization and consistent progression to the Pareto front.

\textbf{Adaptive constraint schedule.} In practice, exact values of constraint thresholds $\bm{b}$ are unknown. We therefore, after every $t_{0}$ steps, set the constraint vector \emph{only} from the policy of the previously optimized iterates: $ \bm{b} = \bm{\beta}^{\top} \bm{\mathcal G} \left(\rho^{(t-t_{0})} \right)$ for some hyperparameter $\bm{\beta} \in (0,1)^{K-1}$ and $\rho^{(t-t_{0})}(\cdot) = \pi_{\psi}^{(t-t_{0})}(\cdot) / \piref(\cdot)$ is the importance sampling ratio. This retains the theoretical lower bound interpretation of the constraints while avoiding a global search across constraint thresholds.

Summarizing the above discussion gives the final
\emph{Multi‑Objective Preference Optimization} (\MOPO{MOPO}) algorithm, shown in
Algorithm \ref{alg:mopo}. At each step we maximize the primary preference objective subject to the time‑varying lower bounds $\bm{b}$, while penalizing divergence from the current reference policy. The result is a scalable, offline algorithm that steadily advances toward Pareto‑optimal solutions.

\begin{figure}[t]
\begin{algorithm}[H]
\caption{Multi-objective Preference Optimization (\MOPO{MOPO})}
\begin{algorithmic}[1]
   \STATE {\bfseries Input:} Dataset $\Dcal$, batch size $M$, learning rate $\eta$, epochs $T$, lag $t_{0}$, relaxation parameter $\bm{\beta}$.
   \STATE Initialize parameter vectors $\bm{\lambda}^{(0)}, \bm{\chi}^{(0)}, \psi^{(0)}$, $\bm{b}$ and $\rho_{\bm{\lambda}^{(0)}}(\cdot)$.
    \FOR{$t = 1,2, \dots ,T$} 
      \STATE Sample $M$ mini-batches from $\Dcal$.
      \STATE $\bm{\chi}^{(t)} = [ \bm{\chi}^{(t-1)} - \eta \nabla_{\bm{\chi}} J(\bm{\chi} \,; \rho_{\bm{\lambda}^{(t-1)}})  ]_{+}$ using Equation \eqref{eq:chi_update}. {\small \texttt{\textcolor{brightblue!80!black}{\hphantom{..} // Lower bound estimation}}}
      \STATE $\bm{\lambda}^{(t)} = [ \bm{\lambda}^{(t-1)} - \eta \nabla_{\bm{\lambda}} J(\bm{\lambda} \, ; \bm{\chi}^{(t)})]_{+}$ using Equation \eqref{eq:lambda_policy_update}. {\small \texttt{ \textcolor{brightblue!80!black}{ //  Solve original dual}}}
      \STATE Compute $\rho_{\bm{\lambda}^{(t)}}(\cdot)$ using Equation \eqref{eq:dual}. {\small \texttt{ \textcolor{brightblue!80!black}{ \hphantom{..} // Retrieve primal solution}}}
      \STATE Update policy $\psi^{(t)} = \psi^{(t-1)} - \eta \nabla_{\psi} J_{\rho_{\bm{\lambda}^{(t)}}}(\pi_{\psi})$ using Equation \eqref{eq:lambda_policy_update}. {\small \texttt{\textcolor{brightblue!80!black}{ \hphantom{..} // Extract policy}}}
      \IF{$t \; \texttt{mod} \; t_{0}$ = 0}
      \STATE $\bm{b} = \bm{\beta}^{\top} \bm{\mathcal G} \left(\rho^{(t-t_{0})} \right)$ {\small \texttt{ \textcolor{brightblue!80!black}{ \hspace{3.5cm} // Update constraint thresholds}}}
      \STATE $\piref \leftarrow \pi_{\psi}^{(t-t_{0})}$ {\small \texttt{ \textcolor{brightblue!80!black}{ \hspace{4cm} // Update reference policy}}}
      \ENDIF
    \ENDFOR
    \STATE {\bfseries Output:} Optimal policy $\pi_{\psi}^{(T)}$.
\end{algorithmic}
\label{alg:mopo}
\end{algorithm}
\end{figure}

\section{Empirical Results}
\label{sec:chapter5-experiments}

We conducted extensive experimental evaluation on the relative empirical performance of the \MOPO{MOPO} algorithm to arrive at the following conclusions: (i) \MOPO{MOPO} exactly recovers the optimal policy under some canonical ordering of preferences, (ii) in contrast to several DPO‑style baselines \cite{ipo}, \MOPO{MOPO} does not overfit to the preference dataset and preserves performance on held‑out comparisons, (iii) it performs better or nearly as well as all baseline algorithms when evaluated on LLMs on real-world data, and (iv) it is more robust to stochastic and complex environments as compared to previous methods as we will see in the ablations. 

\subsection{Synthetic Sanity Check}

Here we verify if \MOPO{MOPO} solves the optimization problem exactly when ground‑truth optimum is known. We consider a two‑objective, context‑free bandit setting with discrete action set $\Ycal=\{y_1,y_2,y_3\}$ and a uniform reference policy $\pi_{\mathrm{ref}}$. Training data $\Dcal=\{(y_{i},y_{i}',\Ibb(y_{i},y_{i}'))\}_{i=1}^{N}$, where $\Ibb(\cdot, \cdot) \in \{0,1\}^{2}$ is the preference indicator vector, i.e., $\Ibb_{k}(y, y') = 1$ if $y \succ y'$ for the $k^{\text{th}}$ objective, and $0$ otherwise for $k \in \{1,2\}$, of size $N$ consists of one of three canonical preference structures:

\small
\begin{align*}
\Dcal_{1}&=\{(y_1,y_2,(1,1)),\,(y_2,y_3,(1,1)),\,(y_1,y_3,(1,1))\} &\text{(total order)}\\
\Dcal_{2}&=\{(y_1,y_2,(1,1)),\,(y_1,y_3,(1,0)),\,(y_2,y_3,(0,0))\} &\text{(partial order)}\\
\Dcal_{3}&=\{(y_1,y_2,(1,1)),\,(y_2,y_1,(0,1))\} &\text{(unobserved preferences)}
\end{align*}
\normalsize

\textbf{Learning Protocol.}  Mini‑batches are drawn uniformly \emph{with replacement} from each of $\Dcal_{j}$ for $j \in \{1,2,3\}$ and optimized with Algorithm \ref{alg:mopo} for 20k steps using Adam \cite{kingma2014adam} with a learning rate of 0.015 and batch size 12. Policy is encoded simply as $\pi_{\psi}(y_{i}) = \text{softmax}(\bm{\psi})_{i}$ using a vector $\bm{\psi} \in \Rbb^{3}$.

\begin{figure*}[t]
\centering

\begin{minipage}[h]{0.57\textwidth}
\vspace{0pt} 

\begin{tcolorbox}[
  width=\linewidth, nobeforeafter,
  coltitle=black, fonttitle=\fontfamily{lmss}\selectfont,
  title={$\tau=0.1$}, halign title=flush center,
  colback=backg_blue!5, colframe=darkgreen!10,
  boxrule=2pt, boxsep=2pt,
  top=-3pt, left=-4pt, right=-4pt, bottom=-1pt
]
  \centering
  {\includegraphics[scale=0.12]{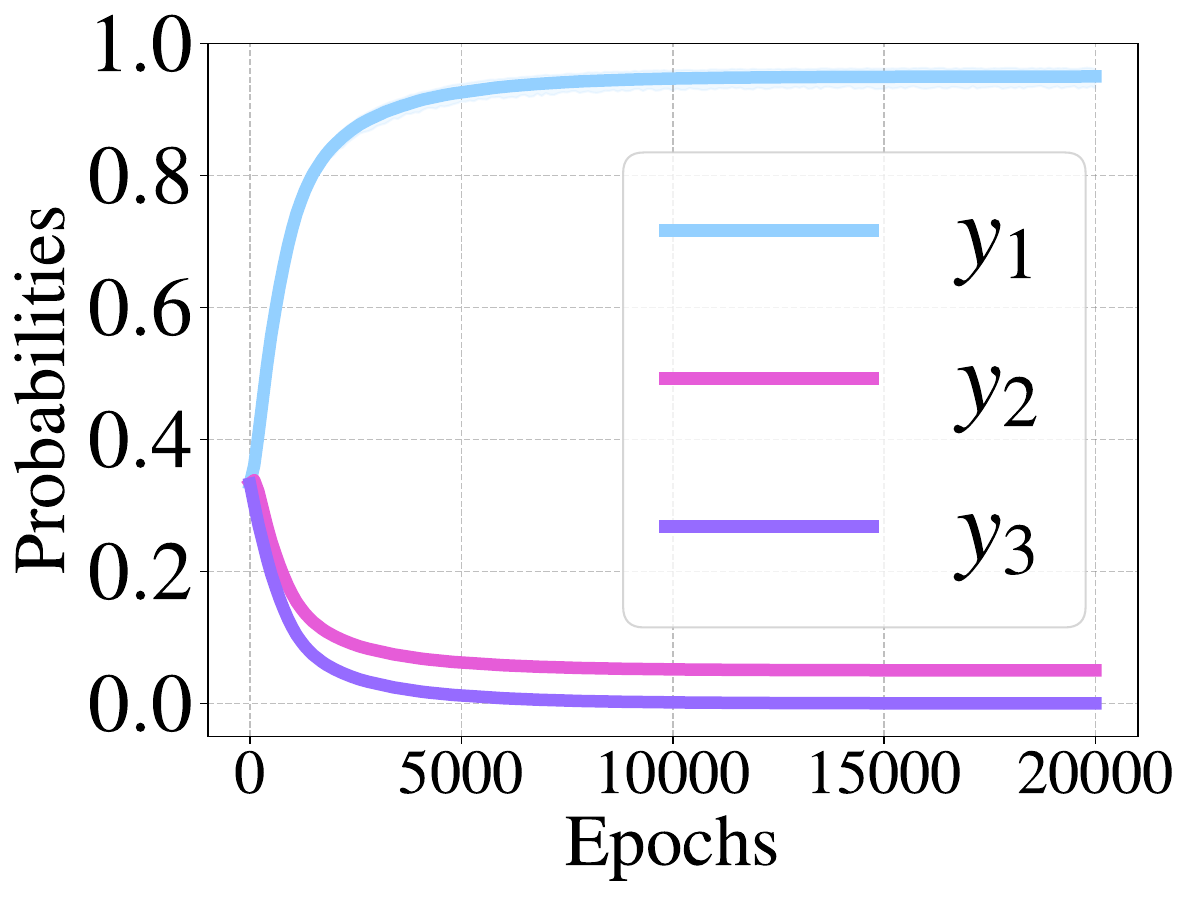}\label{fig:d1_tau-0-1_action_prob}}
  {\includegraphics[scale=0.12]{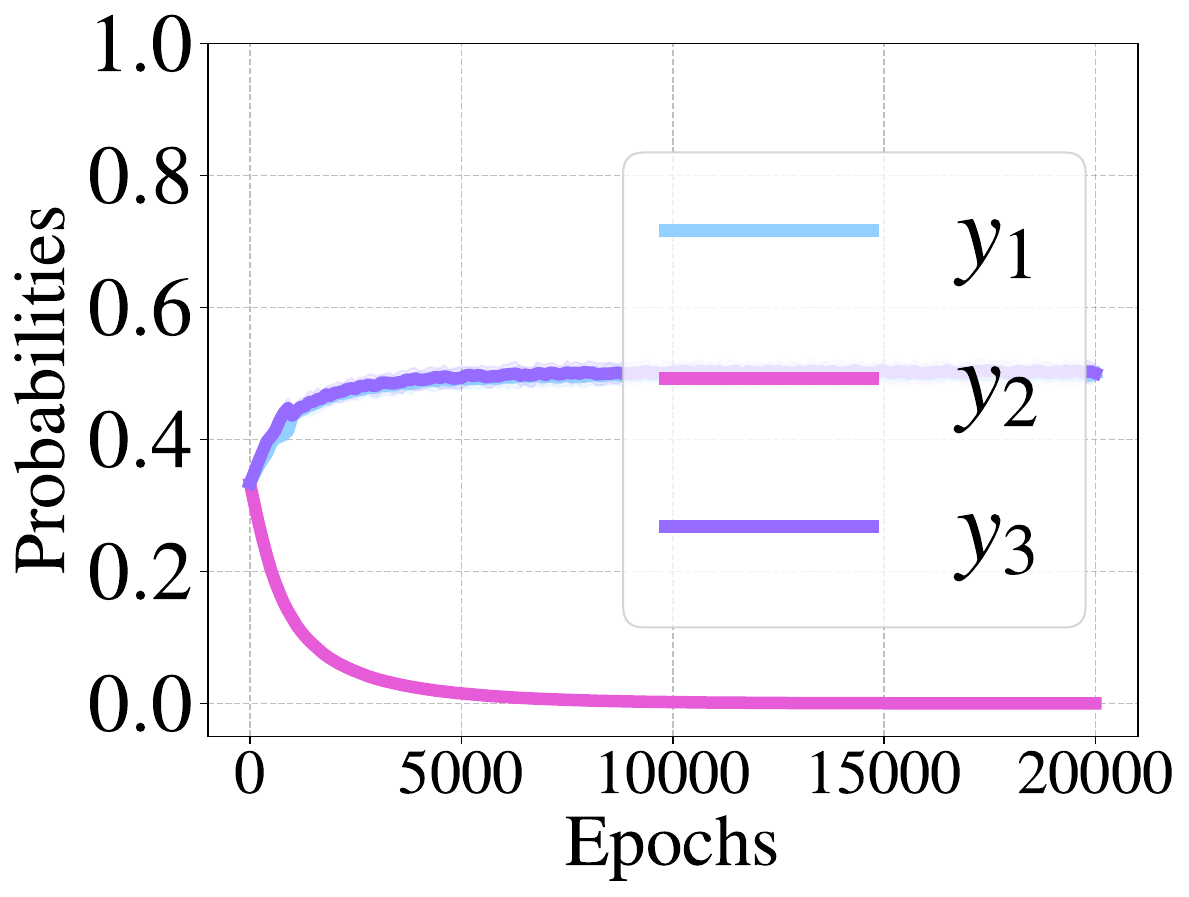}\label{fig:d2_tau-0-1_action_prob}}
  {\includegraphics[scale=0.12]{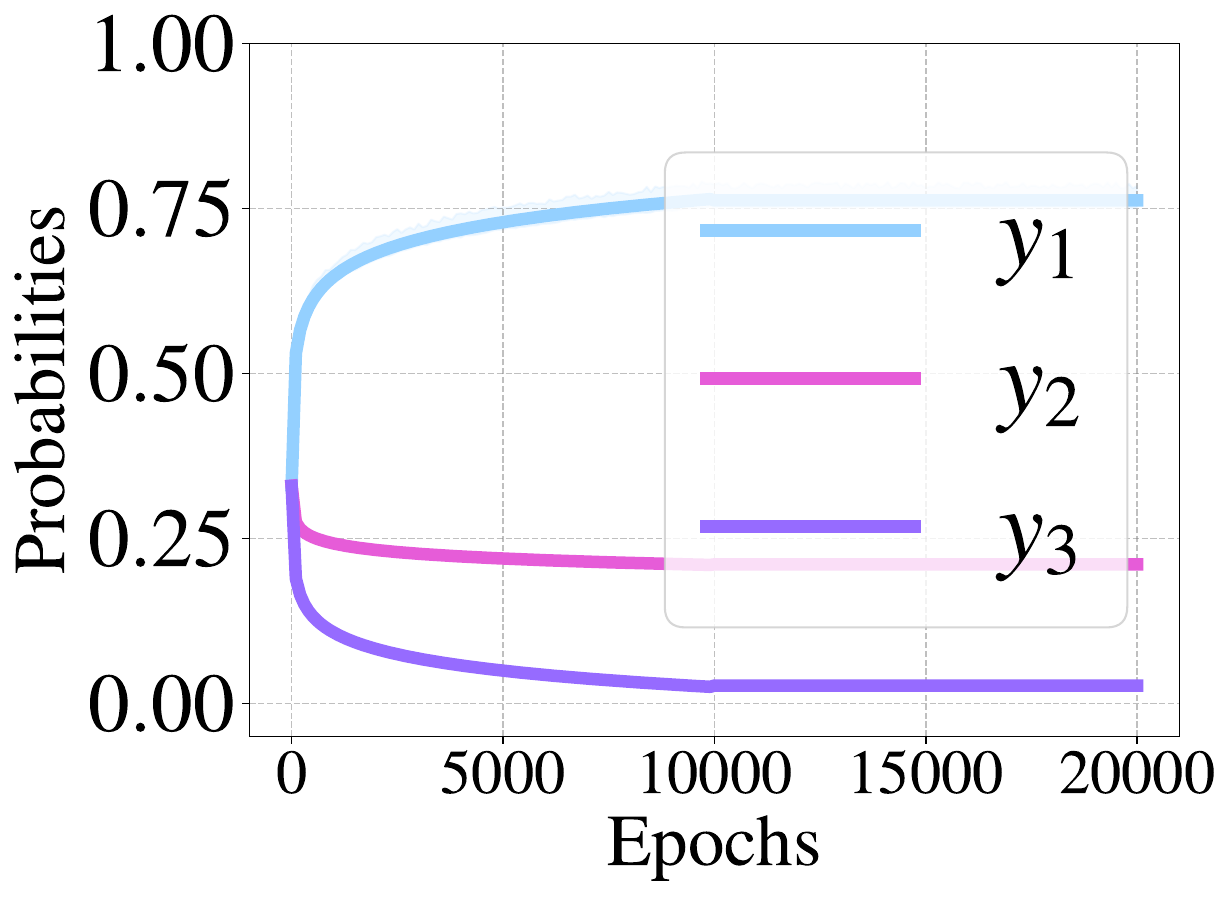}\label{fig:d5_tau-0-1_action_prob}}
\end{tcolorbox}

\begin{tcolorbox}[
  width=\linewidth, nobeforeafter,
  coltitle=black, fonttitle=\fontfamily{lmss}\selectfont,
  title={$\tau=1$}, halign title=flush center,
  colback=backg_blue!5, colframe=skyblue!20,
  boxrule=2pt, boxsep=1pt,
  top=-10pt, left=-4pt, right=-4pt, bottom=-1pt
]
  \centering
  \hspace{0.01cm}
  \subfloat[Dataset $\Dcal_{1}$]{\includegraphics[scale=0.12]{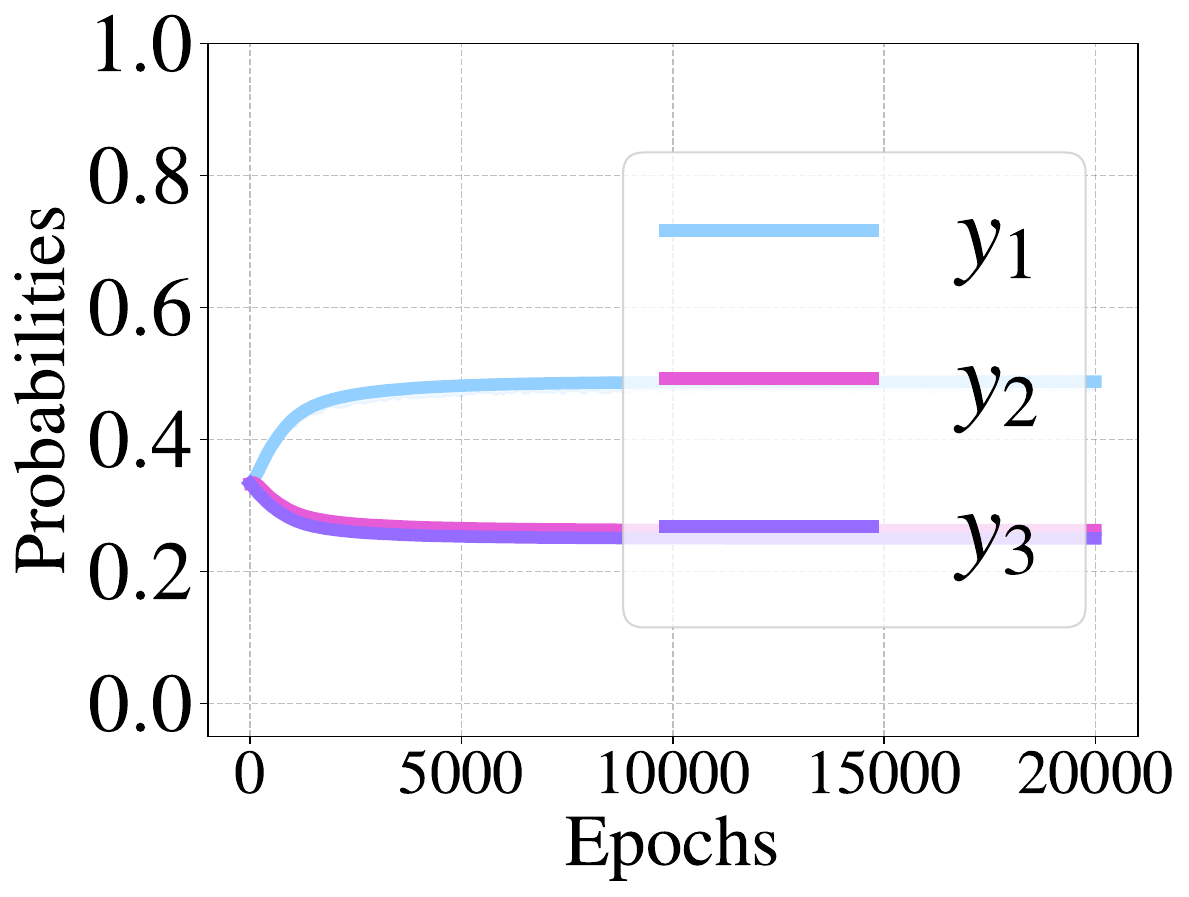}\label{fig:d1_tau-1-0_action_prob}}
  \hspace{0.01cm}
  \subfloat[Dataset $\Dcal_{2}$]{\includegraphics[scale=0.12]{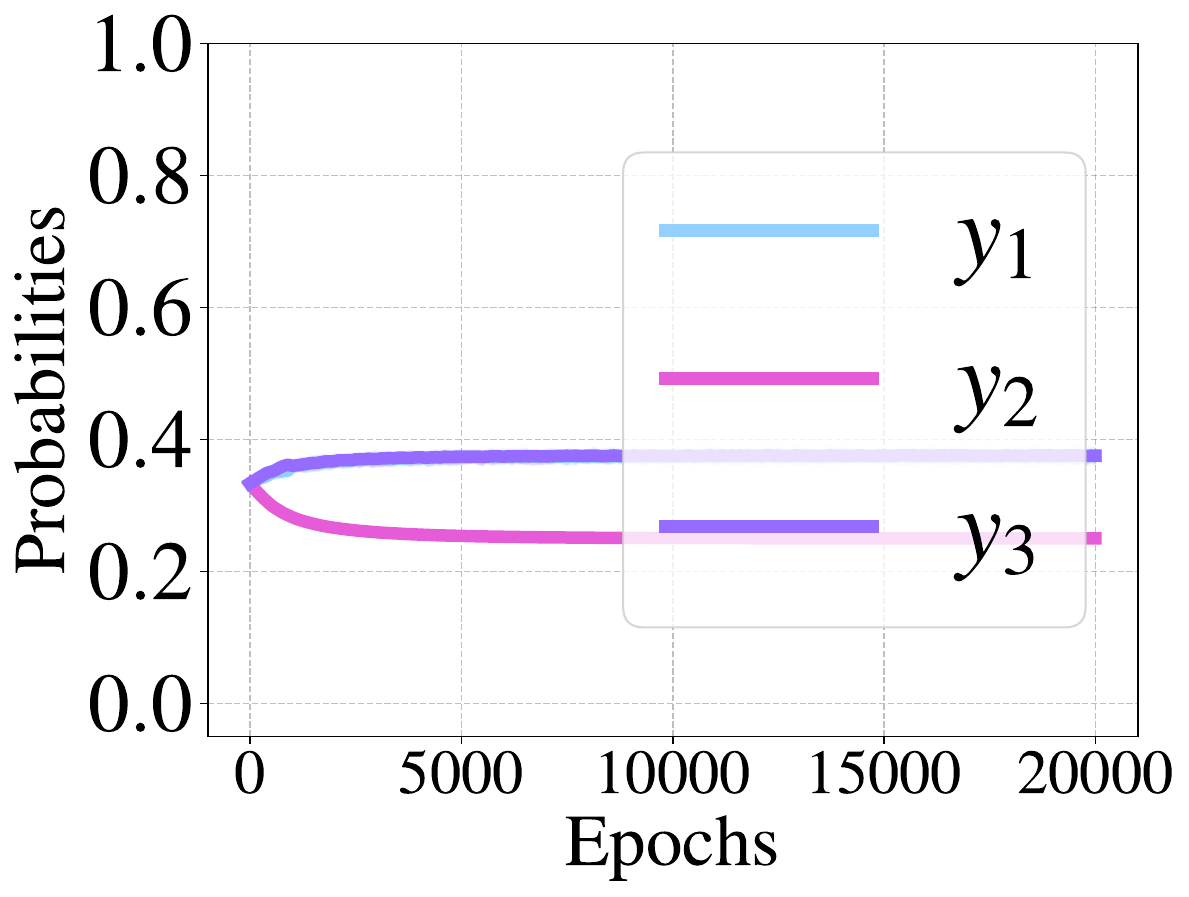}\label{fig:d2_tau-1-0_action_prob}}
  \hspace{0.05cm}
  \subfloat[Dataset $\Dcal_{3}$]{\includegraphics[scale=0.12]{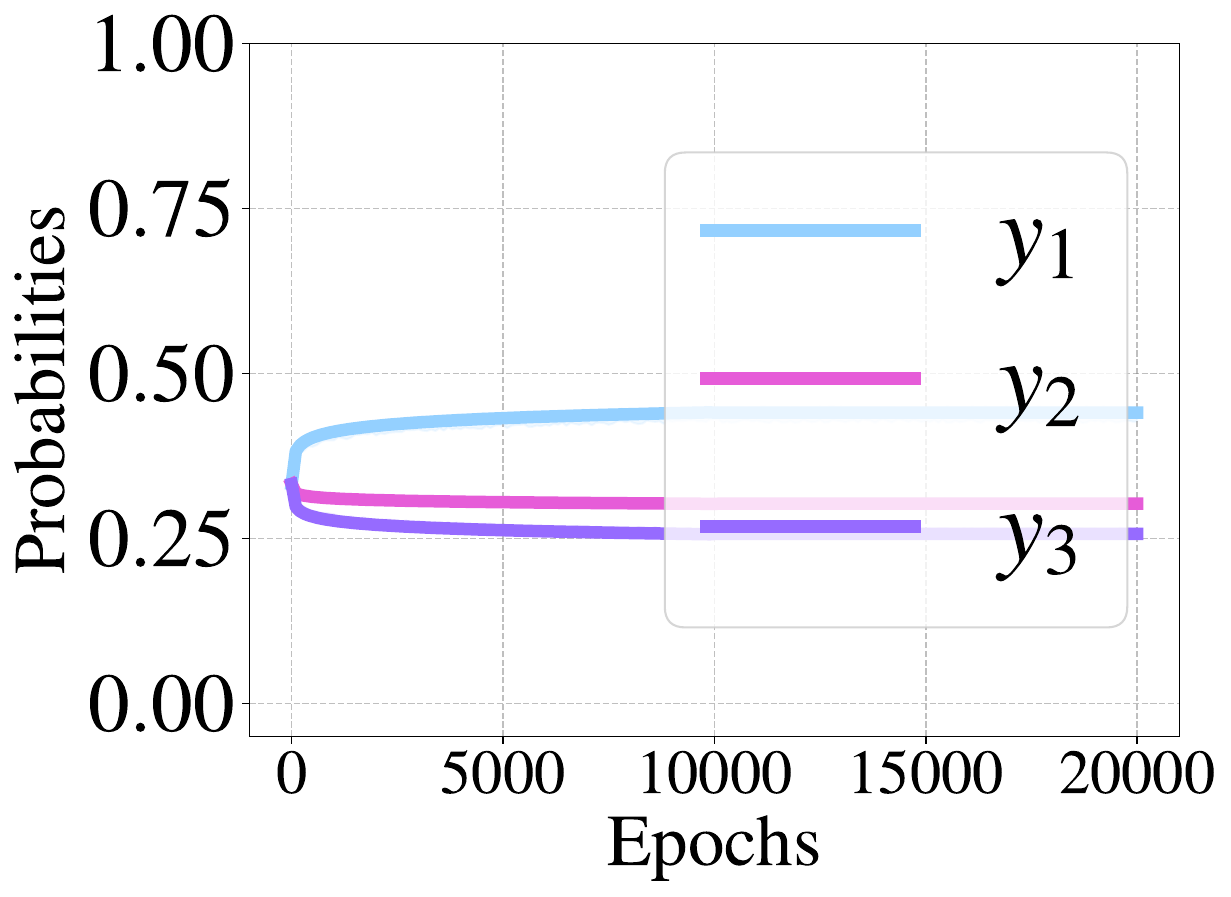}\label{fig:d5_tau-1-0_action_prob}}
\end{tcolorbox}
\caption{Learning curves of action probabilities of \MOPO{MOPO} on various dataset types (read column-wise).}
\label{fig:mopo_action_probs}
\end{minipage} \hfill
\begin{minipage}[h]{0.38\textwidth}
\begin{tcolorbox}[
  width=\linewidth, nobeforeafter,
  coltitle=black, fonttitle=\fontfamily{lmss}\selectfont,
  title={SFT $\to$ Alignment}, halign title=flush center,
  colback=backg_blue!5, colframe=purple!10,
  boxrule=2pt, boxsep=2pt,
  top=-1pt, left=-4pt, right=-4pt, bottom=-1pt
]
  \centering
  \includegraphics[scale=0.28]{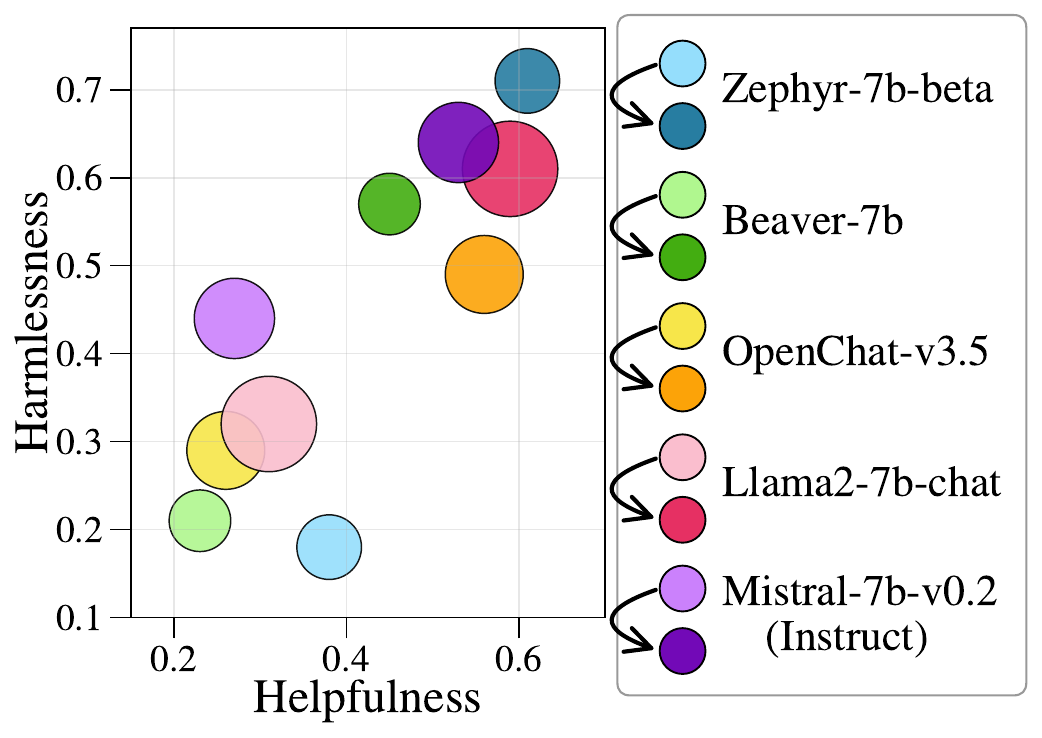}
\end{tcolorbox}
\caption{Open-sourced models before (pastel) and after (saturated) alignment using \MOPO{MOPO}. Circle size represents approximate training data size and annotation cost.}
\label{fig:openmodels}
\end{minipage}
\end{figure*}

\textbf{Results.}  The learned action probabilities for each $\Dcal_{j}$ (column-wise) are seen in Figure \ref{fig:mopo_action_probs}. For $\Dcal_{1}$ the policy converges to the Condorcet winner $y_1$. For $\Dcal_{2}$, in which $y_1$ and $y_3$ are undominated, \MOPO{MOPO} assigns them equal probability. 
Finally, with the inconsistent set $\Dcal_{3}$, \MOPO{MOPO} successively down‑weights the unobserved action $y_3$ as $\tau$ decreases. Across all cases, increasing $\tau$ smoothly interpolates between $\piref$ and the optimal policy, confirming controlled regularization.

\subsection{Experiments on Text Generation Tasks}

Having established the validity of alignment using \MOPO{MOPO}, in this section, we aim to evaluate the performance of our \MOPO{MOPO} algorithm on text generation tasks that involve diverse rewards.

\textbf{Baselines.} We consider (i) Rewards-in-Context (RiC) \cite{yang2024rewards} with in-context rewards and human preferences (this is a state-of-the-art baseline that outperforms Rewarded Soups \cite{rame2023rewarded} and MORLHF \cite{li2020deep}), (ii) 
Preference-aware Autoregressive Reward Model (PARM) \cite{lin2025parm}, a single unified model trained across all preference dimensions, (iii) Multi-objective DPO (MODPO) \cite{zhou2023beyond}, which is a more nuanced version of scalarizing multiple reward models into one (this corresponds to ``DPO on $w^{T}r$" in Figure \ref{fig:dpo_comparison}), and (iv) DPO on \Djoint{$\Dcal_{J}$}, which trains DPO on context-output pairs which are preferred under both reward models similar to our discussion in Section \ref{sec:preliminaries}. We also compare two verions of \MOPO{MOPO}: (i) using the log-barrier formulation of Remark \ref{remark:logbarrier} called \MOPO{MOPO-LB}, and (ii) using the Lagrangian approach called \MOPO{MOPO-Lag}.

\textbf{Benchmarks and Training.} All methods are evaluated based on the quality of their empirical Pareto fronts on the Helpful Assistant task \cite{bai2022training} and the Reddit Summarization task \cite{stiennon2020learning}. The Helpful Assistant task uses the HH-RLHF dataset containing 160k prompts with human preference annotations. Evaluation is performed using three HuggingFace reward models --`helpful', `harmless', and `humor' -- which score responses from different perspectives \cite{wolf2019huggingface}. The Reddit Summary task consists of 14.9k posts and their summaries. We consider three reward models: `pref1' and `no-hallucinate', which evaluate human preference for summaries, and a `faithful' reward that measures the faithfulness of the summary to the original post. We apply these benchmarks to publicly available 7B-level models that have shown strong helpfulness scores \cite{eval-harness, dubois2023alpacafarm}. From the evaluation in Figure \ref{fig:openmodels}, even though we observe that Zephyr-7b-beta \cite{tunstall2023zephyr}, an open-source model fine-tuned over Mistral-7B-v0.1 \cite{jiang2023mistral7b}, Pareto dominates other models when aligned using \MOPO{MOPO}, we conduct experiments on various models to show generalizability of \MOPO{MOPO}. All likelihood maximization problems use parameter efficient fine-tuning with LoRA \cite{hu2022lora} for 10k steps with a batch size of 8. LoRA is applied to the shared transformer backbone, and is optimized together with the policy parameters. See Appendix \ref{appendix:implementation_details} for more details.

\begin{figure*}[h]

\begin{tcolorbox}[width=\textwidth, nobeforeafter, coltitle = black, fonttitle=\fontfamily{lmss}\selectfont, title={RiC vs PARM vs MODPO vs DPO on \Djoint{$\Dcal_{J}$} vs \MOPO{MOPO}}, halign title=flush center, colback=backg_blue!5, colframe=LinTS!10, boxrule=2pt, boxsep=2pt, grow to left by=-0.5mm, top=-1pt, left=-4pt, right=-4pt, bottom=-1pt]
\centering
{

    {\includegraphics[height=0.2\textwidth, width=0.23\textwidth]{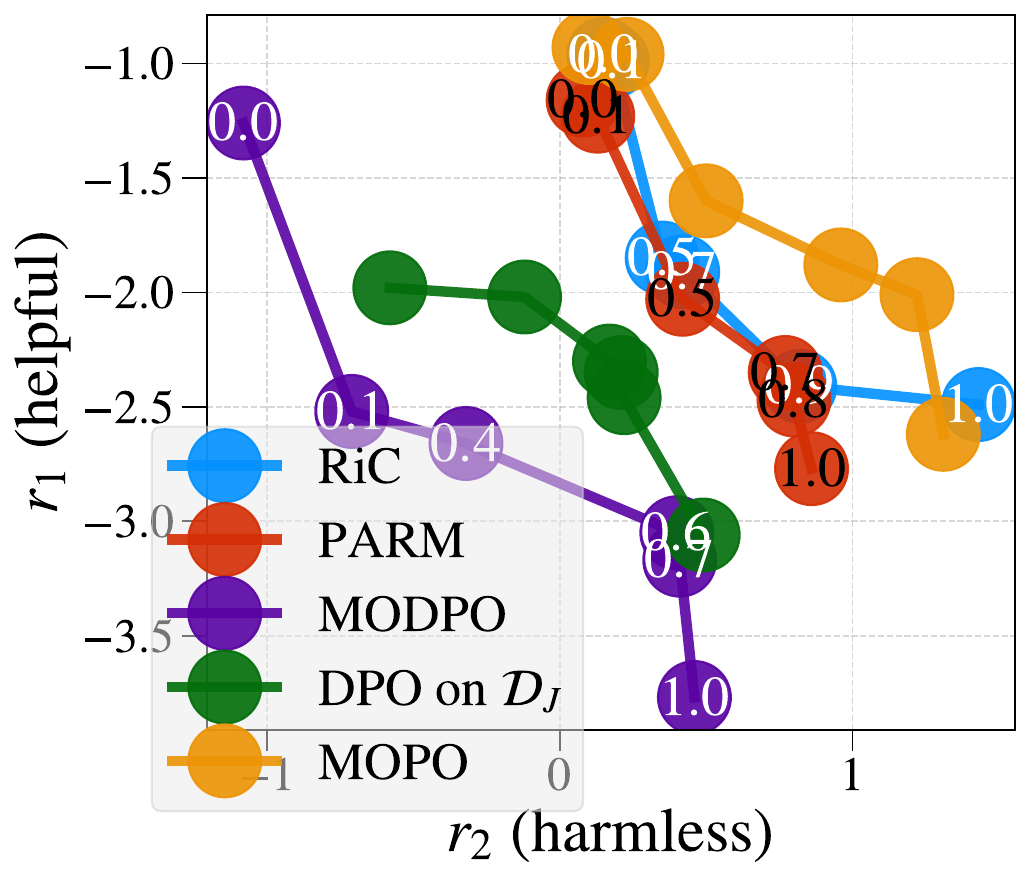}}
}
{
    {\includegraphics[height=0.2\textwidth, width=0.23\textwidth]{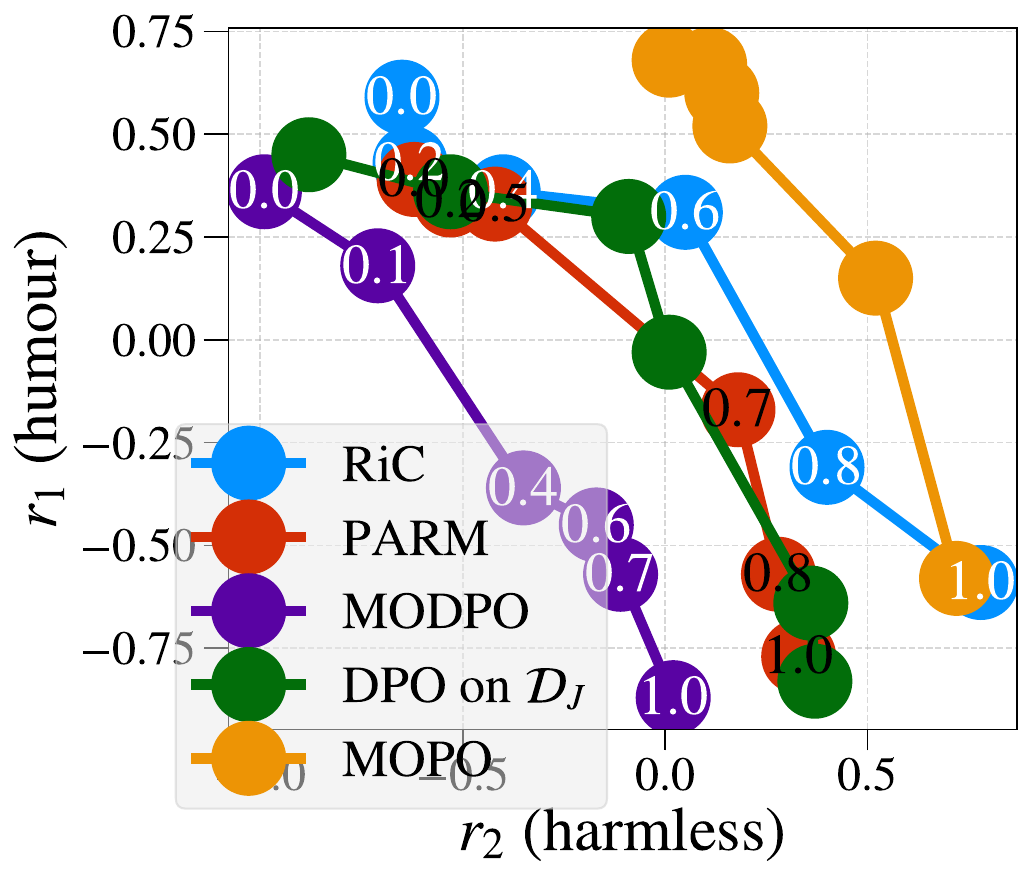}}
}
{
    {\includegraphics[height=0.2\textwidth, width=0.24\textwidth]{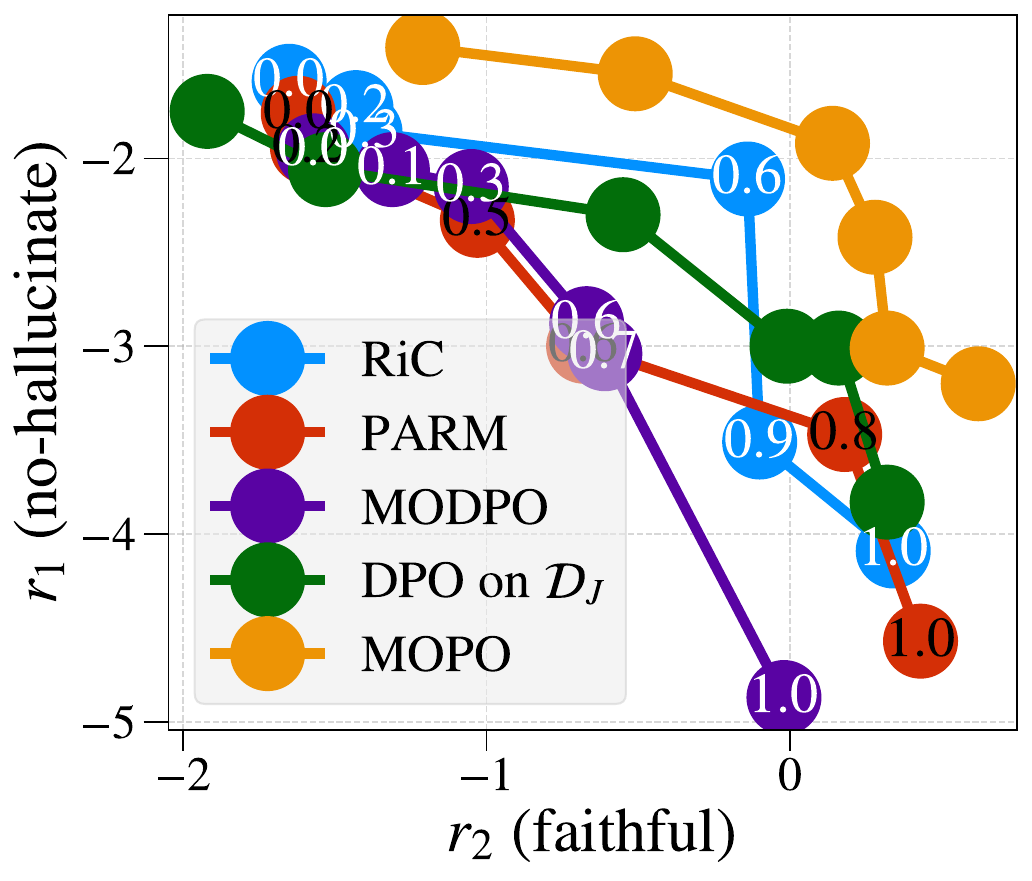}}
}
{
    {\includegraphics[height=0.2\textwidth, width=0.23\textwidth]{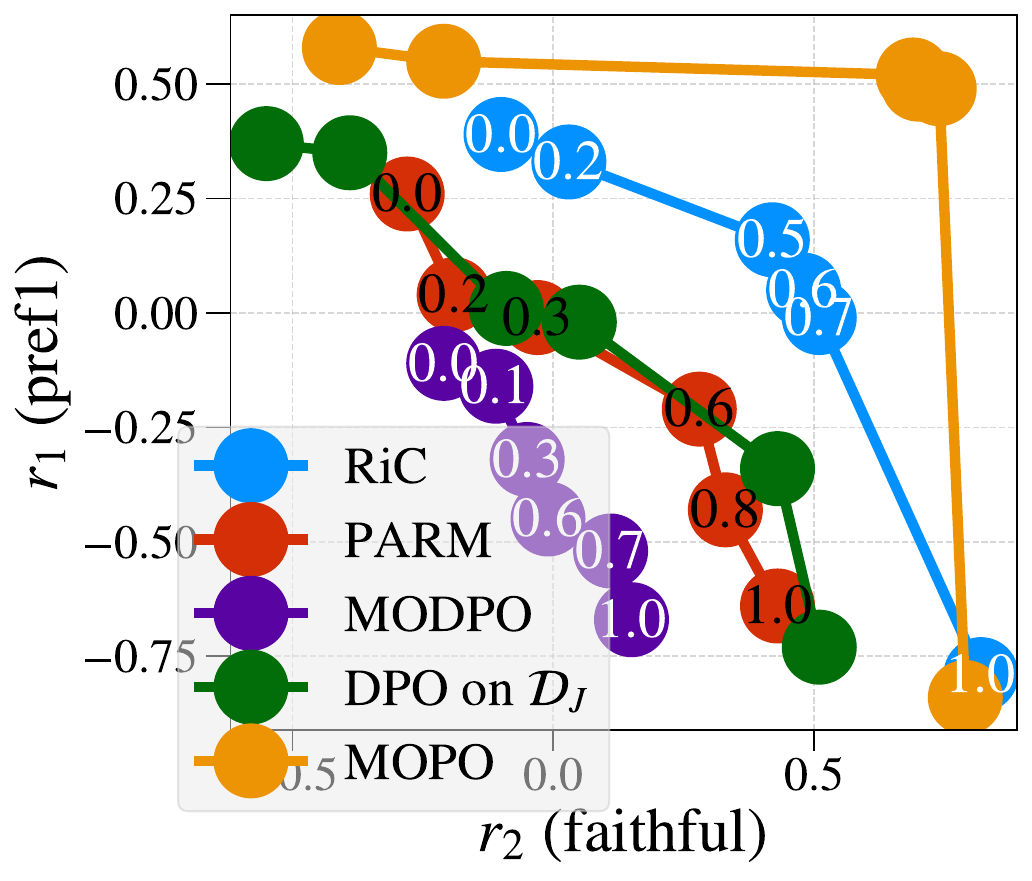}}
}
\end{tcolorbox}
\vspace{-0.2cm}
\caption{Empirical Pareto fronts of phi-1.5 when aligned using baselines on the Helpful Assistant and Reddit Summary tasks.}
\label{fig:phi_pareto}
\end{figure*}

\begin{table*}[t]
\centering
\caption{Results of the Helpful Assistant task and the Reddit Summary task with normalized rewards.
Color intensity reflects per-column magnitude (red = low, green = high).}
\small
\begin{adjustbox}{max width=\textwidth}
\begin{tabular}{cl|c|c|c|c}
\toprule
\textbf{Model} & \textbf{Algorithm} &
\makecell{\texttt{\small helpful-harmless} \\ $(r_{1},r_{2})$} &
\makecell{\texttt{\small humour-harmless} \\ $(r_{1},r_{2})$} &
\makecell{\texttt{\scriptsize no\_hallucinate-faithful} \\ $(r_{1},r_{2})$} &
\makecell{\texttt{\small pref1-faithful} \\ $(r_{1},r_{2})$} \\
\midrule

\multirow{7}{*}{\makecell{\textbf{phi-1.5} \\ \cite{li2023textbooks}}}
& RiC & (\accp{0.23}{0.03}{1}, \accp{0.21}{0.03}{1}) & (\accp{0.31}{0.04}{2}, \accp{0.27}{0.03}{2}) & (\accp{0.40}{0.06}{3}, \accp{0.37}{0.05}{3}) & (\accp{0.39}{0.05}{4}, \accp{0.36}{0.07}{4})  \\
& PARM & (\accp{0.28}{0.02}{1}, \accp{0.24}{0.05}{1}) & (\accp{0.33}{0.05}{2}, \accp{0.32}{0.04}{2}) & \cellcolor{mopogray!60}{(\accp{0.39}{0.05}{3}, \accp{0.41}{0.05}{3})} & \cellcolor{mopogray!60}{(\accp{0.47}{0.07}{4}, \accp{0.49}{0.06}{4})}  \\
& MODPO & (\accp{0.21}{0.04}{1}, \accp{0.20}{0.04}{1}) & (\accp{0.25}{0.04}{2}, \accp{0.24}{0.06}{2}) & (\accp{0.34}{0.07}{3}, \accp{0.35}{0.04}{3}) & (\accp{0.31}{0.06}{4}, \accp{0.30}{0.04}{4})  \\
& DPO on \Djoint{$\Dcal_{J}$} & (\accp{0.22}{0.03}{1}, \accp{0.19}{0.03}{1}) & (\accp{0.27}{0.05}{2}, \accp{0.25}{0.04}{2}) & (\accp{0.35}{0.03}{3}, \accp{0.35}{0.05}{3}) & (\accp{0.31}{0.04}{4}, \accp{0.34}{0.03}{4})  \\
& \MOPO{MOPO-LB} &
(\accp{0.27}{0.04}{1}, \accp{0.24}{0.04}{1}) &
(\accp{0.34}{0.06}{2}, \accp{0.35}{0.05}{2}) &
\cellcolor{mopogray!60}{(\accp{0.46}{0.07}{3}, \accp{0.40}{0.04}{3})} &
(\accp{0.44}{0.03}{4}, \accp{0.47}{0.05}{4}) \\
& \MOPO{MOPO-Lag} &
\cellcolor{mopogray!60}{(\accp{0.29}{0.05}{1}, \accp{0.26}{0.03}{1})} &
\cellcolor{mopogray!60}{(\accp{0.37}{0.05}{2}, \accp{0.36}{0.06}{2})} &
(\accp{0.42}{0.04}{3}, \accp{0.39}{0.05}{3}) &
(\accp{0.47}{0.08}{4}, \accp{0.45}{0.06}{4}) \\
\midrule

\multirow{7}{*}{\makecell{\textbf{OpenChat-v3.5} \\ \cite{wang2023openchat}}}
& RiC & (\accp{0.40}{0.04}{1}, \accp{0.33}{0.03}{1}) & (\accp{0.37}{0.03}{2}, \accp{0.40}{0.04}{2}) & (\accp{0.55}{0.05}{3}, \accp{0.43}{0.04}{3}) & (\accp{0.40}{0.04}{4}, \accp{0.43}{0.03}{4}) \\
& PARM & (\accp{0.38}{0.03}{1}, \accp{0.37}{0.06}{1}) & (\accp{0.36}{0.05}{2}, \accp{0.39}{0.04}{2}) & (\accp{0.60}{0.04}{3}, \accp{0.47}{0.02}{3}) & (\accp{0.39}{0.03}{4}, \accp{0.40}{0.05}{4}) \\
& MODPO & (\accp{0.35}{0.05}{1}, \accp{0.33}{0.04}{1}) & (\accp{0.39}{0.04}{2}, \accp{0.37}{0.05}{2}) & (\accp{0.51}{0.03}{3}, \accp{0.48}{0.04}{3}) & (\accp{0.41}{0.04}{4}, \accp{0.39}{0.02}{4}) \\
& DPO on \Djoint{$\Dcal_{J}$} & (\accp{0.28}{0.02}{1}, \accp{0.26}{0.02}{1}) & (\accp{0.36}{0.03}{2}, \accp{0.35}{0.02}{2}) & (\accp{0.52}{0.04}{3}, \accp{0.47}{0.06}{3}) & (\accp{0.35}{0.01}{4}, \accp{0.37}{0.03}{4}) \\
& \MOPO{MOPO-LB} &
(\accp{0.41}{0.04}{1}, \accp{0.39}{0.03}{1}) &
(\accp{0.40}{0.05}{2}, \accp{0.39}{0.05}{2}) &
\cellcolor{mopogray!60}{(\accp{0.63}{0.02}{3}, \accp{0.50}{0.03}{3})} &
\cellcolor{mopogray!60}{(\accp{0.46}{0.02}{4}, \accp{0.44}{0.03}{4})} \\
& \MOPO{MOPO-Lag} &
\cellcolor{mopogray!60}{(\accp{0.43}{0.05}{1}, \accp{0.41}{0.04}{1})} &
\cellcolor{mopogray!60}{(\accp{0.42}{0.04}{2}, \accp{0.40}{0.04}{2})} &
(\accp{0.61}{0.01}{3}, \accp{0.49}{0.04}{3}) &
(\accp{0.44}{0.04}{4}, \accp{0.42}{0.01}{4}) \\
\midrule

\multirow{7}{*}{\makecell{\textbf{Llama-3.1-8B} \\ \cite{grattafiori2024llama}}}
& RiC & \cellcolor{mopogray!60}{(\accp{0.41}{0.05}{1}, \accp{0.47}{0.03}{1})} & (\accp{0.43}{0.04}{2}, \accp{0.41}{0.02}{2}) & (\accp{0.40}{0.03}{3}, \accp{0.42}{0.05}{3}) & (\accp{0.40}{0.05}{4}, \accp{0.41}{0.03}{4}) \\
& PARM & (\accp{0.40}{0.01}{1}, \accp{0.42}{0.04}{1}) & (\accp{0.35}{0.03}{2}, \accp{0.44}{0.01}{2}) & \cellcolor{mopogray!60}{(\accp{0.44}{0.04}{3}, \accp{0.55}{0.01}{3})} & (\accp{0.49}{0.03}{4}, \accp{0.48}{0.04}{4}) \\
& MODPO & (\accp{0.32}{0.04}{1}, \accp{0.33}{0.03}{1}) & (\accp{0.37}{0.05}{2}, \accp{0.45}{0.04}{2}) & (\accp{0.46}{0.02}{3}, \accp{0.46}{0.03}{3}) & (\accp{0.48}{0.04}{4}, \accp{0.46}{0.06}{4}) \\
& DPO on \Djoint{$\Dcal_{J}$} & (\accp{0.31}{0.03}{1}, \accp{0.30}{0.04}{1}) & (\accp{0.35}{0.04}{2}, \accp{0.40}{0.04}{2}) & (\accp{0.39}{0.04}{3}, \accp{0.40}{0.03}{3}) & (\accp{0.39}{0.01}{4}, \accp{0.39}{0.03}{4}) \\
& \MOPO{MOPO-LB} &
(\accp{0.45}{0.02}{1}, \accp{0.43}{0.05}{1}) &
(\accp{0.42}{0.05}{2}, \accp{0.50}{0.02}{2}) &
(\accp{0.51}{0.03}{3}, \accp{0.49}{0.01}{3}) &
\cellcolor{mopogray!60}{(\accp{0.52}{0.03}{4}, \accp{0.50}{0.05}{4})} \\
& \MOPO{MOPO-Lag} &
\cellcolor{mopogray!60}{(\accp{0.48}{0.02}{1}, \accp{0.46}{0.03}{1})} &
\cellcolor{mopogray!60}{(\accp{0.45}{0.03}{2}, \accp{0.53}{0.05}{2})} &
\cellcolor{mopogray!60}{(\accp{0.54}{0.05}{3}, \accp{0.52}{0.02}{3})} &
(\accp{0.50}{0.05}{4}, \accp{0.49}{0.02}{4}) \\
\midrule

\multirow{7}{*}{\makecell{\textbf{Mistral-7b-v0.2} \\ \textbf{(Instruct)} \\ \cite{jiang2023mistral7b}}}
& RiC & (\accp{0.44}{0.04}{1}, \accp{0.41}{0.03}{1}) & (\accp{0.43}{0.05}{2}, \accp{0.44}{0.02}{2}) & (\accp{0.45}{0.04}{3}, \accp{0.41}{0.03}{3}) & \cellcolor{mopogray!60}{(\accp{0.46}{0.02}{4}, \accp{0.44}{0.05}{4})} \\
& PARM & \cellcolor{mopogray!60}{(\accp{0.48}{0.03}{1}, \accp{0.47}{0.01}{1})} & (\accp{0.41}{0.03}{2}, \accp{0.46}{0.04}{2}) & (\accp{0.43}{0.02}{3}, \accp{0.42}{0.02}{3}) & (\accp{0.43}{0.05}{4}, \accp{0.42}{0.04}{4}) \\
& MODPO & (\accp{0.41}{0.05}{1}, \accp{0.39}{0.04}{1}) & (\accp{0.40}{0.02}{2}, \accp{0.42}{0.02}{2}) & (\accp{0.43}{0.03}{3}, \accp{0.43}{0.05}{3}) & (\accp{0.37}{0.04}{4}, \accp{0.38}{0.03}{4}) \\
& DPO on \Djoint{$\Dcal_{J}$} & (\accp{0.32}{0.01}{1}, \accp{0.30}{0.02}{1}) & (\accp{0.36}{0.04}{2}, \accp{0.39}{0.03}{2}) & (\accp{0.39}{0.05}{3}, \accp{0.40}{0.01}{3}) & (\accp{0.36}{0.03}{4}, \accp{0.35}{0.04}{4}) \\
& \MOPO{MOPO-LB} &
(\accp{0.44}{0.02}{1}, \accp{0.45}{0.04}{1}) &
\cellcolor{mopogray!60}{(\accp{0.49}{0.04}{2}, \accp{0.48}{0.05}{2})} &
(\accp{0.47}{0.04}{3}, \accp{0.45}{0.03}{3}) &
(\accp{0.45}{0.04}{4}, \accp{0.43}{0.03}{4}) \\
& \MOPO{MOPO-Lag} &
(\accp{0.45}{0.04}{1}, \accp{0.47}{0.03}{1}) &
(\accp{0.49}{0.02}{2}, \accp{0.47}{0.04}{2}) &
\cellcolor{mopogray!60}{(\accp{0.48}{0.02}{3}, \accp{0.46}{0.02}{3})} &
(\accp{0.45}{0.02}{4}, \accp{0.42}{0.02}{4}) \\
\midrule

\multirow{7}{*}{\makecell{\textbf{Zephyr-7b-beta} \\ \cite{tunstall2023zephyr}}}
& RiC & (\accp{0.43}{0.02}{1}, \accp{0.46}{0.03}{1}) & (\accp{0.45}{0.02}{2}, \accp{0.47}{0.03}{2}) & (\accp{0.50}{0.03}{3}, \accp{0.48}{0.02}{3}) & (\accp{0.48}{0.02}{4}, \accp{0.44}{0.04}{4}) \\
& PARM & (\accp{0.45}{0.01}{1}, \accp{0.48}{0.02}{1}) & (\accp{0.48}{0.04}{2}, \accp{0.50}{0.02}{2}) & (\accp{0.56}{0.04}{3}, \accp{0.53}{0.03}{3}) & \cellcolor{mopogray!60}{(\accp{0.52}{0.01}{4}, \accp{0.46}{0.03}{4})} \\
& MODPO & (\accp{0.35}{0.01}{1}, \accp{0.33}{0.04}{1}) & (\accp{0.40}{0.03}{2}, \accp{0.40}{0.05}{2}) & (\accp{0.43}{0.01}{3}, \accp{0.44}{0.02}{3}) & (\accp{0.39}{0.03}{4}, \accp{0.40}{0.02}{4}) \\
& DPO on \Djoint{$\Dcal_{J}$} & (\accp{0.34}{0.03}{1}, \accp{0.38}{0.02}{1}) & (\accp{0.38}{0.01}{2}, \accp{0.41}{0.03}{2}) & (\accp{0.41}{0.02}{3}, \accp{0.43}{0.05}{3}) & (\accp{0.36}{0.04}{4}, \accp{0.36}{0.05}{4}) \\
& \MOPO{MOPO-LB} &
\cellcolor{mopogray!60}{(\accp{0.51}{0.04}{1}, \accp{0.52}{0.04}{1})} &
(\accp{0.49}{0.02}{2}, \accp{0.48}{0.04}{2}) &
(\accp{0.55}{0.03}{3}, \accp{0.51}{0.01}{3}) &
(\accp{0.48}{0.02}{4}, \accp{0.43}{0.03}{4}) \\
& \MOPO{MOPO-Lag} &
\cellcolor{mopogray!60}{(\accp{0.53}{0.02}{1}, \accp{0.51}{0.03}{1})} &
\cellcolor{mopogray!60}{(\accp{0.52}{0.03}{2}, \accp{0.53}{0.04}{2})} &
\cellcolor{mopogray!60}{(\accp{0.60}{0.02}{3}, \accp{0.55}{0.03}{3})} &
(\accp{0.50}{0.04}{4}, \accp{0.44}{0.02}{4}) \\
\bottomrule
\end{tabular}
\end{adjustbox}
\label{tab:empirical_tasks_comparison}
\end{table*}

\textbf{Evaluation protocol.}  For both datasets, we uniformly sample 3k prompts from the test sets and cluster them into $j=6$ clusters, and compute the average reward for each objective across cluster groups \cite{mukherjee2024multi, yang2024rewards}. Performance is measured by comparing the resulting multi-objective reward values. 

\begin{wraptable}{r}{0.4\textwidth}
\vspace{-0.5cm}
\caption{Three objective alignment for Helpful Assistant task with normalized rewards.}
\label{tab:threeobjective}
{
\renewcommand{\arraystretch}{1.1}
\fontsize{7}{9}\selectfont
\addtolength{\tabcolsep}{-0.1em}
\begin{tabular}{c|ccc}
\hline
 & helpful & humour & harmless \\
\hline
RLHF-r1 & \cellcolor{accgood!60}{0.76} & -0.42 & -0.23 \\ 
\hline
RLHF-r2 & -0.81 & \cellcolor{accgood!60}{0.53} & -0.40 \\ 
\hline
RLHF-r3 & -0.79 & -0.92 & \cellcolor{accgood!60}{0.42} \\ 
\hline
RiC & 0.25 & 0.15 & 0.11 \\ \hline
PARM & 0.31 & 0.17 & \cellcolor{accgood!20}{0.23} \\ 
\hline
MODPO & 0.04 & -0.09 & 0.08 \\
\hline
DPO on \Djoint{$\Dcal_{J}$} & 0.18 & 0.09 & 0.11 \\
\hline
\MOPO{MOPO-LB} & 0.30 & 0.19 & 0.18 \\
\hline
\MOPO{MOPO-Lag} & \cellcolor{accgood!20}{0.39} & \cellcolor{accgood!20}{0.22} & 0.17 \\ 
\hline
\end{tabular}
}
\vspace{-0.2cm}
\end{wraptable}

See Table \ref{tab:empirical_tasks_comparison} for results over 5 independent runs. Each cell represents reward tuples along with 1 standard deviation of models when aligned with baselines and trained on the corresponding preference dataset. For \MOPO{MOPO}, we let the primary objective be $r_{1}$ and constrain preferences w.r.t. $r_{2}$. For RiC, PARM, and MODPO, values with highest rewards across inference preference vector inputs are shown. According to the results, \MOPO{MOPO} achieves substantial alignment improvement for most objectives and policy models. From the policy model scale perspective, \MOPO{MOPO} is able to scale from tiny-LLMs (phi-1.5) to larger LLMs as well (Llama-3.1-8B). In Figure \ref{fig:phi_pareto} we use phi-1.5 as the SFT model for alignment tasks to empirically visualize Pareto fronts of all baselines. Each point represents the average rewards evaluated on sampled cluster groups from the test set. For \MOPO{MOPO}, we vary the constraint threshold w.r.t. $r_{2}$ to obtain points on the empirical front. For RiC and PARM, the numbers at the centers of the markers indicate the (normalized) preference for $r_{2}$ in each cluster that achieves the highest reward within that cluster. It is clear from the results that \MOPO{MOPO} consistently approximates the Pareto front as well as or better than PARM and RiC.

To assess the scalability of \MOPO{MOPO}, we aim to optimize three objectives in the
Helpful Assistant task, i.e., ‘harmless’, ‘helpful’, and ‘humour’. We use Zephyr-7b-beta as our base SFT model. For easy interpretation, we sample 2k prompts from the test set and plot the average rewards. The results in Table \ref{tab:threeobjective} reveal that RLHF \cite{yang2024rewards}, when optimized for a single reward (see Appendix \ref{appendix:implementation_details} for the problem formulation), achieves high performance on the targeted reward but degrades substantially on the remaining objectives. In contrast, multi-objective algorithms yield more uniform performance across all rewards, with \MOPO{MOPO} achieving the most balanced trade-offs. The results demonstrate the effectiveness of \MOPO{MOPO} in scaling to more than two objectives as well.

\begin{figure*}[ht]
\vspace{-0.1cm}
\centering
\includegraphics[width=0.3\textwidth]{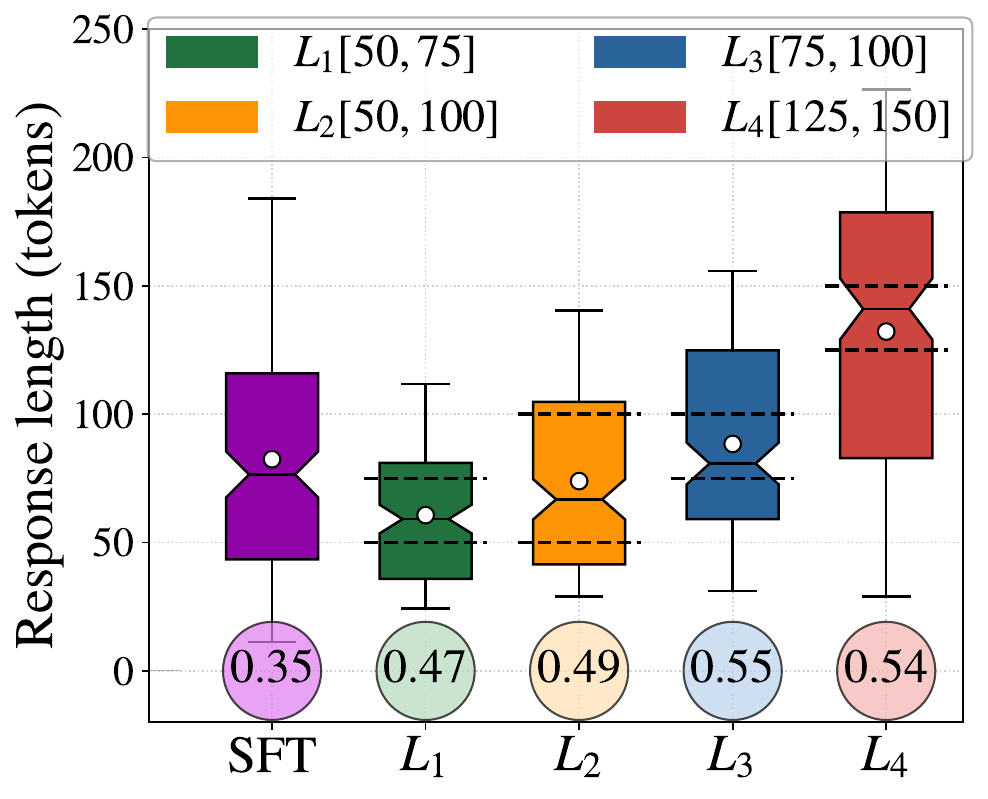}
\caption{\small Response lengths along with helpfulness scores in circles, where notches indicate medians, boxes show $\pm$25\% quantiles, white circles mark the means, and the black dashed lines depict the constraints imposed.}
\label{fig:length}
\vspace{-0.2cm}
\end{figure*}

\begin{figure*}[ht]
\centering
\begin{tcolorbox}[width=0.98\textwidth, nobeforeafter, coltitle = black, fonttitle=\fontfamily{lmss}\selectfont, title={Effect of regularizer $\tau$}, halign title=flush center, colback=backg_blue!5, colframe=darkgreen!10, boxrule=2pt, boxsep=2pt, grow to left by=-0.5mm, top=0pt, left=-4pt, right=-4pt, bottom=-1pt]
    \centering
{
    {\includegraphics[height=0.20\textwidth, width=0.23\textwidth]{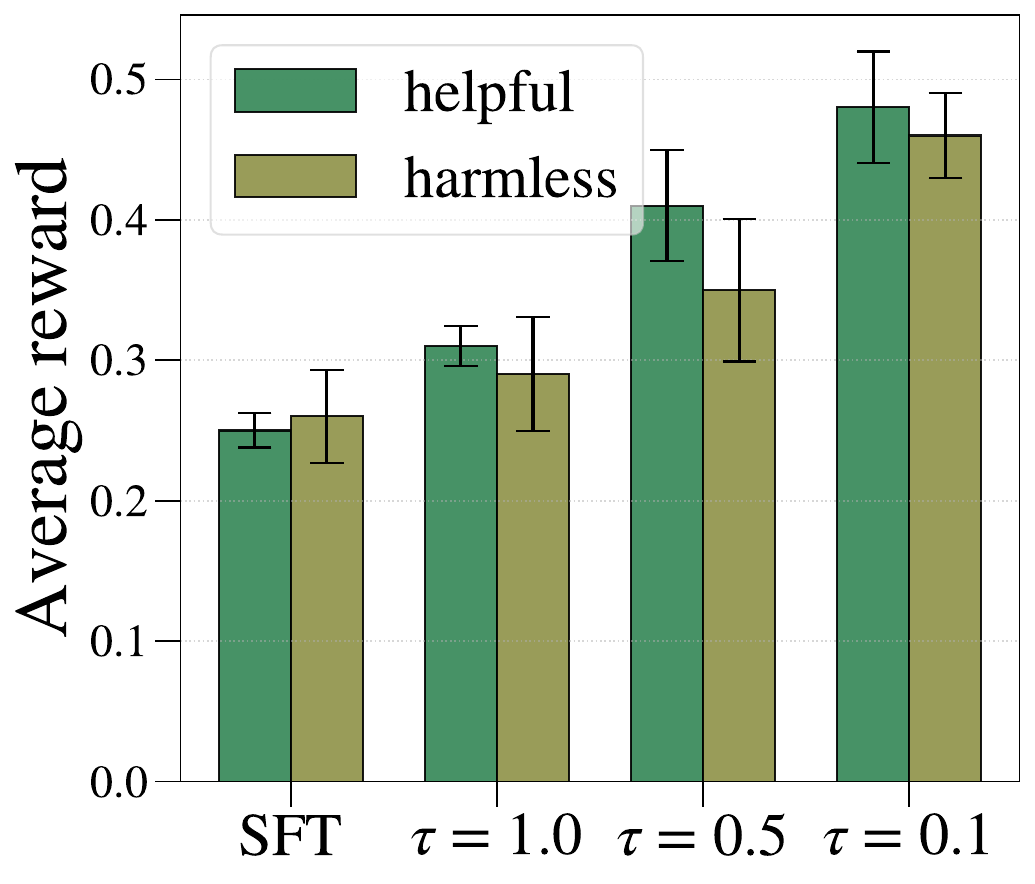}
    }
}
{
    {\includegraphics[height=0.20\textwidth, width=0.23\textwidth]{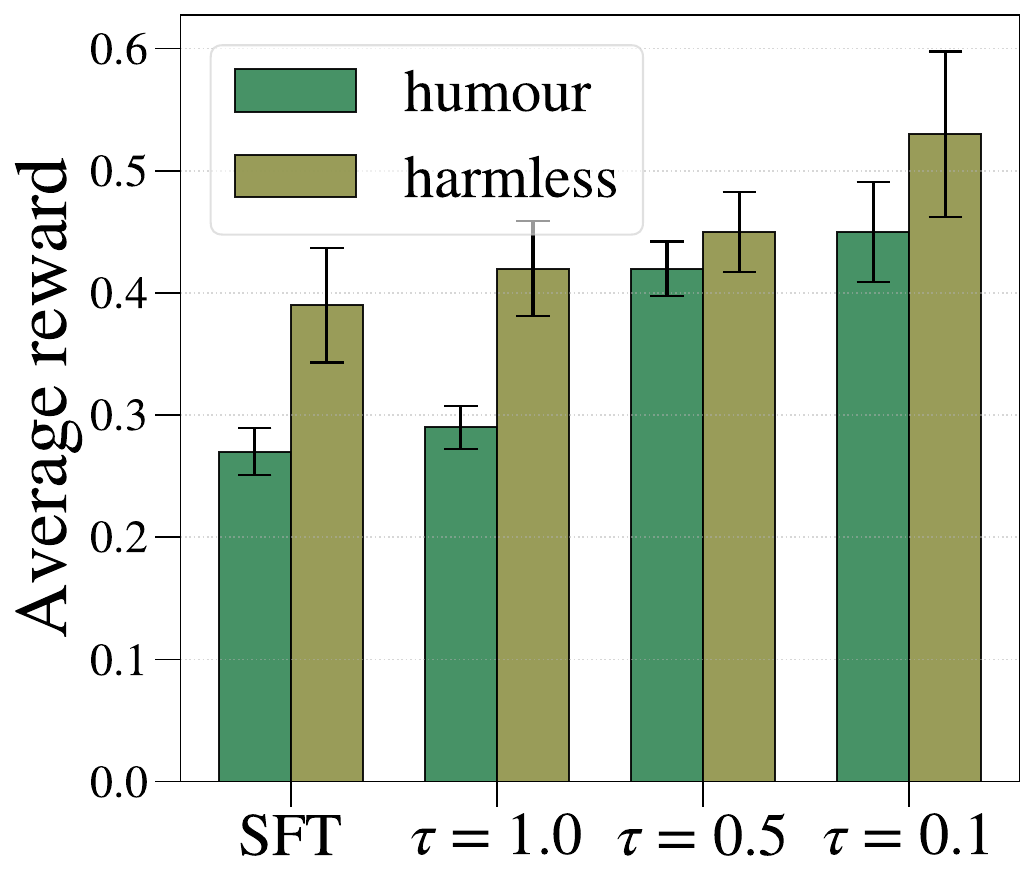}
    }
}
{
    {\includegraphics[height=0.20\textwidth, width=0.23\textwidth]{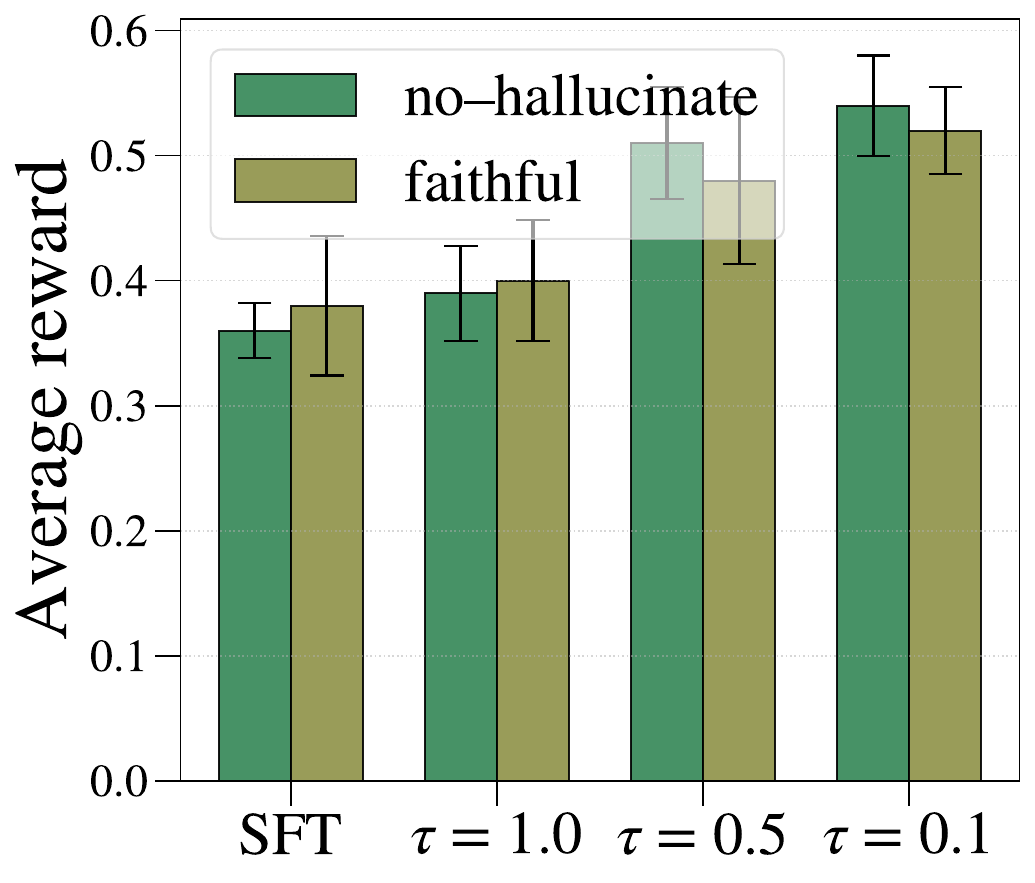}
    }
} 
{
    {\includegraphics[height=0.20\textwidth, width=0.23\textwidth]{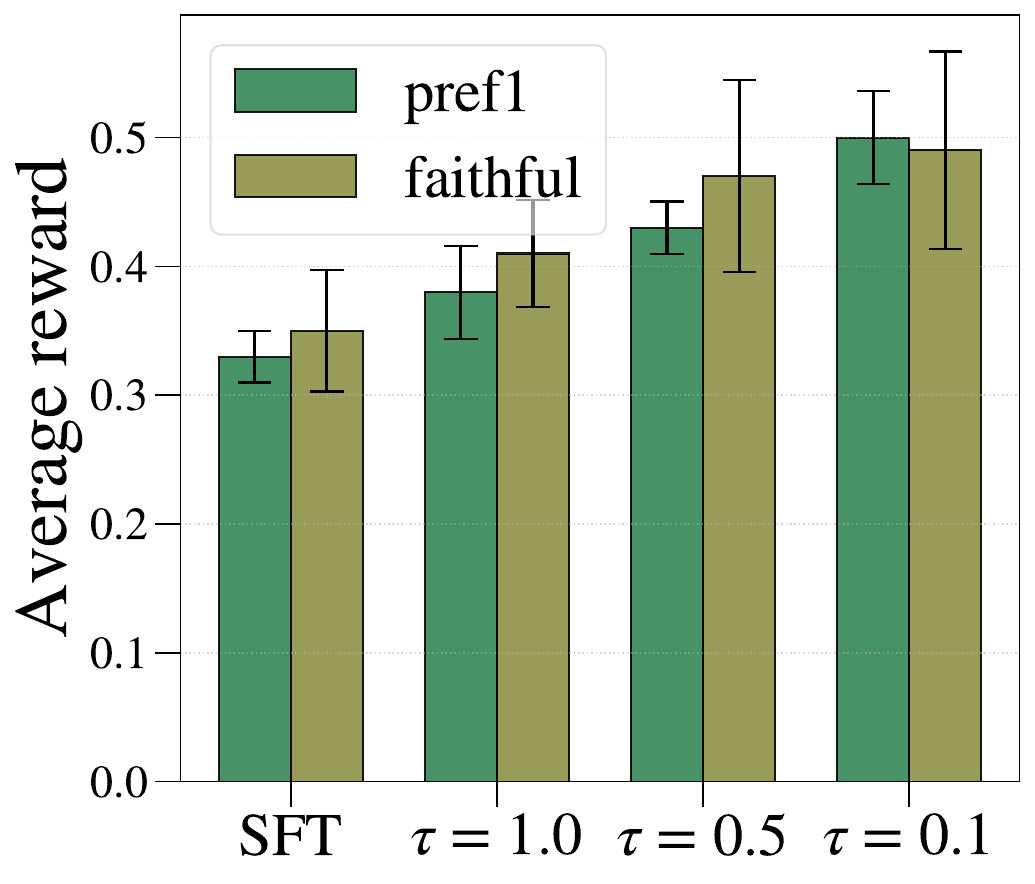}
    }
}
\end{tcolorbox}
\begin{tcolorbox}[width=0.98\textwidth, nobeforeafter, coltitle = black, fonttitle=\fontfamily{lmss}\selectfont, title={Effect of constraint relaxation parameter $\beta$}, halign title=flush center, colback=backg_blue!5, colframe=purple!10, boxrule=2pt, boxsep=2pt, grow to left by=-0.5mm, top=0pt, left=-4pt, right=-4pt, bottom=-1pt]
    \centering
{
    {\includegraphics[height=0.20\textwidth, width=0.23\textwidth]{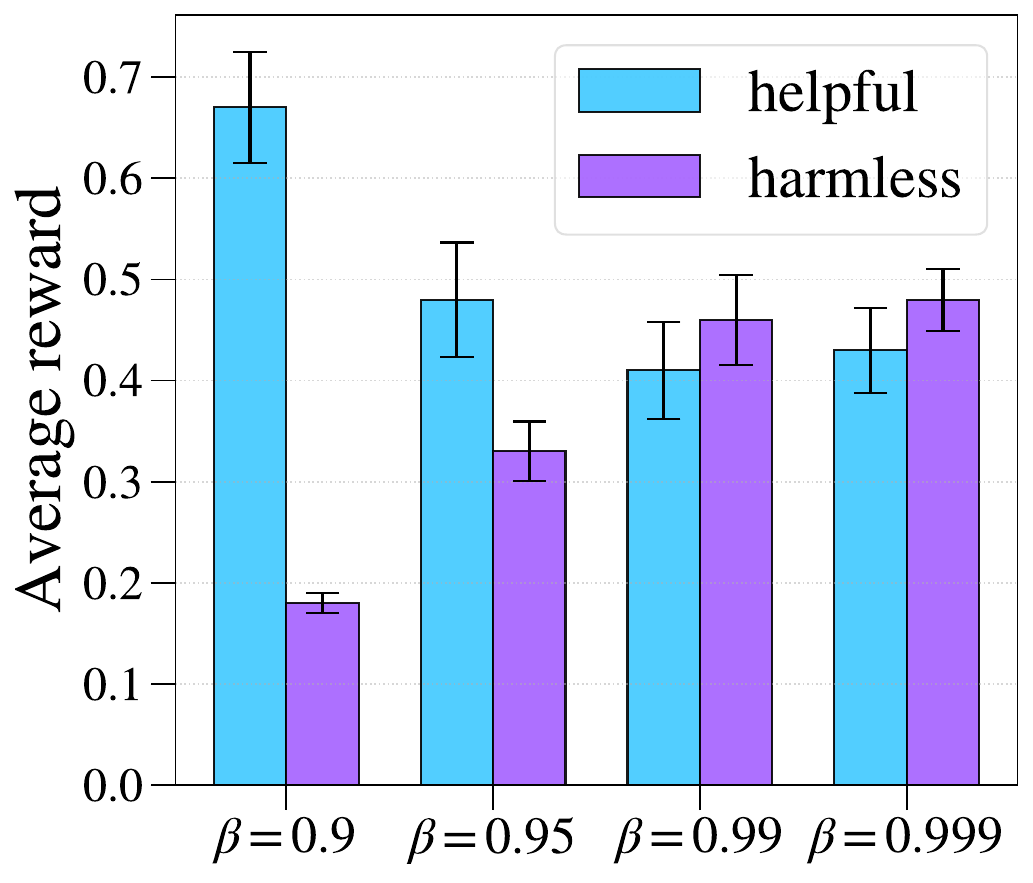}
    }
}
{
    {\includegraphics[height=0.20\textwidth, width=0.23\textwidth]{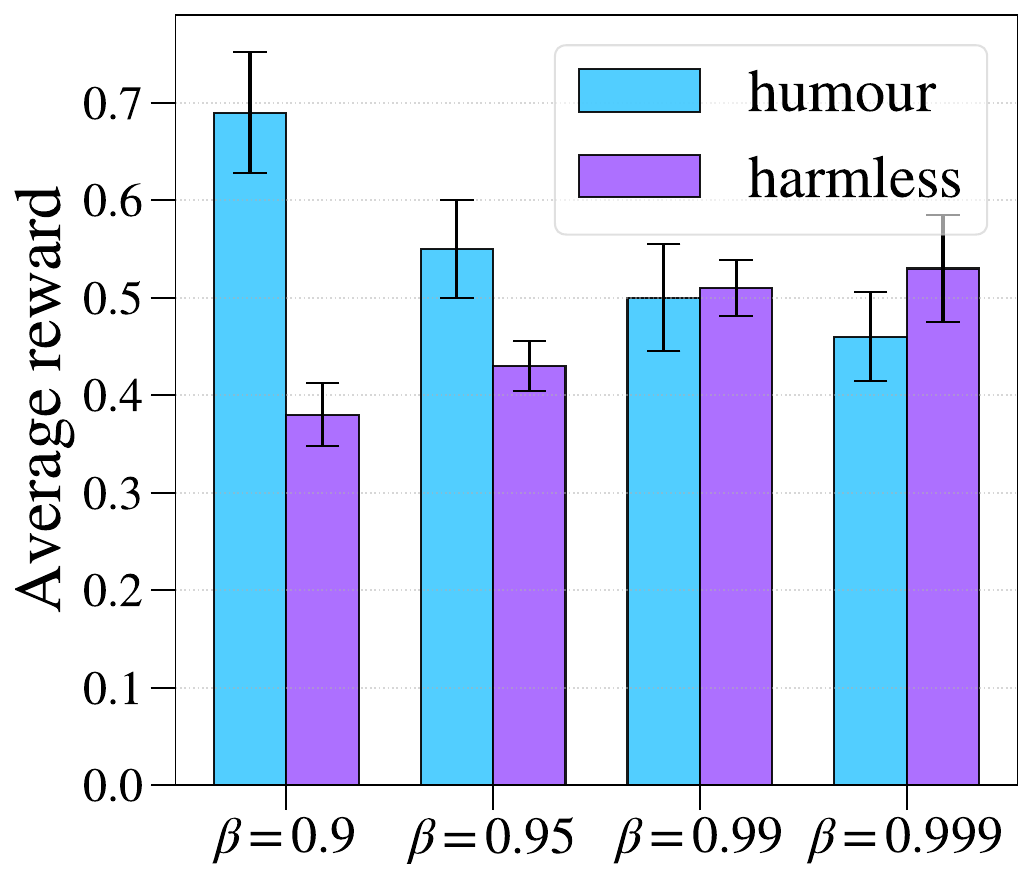}
    }
}
{
    {\includegraphics[height=0.20\textwidth, width=0.23\textwidth]{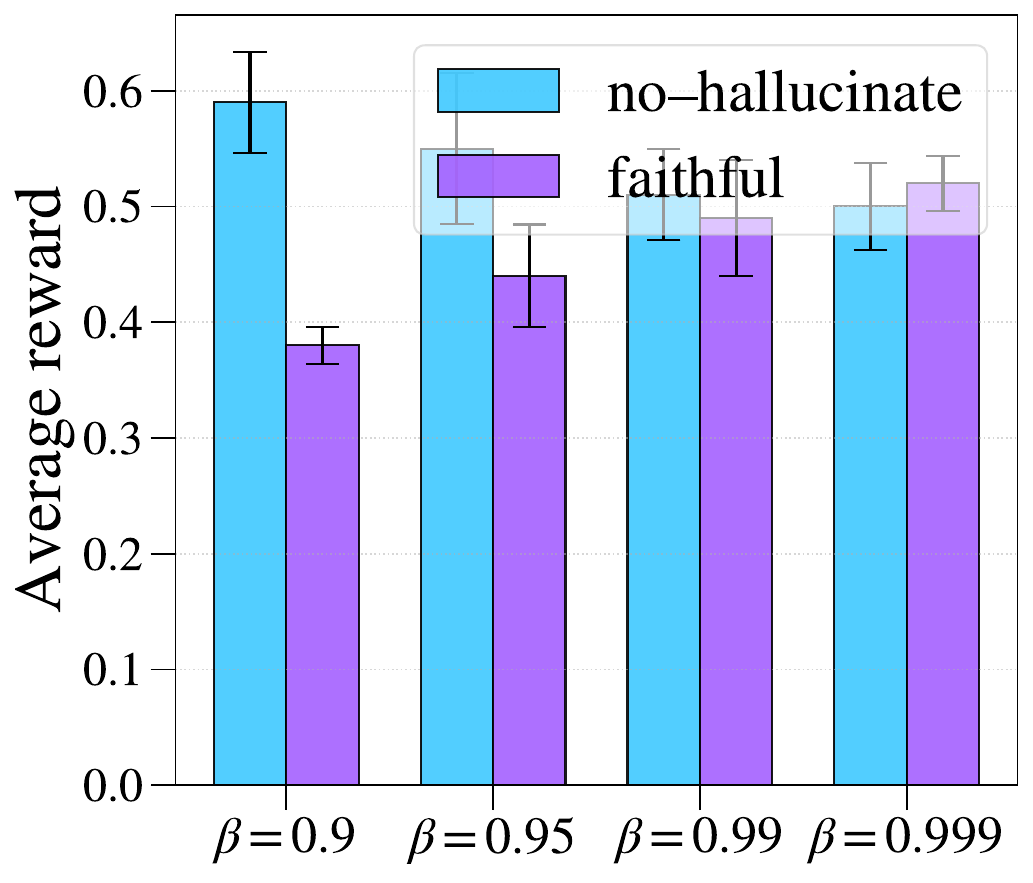}
    }
} 
{
    {\includegraphics[height=0.20\textwidth, width=0.23\textwidth]{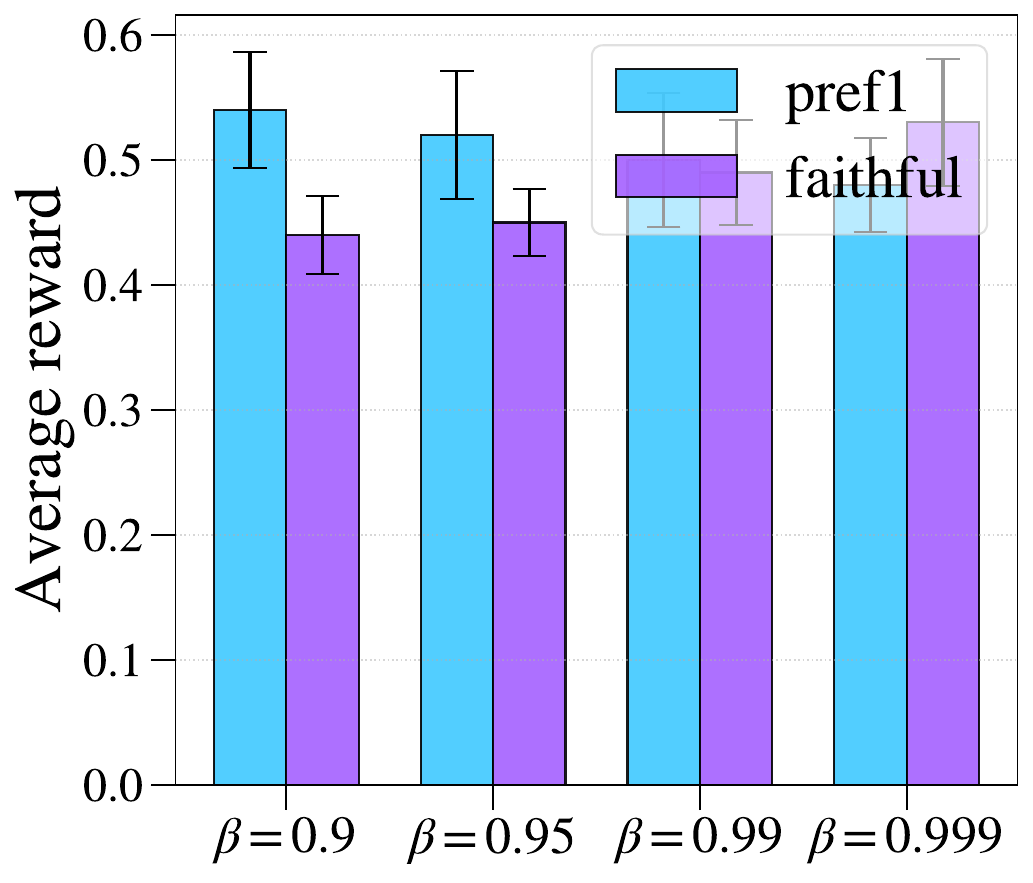}
    }
}
\end{tcolorbox}
\begin{tcolorbox}[width=0.98\textwidth, nobeforeafter, coltitle = black, fonttitle=\fontfamily{lmss}\selectfont, title={Effect of lagged policy updates by $t_{0}$}, halign title=flush center, colback=backg_blue!5, colframe=skyblue!20, boxrule=2pt, boxsep=2pt, grow to left by=-0.5mm, top=0pt, left=-4pt, right=-4pt, bottom=-1pt]
\centering
{
    {\includegraphics[height=0.20\textwidth, width=0.22\textwidth]{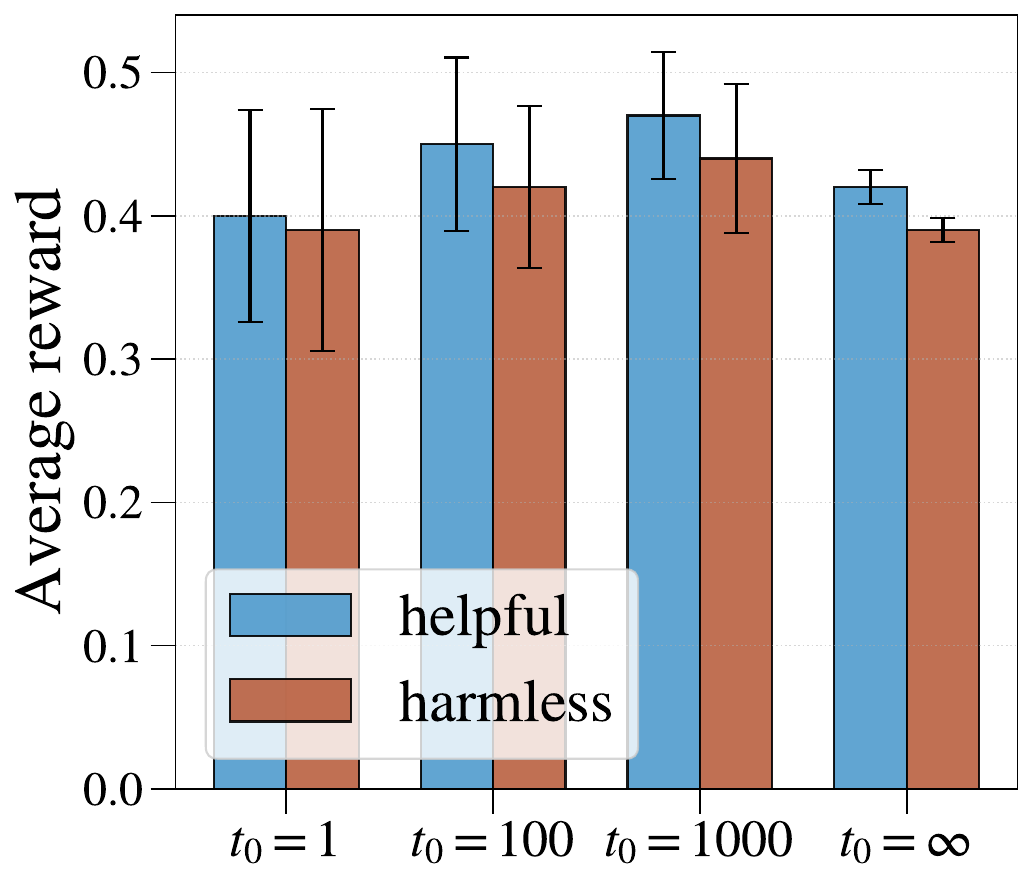}
    }
} \hspace{0.02cm}
{
    {\includegraphics[height=0.20\textwidth, width=0.225\textwidth]{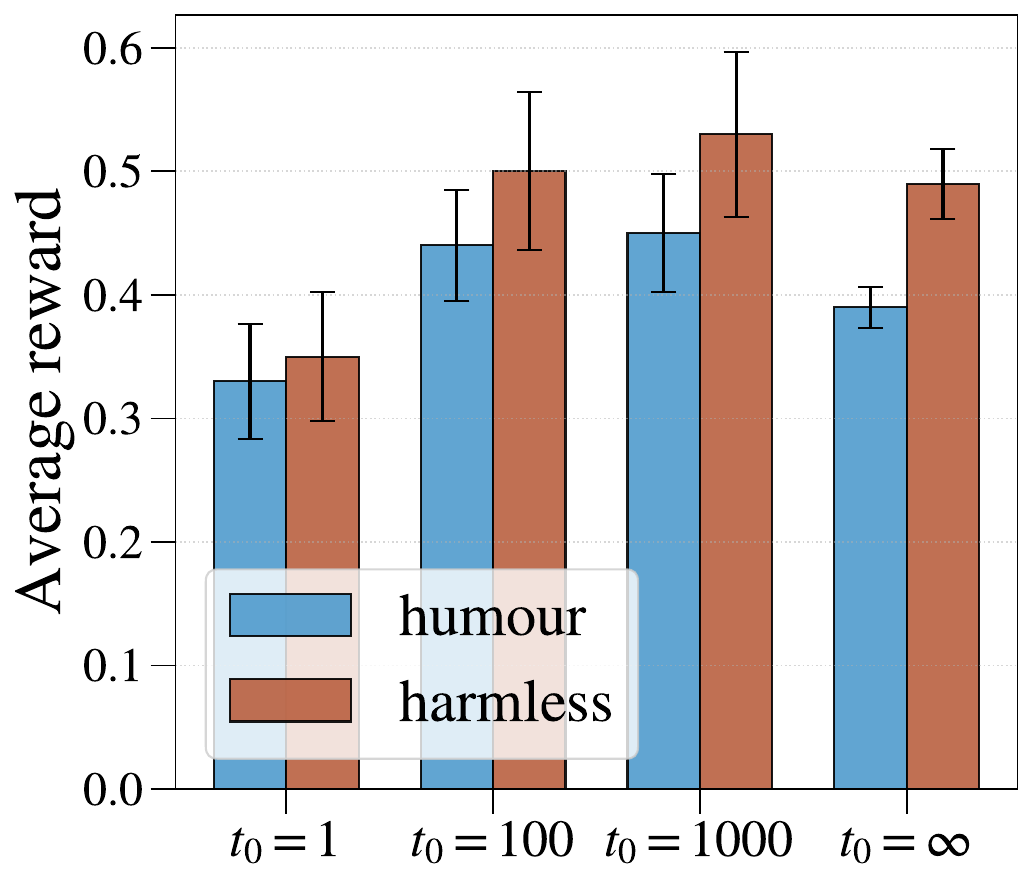}
    }
} \hspace{0.05cm}
{
    {\includegraphics[height=0.20\textwidth, width=0.225\textwidth]{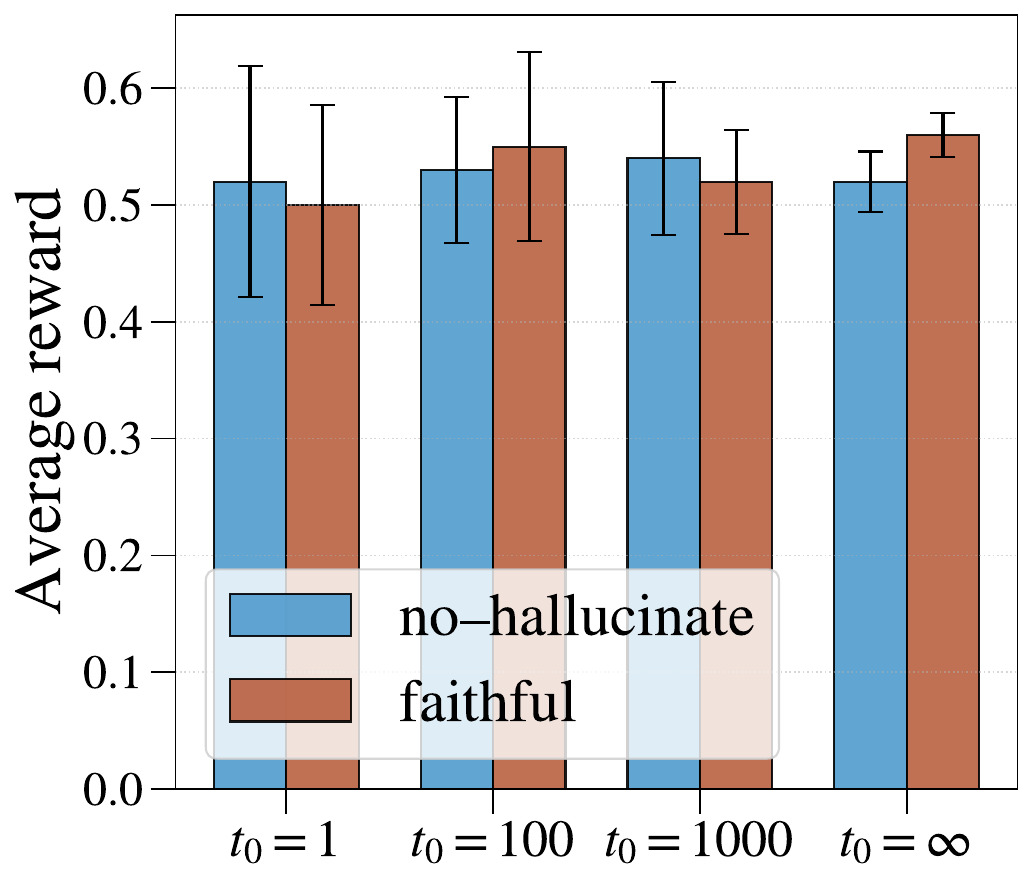}
    }
} \hspace{0.05cm}
{
    {\includegraphics[height=0.20\textwidth, width=0.225\textwidth]{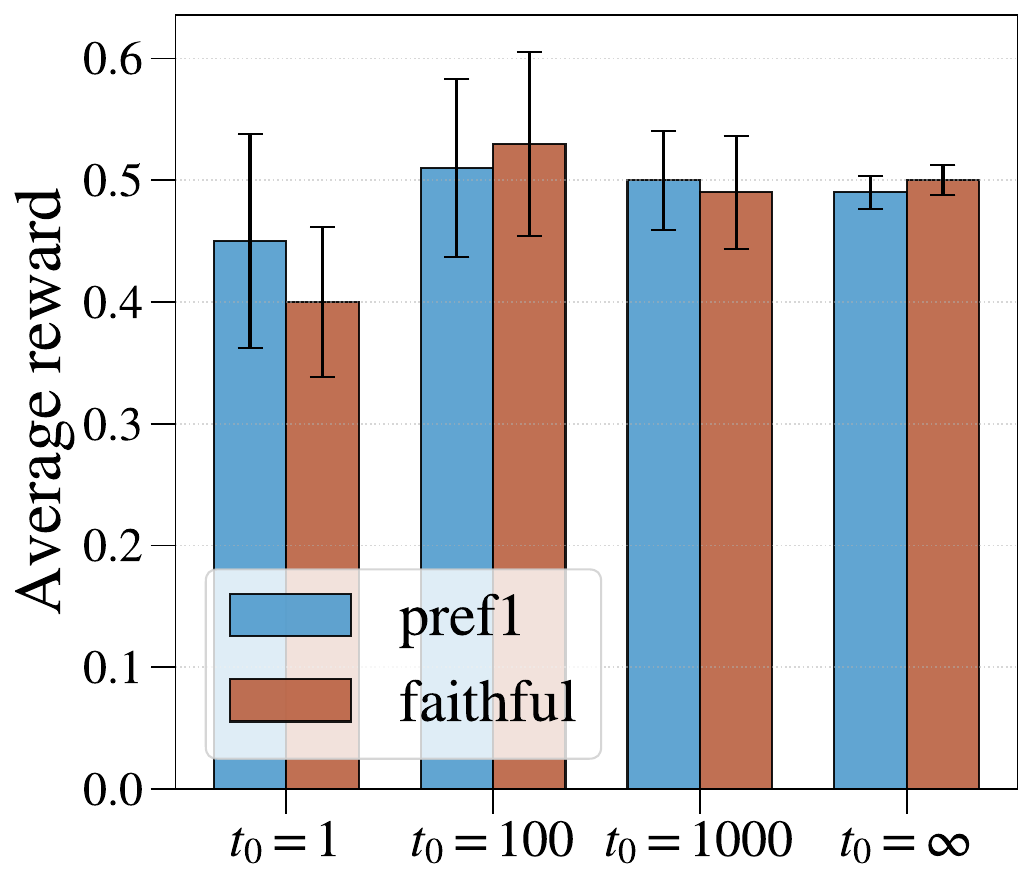}
    }
} 
\end{tcolorbox}
\caption{Normalized rewards of Llama-3.1-8B aligned using \MOPO{MOPO} under varying optimization parameters on the Helpful Assistant and Reddit Summary tasks.}
\label{fig:mopo_ablation}
\vspace{-0.2cm}
\end{figure*}

\subsection{Secondary Evaluation}

We now take a deeper dive into \MOPO{MOPO}'s performance and discuss its dependence on various factors.

\textbf{Verbosity-constrained tasks.} Consider tasks in which the lengths of LLM responses need to be contained in the range $[ l_{\text{low}} , l_{\text{high}} ]$ to control verbosity (while maintaining helpfulness); for example, in summarization tasks \citep{makino2019global}. In this case, the natural choice for reward functions is to directly constrain the rewards $r_{1} ( y | \cdot ) = | y | \geq l_{\text{low}}$ and $r_{2} ( y | \cdot ) = - | y | \geq - l_{\text{high}}$. We illustrate the distributions of the generated response lengths (in tokens) by Zephyr-7b-beta aligned using \MOPO{MOPO}, and report the corresponding helpfulness scores in Figure \ref{fig:length}. We observe that the mean response lengths are in the required range in each case, satisfying the imposed constraints while improving helpfulness.

\textbf{Effect of regularization.} \MOPO{MOPO} employs a \emph{lagged} reference policy to impose a dynamic KL-regularization. This moving baseline stabilizes gradient estimates and prevents overfitting, allowing maximization of multiple reward functions and extension towards the Pareto frontier. In contrast, DPO variants use a static reference and tend to overfit irrespective of regularization \cite{ipo, yang2024rewards}. \MOPO{MOPO}’s adaptive reference updates yield more robust multi-objective alignment as seen in Figure \ref{fig:mopo_ablation}.

\textbf{Sensitivity analysis.} As discussed before, the updates in \MOPO{MOPO} are governed by optimization hyperparameters such as the constraint relaxation factor $\beta$ and the lag interval $t_{0}$. In practice, robustness to these hyperparameters is desirable, as their tuning can significantly affect performance. Figure \ref{fig:mopo_ablation} shows that \MOPO{MOPO} remains stable and effective even under suboptimal choices of $\beta$ and $t_{0}$.

\section{Conclusion}
\label{sec:chapter5-conclusion}

In this chapter, we introduced \MOPO{MOPO}, an offline, multi-objective constrained optimization algorithm that learns from preference data and maximizes a primary objective while enforcing tunable lower bound constraints on secondary objectives. On synthetic benchmarks \MOPO{MOPO} accurately recovers the true Pareto front. Experiments on real-world datasets show that \MOPO{MOPO} matches or surpasses baselines, and ablation studies prove robustness to hyperparameters. An important future direction is to develop a rigorous theoretical analysis of \MOPO{MOPO} to provably advance the field of RLHF.


\clearpage
\newpage

\section{Appendix}
\label{sec:chapter5-appendix}

\begin{definition}
\label{def:kl}
For any two policies $P,Q \in \Delta_{\Ycal}^{\Xcal}$ such that \texttt{Supp}$(P)$ = \texttt{Supp}($Q$), their $\KL$ divergence is defined as:
$$
\KL \left(P \, || \, Q \right)= \E{x \sim \nu \\ y \sim P(\cdot | x)} \left[ \log \left(\frac{P(y \given x)}{Q(y \given x)} \right) \right] \; .
$$
\end{definition}

\subsection{A Detailed Motivating Example}
\label{appendix:motivating_example}

In this example, we empirically demonstrate the necessity of \emph{principled} multi-objective optimization methods that account for multi-dimensional preferences. We benchmark various approaches for multi-objective alignment and show that existing state-of-the-art techniques consistently fail to reach the Pareto front. To ensure clarity, we conduct experiments on synthetic datasets where the true Pareto front is known, allowing for precise evaluation of alignment quality.

Now define the input space $\Xcal := \Ucal [0,1]$ and output space $\Ycal := \Ucal [0,1]$. For input $x \in \Xcal$ and outputs $y, y' \in \Ycal$, the Bradley-Terry preference model $\mathsf{BT}(r(\cdot))$ \citep{bradleyterry} w.r.t. to a reward model $r(\cdot)$ provides preference $z$ as,

\small
\begin{equation}
    \Pr(z = y \,  \given  \, x, y, y') = r(x,y) / \left( r(x,y) + r(x,y') \right)
\label{eq:appendix_bt_model}
\end{equation}
\normalsize

\paragraph{Reward models.} For $(x,y) \in \Xcal \times \Ycal$, we consider two pairs of reward functions $r : \Xcal \times \Ycal \to \Rbb$:
\begin{align}
\label{eq:appendix_reward_models}
    r_{1}^{A}(x,y) = e^{x} + \sqrt{y} - y \quad \text{and} \quad r_{2}^{A}(x,y) = - \sin(x) - y^{2}. \nonumber \\ 
    r_{1}^{B}(x,y) = (x+y)^{2} \quad \text{and} \quad r_{2}^{B}(x,y) = \log  (\frac{1+x}{1+y} ). 
\end{align} 

\paragraph{Dataset construction.} For arbitrary $N \in \Nbb$, generate $x_{i}, y_{i}, y_{i}' \sim \Ucal [0,1]$ for $i \in [N]$. Let $z_{i}^{(1)} \sim \mathsf{BT}(r_{1})$ and $z_{i}^{(2)} \sim \mathsf{BT}(r_{2})$ with $z_{i}^{(1)}, z_{i}^{(2)} \in \{y_{i}, y_{i}'\}$. We now construct four datasets: (i) \Done{$\Dcal_{1}$} = $ \{ (x_{i}, y_{i}, y_{i}', z_{i}^{(1)}  ) \}_{i=1}^{N}$ incorporating preferences w.r.t. reward model $r_{1}(\cdot)$ \textit{only}, (ii) \Dtwo{$\Dcal_{2}$} = $ \{ ( x_{i}, y_{i}, y_{i}', z_{i}^{(2)}  )  \}_{i=1}^{N}$ incorporating preferences w.r.t. reward model $r_{2}(\cdot)$ \textit{only}, (iii) \Djoint{$\Dcal_{J}$} = $ \{  (x_{i},y_{i},y_{i}',z  ) : z = z_{i}^{(1)} = z_{i}^{(2)}  \}_{i=1}^{N}$ incorporating preferences only if they are consistent with reward models $r_{1}(\cdot)$ \textit{and} $r_{2}(\cdot)$, and (iv) \Dcombined{$\Dcal_{C}$} = $ \{  (x_{i}, y_{i}, y_{i}', z_{i}^{(C)}  ) : z_{i}^{(C)} \sim \mathsf{BT}  (w r_{1} + (1-w) r_{2}  )  \}_{i=1}^{N}$ for some $w \in [0,1]$, which incorporates preferences based on some convex weighting of both reward models.

\begin{figure*}[ht]
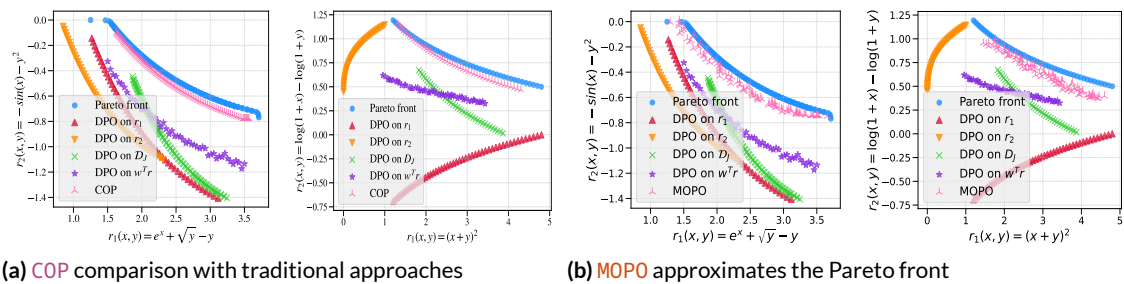

\centering
\subfloat[\COP{COP} comparison with traditional approaches]{{
    \includegraphics[height=0.21\textwidth, width=0.23\textwidth]{chapter5/images/dpo_comparison_set_1.pdf}
}
{
    \includegraphics[height=0.21\textwidth, width=0.23\textwidth]{chapter5/images/dpo_comparison_set_2.pdf}
} 
}
\subfloat[\MOPO{MOPO} approximates the Pareto front]{
{
    \includegraphics[height=0.21\textwidth, width=0.23\textwidth]{chapter5/images/mopo_dpo_comparison_set_A.pdf}
}
{
    \includegraphics[height=0.21\textwidth, width=0.23\textwidth]{chapter5/images/mopo_dpo_comparison_set_B.pdf}
}
}
\caption{Illustration of how a \COP{COP} approach, and hence \MOPO{MOPO}, achieves Pareto-optimal alignment in comparison with DPO on \Done{$\Dcal_{1}$} , \Dtwo{$\Dcal_{2}$} , \Djoint{$\Dcal_{J}$} , and \Dcombined{$\Dcal_{C}$} under two sets of reward models.}
\label{fig:appendix_dpo_comparison}
\end{figure*}

Another approach of solving the multi-objective preference problem is a constrained optimization approach \COP{\small COP}, where we can solve for the optimal policy as $\pi_{\text{\COP{\small COP}}}(x) = \argmax_{y} \; r_{1}(x,y)$,  $\suchthat r_{2}(x,y) \geq b$ for some $b \in \Rbb$. Now, given the four datasets and the constrained optimization approach, we wish to compare learning the optimal policy as described by a Pareto frontier in the $(r_{1}, r_{2})$ space. We train a neural network policy with DPO \cite{rafailov2023direct} for each of the four datasets, and solve a constrained optimization problem for the \COP{\small COP} approach. See Figure \ref{fig:appendix_dpo_comparison}(a) for empirical results under reward model sets $A$ (left) and $B$ (right).

While it is somewhat trivial to see why learning from \Done{$\Dcal_{1}$} and 
\Dtwo{$\Dcal_{2}$} alone yields suboptimal rewards, the case for the jointly preferred dataset \Djoint{$\Dcal_{J}$} and a convex weighted reward model dataset \Dcombined{$\Dcal_{C}$} is not obvious. The issue with \Djoint{$\Dcal_{J}$} is that it only contains samples where $r_{1}$ and $r_{2}$ agree, effectively discarding all points that exhibit a meaningful trade-off between the two objectives. This results in a sparse and biased preference signal that does not span the entire Pareto front. In contrast, \Dcombined{$\Dcal_{C}$} encodes preferences with respect to a \emph{single} scalarized reward model, which inherently biases learning toward one specific convex combination of the objectives. While more sophisticated approaches have been proposed for learning from multi-dimensional preferences - such as RiC \cite{yang2024rewards}, MODPO \cite{zhou2023beyond}, Rewarded Soups \cite{rame2023rewarded}, and SIPO \cite{li2025self} - they remain fundamentally limited in their expressivity. Ultimately, each method relies on learning with respect to a \emph{single} scalarized reward signal of the form $\tilde{r} = f(r_{1}, r_{2})$, where the function $f$ varies across methods. As a result, these approaches do not recover the full structure of the underlying preference landscape and cannot characterize the Pareto front in the multi-objective setting. 

In contrast, constrained optimization (\COP{\small COP}) over $r_{1}$ and $r_{2}$ yields solutions that lie close to the true Pareto frontier. This highlights the need for optimization methods that explicitly account for trade-offs across objectives, rather than collapsing them into a single reward signal, in order to fully leverage multi-dimensional preference data.  \MOPO{MOPO} follows this principle by directly optimizing within the multi-objective space, and empirically achieves solutions that approach the Pareto front as in Figure \ref{fig:appendix_dpo_comparison}(b).

\begin{figure*}[ht]
\noindent
\begin{tcolorbox}[width=0.325\textwidth, nobeforeafter, coltitle = black, fonttitle=\fontfamily{lmss}\selectfont, title={$\tau=0.1$}, halign title=flush center, colback=backg_blue!5, colframe=darkgreen!10, boxrule=2pt, boxsep=2pt, grow to left by=-0.5mm, top=0pt, left=-4pt, right=-4pt, bottom=-1pt]
\centering
{
    \includegraphics[scale=0.25]{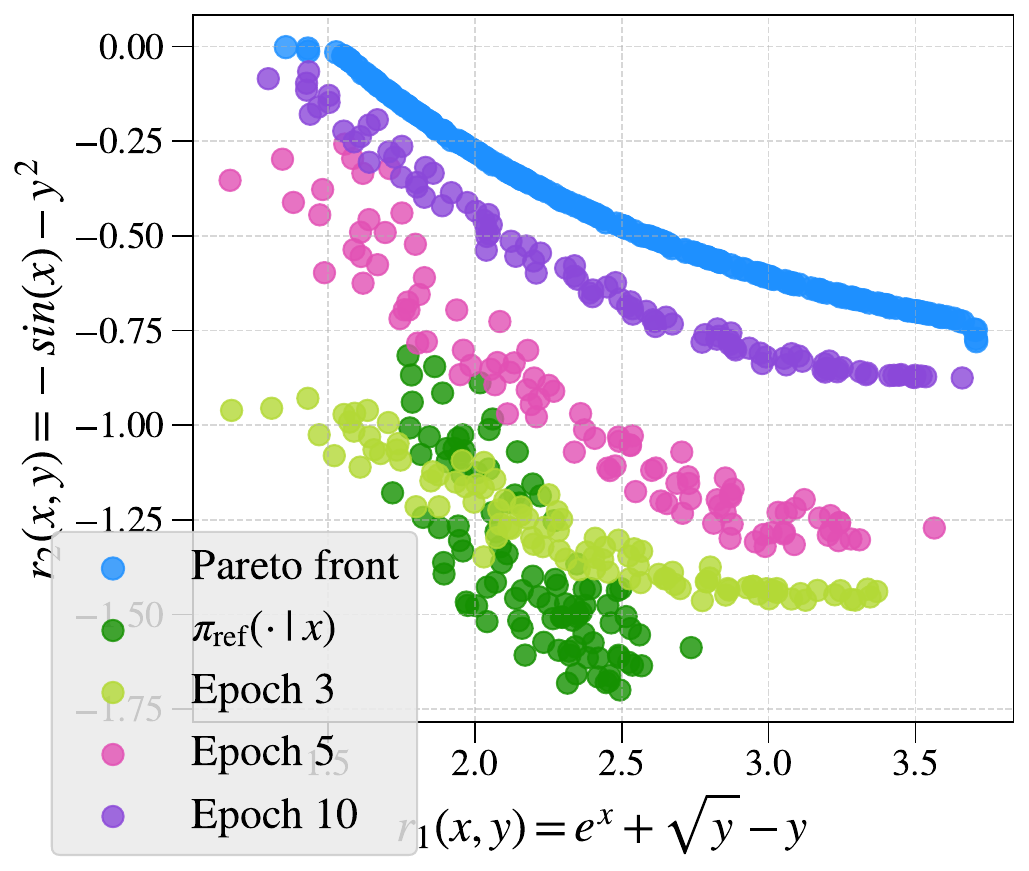}
}
{
    \includegraphics[scale=0.25]{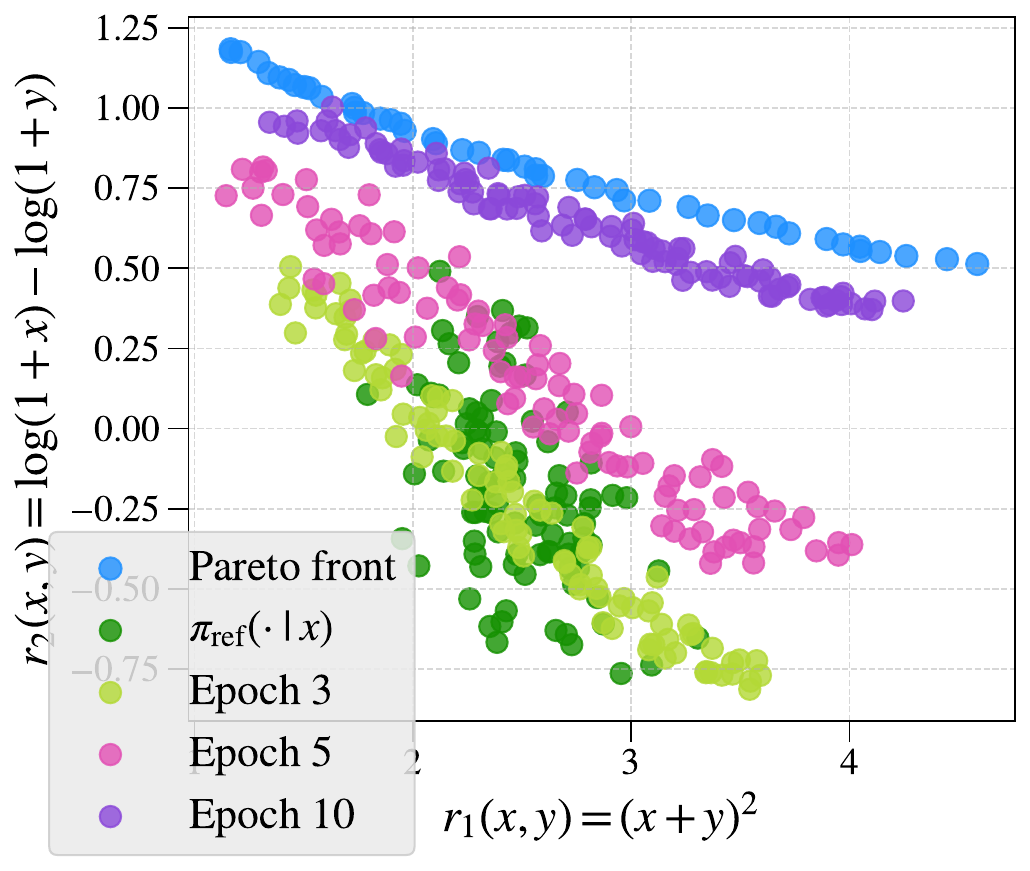}
}
\end{tcolorbox}
\begin{tcolorbox}[width=0.325\textwidth, nobeforeafter, coltitle = black, fonttitle=\fontfamily{lmss}\selectfont, title={$\tau=0.5$}, halign title=flush center, colback=backg_blue!5, colframe=purple!10, boxrule=2pt, boxsep=2pt, grow to left by=-0.5mm, top=0pt, left=-4pt, right=-4pt, bottom=-1pt]
    \centering
{
    \includegraphics[scale=0.25]{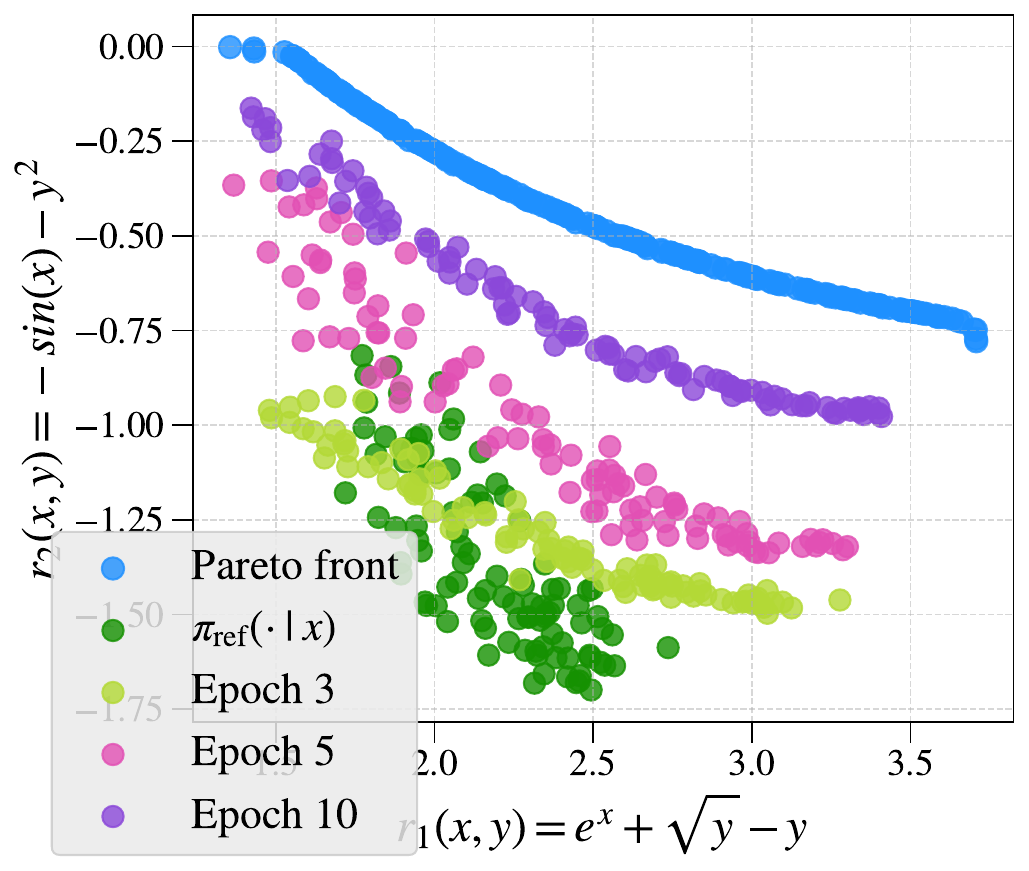}
}
{
    \includegraphics[scale=0.25]{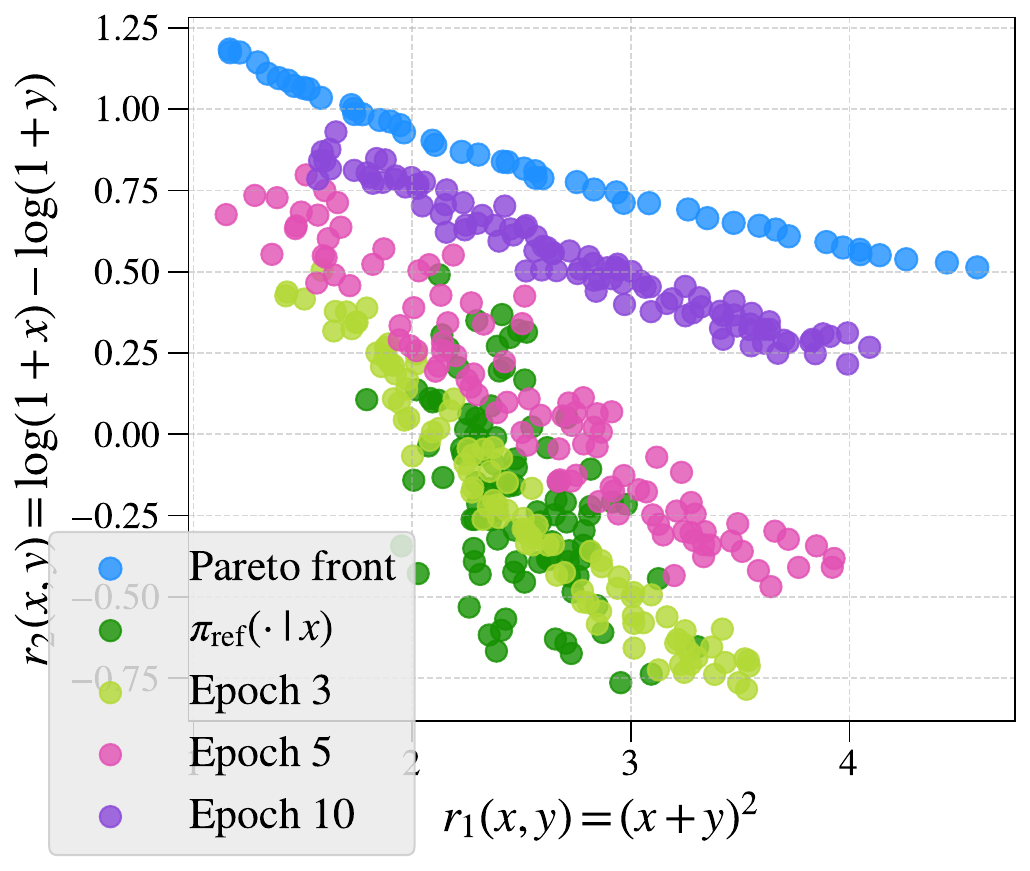}
}
\end{tcolorbox}
\begin{tcolorbox}[width=0.325\textwidth, nobeforeafter, coltitle = black, fonttitle=\fontfamily{lmss}\selectfont, title={$\tau=1$}, halign title=flush center, colback=backg_blue!5, colframe=skyblue!25, boxrule=2pt, boxsep=2pt, grow to left by=-0.5mm, top=0pt, left=-4pt, right=-4pt, bottom=-1pt]
    \centering
{
    \includegraphics[scale=0.25]{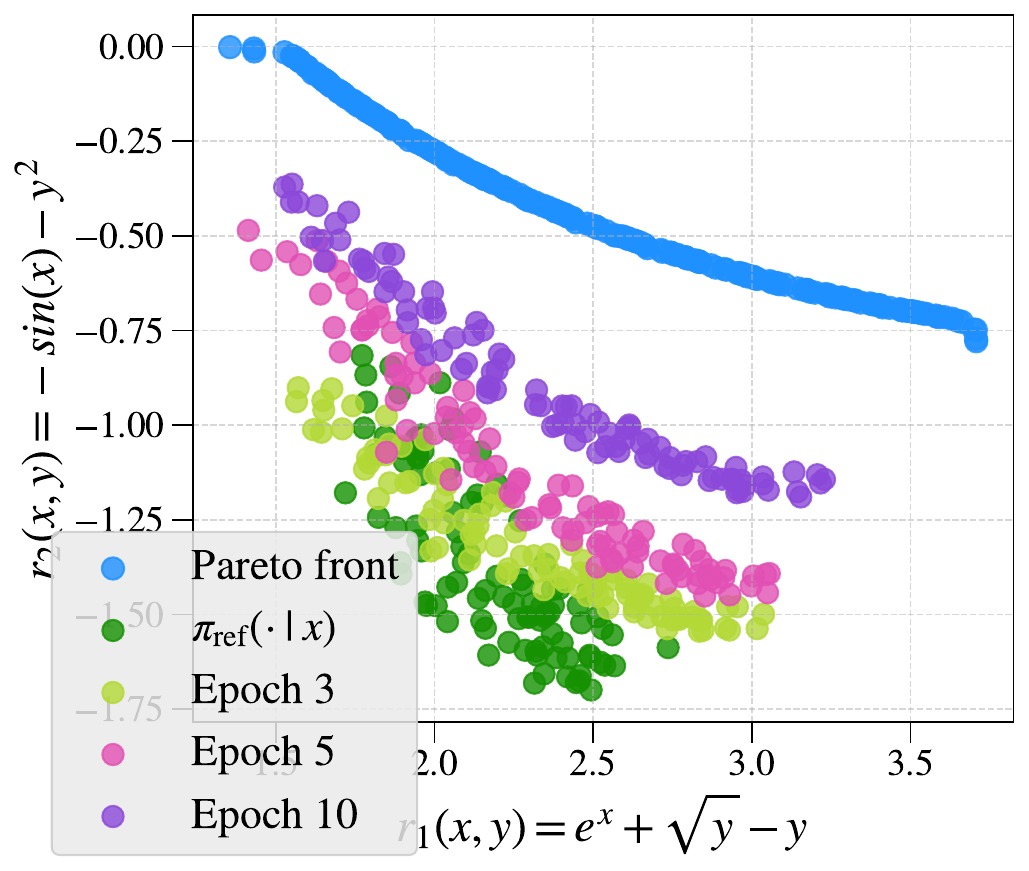}
}
{
    \includegraphics[scale=0.25]{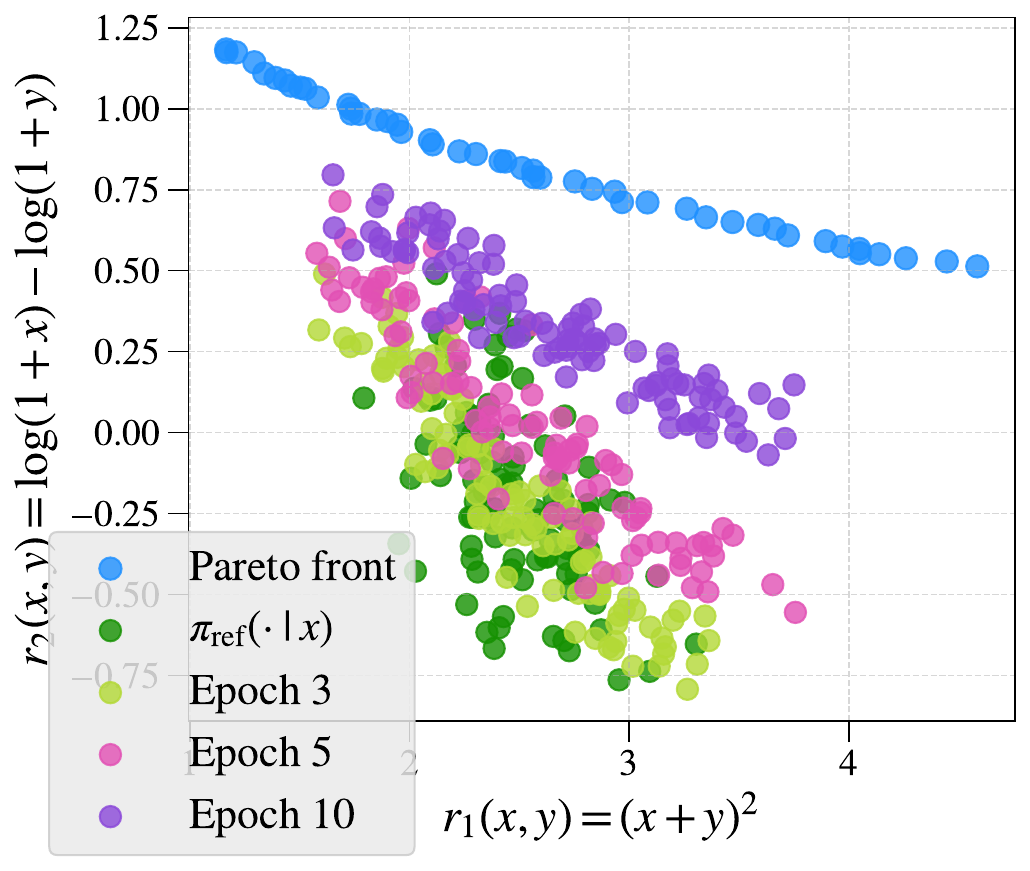}
}
\end{tcolorbox}
\caption{Comparison of the $\KL$-regularized policy learned using \MOPO{MOPO} with the reference policy $\piref$ and the Pareto frontier, visualized in the reward space for two reward model pairs, $A$ (top) and $B$ (bottom).}
\label{fig:mopo_training_pareto}
\end{figure*}

We further empirically validate the correctness of \MOPO{MOPO} and consider whether it is able to generalize and regularize effectively w.r.t. the reference policy. See Figure \ref{fig:mopo_training_pareto} for comparison of the policy learned through \MOPO{MOPO} under various regularization values. We observe that even with an uninformed $\piref$, \MOPO{MOPO} is able to push toward the Pareto frontier and is limited only by the strength of regularization.

\begin{remark}
    Note that reward models are only used for evaluation, and are not assumptions or requirements to finetune policies using \MOPO{MOPO}. \MOPO{MOPO} learns strictly from preference data, without: (i) assuming the existence of a mapping from preferences to pointwise rewards, and (ii) learning this mapping (reward model) from preference data.
\end{remark}

\subsection{Sub-optimality of baselines.}
\label{sec:suboptimalbaselines}

To begin, we list some common divergence metrics that have been used in literature to characterize two probability distributions. Then, we follow with some definitions before listing the main proofs for the sub-optimality of baselines.

\subsubsection{Divergence measures and closed-form policies}
\label{sec:divergence}
We acknowledge that commonly used $f$-divergence measures have been introduced in \cite{fDPO, shi2024decoding} and show them here for completeness:

\begin{center}
\begin{tabular}{lllc}
\toprule
Divergence measure & $f(x)$ & $\nabla f(x)$ & \barrierf \\
\midrule
Reverse KL-divergence & $x\log x$ & $\log x+1$ &\ding{52} \\
Forward KL-divergence & $-\log x$ & $-1/x$ &\ding{52}\\
JSD & $x\log x-(x+1)\log \frac{x+1}{2}$ & $\log\frac{2x}{1+x}$ &\ding{52}\\
$\alpha$-divergence & $\frac{x^{1-\alpha}-(1-\alpha)x-\alpha}{\alpha(1-\alpha)}$ & $(1-x^{-\alpha})/\alpha$&\ding{52}\\
Jeffery divergence & $x\log x-\log x$ & $\log x-\frac{1}{x}+1$&\ding{52}\\
Total Variation & $\vert x-1\vert/2$ & $\operatorname{sgn}(x-1)/2$&\ding{55}\\
Chi-squared & $(x-1)^2$ & $2(x-1)$&\ding{55}\\
\bottomrule
\end{tabular}
\end{center}
Here we show the optimal sampling policies for multi-objective w.r.t. these divergence measures:
\begin{center}
\begin{tabular}{ll}
\toprule
Divergence measure & Optimal policy \\
\midrule
Reverse KL-divergence & $\left(\prod_{i=1}^K\pi_i(y\vert x)^{w_i}\right)\cdot \exp(-Z(x))$ \\
Forward KL-divergence & $\piref(y\vert x)\cdot\left(Z(x)+\sum_{i=1}^K \frac{w_i\piref(y\vert x)}{\pi_i(y\vert x)}\right)^{-1}$ \\
JSD & $\piref(y\vert x)\cdot \left(-1+\exp(Z(x))\prod_{i=1}^K\left(\frac{\piref(y\vert x)}{\pi_i(y\vert x)}+1\right)^{w_i}\right)^{-1}$ \\
$\alpha$-divergence & $\piref(y\vert x)\cdot\left(\alpha Z(x)+\sum_{i=1}^K w_i\left(\frac{\piref(y\vert x)}{\pi_i(y\vert x)}\right)^\alpha\right)^{-\frac{1}{\alpha}}$ \\
\bottomrule
\end{tabular}
\end{center}

\subsubsection{Definitions}
We first begin with some definitions. 

\begin{definition}[$f$-divergence~\cite{fdivergence,fdivergencecsiszar1, fdivergencecsiszar2}]
\label{def:fdivergence}
For probability measures $P$ and $Q$, let $\mu$ be a dominating measure of $P$ and $Q$ (\textit{i.e.} $P,Q\ll \mu$), and let $p,q$ be the Radon-Nikodym derivative~\cite{probtextbook} $\frac{dP}{d\mu}$, $\frac{dQ}{d\mu}$ respectively. For simplicity, here we assume $q>0$ almost surely. Then $f$-divergence from $P$ to $Q$ is defined as
\begin{align*}
    I_f(p\Vert q):=\int qf\left(\frac{p}{q}\right)d\mu\;,
\end{align*}
where $f$ is convex on $\mathbb R_+$, satisfying $f(1)=0$. Most useful divergence measures are included in $f$-divergences, and the commonly used ones and corresponding $f$ are introduced in Appendix~\ref{sec:divergence}.
\end{definition}

\begin{definition}[Barrier function~\cite{convexoptimization}]
\label{def:barrierf}
Given conditions satisfied in Definition \ref{def:fdivergence}, if additionally $0\notin\operatorname{dom}(\nabla f)$, then $f$ is a \barrierf. If a \barrierf $f$ is continuously differentiable and strongly convex on $\mathbb R_+$, then $f$ is a strongly convex and smooth barrier function~(abbreviated as \sbarrierf).
\end{definition}

\begin{definition}[Expected calibration error \cite{guo2017calibration,fDPO}]
\label{def:ece}
Denote the ground truth distribution as $\mathbb P$, context as $X$ and response as $Y$. The expected calibration error of a stochastic policy $\pi$ is defined as
\begin{align*}
    \operatorname{ECE}(\pi):=\underset{\subxy}{\mathbb E} \big\vert\mathbb P(Y=y\vert X=x)-\pi(y\vert x)\big\vert\;.
\end{align*}
\end{definition}

\begin{hypothesis}[Reducible reward misspecification \cite{modelsoup,rame2023rewarded,personalizedsoup}]
\label{def:hypo}
    Let $\theta_{k}$ be the parameter of the optimal policy for objective value $J_{k}$, $\forall k \in [K]$, and $\theta^{\star}_{w}$ be the parameter of the optimal policy for the interpolated objective $\sum_{k=1}^{K} w_k \cdot J_{k}\;$, then this hypothesis claims that
    \begin{align*}
        \theta^*_w\in\left\{\sum_{k=1}^{K} \lambda_{k} \cdot \theta_{k},\lambda \in \Delta^{K-1} \right\} \; , \; \forall w \in \Delta^{K-1}\;.
    \end{align*}
\end{hypothesis}

Extending the results of \cite{fDPO} to the  multi-objective setting, we prove the necessity of $f$ being barrier functions to find an optimal policy $\pi^\star$ for multi-objective alignment. We refer the reader to \cite{shi2024decoding} for a complete discussion.

\begin{restatable}[]{theorem}{thmwontwork}
\label{thm:wontwork}
    If $f$ is not a \barrierf, then for $ \forall C \in \mathbb R_+$, $ N\in\mathbb Z_{\ge 4}$, $K \in \mathbb Z_{\ge 2}$, $\mathcal Y=\{y_i\}_{i=1}^N$, any multi-objective decoding or merging algorithm $\mathcal A:\mathcal S^{K+1}\times \Delta^{K-1}\rightarrow  \mathcal S$, there exists a reference policy $\piref$, policies $\{\mathcal \pi_i\}_{i=1}^K$ and $\pi'$, reward functions $\{\mathcal R_i\}_{i=1}^K$, preference weightings $w\in \Delta^{K-1}$ and $\tau\in\mathbb R_+$, s.t. $\pi_i$ is the optimal policy for $\mathcal R_i$ w.r.t. $\tau\cdot I_f(\cdot\Vert \piref)$, $\forall i \in [K]$, but
    \begin{align*}
        \underset{y\sim \pi_{\mathcal A,w}}{\mathbb E}\left[\sum_{i=1}^K w_i\mathcal R_i(y)\right]\le \underset{y\sim \pi'}{\mathbb E}\left[\sum_{i=1}^K w_i\mathcal R_i(y)\right]-C\;, \mbox{and }
    \end{align*}
    \begin{align*}
        \underset{y\sim \pi_{\mathcal A,w}}{\mathbb E}\left[\sum_{i=1}^K w_i\mathcal R_i(y)\right]-\tau I_f(\pi_{\mathcal A,w}\Vert \piref)\le \underset{y\sim \pi'}{\mathbb E}\left[\sum_{i=1}^K w_i\mathcal R_i(y)\right]-\tau I_f(\pi'\Vert \piref)-C\;,
    \end{align*}
    where $\pi_{\mathcal A,w}(y):=\mathcal A\big(\piref,\pi_1,\pi_2,\ldots,\pi_K,w\big)(y)\;$.
\end{restatable} 

\begin{remark}[Motivating example]
    Here we provide a motivating example where $f\equiv 0$: let $K=4$, $\mathcal R_1(y_1)=\mathcal R_2(y_2)=1$, $\mathcal R_1(y_2)=\mathcal R_2(y_1)=-1$, $\mathcal R_1(y_{3+k})=\mathcal R_2(y_{3+k})=0$, $\mathcal R_1(y_{4-k})=\mathcal R_2(y_{4-k})=1/2$, where $k\in\{0,1\}$. Then the optimal policy for $\mathcal R_1$ is $\pi_1(y_i):=\delta_{1i}$, for $\mathcal R_2$ is $\pi_2(y_i):=\delta_{2i}$, and for $\mathcal R_1/2+\mathcal R_2/2$ is $\pi^\star(y_i):=\delta_{4-k,i}$. Thus $\pi_{\mathcal A,w}$ cannot fit $\pi^\star$ both for $k=0,1$.
\end{remark}

\begin{proof}
Since $f$ is not a \barrierf, $0\in \text{dom}(\nabla f)$. Now we can define $p:=\underset{x\in [0,N]}{\max}\nabla f(x)$, $q:=\underset{x\in [0,N]}{\min}\nabla f(x)$, $r:=\underset{x\in [0,N]}{\max}f(x)-\underset{x\in [0,N]}{\min}f(x)$, $s:=\frac{N-2}{N-3}\cdot C$. Let $w=(0.5, 0.5,\underbrace{0,\ldots,0}_{N-2})$, and we pick $k=\underset{j\in\{3,4,\ldots, N\}}{\argmin}\;\pi_{\mathcal A,w}(y_j)$. Let $\piref(y_i)=\frac{1}{N}$, $\pi_1(y_i)=\delta_{1i}$, $\pi_2(y_i)=\delta_{2i}$, $\pi_j(y_i)=\frac{1}{N}$ and $\pi'(y_i)=\delta_{ik}$, $\forall i\in [N],\;j\in \{3,4,\ldots, K\}$. And set $\mathcal R_1(y_i)=\begin{cases}2p+2r+2s&i=1\\4q-2p-2r-2s&i=2\\p+q+r+s&i=k\\2q&\text{o/w}\end{cases}$, $\mathcal R_2(y_i)=\begin{cases}4q-2p-2r-2s&i=1\\2p+2r+2s&i=2\\p+q+r+s&i=k\\2q&\text{o/w}\end{cases}$, and $\mathcal R_j\equiv 0$, $\forall j\in \{3,4,\ldots, K\}$.

Let $\tau=1$, then the optimization objective for $\mathcal R_1$ w.r.t. $I_f$ is $J_1(\pi):=\underset{y\sim \pi}{\mathbb E} \left[\mathcal R_1(y)\right]-I_f(\pi\Vert\piref)$, and the Lagrangian dual is \begin{align*}
    \mathcal L_1(\pi):=\sum_{i=1}^N \left(-\mathcal R_1(y_i)\cdot\pi(y_i)+\frac{1}{N}f\left(N\cdot \pi(y_i)\right)\right)+\lambda \left(\sum_{i=1}^N \pi(y_i)-1\right)-\sum_{i=1}^N \mu_i\pi(y_i)\;.
\end{align*}
As the objective is convex and the constraints are affine, we can directly apply the \emph{Karush-Kuhn-Tucker conditions}~\cite{convexoptimization}:
\begin{align}
    \nabla \mathcal L_1(\pi_1^\star)&=0\;,\label{eq:6}\\
    \sum_{i=1}^N\pi_1^\star(y_i)&=1\;,\notag\\
    \pi_1^\star(y_i)&\ge 0\;,\notag\\
    \mu_i^\star&\ge 0\;,\notag\\
    \mu_i^\star\pi_1^\star(y_i)&=0\label{eq:17}\;.
\end{align}
Equation \eqref{eq:6} implies
\begin{align*}
    -\mathcal R_1(y_i)+\nabla f(N\cdot \pi_1^\star(y_i))+\lambda^\star-\mu_i^\star=0\;.
\end{align*}
If $\pi_1^\star(y_1)>0$, we have 
\begin{align*}
    \lambda^\star&=\mathcal R_1(y_1)-\nabla f(N\cdot \pi_1^\star(y_1))\\
    &\ge p+2r+2s\;,
\end{align*}
and then for $\forall j\ne 1\;$,
\begin{align*}
\mu_j^\star&=-\mathcal R_1(y_j)+\nabla f(N\cdot \pi_1^\star(y_j))+\lambda^\star\\
&\ge -p-q-r-s+q+p+2r+2s\\
&= r+s\\
&>0\;.
\end{align*}
Combining it with Equation \eqref{eq:17} yields $\pi_1^\star(y_j)=0$ for $\forall j\ne 1$, which is exactly $\pi_1$. Note that we have
\begin{align*}
    J(\pi_1)\ge 2p+2r+2s-\underset{x\in [0,N]}{\max}f(x)\;.
\end{align*}
For any $\pi'$ with $\pi'(y_1)=0$, we have
\begin{align*}
    J(\pi')&\le p+q+r+s-\underset{x\in[0,N]}{\min}f(x)\\
    &=p+q+2r+s-\underset{x\in[0,N]}{\max}f(x)\\&<J(\pi_1)\;.
\end{align*}
Thus $\pi_1$ is the optimal policy for $\mathcal R_1$ w.r.t. $I_f(\cdot\Vert \piref)$. Similarly, $\pi_2$ is the optimal policy for $\mathcal R_2$ w.r.t. $I_f(\cdot\vert \piref)$. By convexity of $f$, the minimum of $I_f(\pi\Vert\piref)$ is obtained when $\pi=\piref$, and thus $\pi_j$ is the optimal policy for $\mathcal R_j$ w.r.t. $I_f(\cdot\Vert\piref)$, for $\forall j\in\{3,4,\ldots, K\}$. Therefore, all conditions are well satisfied by this construction. Note that
\begin{align}
    \underset{y\sim \pi'}{\mathbb E}\left[\sum_{i=1}^K w_i\mathcal R_i(y)\right]&=p+q+r+s\;.\label{eq:7}
\end{align}

While by the selection of $k$, we have
\begin{align}
    \underset{y\sim \pi_{\mathcal A,w}}{\mathbb E}\left[\sum_{i=1}^K w_i\mathcal R_i(y)\right]\le \frac{(N-3)\cdot 2q+p+q+r+s}{N-2}\;.\label{eq:8}
\end{align}

Comparing Equation \eqref{eq:7} with Equation \eqref{eq:8}, we have
\begin{align*}
    \underset{y\sim \pi_{\mathcal A,w}}{\mathbb E}\left[\sum_{i=1}^K w_i\mathcal R_i(y)\right]&\le \underset{y\sim \pi'}{\mathbb E}\left[\sum_{i=1}^K w_i\mathcal R_i(y)\right]-\frac{N-3}{N-2}s\\
    &=\underset{y\sim \pi'}{\mathbb E}\left[\sum_{i=1}^K w_i\mathcal R_i(y)\right]-C\;.
\end{align*}

Note that $\piref$ is a uniform distribution and both $\pi_{\mathcal A,w},\pi'$ are one-point distributions, thus $I_f(\pi_{\mathcal A,w}\Vert \piref)=I_f(\pi'\Vert \piref)$. We have
\begin{align*}
    \underset{y\sim \pi_{\mathcal A,w}}{\mathbb E}\left[\sum_{i=1}^K w_i\mathcal R_i(y)\right]-I_f(\pi_{\mathcal A,w}\Vert\piref)\le \underset{y\sim \pi'}{\mathbb E}\left[\sum_{i=1}^K w_i\mathcal R_i(y)\right]-I_f(\pi'\Vert \piref)-C\;.\tag*{\qedhere}
\end{align*}
\end{proof}

\subsubsection{Baselines are not Pareto-optimal}

Given the necessity of $f$ being a barrier-function, we now show how parameter-merging paradigm algorithms (\cite{guo2024controllable,yang2024rewards,personalizedsoup, zhou2023beyond}) fail to achieve Pareto-optimality. The optimality of parameter-merging paradigm primarily relies on reduced reward mis-specification hypothesis (see Hypothesis \ref{def:hypo}). The following theorem demonstrates that this hypothesis does not hold for almost all $f$-divergence regularized policies. 

\begin{restatable}[]{theorem}{thmrewardsoup}
\label{thm:rewardsoup}
For any $f$-divergence satisfying one of the following conditions: 
(i) $f$ is not a \barrierf; (ii) $I_f$ is Reverse KL-divergence; (iii) $f$ is a \sbarrierf, with finite roots of
    \begin{align*}
        2\nabla f\left(\frac{3\sqrt{1-2x}}{2\sqrt{1-2x}+\sqrt{x}}\right)-2\nabla f\left(\frac{3\sqrt{x}}{2\sqrt{1-2x}+\sqrt{x}}\right)-\nabla f(3-6x)+\nabla f(3x)\;,
    \end{align*}
$\exists N,K\in \mathbb N$, $\mathcal Y=\{y_i\}_{i=1}^N$, $\tau\in\mathbb R_+$, a neural network $nn=\operatorname{softmax}(h_\theta(z_0))$ where $z_0\in \mathbb R^{n}$ and $h_\theta:\mathbb R^{n}\rightarrow \mathbb R^{N}$ is a continuous mapping, preference weightings $w\in\Delta^{K-1}$, reference policy $\piref$, and the objectives $J_1,J_2,\ldots,J_K$ representing reward functions $\mathcal R_1,\mathcal R_2,\ldots,\mathcal R_K$ w.r.t. $\tau\cdot I_f(\cdot\Vert\piref)$, s.t. Hypothesis \ref{def:hypo} does not hold.
\end{restatable} 

\begin{proof}

(i) If $f$ is not a \barrierf, Hypothesis \ref{def:hypo} does not hold immediately from Theorem \ref{thm:wontwork}.

(ii) If $I_f$ is Reverse KL-divergence, we let $N=3$, $K=3$, and $h_\theta(z_0)=W_{\theta}^{(2)}\sigma\left( W_\theta^{(1)}z_0\right)$, where $\sigma$ is $\text{ReLU}(\cdot)$. We set $\mathcal R_i(y_j)=\delta_{ij}$, $\piref(y_i)=1/3$ for $\forall i,j\in [3]$, $z_0=1$ and $\tau=1$. Then the optimal policies are $W_{\theta_1}^{(1)}=e_1$, 
$W_{\theta_1}^{(2)}=\left(\begin{matrix}100\\000\\000\end{matrix}\right)$ for $\mathcal R_1$ w.r.t. $\KL{\cdot}{\piref}$, $W_{\theta_2}^{(1)}=e_2$, 
$W_{\theta_2}^{(2)}=\left(\begin{matrix}000\\010\\000\end{matrix}\right)$ for $\mathcal R_2$ w.r.t. $\KL{\cdot}{\piref}$, and $W_{\theta_3}^{(1)}=e_3$, $W_{\theta_3}^{(2)}=\left(\begin{matrix}000\\000\\001\end{matrix}\right)$ for $\mathcal R_3$ w.r.t. $\KL{\cdot}{\piref}$. Thus we have $h_{\sum_{j=1}^3\lambda_j\theta_j}(z_0)=\left(\lambda_1^2,\lambda_2^2,\lambda_3^2\right)^\top$. Given $w=(0,1/3,2/3)$, the optimal policy $\pi^\star$ should output $\pi^\star(y_1)=\frac{1}{1+\exp(1/3)+\exp(2/3)}$, $\pi^\star(y_2)=\frac{\exp(1/3)}{1+\exp(1/3)+\exp(2/3)}$ and $\pi^\star(y_3)=\frac{\exp(2/3)}{1+\exp(1/3)+\exp(2/3)}$. Note that
\begin{align*}
    \sqrt{t}+\sqrt{t+1/3}+\sqrt{t+2/3}>1\;,\;\forall t\in\mathbb R_+\;,
\end{align*}
thus there is no solution $\lambda\in\Delta^2,t\in\mathbb R_+$ for $\left(\lambda_1^2,\lambda_2^2,\lambda_3^2\right)^\top=\left(t,t+\frac{1}{3},t+\frac{2}{3}\right)^\top$, \textit{i.e.} there is no $\lambda$ s.t. $
    \operatorname{softmax}\left(h_{\sum_{j=1}^3\lambda_j\theta_j}(z_0)\right)=\big(\pi^\star(y_1),\pi^\star(y_2),\pi^\star(y_3)\big)$, \textit{i.e.} Hypothesis \ref{def:hypo} does not hold.

(iii) If $f$ is a \sbarrierf, with finite roots of
    \begin{align*}
        2\nabla f\left(\frac{3\sqrt{1-2x}}{2\sqrt{1-2x}+\sqrt{x}}\right)-2\nabla f\left(\frac{3\sqrt{x}}{2\sqrt{1-2x}+\sqrt{x}}\right)-\nabla f(3-6x)+\nabla f(3x)\;,
    \end{align*} 
we let $N=3$, $K=2$, $h_\theta(z_0)=W_\theta(z_0)$, $z_0=1$, $\mathcal R_1(y_i)=\delta_{1i}$, $\mathcal R_2(y_i)=\delta_{2i}$ and $\piref(y_i)=1/3$, for $\forall i\in [3]$. The optimal policy for $J_1$ is $\pi_{\theta_1}(y_i)=\frac{1}{3}(\nabla f)^{(-1)}\left(\frac{1}{\tau}\delta_{1i}-Z\right)$, and the optimal policy for $J_2$ is $\pi_{\theta_2}(y_i)=\frac{1}{3}(\nabla f)^{(-1)}\left(\frac{1}{\tau}\delta_{2i}-Z\right)$, where $Z$ is the normalization factor. And these policies can be learned by setting $W_{\theta_i}=\big(\log\pi_{\theta_i}(y_1),\log\pi_{\theta_i}(y_2),\log\pi_{\theta_i}(y_3)\big)^\top$. 

We set $a:=\pi_{\theta_1}(y_1)=\frac{1}{3}(\nabla f)^{(-1)}(\frac{1}{\tau}-Z)$, $b:=\pi_{\theta_1}(y_2)=\pi_{\theta_1}(y_3)=\frac{1}{3}(\nabla f)^{(-1)}(-Z)$. Thus we have 
\begin{align}
    \nabla f(3a)-\nabla f(3b)&=\frac{1}{\tau}\label{eq:11}\;,\\
     a+2b&=1\label{eq:12}\;.
\end{align}

The optimal policy for $w_1\cdot J_1+w_2\cdot J_2$ (see \cite{shi2024decoding} for proof) is 
\begin{align}
    \pi^\star_w(y_i)&=\frac{1}{3}(\nabla f)^{(-1)}\left(-Z_w^{\star}+\frac{w_1}{\tau}\delta_{1i}+\frac{w_2}{\tau}\delta_{2i}\right)\;,\label{eq:9}
\end{align}
where $Z_w^\star$ is the normalization factor. By linearly merging the weights of $\pi_{\theta_1}$ and $\pi_{\theta_2}$, we have
\begin{align}
    \pi_{\lambda_1 \theta_1+\lambda_2\theta_2}(y_i)&=\operatorname{softmax}\left(\lambda_1W_{\theta_1}(z_0)+\lambda_2W_{\theta_2}(z_0)\right)(y_i)\notag\\
    &=\frac{1}{Z_\lambda}\left((\nabla f)^{(-1)}\left(\frac{1}{\tau}\delta_{1i}-Z\right)\right)^{\lambda_1}\left((\nabla f)^{(-1)}\left(\frac{1}{\tau}\delta_{2i}-Z\right)\right)^{\lambda_2}\;,\label{eq:10}
\end{align}
where $Z_\lambda$ is the normalization factor. 

With symmetry, Equation \eqref{eq:9}, \eqref{eq:10} and Hypothesis \ref{def:hypo} indicate that $\pi_{\frac{1}{2} \theta_1+\frac{1}{2}\theta_2}=\pi^\star_{(\frac{1}{2},\frac{1}{2})}$, thus
\begin{align*}
    \frac{1}{3}(\nabla f)^{(-1)}\left(-Z_{(0.5,0.5)}^\star+\frac{1}{2\tau}\right)&=\frac{\sqrt{a}}{2\sqrt{a}+\sqrt{b}}\;,\\
    \frac{1}{3}(\nabla f)^{(-1)}\left(-Z_{(0.5,0.5)}^\star\right)&=\frac{\sqrt b}{2\sqrt{a}+\sqrt b}\;,
\end{align*}
and combining them with Equation \eqref{eq:11} yields
\begin{align}
    2\nabla f\left(\frac{3\sqrt{a}}{2\sqrt{a}+\sqrt{b}}\right)-2\nabla f\left(\frac{3\sqrt{b}}{2\sqrt{a}+\sqrt{b}}\right)&=\nabla f(3a)-\nabla f(3b)\;.\label{eq:13}
\end{align}

Given the condition, the solution set $(a,b)$ to Equation \eqref{eq:12},~\eqref{eq:13} is finite, thus there exists $\tau\in \mathbb R_+$ s.t. Equation \eqref{eq:11} does not hold, implying that Hypothesis \ref{def:hypo} does not hold.
\end{proof}

\subsection{Constraint Threshold Initialization}
\label{appendix:constraintspec}

We begin this section by describing an equivalence between the preference-learning based optimization problem and a reward-learning based optimization problem. Note that \MOPO{MOPO} does not assume this equivalence, and directly works with preference data, and this equivalence is established for analysis only. Following this equivalence discussion, we specify a provable method for setting constraint thresholds $\bm{b}$ such that the optimal solution learned by \MOPO{MOPO} is a Pareto-optimal solution.

For all preference instances in the preference dataset, there exists an underlying, unknown reward model based on which preferences are provided. For all contexts-output pairs $(x,y) \in \Xcal \times \Ycal$, let the reward model for the $k^{th}$ objective be $r_{k}(x,y) \in \Rbb$ for $k \in [K]$.

\paragraph{Preference–reward link.} For every objective $k \in [K]$ there exists a \emph{strictly increasing} function $\phi_{k} : \Rbb \to \Rbb$ and a strictly increasing transfer function $\sigma:\Rbb \to (0,1)$ such that for all contexts $x \in \Xcal$ and actions $y,y' \in \Ycal$ we have, 
\begin{equation}
    p_{k}(y \succ y' \mid x) \;=\; \sigma \bigl(\phi_{k}\bigl(r_{k}(x,y)-r_{k}(x,y')\bigr)\bigr).
\label{as:link}
\end{equation}

Now, with respect to the $k^{th}$ objective, let the expected reward $R_{k}(\pi)$ and preference-based objective value $F_{k}(\pi)$ for a policy be given by,

\begin{align*}
 R_{k}(\pi) &:= \mathbb{E}_{x\sim\nu,\;y\sim\pi(\cdot\mid x)}[r_{k}(x,y)] \;\text{for}\,
    \;k \in [K] , \\
 F_{k}(\pi) &:= \mathbb{E}_{x\sim\nu,\;y\sim\pi(\cdot\mid x),\;y'\sim\mu(\cdot\mid x)}
         \bigl[p_{k}(y\succ y' \mid x)\bigr], \quad k \in [K-1], \\
 F_{K}(\pi) &:= \mathbb{E}_{x\sim\nu,\;y\sim\pi(\cdot\mid x),\;y'\sim\mu(\cdot\mid x)}
         \bigl[p(y\succ y' \mid x)\bigr]
        -\tau\, \KL \bigl(\pi\;\|\;\pi^{\mathrm{ref}}\bigr).
\end{align*}

\begin{lemma}[Order preservation]
\label{lem:orderpreservation}
Under Equation \eqref{as:link} and some $u \in \Rbb$, the mapping $ H_{k}(u) :=    \mathbb{E}_{z\sim\mathrm{Unif}[-u,u]}\bigl[\sigma \circ \phi_{k}(z)\bigr]$ is strictly increasing. Moreover, for every policy $\pi$
$$
    F_{k}(\pi)
    \;=\;
    H_{k} \bigl(R_{k}(\pi)-R_{k}(\mu)\bigr), \qquad k \in [K-1],
$$
so that for any $\pi,\pi'$, we have
$
    R_{k}(\pi)\ge R_{k}(\pi')
    \Longleftrightarrow
    F_{k}(\pi)\ge F_{k}(\pi').
$
\end{lemma}

\begin{proof}
Fix $k$.  By Equation \eqref{as:link}, $\sigma \circ \phi_{k}$ is strictly increasing, hence so is its odd
extension $z\mapsto\sigma \circ \phi_{k}(z)$. For $U:=R_{k}(\pi)-R_{k}(\mu)$ let
$z := r_{k}(x,y)-r_{k}(x,y')$. Because $(y,y') \sim (\pi,\mu)$ are independent, $z$ is symmetrically distributed around $U$ and $z \sim \mathrm{Unif}[U-\delta,U+\delta]$ for some $\delta>0$ that does not depend on $U$.  Taking expectation yields $F_{k}(\pi)=H_{k}(U)$, and strict monotonicity of $H_{k}$ follows from strict monotonicity of $\sigma \circ \phi_{k}$.
\end{proof}

Note there that we introduced $H_{k}(u) = \mathbb{E}_{z\sim\mathrm{Unif}[-u,u]}    \bigl[\sigma \circ \phi_{k}(z)\bigr]$. The uniform law is chosen purely for notational brevity;
the proof requires only that the base distribution be symmetric and shifted by the reward gap
$u = R_{k}(\pi) - R_{k}(\mu)$. Consequently, one may replace $\mathrm{Unif}[-u,u]$ by any symmetric density $\rho_{u}(z) = \rho(z-u)$, and define $H_{k}(u) =
\mathbb{E}_{z\sim\rho_{u}} \bigl[\sigma \circ \phi_{k}(z)\bigr]$. Strict monotonicity of $\sigma \circ \phi_{k}$ then guarantees that the Lemma \ref{lem:orderpreservation} holds. Now, let $\bm{b}\in[0,1]^{K-1}$ be the probability thresholds in \COP{COP}, let $\mathbf{F}_{1:K-1}(\pi) = \left( F_{1}(\pi), \dots, F_{K-1}(\pi) \right)$ and $\mathbf{R}_{1:K-1}(\pi) = \left( R_{1}(\pi), \dots, R_{K-1}(\pi) \right)$, and define component-wise

$$
c_{k} := H_{k}^{-1}(b_{k}) + R_{k}(\mu), \quad k\in[K-1], \quad \bm{c}:=(c_{1},\dots,c_{K-1}).
$$ 

By Lemma \ref{lem:orderpreservation}$, F_{k}(\pi)\ge b_{k}\;\Leftrightarrow\;R_{k}(\pi)\ge c_{k}.$ Hence the preference‐space constrained problem
\[
    \max_{\pi}\,F_{K}(\pi) \suchthat \mathbf{F}_{1:K-1}(\pi) \geq \bm{b}
\tag{\COP{COP}}
\]
is \emph{equivalent} to the reward‐space problem
\[
    \max_{\pi}\, \bigl\{R_{K}(\pi)-\tau\,\mathrm{KL}(\pi\;\|\;\pi^{\mathrm{ref}})\bigr\} \suchthat \mathbf{R}_{1:K-1}(\pi) \geq \bm{c}.
\tag{\COP{COP-R}}
\]

Given this equivalence, we now describe the procedure of setting appropriate constraint thresholds $\bm{c}$ for the \COP{COP-R} problem, following which constraint thresholds $\bm{b}$ for the original \COP{COP} can be obtained element-wise via $c_{k} := H_{k}^{-1}(b_{k}) + R_{k}(\mu)$. This setting of constraint thresholds ensures that the optimal solution of the \COP{COP} problem (Problem \eqref{eq:first-obj}) is also a Pareto-optimal solution. We begin with a definition, following which we state the main result for $\bm{c}_{k}$, which holds for all $k \in [K-1]$.

\begin{definition}[Insertion Index]
\label{def:insertion-index}
Let $\mathsf{P}_{k} \;=\; \bigl(\mathsf{P}_{k}(0),\,\mathsf{P}_{k}(1),\,\dots,\,\mathsf{P}_{k}(M-1)\bigr)$ be an ascending (sorted) list of the $k$-th objective values from the Pareto front consisting of $M$ points. For any new value $\alpha \in \Rbb$, the \emph{insertion index} $j$ is the smallest integer $0 \leq j < M$ satisfying $\mathsf{P}_{k}(j) \;\ge\; \alpha \quad (\text{if such a } j \text{ exists})$, and set $j = M$ if no such index exists.
\end{definition}

\begin{restatable}{proposition}{paretooptimal}
\label{prop:pareto_optimal}
For the initial point $\pi_{0}$ of Problem \eqref{eq:first-obj}, let the insertion index of $\alpha_{k} := R_{k}(\pi_{0})$ in $\mathsf{P}_{k}$ be $j_{k}$. If $\bm{c}_k \geq \mathsf{P}_k(\max(0,j_{k}-1)) \forAll k \in [K-1]$, then the optimal solution of Problem \COP{COP-R}, if it exists, is a Pareto-optimal solution.
\end{restatable}
\begin{proof}
\label{proof:pareto_optimal}

We prove by contradiction. First, define a solution element by the tuple $(\pi, \bm{\mathsf{P}}^\pi)$, which refers to a policy $\pi$ along with its corresponding reward vector $\bm{\mathsf{P}}^{\pi} = \left(R_{1}(\pi), \dots, R_{K}(\pi) \right)$. Now suppose that the optimal solution $P^{'}=(\pi^{\prime},\bm{\mathsf{P}}^{\pi^{\prime}})$ of Problem \COP{COP-R} is not a Pareto-optimal solution. By the definition of Pareto-optimal solution, there exists a solution $\hat{P}=(\hat{\pi}, \bm{\mathsf{P}}^{\hat{\pi}})$ in $\Pi_{{P}}$ that dominates $P^{\prime}$, i.e., $R_{k}(\pi') \leq R_{k}(\hat{\pi}) \forAll k \in [K]$. Given $P_{0}=(\pi_{0}, \bm{\mathsf{P}}^{\pi_{0}})$, we have $R_{K}(\hat{\pi}) \geq R_{K}(\pi^{\prime}) \geq R_{K}(\pi_{0})$ by definition. Since both $P_{0}$ and $\hat{P}$ do not dominate each other, since $R_{K}(\hat{\pi}) \geq R_{K}(\pi_{0})$, there exists $k \in [K-1]$ such that $R_{k}(\pi_{0}) \geq R_{k}(\hat{\pi})$.

Now consider the values of $\bm{c}_{k}$ and $R_{k}(\hat{\pi})$ for some objective $k$. Note that $\bm{c}_{k} \geq \mathsf{P}_{k}(\max(0,j_{k}-1))$. If $R_{k}(\pi_{0}) \geq R_{k}(\hat{\pi}) > \bm{c}_{k}$, then $R_{k}(\pi_{0}) \geq R_{k}(\hat{\pi}) > \mathsf{P}_{k}(\max(0,j_{k}-1))$, which is conflicting with the condition that $\mathsf{P}_{k}(\max(0,j_{k}-1))$ is the $(\max(0,j_{k}-1))^{th}$ objective value in $\mathsf{P}_{k}$. If $R_{k}(\hat{\pi}) \leq \bm{c}_{k}$, it conflicts with the condition that $\hat{P}$ dominates $P'$. Therefore, such a $\hat{P}$ does not exist, and hence, $P'$ is a Pareto-optimal solution.
\end{proof}

\begin{wrapfigure}{r}{0.6\textwidth}
\vspace{-0.5cm}
\centering
{
    \includegraphics[height=0.25\textwidth, width=0.55\textwidth]{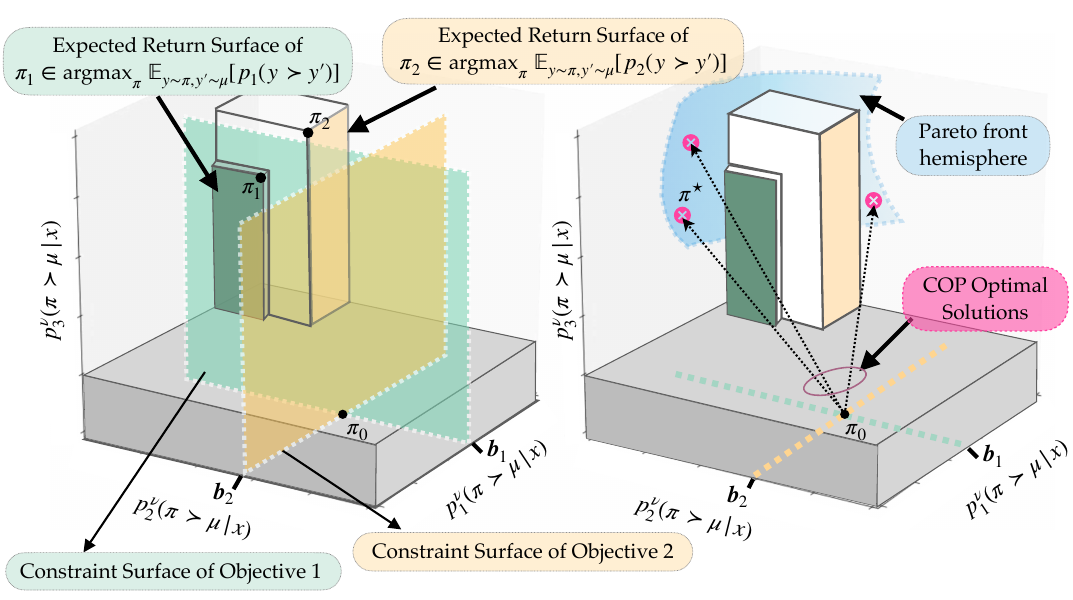}
}
\caption{\small Visualization of criteria for specifying constraint values. 
The expected return surface of $\pi_{1}$ ($\pi_{2}$), call it $\Scal_{1}$ ($\Scal_{2}$), in objective 1(2) is the $\max(0,j_{k} - 1)^{th}$ value in list $\mathsf{P}_{1}$ ($\mathsf{P}_{2}$) respectively. Therefore, specifying constraints values $\bm{b}_{1} \geq \Scal_{1}$ and $\bm{b}_{2} \geq \Scal_{2}$ is sufficient for the optimal solution of Equation \eqref{eq:first-obj} to be a Pareto-optimal solution.}
\label{fig:mopo-constraint-satisfy}
\vspace{-0.25cm}
\end{wrapfigure}

Proposition \ref{prop:pareto_optimal} formalizes the criteria for specifying appropriate constraint values and provides the condition for which the optimal solution of Problem \eqref{eq:first-obj} is a Pareto optimal solution. See Figure \ref{fig:mopo-constraint-satisfy} for the visualization. Proposition \ref{prop:pareto_optimal} gives a \emph{sufficient} condition under which the solution of Problem \eqref{eq:main-obj} is Pareto optimal. However, in practice this condition is (i) overly conservative and may exclude many feasible Pareto points, and (ii) computationally expensive as it requires re‑evaluating \emph{all} policies for non‑dominated sorting at every optimization step. Please see Section \ref{sec:function_approx} for an empirically validated practical constraint specification procedure.

\subsection{Log-barrier Function Optimization}
\label{appendix:logbarrier}

For some $\sigma, s > 0$, consider the following log barrier function,
\begin{equation*}
\Bcal_{\sigma,s}(z) = \begin{cases}
    - \sigma \log(-z) \, , & z \leq -s \\
    \frac{\sigma}{s} z + (1-\log(s))\sigma \, , & z > -s
\end{cases} \;\; , \; \;  \text{with}  \; \; \partial_{z}  \Bcal_{\sigma,s}(z) = \frac{\sigma}{\max(-z,s)} \, .
\end{equation*}

\begin{figure}[ht]
    \centering
    \includegraphics[scale=0.3]{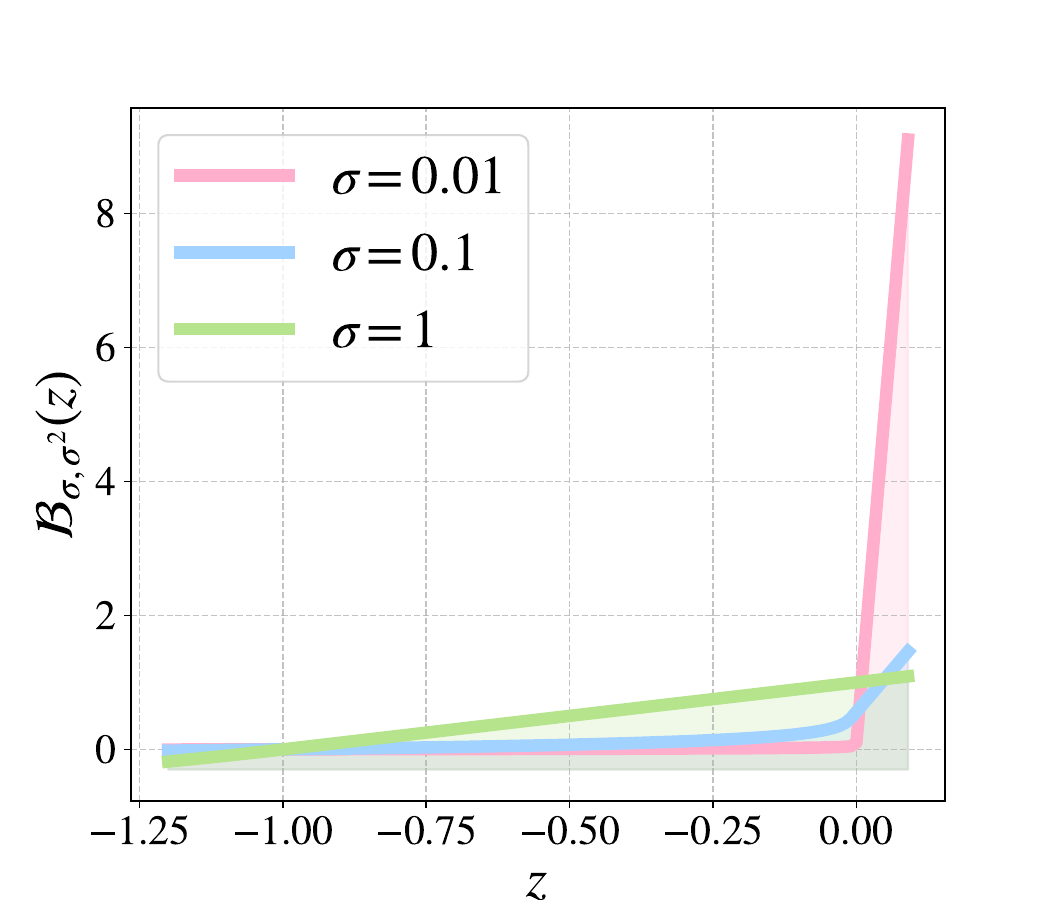}
    \caption{
        \textbf{The relaxed logarithmic barrier.}
        We depict the convergence of the relaxed logarithmic barrier $\mathcal{B}_{\sigma , \sigma^{2}} ( z )$ to the characteristic function $\chi \{ z \leq 0 \}$ as $\sigma \rightarrow 0$.
        We gradually decrease $\sigma$ from 1 to 0.01.
        Consequently, $\mathcal{B}_{\sigma , \sigma^{2}} ( z )$ gets closer to 0 for $z \leq 0$ and increases to $\infty$ otherwise.
    }
    \label{fig:logbarrier}
\end{figure}

For all $z \in \Rbb$, this is a convex, continuous, and differentiable function. Importantly, for $s = \sigma^{2}$, this barrier function converges to the characteristic function $\chi \{ z \leq 0 \}$ as $\sigma \rightarrow 0$, i.e., it takes the value 0 when $z \leq 0$ and $\infty$ otherwise; the condition $s = \sigma^{2}$ is sufficient, but not necessary for constraint satisfaction \citep{kervadec2022constrained}. This convergence to the characteristic function is visually depicted in Figure \ref{fig:logbarrier}, showing the change in the log barrier function as we gradually decrease $\sigma$.

In Section \ref{sec:algorithm}, we had defined

\begin{equation}
\label{eq:logbarrier_appendix}
        \Lcal_{\text{LB}}(\rho, \bm{\sigma}) = \Fcal(\rho) - \sum_{k=1}^{K-1} \Bcal_{\bm{\sigma}_{k},\bm{\sigma}_{k}^{2}}(\bm{b}_{k} - \bm{\Gcal}_{k}(\rho)) \; .
\end{equation}

Since the log barrier converges to the characteristic function as $\sigma \rightarrow 0$, we want to find the maximizer of $\Lcal_{\text{LB}}(\rho, \bm{\sigma})$ for small $\bm{\sigma}$. However, doing so directly leads to instabilities as the objective function is ill-conditioned. Instead, it is common practice to follow an iterative procedure: one finds the maximizer for a fixed $\bm{\sigma}$, reduces $\bm{\sigma}$, and repeats \citep{curtis2024stochastic}. Specifically, the procedure is instantiated with initial values $\rho_{0}$, $\bm{\sigma}_{0}$, and $0 < \bm{\gamma}_{k} < 1$ for $\bm{\gamma} = \{\bm{\gamma}_{k}\}_{k=1}^{K-1}$. On the $t$-th iteration, $\bm{\sigma}_{k}^{(t)} \leftarrow \bm{\gamma}_{k} \bm{\sigma}_{k}^{(t-1)}$ is reduced and $\rho^{(t)} \leftarrow \argmax_{\rho} \Lcal_{\text{LB}}(\rho, \bm{\sigma})$ (with initialization $\rho^{(t-1)}$). In doing so, the constraints are gradually enforced, nudging the LLM to satisfy them over the optimization procedure while avoiding instabilities. As $\{ \bm{\sigma}^{(t)} \} \searrow 0$, the weights $\{ \rho^{(t)} \}$ converge to the maximizer of the constrained problem.

It is impossible to maximize $\Lcal_{\text{LB}}(\rho, \bm{\sigma})$ exactly in many practical applications. Instead, at each iteration, one can take a single optimization step toward the solution.
Doing so is amenable to stochastic gradient methods and mitigates computational overhead: the optimization proceeds as normal while the value of $\bm{\sigma}$ is reduced over the course of the procedure. One can guarantee the convergence of this procedure to the optimal solution in some settings; for example, \cite{curtis2024stochastic} prove convergence when dealing with box constraints.
However, convergence in a scenario like ours is not guaranteed. Nevertheless, we will experimentally demonstrate its use for our constrained problems.

We employ stochastic gradient methods and derive the gradient of our objective function directly:
\begin{equation}
    \label{eq:log-barrier-objective-grad}
    \partial_{\rho} \Lcal_{\text{LB}}(\rho, \bm{\sigma}) =  \partial_{\rho} \Fcal(\rho) \ + \ \sum_{k = 1}^{K-1} \frac{\bm{\sigma}_{k} \partial_{\rho} \bm{\Gcal}_{k}(\rho)}{\max \left(\bm{\Gcal}_{k}(\rho) - \bm{b}_{k} , \bm{\sigma}_{k}^{2} \right)}.
\end{equation}
This follows immediately from Equation \eqref{eq:logbarrier_appendix}. See \cite{schulman2016high} for a more detailed review.

\textbf{Connection to Lagrange multipliers.} The log barrier and the Lagrangian are intrinsically connected; this becomes evident when comparing Equation \eqref{eq:log-barrier-objective-grad} with the (gradient of the) Lagrangian in Equation \eqref{eq:dual}. In particular, we define the multipliers:

\begin{equation*}
    \bm{\lambda}_{k} \ = \ \frac{\bm{\sigma}_{k}}{\max \left(\bm{\Gcal}_{k}(\rho) - \bm{b}_{k} , \bm{\sigma}_{k}^{2} \right)} \; .
\end{equation*}

They can be interpreted as Lagrange multipliers: for active constraints, $\bm{\lambda}_{k} = 1 / \bm{\sigma}_{k}$ is non-zero; for inactive constraints, $\bm{\lambda}_{k} = \bm{\sigma}_{k} / \left(  \bm{\Gcal}_{k}(\rho) - \bm{b}_{k} \right)$ vanishes to $0$ as $\bm{\sigma}_{k} \rightarrow 0$. Hence, the KKT complementary slackness condition is satisfied by design.

\subsection{Main Text Proofs}

\maindual*
\begin{proof}
\label{proof:main_dual}

Given the Lagrangian $\Lcal(\rho, \bm{\lambda})$, the dual formulation is given by

$$
\max_{\rho} \min_{\bm{\lambda} \geq 0} \Lcal(\rho, \bm{\lambda}) \equiv  \min_{\bm{\lambda} \geq 0} \max_{\rho} \Lcal(\rho, \bm{\lambda}) \, .
$$

By the strong duality, it is sufficient to consider KKT conditions for $(\rho^{\star}, \bm{\lambda}^{\star})$. 

(i) Primal feasibility i.e. $\Ebb_{y \sim \piref , y' \sim \mu}[ \rho^{\star}(y) \, \bm{q}(y \succ y')] \geq \bm{b}$.

(ii) Dual feasibility i.e. $\bm{\lambda}^{\star} \geq \bm{0}$.

(iii) Complementary slackness i.e. $(\bm{\lambda}^{\star})^{T} \left(\bm{b} - \Ebb_{y \sim \piref , y' \sim \mu}[ \rho^{\star}(y) \, \bm{q}(y \succ y')] \right) = \bm{0}$.

(iv) Stationarity i.e. $\nabla_{\rho} \Lcal(\rho, \bm{\lambda}) = 0$ i.e.

\begin{align*}
     \E{y' \sim \mu}[p(y \succ y')] - \tau \left(\ln (\rho(y)) + 1 \right) + \bm{\lambda}^{T} \E{y' \sim \mu}[\bm{q}(y \succ y')] = 0 \, . \\
    \implies \rho^{\star}(y) = \exp \left( \frac{1}{\tau} \E{y' \sim \mu} \left[ p(y \succ y') + \bm{\lambda}^{T} \bm{q}(y \succ y') \right] - 1 \right) 
\end{align*}

Now we show that conditions (i)-(iii) hold for the above $\rho^{\star}(y)$. For condition (i), by initialization of $\bm{b}$ using Proposition \ref{prop:pareto_optimal}, we have 

$$
\E{y \sim \piref \\ y' \sim \mu}[ \rho^{\star}(y) \, \bm{q}(y \succ y')] \geq \E{y \sim \piref \\ y' \sim \mu}[ \rho_{0}(y) \, \bm{q}(y \succ y')] \geq \bm{b} \; ,
$$
where $\rho_{0}(y) = \pi_{0}(y) / \piref(y)$,  $\pi_{0}$ is the initialization point of solving Problem \eqref{eq:main-obj}, and the second inequality follows by construction. Condition (ii) also holds by construction of $\bm{\lambda}$. Now, condition (iii) holds by definition if the constraint is active i.e. $\Ebb_{y \sim \piref , y' \sim \mu}[ \rho^{\star}(y) \, \bm{q}(y \succ y')] = \bm{b}$, and if it is inactive, then dual feasibility also ensures that complimentary slackness holds. 
As a consequence, all KKT conditions are always satisfied with the above $\rho^{\star}(y)$, which concludes the proof.
\end{proof}

\lowerbounddual*
\begin{proof}
\label{proof:lowerbound_dual}

For the given constrained optimization problem:

$$
\min_{\pi_{k}} \E{y \sim \pi_{k} \\ y' \sim \mu}[\rho(y) \bm{q}_{k}(y \succ y')] \; \suchthat \; \KL(\pi_{k} \;  || \;  \piref) \leq \epsilon \quad \text{and}\, , \quad \sum_{y \in \Ycal} \pi_{k}(y) = 1 \; ,
$$

we consider its Lagrangian to find its dual problem. By noticing that, $\pi_{k}(y) \rho(y) = \pi_{k}(y) \frac{\pi(y)}{\piref(y)} \approx \pi(y) $, the Lagrangian for some arbitrary multipliers $\chi_{k} \geq 0$ and $\zeta_{k} \in \Rbb$ is given by,

\small
\begin{align*}
\Lcal(\pi_{k}, \chi_{k}, \zeta_{k}) &= \E{y \sim \pi_{k} \\ y' \sim \mu}[\rho(y) \bm{q}_{k}(y \succ y')] + \chi_{k} \left( \E{y \sim \pi_{k}} \left[\ln \left( \frac{\pi_{k}(y)}{\piref(y)} \right) \right] - \epsilon \right) + \zeta_{k} \left( \sum_{y \in \Ycal} \pi_{k}(y) - 1 \right) \\
&= \E{y \sim \pi_{k} \\ y' \sim \mu}[\rho(y) \bm{q}_{k}(y \succ y')] + \chi_{k} \left( \sum_{y \in \Ycal} \pi_{k}(y) \ln \left( \frac{\pi_{k}(y)}{\piref(y)} \right) - \epsilon \right) + \zeta_{k} \left( \sum_{y \in \Ycal} \pi_{k}(y) - 1 \right) \, ,
\end{align*}
\normalsize

where $\chi_{k} \in \Rbb_{+}$ is the Lagrange multiplier for $\KL$ constraint, and $\zeta_{k} \in \Rbb$ is the Lagrange multiplier for the normalization constraint that ensures $\sum_{y \in \Ycal} \pi_{k}(y) = 1$. Hence, the corresponding optimization problem due to strong duality is: 

$$ \min_{\pi_{k}} \max_{\chi_{k} \geq 0, \zeta_{k}} \Lcal(\pi_{k}, \chi_{k}, \zeta_{k}) \equiv \max_{\chi_{k} \geq 0, \zeta_{k}} \min_{\pi_{k}} \Lcal(\pi_{k}, \chi_{k}, \zeta_{k}) .$$  

Now, we can compute the non-parametric closed form solution for each sample $y$ for
the inner minimization problem. Due to the convexity of $\KL$-divergence, it is sufficient to consider $\nabla_{\pi_{k}} \Lcal(\pi_{k}, \chi_{k}, \zeta_{k}) = 0$. For each $y$ we then have,

\begin{align*}
\nabla_{\pi_{k}} \Lcal(\pi_{k}, \chi_{k}, \zeta_{k}) =  \E{y' \sim \mu}[\rho(y) \bm{q}_{k}(y \succ y')] + \chi_{k}^{\star} \left( \ln \left( \frac{\pi_{k}^{\star}(y)}{\piref(y)} \right) + 1 \right) + \zeta_{k} = 0 \\
\implies \pi_{k}^{\star}(y) \propto \piref(y) \exp \left( (\chi_{k}^{\star})^{-1}   \E{y' \sim \mu}[\rho(y) \bm{q}_{k}(y \succ y')] \right) \, 
\end{align*}

with some normalization constant $Z_{k}$ that ensures that $\sum_{y \in \Ycal} \pi_{k}^{\star}(y) = 1$, which is described with respect to $\zeta_{k}$. Then, by plugging the above stationarity condition into the Lagrangian, we have the dual unconstrained optimization problem as,

\small
\begin{align*}
\max_{\chi_{k} \geq 0, \zeta_{k}} \Lcal(\pi_{k}^{\star}, \chi_{k}, \zeta_{k}) &= \E{y \sim \pi_{k}^{\star} \\ y' \sim \mu}[\rho(y) \bm{q}_{k}(y \succ y')] + \chi_{k} \left( \E{y \sim \pi_{k}^{\star}} \left[ \ln \left( \frac{\pi_{k}^{\star}(y)}{\piref(y)} \right) \right] - \epsilon \right) + \zeta_{k} \left( \sum_{y \in \Ycal} \pi_{k}^{\star}(y) - 1 \right) \\ 
&= \E{y \sim \pi_{k}^{\star}} \left[\E{y' \sim \mu}[\rho(y) \bm{q}_{k}(y \succ y')] + \chi_{k} \ln \left( \frac{\pi_{k}^{\star}(y)}{\piref(y)} \right) \right] - \chi_{k} \epsilon \\
&= - \chi_{k} \ln \left( \Ebb_{y \sim \piref , y' \sim \mu} \left[ \exp \left(  \chi_{k}^{-1} \rho(y) \bm{q}_{k}(y \succ y') \right) \right] \right) - \chi_{k} \epsilon \, .
\end{align*}
\normalsize

This concludes the proof.
\end{proof}

\subsection{Empirical Optimization Problem}
\label{appendix:samplingbasedcop}

We now formulate the problem given a fixed offline dataset $\Dcal$ of the form $\Dcal := \left\{ \left(x_{i}, y_{i}, y_{i}', \Ibb(y_{i},y_{i}') \right) \right\}_{i=1}^{N}$, where context $x_{i} \in \Xcal$, $y_{i}, y_{i}'$ are two generations from $\piref$ and $\mu$ respectively, and $\Ibb(\cdot, \cdot) \in \{0,1\}^{K}$ is preference indicator vector over $K$ objectives, i.e., $\Ibb_{k}(y, y') = 1$ if $y \succ_{k} y'$, and $0$ otherwise for $k \in [K]$. With a slight abuse in notation, we let $\Ibb_{p}(\cdot, \cdot) \triangleq \Ibb_{K}(y, y')$ and $\Ibb_{\bm{q}}(\cdot, \cdot) \triangleq \left(\Ibb_{1}(y, y'), \dots, \Ibb_{K-1}(y, y') \right)^{T}$. Then, the empirical optimization problem is given by:

\small
\begin{equation}
    \max_{\rho} \; \underbrace{ \frac{1}{N} \sum_{i=1}^{N} \rho(y_{i}) \, \Ibb_{p}(y_{i}, y_{i}') - \tau  \rho(y_{i}) \, \ln(\rho(y_{i}))}_{\hat{\Fcal}(\rho)} \quad \suchthat \quad \underbrace{\frac{1}{N} \sum_{i=1}^{N}   \rho(y_{i}) \, \Ibb_{\bm{q}}(y_{i}, y_{i}')}_{\hat{\bm{\Gcal}}(\rho)} \geq \bm{b}.
\label{eq:empirical-cop}
\end{equation}
\normalsize

\begin{proof}
We need to show that we can build an unbiased estimate of the optimization function from empirical observations. To this end, consider the sampled \COP{\small COP} as:

\begin{equation}
\label{eq:sampled-cop}
    \max_{\rho} \; \E{y \sim \piref \\ y' \sim \mu} [\rho(y) \, I_{p}(y, y')] - \tau  \E{y \sim \piref}[\rho(y) \, \ln(\rho(y)) \quad \suchthat \quad \E{y \sim \piref \\ y' \sim \mu} [\rho(y) \, \bm{I}_{\bm{q}}(y, y') ] \geq \bm{b}
\end{equation}

, where $\bm{I}(\cdot, \cdot)$ is a Bernoulli random preference vector over $K$ objectives i.e. $\bm{I}_{k}(y,y')$ is a random variable sampled from a Bernoulli distribution with mean $p_{k}(y \succ y')$ such that it is $1$ if $y \succ_{k} y' \,$, and 0 otherwise, where $\succ_{k}$ denotes preference with respect to the $k^{th}$ objective for $k \in [K]$. Following the notation discussed before, we let $I_{p}(\cdot, \cdot)$ to be the preference with respect to the $K^{th}$ objective, and let $\bm{I}_{\bm{q}}(\cdot, \cdot)$ to be the preference vector for the remaining $(K-1)$ objectives i.e. $I_{p}(y, y') = \bm{I}_{K}(y, y')$ and $(\bm{I}_{\bm{q}})_{k} = \bm{I}_{k}(y, y')$ for $k \in [K-1]$. Note that Problem \eqref{eq:main-obj} and Problem \eqref{eq:sampled-cop} are equivalent since $\Ebb_{y \sim \piref, y' \sim \mu}[\rho(y) \, I_{p}(y, y')] = \Ebb_{y \sim \piref, y' \sim \mu}[\rho(y) \, I_{p}(y, y') \given y, y'] = \Ebb_{y \sim \piref, y' \sim \mu}[\rho(y) \, p(y \succ y')]$. Similar argument follows for $\bm{I}_{\bm{q}}$. 
\end{proof}

Note that each data point $\left(x_{i}, y_{i}, y_{i}', \Ibb(y_{i}, y_{i}') \right)$ contributes two terms to the empirical problem above: one with $(x, y, y', \Ibb(y, y')) = (x_i, y_i, y_i', \Ibb(y_i, y_i'))$ and another with $(x, y, y', \Ibb(y, y'))$ = $(x_i, y_i',$ $ y_i, \Ibb(y_i', y_i))$. This symmetry is important to exploit since it reduces gradient variance and improves stability during optimization. For clarity, we omit the symmetric term in notation as incorporating it is trivial -- simply augment the current dataset by swapping $y_{i}$ and $y_{i}'$ and bit flipping $\Ibb(\cdot, \cdot)$ element-wise. 

Now, as before, for some $\bm{\lambda} := \{\lambda_{k}\}_{k=1}^{K-1} \geq \bm{0}$, we have the Lagrangian as $\hat{\Lcal}(\rho, \bm{\lambda}) = \hat{\Fcal}(\rho) - \bm{\lambda}^{T} \left(\bm{b} - \hat{\bm{\Gcal}}(\rho) \right)$, and the dual as,

\small
\begin{equation}
\resizebox{0.7\linewidth}{!}{$
\begin{aligned}
\label{eq:empirical-dual}
    \hat{\dual} & \triangleq \min_{\bm{\lambda} \geq \bm{0}} \max_{\rho} \; \hat{\Lcal}(\rho, \bm{\lambda}) = \min_{\bm{\lambda} \geq \bm{0}} \hat{\Lcal}(\rho^{\star}_{\bm{\lambda}}, \bm{\lambda}) = \min_{\bm{\lambda} \geq 0} \hat{\Fcal}(\rho_{\bm{\lambda}}^{\star}) - \bm{\lambda}^{T}(\bm{b} - \hat{\bm{\Gcal}}(\rho_{\bm{\lambda}}^{\star})) \\
    \text{where},& \;\, \rho^{\star}_{\bm{\lambda}}(y) = \exp\left( (\tau N)^{-1} \sum_{i=1}^{N}[\Ibb_{p}(y,y') + \bm{\lambda}^{T} \Ibb_{\bm{q}}(y,y')] - 1 \right).
\end{aligned}
$}
\end{equation}
\normalsize

Following our earlier discussion on lower bounding the preference probabilities for $[K-1]$ constraints, we wish to constrain $\texttt{LowerBound}(\hat{\bm{\Gcal}}(\rho))$ to be greater than $\bm{b}$. Then, following Equation \eqref{eq:chi_update}, the empirical lower bound is obtained by solving below for $M$ batches of $\Dcal$, where each batch $m$ of size $N_{M} = \lfloor N/M \rfloor$ is of the form $\left(x_{m,j}, y_{m,j}, y_{m,j}', \Ibb(y_{m,j}, y_{m,j}')\right)_{j=1}^{N_{M}}$.

\begin{equation}
\label{eq:empirical-chi-update}
\resizebox{0.82\linewidth}{!}{$
\begin{aligned}
    \min_{\bm{\chi} \geq \bm{0}} \hat{J}(\bm{\chi} \,; \rho) := \frac{1}{M} \sum_{m=1}^{M} \left[ \sum_{k=1}^{K-1} \left[ \chi_{k} \ln \left( \frac{1}{N_{M}} \sum_{j=1}^{N_{M}} 
 \exp \left(  \chi_{k}^{-1} [\rho(y_{m,j}) (\Ibb_{\bm{q}})_{k}(y_{m,j}, y_{m,j}')]  \right) \right) + \chi_{k} \epsilon \right] \right]. 
\end{aligned}
$}
\end{equation}

This transforms the empirical dual Problem \eqref{eq:empirical-dual} into the below optimizations for $\bm{\lambda}$ and policy $\pi_{\psi}$:

\small
\begin{align}
\label{eq:empirical_lambda_policy_update}
    \min_{\bm{\lambda} \geq \bm{0}} \hat{J}(\bm{\lambda} \,; \bm{\chi}) :=  \hat{\Fcal}(\rho) - \bm{\lambda}^{T}( \bm{b} - \hat{J}(\bm{\chi} \,; \rho)) \quad \text{and}, \; 
    \min_{\psi} \hat{J}_{\rho}(\pi_{\psi}) := - \frac{1}{N} \sum_{i=1}^{N} \rho_{\bm{\lambda}}^{\star}(y_{i}) \log(\pi_{\psi}(y_{i})) \, ,
\end{align}
\normalsize

where $\rho_{\bm{\lambda}}^{\star}(\cdot)$ is computed using Equation \eqref{eq:empirical-dual}.

\subsection{Implementation Details}
\label{appendix:implementation_details}

\subsubsection{Background}

\noindent\textbf{SFT.} Supervised fine-tuning (SFT) with labeled demonstrations is widely adopted to fine-tune LLMs \cite{zhang2023llama, peng2023instruction}. Given prompt-response pairs $\{(x, y)\}$ sampled form the dataset $\Dcal$, the SFT loss function is defined as:
\begin{equation}
    \mathcal{L}_{\mathrm{SFT}} = - \mathbb{E}_{(x, y) \sim \Dcal} \left[ \sum_{i} \log \pi_{\rm sft} (y_i|x, y_{<i}) \right], 
\end{equation}
where $\pi_{\mathrm{sft}}$ refers to the LLM policy and $y_{<i}$ indicates all tokens before the $i$-th token in response $y$.

\noindent\textbf{RLHF.} RLHF typically involves two steps \cite{ouyang2022training,wu2023fine}: reward modeling, and RL training. In reward modeling, a reward model $r_{\phi}$ is trained to minimize the loss function

$$
\mathcal{L}_{\mathrm{RM}}(\phi) =  - \mathbb{E}_{(x, y_w, y_l)\sim \Dcal} [ \log(\sigma(r_{\phi}(x, y_w) - r_{\phi}(x, y_l))) ]
$$
, where $\sigma(z)$ is the sigmoid function, $y_w$ and $y_l$ refer to preferred and dispreferred responses, respectively. Generally, RL training uses the PPO algorithm \cite{schulman2017proximal} with an additional KL penalty relative to the SFT policy:
\begin{equation*}
    \arg\max_{\pi_{\theta}} \mathbb{E}_{x\sim \Dcal, y\sim \pi_{\theta}(y|x)} \left[ r_{\phi}(x,y) - \tau \log \frac{\pi_{\theta}(y|x)}{{\pi_{\rm sft}(y|x)}} \right] ,
\end{equation*}
where $\tau > 0$ is the KL penalty coefficient.

\subsubsection{Training details.} We summarize the key implementation details of text generation tasks in Table \ref{tab:exp_details_text_generation}. This table also provides links to the open-sourced datasets and reward models utilized in our study. Implementation is primarily based on trl \cite{vonwerra2022trl}. Especially, SFT fine-tunes the base model, while MORLHF and Rewarded Soups fine-tune the SFT model using the PPO algorithm. In contrast, RiC directly fine-tunes the base model. See \cite{yang2024rewards} for more details. We apply the same 4-bit quantization and LoRA configuration for training all models. During evaluation, we maintain a consistent configuration across different models, generating 64 tokens for the Helpful Assistant task and 32 for the Reddit Summary task. 

For all baselines including \MOPO{MOPO}, we begin by normalizing the rewards using the mean and standard of the offline dataset before incorporating them into the prompts. During online generation (updating the reference policy) and evaluation, we sample a group of 25,000 random samples from a normal distribution and use the maximum and minimum values (generally around $\pm 3$) of these samples to replace the maximum and minimum values of the dataset. This method can prevent the extreme values in the dataset to impact reward selection. 

\paragraph{Incorporating preference vectors}. Pareto fronts are generated as in \cite{yang2024rewards}. One point to consider while evaluating empirical Pareto fronts is to incorporate user preferences for a particular objective. For instance, in the case of 2 objectives, in RiC \cite{yang2024rewards}, scalarization tuples are passed as in-context human preferences. Preference tuple $w = (w_{1},w_{2})$ for the two reward dimensions is passed to the model at inference time to adjust the LLM policy according to the user preferences. It is necessary to map these scalarization tuples $w$ to the desired rewards that will be used as conditioning in prompts. Similarly, for the two objective case, PARM uses prompts at inference time and clusters them. Then, they take two diverse policies, apply each policy to each cluster, and compute a multi-objective reward vector for all policy-cluster pairs. 

\begin{wrapfigure}{r}{0.55\textwidth}
\vspace{-0.5cm}
\centering
{
    \includegraphics[scale=0.43]{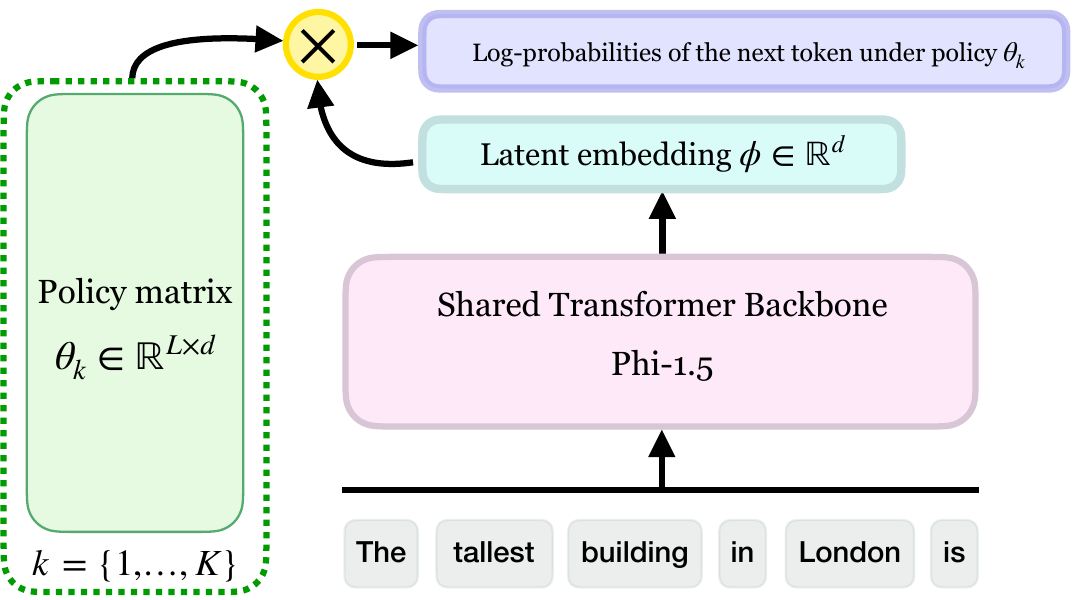}
}
\caption{Multi-headed policy architecture for incorporating preferences at inference time in \MOPO{MOPO}.}
\label{fig:multiplepolicyheads}
\end{wrapfigure}

A single point in the plots is a reward vector for the policy-cluster pair. For a fair comparison, \MOPO{MOPO} should also incorporate the preferences at inference time (in terms of which objective should be primary). For $K$ objectives, one idea would be to solve $K$ separate optimization problems as in Problem \eqref{eq:main-obj}, one for each objective being the ``main" objective. However, this is not efficient since if the LLM had $m$ parameters, the space complexity of representing $K$ policies would be $Km$.

We propose a multi-policy architecture in which all policies share a common LLM backbone. Each policy $k$ is parameterized by a matrix $\theta_k \in \Rbb^{L \times d}$, where $L$ is the number of tokens and $d$ is the transformer embedding dimension. The logit for the next token under policy $k$ is computed as $\theta_k \phi$, where $\phi \in \Rbb^d$ is the final-layer embedding summarizing the input sequence. This multi-headed model is illustrated in Figure \ref{fig:multiplepolicyheads}. The total parameter count is reduced to $m - dL + dKL$, where $m - dL$ parameters are in the shared backbone, and $dKL$ correspond to the $K$ policy heads. The shared backbone enables efficient language modeling, while the separate heads provide sufficient flexibility for each policy to optimize distinct objectives and language styles. During training, each policy matrix is jointly trained with the transformer backbone. At inference time, the user specifies which objective $k_{0} \in [K]$ should be the primary objective, and the model computes logit for the text token using $\theta_{k_{0}}\phi$ to adapt to inference time user specifications.

\begin{table}[ht]
	\centering%
	\caption{Key implementations of the text generation experiments.}%
	\centering
	\resizebox{0.9\textwidth}{!}{
		\begin{tabular}{cc}%
			\toprule
			\multicolumn{2}{c}{\textbf{Basic information}}                                                                                                                                                                             \\
			\midrule

            Architecture         & Transformer                                                                                                                                               \\
			Pre-training (SFT)        & See Section \ref{sec:experiments}                                                                                                                                             \\     
			Hardware             & NVIDIA A100 80GB, 1 accelerator, 12 vCPU     \\
            Quantization for training & 4bit \\
            Fine-tuning strategy & LoRA \cite{hu2022lora}                                                                                                                                                                                     \\
            LoRA $r$             & 16\\
 			LoRA alpha           & 32                                                                                                                                                                                     \\
			LoRA dropout         & 0.05                                                                                                                                                                                                                                                            \\
			Optimizer            & Adam           \\   
   			Batch size           & 8      \\
            Inference tokens for evaluation  & 64 for Helpful Assistant and 32 for Reddit Summary \\

            \midrule
			\multicolumn{2}{c}{\textbf{SFT}}              \\       
			\midrule
             Finetuning steps & 10000   \\
			Initial learning rate        & 1.41e-4  \\
            Learning rate scheduler & Linear \\   
			\midrule
    \multicolumn{2}{c}{\MOPO{MOPO}}              \\       
    			\midrule
                 Finetuning steps & 10000   \\
    			Initial learning rate        & 1.87e-4  \\
                Learning rate scheduler & Linear \\   
                Batch size & 8 \\   
                Regularization $\tau$ & 0.08 \\
                Constraint lower bound ball $\epsilon$ & 0.15 \\
                Constraint relaxation $\beta$ & 0.9995 \\
                Reference policy lag $t_{0}$ & 500 \\
    
            \midrule
			\multicolumn{2}{c}{\textbf{RiC}}              \\       
			\midrule
             Offline finetuning steps & 10000 \\
			Initial learning rate        & 1.41e-4  for offline finetuning, 1e-5 for online finetuning                                                              \\
            Learning rate scheduler & Linear for offline finetuning, constant for online finetuning \\   
            Threshold for MORS &  0.7-quantile for each reward dimension\\
            Online generation sample size per iteration & 10000 \\
            Online finetuning steps per iteration & 4000 \\
            
            \midrule
			\multicolumn{2}{c}{\textbf{RL step for MODPO}}                                                                                                                                                                      \\
			\midrule

			RL algorithm         & PPO \cite{schulman2017proximal}  \\
            Implementation      & trl \cite{vonwerra2022trl} \\
			KL regulaization               & 0.2                          \\
			Epochs               & 1                       \\          
            learning rate        & 1e-5  \\
            lambda for GAE    & 0.95 \\
            gamma             & 1 \\
            cliprange         & 0.2 \\
            Number of optimisation epochs per batch & 4 \\
            Target KL       & 3 \\

            \midrule
			\multicolumn{2}{c}{\textbf{Datasets and Reward Models}}                                                                                                                                                                      \\
			\midrule
            
            Task name            & \textbf{Helpful Assistant}         \\
			Description          & Provide helpful and harmless answers to potentially complex and sensitive questions.    \\
			Prompt               & No prompt, only users' questions.                                                                                                                                                       \\
			Dataset              &  \href{https://huggingface.co/datasets/Anthropic/hh-rlhf}{Anthropic/hh-rlhf}  \cite{bai2022training}  \\
            harmless reward               & \href{https://huggingface.co/Ray2333/gpt2-large-harmless-reward_model}{gpt2-large-harmless-reward\_model}     \\
            helpful reward       & \href{https://huggingface.co/Ray2333/gpt2-large-helpful-reward_model}{gpt2-large-helpful-reward\_model} \\  humor reward & \href{https://huggingface.co/mohameddhiab/humor-no-humor}{humor-no-humor} \\

            Task name            & \textbf{Reddit Summary}         \\
            Description          & Provide a summary to a post from Reddit.\\
            Prompt                & Generate a one-sentence summary of this post. \\
            Dataset & \href{https://huggingface.co/datasets/openai/summarize_from_feedback}{openai/summarize\_from\_feedback} \cite{stiennon2020learning} \\
            pref1 reward      & \href{https://huggingface.co/Tristan/gpt2_reward_summarization}{gpt2\_reward\_summarization} \\
            less-hallucination reward & \href{https://huggingface.co/CogComp/bart-faithful-summary-detector}{bart-summary-detector}      \\
			faithful reward & \href{https://huggingface.co/CogComp/bart-faithful-summary-detector}{bart-faithful-summary-detector}    \\        
			\bottomrule
		\end{tabular}
	}
	\label{tab:exp_details_text_generation}
\end{table}%

\textbf{Inference code.} Here we provide the inference pseudo-code. Notably, to prevent potential precision explosion, we approximate the solution for JSD same as Reverse KL-divergence, as they are inherently similar.
\begin{lstlisting}[
language=Python, breaklines=true, basicstyle=\ttfamily, keywordstyle=\color{blue}, stringstyle=\color{purple}
]
def f_divergence(logp, weights, f_type):
    if f_type in ("reverse_kld", "jsd"):
        return torch.stack([w * lp for w, lp in zip(weights, logp)]).sum(dim=0)

    if f_type == "forward_kld":
        alpha = 1.0
    elif "-divergence" in f_type:
        alpha = float(f_type.split("-", 1)[0])
    else:
        raise ValueError(f"Unknown f_type: {f_type}")

    terms = [
        -alpha * lp + np.log(w)
        for w, lp in zip(weights, logp)
        if w != 0
    ]
    return -torch.logsumexp(torch.stack(terms), dim=0)
\end{lstlisting}

\textbf{Compute resources.} For training RLHF and MODPO models, the number of workers are set as $3$, each taking up $7,000$M of memory, running for $5.1$ and $5.7$ hours respectively; for training RiC and PARM models, the number of workers are set as $2$, each taking up $11,000$M of memory, running for $3.4$ and $3.2$ hours respectively. For \MOPO{MOPO}, see Table \ref{tab:exp_details_text_generation}, which takes $4.1$ hours.

\textbf{Training hyper-parameters.} For PPO, we follow the settings of \cite{yang2024rewards} and train for $25$ batches; for DPO, we follow \cite{zhou2023beyond}, with \texttt{PERDEVICE\_BATCH\_SIZE}$=1$ and \texttt{MAX\_LENGTH}$=64$.

\textbf{Inference hyper-parameters.} For PPO, we follow the settings of \cite{yang2024rewards} with \texttt{NUM\_BEAMS}$=1$; for DPO, we follow \cite{zhou2023beyond} with \texttt{BATCH\_SIZE}$=4$, \texttt{MAX\_LENGTH}$=50$ and \texttt{NUM\_BEAMS}$=1$.

\textbf{Codebase.} Our codebase is mainly based on trl~\cite{trl} (\url{https://github.com/huggingface/trl}), MODPO (\url{https://github.com/ZHZisZZ/modpo}), RiC  (\url{https://github.com/YangRui2015/RiC}) and Finegrained RLHF (\url{https://github.com/allenai/FineGrainedRLHF}), and has referred to f-divergence DPO \cite{fDPO} (\url{https://github.com/alecwangcq/f-divergence-dpo}), PackLLM~\cite{mavromatis2024pack} (\url{https://github.com/cmavro/PackLLM}), and DPA~\cite{haoxiang2024arithmetic} (\url{https://github.com/Haoxiang-Wang/directional-preference-alignment}).

\subsection{Example Outputs}
\label{appendix:exampleoutputs}

Example generations for each dataset are shown in Tables \ref{tab:redditsummary} and \ref{tab:helpfulassistant_example1}. For each dataset, we show a representative prompt in the down-sampled dataset, and one generated response for each model/algorithm, with preference weightings set as $w=(0.5,0.5)$ for MODPO, PARM, and RiC.

\begin{table*}[htbp]
\caption{Examples of \textbf{Reddit Summary}.}
\label{tab:redditsummary}
\footnotesize
    \centering
        \centering
        \small
        \scalebox{.8}{

\begin{tabular}{p{2.7cm}|p{13.3cm}}
\midrule
\textbf{Prompt:} & \textbf{Instruction:} Rewrite the email below in a friendly, concise tone in a single sentence. 

\textbf{Input:} Dear team, due to updated budget constraints, we must reduce remote work days from three to one per week effective next month. Please ensure on-site attendance Monday–Thursday. Thank you for your cooperation.

\textbf{Response:}
\\
\midrule
\multicolumn{2}{l}{\textbf{LM response}} \\
\multicolumn{2}{l}{\ }\\
\textbf{MODPO:} & Starting next month, please work onsite Monday–Thursday; budget changes limit us to one remote day weekly.\\
\\
\textbf{PARM:} & Budget limits mean we’ll shift to one WFH day—please join us on-site Mon–Thu from next month; thanks for flexing with us!\\
\\
\textbf{RiC:} & Heads-up: from next month we’ll be in the office Mon–Thu (just one WFH day) because of budget constraints—appreciate everyone’s flexibility!\\
\\
\MOPO{MOPO}: & Hey team! Budget tweaks mean we’ll have one remote day going forward, so plan to be on-site Mon–Thu starting next month—thanks for rolling with the change!\\
\
\\
\bottomrule[1.2pt]
\end{tabular}
    }    
\end{table*}

\begin{table*}[htbp]
\caption{Examples of \textbf{Helpful Assistants}. Helpfulness \& Humor.}
\label{tab:helpfulassistant_example1}
\footnotesize
\centering
\scalebox{.8}{
\begin{tabular}{p{2.7cm}|p{13.3cm}}
\toprule
\textbf{Prompt:} & \textbf{Human:} What is two factor authentication, and how can it help me stay safe online?\\
\textbf{Assistant:}\\
\midrule
\multicolumn{2}{l}{\textbf{LM response}}\\
\\
\textbf{RLHF (Helpful):} & Two factor authentication is a security system that requires you to provide two pieces of information to verify your identity.  This can help protect you from fraud and identity theft.\\
\\
\textbf{RLHF (Humor):} & What do you mean by two factor authentication?\\
\\
\textbf{MODPO:} & Two-factor authentication (2FA) adds a second proof—usually a code from your phone—on top of your password, so a thief who steals the password still can’t get in.\\
\\
\textbf{PARM:} & Think of 2FA as a double lock: you enter your password, then confirm with a phone code or authenticator app.  That quick extra step blocks most account-takeovers—using it on email first is a great start!\\
\\
\textbf{RiC:} & It’s a “prove-it-twice” check—password *plus* something you have (SMS code, hardware key).  Turn it on for banking, email, socials and you slam the door on 99\% of drive-by hacks.\\
\\
\MOPO{MOPO}: & Picture ordering pizza: the driver needs your address *and* a secret knock.  2FA works the same—password first, then a six-digit code or hardware tap.  Even if crooks guess the password, the knock stays secret, so enable it everywhere (especially email and banking) for huge peace-of-mind gains.\\
\bottomrule
\end{tabular}}
\end{table*}

\subsection{Three-objective Experimental Results}
\label{appendix:full_exp}

Here we present results for the three-objective text generation tasks in Table \ref{tab:threeobjective_appendix}.

\begin{table}[ht]
\centering
\caption{Three-objective alignment for Helpful Assistant task with normalized rewards across different SFT models when aligned using baselines.}
\label{tab:threeobjective_appendix}
\renewcommand{\arraystretch}{1.1}
\fontsize{7}{9}\selectfont
\addtolength{\tabcolsep}{-0.35em}
\begin{adjustbox}{max width=\textwidth}
\begin{tabular}{c|ccc|ccc|ccc|ccc}
\hline
 & \multicolumn{3}{c|}{\textbf{phi-1.5}} & \multicolumn{3}{c|}{\textbf{OpenChat-v3.5}} & 
   \multicolumn{3}{c|}{\textbf{Llama-3.1-8B}} & \multicolumn{3}{c}{\textbf{Mistral-7b-v0.2 (Instruct)}} \\ 
\cline{2-13}
 & helpful & humour & harmless & helpful & humour & harmless & helpful & humour & harmless & helpful & humour & harmless \\
\hline
RLHF-r1 & \cellcolor{accgood!65}{0.65} & -0.73 & -0.47 & \cellcolor{accgood!65}{0.72} & -0.38 & -0.20 & \cellcolor{accgood!65}{0.69} & -0.33 & -0.18 & \cellcolor{accgood!65}{0.74} & -0.41 & -0.22 \\ 
RLHF-r2 & -0.93 & \cellcolor{accgood!65}{0.44} & -0.53 & -0.78 & \cellcolor{accgood!65}{0.49} & -0.37 & -0.80 & \cellcolor{accgood!65}{0.50} & -0.39 & -0.79 & \cellcolor{accgood!65}{0.52} & -0.38 \\ 
RLHF-r3 & -0.88 & -0.97 & \cellcolor{accgood!65}{0.29} & -0.77 & -0.89 & \cellcolor{accgood!65}{0.40} & -0.80 & -0.90 & \cellcolor{accgood!65}{0.43} & -0.78 & -0.91 & \cellcolor{accgood!65}{0.41} \\ 
RiC & 0.18 & 0.10 & 0.05 & 0.27 & 0.16 & 0.12 & 0.26 & 0.14 & 0.11 & 0.25 & 0.15 & 0.10 \\ 
PARM & 0.23 & 0.12 & 0.11 & 0.32 & 0.18 & \cellcolor{accgood!25}{0.24} & 0.30 & 0.17 & \cellcolor{accgood!25}{0.22} & \cellcolor{accgood!25}{0.45} & 0.16 & \cellcolor{accgood!25}{0.23} \\ 
MODPO & -0.04 & -0.17 & 0.00 & 0.05 & -0.08 & 0.09 & 0.03 & -0.10 & 0.07 & 0.04 & -0.09 & 0.08 \\ 
DPO on $\Djoint{\Dcal_J}$ & 0.13 & 0.05 & 0.08 & 0.17 & 0.08 & 0.10 & 0.19 & 0.10 & 0.12 & 0.18 & 0.09 & 0.11 \\ 
\MOPO{MOPO-LB} & \cellcolor{accgood!25}{0.24} & \cellcolor{accgood!25}{0.13} & \cellcolor{accgood!25}{0.14} & 0.29 & 0.18 & 0.17 & 0.31 & 0.20 & 0.19 & 0.30 & 0.19 & 0.18 \\ 
\MOPO{MOPO-Lag} & 0.20 & 0.10 & 0.09 & \cellcolor{accgood!25}{0.40} & \cellcolor{accgood!25}{0.23} & 0.18 & \cellcolor{accgood!25}{0.38} & \cellcolor{accgood!25}{0.21} & 0.16 & 0.39 & \cellcolor{accgood!25}{0.22} & 0.17 \\ 
\hline
\end{tabular}
\end{adjustbox}
\end{table}

\subsection{Extension to Safety Controlled Preference Optimization}

Optimizing large language models for both utility (helpfulness) and safety constitutes an intrinsically multi-objective learning problem. Empirically it is observed that optimizing for one objective often degrades the other: aggressive safety tuning can induce over-refusal on benign inputs, while utility-centric optimization can erode safety guarantees \cite{bai2022training, touvron2023llama, qifine}. These observations motivate a principled multi-objective formulation that treats safety and helpfulness as separate, measurable criteria and seeks Pareto-efficient trade-offs rather than ad-hoc scalarization. Hence, we adapt the \MOPO{MOPO} framework to the setting of balancing helpfulness and safety objectives as part of the preference optimization problem. 

In the setting of incorporating safety as a second objective, we let the true human preference for the primary objective be $p(y \succ y' \given x)$, and for the \textit{safety} objective be $q(y \succ 0 \given x)$ i.e. overloading notation we have, $y \succ 0$ implies that generation $y$ is safe while $0 \succ y$ implies that $y$ is unsafe, where $0$ denotes a \emph{neutral} generation.  In essence, we want the model to output winning and \textit{safe} generations. The goal is to propose a general solution for \rlhf with a primary objective and a safe objective, based on constrained optimization of a function of preferences. To this end, we consider a reference policy $\piref \in \Delta^{\mathcal{X}}_{\mathcal{Y}}$, a real positive regularization parameter $\tau \in \mathbb{R}_{+}$, and let $b \in [0,1]$. The concave constrained optimization problem becomes,

\begin{align}
\label{eq:safe-obj}
    \max_{\pi} \, \E{x \sim \nu \\ y \sim \pi( \cdot \given x) \\ y' \sim \mu( \cdot \given x)}[p(y \succ y'  \given  x)] - \tau \KL(\pi \;  || \; \piref) \,  \suchthat \E{x \sim \nu \\ y \sim \pi( \cdot \given x)}[q(y \succ 0 \given  x)] \geq b.
\end{align}

Problem \eqref{eq:safe-obj} is analogous to Problem \eqref{eq:first-obj}, hence, the derivation carries over by replacing the secondary objective of Problem \eqref{eq:first-obj} with the safety objective of Problem \eqref{eq:safe-obj}.

\paragraph{Safety Benchmark.}
To evaluate safety, we first construct a benchmark that includes discriminative tasks and generative tasks based on previous benchmarks \citep{srivastava2023beyond, eval-harness, tedeschi2024alert,zou2023universal}. Discriminative tasks measure the models' recognition of multiple safety topics, including bias (CrowS-Pairs \citep{nangia2020crows}, BBQ \citep{parrish2021bbq}, WinoGrande \citep{sakaguchi2021winogrande}), ethics (ETHICS \citep{hendrycks2020aligning}, Moral Permissibility \citep{srivastava2023beyond}, Simple Ethics Questions \citep{hendrycks2020aligning,lourie2021scruples}), and toxicity (ToxicGen \citep{hartvigsen2022toxigen}, BigBench HHH Alignment \citep{srivastava2023beyond}).
In generative tasks, we prompt models to generate harmful content using the prompt dataset, AdvBench \citep{zou2023universal}, Real Toxicity Prompts \citep{gehman2020realtoxicityprompts}, ALERT \citep{tedeschi2024alert}. We report the percentage of harmless responses based on the HarmBench-Llama2-13B-Chat safety classifier \citep{mazeika2024harmbench}. Details of the benchmark are given below:

\begin{itemize}
    \item \textbf{CrowS-Pairs}: A challenge set for evaluating the tendency of language models (LMs) to generate biased outputs. We evaluate the English subset and \href{https://github.com/EleutherAI/lm-evaluation-harness/blob/main/lm_eval/tasks/crows_pairs/README.md}{implementation by LM Evaluation Harness} and report the Percentage of Stereotype metric. 
    \item \textbf{Bias Benchmark for QA (BBQ)}: Measures social biases in the output of multiple-choice question-answering models.  We use the Lite version and \href{https://github.com/google/BIG-bench/tree/main/bigbench/benchmark_tasks/bbq_lite}{implementation by BigBench} and report the Accuracy metric.
    \item \textbf{WinoGrande}: A collection of 44K problems inspired by the Winograd Schema Challenge, adjusted to improve scale and robustness against dataset-specific biases. We use the \href{https://github.com/EleutherAI/lm-evaluation-harness/tree/main/lm_eval/tasks/winogrande}{implementation by LM Evaluation Harness} and report the Accuracy metric.
    \item \textbf{ETHICS}: A benchmark spanning concepts in justice, well-being, duties, virtues, and commonsense morality.  We use the \href{https://github.com/EleutherAI/lm-evaluation-harness/tree/main/lm_eval/tasks/winogrande}{implementation by LM Evaluation Harness} and report the Accuracy for all sub-tasks.
    \item  \textbf{BigBench Moral Permissibility}: Tests whether ultra-large language models can read a short story where a moral scenario is presented and answer the question, "Is it morally permissible to do X?" in a manner similar to humans. We use the \href{https://github.com/google/BIG-bench/blob/main/bigbench/benchmark_tasks/moral_permissibility/README.md}{implementation by BigBench} and report the Accuracy metric.
    \item \textbf{BigBench Simple Ethics Questions}: Assesses a language model's responses to hypothetical, consequential, political, and social questions. We use the \href{https://github.com/google/BIG-bench/tree/main/bigbench/benchmark_tasks/simple_ethical_questions}{implementation by BigBench} and report the Accuracy metric.
    \item \textbf{ToxicGen}: A benchmark for evaluating the ability of language models to classify input text as either hateful or not hateful.  We use the \href{https://github.com/EleutherAI/lm-evaluation-harness/tree/main/lm_eval/tasks/toxigen}{implementation by LM Evaluation Harness} and report the Accuracy metric.
    \item \textbf{BigBench HHH Alignment}: Evaluates language models on alignment, pragmatically broken down into the categories of helpfulness, honesty/accuracy, harmlessness, and other aspects.  We use the \href{https://github.com/google/BIG-bench/tree/main/bigbench/benchmark_tasks/hhh_alignment}{implementation by BigBench} and report the Accuracy metric.
    \item \textbf{AdvBench} contains harmful prompts. We use the prompts provided \href{https://github.com/llm-attacks/llm-attacks/blob/main/data/advbench/harmful_behaviors.csv}{here} and generation implementation by  LM Evaluation Harness. We report the percentage of harmless responses measured by \href{https://huggingface.co/cais/HarmBench-Llama-2-13b-cls}{HarmBench-Llama-2-13b-cls} model.  
    \item \textbf{RealToxicityPrompts}: A benchmark for evaluating the ability of language models to continue a prompt in a non-toxic way.  We use the \href{https://github.com/EleutherAI/lm-evaluation-harness/tree/main/lm_eval/tasks/toxigen}{implementation by LM Evaluation Harness} report the percentage of harmless responses measured by \href{https://huggingface.co/cais/HarmBench-Llama-2-13b-cls}{HarmBench-Llama-2-13b-cls} model. 
    \item \textbf{ALERT}: A benchmark to assess the safety of LLMs through red teaming methodologies.   We use the prompts provided \href{https://huggingface.co/datasets/Babelscape/ALERT}{here} and generation implementation by  LM Evaluation Harness. We report the percentage of harmless responses measured by \href{https://huggingface.co/cais/HarmBench-Llama-2-13b-cls}{HarmBench-Llama-2-13b-cls} model. 
    \item \textbf{ALERT Adversarial}: A benchmark to assess the safety of LLMs through red teaming methodologies with adversarial prompts.  We use the prompts provided \href{https://huggingface.co/datasets/Babelscape/ALERT/viewer/alert_adversarial}{here} and generation implementation by  LM Evaluation Harness. We report the percentage of harmless responses measured by \href{https://huggingface.co/cais/HarmBench-Llama-2-13b-cls}{HarmBench-Llama-2-13b-cls} model.  
    \item \textbf{AlpacaEval} Based on the AlpacaFarm evaluation set, which tests the ability of models to follow general user instructions. We employ the \href{https://github.com/tatsu-lab/alpaca_eval}{official implementation} report the LC Win Rate. 
\end{itemize}

\paragraph{Dataset Details.} In the supervised fine-tuning stage, we follow \cite{tunstall2023zephyr,daisafe2023} to use a mix of helpfulness data  from UltraChat \cite{ding2023enhancing} and  safety data from PKU-SafeRLHF \cite{daisafe2023}. In the \MOPO{MOPO} stage, we use 30K helpfulness data from UltraFeedback \cite{cui2024ultrafeedback} and 30K safety data from PKU-SafeRLHF.  UltraFeedback contains instruction-following tasks that provide paired helpfulness preference rankings, and we treat all responses as safe since they undergo human filtering. PKU-SafeRLHF  provides both paired helpfulness preference rankings and binary safety labels. 

\begin{figure*}[ht]
    \centering
    \includegraphics[width=0.4\linewidth]{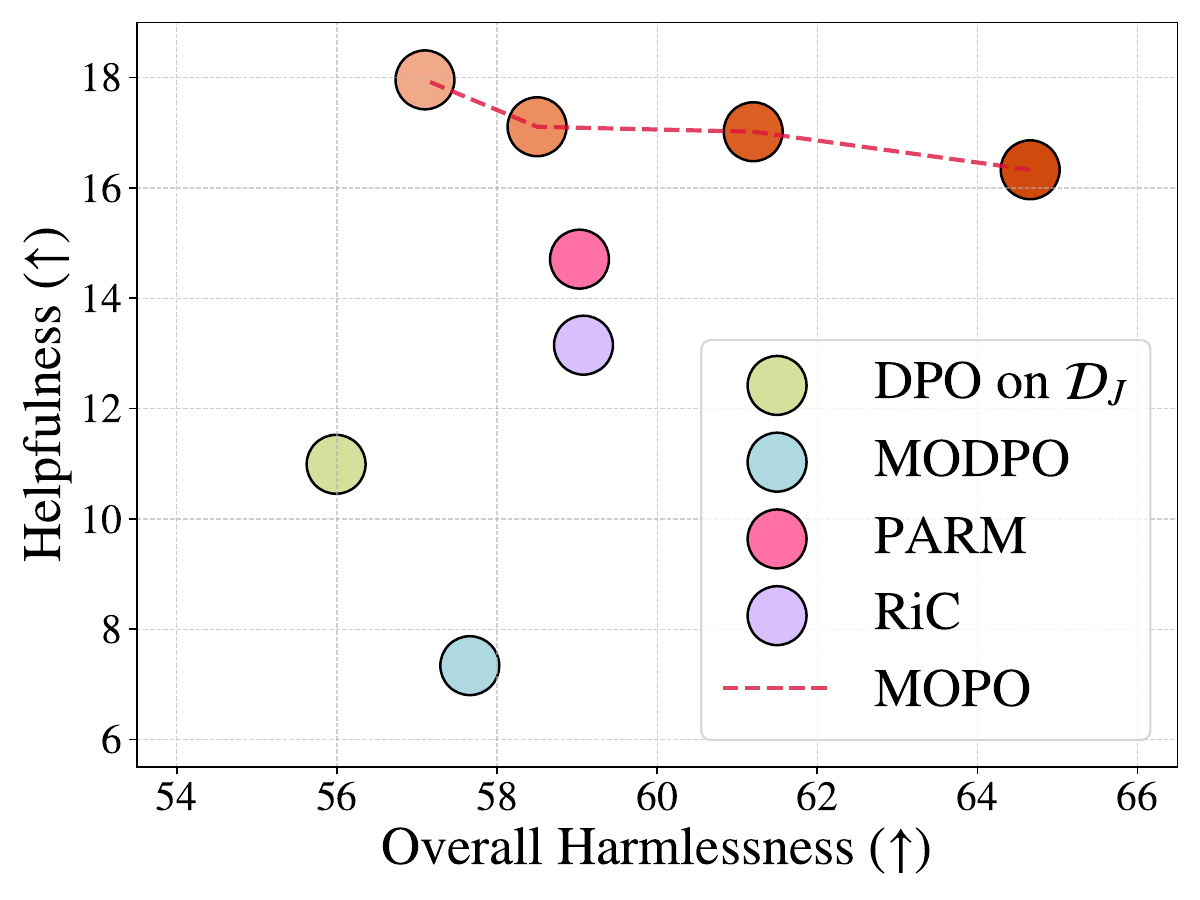}
    \caption{Empirical Pareto front of \MOPO{MOPO} with varying harmless thresholds in comparison with baselines.}
    \label{fig:safety_pareto}
\end{figure*}

\paragraph{Overall Evaluation Metrics.} In the following experiments, we report both the helpfulness and safety performance. Helpfulness is measured using AlpacaEval 2.0 (Alpaca) \citep{dubois2024length,dubois2023alpacafarm}. Safety is assessed using the performance of discriminative tasks (\emph{Disc.}), generative tasks (\emph{Gen.}) from the aforementioned benchmark, and the average safety over these two metrics (\emph{Overall}). The results are presented in Table \ref{tab:safety} and Figure \ref{fig:safety_pareto}.

\begin{table}[ht]
\centering
\caption{Helpfulness and Harmlessness scores.}
\label{tab:safety}
\small
\begin{tabular}{lcccc}
\toprule
& Helpfulness & \multicolumn{3}{c}{Harmlessness} \\
\cmidrule(lr){2-2} \cmidrule(lr){3-5}
& Alpaca ($\uparrow$) & Disc. ($\uparrow$) & Gen. ($\uparrow$) & Overall ($\uparrow$) \\
\midrule
DPO on \Djoint{$\Dcal_{J}$} & 10.99          & 49.05            & 62.94            & 55.99 \\
MODPO                       & 7.34           & 46.42            & 68.91            & 57.66 \\
PARM                        & 14.71          & 48.35            & 69.71            & 59.03 \\
RiC                         & 13.15          & 42.41            & \textbf{75.76}   & 59.08 \\
\MOPO{MOPO}                 & \textbf{16.33} & \textbf{54.09}   & 75.24            & \textbf{64.66} \\
\bottomrule
\end{tabular}
\end{table}

\chapter{Closing Remarks and Future Directions}
\label{chapter:conclusion}

This thesis advances the field of Reinforcement Learning (RL) by addressing critical challenges in safety, scalability, and alignment across constrained and preference-based learning settings. By developing algorithms such as ACPO, \texttt{e-COP}, \texttt{warmPref-PS}, $\PSPL$, and \texttt{MOPO}, this thesis contributes to building robust frameworks for constrained optimization and preference-based learning. These algorithms demonstrate strong theoretical guarantees and empirical performance, tackling diverse challenges like average-reward optimization, episodic learning, efficient exploration, and alignment with human values in safety-critical applications. Below, a brief summary of results and possible future directions of research are provided. 

Chapter \ref{chapter:acpo} studied the problem of learning policies that maximize average-rewards for a given CMDP with average-cost constraints. It showed that the current algorithms with constraint violation bounds for the discounted setting do not generalize to the average setting. Hence, the Average-Constrained Policy Optimization (ACPO) algorithm was proposed that was inspired by the TRPO class of algorithms but based on theoretical sensitivity-type bounds for average-CMDPs. Experimental results on a range of OpenAI Gym  environments (including some high dimensional ones) showed the effectiveness of ACPO on ACMDP RL problems, as well as its superior empirical performance vis-a-vis some current alternatives. A direction for future work is implementation of ACPO to fully exploit the parallelization potential. 

Chapter \ref{chapter:ecop} then introduced an easy to implement, scalable policy optimization algorithm \texttt{e-COP} for episodic RL problems with constraints due to safety or other considerations. It was based on a policy difference lemma for the episodic setting, which surprisingly has quite a different form than the ones for infinite-horizon discounted or average settings. This provided the theoretical foundation for the algorithm, which was designed by incorporating several time-tested, practical as well as novel ideas. Policy optimization algorithms for Constrained MDPs tend to be numerical unstable and non-scalable due to the need for inverting the Fisher information matrix. \texttt{e-COP} sidesteped both of these issues by introducing a quadratic damping penalty term that worked remarkably well. The algorithm development was well supported by theory, as well as with extensive empirical analysis on a range of Safety Gym and Safe MuJoco benchmark environments against a suite of baseline algorithms adapted from their non-episodic roots. 

Next, Chapter \ref{chapter:warmPref-PS} proposed \texttt{warmPref-PS}, a Bayesian posterior sampling-based algorithm that efficiently incorporated offline preference data to warm-start the online learning phase. Theoretical analysis and foundations for bridging the gap between fixed, offline preferences and online learning was also provided. Further, Bootstrapped \texttt{warmPref-PS} was introduced, a computationally tractable extension designed to handle large-scale environments. Theoretical results and empirical evaluations demonstrated the robustness and superior performance of \texttt{warmPref-PS}. While additional work is needed to refine the approach for RLHF, Chapter \ref{chapter:warmPref-PS} provides a promising foundation for further development in this space.

Chapter \ref{chapter:PSPL} then proposed $\PSPL$, a Bayesian Top-two posterior sampling-style algorithm that efficiently incorporated offline preference data to learn a MDP with unknown rewards and transitions. An important attribute of $\PSPL$ was credit assignment to state-action pairs since the learner only received trajectory level feedback. The first theoretical analysis and foundations for bridging the gap between fixed, offline preferences and online learning were provided. Further, Bootstrapped $\PSPL$, an approximate algorithmic version of $\PSPL$ was derived due to loss of conjugacy in posterior updates. Finally, various ablation studies were done to highlight the effect of offline preferences in online learning. 

Finally, Chapter \ref{chapter:mopo} introduced \texttt{MOPO}, a scalable and iterative algorithm for multi-objective preference optimization in LLMs. The chapter formulated alignment as a constrained optimization problem, where a primary objective was maximized while maintaining secondary objectives, such as safety and honesty, above tunable thresholds. \texttt{MOPO} effectively integrated preference-based learning with constraint satisfaction to handle multiple, potentially conflicting objectives in alignment tasks. Theoretical justification for the approach and extensive empirical evaluations demonstrated the scalability of \texttt{MOPO} to multi-billion parameter models and its superior performance compared to existing baselines. While additional research is needed to explore its extensions to dynamic or hierarchical constraints, \texttt{MOPO} provides a robust and generalizable framework for balancing competing alignment objectives in practical LLM training.

Given that, notice there are two key commonalities across this thesis that provide a direction of future research. 

Firstly, Chapters \ref{chapter:warmPref-PS} and \ref{chapter:PSPL} use posterior sampling as the general framework. Simplicity of the algorithms, similarity in various settings, and excellent numerical performance compared to their optimism-based competitors are important factors for practitioners. However, all of the posterior sampling algorithms designed in this dissertation are for the tabular setting where the state and the action spaces are finite. A natural next step would be to extend these algorithms to the continuous state and action spaces, and to propose model-free algorithms for such settings. 

Secondly, while the algorithms developed in this thesis demonstrate strong performance and theoretical guarantees, they have primarily been evaluated using synthetic datasets, which offer controlled environments for benchmarking and analysis. Real-world data, however, often presents additional complexities, such as noise, bias, and variability, which are harder to model but critical for practical deployment. Developing RL algorithms that can effectively handle these challenges without compromising safety, scalability, or alignment is an important avenue for future research. Importantly, the insights and methodologies introduced in this thesis provide a strong foundation for tackling these challenges.

\begin{appendices}
    
\end{appendices}

\clearpage
\addcontentsline{toc}{chapter}{References}
\bibliography{references}
\bibliographystyle{apalike2}

\end{document}